\documentclass{article}

\usepackage[nonatbib,eandd,final]{neurips}%

\usepackage[numbers,sort&compress]{natbib}

\usepackage[utf8]{inputenc} %
\usepackage[T1]{fontenc}    %
\usepackage{hyperref}       %
\usepackage{url}            %
\usepackage{booktabs}       %
\usepackage{amsfonts}       %
\usepackage{nicefrac}       %
\usepackage{microtype}      %
\usepackage{xcolor}         %
\definecolor{darkgreen}{RGB}{41,166,41}

\usepackage{pifont}
\usepackage{todonotes}
\usepackage{xspace}         %
\usepackage{amsmath}
\usepackage[inline]{enumitem}
\usepackage{multirow}
\usepackage{placeins}
\usepackage[nameinlink]{cleveref}
\usepackage{comment}
\usepackage{makecell} %
\usepackage{subcaption} %
\usepackage{pifont} %

\usepackage{textcomp}  %
\usepackage{scalerel}  %

\usepackage{listings}
\definecolor{codegray}{gray}{0.95}
\definecolor{codeblue}{rgb}{0,0,0.6}
\lstdefinestyle{mypython}{
    backgroundcolor=\color{codegray},
    commentstyle=\color{gray},
    keywordstyle=\color{codeblue}\bfseries,
    numberstyle=\tiny\color{gray},
    stringstyle=\color{orange},
    basicstyle=\ttfamily\footnotesize,
    breaklines=true,
    numbers=left,
    numbersep=5pt,
    showstringspaces=false,
    language=Python
}

\usepackage[page]{appendix}
\usepackage{etoolbox}

\renewcommand{\appendixtocname}{Table of Contents.}

\makeatletter
\let\oldappendix\appendices

\renewcommand{\appendices}{%
  \clearpage
  \renewcommand{\thesection}{\Roman{section}}
  \let\tf@toc\tf@app
  \addtocontents{app}{\protect\setcounter{tocdepth}{2}}
  \immediate\write\@auxout{%
    \string\let\string\tf@toc\string\tf@app^^J
  }
  \oldappendix
}%

\newcommand{\listofappendices}{%
  \begingroup
  \renewcommand{\contentsname}{\appendixtocname}
  \let\@oldstarttoc\@starttoc
  \def\@starttoc##1{\@oldstarttoc{app}}
  \tableofcontents%
  \endgroup
}

\makeatother

\renewcommand{\cite}{\citet}

\newcommand{\ColIID}[1]{\textcolor[HTML]{54A24B}{#1}}
\newcommand{\ColTemporal}[1]{\textcolor[HTML]{D55E00}{#1}}
\newcommand{\ColGrouped}[1]{\textcolor[HTML]{6A0DAD}{#1}}

\newcommand{\cIID}{\ColIID{IID}\xspace}
\newcommand{\ctemporal}{\ColTemporal{temporal}\xspace}
\newcommand{\cgrouped}{\ColGrouped{grouped}\xspace}

\newcommand{\benchname}{\texttt{BeyondArena}\xspace}
\newcommand{\df}{\texttt{DataFoundry}\xspace}

\newcommand{\myparagraph}[1]{\textbf{#1}\quad}
\space

\title{Beyond IID: How General Are Tabular\\ Foundation Models, Really?}

\author{
Lennart Purucker$^{1,2}$ \quad
Andrej Tschalzev$^3$ \quad
Nick Erickson$^1$ \quad
Gioia Blayer$^4$ \\
\textbf{David Holzmüller}$^4$ \quad
\textbf{Alan Arazi}$^{5,1}$ \quad
\textbf{Alexander Pfefferle}$^{6,2}$ \quad
\textbf{Mustafa Tajjar}$^{2, 7}$ \\
\textbf{Gaël Varoquaux}$^{4,8}$ \quad
\textbf{Frank Hutter}$^{1,6,2}$
\\
$^1$Prior Labs \quad
$^2$University of Freiburg \quad
$^3$University of Mannheim \quad
$^4$INRIA Saclay \\
$^5$Technion \quad
$^6$ELLIS Institute Tübingen \quad
$^7$Zuse School ELIZA \quad
$^8$Probabl \quad
\\
 $\texttt{mail@tabarena.ai}$
 \vspace{-1em}
}

\newlength{\iconHeight}
\begin{document}

\newcommand{\tree}{\includegraphics[
  height=\iconHeight,
  keepaspectratio,
]{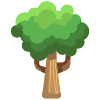}}
\newcommand{\mlp}{\includegraphics[
  height=\iconHeight,
  keepaspectratio,
]{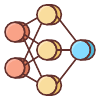}}
\newcommand{\fm}{\includegraphics[
  height=\iconHeight,
  keepaspectratio,
]{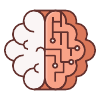}}
\newcommand{\bm}{\includegraphics[
    height=\iconHeight, 
    keepaspectratio,
]{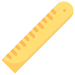}}

\maketitle

\begin{abstract}
\vspace{-1em}

Foundation models for predictive machine learning on tabular data have recently gained significant traction in academia and industry.
Research communities across disciplines are increasingly evaluating tabular foundation models on diverse datasets and tasks.
However, these task- and discipline-specific evaluations remain largely inaccessible to model researchers because benchmark software and evaluation protocols are fragmented. 
As a result, model researchers rely on standard benchmarks, which are mostly defined for tasks where tabular foundation models already excel.
The most challenging scenarios are excluded, limiting meaningful progress in the field by focusing on marginal improvements on IID data rather than on broader, more demanding challenges.
To overcome this, we introduce \benchname, the first unified holistic benchmark for tabular data that supports diverse task types (\cIID, \ctemporal, \cgrouped), across sample size and feature dimensionality scales, with diverse feature types (with text, with high cardinality) from a broad range of disciplines. 
To enable unified benchmarking beyond standard benchmarks, we introduce \df, a Python framework and metadata schema for curating tabular datasets for predictive machine learning.
Our results across $11$ models and $142$ curated datasets show that current open-source tabular foundation models are competitive or excel on datasets up to 100k rows, while traditional tree-based and deep learning models still dominate on grouped or large data. 
\benchname guides model research for the most demanding challenges in tabular data, enabling progress towards truly foundational tabular models. 
\end{abstract}

\begin{figure}[!h]
    \centering
    \includegraphics[width=\textwidth]{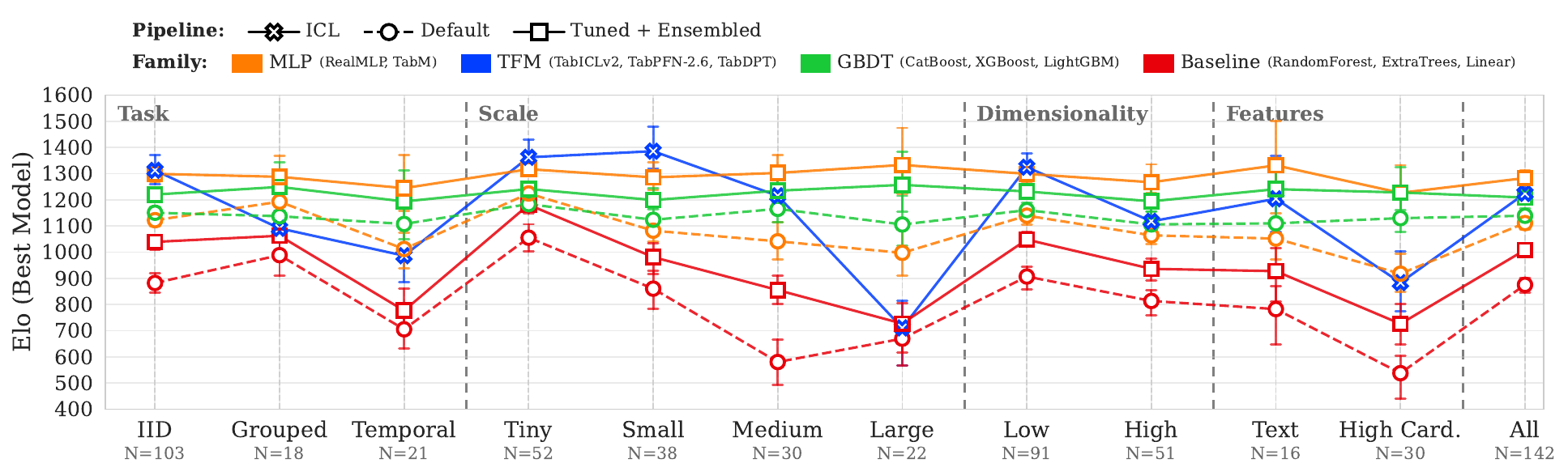}
    \caption{
    \textbf{BeyondArena Results.} We show the Elo of the best model per family across sub-benchmarks, comparing state-of-the-art MLPs, gradient-boosted decision trees (GBDT), tabular foundation models (TFM), and simple baselines. For TFMs, we evaluate in-context learning performance. For traditional models, we evaluate performance with default hyperparameters and with tuning plus post-hoc ensembling. We show results per model in \Cref{fig:results_per_model}, for $n\le100k$ in \Cref{fig:robustness_imputation}.
    } 
    \label{fig:first_page}
\end{figure}

\FloatBarrier
\space
\section{Introduction}
Tabular foundation models (TFMs) have been evaluated across many research communities, ranging from graph learning \citep{choi2025can,eremeev2025turning,hayler2025bringing,liao2026tfmlinker} or surrogate modeling \citep{yu2025git,du2026meta,hu2026foundation}, to real-world applications such as Cybersecurity~\citep{garcia2025foundation}, soil mapping~\citep{barkov2026modern}, or clinical predictions~\citep{shaktah2026established}. 
While the broader research community evaluates TFMs across diverse tasks and datasets, the tabular research community is so far limited to evaluations on IID tasks~\citep{erickson2025tabarena,liu2024talent,zhang2025limix}.
Moreover, historically, projects that broaden evaluation diversity were often developed with independent benchmarking infrastructure and scientific standards, such as PMLB-Mini~\citep{knauer2024pmlbmini} for tiny data, TabReD~\citep{rubachev2024tabred} for tabular-temporal tasks, or TextTabBench~\citep{mraz2025towards} for tabular-text data.
The current state of benchmarking for machine learning on tabular data is not only confusing to practitioners but also impeding researchers. %
\\
We aim to unify benchmarking and accelerate research on TFMs. 
Thus, we introduce \benchname, a holistic benchmark for machine learning on tabular data that unifies independent and identically distributed (IID) and non-IID task types (\cIID, \ctemporal, \cgrouped) across datasets with varying sample scales (100-1k, 1k-10k, 10k-100k, 100k to 1 million), dimensionality scales (${\le}100$, ${>}100$), with challenging feature types (high-cardinality categories, or text) from a broad range of applied disciplines. 
To create \benchname, we curate a large collection of representative datasets and provide reliable baselines. 
Furthermore, we unify the tabular benchmarking community by integrating our benchmark into TabArena's~\citep{erickson2025tabarena} open-source ecosystem.
To summarize, \textbf{our contributions are:}
\begin{itemize}[leftmargin=3em,label=\ding{118}]%
    \vspace{-0.5em}
    \item We manually curate $142$ datasets from $1128$ -- following rigorous protocols -- to obtain high-quality tabular datasets across scales and feature types, and publish \df, a Python framework and metadata schema for reproducible tabular data curation;
    \item We systematically evaluate whether $3$ state-of-the-art open-source TFMs generalize beyond IID settings by comparing them with $8$ strong traditional baselines across diverse tasks; 
    \item We integrate all our contributions into TabArena's open-source benchmark ecosystem, thereby avoiding fragmentation of the research community: code at 
    \href{https://tabarena.ai/code}{\texttt{https://tabarena.ai/code}}, data at 
    \href{https://github.com/TabArena/data-foundry}{\texttt{https://github.com/TabArena/data-foundry}}.
\end{itemize}
\vspace{-0.5em}
Our driving motivation behind \benchname is to understand how well TFMs would perform when practitioners use them in real-world predictive applications, thereby closing the gap between evaluations in academia and real applications.
In this paper, we detail our dataset curation, standardize benchmarking across task types, and evaluate the strengths and weaknesses of existing TFMs.

\textbf{BeyondArena focuses on} evaluating predictive machine learning models for tabular classification and regression on non-IID data, ranging from tiny to large (100-1M), with diverse feature types (high-cardinality categorical and text features).  
Out of scope are few-shot predictions (${<}100$), images as features, and other tasks such as survival analysis or relational learning. 

\textbf{Our results demonstrate} that current state-of-the-art open-source TFMs, evaluated via in-context learning, dominate on tiny, small, and \cIID data, and remain competitive on temporal and high-dimensional datasets with up to 100k training rows.
They fail to compete with traditional tree-based and deep learning models on non-IID \cgrouped or large datasets.
We further ablate our experimental setup to assess the validity of our claims and show the importance of 
\begin{enumerate*}[label=(\roman*)]
\item appropriate outer test splits for grouped data; 
\item appropriate inner validation splits for non-IID or tiny data; 
\item preprocessing for grouped data and text features;
\item and calibrating probabilities.
\end{enumerate*} 

\space
\section{Background: IID and non-IID Tabular Data}
\label{sec:related_work}

Tabular foundation models and, by extension, tabular deep learning have been evaluated in the broader research community across countless applications, from computer science disciplines (cf. \citep{ye2025closer,hoppe2025comparing,cheng2025realistic,schiffman2025foundation,hoo2024tabular,du2026meta,xu2026no,landsgesell2026distributional,zhang2025tabpfn,byun2025risk,hu2026foundation,vu2025adaptation,lourencco2025bridging,choi2025can,berge2026computing,yu2026fire,zheng2025tables,yu2025git,schiff2025gradient,hayler2025bringing,liu2026tabular,kim2026tabular,karabulut2026tabular,marszalek2026tactic,liao2026tfmlinker,eremeev2025turning,pan2026lakemlb}) to applied science disciplines (cf. \citep{garcia2025foundation,dyikanov2024comprehensive,tran2024predicting,dao2025early,karabacak2025data,magadan2023early,alzakari2024artificial,noda2024machine,yang2024comparing,barkov2026modern,wang2025harnessing,li2025mri,li2026tabpfn,liu2026evaluating,perciballi2024adapting,lim2025redefining,zhou2025limitations,schmidinger2026kriging,schwarz2026predicting,zhao2026weakly,varghese2026tabular,ding2025longitudinal,zhu2026method,heinzel2025advancing,johnsson2026predicting,guo2026predicting,wang2026transformer,shang2025context,wu2026panmetai,romano2026machine,rahman2025machine,chen2026data,chen2025coupling,zhang2025boosting,xu5591702multiscale,bellarmino2025minimal,viga2025fuelcast,chu2024deep,saito2025tabular,lal2026evaluating,shaktah2026established}).
To categorize such broad application data, we adopt an application-dependent task definition.
We determine whether a task requires a random \cIID or non-IID split based on what best mirrors the original real-world application, following the principle that test data should reflect the unseen data a model will face once deployed \citep{hornung2026evaluating}.

\myparagraph{Illustration.}
Two practitioners could work with the same data but for fundamentally different applications. 
Practitioner Alice uses the data to predict whether past transactions were fraudulent for follow-up fraud investigations.  
Practitioner Bob uses the data to predict whether new transactions are fraudulent, aiming to prevent fraud in real time.
To estimate model performance, Alice uses a random \cIID split, while Bob uses a non-IID \ctemporal split to simulate the expected distribution of unseen transactions when the model is deployed. 
If Bob used a random split, it would overestimate the performance of models that exploit temporal leakage \citep{rubachev2024tabred,tschalzev2025unreflected}.
In contrast, Alice can use a random split even though transactions are not independent: her test transactions come from the same historical pool as her training transactions, so, under a retrospective view, the training and test samples are exchangeable, and a temporal split would underestimate performance.
The appropriate split depends on the practitioner's application. 
\benchname extends this line of thinking from practitioners to data curation and academic benchmarking.

\myparagraph{IID Tabular Data.}
We treat a dataset's task as \cIID if the associated application does not impose a particular structure on the test samples, allowing a random split.
That is, we hold out randomly sampled data.
Strictly speaking, a random split requires only that training and test samples are exchangeable, i.e., that their joint distribution is invariant under permutation, not that samples are independent; we use \cIID as the conventional label for such tasks.

\myparagraph{Non-IID Tabular Data.}
We treat a dataset's task as non-IID when the application requires a \ctemporal or \cgrouped split.
In a \ctemporal split, a time index is required, and test samples occur strictly after the training data, reflecting prediction for observations in the future (e.g., future transactions).
In a \cgrouped split, also known as grouped resampling for clustered data \citep{hornung2026evaluating}, a group index is required, and all samples with the same index are kept together, so no group appears in both the training and test sets.
Such applications aim to generalize to samples associated with unseen entities (groups).
Grouped tasks can be of type \texttt{label-per-group}, where all samples in a group share the same label, or \texttt{label-per-sample}, where each sample has its own label.
For \texttt{label-per-group}, the objective is to predict a group-level label (e.g., if a customer is fraudulent based on a collection of transactions). 
For \texttt{label-per-sample}, the objective is to predict individual sample labels for data sources not seen during training (e.g., a new country or hospital).
\\
For some datasets, either a temporal or a grouped split is plausible. %
Using a temporal split does not remove group structures, and using a grouped split does not make the task time-invariant.
The split only determines the importance of grouped or temporal dependencies -- it does not remove them. 
\\
\benchname does not include data from time-series forecasting applications, despite the data being tabular-like and non-IID \citep{chatfield2000time, lim2021time}. 
During data curation, we distinguish data from temporal tabular regression and time-series forecasting tasks, as they rely on fundamentally different assumptions and require different validation procedures; see \Cref{appendix:tsfvsttt}.
Nevertheless, tabular models can be used for time-series forecasting tasks \citep{shchur2023autogluon,hoo2024tabular,jayawardhana2026zero,potapczynski2026arrow}, and can be competitive \citep{aksu2024gift,shchur2025fev,garza2026impermanent,qiao2026s}.

\section{Related Work}

As a unified tabular benchmark, \benchname relates to IID and non-IID tabular benchmarks.

\myparagraph{IID Tabular Benchmarks.}
The vast majority of prior benchmarks for predictive machine learning on tabular data focused on IID tasks, often inappropriately treating all data as stemming from IID applications.
Benchmarks were created during model development \citep{holzmuller2024better,gorishniy2024tabr,gorishniy2024tabm,hollmann-nature25a},
for tabular deep learning \citep{gorishniy-neurips21a,shwartz2022tabular,grinsztajn-neurips22a,mcelfresh-neurips23a,shmuel2024comprehensive,zabergja2024tabular,lee2025multitab,jiang2026omnitabbench},
to develop AutoML systems \citep{hanussek2020can,zoller2021benchmark,purucker-automlws22a,jiang2024good,gijsbers-jmlr24a,jurado2025automl},
or from a data-centric view \citep{bischl-arxiv17a,bischl-neuripsdbt21a,olson-biodata17a,romano2022pmlb,fischer-automlws23a,salinas2024tabrepo,tschalzev2024data,kohli-dmlr24a}, all aiming to understand and compare models. 
Recent benchmark projects have focused on broadening evaluation diversity by supporting few-shot or tiny data \citep{knauer2024pmlbmini,lee2025range}, including diverse domains and large data \citep{liu2024talent,ye2024closer}, or by curating datasets representative of real-world tasks \citep{erickson2025tabarena}.
\benchname builds on prior benchmarking efforts by extending their high-quality work beyond IID data.
\\
In parallel, the research community developed benchmarks for multimodal tabular data. 
These efforts concentrated on tabular data with text \citep{shi-neuripsdbt21a,grinsztajn2023vectorizing,lu2023mug,tang2024autogluon,tang2024bag,kim2024carte,kim2025table,mraz2025towards,ressel2025linear,kolomenko2026embedding,blayer2026strablebenchmarkingtabularmachine} or to benchmark LLMs solving predictive tasks by treating tables as text \citep{yin2020tabert,hegselmann2023tabllm,fang2024large,chen2024clinicalbench,bordtelephants,shysheya2025jolt,silvestri2025evaluating,liu2025robustness,pavlidis2025large,gardner2024large,lin2024tab2text,xing2024table,schindler2025tabgemma,gorla2026illusion}.
Tabular data with images has received little benchmarking attention \citep{luo2025time,erickson2022multimodal,tang2024autogluon, arazi2026multabenchbenchmarkingmultimodaltabular}, yet there have been application-specific model comparisons \citep{huang2023multimodal,du2025stil,lu2023mug,hager2023best,tang2024bag,wolf2022daft,jiang2024tabular,polsterl2021combining,du2024tip,malafaia2025learning,ding2025mrlf,hasny2025no}.
Time-series classification or regression benchmarks \citep{bagnall2018uea,tan2021time,dau2019ucr,middlehurst2026multiverse} are also multimodal IID benchmarks. They use random splits, and the state-of-the-art includes tabular models with feature engineering \citep{dempster2019rocket,lines2018time,dempster2021minirocket} or pretrained encoders \citep{goswami2024moment,feofanov2025mantis}.
\benchname makes a first step toward unifying multimodal benchmarking by incorporating IID and non-IID tabular data with text.
We leave tabular data with images and data from time-series classification or regression for future work.

\myparagraph{Non-IID Tabular Benchmarks.}
Benchmarks for non-IID tabular data focused mainly on \ctemporal data. 
While the broader research community worked on non-IID tabular data for decades \citep{stein2002benchmarking,shani2010evaluating,bergmeir2012use,huyen2022designing,ji2023large,baranchuk2023dedrift,ambrosio2025beyond,garcia2026rolling}, only recently \citet{rubachev2024tabred} identified temporal tabular data as a gap in tabular deep learning benchmarks and introduced TabReD, enabling new research \citep{cai2025feature,cai2025understanding}.
\benchname builds upon TabReD by supporting more temporal datasets, more baselines, and tabular foundation models. Moreover, \benchname enables research on IID and temporal tabular data in a single framework. 
\\
There have been no dedicated benchmarks for tabular deep learning on \cgrouped data.
Many applications of grouped tabular data originate from longitudinal data \citep{ding2025longitudinal,diggle2002analysis,weiss2005modeling,diggle2024longitudinal,molenberghs2005models}, algorithm selection problems \citep{tierney2015algorithm,bischl2016aslib,petelin2025pitfalls}, or distribution shift benchmarks \citep{malinin2021shifts,malinin2022shifts,liu2023need,gardner2023benchmarking,yao2022wild,kolesnikov2023wild,shmuel2024machine}.
\benchname is the first benchmark to incorporate grouped data from all of these fields.
Thus, \benchname fills a gap in non-IID benchmarking and enables new research on models that generalize across diverse application domains.

To summarize, compared to related work, we unify IID and non-IID benchmarking in \benchname by curating representative datasets in one shared metadata format (\Cref{sec:data_curation}) and enabling rigorous evaluation of state-of-the-art tabular models on all tasks and multimodal data types (\Cref{sec:experiment_setup}).
\space
\section{Dataset Curation}
\label{sec:data_curation}
Our curation process aims to extend tabular benchmarking to better represent more challenging real-world predictive machine learning applications. 
Therefore, we overhaul the selection criteria from TabArena~\citep{erickson2025tabarena} and extend them to \benchname's focus by supporting:
\begin{enumerate*}[label=(\textbf{\Roman*})]
\item non-IID data (temporal and grouped);
\item data with less than $500$ or more than $100\,000$ training samples; and
\item tabular data with text and date features.
\end{enumerate*}
To ensure selected datasets are high-quality and representative of \benchname's focus, we opted for a manual curation process in which all results were verified by humans\footnote{
We investigated (semi-)automated agentic procedures with limited success; see \Cref{appendix:automated_data_cuartion} for details.}.
While this choice was extremely labor-intensive, it is crucial to the scientific rigor, as the data quality directly affects the correctness of the benchmark's conclusions \citep{tschalzev2024data,rubachev2024tabred,kohli-dmlr24a, tschalzev2025unreflected}.

\myparagraph{Data Foundry.}
To enable such extensive dataset curation and make it reproducible, we introduce \df, a Python framework and metadata schema for tabular datasets and predictive machine learning: \href{https://github.com/TabArena/data-foundry}{\texttt{https://github.com/TabArena/data-foundry}}.
\df contains: (1) a mature Python package for dataset curation, and (2) the exact code we used to process each dataset in easy-to-understand Python notebooks (.ipynb). 
The Python package delivers code for dataset checks, train-test splits, and an extensible metadata schema.
In addition, \df has one Python notebook for each dataset in \benchname.
Each notebook contains reproducible code for metadata edits, preprocessing, dataset checks, train-test splits, and exporting the artifact for consumption by benchmarks, data repositories (such as OpenML, Kaggle, or Hugging Face), or other applications.

\myparagraph{Dataset Sources.}
We gathered datasets from $21$ tabular benchmark studies and searched for new datasets in public data repositories. 
In particular, we re-evaluated $304$ datasets from $14$ prior tabular benchmarks considered for the TabArena curation, comprising all accepted datasets and those rejected for being non-IID, too small, or for other data issues. %
We further collected datasets from $7$ non-IID or multimodal tabular benchmarks: TabReD \citep{rubachev2024tabred}, TableShift \citep{gardner2023benchmarking}, the string vectorizing benchmark \citep{grinsztajn2023vectorizing}, the CARTE benchmark \citep{kim2024carte}, TextTabBench \citep{mraz2025towards}, the TabSTAR benchmark \citep{arazi2025tabstar}, and the AutoGluon Multimodal benchmark \citep{erickson-arxiv20a,erickson2022multimodal,tang2024autogluon}.
Furthermore, we searched for new datasets 
by browsing data repositories or competition websites (UCI Machine Learning Repository \citep{asuncion2007uci}, OpenML \citep{vanschoren-sigkdd13a,bischl2025openml}, Hugging Face \citep{huggingface}, Kaggle \citep{kaggle}, Zindi \citep{zindi}, ASlib \citep{bischl2016aslib}) or public domain government websites. 
Our search was guided by popularity, task, and application domain, while avoiding obvious duplicates, to provide diverse input for our curation process.

\myparagraph{Dataset Selection Criteria.}
A dataset had to fulfill the following criteria to become part of \benchname: 
\begin{enumerate*}[label=(\textbf{\arabic*})]
    \item The dataset and its predictive machine learning task are unique within our benchmark;
    \item The dataset is not a few-shot prediction task, that is, it has at least $100$ train samples; 
    \item The dataset is from a tabular IID or non-IID task, that is, a random, temporal, or grouped split is the appropriate validation protocol;
    \item The dataset was published explicitly for a predictive classification or regression task;
    \item The dataset and its task are representative of a problem in a real-world application that a practitioner would solve with tabular machine learning models, that is, the dataset stems from a real random distribution (e.g., the dataset is not generated from a deterministic function), and tabular models are a suitable, competitive choice (e.g., the dataset is not a vectorized version of ImageNet).    
    \item The dataset and its predictive task do not raise (obvious) ethical concerns.
\end{enumerate*}
\\
Several criteria, in particular (5) and (6), require subjective judgment: each selected dataset was reviewed by at least one of four curators, $51\%$ by two or more, and contested cases were discussed and resolved within the curation team.
We provide a detailed description in \Cref{appendix:dataset_selection_criteria} and share our curation insights per dataset at \href{https://tabarena.github.io/data-foundry/}{\texttt{https://tabarena.github.io/data-foundry/}}.
\\
Compared to the criteria by \citet{erickson2025tabarena}, we made bigger but also nuanced changes.
The bigger changes are that we allow non-IID data, relax the modality requirements, and allow the use of larger datasets with subsampling, making any dataset with at least $100$ training samples in scope. 
Nuanced changes include a more precise definition of duplicates, a clearer distinction from tabular-adjacent predictive tasks, and a sharper definition of what constitutes a representative dataset for predictive classification or regression tasks.
Furthermore, although some of our datasets overlap with the inclusion criteria of the TabArena datasets, we show that ours are more challenging (\Cref{appendix:ta_datasets_comparison}).

\myparagraph{Dataset Processing.}
Each selected dataset was thoroughly investigated and integrated into \df.
By doing so, we unified the data format (e.g., .csv, .parquet, or .xlsx) and feature type annotations (numbers, categorical, dates, strings). 
Moreover, we store all necessary metadata for the task (e.g., train-test splits, target column, columns for the group or time index), as well as annotations from our curation process (e.g., notes on data anomalies), in a machine-readable data schema. 
Whenever necessary, we also apply feature engineering to create the appropriate task, following the task description in the paper or the top solutions on competition websites.
We share details and some general guidelines that we developed during the curation process in \Cref{appendix:dataset_curation_guidelines}.

\begin{figure}
    \centering
    \includegraphics[width=0.75\textwidth]{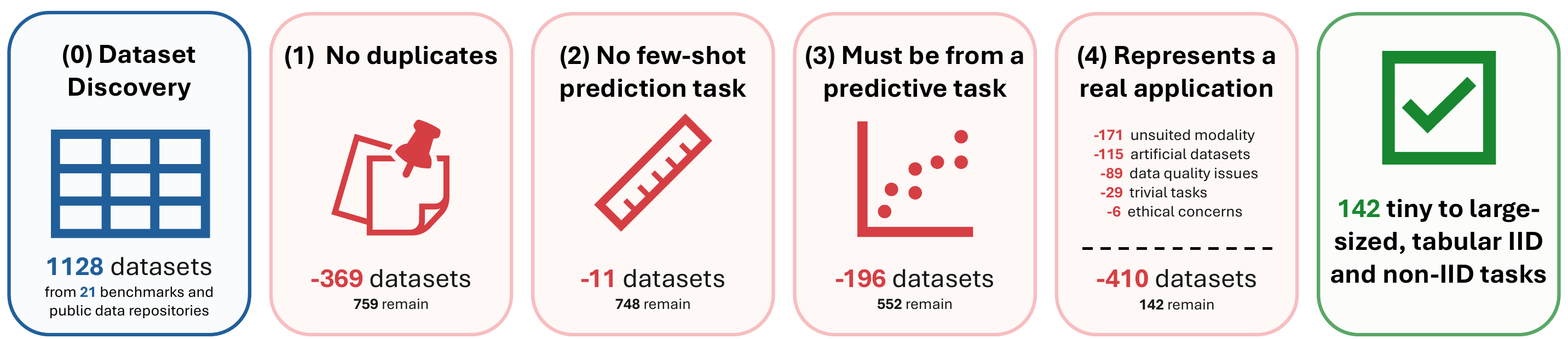}
    \caption{
    \textbf{Dataset Selection Results.} We present the main reasons for filtering datasets (1-4). Within (4), we also show sub-categories. Roughly $12.6\%$ of all investigated datasets were selected.
    }
    \label{fig:curation_filter}
\end{figure}

\myparagraph{Curation Outcome.}
\label{sec:curation_outcome}
We collected $1128$ datasets, of which $142$ met our selection criteria. As shown in \Cref{fig:curation_filter}, the majority of datasets were duplicates, did not represent a real application, or were not originally from a predictive task.
We summarize the characteristics of our dataset collection in
\Cref{fig:ds_dashboard}, with extended overviews in \Cref{appendix:dataset_curation_outcome}.
\benchname includes datasets that range from tiny to large-scale, are recent, and are diverse across feature types, problem types, task types, sources, and application domains. 
We provide details per dataset in \Cref{appendix:datasets} and share their \df artifacts at \href{https://huggingface.co/datasets/TabArena/BeyondArena}{\texttt{https://huggingface.co/datasets/TabArena/BeyondArena}}.
We discuss our distribution policy for datasets in \Cref{appendix:artifact_metadata}.

\begin{figure}
    \centering
    \resizebox{0.8\textwidth}{!}{%
    \begin{minipage}{\textwidth}
        \begin{subfigure}{0.48\textwidth}
            \includegraphics[width=\textwidth]{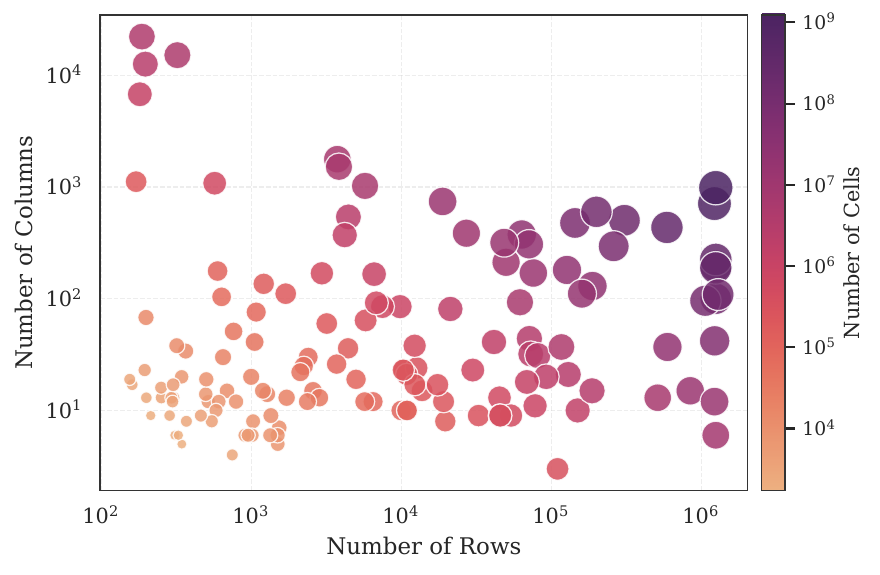}
        \end{subfigure}%
        \hfill
        \begin{subfigure}{0.48\textwidth}
            \includegraphics[width=\textwidth]{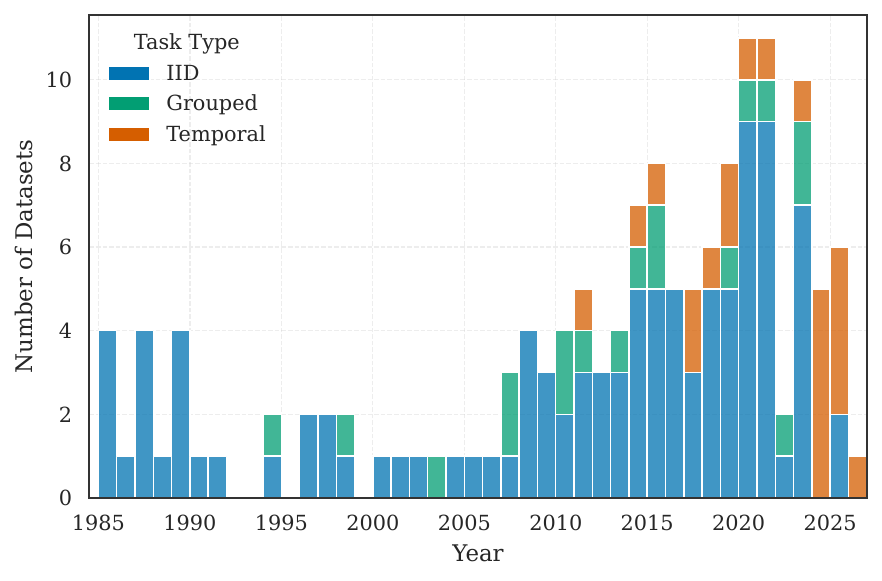}
        \end{subfigure}
        \begin{subfigure}{0.48\textwidth}
        \includegraphics[width=\textwidth]{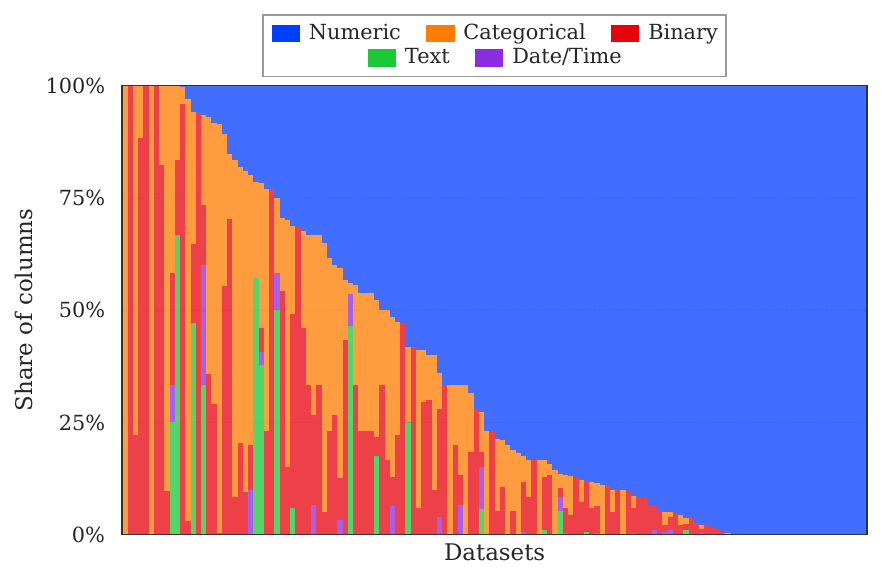}
        \end{subfigure}%
        \hfill
        \begin{subfigure}{0.48\textwidth}
            \includegraphics[width=\textwidth]{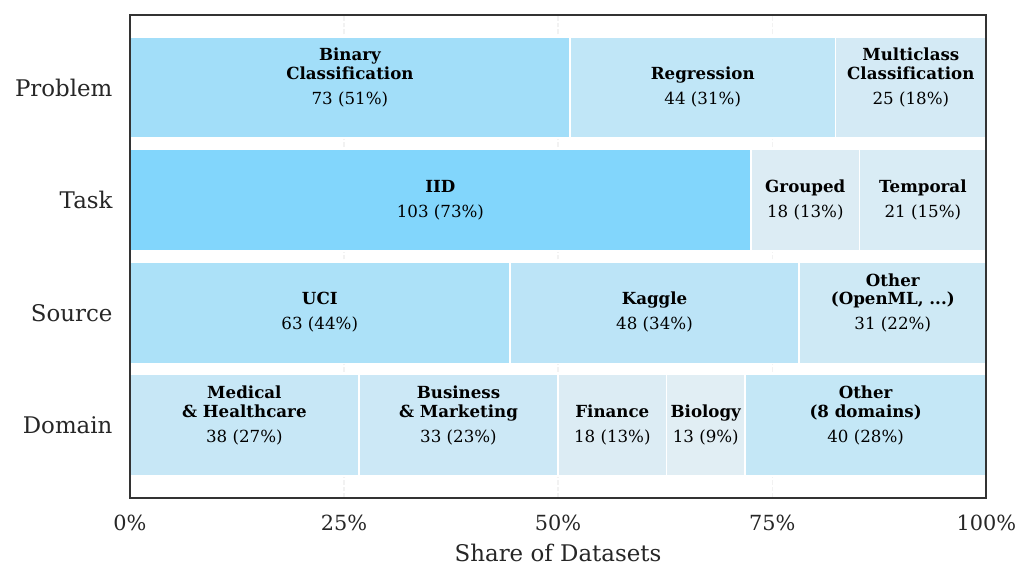}
        \end{subfigure}%
    \end{minipage}
    }
    \caption{
    \textbf{Dataset Dashboard.}
    \textbf{(Top Left):} Dataset sizes (w.r.t. rows, columns, and cells).
    \textbf{(Top Right):} Age distribution.
    \textbf{(Bottom Left):} Distribution of feature types per dataset.
    \textbf{(Bottom Right):} The share of datasets from a specific problem type, task type, dataset source, or application domain.
    }
    \label{fig:ds_dashboard}
\end{figure}

\section{Experimental Design}
\label{sec:experiment_setup}

We follow best practices for evaluating tabular machine learning models.
We compare state-of-the-art models with verified open-source implementations, perform reliable validation and tuning, use ablated preprocessing, employ robust metrics, and rigorously repeat our experiments. 

\myparagraph{BeyondArena Models.}
We compare $11$ tabular machine learning models: Linear / Logistic Regression, Random Forest~\citep{breiman-mlj01a}, Extremely Randomized Trees~\citep{geurts-ml06a}, CatBoost~\citep{prokhorenkova-neurips18a}, LightGBM~\citep{ke-neurips17a}, XGBoost~\citep{chen-kdd16a}, RealMLP~\citep{holzmuller2024better}, TabM~\citep{gorishniy2024tabm}, TabDPT~\citep{ma2024tabdpt}, TabPFN-2.6~\citep{grinsztajn2025tabpfn,priorlabs_tabpfn_2_6}, and TabICLv2~\citep{qu2026tabiclv2}.
We select models that performed competitively on the TabArena live leaderboard~\citep{tabarena_leaderboard}, TALENT~\citep{ye2024closer}, and TabReD~\citep{rubachev2024tabred}, including a linear model, tree-based models, neural networks, and TFMs.
We run TabPFN-2.6 and TabDPT only for datasets with up to $100k$ training samples.
For TabPFN-2.6, we stay within its designated pretraining limits. 
We limit TabDPT due to the infeasible cost of retrieval-based inference for benchmarking: for a dataset with 1 million samples under cross-validation, TabDPT needs 8 million forward passes, 8 per sample.
We imputed missing results for TabPFN-2.6 and TabDPT using the performance of a default Random Forest.

\myparagraph{Open-source Implementation.} We build \benchname on top of TabArena~\citep{erickson2025tabarena}, using its model implementations and benchmark infrastructure. 
As a result, all models we use were unit-tested, implemented in the standardized AutoGluon-based model framework \citep{erickson-arxiv20a}, and are open-source -- ensuring high-quality, reproducible benchmarking. 
We modified implementations to scale them to larger data and ensure proper behavior for non-IID tasks (e.g., handling unseen categories).
Moreover, we added non-IID validation protocols and new (multimodal) preprocessing to TabArena.

\myparagraph{Outer Splits.}
We create train-test splits for a dataset differently across sample sizes to guard against noise biasing the results while remaining conservative with the compute budget. 
We split and subsample datasets with more than 1.25 million samples into 1 million training and 250k test samples.
For \cIID tasks, we use 3-fold cross-validation for datasets with fewer samples.
We perform $n$-repeated 3-fold cross-validation with $n=3$ for 2500-250k training samples; $n=10$ for 500-2500; and $n=20$ for fewer than 500.   
For \cgrouped tasks, we do the same but utilize group-based cross-validation.
We stratify splits on the target variable for IID and grouped classification tasks.  
For \ctemporal tasks, we manually create application-specific temporal splits. The application determines the appropriate prediction time horizon for the test data. 
We roll back the time horizon to create multiple split time points for multiple train-test splits.
At each time point, we use all data before for training and all data after within the time horizon for testing.
We create the same number of splits as for IID or grouped tasks, but we do not use splits with $<50\%$ of the original data as training data. We deem a split with $<50\%$ too unrepresentative of the original application.

\myparagraph{Inner Validation and Tuning Protocol.}
After the outer train-test split, we split the training data into 8 folds for cross-validation (CV) to estimate model performance during tuning.
For datasets with $<500$ training samples, we use 5-repeated 5-fold CV to avoid overtuning~\citep{nagler-neurips24a,schneider2025overtuning}. 
We stratify the splits for classification. 
For IID tasks, we use random splits. For grouped tasks, we use grouped splits based on the group index.
For temporal tasks, we bin the time index and create intervals used as folds for CV, a popular procedure on Kaggle, and following \citet{bergmeir2012use}.
At test time, all models use all available training data: traditional models average the predictions of their per-fold models (a cross-validation ensemble), while TFMs use the full training data, including the validation folds, as in-context data; see \Cref{appendix:validation_usage}.
\\
We tune models using the search spaces by \citet{erickson2025tabarena}, due to a limited compute budget, with $25$ configurations from random search. 
For TFMs (TabDPT, TabPFN-2.6, TabICLv2), we evaluate only in-context learning without (fine-)tuning to assess their pretrained generality using the default preprocessing pipeline and no dataset-specific weight updates.
\\
We set a 4-hour time limit for CV, plus 1 hour of overhead for predicting on test data. 
We increase the time limit to 12 hours for TabPFN-2.6 and TabICLv2, as the 5-hour limit was insufficient for larger datasets. 
We also had to reduce TabICLv2's ensemble member count to $1$ for the $5$ largest datasets.
Unlike traditional models, existing TFMs do not support early stopping to respect predefined time constraints \citep{kuken2025early}. 
Only ${\sim}0.01\%$ of jobs exceed $3.5$ hours; see \Cref{appendix:time_limit_investigation} for details. 

\myparagraph{Multimodal and non-IID Preprocessing.}
We adjust TabArena's preprocessing pipeline to handle multimodal non-IID data.
We use a single general preprocessing pipeline (model-agnostic) and extensions to it per model (model-specific).
We retain the model-specific pipelines from TabArena, which handle, when needed, steps such as scaling or categorical encoding. 
Our new model-agnostic preprocessing pipeline handles date encoding, text encoding, and preprocessing for grouped data.
All preprocessing choices in our pipeline are benchmark design choices, not settled best practices: our ablations show that model rankings are sensitive to them (\Cref{tab:ablations}; \Cref{appendix:ablation_text_encoding,appendix:no_grouped_preprocessing}).
\\
We use skrub~\citep{skrub} to convert datetime features into $10$ numerical features, representing the weekday or a spline-based periodic encoding.  
We encode text features using \texttt{Qwen3-Embedding-8B}~\citep{qwen3embedding}, the best zero-shot multilingual text encoding model based on the MMTEB leaderboard~\citep{muennighoff2022mteb,enevoldsen2025mmtebmassivemultilingualtext}, into a 32-dimensional vector using Qwen3's smallest Matryoshka representation learning~\citep{kusupati2022matryoshka} slice. 
Lastly, we add preprocessing for grouped non-IID data.
For \texttt{label-per-sample} datasets, we drop the group-index column to prevent models from overfitting to it during training.
For \texttt{label-per-group} datasets, we replace the group index with a 50-dimensional group-encoding vector, similar to common practice in Kaggle competitions (e.g., see the top solutions in the AMEX competition \citep{Howard2022AmericanExpressDefaultPrediction}).
For more details on all preprocessing steps, see \Cref{appendix:experiment_setup_details}.

\myparagraph{Compute Resources.} 
We used CPU/GPU virtual machines via GCP~\citep{gcp_platform}; for specifications, see \Cref{appendix:experiment_setup_details}.
The estimated cost of \benchname was ${\sim}\$50k$ and took ${\sim}16.25$ wall-clock years.

\myparagraph{Metrics.} 
We chose metrics aligning with prior benchmarks \citep{gijsbers-jmlr24a,rubachev2024tabred,erickson2025tabarena} and machine learning literature \citep{provost1998case,gneiting2007strictly,fissler2022model}.
We use ROC AUC for binary classification, log-loss for multiclass classification, and RMSE for regression. 
We chose ROC AUC because it is invariant to threshold tuning for binary classification, unlike accuracy as used by TALENT~\citep{liu2024talent}.
For multiclass classification, we use log-loss because it is a proper scoring rule \citep{gneiting2007strictly,gneiting2011making}.
We use RMSE, a point prediction metric, for regression, and do not use probabilistic prediction metrics \citep{gneiting2014probabilistic,landsgesell2026scoringbench,landsgesell2026distributional}.
Point predictions are closer to the application's objectives across almost all datasets we investigated.
Thus, RMSE is more representative for practitioners.
We normalized and aggregated the model performances with Elo~\citep{elo1967proposed} and mean Improvability~\citep{erickson2025tabarena}, using implementations from TabArena.
Improvability ($0\%-100\%$) measures error relative to the best method on each outer split, making it sensitive to performance gaps.
Elo ratings are computed by modeling pairwise comparisons and reflect the probability of winning. 
We always calibrate $1000$ Elo to be the performance of a default XGBoost. 
\space
\begin{figure}[b]
    \centering
    \begin{subfigure}{0.56\textwidth}
        \includegraphics[width=\textwidth]{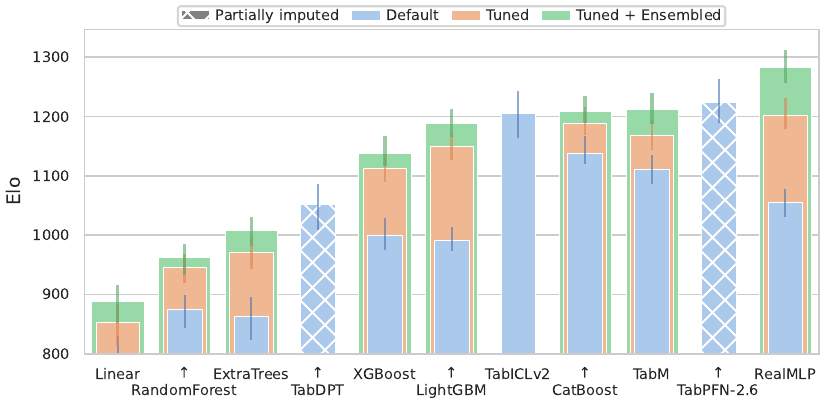}
    \end{subfigure}%
    \hfill
    \begin{subfigure}{0.36\textwidth}
        \includegraphics[width=\textwidth]{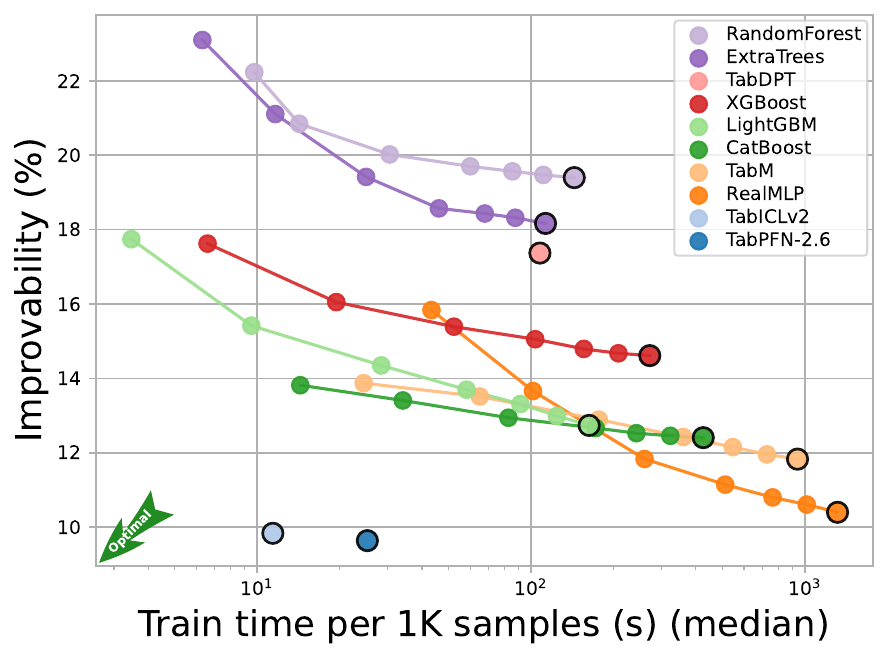}
    \end{subfigure}
    \caption{
    \textbf{BeyondArena Leaderboard (Left=Elo, Right=Improvability).} 
    We show the performance of $11$ models across $142$ datasets with the default hyperparameters, with tuning, and with tuning and post-hoc ensembling. The default performance of TFMs equals their in-context learning performance.
    }
    \label{fig:overall_best_model}
\end{figure}

\section{Results}
\label{sec:results}
\benchname aims to understand how well tabular foundation models would perform when practitioners use them in real-world predictive applications. 
Therefore, we first assess the extent to which current open-source TFMs, used via in-context learning, are truly general, and then investigate the validity of our results through ablation studies.
To investigate the performance with respect to dataset characteristics, we define 12 \textbf{sub-benchmarks}:  
one per task type (\textbf{IID}, \textbf{Grouped}, \textbf{Temporal});
four for dataset scale (\textbf{Tiny} for datasets with $100 \le n \le 1k$ training rows, \textbf{Small} for $1k < n \le 10k$, \textbf{Medium} for $10k < n \le 100k$, \textbf{Large} for $n > 100k$);
two for dataset dimensionality (\textbf{Low} for datasets with $m \le 100$ columns after preprocessing, \textbf{High} for $m > 100$);
two for special feature types (\textbf{Text} for datasets with text features, \textbf{High Card.} for datasets with at least one categorical column that has more than $50$ categories);
and finally the full benchmark containing \textbf{All} datasets.

\subsection{How General Are Current Tabular Foundation Models?}
\label{sec:lb_results}
\begin{figure}[b]
    \centering
    \includegraphics[width=\textwidth]{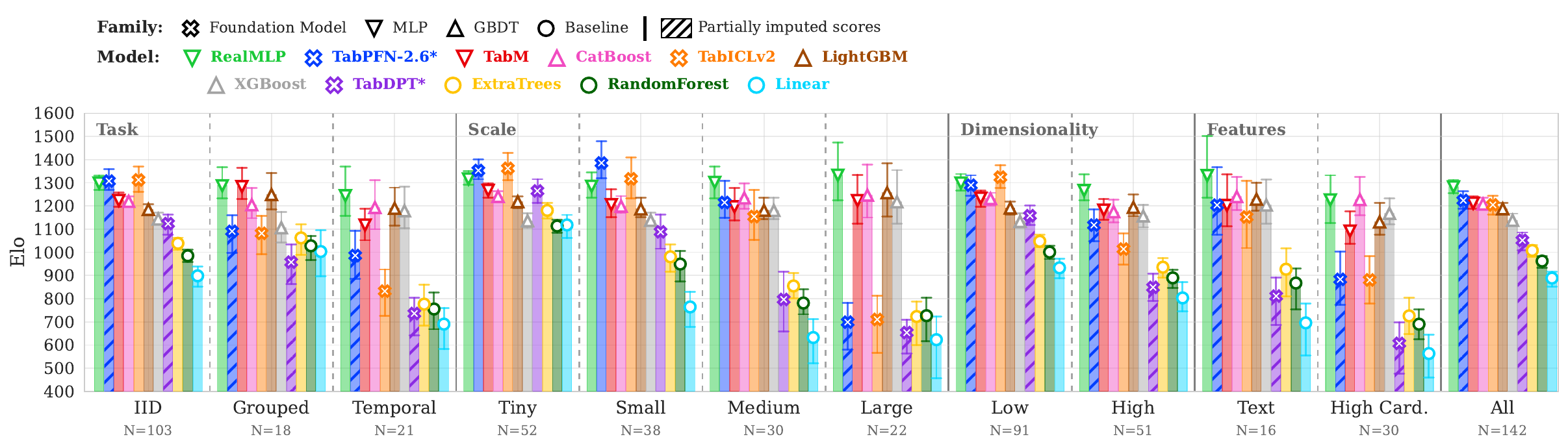}

    \caption{
    \textbf{Elo per model across sub-benchmarks.} 
    For each sub-benchmark of \benchname, we show the performance of each traditional model with tuning and post-hoc ensembling. For TFMs, we show their in-context learning performance.
    We aggregate results per model family in \Cref{fig:first_page}.
    }
    \label{fig:results_per_model}
\end{figure}

\Cref{fig:overall_best_model} shows the Elo and Improvability for all methods on all datasets. \Cref{fig:results_per_model} presents the Elo per sub-benchmark, and \Cref{fig:first_page} shows the Elo of the best model per family.
On the full benchmark, TFMs trail tuned traditional models on the non-IID, large, high-dimensional, and high-cardinality sub-benchmarks. As we show next, most of this gap is explained by dataset size.

\begin{figure}
    \centering
    \begin{subfigure}[b]{0.63\textwidth}
        \includegraphics[width=\linewidth]{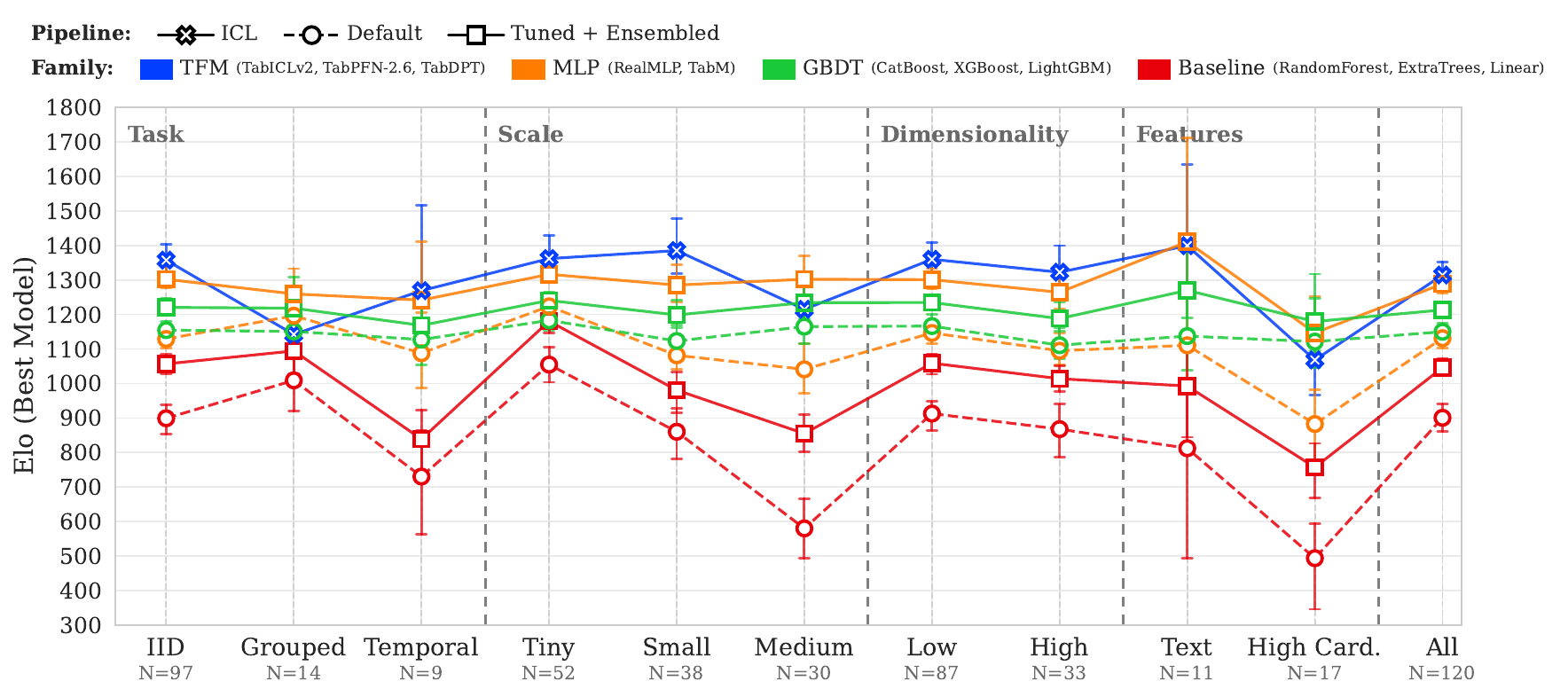}
    \end{subfigure}%
    \hfill
    \begin{subfigure}[b]{0.35\textwidth}
        \includegraphics[width=\linewidth]{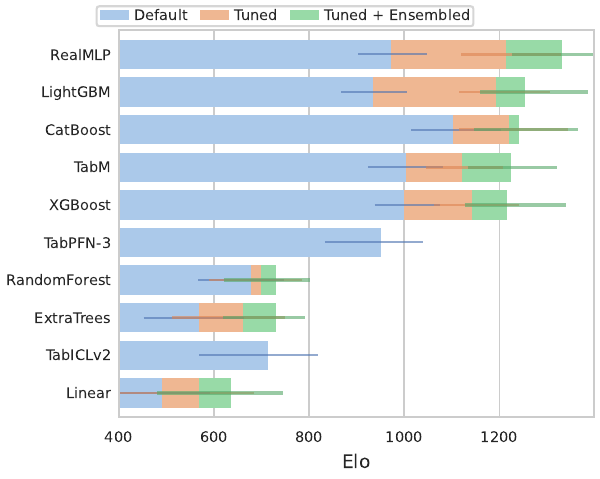}
    \end{subfigure}
    \caption{
    \textbf{Results without Imputation (Left=$n \le 100k$, Right=$n > 100k$).}
    \textbf{Left:} As \Cref{fig:first_page} but only the $120$ datasets with $n \le 100k$, where all TFMs run without imputation.
    \textbf{Right:} Elo on the $22$ datasets with $n > 100k$ with TabPFN-3 and excluding otherwise imputed TabPFN-2.6 and TabDPT.
    }
    \label{fig:robustness_imputation}
\end{figure}

\myparagraph{When does TFM performance match or trail traditional models?}
TFMs perform best on small, tiny, and \cIID datasets (\Cref{fig:first_page}). Moreover, TFMs rank highest on average in Improvability and default performance (\Cref{fig:overall_best_model}, leaderboard as table in \Cref{appendix:result_tab}). 
However, traditional approaches -- particularly deep learning models such as RealMLP -- benefit substantially from tuning and ensembling.
In aggregation, the in-context learning performance of all TFMs fails to compete with traditional models on non-IID (\ctemporal and \cgrouped), large-scale, high-dimensional, and high-cardinality datasets (\Cref{fig:results_per_model}).
To ground our findings in formal statistical evidence, we complement them with a suite of non-parametric tests in \Cref{app:posthoc}, showing that the global ranking and pairwise comparisons across models and sub-benchmarks are significant.
\\
In general, as sub-benchmarks are not independent (\Cref{sec:curation_outcome}), bad performance on one (e.g., large) might also explain bad performance on another (e.g., temporal).
To disentangle this, we investigate how imputed TFMs and correlated sub-benchmarks affect our claims.
Restricting the benchmark to the $120$ datasets with $n \le 100k$ training rows, where no imputation is needed, makes TFMs close the gap to the traditional baselines across several subsets (\Cref{fig:robustness_imputation}, Left).
\Cref{fig:robustness_imputation} (Right) shows that TabICLv2 and TabPFN-3~\citep{grinsztajn2026tabpfn3technicalreport}, which we added for this analysis, fall far behind on large datasets; both run on large datasets without imputation. 
In addition, a multivariate analysis including TabPFN-3, where we fit a linear model predicting the rank of the best TFM per dataset from its task type, size, dimensionality, and feature types, shows that only two properties are significant: more than 100k training rows and grouped splits; for details, see \Cref{appendix:multivariate}.
Hence, after controlling for size, temporal structure, high dimensionality, and high-cardinality features are not significant; their apparent effect reflects that $57\%$, $35\%$, and $43\%$ of these datasets exceed $100k$ rows, compared with $8\%$, $4\%$, and $8\%$ of the remaining datasets.
\textbf{
In summary, current TFMs are competitive below 100k training rows, while large and grouped data remain the main open challenges.
}

We also observe that TFMs' default training costs, including cross-validation, remain significantly lower than those of their tuned competitors (\Cref{fig:overall_best_model}, right), while inference costs remain high (\Cref{appendix:inference_time}).
We present results per dataset in \Cref{appendix:results_per_dataset}.

\myparagraph{How often do TFMs achieve peak performance?}
In \Cref{fig:rank_one_plot}, we move past results in aggregation and investigate the ability of TFMs to achieve peak performance (rank 1) on individual datasets, showing that  
\begin{enumerate*}[label=(\textbf{\arabic*})]
    \item TabICLv2 ranks first on $19\%$ of all datasets, followed by TabPFN-2.6 ($10.5\%$).
    \item TFMs significantly outperform other models on $50\%$ of datasets, and at least one TFM matches the best non-TFM on an additional $20\%$, making TFMs a viable way to achieve peak performance on $70\%$ of all datasets.
    \item On grouped, temporal, and large datasets, TFMs attain peak performance only on a small fraction of datasets.
\end{enumerate*} 
Overall, TFMs are clearly outperformed on $42$ datasets that are large, high-dimensional, non-IID, or include high-cardinality categorical features, underscoring the need for unified, holistic benchmarks to improve TFMs across a wider variety of datasets. 


\subsection{Ablations on Benchmark Design Choices}
\label{sec:benchmark_science}

\begin{figure*}
    \centering

    \begin{minipage}[c]{0.48\textwidth}
        \centering
        \includegraphics[width=\linewidth]{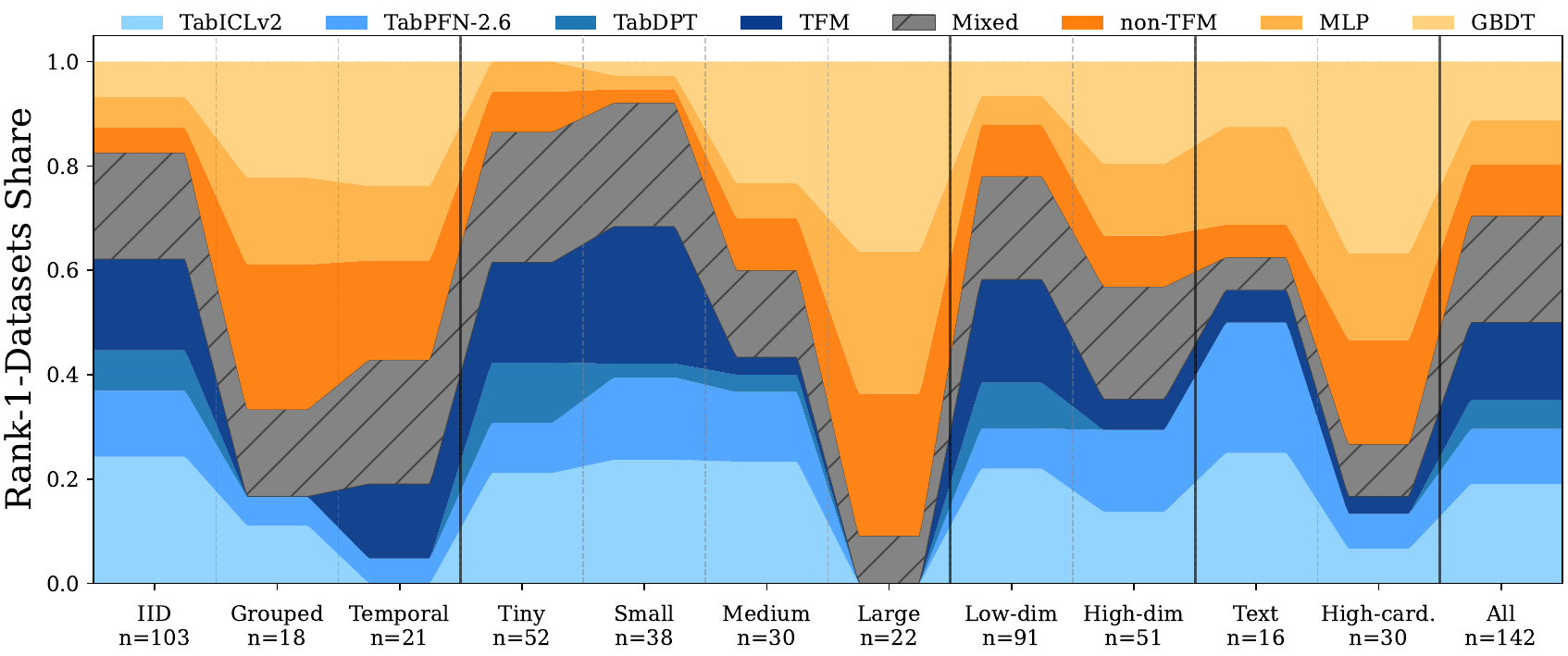}
        \caption{\textbf{Rank-1 share across sub-benchmarks}. 
        We show the share of datasets where each model (family) significantly outperforms the others. 
        Significance is measured by a Wilcoxon signed-rank test with $\alpha=0.05$ for datasets with ${>}3$ outer splits and by whether a model wins all splits otherwise. 
        \textbf{MLP}, \textbf{GBDT}, win for the family;
        \textbf{TFM}, win for the family, no winner within the family;        \textbf{non-TFM} no clear winner for the best GBDT and MLP; 
        \textbf{Mixed} no clear winner for the best TFM and non-TFM.
        }
        \label{fig:rank_one_plot}
    \end{minipage}
    \hfill
    \begin{minipage}[c]{0.48\textwidth}
        \centering

        \captionof{table}{\textbf{Ablation Results.} 
        We show Kendall's tau ($\tau$) \citep{kendall1938new} and the win rate when ablating our experimental setup. 
        $\tau \in [-1, 1]$ measures the difference in ranking, with $1.0$ being identical ranks.
        Win rate shows how often \benchname's setting leads to a higher average rank across models. 
        We split ablations for grouped data: 
        \texttt{label-per-group} (\texttt{L-P-G}) / 
        \texttt{label-per-sample} (\texttt{L-P-S}); 
        and text data: $\le50$ characters on average (short) / $>50$ (long).
        We share details in \Cref{appendix:ablations}.}

        \label{tab:ablations}

        \vspace{0.5em}

        \resizebox{0.85\textwidth}{!}{%
        \begin{tabular}{lllcc}
        \toprule
        Part & Ablation & Setting (\benchname $\to$ new) & Kendall $\tau$ & Win Rate \\
        \midrule
        
        \multirow{2}{*}{Inner Splits}
            & \hyperref[ablation:b1]{\textbf{(B.1)}} & 5$\times$5-fold CV $\to$ 8-fold CV & 0.93 & 100\% \\
            & \hyperref[ablation:b2]{\textbf{(B.2)}} & Non-IID $\to$ IID & 1.00 & 100\% \\
        
        \midrule
        
        \multirow{2}{*}{\shortstack[l]{Grouped Data\\Pre-processing}}
            & \hyperref[ablation:c1]{\textbf{(C.1a)}} & (L-P-G) Agg. Index $\to$ N/A & 0.81 & 71\% \\
            & \hyperref[ablation:c1]{\textbf{(C.1b)}} & (L-P-S) Drop Index \,$\to$ N/A & 0.43 & 71\% \\
        
        \midrule
        
        \multirow{2}{*}{\shortstack[l]{Text Data\\Pre-processing}}
            & \hyperref[ablation:c2]{\textbf{(C.2a)}} & (Short Text) Qwen3 $\to$ TF-IDF & 1.00 & 78\% \\
            & \hyperref[ablation:c2]{\textbf{(C.2b)}} & (Long Text) Qwen3 \,$\to$ TF-IDF & 0.89 & 6\% \\
        
        \midrule
        
        \multirow{1}{*}{Post-processing}
            & \hyperref[ablation:d]{\textbf{(D)}} & N/A $\to$ Probability Calibration & 0.85 & 18\% \\
        
        \bottomrule
        \end{tabular}
        }
    \end{minipage}

\end{figure*}

To better understand \benchname and inform future benchmarks, we ablate our experiment setup as described in \Cref{sec:experiment_setup}.
We summarize the results for the majority of our ablations in \Cref{tab:ablations}.

\myparagraph{Different Outer Splits.}
\textbf{(A.1)} We use up to $60$ outer validation splits in \benchname, which drastically increases compute costs. 
To reduce environmental impact and enable compute-constrained researchers, we would like to use fewer outer splits.
\cite{erickson2025tabarena} introduced \texttt{TabArena-Lite}, saving compute by using only the first split per dataset, which produces a comparable ranking \textit{on average}. 
We instead introduce \texttt{BeyondArena-Core}, by automatically selecting a subset of splits per dataset, ensuring that the ranking \textit{per dataset} on the subset is comparable to that obtained with all splits.  
Compared to \benchname, \texttt{-Lite} would be $\times9$ and \texttt{-Core} $\times5$ faster. 
Yet, the per dataset win rates of \texttt{-Core} are $\times2.3$ more stable than \texttt{-Lite}; see \Cref{appendix:ablation_fewer_splits} for details.
\\
\textbf{(A.2)} Besides using fewer outer splits, we ablate using IID splits for non-IID data. 
Prior work has shown that inappropriate splits confound the evaluation of \ctemporal data, cf. \citep{rubachev2024tabred,tschalzev2025unreflected}.
In \Cref{appendix:ablation_iid_splits}, we show that, for two exemplary grouped datasets, inappropriate outer splits can drastically distort model rankings ($\tau=0.49$ and $\tau=0.60$) and raw performance estimates. 
Hence, it is pivotal to use appropriate non-IID splits when evaluating \cgrouped data.

\myparagraph{Different Inner Splits.}
\textbf{(B.1)\label{ablation:b1}} \benchname uses 5-repeated 5-fold inner cross-validation ($5 \times 5$ CV) for datasets with fewer than 500 training samples, and 8-fold CV otherwise, to avoid overtuning hurting test performance \citep{nagler-neurips24a,schneider2025overtuning}.
\Cref{tab:ablations} and \Cref{appendix:ablation_less_inner_splits} show that $5 \times 5$ CV consistently achieves a higher test performance, verifying the correctness of our setup. 
\\
\textbf{(B.2)\label{ablation:b2}}
We use non-IID inner splits for non-IID tasks. 
\Cref{tab:ablations} shows that using IID inner splits results in similar ranks, but non-IID splits perform better on average and can reach significantly higher peaks (see \Cref{appendix:ablation_iid_inner_splits}).
Thus, using non-IID inner splits for temporal and grouped data was appropriate.

\myparagraph{Different Preprocessing.}
We use preprocessing for grouped data to handle the group index column.
We ablate using only the raw data in \Cref{tab:ablations} and \Cref{appendix:no_grouped_preprocessing}.
\textbf{(C.1a\label{ablation:c1})} For \texttt{L-P-G} data, the model rankings are mostly stable, while \textbf{(C.1b)} the rank ordering is mixed up for \texttt{L-P-S} data.
In both cases, \benchname's preprocessing is better on average, yet TabPFN-2.6 improves substantially when preprocessing is disabled (\ref{appendix:no_grouped_preprocessing}). 
The results invite future work.
For \benchname, we conclude that the performance of most models is representative, whereas TabPFN-2.6 is negatively biased. 
\\
\textbf{(C.2.a/b)\label{ablation:c2}} We used \texttt{Qwen3-Embedding-8B} to encode text features. 
We compare Qwen3 to TF-IDF from skrub~\citep{skrub}, a classical text encoder in \Cref{tab:ablations} and \Cref{appendix:ablation_text_encoding}.
For short-text datasets, Qwen3 performs better than TF-IDF.
For long texts, TF-IDF generally yields better predictive performance.
Yet the model rankings are only slightly affected; thus, our results remain representative.
\\
In the future, we recommend including preprocessing in the search space for grouped and text data. 

\myparagraph{Probability Calibration for Log-loss.}
\textbf{(D)\label{ablation:d}} We use log-loss as a metric for multiclass classification.
Log-loss is affected by probability calibration, and thus, post hoc calibration methods can often improve performance. 
\benchname does not use post hoc calibration by default; we ablate adding it based on the saved predictions. 
In \Cref{tab:ablations}, we show that calibration would have performed better most of the time. Across all models, only TabPFN-2.6 and RealMLP perform worse with calibration (\Cref{appendix:ablation_probabiltily_calibration}).
Thus, we recommend using calibration by default for other models going forward.

\space
\section{Conclusion}
\label{sec:conclusion}

We introduced \benchname, a unified benchmark for predictive modeling on tabular data that consolidates diverse task types (\cIID, \ctemporal, \cgrouped), dataset scales, feature dimensionality, and feature types. Together with \df, a framework for reproducible dataset curation, we enable standardized evaluation across $142$ datasets and $11$ models. 
Our results show that current open-source tabular foundation models (TFMs), used via in-context learning, perform well on datasets with up to 100k training rows but are outperformed by tuned traditional methods on large-scale or grouped data. 
In summary, TFMs generalize well to existing paradigms and show very promising in-context learning performance, but they are not yet fully general for tabular data.

\myparagraph{How to Use BeyondArena in Research.}
\label{sec:how_to_use}
\benchname's protocol reflects one recommended setting for the research question we study, not the only valid use of our curated datasets. \df provides each dataset and its metadata independently of our protocol, and our benchmark code supports alternative inner validation, tuning, or preprocessing. Following TabArena~\citep{erickson2025tabarena}, we value a method even if it is not state-of-the-art on the full benchmark: any sub-benchmark or subset large enough to support a claim is a valid basis for a contribution. 
Likewise, no single metric is mandatory, and authors should report whichever of Elo, Improvability, rank-1 rate, or time-to-performance best fits their question; absent a specific preference, we recommend Elo. 
Finally, we recommend using \benchname-Core, which approximately preserves per-dataset rankings and requires roughly only one-fifth of the compute (\Cref{appendix:ablation_fewer_splits}).

\myparagraph{Limitations and Societal Impact.}
Our conclusions are limited by:
\begin{enumerate*}[label=(\textbf{\arabic*})]
    \item We tune models for $25$ random configurations due to compute constraints.
    \item We evaluate in-context learning performance of TFMs, but do not investigate fine-tuning TFMs.
    \item We include the top three TFMs from TabArena, plus TabPFN-3, in our analysis without imputation; however, we do not explore other open- or closed-source models.
    \item Across the $142$ datasets, some sub-benchmarks have more datasets than others and are correlated, so some are under- and over-represented in the leaderboard computed across all datasets.   
    \item We make the first step towards a unified pipeline for preprocessing, validation, and tuning across task types and feature types. 
    Yet, the best practices for validation protocols for temporal and grouped data, as well as for preprocessing grouped and text data, remain to be explored.
\end{enumerate*}
On the societal side, \benchname improves rigor and reproducibility, narrowing the gap between academics and practitioners at non-trivial computational cost -- traded off by using \texttt{BeyondArena-Core}.

\myparagraph{Future Work.}
To broaden the scope of evaluations of TFMs and further explore their generalizability, future directions could extend \benchname to few-shot predictions, more multimodal tabular data (e.g., tabular-image-text), and other tasks such as relational learning or survival analysis.

\textbf{To conclude,} \benchname surfaces critical limitations of current TFMs and establishes a new evaluation standard for tabular machine learning on broader, more demanding challenges.
\space

\newpage
\begin{ack}

L.P. acknowledges funding by the Deutsche Forschungsgemeinschaft (DFG, German Research Foundation) under SFB 1597 (SmallData), grant number 499552394.
A.T. acknowledges funding by the German Federal Ministry for Economic Affairs and Energy (BMWE).
A.A. is supported by an Israel Ministry of Science and Technology (MOST) grant on multi-modal AI.
A.P. was funded by the European Union. Views and opinions expressed are however those of the author(s) only and do not necessarily reflect those of the European Union or the European Commission. Neither the European Union nor the European Commission can be held responsible for them. This work was supported by the European Union’s Horizon Europe research and innovation programme under grant agreement No 101214398 (ELLIOT).
\begin{center}
\begin{minipage}{0.3\textwidth}
\centering
\includegraphics[width=\textwidth]{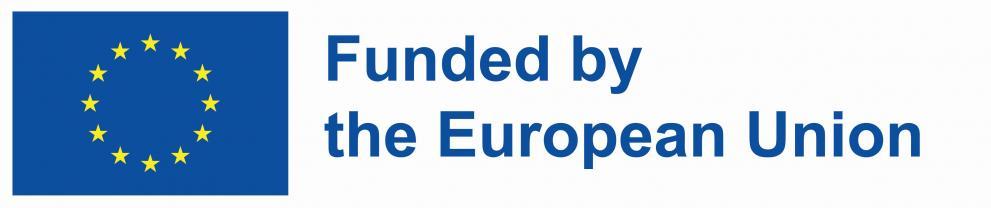}
\end{minipage}
\begin{minipage}{0.3\textwidth}
\centering
\includegraphics[width=\textwidth]{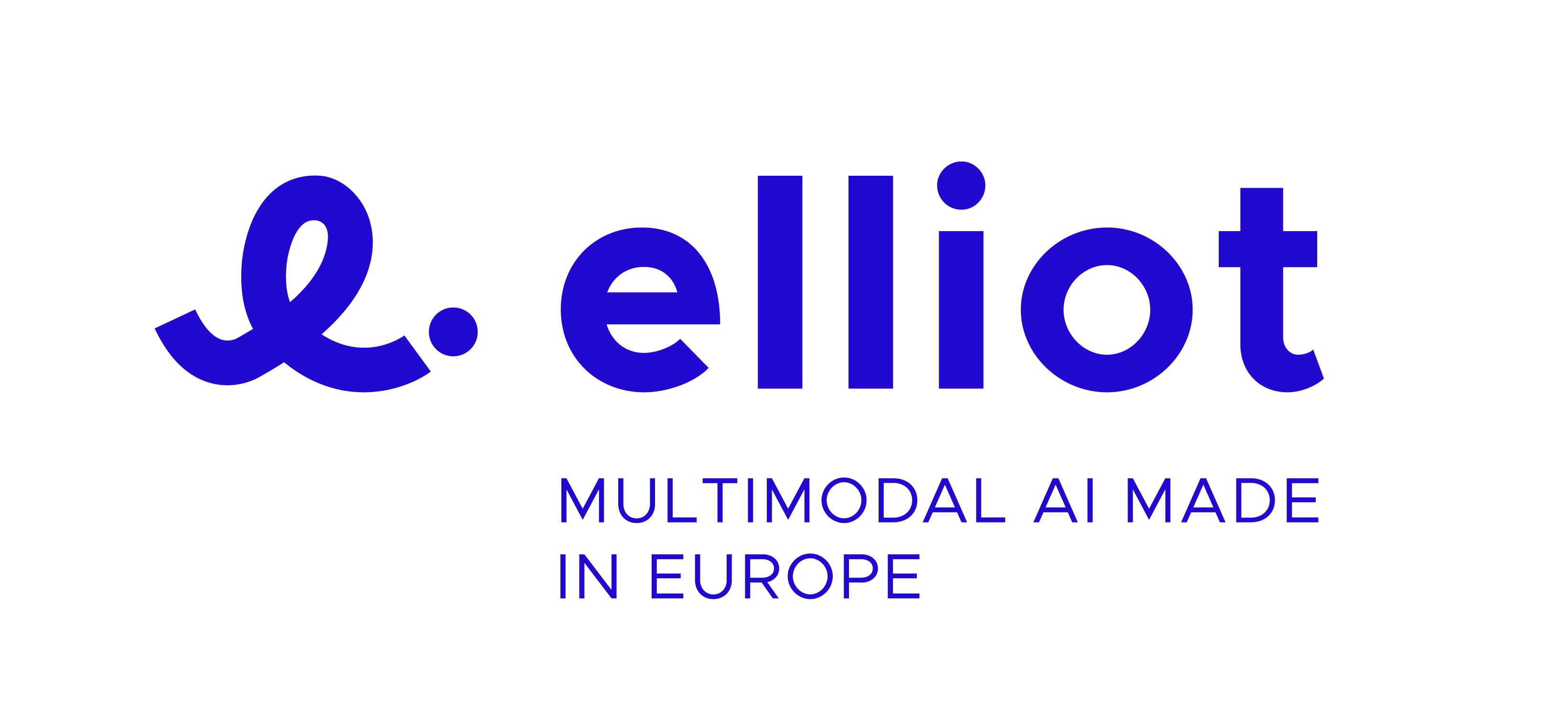}
\end{minipage}
\end{center}
M.T. is supported by the Konrad Zuse School of Excellence in Learning and Intelligent Systems (https://eliza.school/) through the DAAD programme Konrad Zuse Schools of Excellence in Artificial Intelligence, sponsored by the Federal Ministry of Education and Research.
G.V. acknowledges support from ANR via grant TaFoMo (ANR-25-CE23-1822). This work is partly supported by Hi! PARIS and ANR/France 2030 program (ANR-23-IACL-0005).
F.H. acknowledges the financial support of the Hector Foundation.
Finally, we thank the reviewers for their constructive feedback and contribution to improving the paper.

\section*{Author Contributions}
L.P. led the code development, led the experiments, led the writing of the paper, led the evaluation and visualizations, led dataset curation, and led the management of the collaboration.
A.T. contributed to the writing of the paper, contributed to evaluation and visualizations, and contributed to dataset curation.
N.E. contributed to the writing of the paper and contributed to the evaluation and visualizations.
G.B. helped with writing the paper, contributed to the evaluation and visualizations, helped with dataset curation, and helped with reviewing and editing.
D.H. contributed to writing the paper, helped with the management of the collaboration, and helped with supervising the project. 
A.A. helped with writing the paper, and contributed to reviewing and editing.
A.P. helped with dataset curation.
M.T. helped with dataset curation.
G.V. helped with evaluation and visualizations, helped with management of the collaboration, and supervised G.B. during the project.
F.H. supervised L.P. during the project.

\section*{Competing Interests}
D.H. is one of the authors of RealMLP. 
L.P. and F.H. are a subset of the authors of TabPFN-2.6 and TabPFN-3. 
D.H. and G.V. are a subset of the authors of TabICLv2.
L.P., A.T., N.E., and D.H. are the maintainers of TabArena.  
L.P. and N.E. are maintainers of AutoGluon.
G.V. is a core contributor to skrub.
G.V. is (co)founder of scikit-learn.
L.P., N.E., A.A., and F.H. are affiliated with Prior Labs, a company focused on developing tabular foundation models. 
G.V. is affiliated with probabl, a company focused on developing software for industrial-grade data science. 
The authors declare no other competing interests.
\end{ack}
\space
\bibliography{strings,lib,proc,local,dataset_references}
\bibliographystyle{unsrtnat}

\newpage
\begin{appendices}

\listofappendices
\counterwithin{figure}{section}
\counterwithin{table}{section}
\crefalias{section}{appendix}
\crefalias{subsection}{appendix}

\newpage

\section{Background}

\subsection{Time-series Forecasting Tasks vs. Temporal Tabular Tasks}
\label{appendix:tsfvsttt}
We argue that there is a fundamental difference between the definitions of time-series forecasting and temporal tabular data tasks, yielding distinct validation requirements. In short: time-series forecasting tasks predict \textit{for} the future; temporal tabular tasks predict \textit{in} the future -- thus, when validating a model, it cannot or can use data from the future depending on the task. 
\\
To illustrate, assume, we have a time index $t \in T$ and are given training data $D_{\text{past}} = (\{ x_s \in X : s \le t \}, \{ y_s \in Y : s \le t \})$ with the goal to learn to predict $Y$. 
In time-series forecasting tasks with a forecast horizon $h$, we fit on $D_{\text{past}}$ and predict $Y_{\text{future}} = \{\, y_s \in Y \;:\; t < s \le t + h \,\}$ -- at time point $t$.
In tabular temporal data tasks, we fit on $D_{\text{past}}$ at time point $t$ and predict $Y_{\text{future}} = \{\, y_s \in Y \;:\;  t < s \,\}$ at time point $s$, with $s \in T,\, s> t$, using $D_{\text{past}}$ and $X_{\text{future}} = \{\, x_s \,\}$. 
That is, in temporal tabular tasks, we fit a model at time point $t$ and deploy it to predict at time point $s$ using data from $s$.
Moreover, such tasks do not have a forecast horizon but a refit horizon, which determines the time until we refit the model on a new $D_{\text{past}}$.

While the many nuances and language differences between the two fields of structured data are out of scope here, the key difference for us lies in how the task affects the data. 
When naively using data from time-series forecasting tasks, one would violate the assumptions underlying the data and its associated validation process.
For example, if we were to use a forecasting dataset and validate a model as if we were predicting at time points $s$ instead of $t$, the validation would be unrepresentative of the dataset's task. 
Moreover, it becomes questionable what the validation would tell us; for many forecasting tasks, one could just read off the target variable (e.g., the stock price) at time points $s$. 
When this is the original intended task, we argue that it is a temporal tabular task and not a forecasting problem.  

For us, the distinction between time-series forecasting data and temporal tabular data is important during data curation.
If we naively use time-series forecasting data for temporal tabular regression, we ignore the real-world context of the data.
Moreover, the validation procedure would not accurately reflect how well a real-world tabular temporal model performs, but instead estimates how well the model can solve a task, with at best questionable relevance in reality. 
As we aim to curate data and benchmark models such that the results are as representative of the real-world as possible, we create and use this distinction between data from time-series forecasting and temporal tabular task.

\newpage
\section{Dataset Curation}

\subsection{Towards Automated Data Curation?}
\label{appendix:automated_data_cuartion}
We investigated (semi-)automated procedures to aid dataset selection, but found that most of our selection criteria cannot be faithfully applied -- not due to a lack of tooling, but to a significant lack of documented, structured metadata required for curation.
In most cases, the data and its context from the data repository do not contain the necessary information to make decisions. 
Moreover, asking agentic LLMs to determine the metadata led to hallucination and faulty judgment. 
Instead, subjective human judgment and collaborative discussion were required to obtain the metadata for our final decisions. 
We found agentic tooling most useful for documenting metadata created by humans. 
In particular, we used Agent Skills~\citep{agentskills_repo} via Claude Code~\citep{anthropic_claude_code} to fill metadata templates from human input.
Nevertheless, the future of high-quality automated data curation is promising, and we strongly believe that human-curated metadata and standardized schema, like in \df, will enable us to codify data curation. 
Therefore, we also share our agent skill markdown file as part of \df.

\subsection{Details on Dataset Selection Criteria}
\label{appendix:dataset_selection_criteria}

Below, we provide a detailed description of each criterion used in \benchname.  
Our selection criteria are based on the work by \citet{erickson2025tabarena}. 
Moreover, our criteria involve subjective human interpretation. Thus, we publicly share our curation insights and decisions for each dataset at \href{https://tabarena.github.io/data-foundry/}{\texttt{https://tabarena.github.io/data-foundry/}}. 

\begin{description}
    \item[Unique Dataset:] We ensure datasets in \benchname are unique to avoid biasing our results towards a particular distribution. 
    To do so, we ensure that each dataset in \benchname has a unique original data source.
    \\
    When we started evaluating new datasets from OpenML and Kaggle, we quickly realized that the only way to identify duplicates was to determine the original source of each dataset, because many datasets were re-uploaded to OpenML and Kaggle under different names and without proper attribution to the original source.     
    In particular, on Kaggle, we noticed that gamified incentives (such as upvotes and medals) might have even prompted users to avoid making their data look like duplicates, so that new, original contributions result in a larger reward. At the same time, Kaggle offers a discussion board for datasets, which enables the community to call out bad behavior and data quality issues.
    As a result, given a dataset from any data repository, our first priority is to determine its original source, or if it's an original contribution. 
    In many cases, this is trivial; in others, it requires prolonged investigation, including comparing data statistics and comparing the dataset to popular datasets from the same domain. 
    When the source remains unclear even after a thorough investigation, we leave it to the curator team's subjective judgment to assess the dataset's authenticity and uniqueness. 

    \item[No Few-shot Prediction Tasks:] We exclude datasets that have fewer than $100$ train samples after creating the train-test split. 
    Such data, often called few-shot prediction tasks, require a fundamentally different validation protocol, as data scarcity creates unique challenges, such as the inability to split the data for cross-validation. 
    At the same time, it enables unique solutions to be highly competitive, such as using LLMs for tabular machine learning \citep{hegselmann2023tabllm,gardner2024large,shysheya2025jolt,schindler2025tabgemma,lee2025range}.
    While our benchmark focuses on $1\,000\,000$ training samples, we do not exclude larger datasets; instead, we sub-sample them. 
    
    \item[IID and non-IID Tabular Datasets:] We include any tabular dataset whose original task requires a random, temporal, or grouped split. 
    We exclude datasets from time-series forecasting tasks because benchmarking such data in a representative way for their real-world context would require a fundamentally different validation protocol than data from temporal tabular tasks (see \Cref{appendix:tsfvsttt}).
    We leave benchmarking tabular machine learning models for time-series forecasting tasks to benchmarks from the time-series forecasting community \citep{aksu2024gift,shchur2025fev,garza2026impermanent,qiao2026s}.  
    
    \item[Predictive Machine Learning Tasks for Classification or Regression:] 
    We exclude datasets that do not originally stem from a predictive machine learning task for classification or regression.
    Following TabArena, we exclude data from scientific discovery tasks (e.g., survey data or non-predictive tables).
    For 
    \benchname, we also exclude data from other tabular-adjacent tasks, in particular, click-through rate prediction tasks \citep{guo2017deepfm,zhu2021open,zhang2021deep} and ranking prediction or information retrieval tasks used in recommender systems \citep{shani2010evaluating,steck2013evaluation,said2014comparative,zhang2019deep}. 
    Like for data from time-series forecasting tasks, datasets from such tasks require specific validation protocols to represent their real-world context appropriately. Furthermore, such tasks have their own benchmarking communities.

    \item[Representative for Real-world Applications of Tabular Machine learning:] 
    \benchname aims to understand how well tabular machine learning models would perform when practitioners use them in real-world applications.
    Consequently, we exclude datasets that represent tasks for which tabular machine learning models would not be used.
    Moreover, we exclude datasets that do not represent real-world applications or cannot be preprocessed to be representative. 
    Practically speaking, this means we exclude datasets that 
    \textbf{(A)} are from non-tabular modalities, where modality-specific models are clearly superior;
    \textbf{(B)} do not stem from a real random distribution;
    \textbf{(C)} are trivial;
    \textbf{(D)} have irreversible data quality issues;
    \textbf{(E)} and for which we did not find sufficient information to make an informed decision. 
    \\
    \textbf{(A)} We exclude datasets from non-tabular modalities, where modality-specific models are clearly superior.
    However, unlike TabArena, we determine whether modality-specific models are superior for each dataset rather than excluding all datasets from non-tabular modalities. 
    Consequently, we allow the inclusion of vectorized image, text, audio, or time-series data if tabular machine learning models are competitive with modality-specific solutions. 
    We investigate recent benchmarks related to the application of the dataset or introductory paper to determine if modality-specific solutions are clearly superior. 
    Even if we're not domain experts, we believe our curation team is sufficiently well-versed in interpreting academic papers to determine whether tabular models are competitive with modality-specific solutions.    
    \\
    \textbf{(B)} We exclude datasets that do not stem from a real random distribution.
    In line with the TabArena definition, we exclude artificial data and data generated by a deterministic function.
    Unlike TabArena, we believe that simulated data (e.g., the Higgs bosons dataset \citep{baldi2014searching}) represent real random distributions from important real-world applications. 
    Nevertheless, we do not include such datasets in \benchname, as such tasks usually have dedicated domain-specific benchmarks \citep{adam2015higgs,bhimji2025fair}.
    \\
    \textbf{(C)} We exclude datasets that are trivial because practitioners would not use machine learning to solve trivial datasets.
    A dataset is deemed trivial if all models, without tuning, consistently achieve the same better-than-random score. 
    That is, there is no variance, or all models solve the dataset perfectly (error of $0$). 
    We utilize the performance of models from \benchname.
    \\
    \textbf{(D)} We exclude datasets that have irreversible data quality issues. Many datasets shared online are already preprocessed; some have been processed in ways that make them unusable for benchmarking. Thus, we follow TabArena and exclude datasets where irreversible preprocessing leaks the target or test feature distribution (e.g., PCA-transformed data).
    \\
    \textbf{(E)} We exclude datasets for which we did not find sufficient information to determine whether the datasets would pass our selection criteria. Datasets shared online often come with only limited information about their source and significant ambiguity about their authenticity. In such cases, if a prolonged investigation failed to identify any source information, it was up to the curation team to decide whether the data should be used for benchmarking. 

    \item[Ethically Unambiguous Tasks:] We exclude datasets with tasks that pose ethical concerns. As part of \benchname, we extend the scope of ethical concerns to include cases in which data subjects or data creators request that the data not be used for machine learning research, such as the Pima Indian Diabetes Dataset (PIDD) \citep{radin2017digital}.
\end{description}

\subsection{Details on Dataset Processing}
\label{appendix:dataset_curation_guidelines}

All datasets underwent a processing procedure designed to create consistency and add reproducibility. Across datasets, we observed some recurring processing patterns, which we share below.
Case-by-case details and comments for the processing can be found in \df. 

\begin{description}
    \item[General:] 
    Many datasets come with unique sample identifiers. We drop all uninformative sample identifiers. Informative sample identifiers, such as a time index, were kept and processed to represent their original meaning (e.g., time) where possible. In many cases, sample identifiers were an artifact of data storage. 
    \\
    For similar reasons, many datasets contained missing values represented by proxy values (e.g., \texttt{999}, \texttt{-1}).
    We converted these to explicit missing values (\texttt{NA}) whenever such encoding could be reliably inferred from the data description and task context.
    \\
    For numerical target variables with strong skewness and heavy tails (e.g., housing prices), logarithmic scaling was considered on a case-by-case basis.
    \\
    Dataset names were standardized using \texttt{snake\_case} naming conventions.
    \\
    We always shuffle the sample order for both IID and grouped data to avoid methods exploiting implicit order leakage.
    For temporal data, samples were sorted chronologically by their time index. 
    \item[Feature Type Assignments (Dtypes):]
    We transformed features with object or string representations to categorical or string type. We used a categorical type when the number of possible values was fixed and finite (as inferred from the data description or task context). Otherwise, we used a string type.
    \\
    We converted date features to standardized \texttt{YYYY-MM-DD} datetime representations whenever possible. For date features originally encoded numerically, we reconstructed as much of the date information as possible.
    \\
    All other features were encoded as numerical types.
    \item[Creating Temporal Tasks and Splits:]
    For temporal prediction tasks, we manually define the prediction horizon and the associated test time points. 
    Then, we verified that each feature was valid to use, i.e., the information from the feature would have been available at prediction time, thereby preventing target leakage from future information.
    We filtered the samples and features as needed to avoid any temporal leakage
    \\
    Datasets were additionally analyzed for grouped temporal structure, such as repeated observations from the same entities over time (e.g., multiple transactions from a single customer). In such cases, using a group split was carefully considered during split construction.
    \\
    We ensure that the first split of a temporal task is always the one with the most recent time test point. Thus, it is also the split with the most training data, and we deem it the most representative split.
    We order all other splits by descending time test points. 
\end{description}

\subsection{Details on Curation Outcome}
\label{appendix:dataset_curation_outcome}

\begin{figure}
    \centering
    \includegraphics[width=0.6\textwidth]{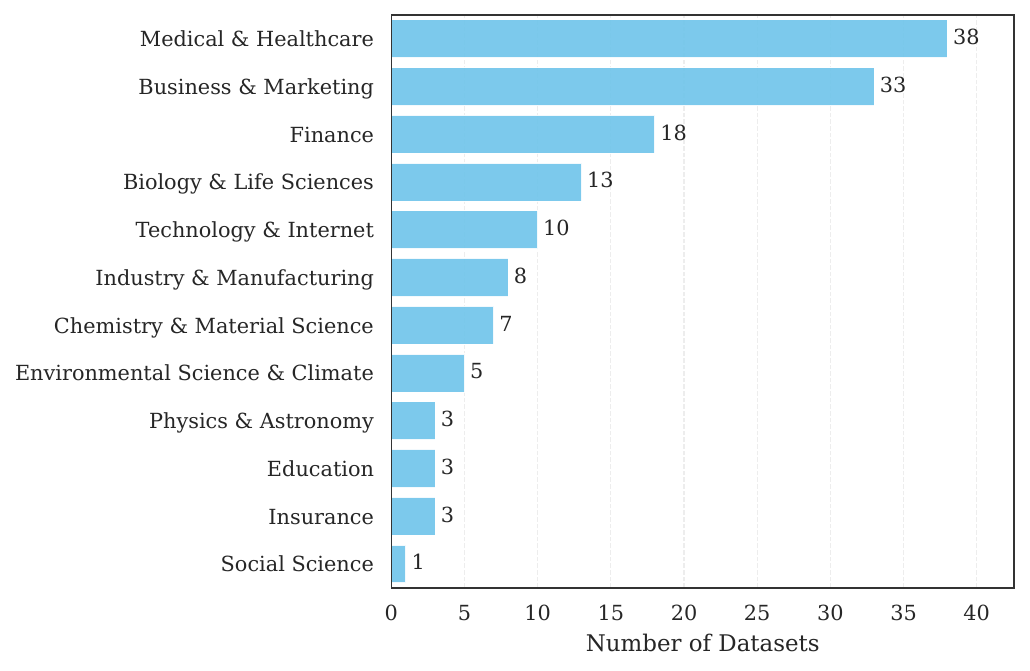}
    \caption{
    \textbf{Dataset Domain Breakdown.} We use the domain categories from TALENT~\citep{ye2024closer} and indicate how many out of our $142$ datasets stem from each domain. 
    }
    \label{fig:domain_Breakdown}
\end{figure}

\begin{figure}
    \centering
    \includegraphics[width=0.7\textwidth]{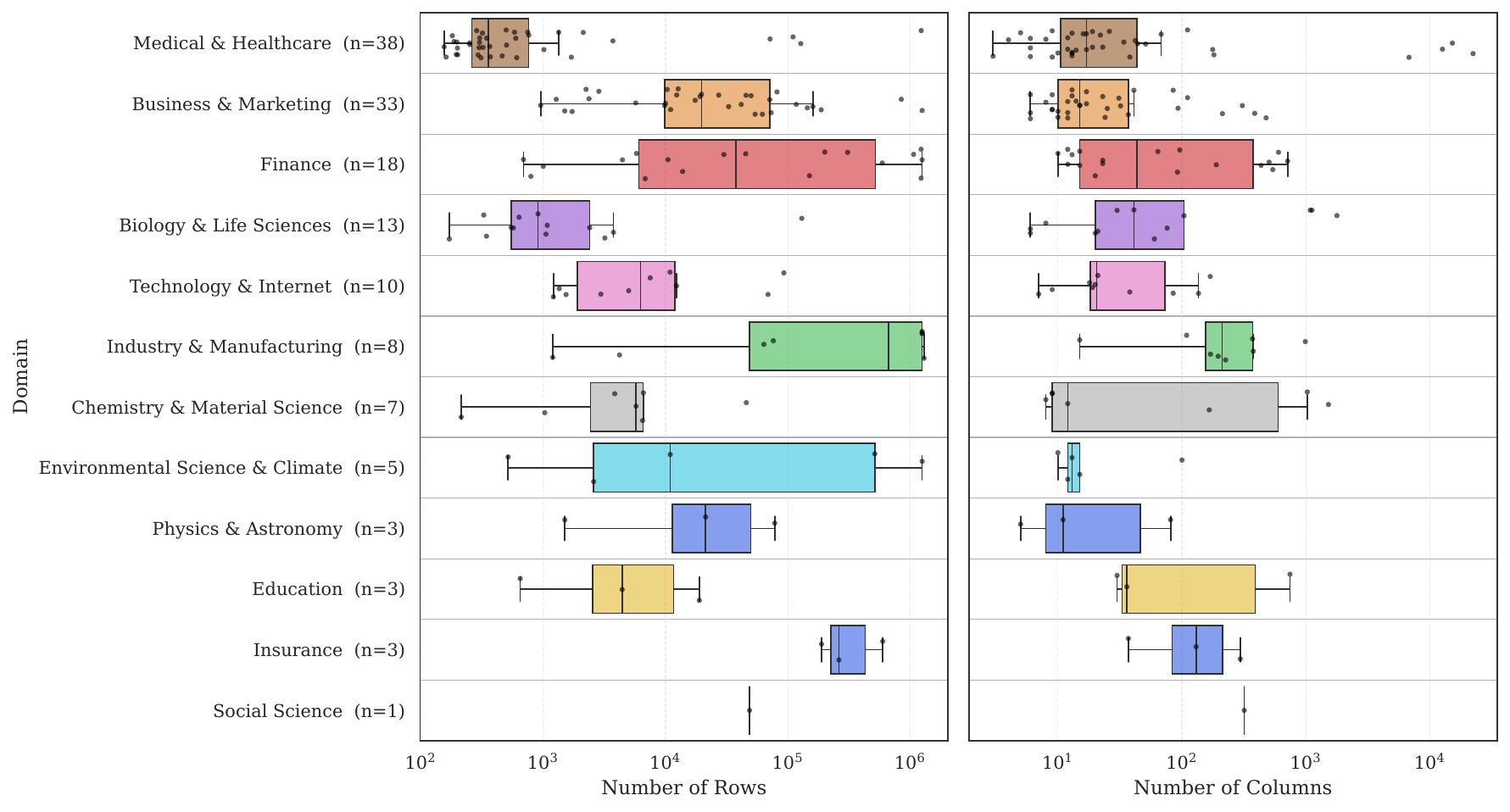}
    \caption{
    \textbf{Dataset Scale Per Domain.} We show the number of rows and columns for each dataset binned by their domain.
    }
    \label{fig:domain_scales}
\end{figure}

\begin{figure}
    \centering
    \includegraphics[width=0.7\textwidth]{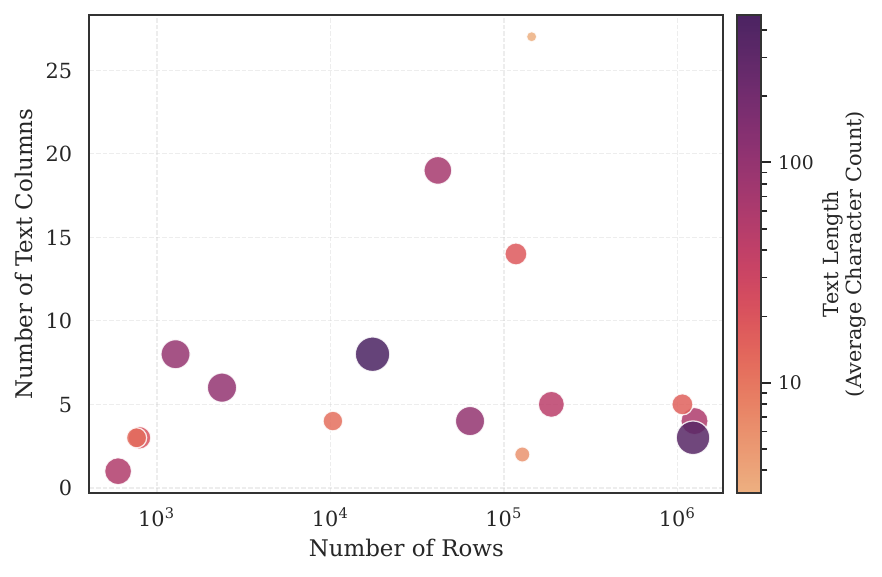}
    \caption{
    \textbf{Characteristics Tabular Data with Text Features.} We show the number of rows and the number of text columns. The hue highlights the average text length. We have an even mix of short and long text datasets, spanning several orders of magnitude in sample size and a diverse number of text columns. 
    }
    \label{fig:text_char}
\end{figure}

\begin{figure}
    \centering
    \includegraphics[width=0.7\textwidth]{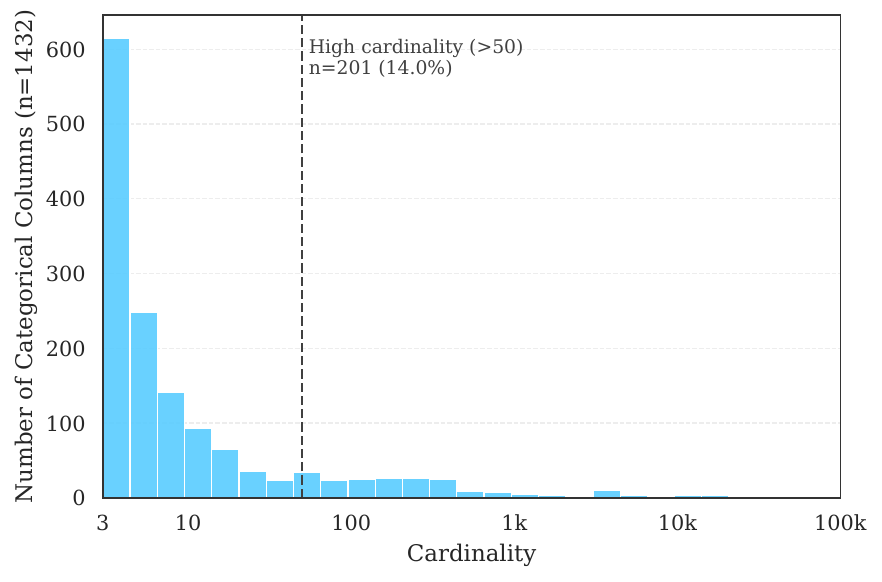}
    \caption{
    \textbf{Distribution Cardinality of Categorical.} The distribution of the cardinality of all categorical columns across all datasets in our dataset collection. 
    }
    \label{fig:categorical_cardinality}
\end{figure}

\begin{figure}
    \centering
    \begin{subfigure}{0.48\textwidth}
        \includegraphics[width=\textwidth]{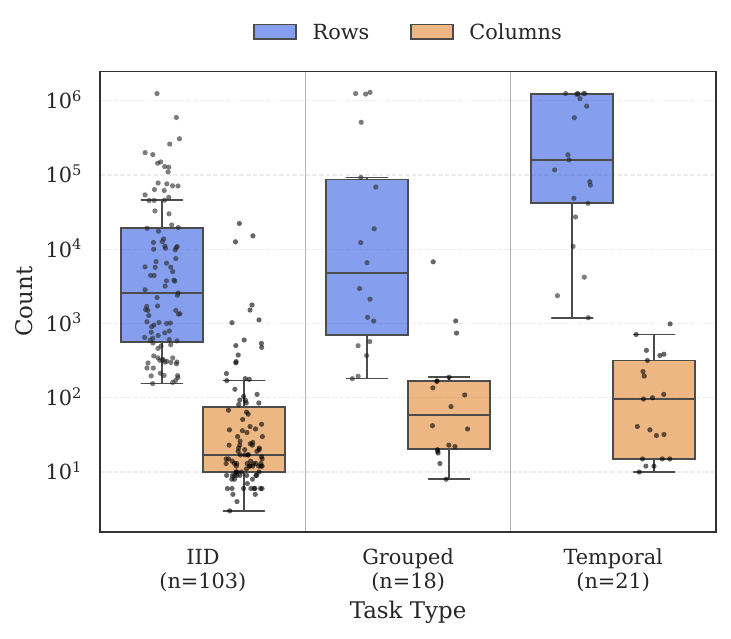}
    \end{subfigure}%
    \hfill
    \begin{subfigure}{0.48\textwidth}
        \includegraphics[width=\textwidth]{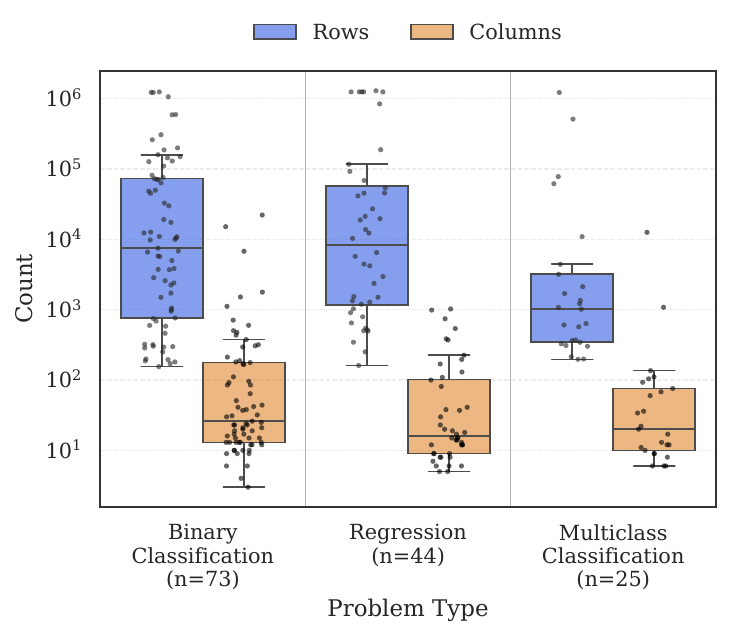}
    \end{subfigure}
    \caption{
    \textbf{(Left) Data Scales Per Task.} 
    The row and column count for each dataset, binned by task type. We see a trend towards larger temporal datasets. 
    \textbf{(Right) Data Scales Per Problem.} 
    Datasets binned by problem type show that multiclass classification problems are smaller on average, while binary classification and regression are similarly distributed in size. 
    }
    \label{fig:data_scales_more}
\end{figure}

\begin{figure}
    \centering
    \includegraphics[width=0.6\textwidth]{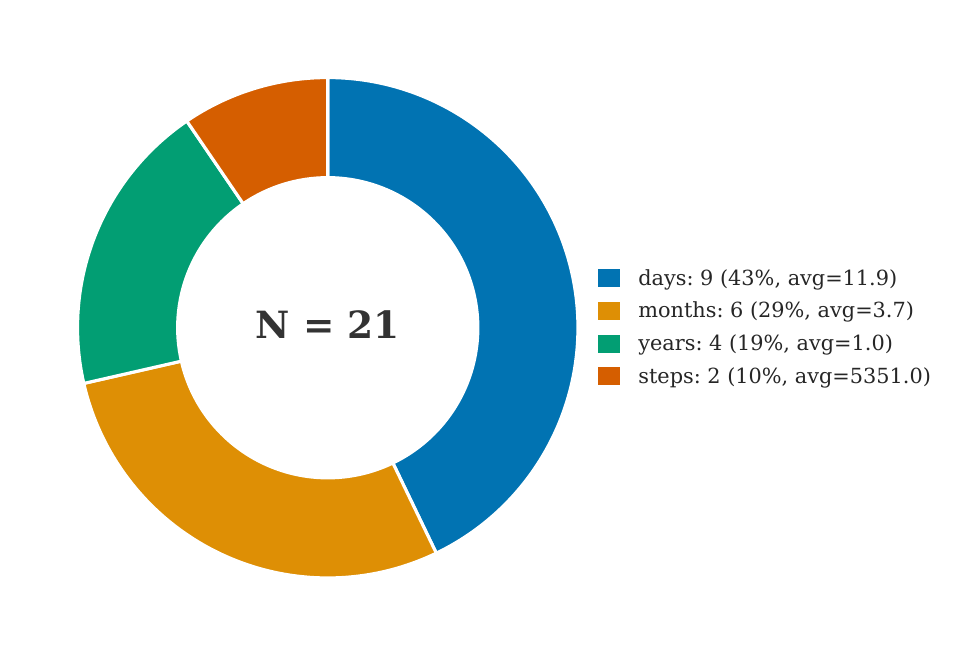}
    \caption{
    \textbf{Distribution of Time Horizons for Temporal Dataset.} We show the count of time horizon units (days, months, years, steps) and the average number of values per unit (e.g. $11.9$ days or $3.7$ months).
    The time horizon indicates the time until we refit or get new labeled data.    
    }
    \label{fig:time_horizons}
\end{figure}

We investigate additional characteristics of our curated dataset.
\\
In \Cref{fig:domain_Breakdown}, we show how many datasets we have per domain, and in \Cref{fig:domain_scales}, we show how the dataset scales (in terms of number of rows and columns) varies across domains. 
Our datasets cover a broad range of domains with a large focus on some of the most common applied disciplines such as healthcare, marketing and finance. 
Interestingly, we noticed that medical and healthcare datasets are generally smaller, whereas finance datasets are several orders of magnitude larger on average. 
\\
In \Cref{fig:text_char}, we show the distribution of tabular datasets with text in our collection. 
Our text datasets span a wide range of scales, with multiple text columns and varying average text lengths. 
\\ 
In \Cref{fig:categorical_cardinality}, we show the distribution of the cardinality across all categorical features. We see a clear long-tail for high cardinality, making up $14\%$ of all categorical columns. 
\\
In \Cref{fig:data_scales_more}, we investigate how the scale and dimensionality of our datasets change based on the task type and problem type. We observe that temporal data tends to have larger average sizes, while multiclass classification often stems from smaller problems. 
\\
In \Cref{fig:time_horizons}, we show the distribution of time horizons for temporal tasks. 
In most cases, the time horizon is days or months. 
Two temporal datasets have only a time index, decoupled from dates. Both used higher step sizes for the prediction horizon.  

\myparagraph{TabArena Reinvestigation.}
We re-investigated all accepted datasets from TabArena during our data curation process and filtered $2$ out of the $51$ datasets.
We removed the \texttt{anneal} dataset because it is trivial (all models achieved perfect accuracy).
Moreover, we removed the \texttt{diabetes} dataset due to ethical concerns, already raised by \citet{radin2017digital} in 2017. 
Finally, while ingesting the datasets into \df, we fixed several minor issues, including dropping constant columns, ensuring correct feature types, and using log scaling for price prediction tasks when appropriate.   


\subsection{Artifact Metadata and Distribution Policy}
\label{appendix:artifact_metadata}

Several datasets in \benchname are from sources that do not support redistribution  (e.g., Kaggle). 
For each dataset, we share the exact code to reproduce our data state and exhaustive metadata in \df (\href{https://github.com/TabArena/data-foundry}{\texttt{https://github.com/TabArena/data-foundry}}), allowing others to reproduce our steps for obtaining all datasets.
Moreover, we release all processed data artifacts used for benchmarking in the form of a single combined backup on Hugging Face (\href{https://huggingface.co/datasets/TabArena/BeyondArena}{\texttt{https://huggingface.co/datasets/TabArena/BeyondArena}}), enabling continuous reproducibility of the exact benchmark setup.  
For Kaggle-sourced datasets specifically, which require per-user license acceptance, this combined backup serves as a fallback should a competition dataset later be removed from Kaggle (as previously happened with, e.g., the JaneStreet competition).
We do not provide primary hosting or redistribution of the original datasets.

\subsection{BeyondArena Datasets Overview}
\label{appendix:datasets}

\Cref{tab:all-datasets} provides a per-dataset overview of all datasets used in \benchname. 

\begin{table}[ht]
\centering
\caption{Per-dataset metadata for \benchname benchmark, sorted by number of rows ($N$). Columns: $N$ rows, $d$ columns before preprocessing, $C$ classes (regression: ---), problem type (\textbf{Binary} classification / \textbf{Multi}class / \textbf{Reg}ression ), task type (IID / Temporal / Grouped). Domain abbreviations: M \& H = Medical \& Healthcare; B \& M = Business \& Marketing; B \& L = Biology \& Life Sciences; T \& I = Technology \& Internet; I \& M = Industry \& Manufacturing; C \& M = Chemistry \& Material Science; E \& C = Environmental Science \& Climate; P \& A = Physics \& Astronomy.}
\label{tab:all-datasets}
\resizebox{\textwidth}{!}{%
\begin{tabular}{l l l r r r r r l l}
\toprule
Dataset & Domain & Source & Year & Age & N & d & C & Prob. & Task \\
\midrule
hepatitis\_survival\_prediction \citep{efron1981statistical} & M \& H & UCI & 1981 & 45 & 155 & 19 & 2 & Binary & IID \\
cirrhosis\_patient\_survival\_prediction \citep{dickson1989prognosis} & M \& H & UCI & 1984 & 42 & 161 & 17 & -- & Reg & IID \\
clock\_protein\_toxicity \citep{gul2021structure} & B \& L & UCI & 2021 & 5 & 171 & 1,117 & 2 & Binary & IID \\
pancreatic\_cancer\_mouse\_detection \citep{hingorani2003preinvasive} & M \& H & Other & 2003 & 23 & 181 & 6,771 & 2 & Binary & Grouped \\
lung\_cancer\_epithelial\_genexp \citep{spira2007airway} & M \& H & GOV Website & 2006 & 20 & 187 & 22,215 & 2 & Binary & IID \\
parkinsons\_biomedical\_voice\_measurements \citep{little2007exploiting} & M \& H & UCI & 2007 & 19 & 195 & 23 & 2 & Binary & Grouped \\
lung\_cancer \citep{bhattacharjee2001classification} & M \& H & Other & 2001 & 25 & 197 & 12,600 & 4 & Multi & IID \\
audiology\_diagnosis \citep{bareiss1990protos} & M \& H & UCI & 1987 & 39 & 199 & 68 & 3 & Multi & IID \\
heart\_disease\_va\_long\_beach \citep{detrano1989international} & M \& H & UCI & 1989 & 37 & 200 & 13 & 2 & Binary & IID \\
forensic\_glass\_identification \citep{German1987glass} & C \& M & UCI & 1987 & 39 & 214 & 9 & 6 & Multi & IID \\
early\_stage\_diabetes\_risk\_prediction \citep{islam2019likelihood} & M \& H & UCI & 2019 & 7 & 251 & 16 & 2 & Binary & IID \\
body\_density\_prediction \citep{penrose1985generalized} & M \& H & Kaggle & 1985 & 41 & 252 & 13 & -- & Reg & IID \\
ljubljana\_breast\_cancer \citep{Zwitter1988BreastCancer} & M \& H & UCI & 1988 & 38 & 286 & 9 & 2 & Binary & IID \\
heart\_disease\_hungary \citep{detrano1989international} & M \& H & UCI & 1989 & 37 & 294 & 13 & 2 & Binary & IID \\
heart\_failure\_followup\_survival \citep{chicco2020machine} & M \& H & UCI & 2020 & 6 & 299 & 12 & 2 & Binary & IID \\
ljubljana\_primary\_tumor \citep{Zwitter1987primarytumor} & M \& H & UCI & 1987 & 39 & 302 & 17 & 11 & Multi & IID \\
heart\_disease\_cleveland \citep{detrano1989international} & M \& H & UCI & 1989 & 37 & 303 & 13 & 2 & Binary & IID \\
biomechanical\_orthopaedic\_prediction \citep{Barreto2005Vertebral} & M \& H & UCI & 2011 & 15 & 310 & 6 & 3 & Multi & IID \\
gallstone\_disease \citep{esen2024early} & M \& H & UCI & 2023 & 3 & 319 & 38 & 2 & Binary & IID \\
prostate\_cancer\_detection \citep{petricoin2002serum} & M \& H & Other & 2002 & 24 & 322 & 15,154 & 2 & Binary & IID \\
ecoli\_proteins \citep{horton1996probabilistic} & B \& L & UCI & 1996 & 30 & 327 & 6 & 5 & Multi & IID \\
horse\_colic\_survival \citep{McLeish1989HorseColic} & B \& L & UCI & 1989 & 37 & 344 & 20 & 3 & Multi & IID \\
blood\_tests\_drink\_prediction \citep{UCILiverDisorders2016} & M \& H & UCI & 1996 & 30 & 345 & 5 & -- & Reg & IID \\
eryhemato\_squamous\_disease \citep{guvenir1998learning} & M \& H & UCI & 1997 & 29 & 366 & 34 & 6 & Multi & IID \\
dementia\_prediction \citep{marcus2010open} & M \& H & Other & 2010 & 16 & 370 & 8 & 3 & Multi & Grouped \\
south\_africa\_coronary\_heart\_disease \citep{rossouw1983coronary} & M \& H & Kaggle & 1983 & 43 & 462 & 9 & 2 & Binary & IID \\
obesity\_estimation \citep{palechor2019dataset} & M \& H & UCI & 2019 & 7 & 498 & 14 & -- & Reg & IID \\
telemonitoring\_parkinsons\_biomedical\_voice\_measurements \citep{tsanas2009accurate} & M \& H & UCI & 2007 & 19 & 502 & 19 & -- & Reg & Grouped \\
forest\_fires \citep{cortez2007data} & E \& C & UCI & 2008 & 18 & 517 & 12 & -- & Reg & IID \\
qsar\_aquatic\_toxicity \citep{cassotti2014prediction} & B \& L & UCI & 2014 & 12 & 546 & 8 & -- & Reg & IID \\
micro\_mass \citep{mahe2014automatic} & B \& L & UCI & 2013 & 13 & 571 & 1,082 & 20 & Multi & Grouped \\
indian\_liver\_patient\_dataset \citep{ramana2012critical} & M \& H & UCI & 2012 & 14 & 583 & 10 & 2 & Binary & IID \\
drug\_induced\_autoimmunity\_prediction \citep{huang2025interdia} & M \& H & UCI & 2025 & 1 & 597 & 177 & 2 & Binary & IID \\
hepatitis\_c\_prediction \citep{hoffmann2018using} & M \& H & UCI & 2018 & 8 & 608 & 12 & 4 & Multi & IID \\
biogeographical\_ancestry\_prediction \citep{heinzel2025advancing,ruiz2023development,xavier2020development} & B \& L & GitHub & 2025 & 1 & 635 & 104 & 10 & Multi & IID \\
student\_portuguese\_performance \citep{silva2008using} & Education & UCI & 2008 & 18 & 649 & 30 & -- & Reg & IID \\
credit\_approval \citep{quinlan1987simplifying} & Finance & UCI & 1987 & 39 & 690 & 15 & 2 & Binary & IID \\
blood\_transfusion \citep{yeh2009knowledge} & M \& H & UCI & 2008 & 18 & 748 & 4 & 2 & Binary & IID \\
regensburg\_pediatric\_appendicitis \citep{marcinkevivcs2024interpretable} & M \& H & Other & 2021 & 5 & 763 & 51 & 2 & Binary & IID \\
mutual\_funds\_india \citep{Barnawal2022MutualFundsIndiaDetailed} & Finance & Kaggle & 2023 & 3 & 793 & 12 & -- & Reg & IID \\
qsar\_fish\_toxicity \citep{cassotti2015similarity} & B \& L & UCI & 2015 & 11 & 908 & 6 & -- & Reg & IID \\
tour\_travels\_churn \citep{Tejashvi2023TourTravelsCustomerChurnPrediction} & B \& M & Kaggle & 2021 & 5 & 954 & 6 & 2 & Binary & IID \\
credit\_g \citep{hofmann1994statlog} & Finance & UCI & 1994 & 32 & 1,000 & 20 & 2 & Binary & IID \\
maternal\_health\_risk \citep{ahmed2020review} & M \& H & UCI & 2020 & 6 & 1,014 & 6 & 3 & Multi & IID \\
concrete\_compressive\_strength \citep{yeh1998modeling} & C \& M & UCI & 1998 & 28 & 1,030 & 8 & -- & Reg & IID \\
qsar\_biodeg \citep{mansouri2013quantitative} & B \& L & UCI & 2013 & 13 & 1,054 & 41 & 2 & Binary & IID \\
mice\_protein\_trisomy\_discriminant \citep{higuera2015self} & B \& L & UCI & 2015 & 11 & 1,080 & 76 & 8 & Multi & Grouped \\
garments\_worker\_productivity \citep{imran2021mining} & I \& M & UCI & 2020 & 6 & 1,197 & 15 & -- & Reg & Temporal \\
asp\_potassco\_classification \citep{hoos2014claspfolio,bischl_aslib_2016} & T \& I & ASlib & 2014 & 12 & 1,212 & 136 & 11 & Multi & Grouped \\
wine\_world\_cost \citep{Rustamov2023WineDataset} & B \& M & Kaggle & 2023 & 3 & 1,279 & 14 & -- & Reg & IID \\
healthcare\_insurance\_expenses \citep{arunjangir2452023insurance} & M \& H & Kaggle & 2023 & 3 & 1,338 & 6 & -- & Reg & IID \\
website\_phishing \citep{abdelhamid2014phishing} & T \& I & UCI & 2014 & 12 & 1,353 & 9 & 3 & Multi & IID \\
fitness\_club \citep{ddosad2023fitness} & B \& M & Kaggle & 2023 & 3 & 1,500 & 6 & 2 & Binary & IID \\
airfoil\_self\_noise \citep{brooks1989airfoil} & P \& A & UCI & 2014 & 12 & 1,503 & 5 & -- & Reg & IID \\
fiat\_500 \citep{paolocons2020fiat} & T \& I & Kaggle & 2020 & 6 & 1,538 & 7 & -- & Reg & IID \\
mic \citep{golovenkin2020trajectories} & M \& H & UCI & 2020 & 6 & 1,699 & 111 & 8 & Multi & IID \\
bad\_customer\_detection \citep{Podsyp2020IsThisAGoodCustomer} & B \& M & Kaggle & 2020 & 6 & 1,723 & 13 & 2 & Binary & IID \\
cardiotocography \citep{campos2010cardiotocography} & M \& H & UCI & 2010 & 16 & 2,126 & 22 & 3 & Multi & Grouped \\
marketing\_campaign \citep{saldanha2020marketing} & B \& M & Kaggle & 2020 & 6 & 2,240 & 25 & 2 & Binary & IID \\
coffee\_rating\_prediction \citep{AlIrsyad2023CoffeeDataCoffeeReview} & B \& M & Kaggle & 2023 & 3 & 2,369 & 12 & -- & Reg & Temporal \\
hazelnut\_spread\_contaminant\_detection \citep{ricci2021machine} & B \& L & OpenML & 2020 & 6 & 2,400 & 30 & 2 & Binary & IID \\
seismic\_bumps \citep{sikora2010application} & E \& C & UCI & 2013 & 13 & 2,584 & 15 & 2 & Binary & IID \\
iranian\_churn \citep{keramati2011churn} & B \& M & UCI & 2011 & 15 & 2,850 & 13 & 2 & Binary & IID \\
sat11\_hand\_algo\_runtime \citep{xu-sat12a,sat12,bischl_aslib_2016} & T \& I & ASlib & 2011 & 15 & 2,960 & 169 & -- & Reg & Grouped \\
splice \citep{towell1994knowledge} & B \& L & UCI & 1991 & 35 & 3,190 & 60 & 3 & Multi & IID \\
thyroid\_discordant \citep{quinlan1987simplifying} & M \& H & UCI & 1986 & 40 & 3,711 & 26 & 2 & Binary & IID \\
bioresponse \citep{bioresponse2012hamner} & B \& L & Kaggle & 2012 & 14 & 3,751 & 1,776 & 2 & Binary & IID \\
hiva\_agnostic \citep{guyon2007agnostic} & C \& M & Other & 2007 & 19 & 3,845 & 1,518 & 2 & Binary & IID \\
mercedes\_benz\_greener\_manufacturing \citep{Novy2017MercedesBenzGreenerManufacturing} & I \& M & Kaggle & 2017 & 9 & 4,204 & 371 & -- & Reg & Temporal \\
predict\_students\_dropout\_and\_academic\_success \citep{martins2021early} & Education & UCI & 2021 & 5 & 4,424 & 36 & 3 & Multi & IID \\
santander\_transaction\_value \citep{McDonald2018SantanderValuePredictionChallenge} & Finance & Kaggle & 2018 & 8 & 4,447 & 540 & -- & Reg & IID \\
\bottomrule
\end{tabular}
}
\end{table}

\begin{table}[ht]
\ContinuedFloat
\centering
\caption[]{(continued) Per-dataset metadata for \benchname benchmark.}
\resizebox{\textwidth}{!}{%
\begin{tabular}{l l l r r r r r l l}
\toprule
Dataset & Domain & Source & Year & Age & N & d & C & Prob. & Task \\
\midrule
churn \citep{marcoulides2005churn} & T \& I & OpenML & 2005 & 21 & 5,000 & 19 & 2 & Binary & IID \\
homeq\_default\_prediction \citep{baesens2016credit} & B \& M & Other & 2016 & 10 & 5,708 & 12 & 2 & Binary & IID \\
qsar\_tid\_11 \citep{olier2018meta} & C \& M & OpenML & 2015 & 11 & 5,741 & 1,024 & -- & Reg & IID \\
polish\_companies\_bankruptcy \citep{zikeba2016ensemble} & Finance & UCI & 2010 & 16 & 5,790 & 64 & 2 & Binary & IID \\
wine\_quality \citep{cortez2009modeling} & C \& M & UCI & 2009 & 17 & 6,497 & 12 & -- & Reg & IID \\
musk \citep{dietterich1993comparison} & C \& M & UCI & 1994 & 32 & 6,598 & 166 & 2 & Binary & Grouped \\
taiwanese\_bankruptcy\_prediction \citep{liang2016financial} & Finance & UCI & 2009 & 17 & 6,819 & 92 & 2 & Binary & IID \\
naticusdroid\_android\_permissions\_dataset \citep{mathur2021naticusdroid} & T \& I & UCI & 2021 & 5 & 7,491 & 85 & 2 & Binary & IID \\
coil\_2000 \citep{van2000coil} & B \& M & UCI & 2000 & 26 & 9,822 & 85 & 2 & Binary & IID \\
bank\_customer\_churn \citep{Topre2022BankCustomerChurn} & B \& M & Kaggle & 2020 & 6 & 10,000 & 10 & 2 & Binary & IID \\
immoscout\_german\_house\_prices \citep{Shritech2019GermanHousingPricePrediction,OpenML43342Dataset} & B \& M & Kaggle & 2019 & 7 & 10,317 & 23 & -- & Reg & IID \\
heloc \citep{averkiyoliabev2021heloc} & Finance & Kaggle & 2021 & 5 & 10,459 & 23 & 2 & Binary & IID \\
jm1 \citep{menzies2004good} & T \& I & OpenML & 2004 & 22 & 10,885 & 21 & 2 & Binary & IID \\
ghanas\_indigenous\_intel \citep{zindi_ghana_indigenous_intel_2025} & E \& C & Zindi & 2025 & 1 & 10,928 & 10 & 4 & Multi & Temporal \\
ecommerce\_shipping \citep{gopalani2021ecommerce} & B \& M & Kaggle & 2021 & 5 & 10,999 & 10 & 2 & Binary & IID \\
video\_game\_fps\_prediction \citep{peeters2021performance} & T \& I & OpenML & 2020 & 6 & 12,288 & 38 & -- & Reg & Grouped \\
online\_shoppers\_purchasing\_intention\_dataset \citep{sakar2019real} & B \& M & UCI & 2017 & 9 & 12,330 & 17 & 2 & Binary & IID \\
in\_vehicle\_coupon\_recommendation \citep{wang2017bayesian} & B \& M & UCI & 2017 & 9 & 12,684 & 24 & 2 & Binary & IID \\
miami\_housing \citep{bourassa2021big} & Finance & Kaggle & 2016 & 10 & 13,776 & 15 & -- & Reg & IID \\
emscad \citep{vidros2017automatic} & B \& M & Other & 2014 & 12 & 17,460 & 17 & 2 & Binary & IID \\
early\_learning\_predictors \citep{DataDrive2030_2024_elom_thrivebyfive} & Education & Other & 2023 & 3 & 18,874 & 743 & -- & Reg & Grouped \\
hr\_analytics \citep{arashnic2021hr} & B \& M & Kaggle & 2021 & 5 & 19,158 & 12 & 2 & Binary & IID \\
houses \citep{pace1997sparse} & B \& M & Other & 1990 & 36 & 19,675 & 8 & -- & Reg & IID \\
superconductivity \citep{hamidieh2018data} & P \& A & UCI & 2018 & 8 & 21,263 & 81 & -- & Reg & IID \\
sberbank\_housing\_market\_forecasting \citep{Herman2024HomeCreditCreditRiskModelStability} & B \& M & Kaggle & 2017 & 9 & 27,195 & 386 & -- & Reg & Temporal \\
credit\_card\_clients\_default \citep{yeh2009comparisons} & Finance & UCI & 2009 & 17 & 30,000 & 23 & 2 & Binary & IID \\
amazon\_employee\_access \citep{hamner2013amazon} & B \& M & Kaggle & 2010 & 16 & 32,769 & 9 & 2 & Binary & IID \\
california\_house\_prices\_2020 \citep{d2lcourse2021california_house_prices} & B \& M & Kaggle & 2021 & 5 & 41,528 & 41 & -- & Reg & Temporal \\
bank\_marketing \citep{moro2014bank-marketing} & Finance & UCI & 2012 & 14 & 45,211 & 13 & 2 & Binary & IID \\
food\_delivery\_time \citep{rajatkumar302023food} & B \& M & Kaggle & 2023 & 3 & 45,451 & 9 & -- & Reg & IID \\
physiochemical\_protein \citep{rana2013protein} & C \& M & UCI & 2013 & 13 & 45,730 & 9 & -- & Reg & IID \\
anes\_voting\_2026 \citep{anes2026timeseries} & Social Science & Other & 2026 & 0 & 48,587 & 318 & 2 & Binary & Temporal \\
kdd\_cup\_09\_appetency \citep{guyon2009analysis} & B \& M & Other & 2008 & 18 & 50,000 & 212 & 2 & Binary & IID \\
diamonds \citep{wickham2016data} & B \& M & Other & 2015 & 11 & 53,940 & 9 & -- & Reg & IID \\
otto\_group\_product\_classification\_challenge \citep{Bossan2015OttoGroupProductClassificationChallenge} & B \& M & Kaggle & 2015 & 11 & 61,878 & 93 & 9 & Multi & IID \\
labour\_inspection\_compliance \citep{flogard2022dataset} & I \& M & Other & 2019 & 7 & 63,634 & 376 & 2 & Binary & IID \\
video\_transcoding\_time\_prediction \citep{deneke2014video} & T \& I & UCI & 2015 & 11 & 68,784 & 18 & -- & Reg & Grouped \\
santander\_customer\_satisfaction \citep{Jimenez2016SantanderCustomerSatisfaction} & B \& M & Kaggle & 2016 & 10 & 71,080 & 307 & 2 & Binary & IID \\
diabetes\_130\_us \citep{strack2014impact} & M \& H & UCI & 2014 & 12 & 71,518 & 44 & 2 & Binary & IID \\
kick \citep{DontGetKicked} & B \& M & Kaggle & 2011 & 15 & 72,983 & 32 & 2 & Binary & Temporal \\
aps\_failure \citep{ida2016challenge} & I \& M & UCI & 2016 & 10 & 76,000 & 170 & 2 & Binary & IID \\
sdss\_17 \citep{accetta2022seventeenth} & P \& A & Kaggle & 2022 & 4 & 78,053 & 11 & 3 & Multi & IID \\
hotel\_booking\_demand \citep{antonio2019hotel} & B \& M & Other & 2019 & 7 & 81,418 & 31 & 2 & Binary & Temporal \\
5g\_energy\_consumption \citep{huawei_netop_5g_energy_consumption} & T \& I & HuggingFace & 2023 & 3 & 92,629 & 20 & -- & Reg & Grouped \\
sepsis\_survival\_minimal\_clinical\_records \citep{chicco2020survival} & M \& H & UCI & 2020 & 6 & 110,204 & 3 & 2 & Binary & IID \\
sf\_permit\_time \citep{SanFrancisco2026BuildingPermits} & B \& M & GOV Website & 2025 & 1 & 116,954 & 37 & -- & Reg & Temporal \\
wids\_diabetes\_mellitus \citep{Matthys2021WiDSDatathon2021} & M \& H & Kaggle & 2021 & 5 & 127,358 & 181 & 2 & Binary & IID \\
customer\_satisfaction\_in\_airline \citep{yakhyojon2023airlinesatisfaction} & B \& L & Kaggle & 2023 & 3 & 129,880 & 21 & 2 & Binary & IID \\
pva\_revenue\_prediction\_kddcup98 \citep{Parsa1998KDDCup1998} & B \& M & Other & 1997 & 29 & 144,095 & 477 & 2 & Binary & IID \\
give\_me\_some\_credit \citep{cukierski2011credit} & Finance & Kaggle & 2011 & 15 & 150,000 & 10 & 2 & Binary & IID \\
acquire\_valued\_shoppers\_challenge \citep{DMDave2014AcquireValuedShoppersChallenge} & B \& M & Kaggle & 2014 & 12 & 160,057 & 111 & 2 & Binary & Temporal \\
kickstarter \citep{webrobots2026kickstarter} & B \& M & Other & 2025 & 1 & 187,118 & 15 & 2 & Binary & Temporal \\
allstate\_claims\_severity \citep{Ferguson2016AllstateClaimsSeverity} & Insurance & Kaggle & 2016 & 10 & 188,317 & 130 & -- & Reg & IID \\
santander\_customer\_transaction\_prediction \citep{Piedra2019SantanderCustomerTransactionPrediction} & Finance & Kaggle & 2019 & 7 & 200,000 & 600 & 2 & Binary & IID \\
homesite\_quote\_conversion \citep{Darrel2015HomesiteQuoteConversion} & Insurance & Kaggle & 2015 & 11 & 260,753 & 295 & 2 & Binary & IID \\
home\_credit\_default\_risk \citep{Montoya2018HomeCreditDefaultRisk} & Finance & Kaggle & 2018 & 8 & 307,507 & 504 & 2 & Binary & IID \\
covertype \citep{blackard1999comparative} & E \& C & UCI & 1998 & 28 & 512,625 & 13 & 3 & Multi & Grouped \\
ieee\_fraud\_detection \citep{ieee-fraud-detection} & Finance & Kaggle & 2019 & 7 & 590,540 & 435 & 2 & Binary & Temporal \\
porto\_seguro \citep{Howard2017PortoSegurosSafeDriverPrediction} & Insurance & Kaggle & 2017 & 9 & 595,206 & 37 & 2 & Binary & IID \\
rossmann\_store\_sales \citep{kaggle_rossmann_store_sales} & B \& M & Kaggle & 2015 & 11 & 844,392 & 15 & -- & Reg & Temporal \\
lending\_club\_1m \citep{sanz2025credit} & Finance & Kaggle & 2018 & 8 & 1,064,751 & 96 & 2 & Binary & Temporal \\
home\_credit\_default\_stability\_1m \citep{Herman2024HomeCreditCreditRiskModelStability} & Finance & Kaggle & 2024 & 2 & 1,224,927 & 711 & 2 & Binary & Temporal \\
consumer\_complaints\_1m \citep{cfpb2025ConsumerComplaintDatabase} & Finance & GOV Website & 2025 & 1 & 1,226,140 & 12 & 3 & Multi & Temporal \\
sepsis\_prediction\_1m \citep{reyna2020early} & M \& H & Other & 2019 & 7 & 1,228,686 & 42 & 2 & Binary & Grouped \\
amex\_non\_iid\_1m \citep{howard2022amex} & Finance & Kaggle & 2022 & 4 & 1,249,605 & 189 & 2 & Binary & Grouped \\
delivery\_eta\_1m \citep{rubachev2025tabred} & I \& M & Kaggle & 2024 & 2 & 1,250,000 & 225 & -- & Reg & Temporal \\
cooking\_time\_1m \citep{rubachev2025tabred} & I \& M & Kaggle & 2024 & 2 & 1,250,000 & 196 & -- & Reg & Temporal \\
climate\_model\_weather\_forecasting\_1m \citep{rubachev2025tabred} & E \& C & Kaggle & 2024 & 2 & 1,250,000 & 100 & -- & Reg & Temporal \\
maps\_router\_eta\_1m \citep{rubachev2025tabred} & I \& M & Kaggle & 2024 & 2 & 1,250,000 & 988 & -- & Reg & Temporal \\
mercari\_price\_suggestion\_1m \citep{Howard2017MercariPriceSuggestionChallenge} & B \& M & Kaggle & 2018 & 8 & 1,250,000 & 6 & -- & Reg & IID \\
electric\_motor\_temperature\_prediction \citep{kirchgassner2020estimating} & I \& M & Kaggle & 2021 & 5 & 1,296,316 & 109 & -- & Reg & Grouped \\
\bottomrule
\end{tabular}
}
\end{table}

\clearpage
\newpage
\section{Experimental Setup}

\subsection{Time Limit Impact}
\label{appendix:time_limit_investigation}

In our experiments, we limit the evaluation of a single configuration on a single train split of a dataset to $4$ hours. Thus, a model must complete training across all inner folds within $4$ hours; otherwise, training is gracefully stopped early. We use a $4$-hour limit instead of the $1$-hour limit used in TabArena because our benchmark includes larger datasets.

\begin{figure}
    \centering
    \includegraphics[width=\textwidth]{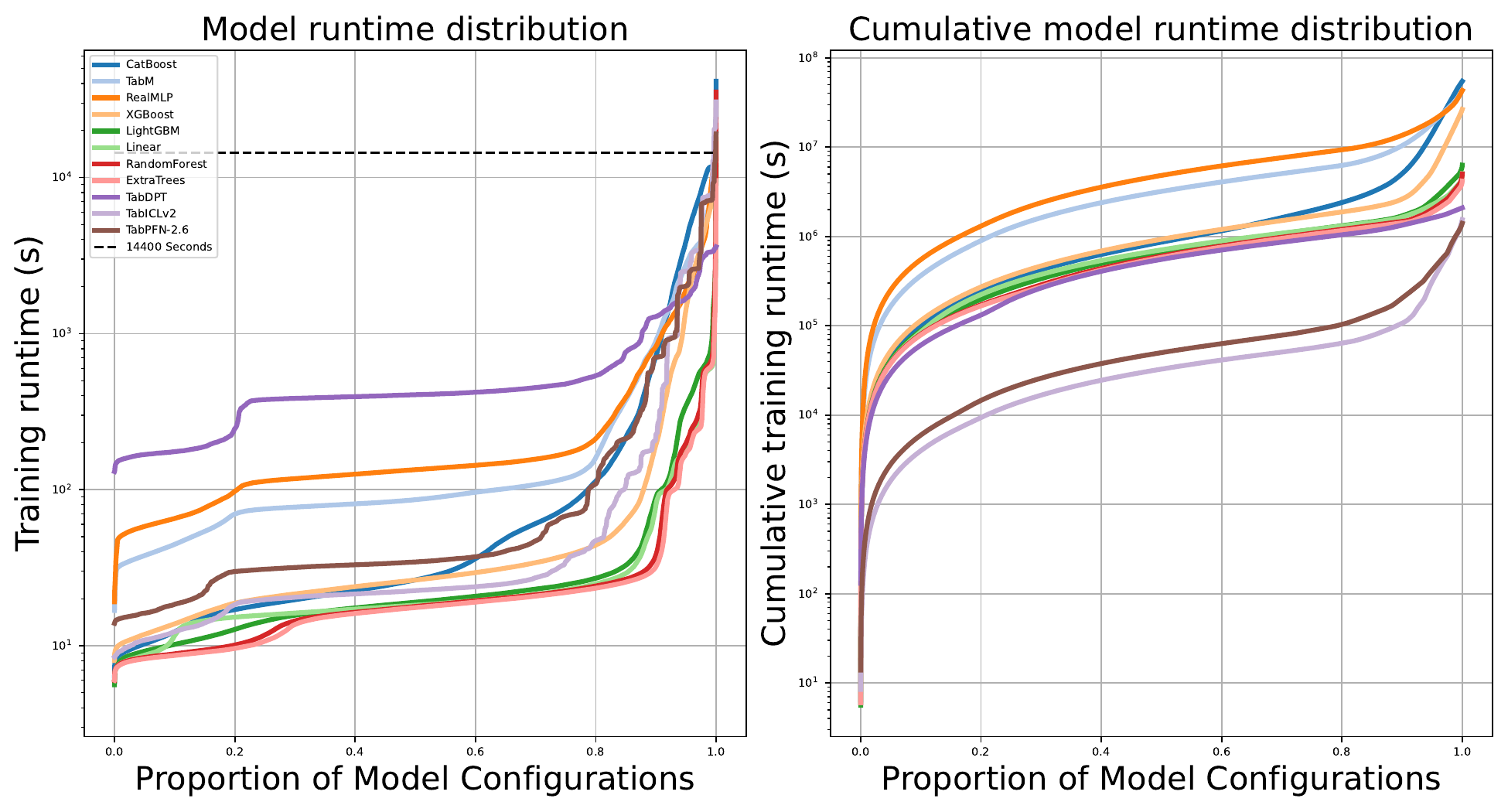}
    \caption{
    \textbf{Training runtime distribution across hyperparameter configurations.}
        Each point shows the training runtime, in seconds, for a hyperparameter configuration evaluated on one train split of a dataset, aggregated across all models. The x-axis shows the cumulative proportion of configurations, and the y-axis shows training time in seconds. The vast majority of configurations finish well below the $4$-hour time limit: only ${\sim}1.72\%$ exceed $1$ hour, ${\sim}0.01\%$ exceed $3.5$ hours, and ${\sim}0.003\%$ ($31$ out of $785{,}208$) exceed the $4$-hour limit.
        }
    \label{fig:time_limit_plot}
\end{figure}

\Cref{fig:time_limit_plot} shows the training runtime distribution across all hyperparameter configurations and models, visualizing the proportion of configurations (x-axis) that required a given training time in seconds (y-axis).

We find that the time limit is rarely reached. Only ${\sim}1.72\%$ of jobs exceeded $1$ hour, ${\sim}0.01\%$ exceeded $3.5$ hours, and only ${\sim}0.003\%$ of jobs, corresponding to $31$ out of $785{,}208$, exceeded the $4$-hour time limit.

\subsection{Additional Details}
\label{appendix:experiment_setup_details}

Below, we provide more details on specific parts of our experimental setup. 

\myparagraph{Validation Splits \& Insufficient Samples per Class.}
\label{appendix:small_class_count_splits}
To avoid errors when evaluating validation metrics due to missing classes, we dynamically reduce the number of folds to match the minimum number of samples per class when a dataset has an insufficient number of minority-class samples for cross-validation (i.e., fewer than 5 or 8).

\myparagraph{Date Encoding}
We use the \texttt{DatetimeEncoder} from Skrub~\citep{skrub} to convert datetime features into $10$ numerical features that represent values such as the year or weekday, as well as a spline-based periodic encoding of the timestamp.  

\myparagraph{Text Encoding}
We use Qwen3-Embedding-8B to transform each text feature into a 4096-dimensional vector, from which we take the first 32 dimensions using Qwen's supported minimal Matryoshka representation learning~\citep{kusupati2022matryoshka} slice. 
We also add caching, so we only need to compute the text encoding once. 

\myparagraph{Grouped Preprocessing}
We add preprocessing for grouped non-IID data, executed at the end of the model-agnostic preprocessing.
For \texttt{label-per-group} datasets, we replace the group index with a transductive 50-dimensional group-encoding.
\\
Per group: we
(a) sort all samples by a group time index, when available; 
(b) aggregate all features across samples (mean/std/min/max/last for numeric, count/last/nunique for non-numeric);
(c) select the top $50$ aggregations by variance when fitting the preprocessing;
and (d) join the top $50$ aggregated features back to all original samples.
At test time, when given a new group as input, we replace the group's index with these $50$ aggregations computed over the group's samples. 
This preprocessing assumes a predictive machine learning task in which all samples from a group are available when predicting. 
This is the case for all \texttt{label-per-group} datasets in \benchname. 

\myparagraph{Compute Hardware.}
We run all our experiments on the Google Cloud Platform (GCP) \citep{gcp_platform}.
We selected virtual machines (VMs) from GCP's offering that might be representative of the hardware of an average practitioner, while being reasonably affordable for running large-scale experiments.
For datasets with fewer than 250k samples, we ran MLPs (TabM, RealMLP) on a2-highgpu-1g~\citep{gcp_a2_highgpu_1g} VMs with an NVIDIA A100 GPU with 40GB VRAM, 12 vCPUs, and 85 GB RAM.
For MLPs on larger datasets, and for any TFM (TabDPT, TabPFN-2.6, TabICLv2), we use a2-ultragpu-1g~\citep{gcp_a2_ultragpu_1g} VMs with an NVIDIA A100 GPU with 80 GB VRAM, 12 vCPUs, and 170 GB RAM.
For all other models, we use n4-standard-16~\citep{gcp_n4_standard_16} VMs with 16 vCPUs and 64 GB RAM.

\clearpage
\newpage

\subsection{Use of Validation Data at Test Time}
\label{appendix:validation_usage}

In \benchname, all models use all available training data at test time, through different mechanisms by model family. 
Following TabArena~\citep{erickson2025tabarena}, we train each traditional model configuration once per inner fold, with early stopping on that fold's validation data where the model supports it. At test time, we do not refit a single model on the full training data. Instead, we average the predictions of all fold models, so the resulting cross-validation ensemble has collectively seen every training sample. Post-hoc ensembling then combines such configurations using their out-of-fold predictions. TFMs require no training or early stopping, so at test time we pass the full training data, including all validation folds, directly as in-context data.

The main alternative to a cross-validation ensemble is refitting: selecting a configuration and retraining one model on all training data. 
\citet{zabergja2024tabular} report gains from refitting, but compare it against a single model trained on a fixed 70\% split of the training data and tuned on the remaining 30\%, so their comparison changes both the amount of training data and the HPO validation scheme, and includes no ensembling over folds; it therefore speaks to a different choice than ours.
A closer comparison is the ablation in \citet{holzmuller2024better} (Appendix~B.3), which contrasts bagging and refitting with one or five models for RealMLP and LightGBM, using tuned default hyperparameters rather than per-dataset HPO. There, five models clearly outperform one under either strategy. Averaging five bagged models outperforms a single refit for RealMLP, but not consistently for LightGBM; refitting five models is mostly better than bagging five. The authors also note that it is unclear how HPO would affect this comparison.
\\
Another practical drawback of refitting is that it leaves no validation data for early stopping, so we must transfer the number of training iterations from the cross-validation runs. These considerations motivated TabArena's choice of cross-validation ensembles, which we adopt. How best to combine early stopping, tuning, and refitting remains, in our view, an open and underexplored question, and \benchname offers a natural testbed for studying it.

\section{Results}

\subsection{Inference Time Overview}
\label{appendix:inference_time}

\begin{figure}
    \centering
    \includegraphics[width=0.8\textwidth]{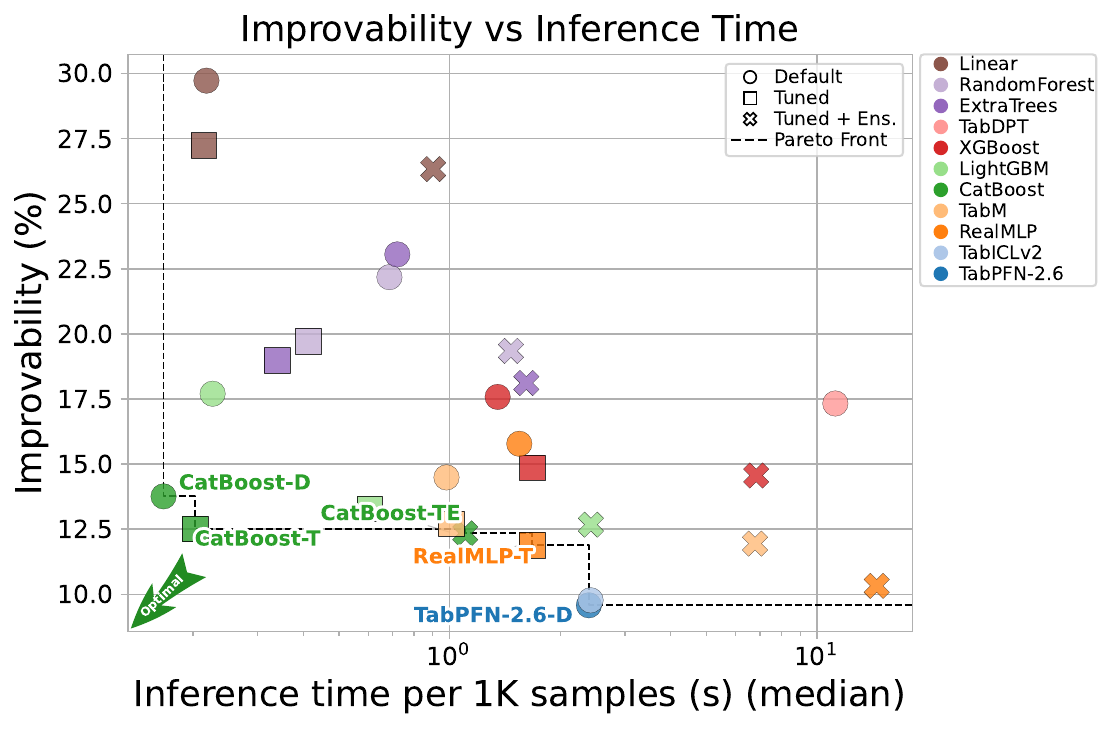}
    \caption{
    \textbf{Improvability vs. Inference Time Pareto Front.} 
    We show the Improvability and inference time of models with the default hyperparameters, with tuning, and with tuning and post-hoc ensembling. The default performance of TFMs equals their in-context learning performance.
    }
    \label{fig:infer_time_plot}
\end{figure}

We show the Pareto front of Improvability vs. inference time in \Cref{fig:infer_time_plot}.
We observe that CatBoost dominates the inference time Pareto front, while RealMLP is competitive by default. 
TabICLv2 and TabPFN-2.6 both require significantly longer inference time while achieving $2.5\%$ Improvement over a tuned CatBoost.
A tuned RealMLP incurs an even larger inference time. 

\subsection{BeyondArena Leaderboard as a Table}
\label{appendix:result_tab}
\Cref{tab:leaderboard} shows the per-model performance across all datasets. 

\begin{table}[t]
    \definecolor{gold}{RGB}{220, 190, 0}
    \definecolor{silver}{RGB}{160, 160, 160}
    \definecolor{bronze}{RGB}{170, 105, 40}
    \centering
    \caption{\textbf{BeyondArena Leaderboard.} We show default (D), tuned (T), and tuned + ensembled (T+E) performances. 
    Results for TabPFN-2.6 and TabDPT are imputed for datasets with more than 100k samples.
    The best three values in columns are highlighted with \textcolor{gold}{gold}, \textcolor{silver}{silver}, and \textcolor{bronze}{bronze} colors. For Elo values, we also indicate their approximate 95\% confidence intervals obtained through bootstrapping.} 
    \label{tab:leaderboard}
    \resizebox{\textwidth}{!}{
        \addtolength{\tabcolsep}{-0.4em}
    \begin{tabular}{lcccccrr}
\toprule
\textbf{Model} & \textbf{Elo ($\uparrow$)} & \textbf{Improva-} & \textbf{Avg.} & \textbf{Harm.} & \textbf{\#wins ($\uparrow$)} & \textbf{Train time} & \textbf{Predict time} \\
 &  & \textbf{bility ($\downarrow$)} & \textbf{rank ($\downarrow$)} & \textbf{mean} &  & \textbf{per 1K [s]} & \textbf{per 1K [s]} \\
 &  &  &  & \textbf{rank ($\downarrow$)} &  &  &  \\
\midrule
RealMLP (T+E) & \textcolor{gold}{\textbf{1283${}_{-27,+30}$}} & \textcolor{bronze}{\textbf{10.3\%}} & \textcolor{gold}{\textbf{7.2}} & \textcolor{bronze}{\textbf{3.8}} & \textcolor{bronze}{\textbf{9.9}} & 1316.51 & 14.54 \\
TabPFN-2.6 (D) & \textcolor{silver}{\textbf{1224${}_{-35,+39}$}} & \textcolor{gold}{\textbf{9.6\%}} & \textcolor{silver}{\textbf{8.9}} & \textcolor{silver}{\textbf{3.0}} & \textcolor{silver}{\textbf{24.5}} & 25.26 & 2.39 \\
TabM (T+E) & \textcolor{bronze}{\textbf{1212${}_{-25,+29}$}} & 11.8\% & \textcolor{bronze}{\textbf{9.3}} & 5.1 & 5.1 & 940.25 & 6.13 \\
CatBoost (T+E) & 1208${}_{-19,+27}$ & 12.4\% & 9.4 & 5.4 & 6.4 & 425.71 & 1.10 \\
TabICLv2 (D) & 1205${}_{-42,+39}$ & \textcolor{silver}{\textbf{9.8\%}} & 9.5 & \textcolor{gold}{\textbf{2.7}} & \textcolor{gold}{\textbf{33.3}} & 11.41 & 2.42 \\
RealMLP (T) & 1202${}_{-23,+31}$ & 11.9\% & 9.6 & 5.1 & 6.9 & 1316.51 & 1.68 \\
CatBoost (T) & 1189${}_{-21,+27}$ & 12.5\% & 10.0 & 5.7 & 5.7 & 425.71 & 0.20 \\
LightGBM (T+E) & 1188${}_{-22,+26}$ & 12.7\% & 10.0 & 5.9 & 5.7 & 163.08 & 2.42 \\
TabM (T) & 1169${}_{-26,+26}$ & 12.5\% & 10.6 & 5.8 & 5.1 & 940.25 & 0.97 \\
LightGBM (T) & 1149${}_{-23,+23}$ & 13.3\% & 11.3 & 6.9 & 3.0 & 163.08 & 0.61 \\
CatBoost (D) & 1139${}_{-20,+29}$ & 13.8\% & 11.6 & 7.3 & 2.8 & 14.38 & 0.17 \\
XGBoost (T+E) & 1138${}_{-22,+29}$ & 14.6\% & 11.7 & 7.2 & 2.9 & 271.48 & 6.83 \\
XGBoost (T) & 1112${}_{-22,+26}$ & 14.9\% & 12.5 & 8.2 & 2.6 & 271.48 & 1.68 \\
TabM (D) & 1111${}_{-24,+25}$ & 13.8\% & 12.6 & 7.9 & 2.7 & 24.50 & 0.89 \\
RealMLP (D) & 1056${}_{-25,+23}$ & 15.8\% & 14.4 & 8.8 & 2.5 & 43.31 & 1.55 \\
TabDPT (D) & 1052${}_{-44,+34}$ & 17.3\% & 14.6 & 5.6 & 9.2 & 107.79 & 11.22 \\
ExtraTrees (T+E) & 1008${}_{-27,+23}$ & 18.1\% & 16.0 & 11.0 & 1.1 & 113.01 & 1.62 \\
XGBoost (D) & 1000${}_{-25,+29}$ & 17.6\% & 16.3 & 11.8 & 1.0 & 6.59 & 1.35 \\
LightGBM (D) & 991${}_{-18,+23}$ & 17.7\% & 16.6 & 12.2 & 1.4 & 3.47 & 0.23 \\
ExtraTrees (T) & 970${}_{-29,+26}$ & 19.0\% & 17.2 & 11.8 & 1.0 & 113.01 & 0.34 \\
RandomForest (T+E) & 962${}_{-30,+23}$ & 19.4\% & 17.5 & 12.9 & 0.8 & 143.88 & 1.47 \\
RandomForest (T) & 945${}_{-27,+24}$ & 19.7\% & 18.0 & 13.3 & 0.8 & 143.88 & 0.41 \\
Linear (T+E) & 889${}_{-37,+28}$ & 26.3\% & 19.7 & 11.5 & 2.0 & 204.03 & 0.90 \\
RandomForest (D) & 875${}_{-32,+25}$ & 22.2\% & 20.1 & 15.5 & 0.6 & 9.76 & 0.69 \\
ExtraTrees (D) & 863${}_{-41,+34}$ & 23.1\% & 20.4 & 14.1 & 1.2 & 6.31 & 0.72 \\
Linear (T) & 854${}_{-41,+31}$ & 27.2\% & 20.7 & 12.5 & 1.8 & 204.03 & 0.21 \\
Linear (D) & 792${}_{-50,+39}$ & 29.7\% & 22.2 & 13.6 & 2.0 & 6.44 & 0.22 \\
\bottomrule
\end{tabular}
\space
}
\end{table}

\subsection{TabArena-v0.1 Datasets vs. BeyondArena Datasets in the Same Scope} \label{appendix:ta_datasets_comparison}
\begin{figure}
    \centering
    \includegraphics[width=0.75\columnwidth]{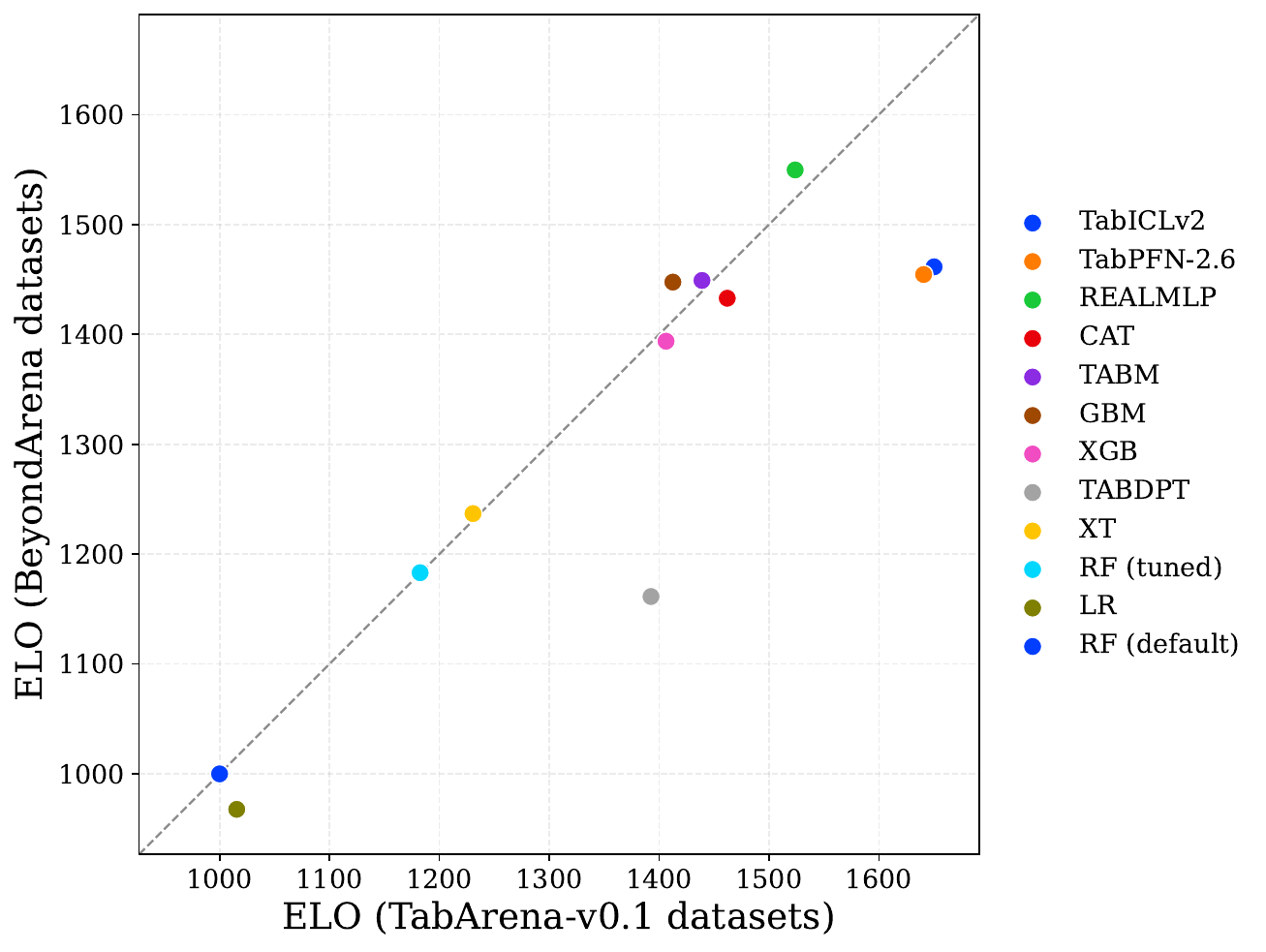}

    \caption{\textbf{Elo parity scatterplot comparing the $\mathbf{49}$ accepted TabArena datasets with $\mathbf{20}$ BeyondArena datasets within the same scope.} The dashed diagonal indicates equal performance relative to a default random forest across the two dataset groups.}
    
    \label{fig:tabarena_compare}
\end{figure}

To assess whether the newly added datasets that were in the same scope as the TabArena datasets pose a bigger challenge, we compare ELO on two dataset subsets: (1) the $49$ accepted TabArena datasets, (2) $20$ datasets which fall in the selection criteria of TabArena (IID, $500-250,000$ samples). 
In \Cref{fig:tabarena_compare}, we see that while tabular foundation models (TFMs) dominate on the TabArena datasets, they are outperformed by RealMLP on the new datasets. While Elo performance is stable for non-foundation models, all three evaluated TFMs degrade substantially on the new datasets. This can be explained by the fact that several of the new dataset additions are at the upper end of TabArena's sample-size limits, are often high-dimensional, and frequently contain high-cardinality categorical features. 
Overall, this confirms that even datasets within the scope of previous benchmarks can help measure future improvements of TFM on more challenging datasets.

\clearpage
\newpage

\subsection{Which dataset properties explain TFM performance when controlling for the others?}
\label{appendix:multivariate}

BeyondArena's sub-benchmarks overlap (e.g., large datasets are over-represented among temporal and high-dimensional datasets), so a TFM weakness on one sub-benchmark may reflect a weakness on another.
To separate these effects, we regress the rank of the best TFM per dataset on eight binary dataset properties, using the same definitions as our sub-benchmarks (temporal, grouped, tiny, small, large, high-dim, text, high-cardinality; reference: IID, medium, low-dim, no text, no high-cardinality).
We rank $28$ entries per outer split (our $27$ leaderboard entries plus TabPFN-3), take the best rank among TFMs (TabICLv2, TabPFN-2.6, TabDPT, TabPFN-3), and average it over a dataset's splits; higher ranks are worse for TFMs.
We exclude imputed results, so on large datasets the TFMs are TabICLv2 and TabPFN-3.
We fit ordinary least squares over the $142$ datasets with heteroskedasticity-robust (HC3) standard errors~\citep{mackinnon1985hc3}.

\Cref{tab:multivariate_results} shows that only two properties are significant: more than $100$k training rows (the best TFM's rank worsens by $8.0$ of $28$, $p < 0.001$) and grouped splits ($5.9$, $p < 0.001$); both remain significant after Bonferroni correction.
Temporal splits, high dimensionality, and high-cardinality features are significant on their own, but not once size is controlled for, because they co-occur with large datasets: $57\%$, $35\%$, and $43\%$ of temporal, high-dimensional, and high-cardinality datasets are large, compared with $8\%$, $4\%$, and $8\%$ of the remaining datasets.
Grouped splits are almost uncorrelated with size ($r = 0.07$) and remain significant when restricting to the $120$ datasets with at most $100$k training rows ($5.6$, $p < 0.001$).
The conclusions remain unchanged when we use the margin between the best TFM and the best non-TFM as the target, count imputed results as TFM results, or use only TabICLv2 and TabPFN-3 as TFMs.

\begin{table}[h]
\centering
\caption{\textbf{Multivariate analysis.} OLS coefficients for the best-TFM rank (of $28$; higher is worse for TFMs) with HC3 $95\%$ CIs; univariate coefficients fit each property alone. $R^2 = 0.59$, $n = 142$ datasets.}
\label{tab:multivariate_results}
\small
\begin{tabular}{lrrrr}
\toprule
Property & Coefficient & 95\% CI & $p$ & Univariate \\
\midrule
Temporal   & $0.18$  & [$-2.77$, $3.12$] & $0.91$ & $3.78$ \\
Grouped    & $\mathbf{5.94}$ & [$3.22$, $8.66$] & $<0.001$ & $6.35$ \\
Tiny       & $-0.29$ & [$-2.25$, $1.67$] & $0.77$ & $-2.05$ \\
Small      & $-1.38$ & [$-3.24$, $0.49$] & $0.15$ & $-3.39$ \\
Large      & $\mathbf{7.98}$ & [$4.68$, $11.27$] & $<0.001$ & $9.33$ \\
High-dim   & $-0.64$ & [$-2.34$, $1.07$] & $0.46$ & $2.44$ \\
Text       & $-1.27$ & [$-4.65$, $2.10$] & $0.46$ & $-0.52$ \\
High-card. & $1.87$  & [$-0.99$, $4.73$] & $0.20$ & $4.23$ \\
\bottomrule
\end{tabular}
\end{table}

\section{Statistical Analysis}
\label{app:posthoc}

We follow the standard practice for comparing multiple learning algorithms across many datasets \citep{demvsar2006statistical}. All tests are based on \emph{normalized error} $e_{\text{norm}} \in [0,1]$: on each dataset, methods are scored from best (0) to worst (1) among the 27 methods (11 default, 8 tuned, 8 tuned + ensembled), then averaged across cross-validation folds. Normalized errors let us work on a common scale, even though the benchmark spans heterogeneous tasks with their respective metrics: ROC AUC, log-loss, and RMSE. 
\\
Whenever a test produces multiple comparisons, we control the family-wise error rate with the \textbf{Holm--\v{S}id\'{a}k correction} \citep{holm1979simple,vsidak1967rectangular}. Holm makes no assumption that the individual tests are independent, which matters here because comparisons across the same 142 datasets yield p-values that are inherently correlated rather than independent.
We report effect sizes alongside corrected $p$-values.

\subsection{Test 1: Are the methods really different from each other?}
\label{app:posthoc:t1}

The first question to settle is whether the leaderboard ranking is genuinely informative or whether it could plausibly arise from random fluctuations across the 142 datasets. We answer this with a \textbf{Friedman test} \citep{friedman1937use}. The Friedman test ranks the 27 methods within each dataset and asks whether the resulting rank distributions across datasets are consistent with all methods being equivalent. Because ranks are computed within each dataset, the test naturally accounts for the fact
that the same 142 datasets are scored under every method.
\\
The Friedman test rejects the null hypothesis: the observed test statistic is 
$\chi^2_{26} = 1374.9$, giving $p < 10^{-273}$. 

To go beyond this global statement and identify which specific pairs of methods differ, we follow up with \textbf{pairwise Wilcoxon signed-rank} tests~\citep{wilcoxon1945individual} over all
$\binom{27}{2}=351$ method pairs, with Holm correction. For each pair of methods $(A, B)$ with normalized errors
$\{e_i^A\}_{i=1}^{N}$ and $\{e_i^B\}_{i=1}^{N}$ on the $N=142$ datasets,
let $d_i = e_i^A - e_i^B$ be the per-dataset paired difference. The
Wilcoxon signed-rank test drops the magnitudes of the $d_i$ and works only with the ranks of their absolute values $|d_i|$, keeping their signs. The hypotheses are:
\begin{align*}
  H_0 &: \text{the differences } d_i \text{ are symmetrically distributed around zero}; \\
  H_1 &: \text{the differences are not symmetric around zero (two-sided).}
\end{align*}

We reject $H_0$ at level $\alpha=0.05$ when the corresponding two-sided $p$-value falls below $0.05$.
Of those 351 pairs, \textbf{260 (74.6\%)} show performance differences that survive Holm correction. This sharpens the global Friedman result: not only is the leaderboard ranking real, but most individual pairs of methods can be reliably distinguished from one another.

\subsection{Test 2: At what training-set size does using a TFM stop being a reasonable choice?}
\label{app:posthoc:t4}

We split the 142 datasets into the four scale categories used throughout the
paper --- tiny (100-1k rows), small (1k-10k), medium (10k-100k), and
large ($> 10^5$) --- and compare the distribution of TFM normalized error
across them with a \textbf{Kruskal-Wallis test} \citep{kruskal1952use}. In Test~\ref{app:posthoc:t1}, every dataset is scored under all 27 methods, so the methods can be ranked \emph{within each dataset} and Friedman aggregates those within-dataset rankings across the 142 rows. Here, by contrast, each dataset belongs to exactly one scale category: the four scale groups are disjoint sets of datasets so Wilcoxon and Friedman are not suitable.

Formally, let $g_i \in \{1, 2, 3, 4\}$ denote the scale category of
dataset $i \in \{1, \dots, N\}$ with $N = 142$, and let $e_i$ be the
TFM normalized error on dataset $i$. We rank the $N$ datasets globally
on $e_i$ from best ($1$) to worst ($N$) across all four groups
combined, and let $\bar{R}_g$ denote the mean rank within scale group
$g$, with $n_g$ the number of datasets in that group. The hypotheses are:
\begin{align*}
  H_0 &: \text{all four scale groups have the same distribution of TFM normalized error}; \\
  &\quad\text{i.e., dataset size has no systematic effect on TFM error.} \\
  H_1 &: \text{at least one group is stochastically different from the others;} \\
  &\quad\text{i.e., TFM error tends to be larger (or smaller) on datasets of some sizes than others.}
\end{align*}
The Kruskal-Wallis test statistic is
\begin{equation*}
  H \;=\; \frac{12}{N(N+1)} \sum_{g=1}^{4} n_g \left(\bar{R}_g - \frac{N+1}{2}\right)^{2},
\end{equation*}
which under $H_0$ is approximately $\chi^2$-distributed with $k - 1 = 3$ degrees of freedom. We reject $H_0$ at significance level $\alpha = 0.05$ when the corresponding $p$-value falls below $0.05$. 
\\
The Kruskal-Wallis statistic is $H=30.6$ on 3 degrees of freedom ($p=1.0 \times 10^{-6}$), so the four scale categories do \emph{not} share
the same distribution of TFM errors. 

Kruskal-Wallis tells us that the four scale groups are not identically
distributed, but it does not say \emph{which} pairs differ or in which
direction. To localize where the
change actually happens, we follow up with the two-sample analogue of
Kruskal-Wallis: \textbf{p   airwise Mann-Whitney $U$ tests} \citep{mann1947test}
between every pair of scale categories, with Holm correction over the
$\binom{4}{2} = 6$ comparisons (Table~\ref{tab:app:kw_scale}).
\\
The pattern is clear. Tiny, small, and medium scales are statistically
indistinguishable from each other (every pairwise $p_\text{adj} > 0.5$): a
TFM does not perform worse on a 50k-row dataset than
on a 500-row one. 
This pattern breaks at the large-scale boundary: every comparison involving large datasets is highly significant ($p_\text{adj}<0.001$) and the rank-biserial effect sizes are large ($r$ between $0.68$ and $0.79$).
The takeaway is that TFM degradation happens at \emph{$\sim100,000$ rows}.

\begin{table}[h]
\centering
\caption{Pairwise Mann-Whitney $U$ tests for TFM normalized error across
         dataset scales, Holm-corrected over six comparisons. $\bar{e}_A$
         and $\bar{e}_B$ are the mean TFM normalized errors in scales
         $A$ and $B$. The rank-biserial correlation $r$ measures the
         size of the gap on a $[-1, +1]$ scale: a positive $r$ means scale
         $B$ has the higher (worse) error. \textbf{The three rows comparing
         \emph{large} against the other scales are statistically
         significant after correction (in bold)}; the other three are not.
         \emph{For example:} the third row says that
         comparing tiny ($n_A=52$) against large ($n_B=22$), the mean TFM
         error rises from $0.183$ to $0.528$, with a large rank-biserial
         effect size of $r = +0.71$ (TFM errors on tiny datasets are
         systematically smaller than on large datasets), and the
         corresponding Holm-corrected $p$-value is below $0.001$.}
\label{tab:app:kw_scale}
\small
\begin{tabular}{llccccc}
\toprule
Scale A & Scale B & $n_A$ & $\bar{e}_A$ & $\bar{e}_B$ & $r$ & $p_\text{adj}$ \\
\midrule
tiny  ($n=52$) & small  ($n=38$) & & 0.183 & 0.135 & $-0.11$ & 0.769 \\
tiny           & medium ($n=30$) & & 0.183 & 0.212 & $+0.06$ & 0.769 \\
tiny           & large  ($n=22$) & & 0.183 & 0.528 & $+0.71$ & \textbf{$<0.001$\,***} \\
small          & medium          & & 0.135 & 0.212 & $+0.20$ & 0.505 \\
small          & large           & & 0.135 & 0.528 & $+0.79$ & \textbf{$<0.001$\,***} \\
medium         & large           & & 0.212 & 0.528 & $+0.68$ & \textbf{$<0.001$\,***} \\
\bottomrule
\end{tabular}
\end{table}

\subsection{Test 3: How much do tuning and ensembling help non-TFM models?}
\label{app:posthoc:t5}

Traditional models seem to require tuning and ensembling to reach their full potential, while TFMs rely on in-context learning by default. 
We now test each of the eight traditional algorithms in BeyondArena to determine whether tuning and post hoc ensembling are actually required. 
\\
For each algorithm, across all 142 datasets, we have three matched observations: default error, tuned error, and tuned-plus-ensembled error.
The matched structure is the right setting for a \textbf{Wilcoxon signed-rank
test}. We run two such tests per algorithm --- (default $\to$ tuned) and (tuned $\to$ tuned+ensemble) --- for a total of 16 tests jointly Holm-corrected.

Formally, fix a traditional algorithm $a \in \{$CatBoost, LightGBM, Linear,
RandomForest, RealMLP, TabM, XGBoost, ExtraTrees$\}$, and let
$e_i^{\text{def}}, e_i^{\text{tun}}, e_i^{\text{ens}}$ denote its
normalized errors on dataset $i \in \{1,\dots,142\}$ when run with default
hyperparameters, with tuned hyperparameters, and with tuned-plus-ensembled
hyperparameters respectively. We test two paired hypotheses per algorithm:
\begin{align*}
  H_0^{(\text{tune})} &: \text{the per-dataset differences } d_i^{(\text{tune})} = e_i^{\text{def}} - e_i^{\text{tun}} \text{ are symmetric around } 0; \\
  H_0^{(\text{ens})}  &: \text{the per-dataset differences } d_i^{(\text{ens})}  = e_i^{\text{tun}} - e_i^{\text{ens}} \text{ are symmetric around } 0.
\end{align*}
Under each $H_0$, the matched pairs convey no systematic improvement; under the corresponding $H_1$, one pipeline variant is consistently better
than the other.
As in Test~\ref{app:posthoc:t1}, we compute the signed-rank statistic on the differences. %
We reject $H_0$ at level $\alpha = 0.05$ when the corresponding two-sided $p$-value falls below $0.05$ \emph{after Holm correction across all
$2 \times 8 = 16$ tests jointly}.
\\
Every single one of the 16 tests is significant at $p_\text{adj}<0.001$
(Table~\ref{tab:app:wilcoxon_tuning}). \textbf{Both HPO and ensembling consistently
improve traditional models across the entire benchmark}. RealMLP shows the largest jump from tuning to ensembling ($r=0.87$), while LightGBM shows the largest jump from the default to its tuned version ($r=0.89$).

\begin{table}[h]
\centering
\caption{Wilcoxon signed-rank tests for the impact of tuning and post-hoc ensembling, computed independently for each of the eight non-TFM algorithms ($n=142$ datasets per algorithm). The first three columns show the mean normalized error of each pipeline variant. The last two columns give the rank-biserial effect size $r$ for the two transitions; a positive $r$ indicates improvement. Every cell marked *** is significant at $p_\text{adj}<0.001$ after Holm correction over all 16 tests. The entry shown in bold per column is the largest effect.}
\label{tab:app:wilcoxon_tuning}
\small
\begin{tabular}{lcccccc}
\toprule
& \multicolumn{3}{c}{Mean normalized error}
& \multicolumn{2}{c}{Effect size $r$} \\
\cmidrule(lr){2-4}\cmidrule(lr){5-6}
Algorithm & Default & Tuned & +Ensemble
          & Default$\to$Tuned & Tuned$\to$Ens.\ \\
\midrule
CatBoost  & 0.204 & 0.174 & 0.167 & 0.48\,*** & 0.51\,*** \\
LightGBM  & 0.343 & 0.195 & 0.176 & \textbf{0.89\,***} & 0.63\,*** \\
Linear    & 0.713 & 0.614 & 0.571 & 0.53\,*** & 0.77\,*** \\
RandomForest & 0.497 & 0.393 & 0.379 & 0.57\,*** & 0.40\,*** \\
RealMLP   & 0.271 & 0.166 & 0.121 & 0.80\,*** & \textbf{0.87\,***} \\
TabM      & 0.237 & 0.189 & 0.173 & 0.46\,*** & 0.79\,*** \\
XGBoost   & 0.346 & 0.236 & 0.224 & 0.84\,*** & 0.58\,*** \\
ExtraTrees & 0.517 & 0.370 & 0.341 & 0.65\,*** & 0.65\,*** \\
\bottomrule
\end{tabular}
\end{table}

\subsection{Test 4: What dataset properties predict that GBDTs will beat TFMs?}
\label{app:posthoc:t8}

For each of the 142 datasets, we summarize the head-to-head outcome with a single signed quantity, $\Delta = \bar{e}_{\text{TFM}}^{\star} -\bar{e}_{\text{GBDT}}^{\star}$, where $\bar{e}^{\star}$ is the minimum normalized error achieved by any member of that family on the dataset. Negative $\Delta$ means the best TFM beat the best GBDT; positive $\Delta$ means the opposite. We then correlate $\Delta$ with a set of meta-features of our datasets using \textbf{Spearman's rank correlation}
$\rho$ \citep{spearman1961proof}. The eight meta-feature tests are jointly Holm-corrected.
\\
\textbf{Two meta-features stand out (Table~\ref{tab:app:metafeature}): dataset size and the number of high-cardinality categorical columns.} Both correlations are univariate, and the two meta-features are correlated ($43\%$ of high-cardinality datasets are large). Once size is controlled for in a multivariate analysis, high-cardinality features are no longer significant (\Cref{appendix:multivariate}). 
The maximum cardinality of any single categorical column also predicts a GBDT advantage. Missing values, while a weaker signal, also push the balance toward GBDTs.

\begin{table}[h]
\centering
\caption{Spearman rank correlation between dataset meta-features and the per-dataset TFM-vs-GBDT advantage $\Delta = e_\text{TFM} - e_\text{GBDT}$. \textbf{A positive $\rho$ means the meta-feature predicts a \emph{larger} GBDT advantage}. The sample size $n$ varies because a few features (e.g.\ minority-class share, max categorical cardinality) are undefined on regression tasks or on datasets with no categorical columns. We used Holm correction over all features.}
\label{tab:app:metafeature}
\small
\begin{tabular}{lccccl}
\toprule
Meta-feature & $n$ & $\rho$ & $p_\text{raw}$ & $p_\text{adj}$ & Sig.\ \\
\midrule
\texttt{num\_rows}                    & 142 & $+0.603$ & $< 10^{-14}$ & $< 10^{-13}$ & *** \\
\texttt{num\_high\_cardinality\_cats} & 142 & $+0.466$ & $< 10^{-8}$  & $< 10^{-7}$  & *** \\
\texttt{max\_categorical\_cardinality}& 105 & $+0.413$ & $< 10^{-4}$  & $< 10^{-3}$  & *** \\
\texttt{missing\_value\_fraction}     & 142 & $+0.286$ & $0.0006$     & $0.003$      & **  \\
\texttt{num\_cols\_after\_preproc}    & 142 & $+0.212$ & $0.011$      & $0.045$      & *   \\
\texttt{age}                          & 142 & $-0.242$ & $0.004$      & $0.019$      & *   \\
\texttt{minority\_class\_pct}         &  98 & $-0.211$ & $0.038$      & $0.075$      & ns  \\
\texttt{num\_cols}                    & 142 & $+0.197$ & $0.019$      & $0.056$      & ns  \\
\texttt{num\_text\_cols}              & 142 & $+0.051$ & $0.545$      & $0.545$      & ns  \\
\bottomrule
\end{tabular}
\end{table}

\subsection{Test 5: Which method should a practitioner pick, given their data regime?}
\label{app:posthoc:t9}

Test~\ref{app:posthoc:t1} established that there is a real global ranking across all 27 methods that is not noise, and Tests~\ref{app:posthoc:t4}--\ref{app:posthoc:t8} answered specific sub-questions about how dataset size, tuning, and dataset meta-features affect performance. None of these directly answers the question a practitioner actually faces: \emph{given a dataset with a specific combination of properties, which method should I deploy?}
A global ranking averaged across all 142 datasets may not match the ranking inside any one regime --- TFMs lead on average but lose to neural networks on large or grouped data, for instance. 
\\
To turn the benchmark into actionable model-selection guidance, we run Test~\ref{app:posthoc:t1}'s procedure separately within each of the 11 sub-benchmarks. The subsets correspond to one Friedman test per axis level along four axes: task type (IID, temporal, grouped — 3 subsets), dataset scale (tiny, small, medium, large — 4 subsets), dimensionality (low-dim, high-dim — 2 subsets), and special feature types (text, high-cardinality — 2 subsets), for 11 sub-benchmarks in total. 
Each subset gets its own \textbf{Friedman test} on the methods restricted to its datasets, followed by Holm-corrected \textbf{pairwise Wilcoxon tests} within that subset.

For each subset $s$ with $N_s$ datasets, the Friedman test asks whether the 27 methods are equivalent within that subset:
\begin{align*}
  H_0^{(s)} &: \text{all 27 methods have the same expected within-subset rank}; \\
  H_1^{(s)} &: \text{at least one method has a different expected rank within subset } s.
\end{align*}
When $H_0^{(s)}$ is rejected, we localize which method pairs differ by running pairwise Wilcoxon signed-rank tests within the subset. For each pair of methods $(A, B)$ with paired per-dataset normalized-error differences $d_i = e_i^A - e_i^B$ on the $N_s$ datasets in subset $s$:
\begin{align*}
  H_0^{(s, A, B)} &: \text{the differences } d_i \text{ are symmetrically distributed around zero}; \\
  H_1^{(s, A, B)} &: \text{the differences are not symmetric around zero (two-sided).}
\end{align*}
Holm correction is applied jointly over all $\binom{27}{2} = 351$ pairs within each subset.

A coherent picture emerges across the sub-benchmarks (\autoref{tab:app:t9_summary}): 
\textbf{TFMs} are the top-1 method on IID, tiny/small, and low-dimensional data. 
\textbf{Tuned and ensembled neural networks} (RealMLP) lead on temporal, grouped, medium, large, high-dimensional, and text subsets. 
\textbf{Tuned GBDTs} (CatBoost) lead on high-cardinality categorical data. 
The grouped subset is the one place where this narrative deserves a caveat: although RealMLP is the top-ranked method, only $5.1\%$ of pairwise comparisons within the grouped subset survive Holm correction, meaning the ranking among the remaining methods is uncertain. 

\begin{table}
\centering
\caption{The Friedman test is significant ($p<0.001$) in every one of the 11 subsets: the per-subset rankings are real. The fraction of pairwise comparisons that survive Holm correction within a subset (column ``\% sig.\ pairs'') varies considerably, from $75.5\%$ down to $5.1\%$. A high fraction means the subset is large and homogeneous enough that most methods can be reliably ordered against each other; a low fraction (the grouped subset, with only 18 datasets) means that even though some method is clearly best, the ranking among the others is too noisy. We report the top method and its family in each subset; for subsets with low sig.\ pairs, only the top method itself should be read as a robust recommendation. ``Top-1 method'' is the method with the smallest mean Friedman rank in the subset, with [D] denoting default-mode and [T+E] tuned-plus-ensembled.}
\label{tab:app:t9_summary}
\small
\begin{tabular}{llrrrrll}
\toprule
Subset & $n$ & $\chi^2$ & $p$ & Sig.\ pairs & \% sig.\ & Top-1 method & Family \\
\midrule
\multicolumn{8}{l}{\textit{Task type}} \\
task=random   & 103 & 1086.8 & $<10^{-3}$ & 265/351 & 75.5 & TabICLv2 (default) & TFM \\
task=temporal &  21 &  345.4 & $<10^{-3}$ & 154/351 & 43.9 & RealMLP (t+ens)    & Neural \\
task=grouped  &  18 &  147.4 & $<10^{-3}$ &  18/351 &  5.1 & RealMLP (t+ens)    & Neural \\
\midrule
\multicolumn{8}{l}{\textit{Dataset scale}} \\
scale=tiny    &  52 &  408.9 & $<10^{-3}$ & 163/351 & 46.4 & TabICLv2 (default) & TFM \\
scale=small   &  38 &  500.2 & $<10^{-3}$ & 214/351 & 61.0 & TabPFN-2.6 (default)  & TFM \\
scale=medium  &  30 &  481.9 & $<10^{-3}$ & 196/351 & 55.8 & RealMLP (t+ens)    & Neural \\
scale=large   &  22 &  398.5 & $<10^{-3}$ & 196/351 & 55.8 & RealMLP (t+ens)    & Neural \\
\midrule
\multicolumn{8}{l}{\textit{Dimensionality}} \\
dim=low-dim   &  91 &  927.5 & $<10^{-3}$ & 254/351 & 72.4 & TabICLv2 (default) & TFM \\
dim=high-dim  &  51 &  568.4 & $<10^{-3}$ & 207/351 & 59.0 & RealMLP (t+ens)    & Neural \\
\midrule
\multicolumn{8}{l}{\textit{Feature type}} \\
feature=text          &  16 &  229.2 & $<10^{-3}$ &  85/351 & 24.2 & RealMLP (t+ens) & Neural \\
feature=high-card.    &  30 &  502.5 & $<10^{-3}$ & 208/351 & 59.3 & CAT (t+ens)        & Tree \\
\bottomrule
\end{tabular}
\end{table}

\clearpage
\newpage

\section{Ablations}
\label{appendix:ablations}

We ablate \benchname to determine the validity of our results. We summarize the results in \Cref{tab:ablations} in the main paper. Here, we provide more details on the experimental setup for each ablation. 
\begin{enumerate}
    \item In \Cref{appendix:ablation_fewer_splits}, we introduce \texttt{BeyondArena-Core} and show that one can use fewer splits while obtaining similar representative results per dataset.
    \item In \Cref{appendix:ablation_iid_splits}, we show that using IID outer splits for non-IID grouped data significantly distorts the results. Benchmarks should use the correct splits for non-IID data.
    \item In \Cref{appendix:ablation_less_inner_splits}, we show that using $5 \times 5$ cross-validation instead of $8$-fold cross-validation for smaller data did lead to better performance for all models, making it a valid pipeline choice.
    \item In \Cref{appendix:ablation_iid_inner_splits}, we show that using non-IID inner splits for grouped and temporal data leads to better performance, solidifying the importance of using the appropriate splits per task.
    \item In \Cref{appendix:no_grouped_preprocessing}, we show that our grouped preprocessing leads to big gains, but not for all models. 
    \item In \Cref{appendix:ablation_text_encoding}, we show that our default text encoding method (Qwen3) is better than a classical TF-IDF method for short text. Yet, it is consistently beaten on large text. Our results remain representative, as the model rankings are highly correlated.
    \item In \Cref{appendix:ablation_probabiltily_calibration}, we show that probability calibration can significantly boost the performance, mostly for tree-based models. While \benchname did not use calibration, future benchmarks should.  
\end{enumerate}

\subsection{(Outer Splits, A.1) Using Fewer Splits with BeyondArena-Core}
\label{appendix:ablation_fewer_splits}

We use up to $60$ splits per dataset. Inspired by TabArena's Lite subset, which established that benchmark compute costs can be substantially reduced while preserving the aggregate ranking by using only the first split of each dataset, we introduce \texttt{BeyondArena-Core}.
We refine the concept of TabArena-Lite to not only preserve aggregate rankings but also to preserve per-dataset rankings, by adjusting the split budget on a per-dataset basis to achieve a per-dataset stability of $\Phi=0.8$.

Formally, for each dataset $d$ with $F_d$ available splits, we define $\rho_d$ as the mean off-diagonal Spearman correlation of dataset $d$'s $F_d{\times}F_d$ split-similarity matrix, i.e., averaged pairwise rank agreement across errors of methods between splits. 
For each stability target $\Phi \in \{0, 0.6, 0.8, 0.9, 0.95, 1\}$ and each dataset $d$, we determine a per-dataset split budget via Spearman-Brown extrapolation:
\begin{equation}
  k_d(\Phi) \;=\; \min\!\Bigl(F_d,\;
  \bigl\lceil \tfrac{\Phi\,(1-\rho_d)}{\rho_d\,(1-\Phi)} \bigr\rceil
  \Bigr),
  \qquad k_d(1) := F_d.
  \label{eq:k_needed}
\end{equation}
We then recompute the leaderboard restricted to those tasks and overlay two views on a shared percentage-points axis: (i)~aggregate per-method $|\Delta\mathrm{Winrate}|$ against the $\Phi=1$ reference (Mean and Max
curves), and (ii)~per-dataset bootstrap winrate jitter at $k_d(\Phi)$ as a strip plot with cross-dataset medians. The bootstrap is taken \emph{with replacement} so the $\Phi=1$ column reflects the irreducible variance of the all-splits estimate rather than collapsing to zero, providing a precision floor against which fidelity loss at
smaller $\Phi$ can be judged. The dashed reference line marks the per-split mean absolute deviation (MAD) under uniform-random rankings (an upper bound for the
$k>1$ regime).

\begin{figure}
    \centering
    \includegraphics[width=\textwidth]{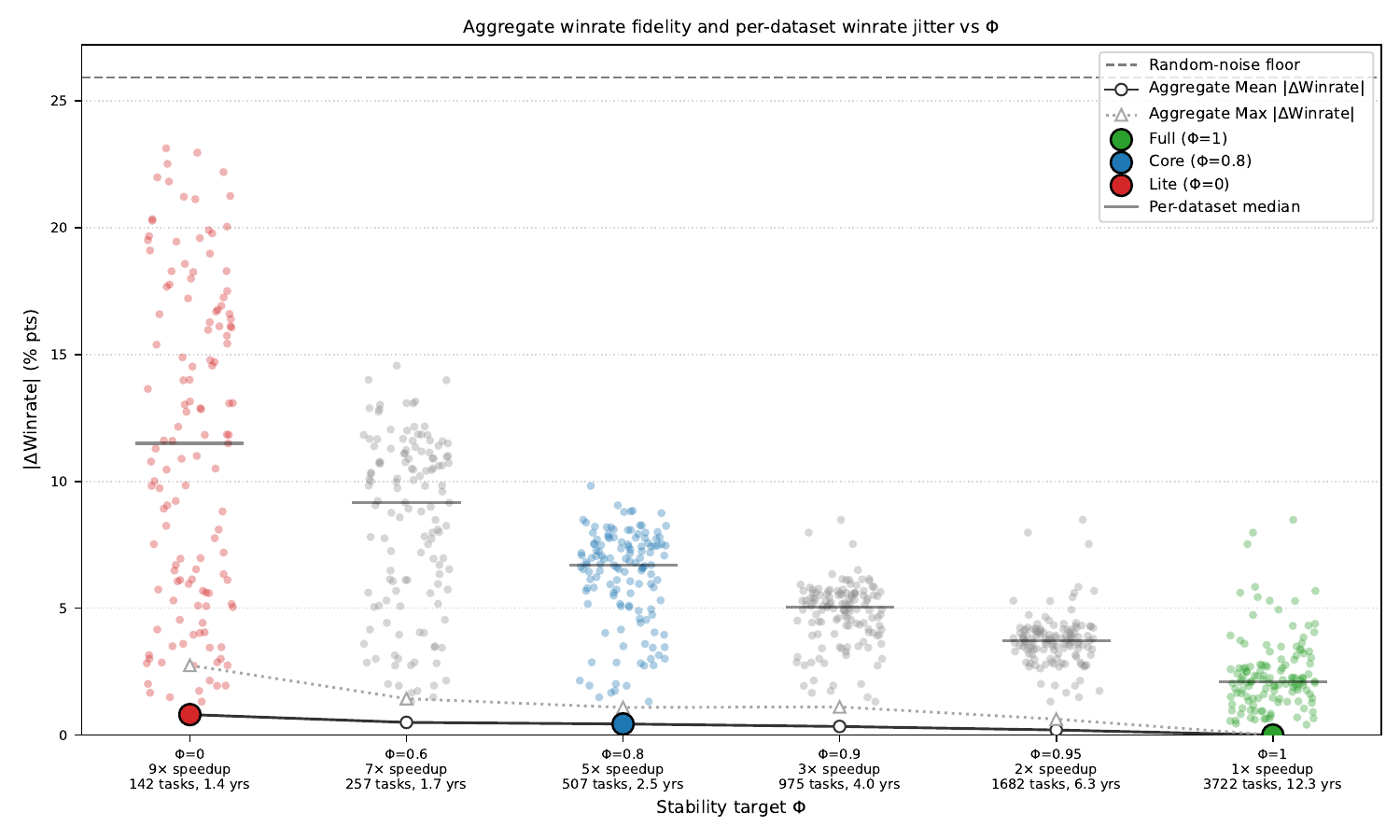}
    \caption{
    \textbf{Compute--fidelity trade-off across stability targets $\Phi$.}
    Aggregate per-method $|\Delta\mathrm{Winrate}|$ vs.\ the $\Phi=1$
    reference (Mean: solid, Max: dotted) overlaid on per-dataset
    bootstrap winrate jitter at the per-dataset split budget $k_d(\Phi)$
    (Eq.~\ref{eq:k_needed}); one translucent dot per dataset, black bars
    mark the cross-dataset median. The dashed line is the per-split MAD
    under uniform-random rankings (upper bound for $k>1$).
    }
    \label{fig:stability_combined}
\end{figure}

\Cref{fig:stability_combined} shows that the aggregate mean winrate delta is well below $1\%$ regardless of $\Phi$, meaning that even a single split per dataset is sufficient to faithfully represent the overall leaderboard and model rankings, aligning with TabArena's insights.
The per-dataset jitter median, however, grows substantially from Full through Lite: individual datasets' orderings become noticeably less stable as compute is reduced, even when the benchmark-wide ranking is
preserved. In the worst case ($\Phi=0$), several datasets have a winrate delta of over $20\%$, which is approaching random noise and thus insufficient for effective dataset-level analysis. By adding splits to satisfy $\Phi$, we efficiently increase compute for the most unstable datasets, resulting in Core ($\Phi=0.8$) having all datasets with a mean winrate delta below $10\%$ while costing only $60\%$ more than Lite.
\\
Benchmark users and creators should use a Core-like subset ($\Phi=0.8$) when reporting per-dataset results (rank tables, per-dataset wins/losses, dataset-conditioned analyses), which preserves per-dataset rank stability while being 5x cheaper than Full; those reporting only aggregate metrics (overall mean rank, headline winrate) can drop all the way to Lite ($\Phi=0$) and recover the full speedup (9x faster than Full) with no meaningful loss in fidelity.
\space

\subsection{(Outer Splits, A.2) IID Splits for non-IID Grouped Data}
\label{appendix:ablation_iid_splits}

\begin{figure}
    \centering
    \includegraphics[width=\textwidth]{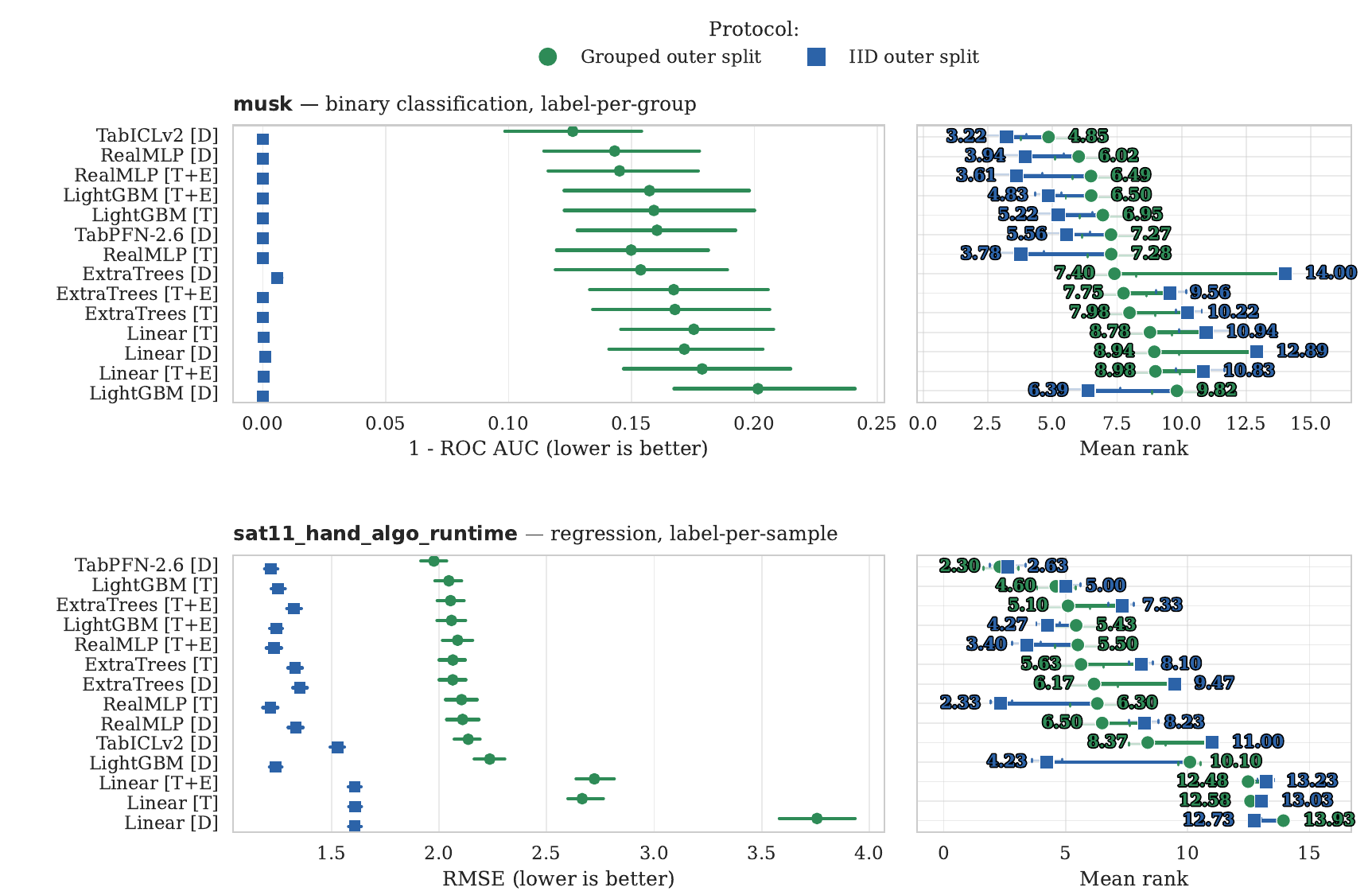}
    \caption{
    \textbf{Grouped vs. IID Outer Splits.}
    We compare changing the outer splits for two \cgrouped datasets. As a result, \textbf{(left)} the raw metric values and their variances change in a drastic manner, while \textbf{(right)} model ranks are heavily distorted (\texttt{musk} $\tau=0.60$, \texttt{sat11\_hand\_algo\_runtime} $\tau=0.49$).
    }
    \label{fig:ablation_iid_splits_non_iid_data}
\end{figure}

We investigate the impact on benchmark results when using IID outer train-test splits for two grouped datasets (\texttt{musk} and \texttt{sat11\_hand\_algo\_runtime}). 
We use the same experimental setup as described in \Cref{sec:experiment_setup}, but reduce the costs by using a representative subset of models: Linear, ExtraTrees, LightGBM, RealMLP, TabPFN-2.6, and TabICLv2.
\Cref{fig:ablation_iid_splits_non_iid_data} shows the results of running all models on the Grouped vs. IID outer splits. 
For \texttt{musk}, the task becomes trivial under IID splits, with almost all models having an error of 0. 
At the same time, the variance across folds and the model rankings are heavily distorted. 
\texttt{sat11\_hand\_algo\_runtime} it looks similiar. While the task does not become trivial, it becomes much simpler. Ranks are distorted again.
\\
We conclude, as with temporal data, that the appropriate outer split is crucial for comparing models. 

\subsection{(Inner Splits, B.1) Using 8-fold Cross-Validation for Tiny Data}
\label{appendix:ablation_less_inner_splits}

\begin{figure}
    \centering
    \includegraphics[width=\textwidth]{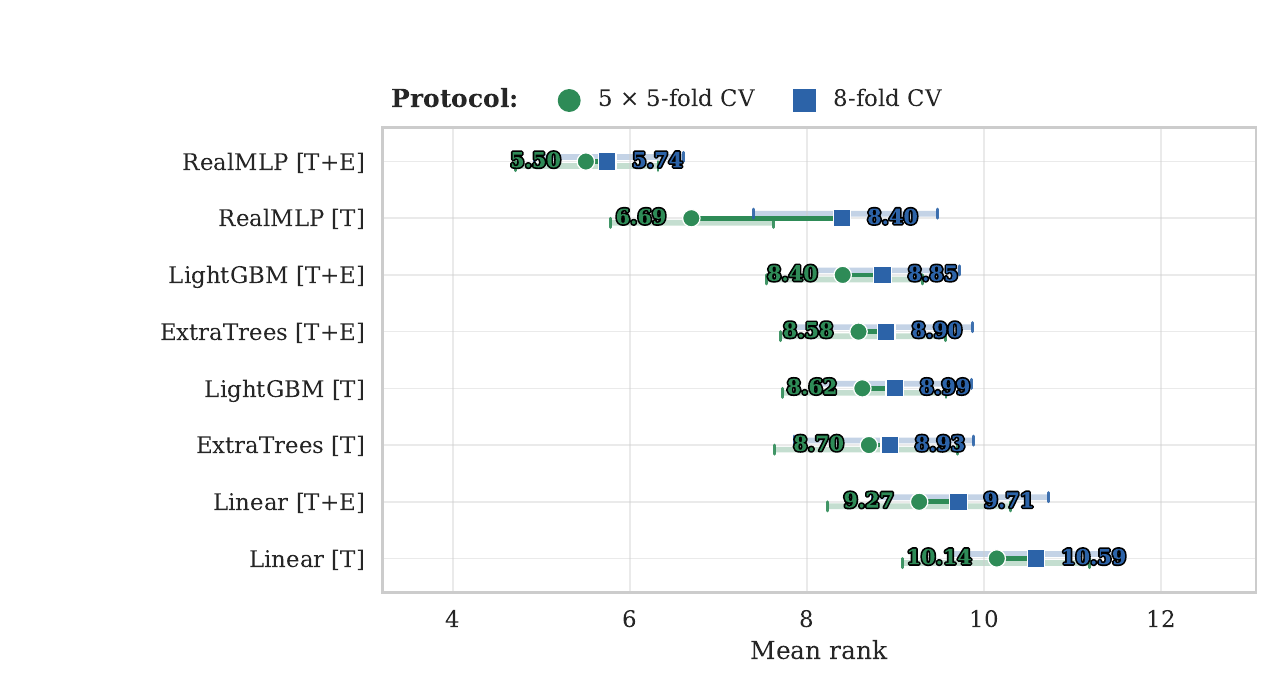}
    \caption{
    \textbf{Performance of Cross-validation (CV) Strategies for Tiny Data.}
    We compare Linear, ExtraTrees, LightGBM, and RealMLP with tuning and with post-hoc ensembling on datasets with fewer than 500 samples, when changing the inner splits from 5-repeated 5-fold ($5 \times 5$) CV to 8-fold CV; all other benchmark settings are identical.
    $5 \times 5$ CV consistently performs better than 8-fold. 
    Moreover, tuned RealMLP gains a drastic boost, which aligns with it being the most overtuned model in TabArena \citep{erickson2025tabarena}.
    }
    \label{fig:ablation_less_inner_splits}
\end{figure}

We investigate the impact of changing the inner splits from 5-repeated 5-fold ($5 \times 5$) CV to 8-fold CV
This ablation only uses datasets with $\le 500$ training samples. 
To reduce the cost of the ablation, we (a) compare on IID datasets with $\le100$ features, and (b) only run the first three outer folds. 
We then run the full tuning + ensembling for Linear, ExtraTrees, LightGBM, and RealMLP. 
We ignore TFMs because they are unaffected by the validation protocol; we don't tune them and use the full training data as in-context data at test time. 
\Cref{fig:ablation_less_inner_splits} shows that $5 \times 5$-fold CV performs better in all cases and gives the largest boost to RealMLP.
The rankings are comparable across both protocols.

\subsection{(Inner Splits, B.2) Using IID Splits for non-IID Data}
\label{appendix:ablation_iid_inner_splits}
\begin{figure}
    \centering
    \includegraphics[width=\textwidth]{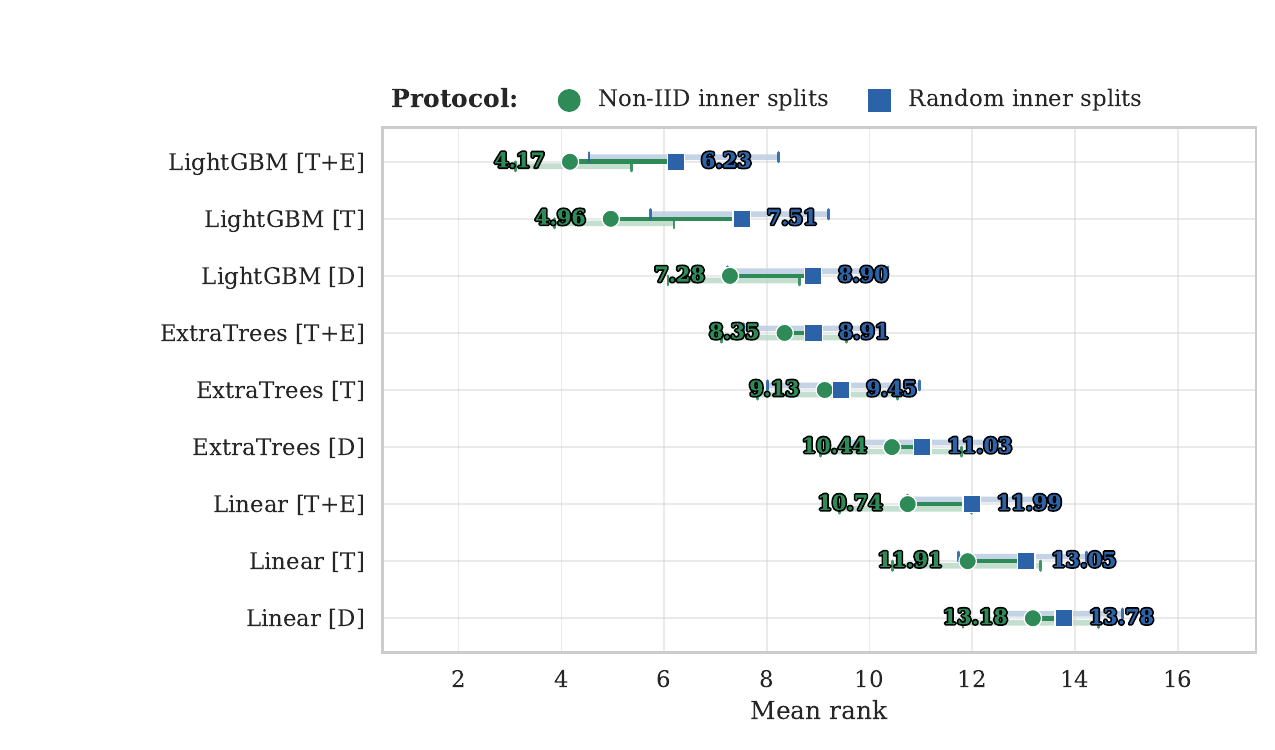}
    \caption{
    \textbf{Performance of Non-IID vs IID Inner Splits for Non-IID Data.}
    We compare Linear, ExtraTrees, and LightGBM when using non-IID inner splits (grouped or temporal) as a validation protocol against always using random inner splits. 
    We observe that Linear and LightGBM benefit significantly from non-IID splits, whereas ExtraTrees remains largely unaffected.
    Moreover, the ranking order is getting mixed: a tuned plus ensembled ExtraTrees model would beat a default LightGBM, only when using random inner splits.
    }
    \label{fig:ablation_iid_inner_splits}
\end{figure}

We used non-IID validation splits for all non-IID datasets in \benchname. 
Here, we investigate how performance changes when using IID splits for validation. 
We follow the experimental setup from \Cref{sec:experiment_setup}, using only Linear, ExtraTrees, and LightGBM, and running only the first outer split. 
In \Cref{fig:ablation_iid_inner_splits}, we observe that non-IID splits led to better performance in almost all cases, and in such cases, providing a significant boost in performance. 

\subsection{(Grouped Data Preprocessing, C.1a/b) Disabling Preprocessing for Grouped Data.}
\label{appendix:no_grouped_preprocessing}

\begin{figure}
    \centering
    \includegraphics[width=\textwidth]{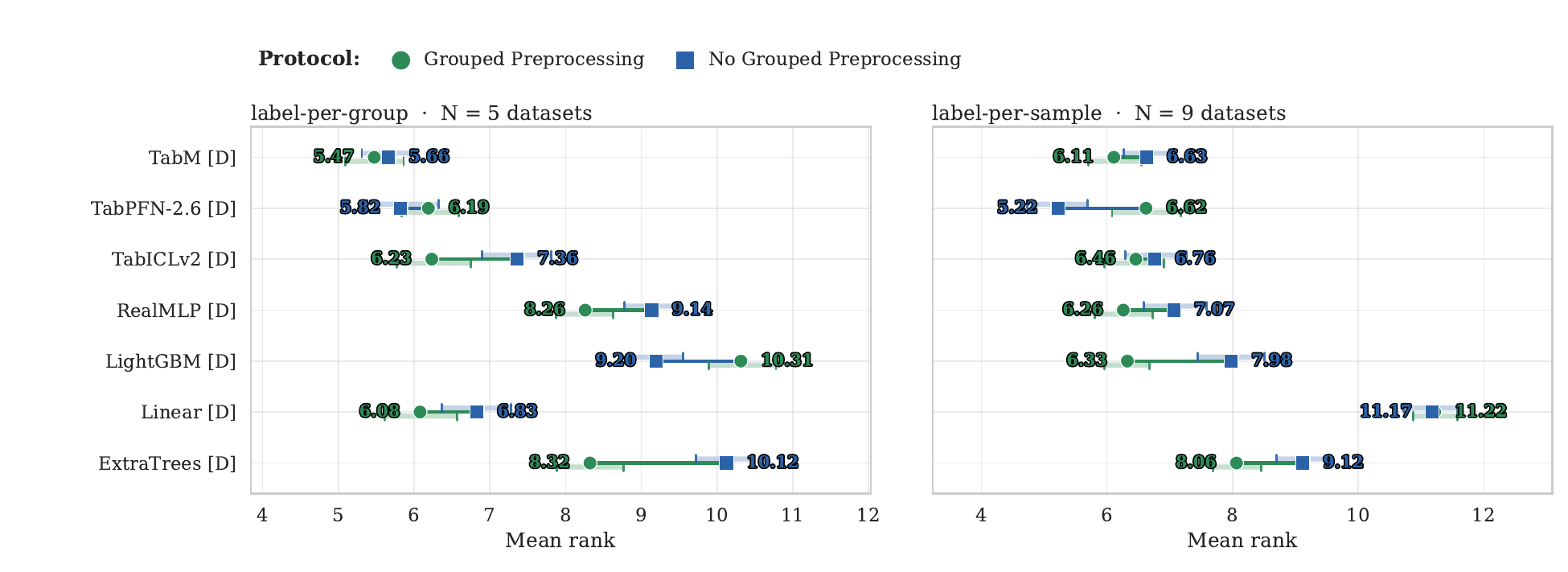}
    \caption{
    \textbf{Impact of Preprocessing for Grouped Data.}
    We compare the default performance of Linear, ExtraTrees, LightGBM, RealMLP, TabICLv2, TabPFN-2.6, and TabM with and without the group preprocessing introduced by \benchname, split by \texttt{label-per-group} (\texttt{L-P-G}) and \texttt{label-per-sample} (\texttt{L-P-S}) data.
    We observe that the performance gains depend on the model, while, on average, enabling grouped preprocessing led to better performance.
    TabPFN-2.6 is a distinguishable outlier compared to all other models, as it benefited the most from not using grouped preprocessing. 
    }
    \label{fig:no_grouped_preprocessing}
\end{figure}

In \benchname, we introduce a default model-agnostic preprocessing step for grouped data that follows industry practice. Here, we examine how the performance would change if we disable this preprocessing. 
We use the same experimental setup as in \Cref{sec:experiment_setup}, but only compare the default performance of Linear, ExtraTrees, LightGBM, RealMLP, TabM, TabPFN-2.6, and TabICLv2. 
Moreover, we restrict the ablation to only using datasets with $\le100k$ training samples. 
In \Cref{fig:no_grouped_preprocessing}, we split the ablation by \texttt{label-per-group} (\texttt{L-P-G}) and \texttt{label-per-sample} (\texttt{L-P-S}) datasets, showing that for both, the rankings can change a lot while performance gains with grouped preprocessing are higher on average. For TabPFN-2.6, no preprocessing would have been better. 

\subsection{(Text Data Preprocessing, C.2a/b) TF-IDF for Text Encoding.}
\label{appendix:ablation_text_encoding}

\begin{figure}
    \centering
    \includegraphics[width=\textwidth]{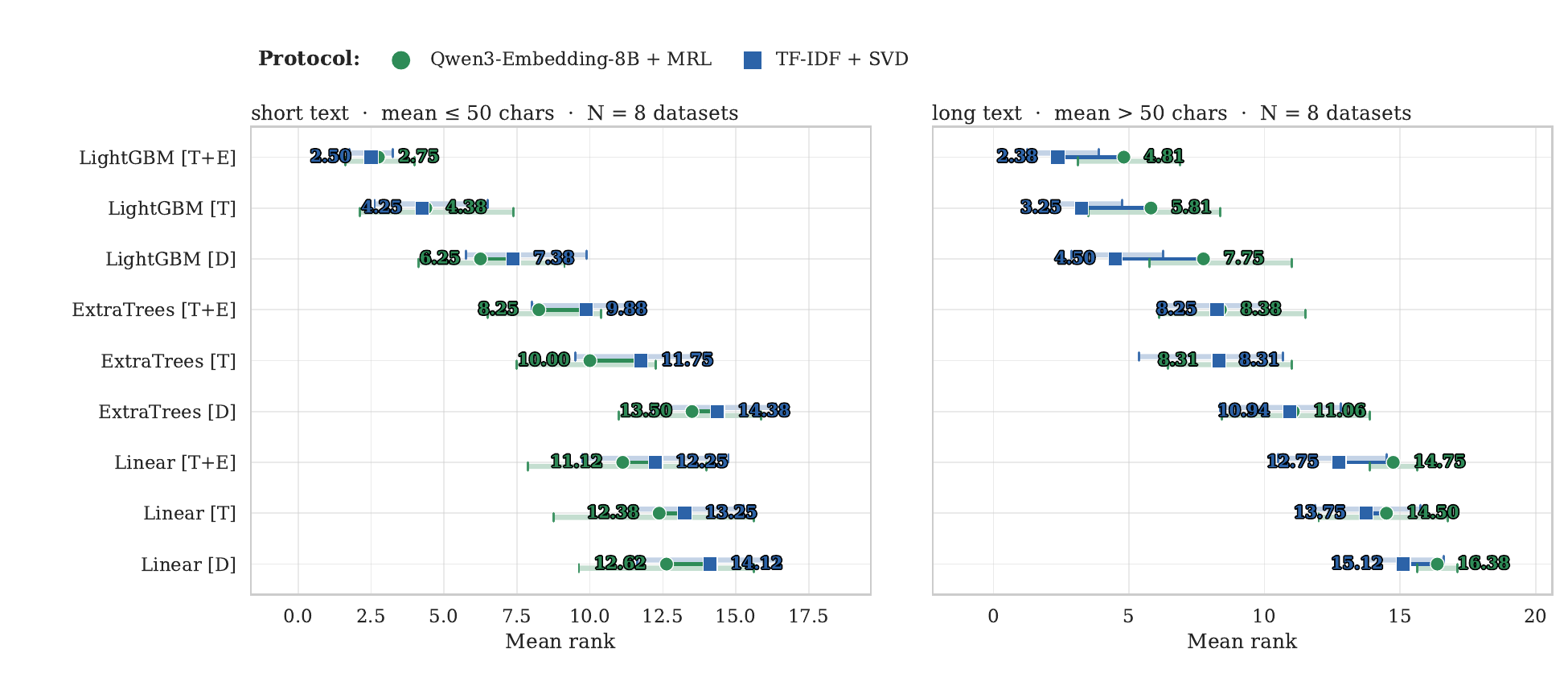}
    \caption{
    \textbf{LM vs. TF-IDF for Encoding Text Features.}
    We compare the performance of Linear, ExtraTrees, and LightGBM when changing the preprocessing of text features from a language model (LM), in the form of Qwen3-Embedding-8B with Matryoshka representation learning (MRL), to a classical encoding method, in the form of TF-IDF with SVD.
    Both approaches create a 32-dimensional output vector for each text feature. 
    We split the results across short-text ($\le50$ characters on average across all text cells) and long-text datasets. 
    For short texts, the LM consistently performs better, while for long texts, TF-IDF dominates.  
    In both cases, the model rankings are not affected by the preprocessing choice. 
    }
    \label{fig:ablation_text_encoding}
\end{figure}

\benchname contains tabular data with text, which requires encoding for tabular machine learning models. 
By default, we used Qwen3-Embedding-8B with Matryoshka representation learning (MRL) for text encoding. Here, we ablate using TF-IDF from Skrub with SVD.
We run the ablation across all datasets with text, following the experimental setup in \Cref{sec:experiment_setup}. We use only the first split of each dataset and restrict the runs to the cheaper CPU models: Linear, ExtraTrees, and LightGBM. 
\Cref{fig:ablation_text_encoding} shows mixed results. For short text, Qwen3 was the better choice. For long text, TF-IDF was the better choice. In the future, one would need to tune the preprocessing method for each dataset to achieve peak performance. 

\subsection{(Post-processing, D) Using Probability Calibration for Log-loss.}
\label{appendix:ablation_probabiltily_calibration}

\begin{figure}
    \centering
    \includegraphics[width=\textwidth]{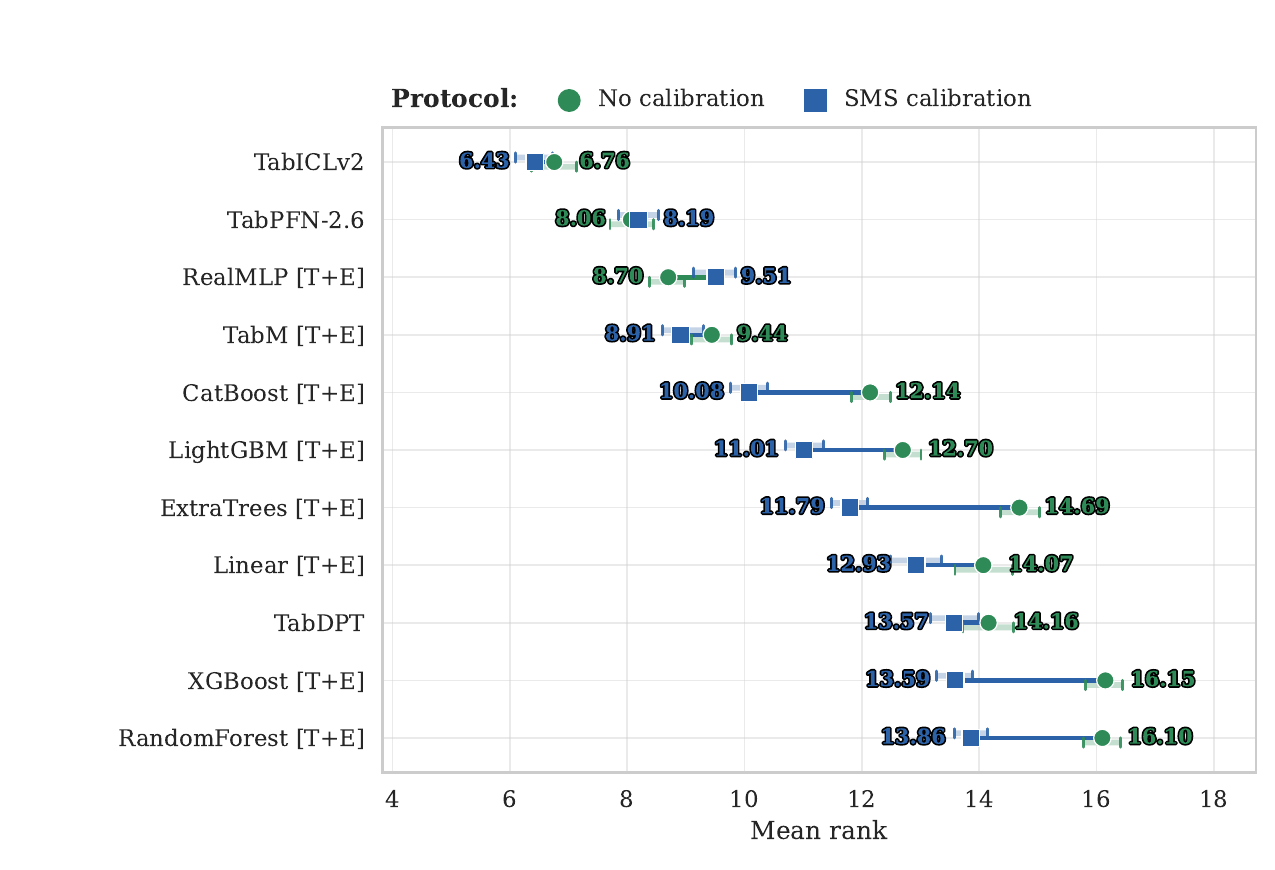}
    \caption{
    \textbf{No Calibration vs. Probability Calibration for Multiclass Classification.}
    We compare the performance of all models when adding post hoc probability calibration (via SMS calibration \citep{berta2025structured}) after tuning and ensembling for multiclass classification. 
    SMS calibration yields significant gains in average rank for several models and alters the model rank order. 
    }
    \label{fig:ablation_calibration}
\end{figure}

\benchname uses log-loss as a metric for multiclass classification.
Post hoc calibration can improve the model's performance when measured by log-loss. 
Since \benchname does not use post hoc calibration, we ablate using structured matrix scaling (SMS) for calibration \citep{berta2025structured}, the default of the \texttt{probmetrics} framework \citep{berta2025rethinking}.
We recompute the log-loss of all methods after SMS using the saved predictions from the main benchmark. 
In \Cref{fig:ablation_calibration}, we show that SMS calibration improved the performance for most methods. 
In particular, tree-based models improved significantly, while only TabPFN-2.6 and RealMLP performed worse in calibration.

\clearpage
\newpage
\section{Per Dataset Results}
\label{appendix:results_per_dataset}
\noindent \textbf{Performance Per Dataset.} We show the average predictive performance per dataset with the standard deviation over folds. We show the performance for the default hyperparameter configuration (\texttt{Default}), for the model after tuning (\texttt{Tuned}), and for the ensemble after tuning (\texttt{Tuned + Ens.}). We highlight the best-performing methods with significance on three levels: (1) \textcolor{green!50!black}{Green}: The best performing method on average; (2) \textbf{Bold}: Methods that are not significantly worse than the best method on average, based on a Wilcoxon Signed-Rank test for paired samples with Holm-Bonferroni correction and $\alpha=0.05$. (3) \underline{Underlined}: Methods that are not significantly worse than the best method in the same pipeline regime (\texttt{Default}, \texttt{Tuned}, or \texttt{Tuned + Ens.}), based on a Wilcoxon Signed-Rank test for paired samples with Holm-Bonferroni correction and $\alpha=0.05$. Datasets with only a single split do not receive bold/underline annotations because the significance test degenerates without paired samples.
\par\medskip

\noindent \textbf{HPO Pareto Trajectory.} Beside each per-dataset table we plot the metric error ($y$-axis) of each method's tuned configuration against cumulative training time in seconds ($x$-axis), as we increase the number of HPO trials. Each curve traces the Pareto frontier of validation error versus training-time budget for a single method, so points further down and to the left dominate. Reading the plot together with the table makes it possible to compare the cost (compute budget needed to reach a given error) and the achievable error of each method on that dataset, beyond the single endpoint summary in the table.
\par\medskip

\begin{center}
  \begin{minipage}[t]{0.48\textwidth}
    \centering
    \vspace{0pt}
    \scriptsize
\begin{tabular}{l@{\quad}lll}
\toprule
 & Default & Tuned & Tuned + Ens. \\
\midrule
Linear & 5.464 $\pm$ 0.234 & 4.961 $\pm$ 0.239 & 4.886 $\pm$ 0.183 \\
RandomForest & 4.251 $\pm$ 0.092 & 4.024 $\pm$ 0.105 & 4.024 $\pm$ 0.104 \\
ExtraTrees & 4.155 $\pm$ 0.102 & 4.014 $\pm$ 0.104 & 3.991 $\pm$ 0.110 \\
CatBoost & \underline{\textbf{3.904 $\pm$ 0.110}} & \underline{\textbf{3.911 $\pm$ 0.110}} & \underline{\textbf{3.905 $\pm$ 0.110}} \\
LightGBM & 3.950 $\pm$ 0.108 & \underline{3.914 $\pm$ 0.112} & \underline{\textbf{3.908 $\pm$ 0.104}} \\
XGBoost & 3.963 $\pm$ 0.117 & \underline{3.918 $\pm$ 0.117} & \underline{\textbf{3.909 $\pm$ 0.108}} \\
RealMLP & 3.919 $\pm$ 0.104 & \underline{\textbf{3.910 $\pm$ 0.129}} & \underline{\textbf{3.899 $\pm$ 0.109}} \\
TabM & \underline{\textbf{\textcolor{green!50!black}{3.889 $\pm$ 0.102}}} & \underline{\textbf{3.896 $\pm$ 0.114}} & \underline{\textbf{3.892 $\pm$ 0.115}} \\
TabDPT & 4.339 $\pm$ 0.110 & - & - \\
TabPFN-2.6 & 3.977 $\pm$ 0.165 & - & - \\
TabICLv2 & 3.986 $\pm$ 0.149 & - & - \\
\bottomrule
\end{tabular}
\space
  \end{minipage}\hfill
  \begin{minipage}[t]{0.48\textwidth}
    \centering
    \vspace{0pt}
    \includegraphics[width=\linewidth]{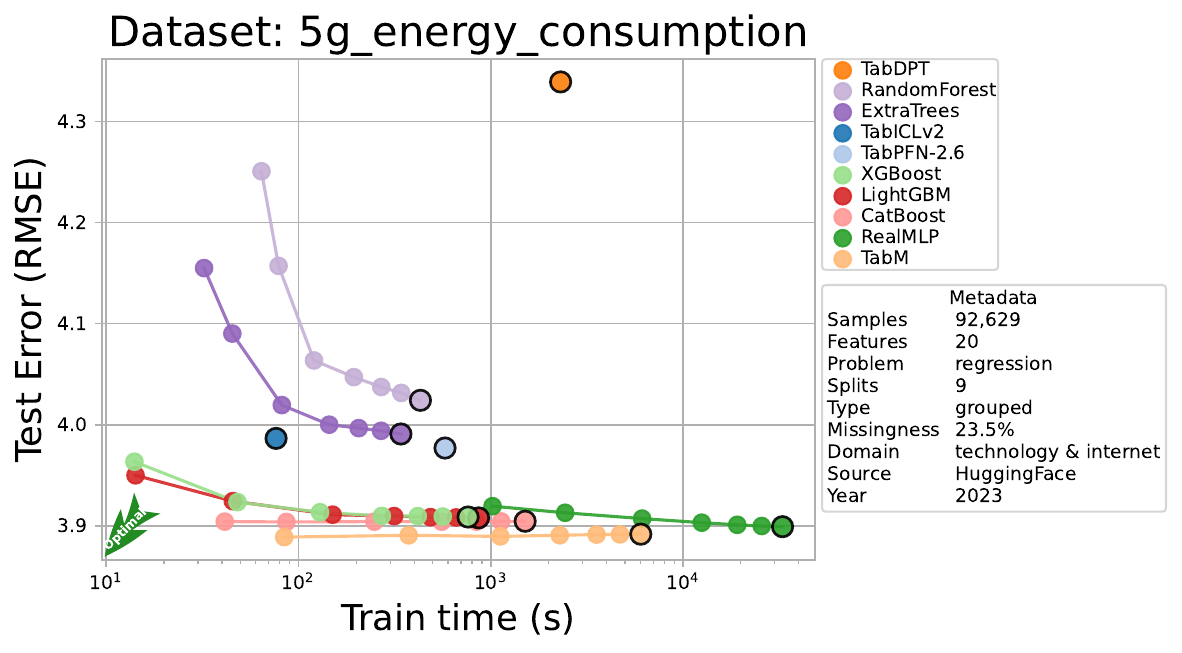}
  \end{minipage}

  \captionof{figure}{\textbf{5g\_energy\_consumption}: per-method test error (left) and HPO Pareto trajectory (right).}
  \label{fig:perdataset_5g_energy_consumption}
\end{center}
\space
\begin{center}
  \begin{minipage}[t]{0.48\textwidth}
    \centering
    \vspace{0pt}
    \scriptsize
\begin{tabular}{l@{\quad}lll}
\toprule
 & Default & Tuned & Tuned + Ens. \\
\midrule
Linear & \underline{\textbf{0.372 $\pm$ 0.038}} & \underline{\textbf{0.376 $\pm$ 0.042}} & \underline{\textbf{0.375 $\pm$ 0.041}} \\
RandomForest & \underline{\textbf{0.365 $\pm$ 0.031}} & \underline{\textbf{0.352 $\pm$ 0.045}} & \underline{\textbf{0.353 $\pm$ 0.047}} \\
ExtraTrees & \underline{\textbf{0.362 $\pm$ 0.035}} & \underline{\textbf{0.357 $\pm$ 0.045}} & \underline{\textbf{0.356 $\pm$ 0.044}} \\
CatBoost & \underline{\textbf{0.355 $\pm$ 0.041}} & \underline{\textbf{0.356 $\pm$ 0.042}} & \underline{\textbf{0.356 $\pm$ 0.043}} \\
LightGBM & \underline{\textbf{0.354 $\pm$ 0.048}} & \underline{\textbf{0.352 $\pm$ 0.045}} & \underline{\textbf{0.352 $\pm$ 0.044}} \\
XGBoost & \underline{\textbf{0.354 $\pm$ 0.048}} & \underline{\textbf{0.353 $\pm$ 0.046}} & \underline{\textbf{0.353 $\pm$ 0.047}} \\
RealMLP & \underline{\textbf{0.353 $\pm$ 0.049}} & \underline{\textbf{0.351 $\pm$ 0.042}} & \underline{\textbf{0.352 $\pm$ 0.045}} \\
TabM & \underline{\textbf{0.353 $\pm$ 0.043}} & \underline{\textbf{0.349 $\pm$ 0.042}} & \underline{\textbf{\textcolor{green!50!black}{0.348 $\pm$ 0.041}}} \\
TabDPT & - & - & - \\
TabPFN-2.6 & - & - & - \\
TabICLv2 & \underline{\textbf{0.368 $\pm$ 0.041}} & - & - \\
\bottomrule
\end{tabular}
\space
  \end{minipage}\hfill
  \begin{minipage}[t]{0.48\textwidth}
    \centering
    \vspace{0pt}
    \includegraphics[width=\linewidth]{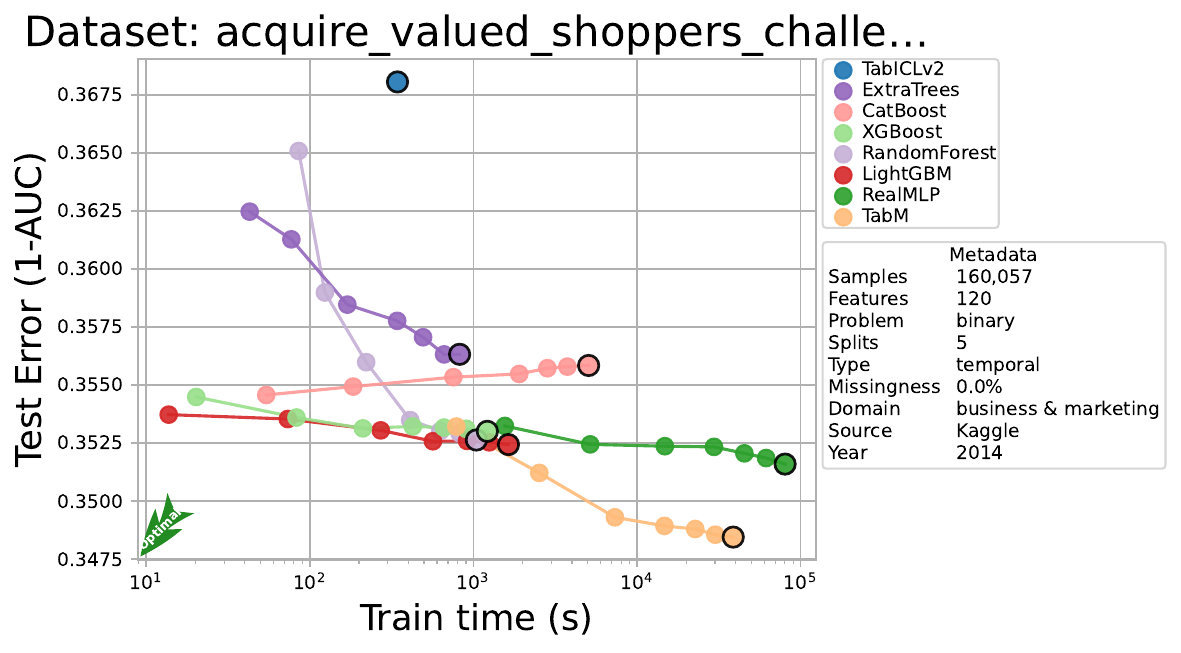}
  \end{minipage}

  \captionof{figure}{\textbf{acquire\_valued\_shoppers\_challenge}: per-method test error (left) and HPO Pareto trajectory (right).}
  \label{fig:perdataset_acquire_valued_shoppers_challenge}
\end{center}
\space
\begin{center}
  \begin{minipage}[t]{0.48\textwidth}
    \centering
    \vspace{0pt}
    \scriptsize
\begin{tabular}{l@{\quad}lll}
\toprule
 & Default & Tuned & Tuned + Ens. \\
\midrule
Linear & 5.337 $\pm$ 0.152 & 4.745 $\pm$ 0.148 & 4.729 $\pm$ 0.135 \\
RandomForest & 1.945 $\pm$ 0.115 & 1.928 $\pm$ 0.113 & 1.874 $\pm$ 0.119 \\
ExtraTrees & 1.898 $\pm$ 0.114 & 1.677 $\pm$ 0.121 & 1.677 $\pm$ 0.123 \\
CatBoost & 1.602 $\pm$ 0.102 & 1.356 $\pm$ 0.122 & 1.361 $\pm$ 0.122 \\
LightGBM & 1.575 $\pm$ 0.130 & 1.530 $\pm$ 0.147 & 1.528 $\pm$ 0.135 \\
XGBoost & 1.574 $\pm$ 0.137 & 1.450 $\pm$ 0.126 & 1.456 $\pm$ 0.127 \\
RealMLP & 1.218 $\pm$ 0.101 & \underline{1.185 $\pm$ 0.113} & \underline{1.170 $\pm$ 0.110} \\
TabM & 1.283 $\pm$ 0.129 & 1.216 $\pm$ 0.107 & 1.195 $\pm$ 0.105 \\
TabDPT & 1.203 $\pm$ 0.092 & - & - \\
TabPFN-2.6 & \underline{\textbf{1.117 $\pm$ 0.096}} & - & - \\
TabICLv2 & \underline{\textbf{\textcolor{green!50!black}{1.112 $\pm$ 0.092}}} & - & - \\
\bottomrule
\end{tabular}
\space
  \end{minipage}\hfill
  \begin{minipage}[t]{0.48\textwidth}
    \centering
    \vspace{0pt}
    \includegraphics[width=\linewidth]{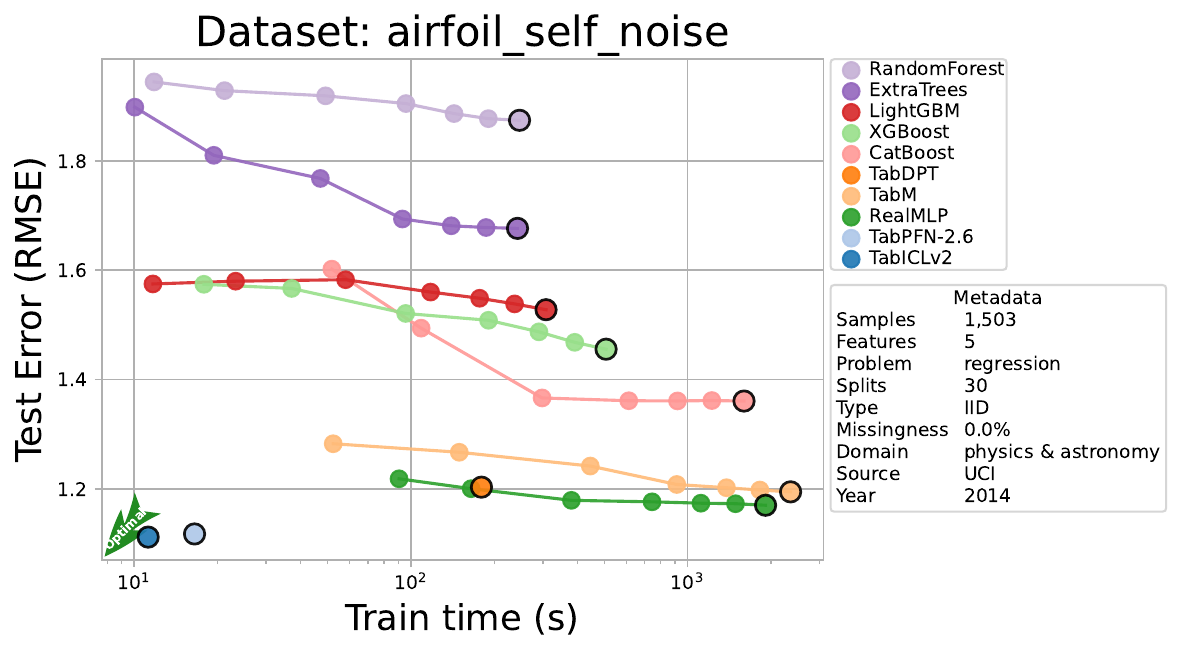}
  \end{minipage}

  \captionof{figure}{\textbf{airfoil\_self\_noise}: per-method test error (left) and HPO Pareto trajectory (right).}
  \label{fig:perdataset_airfoil_self_noise}
\end{center}
\space
\begin{center}
  \begin{minipage}[t]{0.48\textwidth}
    \centering
    \vspace{0pt}
    \scriptsize
\begin{tabular}{l@{\quad}lll}
\toprule
 & Default & Tuned & Tuned + Ens. \\
\midrule
Linear & 0.565 $\pm$ 0.001 & 0.565 $\pm$ 0.001 & 0.564 $\pm$ 0.001 \\
RandomForest & 0.560 $\pm$ 0.001 & 0.558 $\pm$ 0.001 & 0.557 $\pm$ 0.001 \\
ExtraTrees & 0.558 $\pm$ 0.001 & 0.558 $\pm$ 0.001 & 0.557 $\pm$ 0.001 \\
CatBoost & 0.534 $\pm$ 0.001 & 0.534 $\pm$ 0.001 & 0.534 $\pm$ 0.001 \\
LightGBM & 0.540 $\pm$ 0.001 & 0.536 $\pm$ 0.001 & 0.535 $\pm$ 0.001 \\
XGBoost & 0.538 $\pm$ 0.001 & 0.538 $\pm$ 0.001 & 0.536 $\pm$ 0.001 \\
RealMLP & 0.536 $\pm$ 0.001 & 0.533 $\pm$ 0.001 & \underline{\textbf{\textcolor{green!50!black}{0.532 $\pm$ 0.001}}} \\
TabM & \underline{0.533 $\pm$ 0.001} & \underline{0.532 $\pm$ 0.001} & 0.532 $\pm$ 0.001 \\
TabDPT & - & - & - \\
TabPFN-2.6 & - & - & - \\
TabICLv2 & 0.542 $\pm$ 0.001 & - & - \\
\bottomrule
\end{tabular}
\space
  \end{minipage}\hfill
  \begin{minipage}[t]{0.48\textwidth}
    \centering
    \vspace{0pt}
    \includegraphics[width=\linewidth]{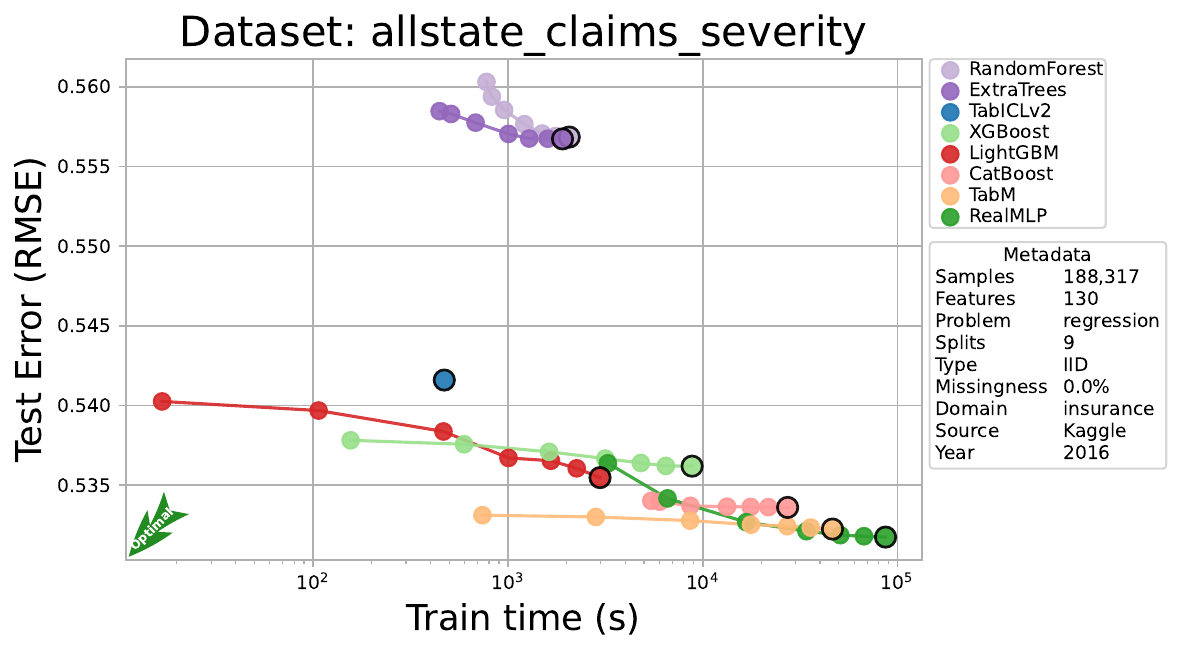}
  \end{minipage}

  \captionof{figure}{\textbf{allstate\_claims\_severity}: per-method test error (left) and HPO Pareto trajectory (right).}
  \label{fig:perdataset_allstate_claims_severity}
\end{center}
\space
\begin{center}
  \begin{minipage}[t]{0.48\textwidth}
    \centering
    \vspace{0pt}
    \scriptsize
\begin{tabular}{l@{\quad}lll}
\toprule
 & Default & Tuned & Tuned + Ens. \\
\midrule
Linear & 0.149 $\pm$ 0.009 & 0.147 $\pm$ 0.009 & 0.146 $\pm$ 0.010 \\
RandomForest & 0.172 $\pm$ 0.006 & 0.172 $\pm$ 0.006 & 0.172 $\pm$ 0.006 \\
ExtraTrees & 0.168 $\pm$ 0.007 & 0.160 $\pm$ 0.007 & 0.159 $\pm$ 0.007 \\
CatBoost & \underline{0.120 $\pm$ 0.006} & \underline{\textbf{\textcolor{green!50!black}{0.118 $\pm$ 0.006}}} & \underline{\textbf{0.118 $\pm$ 0.006}} \\
LightGBM & 0.154 $\pm$ 0.007 & 0.145 $\pm$ 0.009 & 0.141 $\pm$ 0.008 \\
XGBoost & 0.166 $\pm$ 0.008 & 0.152 $\pm$ 0.011 & 0.143 $\pm$ 0.008 \\
RealMLP & 0.153 $\pm$ 0.009 & 0.132 $\pm$ 0.005 & 0.130 $\pm$ 0.007 \\
TabM & 0.163 $\pm$ 0.007 & 0.159 $\pm$ 0.011 & 0.151 $\pm$ 0.008 \\
TabDPT & 0.189 $\pm$ 0.008 & - & - \\
TabPFN-2.6 & 0.152 $\pm$ 0.007 & - & - \\
TabICLv2 & 0.145 $\pm$ 0.008 & - & - \\
\bottomrule
\end{tabular}
\space
  \end{minipage}\hfill
  \begin{minipage}[t]{0.48\textwidth}
    \centering
    \vspace{0pt}
    \includegraphics[width=\linewidth]{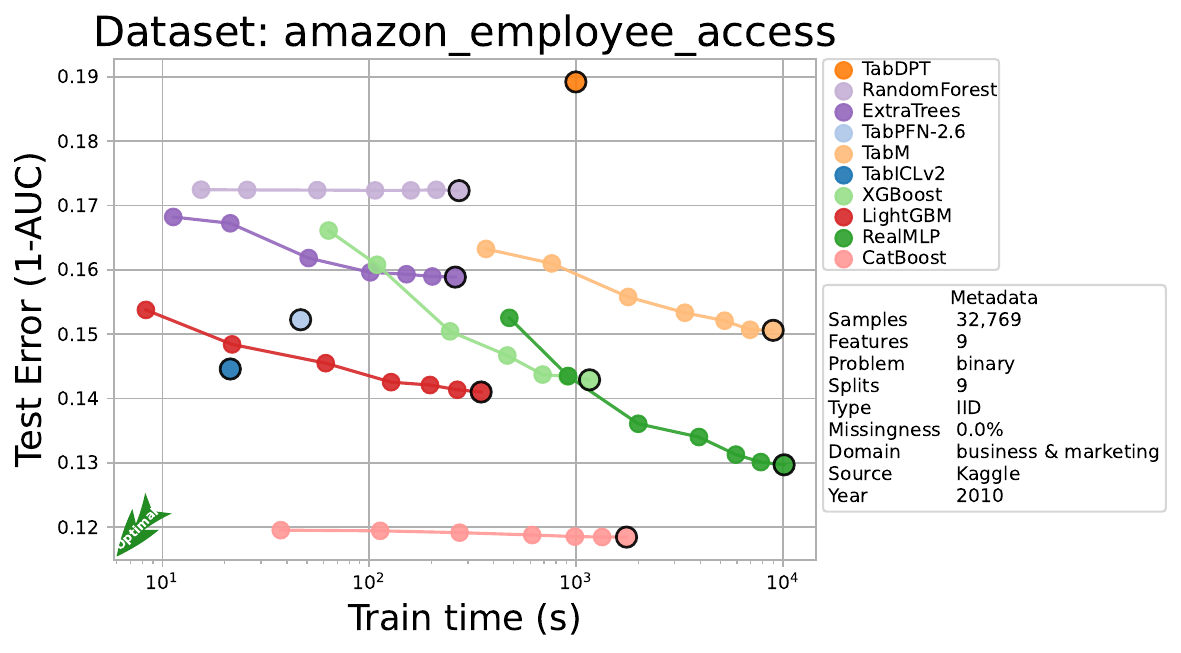}
  \end{minipage}

  \captionof{figure}{\textbf{amazon\_employee\_access}: per-method test error (left) and HPO Pareto trajectory (right).}
  \label{fig:perdataset_amazon_employee_access}
\end{center}
\space
\begin{center}
  \begin{minipage}[t]{0.48\textwidth}
    \centering
    \vspace{0pt}
    \scriptsize
\begin{tabular}{l@{\quad}lll}
\toprule
 & Default & Tuned & Tuned + Ens. \\
\midrule
Linear & 0.061 & 0.061 & 0.058 \\
RandomForest & 0.057 & 0.057 & 0.058 \\
ExtraTrees & 0.060 & 0.061 & 0.060 \\
CatBoost & 0.057 & 0.056 & 0.056 \\
LightGBM & 0.057 & 0.055 & 0.055 \\
XGBoost & 0.056 & 0.055 & 0.055 \\
RealMLP & 0.058 & 0.055 & 0.055 \\
TabM & 0.055 & \textcolor{green!50!black}{0.054} & 0.055 \\
TabDPT & - & - & - \\
TabPFN-2.6 & - & - & - \\
TabICLv2 & 0.063 & - & - \\
\bottomrule
\end{tabular}
\space
  \end{minipage}\hfill
  \begin{minipage}[t]{0.48\textwidth}
    \centering
    \vspace{0pt}
    \includegraphics[width=\linewidth]{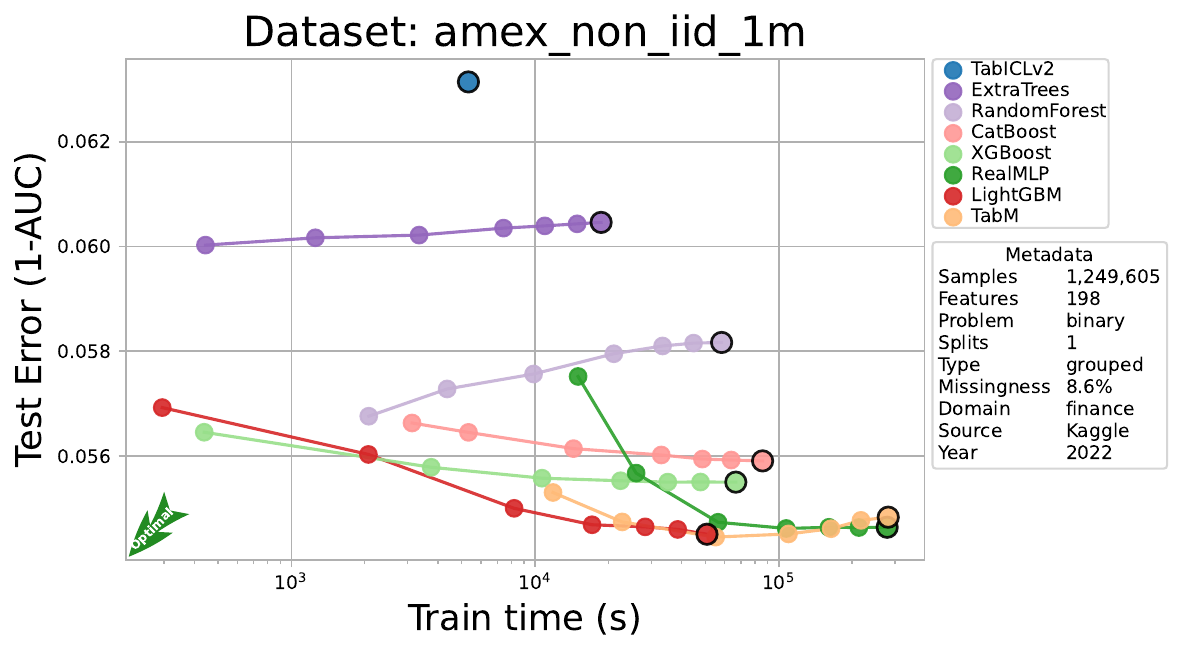}
  \end{minipage}

  \captionof{figure}{\textbf{amex\_non\_iid\_1m}: per-method test error (left) and HPO Pareto trajectory (right).}
  \label{fig:perdataset_amex_non_iid_1m}
\end{center}
\space
\begin{center}
  \begin{minipage}[t]{0.48\textwidth}
    \centering
    \vspace{0pt}
    \scriptsize
\begin{tabular}{l@{\quad}lll}
\toprule
 & Default & Tuned & Tuned + Ens. \\
\midrule
Linear & 0.086 $\pm$ 0.044 & 0.077 $\pm$ 0.043 & 0.077 $\pm$ 0.043 \\
RandomForest & 0.081 $\pm$ 0.035 & 0.085 $\pm$ 0.058 & 0.077 $\pm$ 0.046 \\
ExtraTrees & 0.096 $\pm$ 0.028 & 0.082 $\pm$ 0.053 & 0.078 $\pm$ 0.048 \\
CatBoost & \underline{0.068 $\pm$ 0.041} & \underline{\textbf{0.067 $\pm$ 0.039}} & \underline{\textbf{\textcolor{green!50!black}{0.067 $\pm$ 0.039}}} \\
LightGBM & 0.070 $\pm$ 0.039 & \underline{\textbf{0.068 $\pm$ 0.040}} & \underline{\textbf{0.067 $\pm$ 0.038}} \\
XGBoost & \underline{\textbf{0.067 $\pm$ 0.038}} & \underline{\textbf{0.067 $\pm$ 0.038}} & \underline{\textbf{0.067 $\pm$ 0.038}} \\
RealMLP & \underline{0.069 $\pm$ 0.041} & \underline{\textbf{0.067 $\pm$ 0.040}} & \underline{\textbf{0.067 $\pm$ 0.040}} \\
TabM & \underline{\textbf{0.067 $\pm$ 0.039}} & \underline{\textbf{0.067 $\pm$ 0.040}} & \underline{\textbf{0.067 $\pm$ 0.039}} \\
TabDPT & 0.184 $\pm$ 0.063 & - & - \\
TabPFN-2.6 & \underline{\textbf{0.067 $\pm$ 0.040}} & - & - \\
TabICLv2 & 0.069 $\pm$ 0.040 & - & - \\
\bottomrule
\end{tabular}
\space
  \end{minipage}\hfill
  \begin{minipage}[t]{0.48\textwidth}
    \centering
    \vspace{0pt}
    \includegraphics[width=\linewidth]{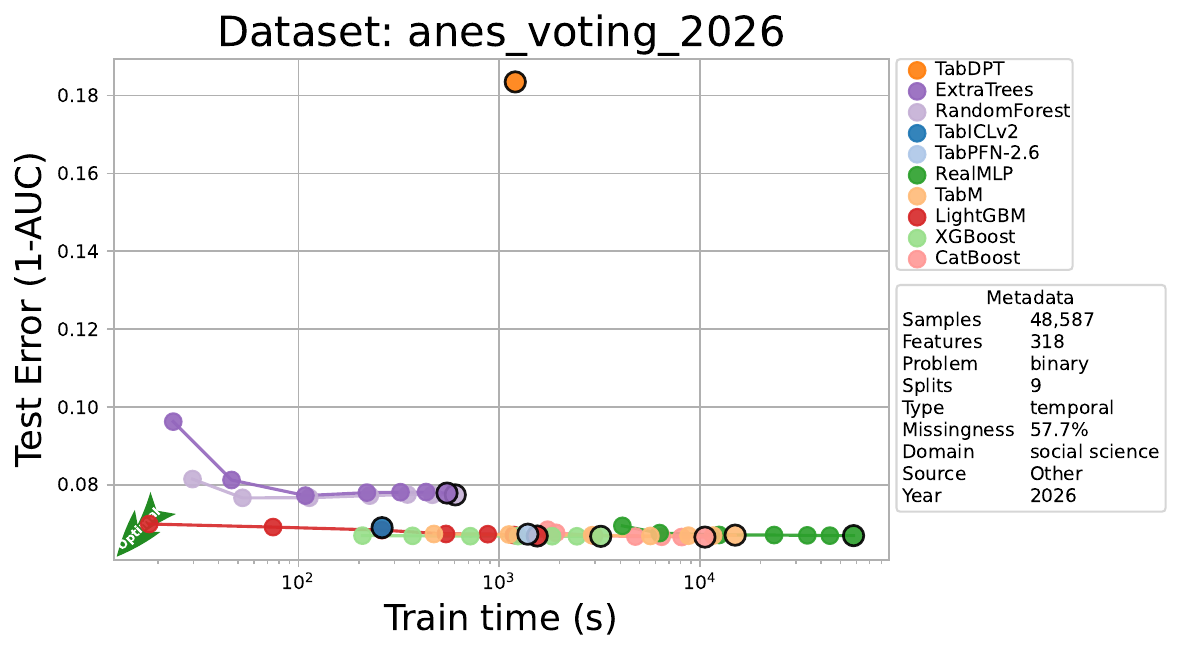}
  \end{minipage}

  \captionof{figure}{\textbf{anes\_voting\_2026}: per-method test error (left) and HPO Pareto trajectory (right).}
  \label{fig:perdataset_anes_voting_2026}
\end{center}
\space
\begin{center}
  \begin{minipage}[t]{0.48\textwidth}
    \centering
    \vspace{0pt}
    \scriptsize
\begin{tabular}{l@{\quad}lll}
\toprule
 & Default & Tuned & Tuned + Ens. \\
\midrule
Linear & 0.012 $\pm$ 0.003 & 0.012 $\pm$ 0.003 & 0.010 $\pm$ 0.002 \\
RandomForest & 0.009 $\pm$ 0.002 & \underline{0.009 $\pm$ 0.002} & \underline{0.009 $\pm$ 0.002} \\
ExtraTrees & 0.010 $\pm$ 0.002 & 0.010 $\pm$ 0.003 & \underline{0.009 $\pm$ 0.002} \\
CatBoost & 0.008 $\pm$ 0.002 & \underline{0.008 $\pm$ 0.002} & \underline{0.008 $\pm$ 0.002} \\
LightGBM & 0.008 $\pm$ 0.002 & \underline{0.008 $\pm$ 0.002} & \underline{0.008 $\pm$ 0.002} \\
XGBoost & 0.008 $\pm$ 0.002 & \underline{0.008 $\pm$ 0.002} & \underline{0.008 $\pm$ 0.002} \\
RealMLP & 0.010 $\pm$ 0.003 & 0.010 $\pm$ 0.002 & \underline{0.008 $\pm$ 0.002} \\
TabM & 0.008 $\pm$ 0.002 & \underline{0.008 $\pm$ 0.002} & \underline{0.008 $\pm$ 0.002} \\
TabDPT & 0.010 $\pm$ 0.003 & - & - \\
TabPFN-2.6 & 0.008 $\pm$ 0.002 & - & - \\
TabICLv2 & \underline{\textbf{\textcolor{green!50!black}{0.006 $\pm$ 0.002}}} & - & - \\
\bottomrule
\end{tabular}
\space
  \end{minipage}\hfill
  \begin{minipage}[t]{0.48\textwidth}
    \centering
    \vspace{0pt}
    \includegraphics[width=\linewidth]{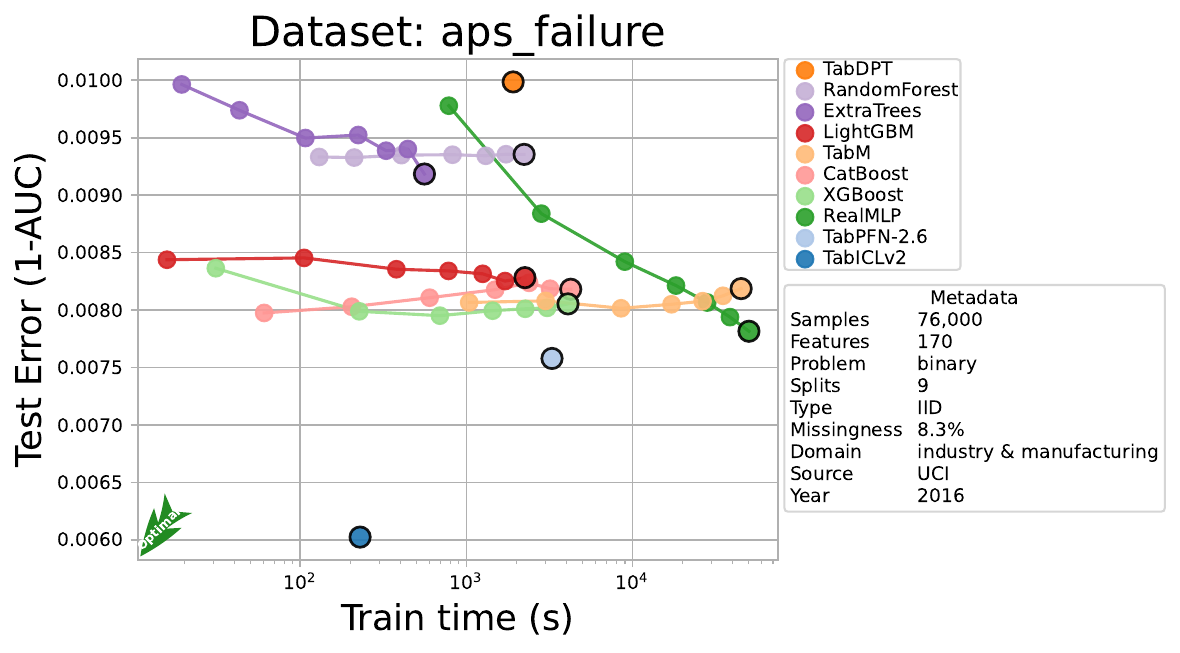}
  \end{minipage}

  \captionof{figure}{\textbf{aps\_failure}: per-method test error (left) and HPO Pareto trajectory (right).}
  \label{fig:perdataset_aps_failure}
\end{center}
\space
\begin{center}
  \begin{minipage}[t]{0.48\textwidth}
    \centering
    \vspace{0pt}
    \scriptsize
\begin{tabular}{l@{\quad}lll}
\toprule
 & Default & Tuned & Tuned + Ens. \\
\midrule
Linear & 3.815 $\pm$ 0.655 & 3.063 $\pm$ 0.444 & 2.805 $\pm$ 0.401 \\
RandomForest & 2.286 $\pm$ 0.149 & 2.287 $\pm$ 0.149 & 2.279 $\pm$ 0.163 \\
ExtraTrees & 2.306 $\pm$ 0.144 & \underline{2.265 $\pm$ 0.146} & 2.259 $\pm$ 0.139 \\
CatBoost & 2.231 $\pm$ 0.113 & \underline{2.245 $\pm$ 0.110} & \underline{2.242 $\pm$ 0.110} \\
LightGBM & 2.289 $\pm$ 0.078 & 2.267 $\pm$ 0.095 & 2.265 $\pm$ 0.091 \\
XGBoost & 2.282 $\pm$ 0.066 & \underline{2.251 $\pm$ 0.079} & \underline{2.249 $\pm$ 0.078} \\
RealMLP & 2.546 $\pm$ 0.317 & \underline{2.257 $\pm$ 0.133} & \underline{2.250 $\pm$ 0.117} \\
TabM & 2.307 $\pm$ 0.183 & \underline{2.251 $\pm$ 0.129} & \underline{2.242 $\pm$ 0.119} \\
TabDPT & 2.295 $\pm$ 0.114 & - & - \\
TabPFN-2.6 & \underline{\textbf{2.187 $\pm$ 0.101}} & - & - \\
TabICLv2 & \underline{\textbf{\textcolor{green!50!black}{2.175 $\pm$ 0.097}}} & - & - \\
\bottomrule
\end{tabular}
\space
  \end{minipage}\hfill
  \begin{minipage}[t]{0.48\textwidth}
    \centering
    \vspace{0pt}
    \includegraphics[width=\linewidth]{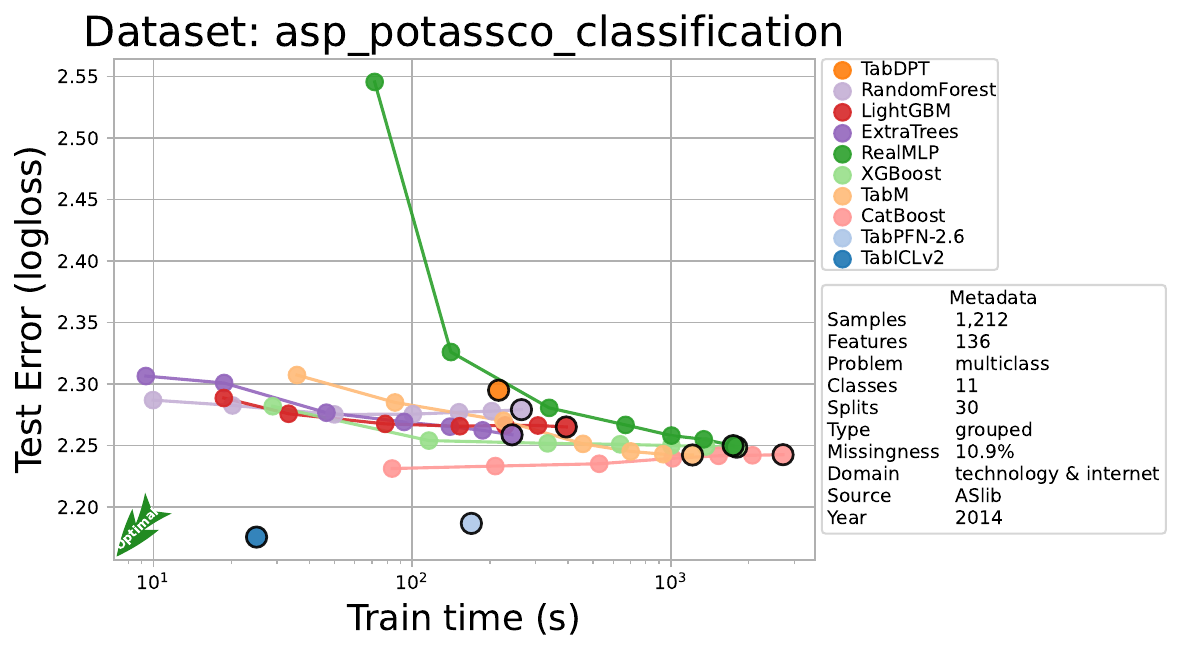}
  \end{minipage}

  \captionof{figure}{\textbf{asp\_potassco\_classification}: per-method test error (left) and HPO Pareto trajectory (right).}
  \label{fig:perdataset_asp_potassco_classification}
\end{center}
\space
\begin{center}
  \begin{minipage}[t]{0.48\textwidth}
    \centering
    \vspace{0pt}
    \scriptsize
\begin{tabular}{l@{\quad}lll}
\toprule
 & Default & Tuned & Tuned + Ens. \\
\midrule
Linear & 0.283 $\pm$ 0.086 & \underline{0.293 $\pm$ 0.092} & \underline{\textbf{0.276 $\pm$ 0.084}} \\
RandomForest & 0.361 $\pm$ 0.054 & 0.381 $\pm$ 0.086 & 0.365 $\pm$ 0.060 \\
ExtraTrees & 0.361 $\pm$ 0.052 & 0.380 $\pm$ 0.113 & 0.361 $\pm$ 0.066 \\
CatBoost & 0.427 $\pm$ 0.059 & 0.338 $\pm$ 0.065 & 0.336 $\pm$ 0.065 \\
LightGBM & 0.433 $\pm$ 0.053 & 0.359 $\pm$ 0.057 & 0.360 $\pm$ 0.056 \\
XGBoost & 0.368 $\pm$ 0.066 & 0.350 $\pm$ 0.057 & 0.349 $\pm$ 0.058 \\
RealMLP & 0.347 $\pm$ 0.052 & 0.310 $\pm$ 0.072 & 0.302 $\pm$ 0.066 \\
TabM & 0.342 $\pm$ 0.067 & 0.349 $\pm$ 0.070 & 0.347 $\pm$ 0.069 \\
TabDPT & 0.402 $\pm$ 0.053 & - & - \\
TabPFN-2.6 & 0.309 $\pm$ 0.070 & - & - \\
TabICLv2 & \underline{\textbf{\textcolor{green!50!black}{0.270 $\pm$ 0.093}}} & - & - \\
\bottomrule
\end{tabular}
\space
  \end{minipage}\hfill
  \begin{minipage}[t]{0.48\textwidth}
    \centering
    \vspace{0pt}
    \includegraphics[width=\linewidth]{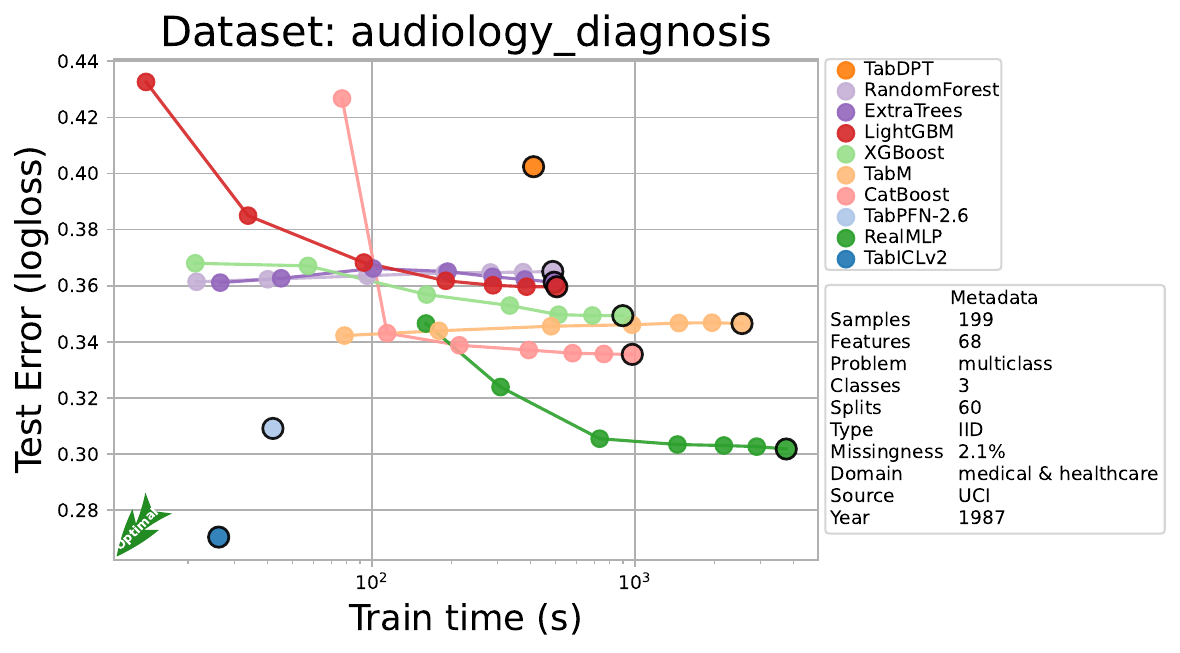}
  \end{minipage}

  \captionof{figure}{\textbf{audiology\_diagnosis}: per-method test error (left) and HPO Pareto trajectory (right).}
  \label{fig:perdataset_audiology_diagnosis}
\end{center}
\space
\begin{center}
  \begin{minipage}[t]{0.48\textwidth}
    \centering
    \vspace{0pt}
    \scriptsize
\begin{tabular}{l@{\quad}lll}
\toprule
 & Default & Tuned & Tuned + Ens. \\
\midrule
Linear & 0.268 $\pm$ 0.017 & 0.269 $\pm$ 0.017 & 0.266 $\pm$ 0.018 \\
RandomForest & 0.269 $\pm$ 0.022 & 0.273 $\pm$ 0.026 & 0.269 $\pm$ 0.025 \\
ExtraTrees & 0.296 $\pm$ 0.022 & 0.279 $\pm$ 0.024 & 0.277 $\pm$ 0.025 \\
CatBoost & \underline{\textbf{0.252 $\pm$ 0.020}} & \underline{0.253 $\pm$ 0.022} & \underline{\textbf{\textcolor{green!50!black}{0.249 $\pm$ 0.020}}} \\
LightGBM & 0.273 $\pm$ 0.023 & 0.264 $\pm$ 0.021 & 0.255 $\pm$ 0.019 \\
XGBoost & 0.269 $\pm$ 0.026 & \underline{\textbf{0.253 $\pm$ 0.021}} & \underline{\textbf{0.252 $\pm$ 0.021}} \\
RealMLP & 0.269 $\pm$ 0.025 & 0.264 $\pm$ 0.023 & 0.257 $\pm$ 0.020 \\
TabM & 0.261 $\pm$ 0.021 & 0.261 $\pm$ 0.023 & 0.261 $\pm$ 0.022 \\
TabDPT & \textbf{0.254 $\pm$ 0.021} & - & - \\
TabPFN-2.6 & \underline{\textbf{0.250 $\pm$ 0.021}} & - & - \\
TabICLv2 & \underline{\textbf{0.252 $\pm$ 0.020}} & - & - \\
\bottomrule
\end{tabular}
\space
  \end{minipage}\hfill
  \begin{minipage}[t]{0.48\textwidth}
    \centering
    \vspace{0pt}
    \includegraphics[width=\linewidth]{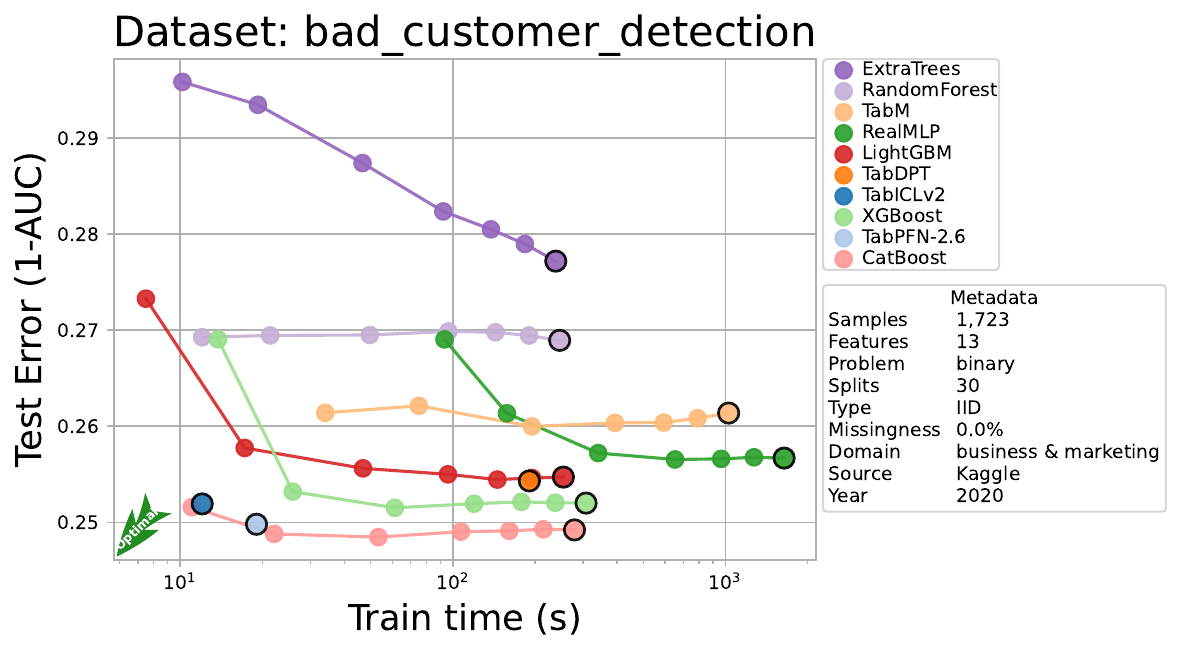}
  \end{minipage}

  \captionof{figure}{\textbf{bad\_customer\_detection}: per-method test error (left) and HPO Pareto trajectory (right).}
  \label{fig:perdataset_bad_customer_detection}
\end{center}
\space
\begin{center}
  \begin{minipage}[t]{0.48\textwidth}
    \centering
    \vspace{0pt}
    \scriptsize
\begin{tabular}{l@{\quad}lll}
\toprule
 & Default & Tuned & Tuned + Ens. \\
\midrule
Linear & 0.229 $\pm$ 0.005 & 0.228 $\pm$ 0.005 & 0.228 $\pm$ 0.005 \\
RandomForest & 0.147 $\pm$ 0.007 & 0.143 $\pm$ 0.007 & 0.142 $\pm$ 0.007 \\
ExtraTrees & 0.146 $\pm$ 0.007 & 0.142 $\pm$ 0.006 & 0.140 $\pm$ 0.007 \\
CatBoost & \underline{\textbf{0.130 $\pm$ 0.006}} & \underline{\textbf{0.129 $\pm$ 0.005}} & \underline{\textbf{0.129 $\pm$ 0.005}} \\
LightGBM & 0.135 $\pm$ 0.007 & 0.133 $\pm$ 0.006 & 0.132 $\pm$ 0.006 \\
XGBoost & 0.136 $\pm$ 0.007 & 0.135 $\pm$ 0.006 & 0.134 $\pm$ 0.006 \\
RealMLP & 0.134 $\pm$ 0.006 & 0.131 $\pm$ 0.005 & \underline{0.130 $\pm$ 0.006} \\
TabM & \underline{\textbf{0.130 $\pm$ 0.007}} & \underline{\textbf{0.130 $\pm$ 0.007}} & \underline{\textbf{0.129 $\pm$ 0.006}} \\
TabDPT & 0.134 $\pm$ 0.007 & - & - \\
TabPFN-2.6 & \underline{\textbf{0.129 $\pm$ 0.006}} & - & - \\
TabICLv2 & \underline{\textbf{\textcolor{green!50!black}{0.129 $\pm$ 0.006}}} & - & - \\
\bottomrule
\end{tabular}
\space
  \end{minipage}\hfill
  \begin{minipage}[t]{0.48\textwidth}
    \centering
    \vspace{0pt}
    \includegraphics[width=\linewidth]{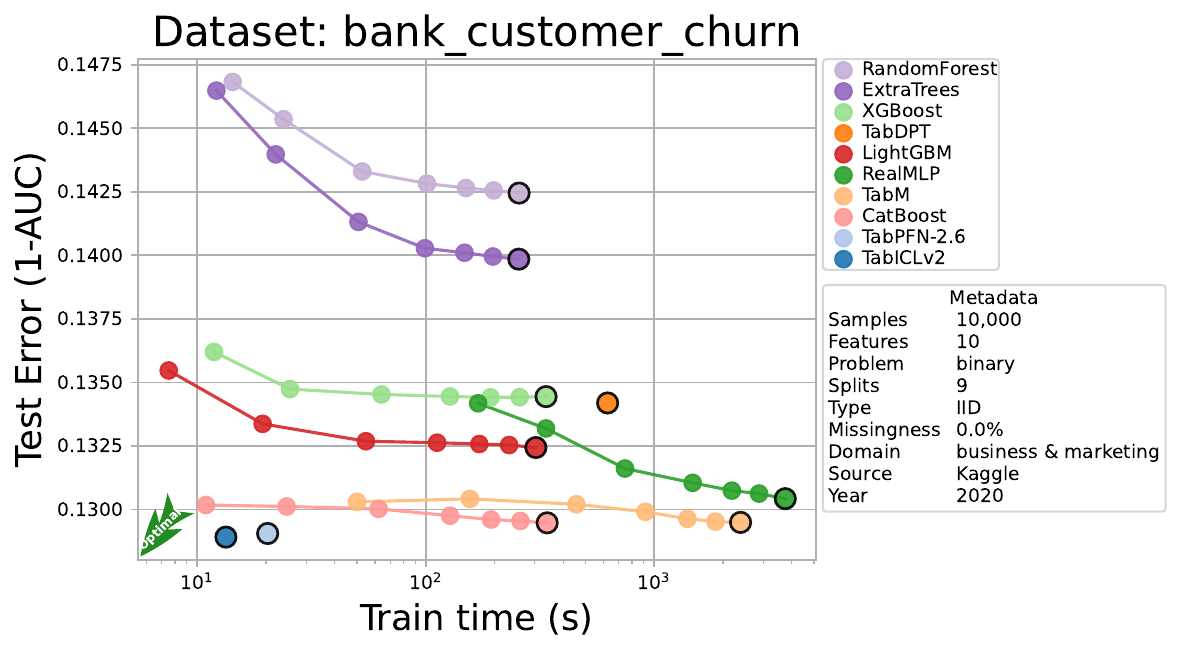}
  \end{minipage}

  \captionof{figure}{\textbf{bank\_customer\_churn}: per-method test error (left) and HPO Pareto trajectory (right).}
  \label{fig:perdataset_bank_customer_churn}
\end{center}
\space
\begin{center}
  \begin{minipage}[t]{0.48\textwidth}
    \centering
    \vspace{0pt}
    \scriptsize
\begin{tabular}{l@{\quad}lll}
\toprule
 & Default & Tuned & Tuned + Ens. \\
\midrule
Linear & 0.260 $\pm$ 0.008 & 0.259 $\pm$ 0.008 & 0.259 $\pm$ 0.008 \\
RandomForest & 0.270 $\pm$ 0.006 & 0.241 $\pm$ 0.006 & 0.240 $\pm$ 0.005 \\
ExtraTrees & 0.275 $\pm$ 0.006 & 0.244 $\pm$ 0.006 & 0.243 $\pm$ 0.006 \\
CatBoost & \underline{\textbf{0.235 $\pm$ 0.006}} & \underline{\textbf{0.235 $\pm$ 0.006}} & \underline{\textbf{\textcolor{green!50!black}{0.235 $\pm$ 0.006}}} \\
LightGBM & 0.237 $\pm$ 0.006 & 0.237 $\pm$ 0.006 & \underline{\textbf{0.236 $\pm$ 0.006}} \\
XGBoost & 0.238 $\pm$ 0.007 & \underline{\textbf{0.236 $\pm$ 0.006}} & \underline{\textbf{0.235 $\pm$ 0.006}} \\
RealMLP & 0.240 $\pm$ 0.007 & 0.237 $\pm$ 0.006 & \underline{\textbf{0.236 $\pm$ 0.006}} \\
TabM & 0.237 $\pm$ 0.006 & 0.237 $\pm$ 0.006 & 0.236 $\pm$ 0.006 \\
TabDPT & 0.238 $\pm$ 0.006 & - & - \\
TabPFN-2.6 & 0.236 $\pm$ 0.006 & - & - \\
TabICLv2 & 0.239 $\pm$ 0.007 & - & - \\
\bottomrule
\end{tabular}
\space
  \end{minipage}\hfill
  \begin{minipage}[t]{0.48\textwidth}
    \centering
    \vspace{0pt}
    \includegraphics[width=\linewidth]{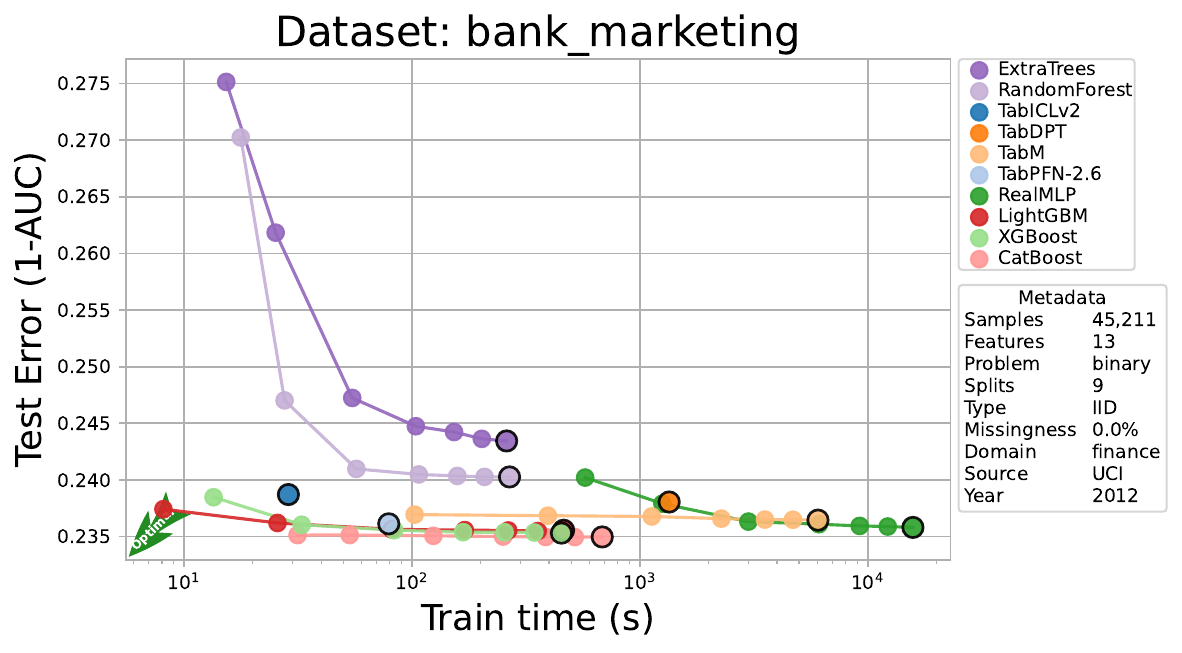}
  \end{minipage}

  \captionof{figure}{\textbf{bank\_marketing}: per-method test error (left) and HPO Pareto trajectory (right).}
  \label{fig:perdataset_bank_marketing}
\end{center}
\space
\begin{center}
  \begin{minipage}[t]{0.48\textwidth}
    \centering
    \vspace{0pt}
    \scriptsize
\begin{tabular}{l@{\quad}lll}
\toprule
 & Default & Tuned & Tuned + Ens. \\
\midrule
Linear & 1.842 $\pm$ 0.075 & 1.647 $\pm$ 0.048 & 1.647 $\pm$ 0.048 \\
RandomForest & 1.670 $\pm$ 0.013 & 1.573 $\pm$ 0.037 & 1.563 $\pm$ 0.032 \\
ExtraTrees & 1.684 $\pm$ 0.011 & 1.535 $\pm$ 0.036 & 1.535 $\pm$ 0.034 \\
CatBoost & 1.556 $\pm$ 0.028 & 1.527 $\pm$ 0.033 & 1.529 $\pm$ 0.032 \\
LightGBM & 1.694 $\pm$ 0.034 & 1.555 $\pm$ 0.034 & 1.552 $\pm$ 0.033 \\
XGBoost & 1.634 $\pm$ 0.040 & 1.571 $\pm$ 0.033 & 1.572 $\pm$ 0.034 \\
RealMLP & 1.576 $\pm$ 0.039 & 1.427 $\pm$ 0.037 & \underline{1.420 $\pm$ 0.032} \\
TabM & 1.427 $\pm$ 0.048 & \underline{1.418 $\pm$ 0.046} & \underline{1.417 $\pm$ 0.045} \\
TabDPT & 1.511 $\pm$ 0.028 & - & - \\
TabPFN-2.6 & \underline{\textbf{1.373 $\pm$ 0.045}} & - & - \\
TabICLv2 & \underline{\textbf{\textcolor{green!50!black}{1.369 $\pm$ 0.050}}} & - & - \\
\bottomrule
\end{tabular}
\space
  \end{minipage}\hfill
  \begin{minipage}[t]{0.48\textwidth}
    \centering
    \vspace{0pt}
    \includegraphics[width=\linewidth]{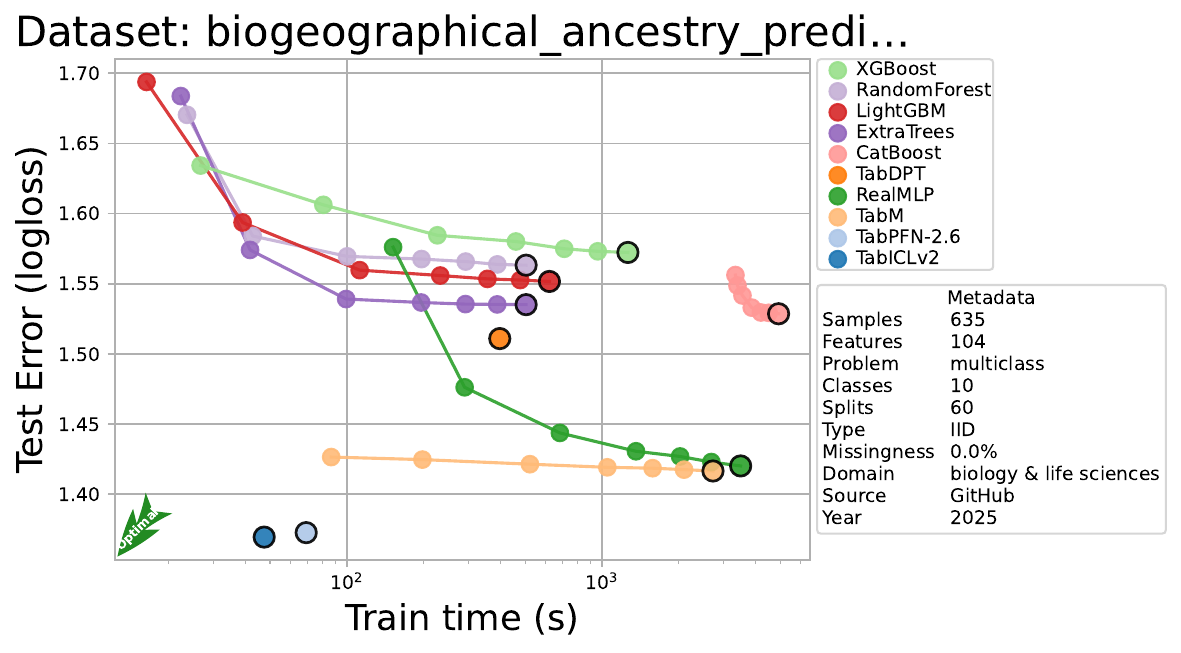}
  \end{minipage}

  \captionof{figure}{\textbf{biogeographical\_ancestry\_prediction}: per-method test error (left) and HPO Pareto trajectory (right).}
  \label{fig:perdataset_biogeographical_ancestry_prediction}
\end{center}
\space
\begin{center}
  \begin{minipage}[t]{0.48\textwidth}
    \centering
    \vspace{0pt}
    \scriptsize
\begin{tabular}{l@{\quad}lll}
\toprule
 & Default & Tuned & Tuned + Ens. \\
\midrule
Linear & 0.373 $\pm$ 0.044 & \underline{0.341 $\pm$ 0.057} & 0.340 $\pm$ 0.053 \\
RandomForest & 0.370 $\pm$ 0.029 & 0.368 $\pm$ 0.037 & 0.367 $\pm$ 0.037 \\
ExtraTrees & 0.411 $\pm$ 0.023 & 0.364 $\pm$ 0.030 & 0.364 $\pm$ 0.030 \\
CatBoost & 0.367 $\pm$ 0.043 & 0.349 $\pm$ 0.044 & 0.349 $\pm$ 0.044 \\
LightGBM & 0.377 $\pm$ 0.044 & 0.367 $\pm$ 0.044 & 0.364 $\pm$ 0.041 \\
XGBoost & 0.403 $\pm$ 0.050 & 0.376 $\pm$ 0.042 & 0.377 $\pm$ 0.041 \\
RealMLP & 0.355 $\pm$ 0.080 & \underline{\textbf{0.336 $\pm$ 0.065}} & \underline{\textbf{\textcolor{green!50!black}{0.332 $\pm$ 0.054}}} \\
TabM & 0.348 $\pm$ 0.036 & 0.341 $\pm$ 0.049 & 0.341 $\pm$ 0.048 \\
TabDPT & 0.340 $\pm$ 0.054 & - & - \\
TabPFN-2.6 & 0.341 $\pm$ 0.056 & - & - \\
TabICLv2 & \underline{\textbf{0.332 $\pm$ 0.059}} & - & - \\
\bottomrule
\end{tabular}
\space
  \end{minipage}\hfill
  \begin{minipage}[t]{0.48\textwidth}
    \centering
    \vspace{0pt}
    \includegraphics[width=\linewidth]{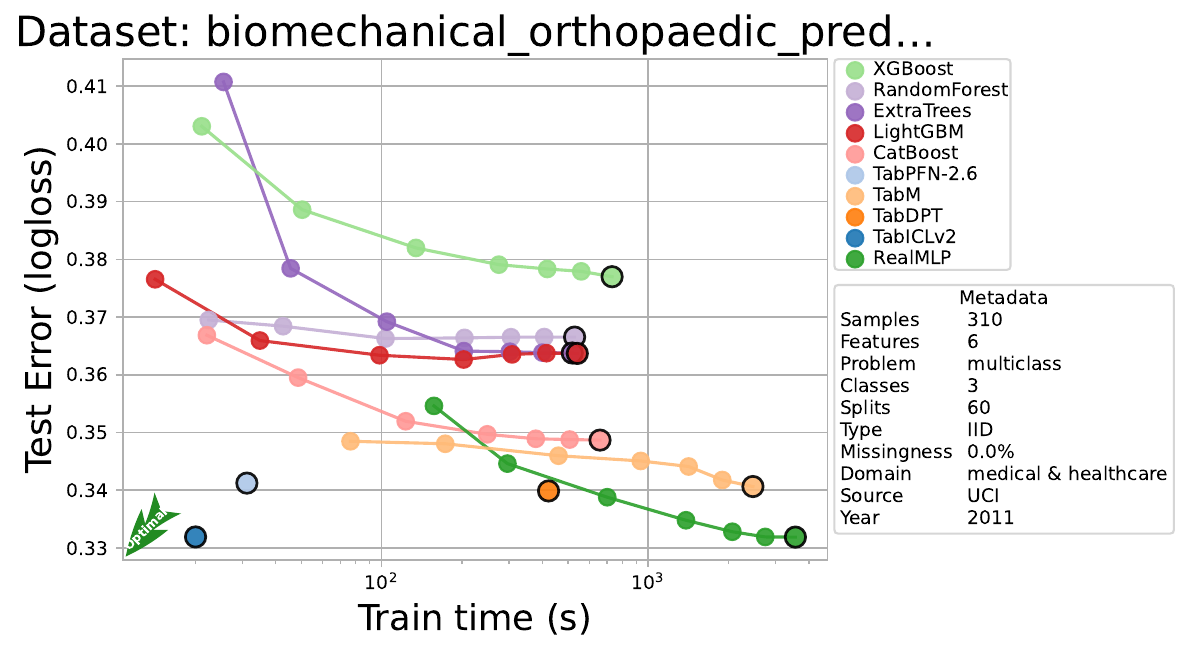}
  \end{minipage}

  \captionof{figure}{\textbf{biomechanical\_orthopaedic\_prediction}: per-method test error (left) and HPO Pareto trajectory (right).}
  \label{fig:perdataset_biomechanical_orthopaedic_prediction}
\end{center}
\space
\begin{center}
  \begin{minipage}[t]{0.48\textwidth}
    \centering
    \vspace{0pt}
    \scriptsize
\begin{tabular}{l@{\quad}lll}
\toprule
 & Default & Tuned & Tuned + Ens. \\
\midrule
Linear & 0.211 $\pm$ 0.012 & 0.195 $\pm$ 0.011 & 0.186 $\pm$ 0.011 \\
RandomForest & 0.130 $\pm$ 0.012 & \underline{0.130 $\pm$ 0.012} & \underline{0.129 $\pm$ 0.012} \\
ExtraTrees & 0.133 $\pm$ 0.012 & 0.135 $\pm$ 0.012 & \underline{0.132 $\pm$ 0.011} \\
CatBoost & 0.130 $\pm$ 0.009 & \underline{\textbf{0.128 $\pm$ 0.009}} & \underline{\textbf{0.128 $\pm$ 0.009}} \\
LightGBM & 0.132 $\pm$ 0.010 & 0.130 $\pm$ 0.011 & \underline{0.129 $\pm$ 0.011} \\
XGBoost & 0.130 $\pm$ 0.009 & \underline{\textbf{0.127 $\pm$ 0.011}} & \underline{\textbf{0.127 $\pm$ 0.011}} \\
RealMLP & 0.142 $\pm$ 0.011 & 0.133 $\pm$ 0.009 & \underline{\textbf{0.126 $\pm$ 0.010}} \\
TabM & 0.137 $\pm$ 0.010 & \underline{\textbf{0.127 $\pm$ 0.010}} & \underline{\textbf{0.127 $\pm$ 0.011}} \\
TabDPT & 0.135 $\pm$ 0.011 & - & - \\
TabPFN-2.6 & \underline{\textbf{\textcolor{green!50!black}{0.125 $\pm$ 0.010}}} & - & - \\
TabICLv2 & 0.141 $\pm$ 0.011 & - & - \\
\bottomrule
\end{tabular}
\space
  \end{minipage}\hfill
  \begin{minipage}[t]{0.48\textwidth}
    \centering
    \vspace{0pt}
    \includegraphics[width=\linewidth]{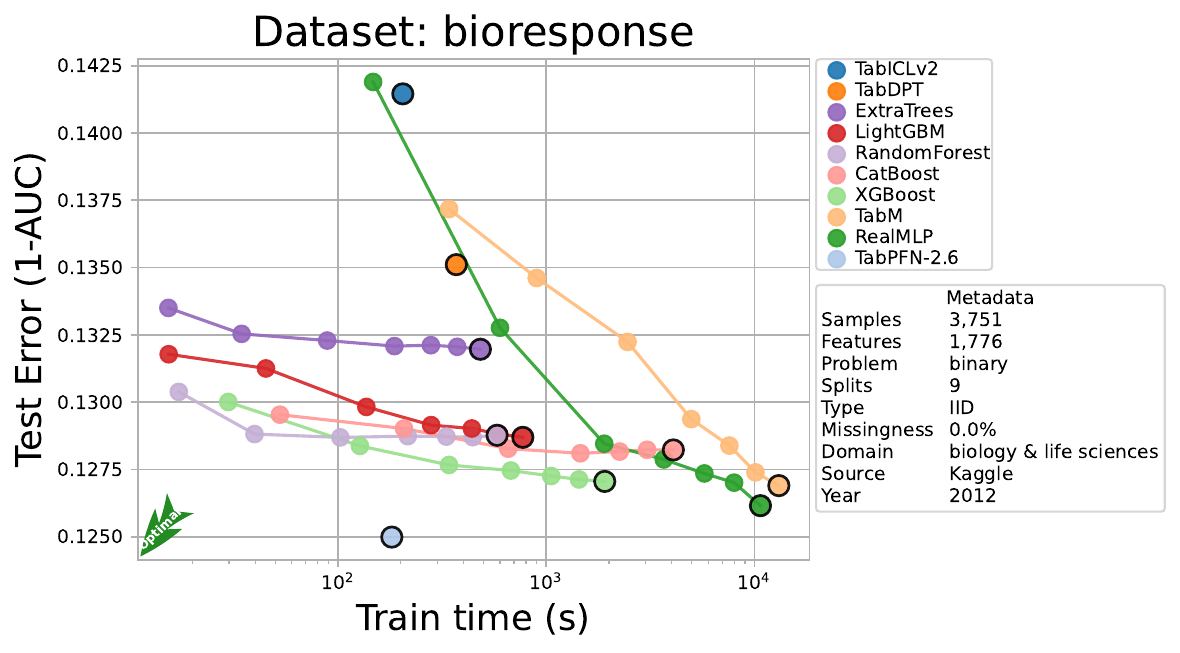}
  \end{minipage}

  \captionof{figure}{\textbf{bioresponse}: per-method test error (left) and HPO Pareto trajectory (right).}
  \label{fig:perdataset_bioresponse}
\end{center}
\space
\begin{center}
  \begin{minipage}[t]{0.48\textwidth}
    \centering
    \vspace{0pt}
    \scriptsize
\begin{tabular}{l@{\quad}lll}
\toprule
 & Default & Tuned & Tuned + Ens. \\
\midrule
Linear & 0.681 $\pm$ 0.030 & \underline{0.683 $\pm$ 0.033} & \underline{0.683 $\pm$ 0.030} \\
RandomForest & 0.699 $\pm$ 0.032 & 0.685 $\pm$ 0.031 & 0.686 $\pm$ 0.031 \\
ExtraTrees & 0.688 $\pm$ 0.029 & \underline{0.683 $\pm$ 0.030} & 0.683 $\pm$ 0.029 \\
CatBoost & 0.684 $\pm$ 0.029 & 0.686 $\pm$ 0.028 & 0.685 $\pm$ 0.029 \\
LightGBM & 0.689 $\pm$ 0.029 & 0.686 $\pm$ 0.030 & 0.686 $\pm$ 0.030 \\
XGBoost & 0.701 $\pm$ 0.029 & 0.687 $\pm$ 0.030 & 0.687 $\pm$ 0.030 \\
RealMLP & 0.681 $\pm$ 0.031 & \underline{0.681 $\pm$ 0.030} & \underline{0.681 $\pm$ 0.030} \\
TabM & 0.685 $\pm$ 0.030 & \underline{0.684 $\pm$ 0.031} & 0.684 $\pm$ 0.031 \\
TabDPT & \underline{\textbf{\textcolor{green!50!black}{0.673 $\pm$ 0.031}}} & - & - \\
TabPFN-2.6 & 0.678 $\pm$ 0.031 & - & - \\
TabICLv2 & 0.678 $\pm$ 0.031 & - & - \\
\bottomrule
\end{tabular}
\space
  \end{minipage}\hfill
  \begin{minipage}[t]{0.48\textwidth}
    \centering
    \vspace{0pt}
    \includegraphics[width=\linewidth]{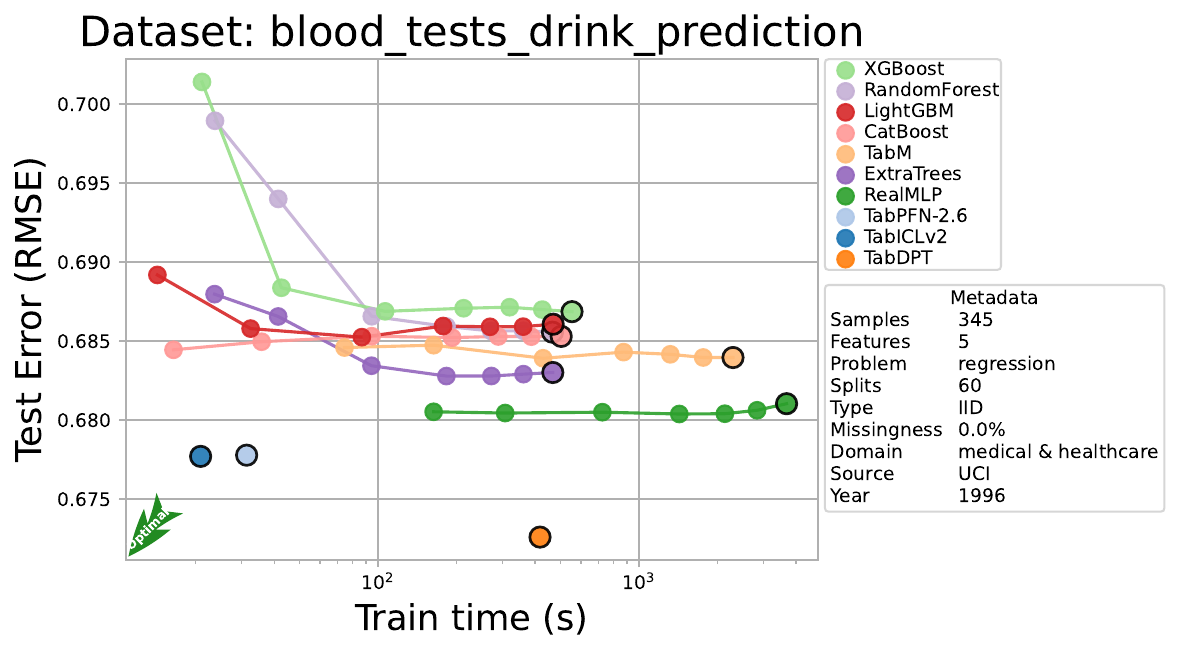}
  \end{minipage}

  \captionof{figure}{\textbf{blood\_tests\_drink\_prediction}: per-method test error (left) and HPO Pareto trajectory (right).}
  \label{fig:perdataset_blood_tests_drink_prediction}
\end{center}
\space
\begin{center}
  \begin{minipage}[t]{0.48\textwidth}
    \centering
    \vspace{0pt}
    \scriptsize
\begin{tabular}{l@{\quad}lll}
\toprule
 & Default & Tuned & Tuned + Ens. \\
\midrule
Linear & 0.269 $\pm$ 0.032 & \underline{0.252 $\pm$ 0.034} & \underline{\textbf{0.247 $\pm$ 0.032}} \\
RandomForest & 0.307 $\pm$ 0.029 & 0.289 $\pm$ 0.032 & 0.289 $\pm$ 0.032 \\
ExtraTrees & 0.304 $\pm$ 0.028 & 0.271 $\pm$ 0.032 & 0.272 $\pm$ 0.032 \\
CatBoost & 0.257 $\pm$ 0.032 & 0.256 $\pm$ 0.033 & 0.257 $\pm$ 0.032 \\
LightGBM & 0.269 $\pm$ 0.035 & \underline{0.251 $\pm$ 0.033} & 0.252 $\pm$ 0.034 \\
XGBoost & 0.285 $\pm$ 0.035 & 0.264 $\pm$ 0.034 & 0.264 $\pm$ 0.034 \\
RealMLP & \underline{\textbf{0.248 $\pm$ 0.032}} & \underline{0.250 $\pm$ 0.030} & \underline{0.249 $\pm$ 0.031} \\
TabM & 0.259 $\pm$ 0.031 & 0.262 $\pm$ 0.033 & 0.260 $\pm$ 0.032 \\
TabDPT & 0.251 $\pm$ 0.032 & - & - \\
TabPFN-2.6 & 0.251 $\pm$ 0.032 & - & - \\
TabICLv2 & \underline{\textbf{\textcolor{green!50!black}{0.247 $\pm$ 0.033}}} & - & - \\
\bottomrule
\end{tabular}
\space
  \end{minipage}\hfill
  \begin{minipage}[t]{0.48\textwidth}
    \centering
    \vspace{0pt}
    \includegraphics[width=\linewidth]{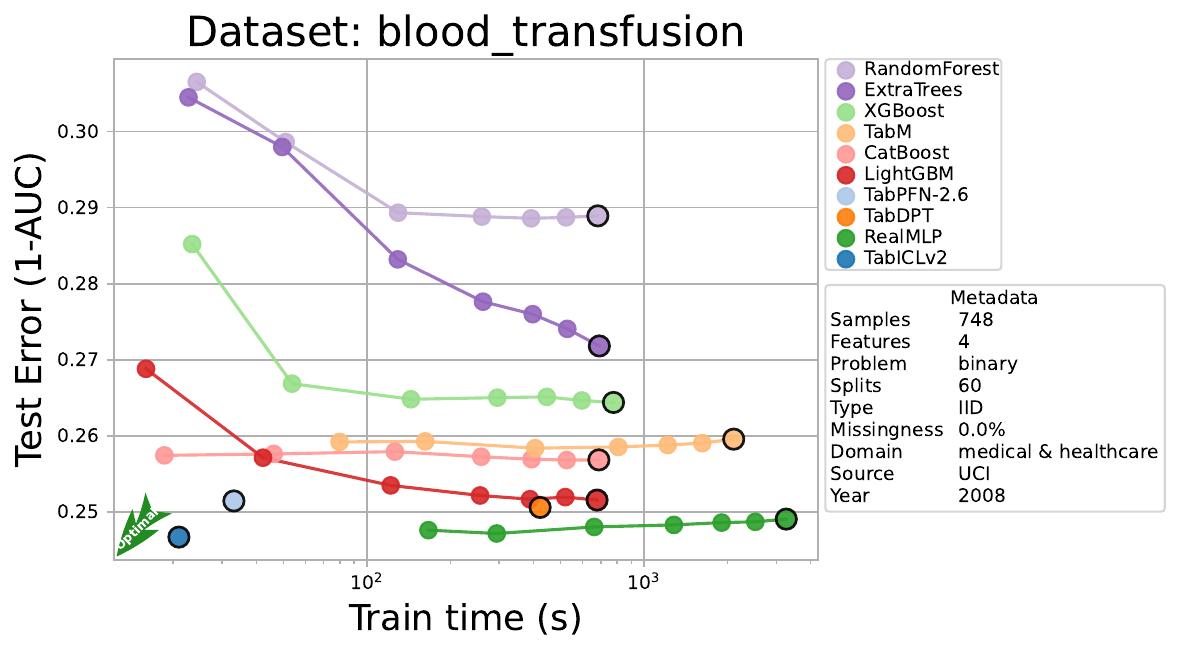}
  \end{minipage}

  \captionof{figure}{\textbf{blood\_transfusion}: per-method test error (left) and HPO Pareto trajectory (right).}
  \label{fig:perdataset_blood_transfusion}
\end{center}
\space
\begin{center}
  \begin{minipage}[t]{0.48\textwidth}
    \centering
    \vspace{0pt}
    \scriptsize
\begin{tabular}{l@{\quad}lll}
\toprule
 & Default & Tuned & Tuned + Ens. \\
\midrule
Linear & 0.011 $\pm$ 0.001 & 0.011 $\pm$ 0.001 & \underline{0.011 $\pm$ 0.001} \\
RandomForest & 0.011 $\pm$ 0.001 & 0.011 $\pm$ 0.001 & 0.011 $\pm$ 0.001 \\
ExtraTrees & 0.011 $\pm$ 0.001 & 0.011 $\pm$ 0.001 & 0.011 $\pm$ 0.001 \\
CatBoost & 0.011 $\pm$ 0.001 & 0.011 $\pm$ 0.001 & 0.011 $\pm$ 0.001 \\
LightGBM & 0.011 $\pm$ 0.001 & 0.011 $\pm$ 0.001 & 0.011 $\pm$ 0.001 \\
XGBoost & 0.012 $\pm$ 0.001 & 0.012 $\pm$ 0.001 & 0.012 $\pm$ 0.001 \\
RealMLP & 0.010 $\pm$ 0.001 & \underline{0.010 $\pm$ 0.001} & \underline{0.010 $\pm$ 0.001} \\
TabM & 0.011 $\pm$ 0.001 & 0.011 $\pm$ 0.001 & 0.011 $\pm$ 0.001 \\
TabDPT & \underline{\textbf{\textcolor{green!50!black}{0.010 $\pm$ 0.001}}} & - & - \\
TabPFN-2.6 & 0.010 $\pm$ 0.001 & - & - \\
TabICLv2 & \underline{\textbf{0.010 $\pm$ 0.001}} & - & - \\
\bottomrule
\end{tabular}
\space
  \end{minipage}\hfill
  \begin{minipage}[t]{0.48\textwidth}
    \centering
    \vspace{0pt}
    \includegraphics[width=\linewidth]{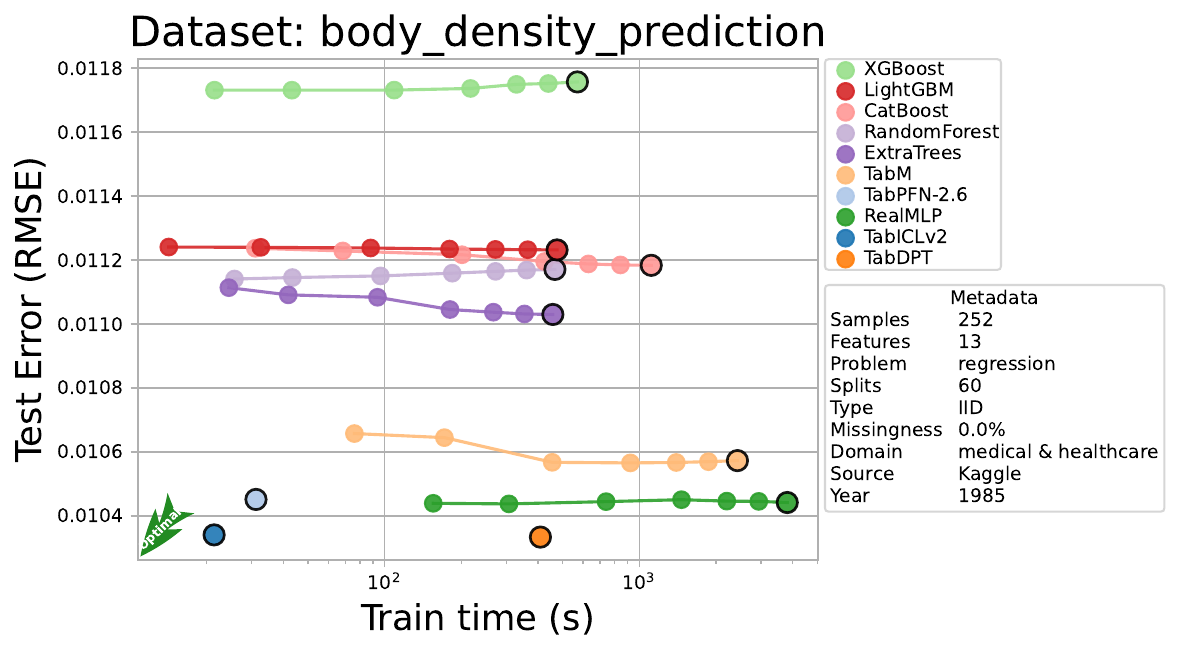}
  \end{minipage}

  \captionof{figure}{\textbf{body\_density\_prediction}: per-method test error (left) and HPO Pareto trajectory (right).}
  \label{fig:perdataset_body_density_prediction}
\end{center}
\space
\begin{center}
  \begin{minipage}[t]{0.48\textwidth}
    \centering
    \vspace{0pt}
    \scriptsize
\begin{tabular}{l@{\quad}lll}
\toprule
 & Default & Tuned & Tuned + Ens. \\
\midrule
Linear & \underline{\textbf{0.245 $\pm$ 0.012}} & \underline{\textbf{0.242 $\pm$ 0.009}} & \underline{\textbf{0.240 $\pm$ 0.009}} \\
RandomForest & \underline{\textbf{0.211 $\pm$ 0.009}} & \underline{\textbf{0.202 $\pm$ 0.009}} & \underline{\textbf{0.202 $\pm$ 0.009}} \\
ExtraTrees & \underline{\textbf{0.202 $\pm$ 0.011}} & \underline{\textbf{0.201 $\pm$ 0.011}} & \underline{\textbf{0.200 $\pm$ 0.011}} \\
CatBoost & \underline{\textbf{0.196 $\pm$ 0.008}} & \underline{\textbf{0.195 $\pm$ 0.008}} & \underline{\textbf{0.195 $\pm$ 0.008}} \\
LightGBM & \underline{\textbf{0.196 $\pm$ 0.010}} & \underline{\textbf{0.195 $\pm$ 0.010}} & \underline{\textbf{0.193 $\pm$ 0.010}} \\
XGBoost & \underline{\textbf{0.198 $\pm$ 0.009}} & \underline{\textbf{0.195 $\pm$ 0.010}} & \underline{\textbf{0.194 $\pm$ 0.010}} \\
RealMLP & \underline{\textbf{0.197 $\pm$ 0.009}} & \underline{\textbf{0.193 $\pm$ 0.009}} & \underline{\textbf{0.193 $\pm$ 0.009}} \\
TabM & \underline{\textbf{0.199 $\pm$ 0.010}} & \underline{\textbf{0.198 $\pm$ 0.012}} & \underline{\textbf{0.197 $\pm$ 0.011}} \\
TabDPT & \underline{\textbf{0.322 $\pm$ 0.018}} & - & - \\
TabPFN-2.6 & \underline{\textbf{\textcolor{green!50!black}{0.189 $\pm$ 0.010}}} & - & - \\
TabICLv2 & \underline{\textbf{0.197 $\pm$ 0.011}} & - & - \\
\bottomrule
\end{tabular}
\space
  \end{minipage}\hfill
  \begin{minipage}[t]{0.48\textwidth}
    \centering
    \vspace{0pt}
    \includegraphics[width=\linewidth]{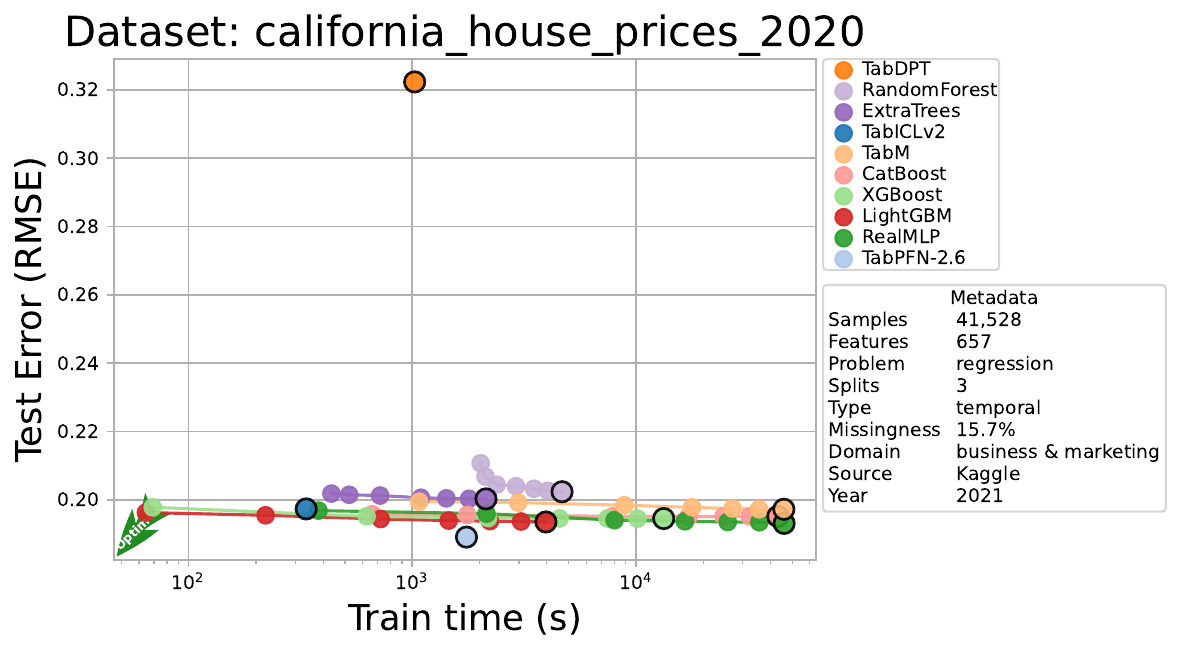}
  \end{minipage}

  \captionof{figure}{\textbf{california\_house\_prices\_2020}: per-method test error (left) and HPO Pareto trajectory (right).}
  \label{fig:perdataset_california_house_prices_2020}
\end{center}
\space
\begin{center}
  \begin{minipage}[t]{0.48\textwidth}
    \centering
    \vspace{0pt}
    \scriptsize
\begin{tabular}{l@{\quad}lll}
\toprule
 & Default & Tuned & Tuned + Ens. \\
\midrule
Linear & 0.341 $\pm$ 0.065 & 0.345 $\pm$ 0.073 & 0.305 $\pm$ 0.062 \\
RandomForest & 0.253 $\pm$ 0.045 & 0.246 $\pm$ 0.052 & 0.241 $\pm$ 0.049 \\
ExtraTrees & 0.281 $\pm$ 0.046 & 0.245 $\pm$ 0.047 & 0.243 $\pm$ 0.046 \\
CatBoost & \underline{0.225 $\pm$ 0.052} & \underline{\textbf{\textcolor{green!50!black}{0.222 $\pm$ 0.052}}} & \underline{\textbf{0.223 $\pm$ 0.052}} \\
LightGBM & 0.230 $\pm$ 0.051 & 0.232 $\pm$ 0.057 & \underline{\textbf{0.225 $\pm$ 0.050}} \\
XGBoost & \underline{0.231 $\pm$ 0.051} & 0.231 $\pm$ 0.056 & \underline{\textbf{0.227 $\pm$ 0.052}} \\
RealMLP & 0.278 $\pm$ 0.062 & 0.263 $\pm$ 0.054 & 0.245 $\pm$ 0.051 \\
TabM & 0.270 $\pm$ 0.066 & 0.259 $\pm$ 0.060 & 0.252 $\pm$ 0.058 \\
TabDPT & 0.258 $\pm$ 0.059 & - & - \\
TabPFN-2.6 & 0.254 $\pm$ 0.070 & - & - \\
TabICLv2 & \underline{\textbf{0.230 $\pm$ 0.064}} & - & - \\
\bottomrule
\end{tabular}
\space
  \end{minipage}\hfill
  \begin{minipage}[t]{0.48\textwidth}
    \centering
    \vspace{0pt}
    \includegraphics[width=\linewidth]{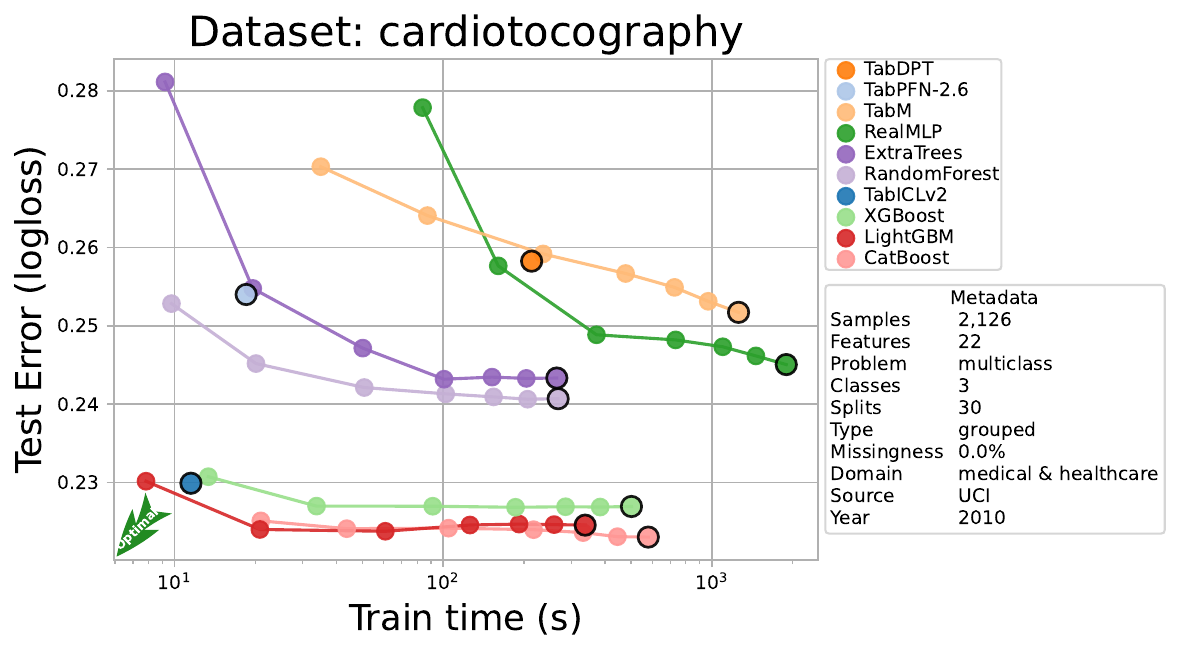}
  \end{minipage}

  \captionof{figure}{\textbf{cardiotocography}: per-method test error (left) and HPO Pareto trajectory (right).}
  \label{fig:perdataset_cardiotocography}
\end{center}
\space
\begin{center}
  \begin{minipage}[t]{0.48\textwidth}
    \centering
    \vspace{0pt}
    \scriptsize
\begin{tabular}{l@{\quad}lll}
\toprule
 & Default & Tuned & Tuned + Ens. \\
\midrule
Linear & 0.223 $\pm$ 0.014 & 0.175 $\pm$ 0.006 & 0.175 $\pm$ 0.006 \\
RandomForest & 0.083 $\pm$ 0.012 & \underline{0.082 $\pm$ 0.012} & \underline{0.081 $\pm$ 0.012} \\
ExtraTrees & 0.085 $\pm$ 0.010 & 0.082 $\pm$ 0.011 & \underline{0.080 $\pm$ 0.011} \\
CatBoost & 0.076 $\pm$ 0.005 & \underline{0.076 $\pm$ 0.005} & \underline{0.076 $\pm$ 0.005} \\
LightGBM & 0.083 $\pm$ 0.011 & \underline{0.080 $\pm$ 0.011} & 0.081 $\pm$ 0.010 \\
XGBoost & 0.077 $\pm$ 0.008 & \underline{0.077 $\pm$ 0.008} & \underline{0.079 $\pm$ 0.010} \\
RealMLP & 0.081 $\pm$ 0.009 & \underline{0.077 $\pm$ 0.010} & \underline{0.077 $\pm$ 0.009} \\
TabM & 0.078 $\pm$ 0.010 & \underline{0.079 $\pm$ 0.011} & \underline{0.080 $\pm$ 0.011} \\
TabDPT & 0.074 $\pm$ 0.012 & - & - \\
TabPFN-2.6 & 0.074 $\pm$ 0.010 & - & - \\
TabICLv2 & \underline{\textbf{\textcolor{green!50!black}{0.067 $\pm$ 0.010}}} & - & - \\
\bottomrule
\end{tabular}
\space
  \end{minipage}\hfill
  \begin{minipage}[t]{0.48\textwidth}
    \centering
    \vspace{0pt}
    \includegraphics[width=\linewidth]{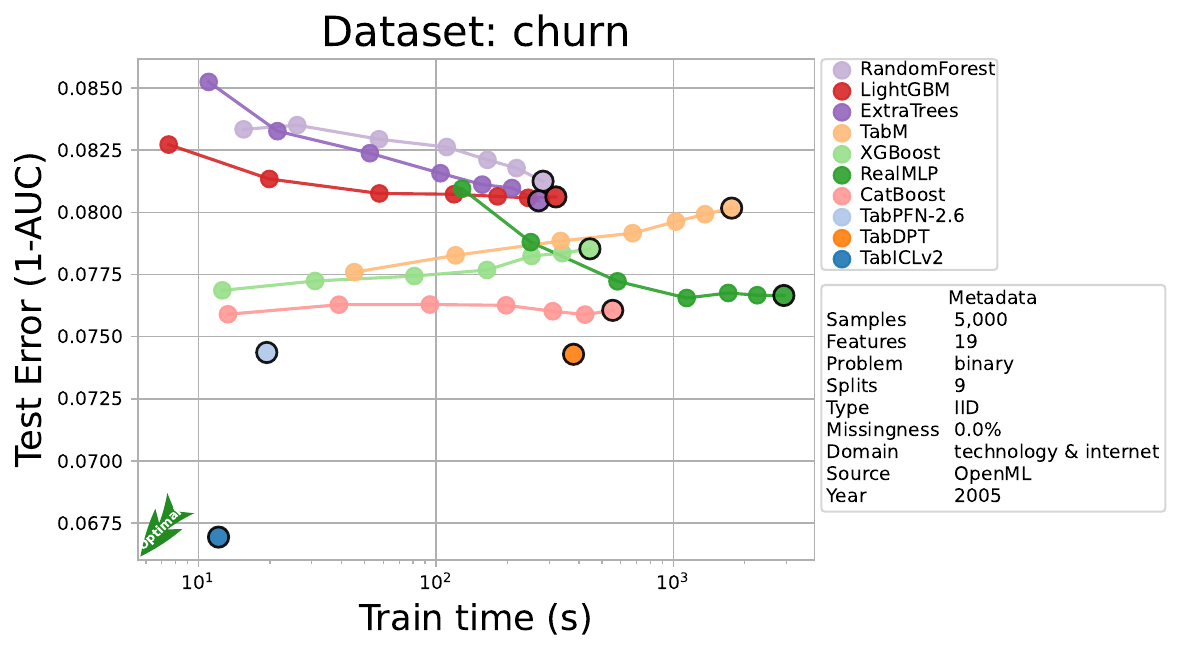}
  \end{minipage}

  \captionof{figure}{\textbf{churn}: per-method test error (left) and HPO Pareto trajectory (right).}
  \label{fig:perdataset_churn}
\end{center}
\space
\begin{center}
  \begin{minipage}[t]{0.48\textwidth}
    \centering
    \vspace{0pt}
    \scriptsize
\begin{tabular}{l@{\quad}lll}
\toprule
 & Default & Tuned & Tuned + Ens. \\
\midrule
Linear & 0.905 $\pm$ 0.089 & 0.878 $\pm$ 0.084 & 0.870 $\pm$ 0.083 \\
RandomForest & 0.879 $\pm$ 0.082 & 0.867 $\pm$ 0.077 & 0.867 $\pm$ 0.078 \\
ExtraTrees & 0.877 $\pm$ 0.078 & 0.883 $\pm$ 0.079 & 0.882 $\pm$ 0.079 \\
CatBoost & 0.875 $\pm$ 0.080 & 0.863 $\pm$ 0.078 & 0.863 $\pm$ 0.078 \\
LightGBM & 0.884 $\pm$ 0.080 & 0.869 $\pm$ 0.081 & 0.868 $\pm$ 0.082 \\
XGBoost & 0.909 $\pm$ 0.085 & 0.873 $\pm$ 0.081 & 0.872 $\pm$ 0.080 \\
RealMLP & 0.840 $\pm$ 0.078 & \underline{\textbf{0.839 $\pm$ 0.076}} & \underline{\textbf{0.833 $\pm$ 0.075}} \\
TabM & \underline{\textbf{\textcolor{green!50!black}{0.829 $\pm$ 0.080}}} & \underline{0.836 $\pm$ 0.079} & \underline{0.835 $\pm$ 0.078} \\
TabDPT & 0.848 $\pm$ 0.083 & - & - \\
TabPFN-2.6 & 0.849 $\pm$ 0.082 & - & - \\
TabICLv2 & 0.855 $\pm$ 0.086 & - & - \\
\bottomrule
\end{tabular}
\space
  \end{minipage}\hfill
  \begin{minipage}[t]{0.48\textwidth}
    \centering
    \vspace{0pt}
    \includegraphics[width=\linewidth]{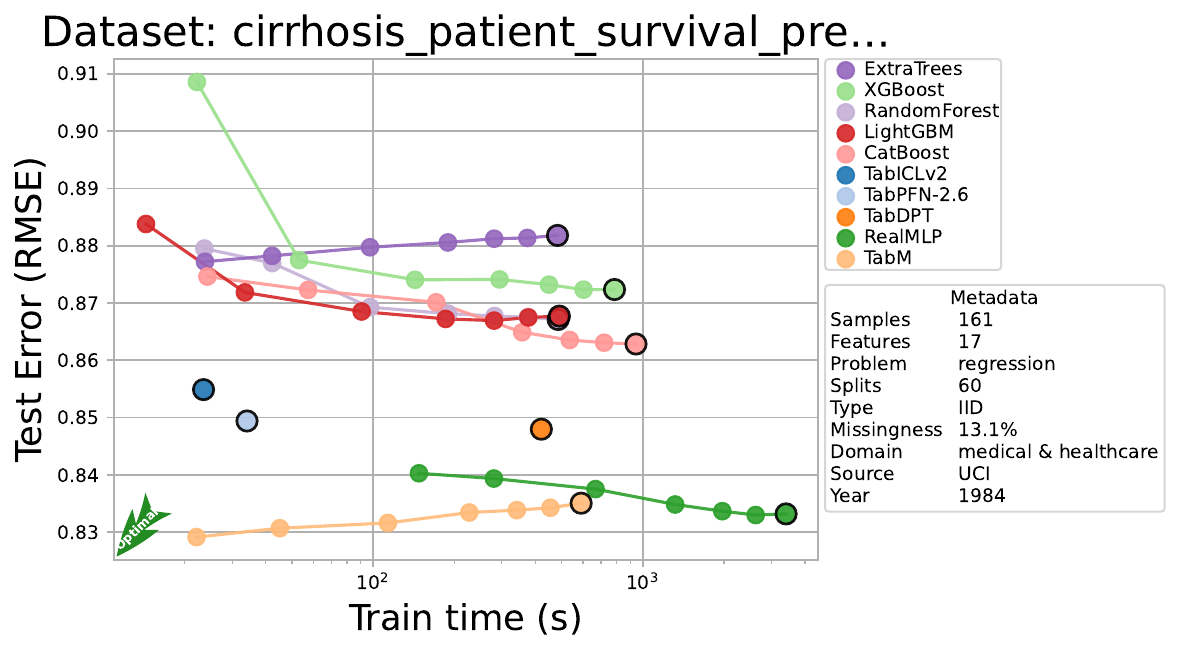}
  \end{minipage}

  \captionof{figure}{\textbf{cirrhosis\_patient\_survival\_prediction}: per-method test error (left) and HPO Pareto trajectory (right).}
  \label{fig:perdataset_cirrhosis_patient_survival_prediction}
\end{center}
\space
\begin{center}
  \begin{minipage}[t]{0.48\textwidth}
    \centering
    \vspace{0pt}
    \scriptsize
\begin{tabular}{l@{\quad}lll}
\toprule
 & Default & Tuned & Tuned + Ens. \\
\midrule
Linear & 1.538 & 1.545 & 1.537 \\
RandomForest & 1.496 & 1.496 & 1.494 \\
ExtraTrees & 1.526 & 1.519 & 1.519 \\
CatBoost & 1.339 & 1.330 & 1.325 \\
LightGBM & 1.353 & 1.329 & 1.321 \\
XGBoost & 1.341 & 1.341 & 1.327 \\
RealMLP & 1.332 & 1.300 & \textcolor{green!50!black}{1.293} \\
TabM & 1.340 & 1.312 & 1.312 \\
TabDPT & - & - & - \\
TabPFN-2.6 & - & - & - \\
TabICLv2 & 1.392 & - & - \\
\bottomrule
\end{tabular}
\space
  \end{minipage}\hfill
  \begin{minipage}[t]{0.48\textwidth}
    \centering
    \vspace{0pt}
    \includegraphics[width=\linewidth]{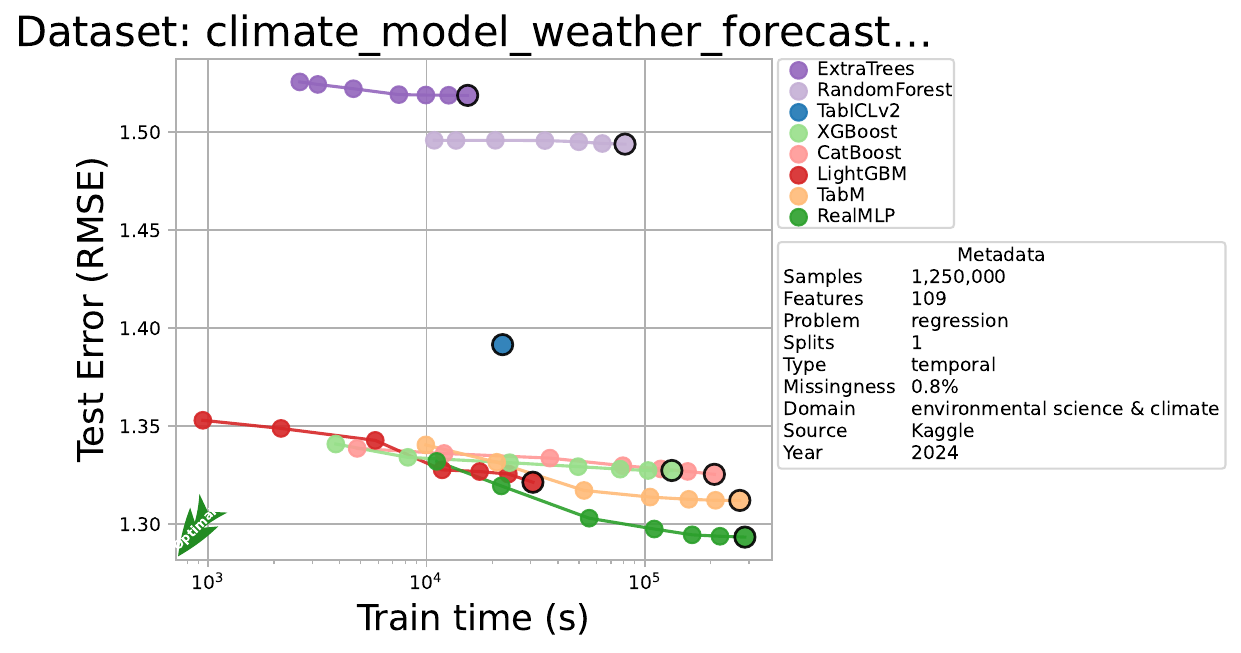}
  \end{minipage}

  \captionof{figure}{\textbf{climate\_model\_weather\_forecasting\_1m}: per-method test error (left) and HPO Pareto trajectory (right).}
  \label{fig:perdataset_climate_model_weather_forecasting_1m}
\end{center}
\space
\begin{center}
  \begin{minipage}[t]{0.48\textwidth}
    \centering
    \vspace{0pt}
    \scriptsize
\begin{tabular}{l@{\quad}lll}
\toprule
 & Default & Tuned & Tuned + Ens. \\
\midrule
Linear & 0.528 $\pm$ 0.075 & 0.541 $\pm$ 0.069 & 0.542 $\pm$ 0.069 \\
RandomForest & 0.518 $\pm$ 0.069 & 0.511 $\pm$ 0.080 & 0.510 $\pm$ 0.079 \\
ExtraTrees & 0.509 $\pm$ 0.069 & 0.500 $\pm$ 0.071 & 0.498 $\pm$ 0.071 \\
CatBoost & 0.482 $\pm$ 0.065 & 0.477 $\pm$ 0.067 & 0.474 $\pm$ 0.067 \\
LightGBM & \underline{\textbf{0.461 $\pm$ 0.062}} & 0.472 $\pm$ 0.058 & 0.470 $\pm$ 0.059 \\
XGBoost & 0.477 $\pm$ 0.066 & \underline{\textbf{\textcolor{green!50!black}{0.461 $\pm$ 0.066}}} & \underline{\textbf{0.461 $\pm$ 0.063}} \\
RealMLP & 0.503 $\pm$ 0.069 & 0.479 $\pm$ 0.066 & \underline{\textbf{0.465 $\pm$ 0.063}} \\
TabM & 0.477 $\pm$ 0.063 & \underline{\textbf{0.468 $\pm$ 0.062}} & \underline{\textbf{0.469 $\pm$ 0.065}} \\
TabDPT & 0.544 $\pm$ 0.067 & - & - \\
TabPFN-2.6 & 0.478 $\pm$ 0.068 & - & - \\
TabICLv2 & 0.508 $\pm$ 0.068 & - & - \\
\bottomrule
\end{tabular}
\space
  \end{minipage}\hfill
  \begin{minipage}[t]{0.48\textwidth}
    \centering
    \vspace{0pt}
    \includegraphics[width=\linewidth]{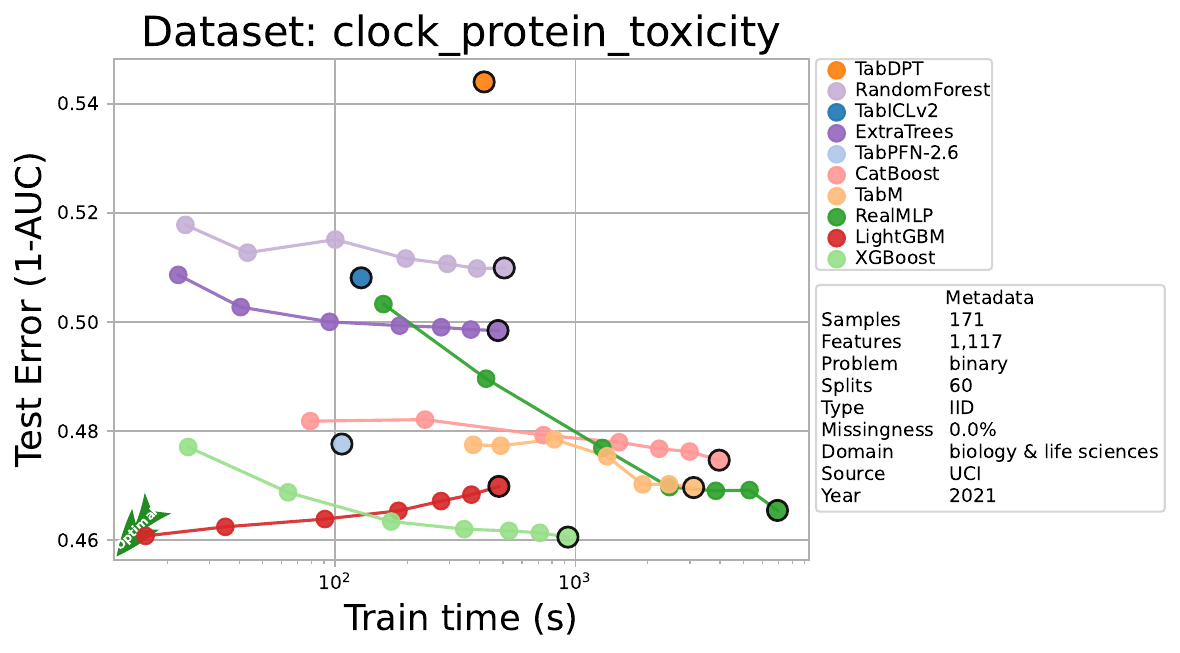}
  \end{minipage}

  \captionof{figure}{\textbf{clock\_protein\_toxicity}: per-method test error (left) and HPO Pareto trajectory (right).}
  \label{fig:perdataset_clock_protein_toxicity}
\end{center}
\space
\begin{center}
  \begin{minipage}[t]{0.48\textwidth}
    \centering
    \vspace{0pt}
    \scriptsize
\begin{tabular}{l@{\quad}lll}
\toprule
 & Default & Tuned & Tuned + Ens. \\
\midrule
Linear & \underline{\textbf{1.205 $\pm$ 0.175}} & \underline{\textbf{1.145 $\pm$ 0.150}} & \underline{\textbf{1.148 $\pm$ 0.187}} \\
RandomForest & \underline{\textbf{1.212 $\pm$ 0.205}} & \underline{\textbf{1.206 $\pm$ 0.194}} & \underline{\textbf{1.204 $\pm$ 0.204}} \\
ExtraTrees & \underline{\textbf{1.247 $\pm$ 0.257}} & \underline{\textbf{1.230 $\pm$ 0.250}} & \underline{\textbf{1.229 $\pm$ 0.251}} \\
CatBoost & \underline{\textbf{1.105 $\pm$ 0.183}} & \underline{\textbf{1.093 $\pm$ 0.165}} & \underline{\textbf{1.093 $\pm$ 0.168}} \\
LightGBM & \underline{\textbf{1.136 $\pm$ 0.185}} & \underline{\textbf{1.085 $\pm$ 0.186}} & \underline{\textbf{1.085 $\pm$ 0.186}} \\
XGBoost & \underline{\textbf{1.133 $\pm$ 0.196}} & \underline{\textbf{1.091 $\pm$ 0.182}} & \underline{\textbf{1.091 $\pm$ 0.182}} \\
RealMLP & \underline{\textbf{1.079 $\pm$ 0.175}} & \underline{\textbf{1.079 $\pm$ 0.184}} & \underline{\textbf{1.070 $\pm$ 0.190}} \\
TabM & \underline{\textbf{1.104 $\pm$ 0.179}} & \underline{\textbf{1.104 $\pm$ 0.158}} & \underline{\textbf{1.100 $\pm$ 0.168}} \\
TabDPT & \underline{\textbf{1.111 $\pm$ 0.140}} & - & - \\
TabPFN-2.6 & \underline{\textbf{\textcolor{green!50!black}{1.042 $\pm$ 0.183}}} & - & - \\
TabICLv2 & \underline{\textbf{1.080 $\pm$ 0.172}} & - & - \\
\bottomrule
\end{tabular}
\space
  \end{minipage}\hfill
  \begin{minipage}[t]{0.48\textwidth}
    \centering
    \vspace{0pt}
    \includegraphics[width=\linewidth]{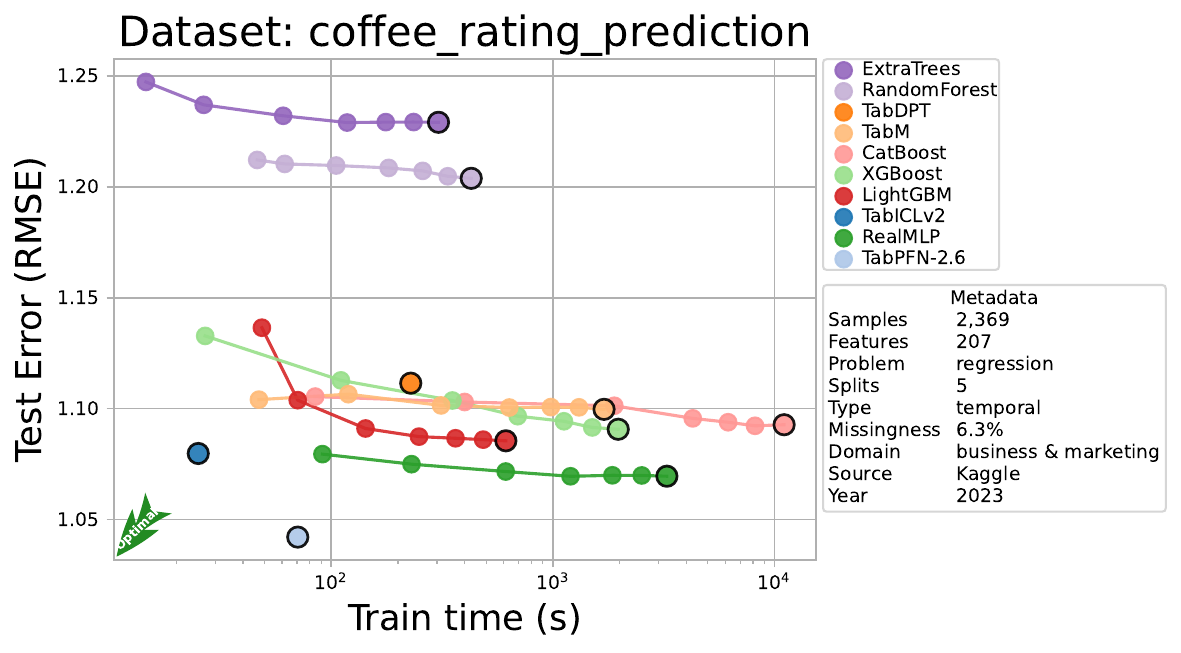}
  \end{minipage}

  \captionof{figure}{\textbf{coffee\_rating\_prediction}: per-method test error (left) and HPO Pareto trajectory (right).}
  \label{fig:perdataset_coffee_rating_prediction}
\end{center}
\space
\begin{center}
  \begin{minipage}[t]{0.48\textwidth}
    \centering
    \vspace{0pt}
    \scriptsize
\begin{tabular}{l@{\quad}lll}
\toprule
 & Default & Tuned & Tuned + Ens. \\
\midrule
Linear & 0.267 $\pm$ 0.020 & 0.264 $\pm$ 0.022 & 0.264 $\pm$ 0.022 \\
RandomForest & 0.296 $\pm$ 0.019 & 0.257 $\pm$ 0.021 & 0.257 $\pm$ 0.020 \\
ExtraTrees & 0.296 $\pm$ 0.020 & 0.261 $\pm$ 0.023 & 0.255 $\pm$ 0.023 \\
CatBoost & 0.241 $\pm$ 0.021 & \underline{0.240 $\pm$ 0.018} & \underline{0.240 $\pm$ 0.021} \\
LightGBM & 0.251 $\pm$ 0.019 & \underline{0.239 $\pm$ 0.019} & \underline{0.238 $\pm$ 0.020} \\
XGBoost & 0.242 $\pm$ 0.016 & 0.247 $\pm$ 0.021 & 0.243 $\pm$ 0.018 \\
RealMLP & 0.263 $\pm$ 0.021 & \underline{0.244 $\pm$ 0.019} & 0.242 $\pm$ 0.017 \\
TabM & 0.246 $\pm$ 0.019 & \underline{0.239 $\pm$ 0.022} & \underline{0.239 $\pm$ 0.021} \\
TabDPT & 0.276 $\pm$ 0.020 & - & - \\
TabPFN-2.6 & \underline{\textbf{\textcolor{green!50!black}{0.225 $\pm$ 0.021}}} & - & - \\
TabICLv2 & \underline{\textbf{0.229 $\pm$ 0.022}} & - & - \\
\bottomrule
\end{tabular}
\space
  \end{minipage}\hfill
  \begin{minipage}[t]{0.48\textwidth}
    \centering
    \vspace{0pt}
    \includegraphics[width=\linewidth]{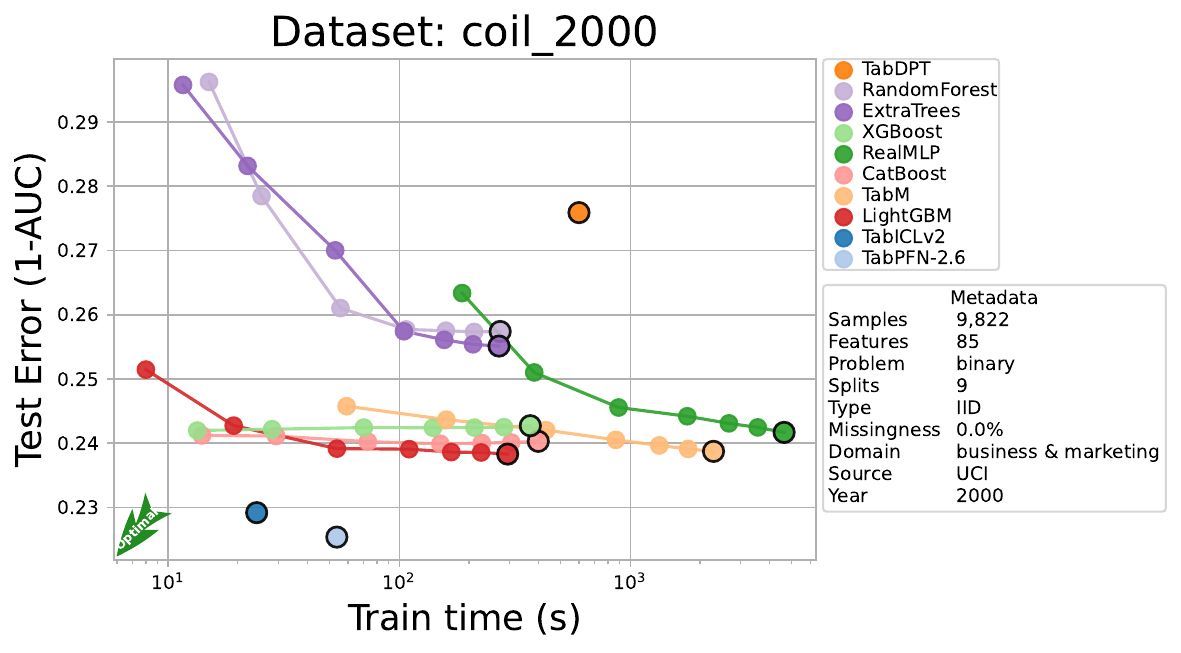}
  \end{minipage}

  \captionof{figure}{\textbf{coil\_2000}: per-method test error (left) and HPO Pareto trajectory (right).}
  \label{fig:perdataset_coil_2000}
\end{center}
\space
\begin{center}
  \begin{minipage}[t]{0.48\textwidth}
    \centering
    \vspace{0pt}
    \scriptsize
\begin{tabular}{l@{\quad}lll}
\toprule
 & Default & Tuned & Tuned + Ens. \\
\midrule
Linear & 8.221 $\pm$ 0.388 & 8.161 $\pm$ 0.395 & 8.158 $\pm$ 0.392 \\
RandomForest & 5.212 $\pm$ 0.367 & 5.018 $\pm$ 0.409 & 5.008 $\pm$ 0.410 \\
ExtraTrees & 5.153 $\pm$ 0.364 & 4.946 $\pm$ 0.388 & 4.927 $\pm$ 0.401 \\
CatBoost & 4.127 $\pm$ 0.356 & \underline{4.143 $\pm$ 0.364} & 4.118 $\pm$ 0.367 \\
LightGBM & 4.389 $\pm$ 0.347 & 4.170 $\pm$ 0.363 & 4.165 $\pm$ 0.340 \\
XGBoost & 4.609 $\pm$ 0.373 & 4.173 $\pm$ 0.325 & 4.151 $\pm$ 0.327 \\
RealMLP & 4.683 $\pm$ 0.321 & \underline{4.089 $\pm$ 0.311} & \underline{4.026 $\pm$ 0.294} \\
TabM & 4.206 $\pm$ 0.304 & 4.176 $\pm$ 0.305 & 4.095 $\pm$ 0.301 \\
TabDPT & 4.195 $\pm$ 0.399 & - & - \\
TabPFN-2.6 & 3.978 $\pm$ 0.333 & - & - \\
TabICLv2 & \underline{\textbf{\textcolor{green!50!black}{3.901 $\pm$ 0.351}}} & - & - \\
\bottomrule
\end{tabular}
\space
  \end{minipage}\hfill
  \begin{minipage}[t]{0.48\textwidth}
    \centering
    \vspace{0pt}
    \includegraphics[width=\linewidth]{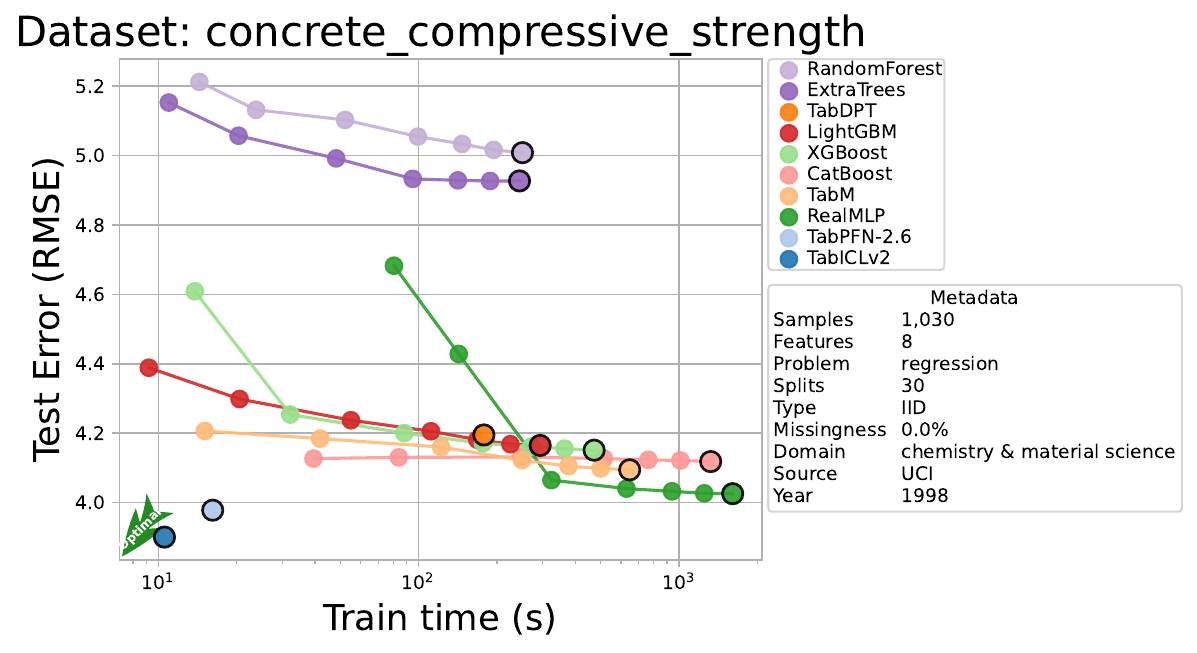}
  \end{minipage}

  \captionof{figure}{\textbf{concrete\_compressive\_strength}: per-method test error (left) and HPO Pareto trajectory (right).}
  \label{fig:perdataset_concrete_compressive_strength}
\end{center}
\space
\begin{center}
  \begin{minipage}[t]{0.48\textwidth}
    \centering
    \vspace{0pt}
    \scriptsize
\begin{tabular}{l@{\quad}lll}
\toprule
 & Default & Tuned & Tuned + Ens. \\
\midrule
Linear & 0.561 & 0.601 & 0.565 \\
RandomForest & 0.437 & 0.437 & 0.432 \\
ExtraTrees & 0.479 & 0.427 & 0.430 \\
CatBoost & 0.431 & \textcolor{green!50!black}{0.423} & 0.425 \\
LightGBM & 0.429 & 0.430 & 0.428 \\
XGBoost & 0.432 & 0.433 & 0.431 \\
RealMLP & 0.446 & 0.446 & 0.435 \\
TabM & 0.443 & 0.426 & 0.424 \\
TabDPT & - & - & - \\
TabPFN-2.6 & - & - & - \\
TabICLv2 & 0.445 & - & - \\
\bottomrule
\end{tabular}
\space
  \end{minipage}\hfill
  \begin{minipage}[t]{0.48\textwidth}
    \centering
    \vspace{0pt}
    \includegraphics[width=\linewidth]{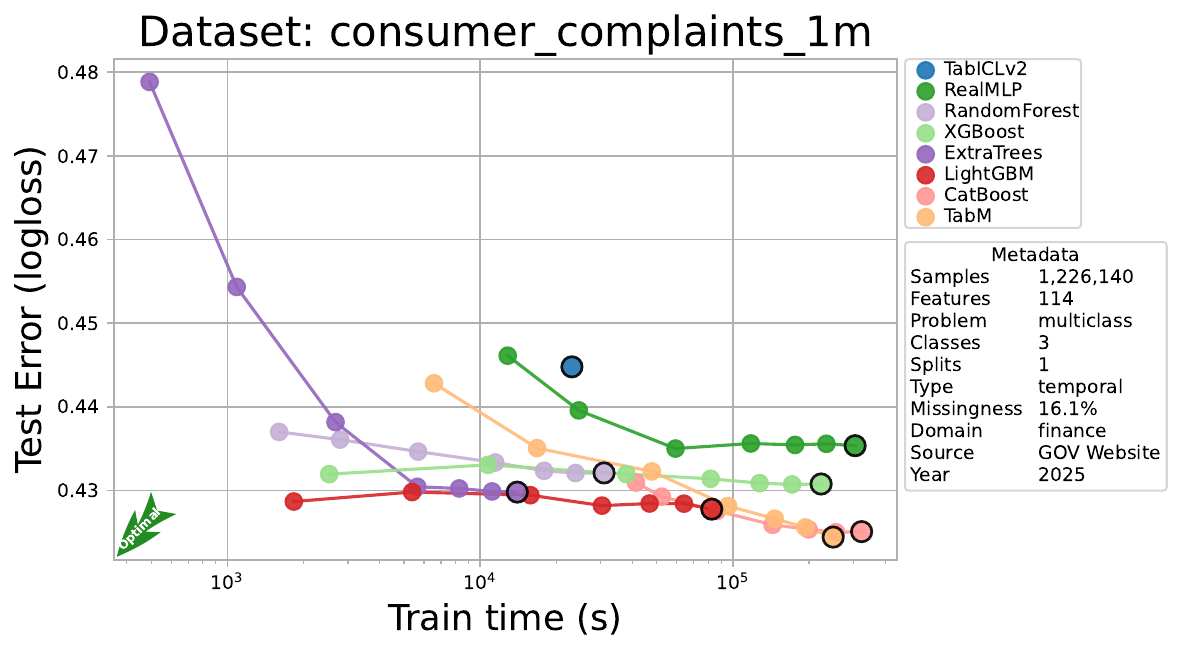}
  \end{minipage}

  \captionof{figure}{\textbf{consumer\_complaints\_1m}: per-method test error (left) and HPO Pareto trajectory (right).}
  \label{fig:perdataset_consumer_complaints_1m}
\end{center}
\space
\begin{center}
  \begin{minipage}[t]{0.48\textwidth}
    \centering
    \vspace{0pt}
    \scriptsize
\begin{tabular}{l@{\quad}lll}
\toprule
 & Default & Tuned & Tuned + Ens. \\
\midrule
Linear & 0.487 & 0.485 & 0.483 \\
RandomForest & 0.483 & 0.483 & 0.483 \\
ExtraTrees & 0.485 & 0.485 & 0.485 \\
CatBoost & 0.476 & 0.476 & \textcolor{green!50!black}{0.476} \\
LightGBM & 0.481 & 0.477 & 0.476 \\
XGBoost & 0.478 & 0.478 & 0.477 \\
RealMLP & 0.480 & 0.477 & 0.476 \\
TabM & 0.482 & 0.479 & 0.478 \\
TabDPT & - & - & - \\
TabPFN-2.6 & - & - & - \\
TabICLv2 & 0.488 & - & - \\
\bottomrule
\end{tabular}
\space
  \end{minipage}\hfill
  \begin{minipage}[t]{0.48\textwidth}
    \centering
    \vspace{0pt}
    \includegraphics[width=\linewidth]{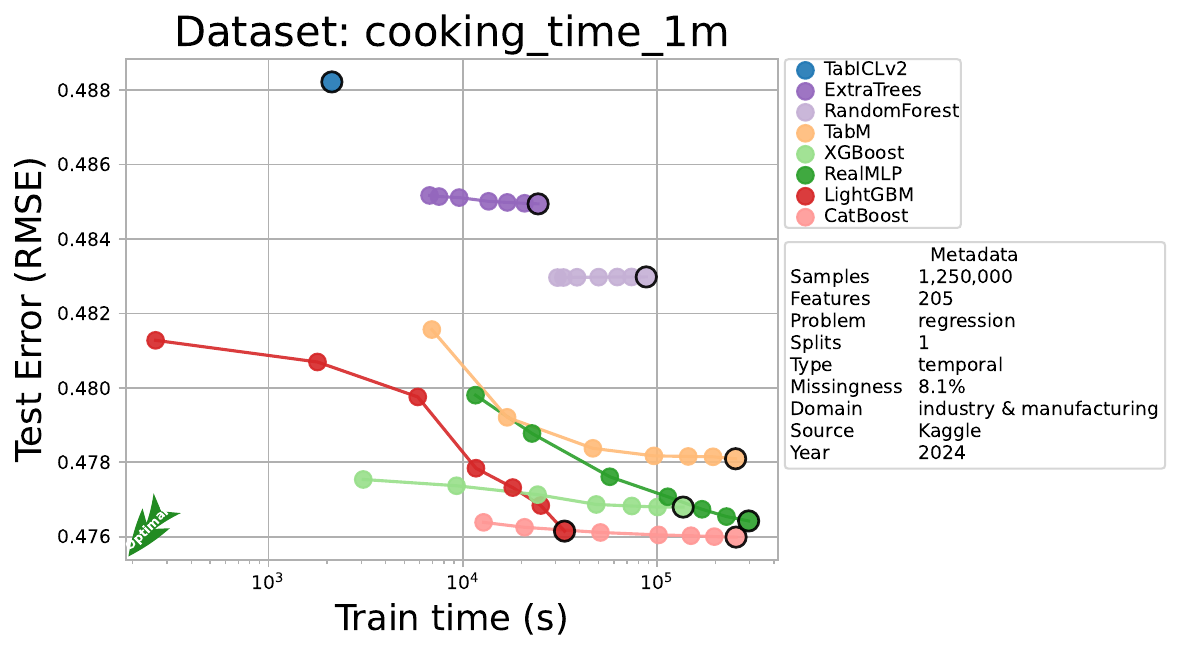}
  \end{minipage}

  \captionof{figure}{\textbf{cooking\_time\_1m}: per-method test error (left) and HPO Pareto trajectory (right).}
  \label{fig:perdataset_cooking_time_1m}
\end{center}
\space
\begin{center}
  \begin{minipage}[t]{0.48\textwidth}
    \centering
    \vspace{0pt}
    \scriptsize
\begin{tabular}{l@{\quad}lll}
\toprule
 & Default & Tuned & Tuned + Ens. \\
\midrule
Linear & \underline{\textbf{0.819 $\pm$ 0.128}} & \underline{\textbf{0.805 $\pm$ 0.143}} & \underline{\textbf{0.805 $\pm$ 0.140}} \\
RandomForest & \underline{\textbf{0.769 $\pm$ 0.056}} & \underline{\textbf{0.754 $\pm$ 0.049}} & \underline{\textbf{0.752 $\pm$ 0.053}} \\
ExtraTrees & \underline{\textbf{0.783 $\pm$ 0.034}} & \underline{\textbf{0.754 $\pm$ 0.025}} & \underline{\textbf{0.762 $\pm$ 0.019}} \\
CatBoost & \underline{\textbf{0.739 $\pm$ 0.026}} & \underline{\textbf{0.715 $\pm$ 0.046}} & \underline{\textbf{0.717 $\pm$ 0.034}} \\
LightGBM & \underline{\textbf{0.807 $\pm$ 0.016}} & \underline{\textbf{0.720 $\pm$ 0.058}} & \underline{\textbf{0.724 $\pm$ 0.051}} \\
XGBoost & \underline{\textbf{0.812 $\pm$ 0.038}} & \underline{\textbf{0.743 $\pm$ 0.046}} & \underline{\textbf{0.744 $\pm$ 0.045}} \\
RealMLP & \underline{\textbf{0.776 $\pm$ 0.088}} & \underline{\textbf{0.733 $\pm$ 0.055}} & \underline{\textbf{0.712 $\pm$ 0.061}} \\
TabM & \underline{\textbf{0.850 $\pm$ 0.102}} & \underline{\textbf{\textcolor{green!50!black}{0.696 $\pm$ 0.109}}} & \underline{\textbf{0.707 $\pm$ 0.098}} \\
TabDPT & - & - & - \\
TabPFN-2.6 & - & - & - \\
TabICLv2 & \underline{\textbf{1.025 $\pm$ 0.256}} & - & - \\
\bottomrule
\end{tabular}
\space
  \end{minipage}\hfill
  \begin{minipage}[t]{0.48\textwidth}
    \centering
    \vspace{0pt}
    \includegraphics[width=\linewidth]{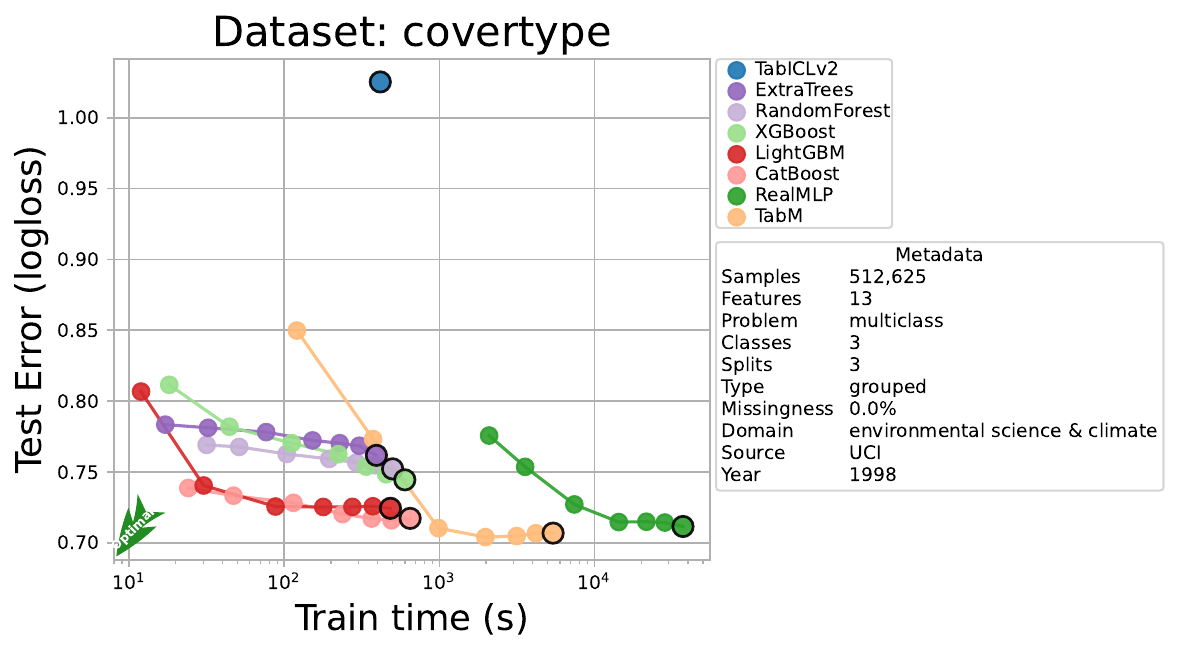}
  \end{minipage}

  \captionof{figure}{\textbf{covertype}: per-method test error (left) and HPO Pareto trajectory (right).}
  \label{fig:perdataset_covertype}
\end{center}
\space
\begin{center}
  \begin{minipage}[t]{0.48\textwidth}
    \centering
    \vspace{0pt}
    \scriptsize
\begin{tabular}{l@{\quad}lll}
\toprule
 & Default & Tuned & Tuned + Ens. \\
\midrule
Linear & 0.079 $\pm$ 0.016 & 0.076 $\pm$ 0.017 & 0.075 $\pm$ 0.017 \\
RandomForest & 0.066 $\pm$ 0.014 & 0.065 $\pm$ 0.015 & 0.065 $\pm$ 0.015 \\
ExtraTrees & 0.074 $\pm$ 0.016 & 0.076 $\pm$ 0.016 & 0.075 $\pm$ 0.016 \\
CatBoost & 0.062 $\pm$ 0.014 & 0.065 $\pm$ 0.015 & 0.064 $\pm$ 0.014 \\
LightGBM & 0.063 $\pm$ 0.014 & \underline{0.064 $\pm$ 0.015} & \underline{0.062 $\pm$ 0.014} \\
XGBoost & 0.067 $\pm$ 0.014 & \underline{0.063 $\pm$ 0.014} & \underline{0.062 $\pm$ 0.014} \\
RealMLP & 0.075 $\pm$ 0.016 & 0.069 $\pm$ 0.015 & 0.068 $\pm$ 0.015 \\
TabM & 0.064 $\pm$ 0.015 & 0.065 $\pm$ 0.015 & 0.064 $\pm$ 0.015 \\
TabDPT & 0.066 $\pm$ 0.016 & - & - \\
TabPFN-2.6 & \underline{\textbf{\textcolor{green!50!black}{0.060 $\pm$ 0.014}}} & - & - \\
TabICLv2 & 0.061 $\pm$ 0.013 & - & - \\
\bottomrule
\end{tabular}
\space
  \end{minipage}\hfill
  \begin{minipage}[t]{0.48\textwidth}
    \centering
    \vspace{0pt}
    \includegraphics[width=\linewidth]{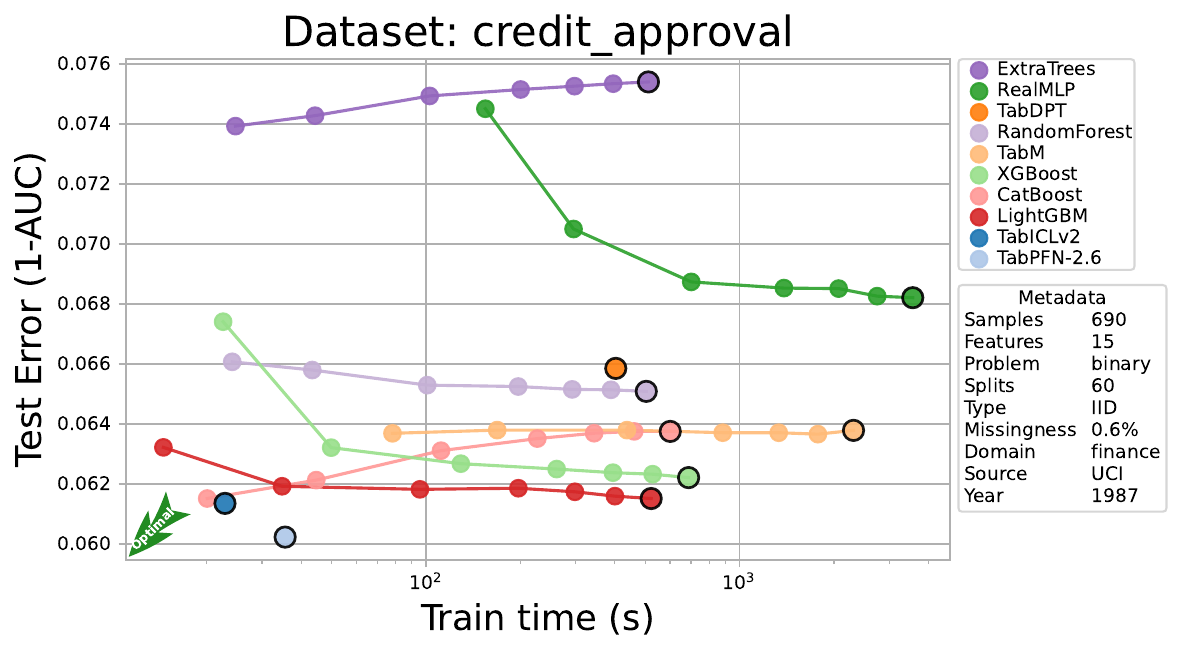}
  \end{minipage}

  \captionof{figure}{\textbf{credit\_approval}: per-method test error (left) and HPO Pareto trajectory (right).}
  \label{fig:perdataset_credit_approval}
\end{center}
\space
\begin{center}
  \begin{minipage}[t]{0.48\textwidth}
    \centering
    \vspace{0pt}
    \scriptsize
\begin{tabular}{l@{\quad}lll}
\toprule
 & Default & Tuned & Tuned + Ens. \\
\midrule
Linear & 0.254 $\pm$ 0.005 & 0.254 $\pm$ 0.005 & 0.255 $\pm$ 0.005 \\
RandomForest & 0.231 $\pm$ 0.004 & 0.219 $\pm$ 0.004 & 0.219 $\pm$ 0.004 \\
ExtraTrees & 0.229 $\pm$ 0.003 & 0.220 $\pm$ 0.002 & 0.218 $\pm$ 0.003 \\
CatBoost & 0.215 $\pm$ 0.003 & 0.215 $\pm$ 0.003 & 0.215 $\pm$ 0.003 \\
LightGBM & 0.216 $\pm$ 0.003 & 0.215 $\pm$ 0.003 & 0.214 $\pm$ 0.003 \\
XGBoost & 0.217 $\pm$ 0.003 & 0.215 $\pm$ 0.003 & 0.214 $\pm$ 0.003 \\
RealMLP & 0.215 $\pm$ 0.003 & 0.214 $\pm$ 0.003 & 0.214 $\pm$ 0.003 \\
TabM & 0.216 $\pm$ 0.004 & \underline{0.211 $\pm$ 0.003} & \underline{0.212 $\pm$ 0.003} \\
TabDPT & 0.216 $\pm$ 0.002 & - & - \\
TabPFN-2.6 & 0.210 $\pm$ 0.003 & - & - \\
TabICLv2 & \underline{\textbf{\textcolor{green!50!black}{0.209 $\pm$ 0.003}}} & - & - \\
\bottomrule
\end{tabular}
\space
  \end{minipage}\hfill
  \begin{minipage}[t]{0.48\textwidth}
    \centering
    \vspace{0pt}
    \includegraphics[width=\linewidth]{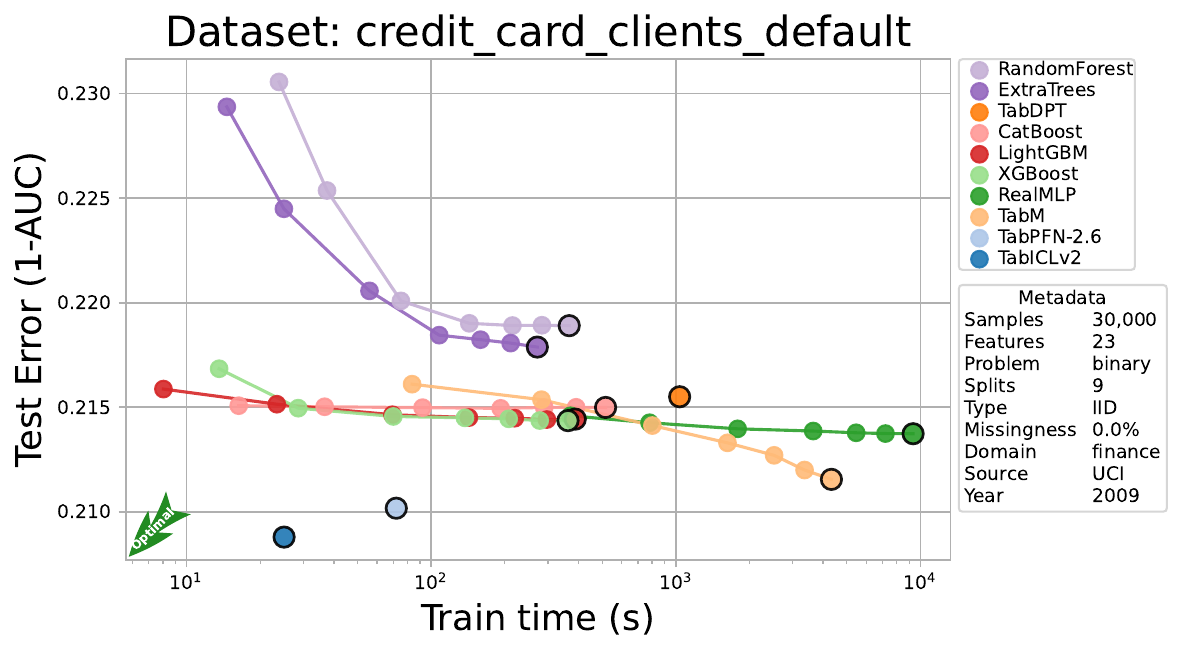}
  \end{minipage}

  \captionof{figure}{\textbf{credit\_card\_clients\_default}: per-method test error (left) and HPO Pareto trajectory (right).}
  \label{fig:perdataset_credit_card_clients_default}
\end{center}
\space
\begin{center}
  \begin{minipage}[t]{0.48\textwidth}
    \centering
    \vspace{0pt}
    \scriptsize
\begin{tabular}{l@{\quad}lll}
\toprule
 & Default & Tuned & Tuned + Ens. \\
\midrule
Linear & 0.217 $\pm$ 0.026 & \underline{0.211 $\pm$ 0.026} & 0.211 $\pm$ 0.027 \\
RandomForest & 0.215 $\pm$ 0.029 & 0.218 $\pm$ 0.028 & 0.218 $\pm$ 0.028 \\
ExtraTrees & 0.219 $\pm$ 0.028 & 0.221 $\pm$ 0.027 & 0.220 $\pm$ 0.027 \\
CatBoost & 0.210 $\pm$ 0.028 & \underline{0.207 $\pm$ 0.028} & \underline{\textbf{\textcolor{green!50!black}{0.205 $\pm$ 0.028}}} \\
LightGBM & 0.234 $\pm$ 0.025 & \underline{\textbf{0.208 $\pm$ 0.029}} & \underline{\textbf{0.206 $\pm$ 0.028}} \\
XGBoost & 0.218 $\pm$ 0.030 & \underline{0.208 $\pm$ 0.030} & 0.208 $\pm$ 0.030 \\
RealMLP & 0.222 $\pm$ 0.025 & 0.214 $\pm$ 0.029 & 0.208 $\pm$ 0.028 \\
TabM & \underline{\textbf{0.207 $\pm$ 0.029}} & \underline{\textbf{0.209 $\pm$ 0.030}} & \underline{\textbf{0.205 $\pm$ 0.029}} \\
TabDPT & 0.215 $\pm$ 0.028 & - & - \\
TabPFN-2.6 & \underline{\textbf{0.207 $\pm$ 0.026}} & - & - \\
TabICLv2 & \underline{\textbf{0.205 $\pm$ 0.026}} & - & - \\
\bottomrule
\end{tabular}
\space
  \end{minipage}\hfill
  \begin{minipage}[t]{0.48\textwidth}
    \centering
    \vspace{0pt}
    \includegraphics[width=\linewidth]{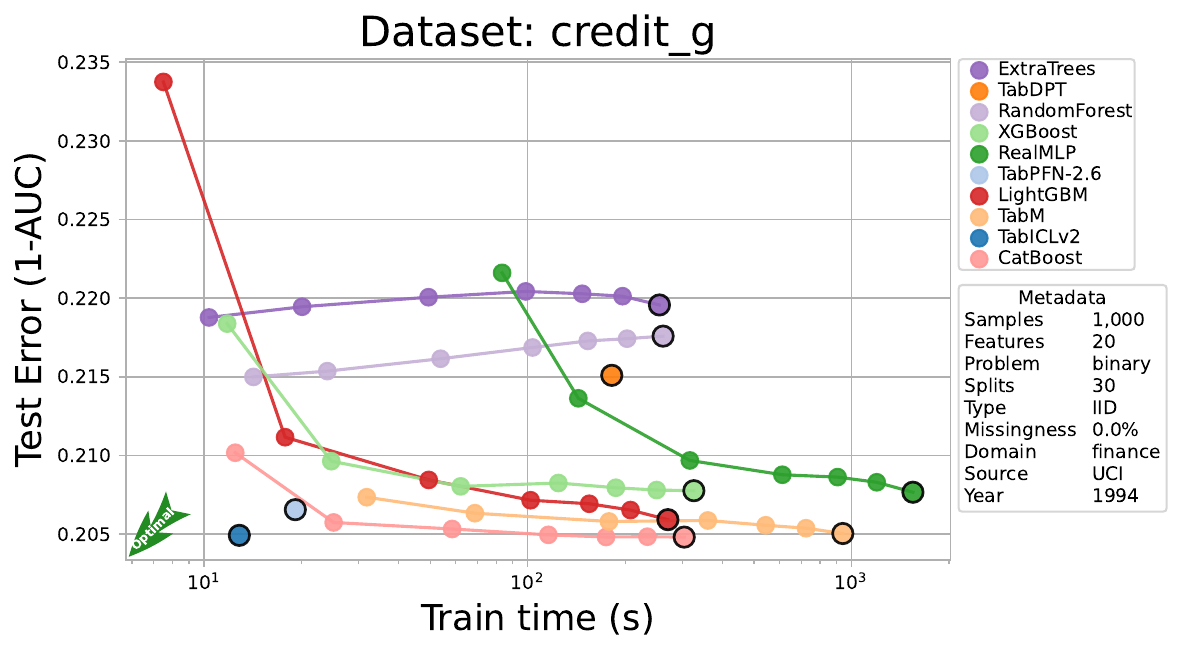}
  \end{minipage}

  \captionof{figure}{\textbf{credit\_g}: per-method test error (left) and HPO Pareto trajectory (right).}
  \label{fig:perdataset_credit_g}
\end{center}
\space
\begin{center}
  \begin{minipage}[t]{0.48\textwidth}
    \centering
    \vspace{0pt}
    \scriptsize
\begin{tabular}{l@{\quad}lll}
\toprule
 & Default & Tuned & Tuned + Ens. \\
\midrule
Linear & 0.038 $\pm$ 0.002 & 0.038 $\pm$ 0.002 & 0.038 $\pm$ 0.002 \\
RandomForest & 0.008 $\pm$ 0.000 & 0.008 $\pm$ 0.000 & 0.007 $\pm$ 0.000 \\
ExtraTrees & 0.008 $\pm$ 0.000 & 0.006 $\pm$ 0.000 & 0.006 $\pm$ 0.000 \\
CatBoost & 0.005 $\pm$ 0.000 & 0.005 $\pm$ 0.000 & 0.005 $\pm$ 0.000 \\
LightGBM & 0.006 $\pm$ 0.000 & 0.006 $\pm$ 0.000 & 0.006 $\pm$ 0.000 \\
XGBoost & 0.006 $\pm$ 0.000 & 0.006 $\pm$ 0.000 & 0.006 $\pm$ 0.000 \\
RealMLP & 0.005 $\pm$ 0.000 & 0.005 $\pm$ 0.000 & 0.005 $\pm$ 0.000 \\
TabM & 0.005 $\pm$ 0.000 & \underline{\textbf{0.005 $\pm$ 0.000}} & \underline{\textbf{\textcolor{green!50!black}{0.005 $\pm$ 0.000}}} \\
TabDPT & 0.006 $\pm$ 0.000 & - & - \\
TabPFN-2.6 & \underline{\textbf{0.005 $\pm$ 0.000}} & - & - \\
TabICLv2 & \underline{0.005 $\pm$ 0.000} & - & - \\
\bottomrule
\end{tabular}
\space
  \end{minipage}\hfill
  \begin{minipage}[t]{0.48\textwidth}
    \centering
    \vspace{0pt}
    \includegraphics[width=\linewidth]{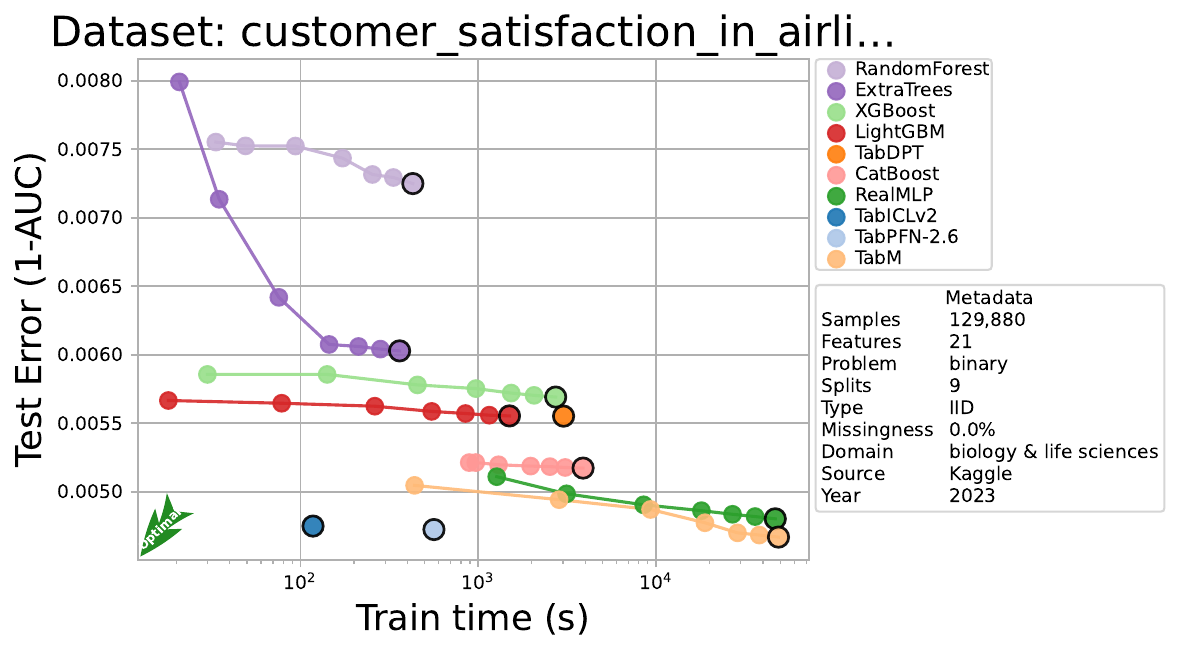}
  \end{minipage}

  \captionof{figure}{\textbf{customer\_satisfaction\_in\_airline}: per-method test error (left) and HPO Pareto trajectory (right).}
  \label{fig:perdataset_customer_satisfaction_in_airline}
\end{center}
\space
\begin{center}
  \begin{minipage}[t]{0.48\textwidth}
    \centering
    \vspace{0pt}
    \scriptsize
\begin{tabular}{l@{\quad}lll}
\toprule
 & Default & Tuned & Tuned + Ens. \\
\midrule
Linear & 1.004 & 0.999 & 0.874 \\
RandomForest & 0.530 & 0.530 & 0.531 \\
ExtraTrees & 0.533 & 0.533 & 0.533 \\
CatBoost & 0.516 & 0.515 & 0.515 \\
LightGBM & 0.517 & 0.515 & 0.515 \\
XGBoost & 0.516 & 0.516 & 0.514 \\
RealMLP & 0.517 & \textcolor{green!50!black}{0.514} & 0.514 \\
TabM & 0.517 & 0.517 & 0.515 \\
TabDPT & - & - & - \\
TabPFN-2.6 & - & - & - \\
TabICLv2 & 0.524 & - & - \\
\bottomrule
\end{tabular}
\space
  \end{minipage}\hfill
  \begin{minipage}[t]{0.48\textwidth}
    \centering
    \vspace{0pt}
    \includegraphics[width=\linewidth]{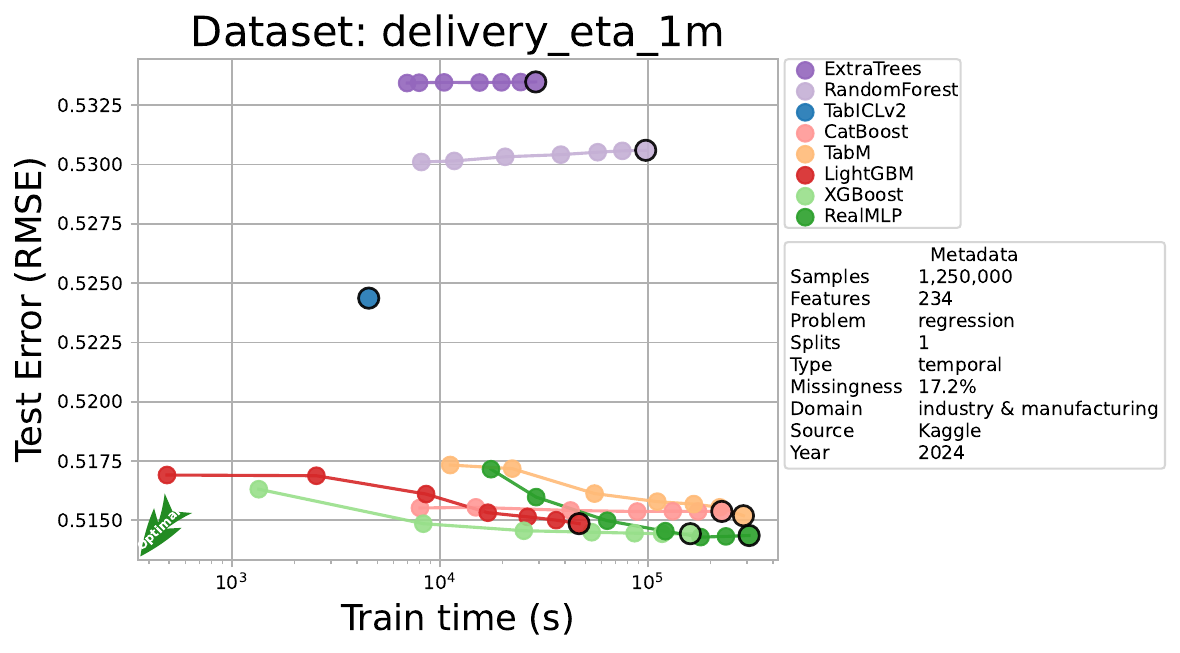}
  \end{minipage}

  \captionof{figure}{\textbf{delivery\_eta\_1m}: per-method test error (left) and HPO Pareto trajectory (right).}
  \label{fig:perdataset_delivery_eta_1m}
\end{center}
\space
\begin{center}
  \begin{minipage}[t]{0.48\textwidth}
    \centering
    \vspace{0pt}
    \scriptsize
\begin{tabular}{l@{\quad}lll}
\toprule
 & Default & Tuned & Tuned + Ens. \\
\midrule
Linear & 0.719 $\pm$ 0.071 & \underline{\textbf{0.625 $\pm$ 0.062}} & \underline{\textbf{0.624 $\pm$ 0.062}} \\
RandomForest & 0.687 $\pm$ 0.048 & 0.663 $\pm$ 0.058 & 0.662 $\pm$ 0.053 \\
ExtraTrees & 0.714 $\pm$ 0.044 & 0.645 $\pm$ 0.051 & 0.645 $\pm$ 0.051 \\
CatBoost & 0.668 $\pm$ 0.052 & 0.645 $\pm$ 0.051 & 0.646 $\pm$ 0.051 \\
LightGBM & 0.663 $\pm$ 0.053 & 0.636 $\pm$ 0.061 & 0.635 $\pm$ 0.057 \\
XGBoost & 0.714 $\pm$ 0.052 & 0.685 $\pm$ 0.058 & 0.683 $\pm$ 0.058 \\
RealMLP & \underline{0.649 $\pm$ 0.083} & \underline{\textbf{0.624 $\pm$ 0.071}} & \underline{\textbf{\textcolor{green!50!black}{0.619 $\pm$ 0.066}}} \\
TabM & \underline{0.653 $\pm$ 0.066} & 0.647 $\pm$ 0.072 & 0.644 $\pm$ 0.068 \\
TabDPT & 0.706 $\pm$ 0.079 & - & - \\
TabPFN-2.6 & 0.671 $\pm$ 0.074 & - & - \\
TabICLv2 & 0.682 $\pm$ 0.077 & - & - \\
\bottomrule
\end{tabular}
\space
  \end{minipage}\hfill
  \begin{minipage}[t]{0.48\textwidth}
    \centering
    \vspace{0pt}
    \includegraphics[width=\linewidth]{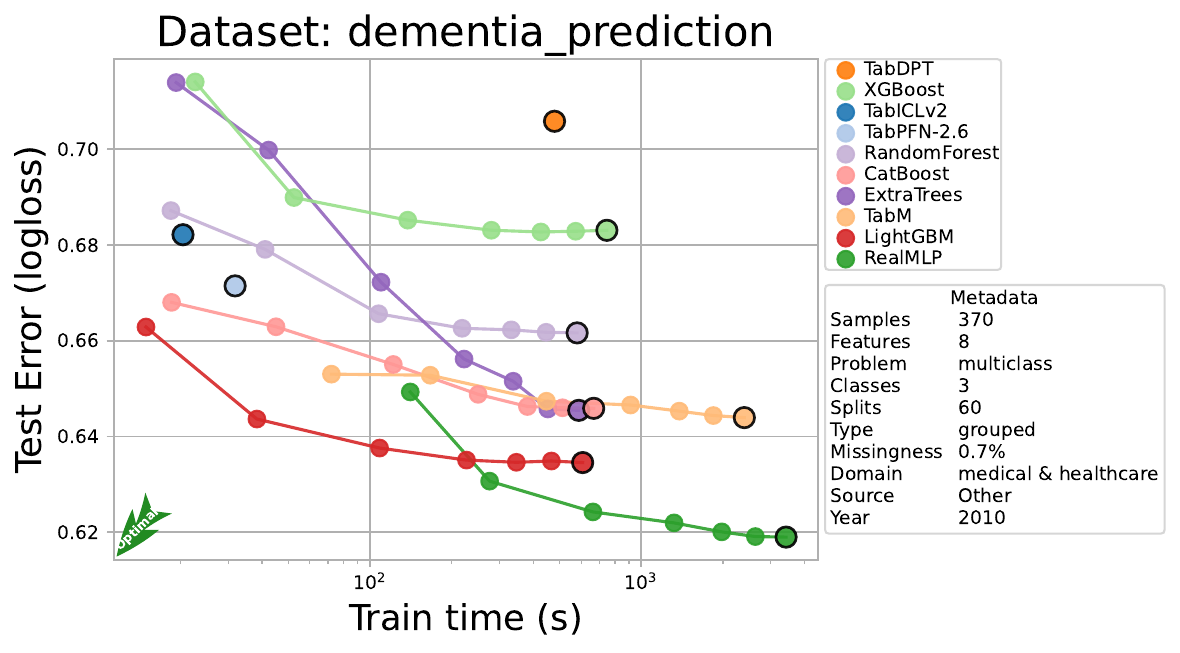}
  \end{minipage}

  \captionof{figure}{\textbf{dementia\_prediction}: per-method test error (left) and HPO Pareto trajectory (right).}
  \label{fig:perdataset_dementia_prediction}
\end{center}
\space
\begin{center}
  \begin{minipage}[t]{0.48\textwidth}
    \centering
    \vspace{0pt}
    \scriptsize
\begin{tabular}{l@{\quad}lll}
\toprule
 & Default & Tuned & Tuned + Ens. \\
\midrule
Linear & 0.352 $\pm$ 0.002 & 0.342 $\pm$ 0.002 & 0.342 $\pm$ 0.002 \\
RandomForest & 0.364 $\pm$ 0.002 & 0.346 $\pm$ 0.002 & 0.344 $\pm$ 0.003 \\
ExtraTrees & 0.369 $\pm$ 0.003 & 0.350 $\pm$ 0.003 & 0.347 $\pm$ 0.003 \\
CatBoost & \underline{0.328 $\pm$ 0.003} & \underline{\textbf{0.327 $\pm$ 0.003}} & \underline{\textbf{\textcolor{green!50!black}{0.327 $\pm$ 0.003}}} \\
LightGBM & 0.350 $\pm$ 0.003 & 0.338 $\pm$ 0.002 & 0.333 $\pm$ 0.002 \\
XGBoost & 0.335 $\pm$ 0.003 & 0.330 $\pm$ 0.003 & 0.329 $\pm$ 0.003 \\
RealMLP & 0.341 $\pm$ 0.004 & \underline{0.330 $\pm$ 0.004} & 0.330 $\pm$ 0.002 \\
TabM & 0.339 $\pm$ 0.003 & 0.338 $\pm$ 0.003 & 0.337 $\pm$ 0.003 \\
TabDPT & 0.377 $\pm$ 0.005 & - & - \\
TabPFN-2.6 & 0.332 $\pm$ 0.002 & - & - \\
TabICLv2 & 0.335 $\pm$ 0.002 & - & - \\
\bottomrule
\end{tabular}
\space
  \end{minipage}\hfill
  \begin{minipage}[t]{0.48\textwidth}
    \centering
    \vspace{0pt}
    \includegraphics[width=\linewidth]{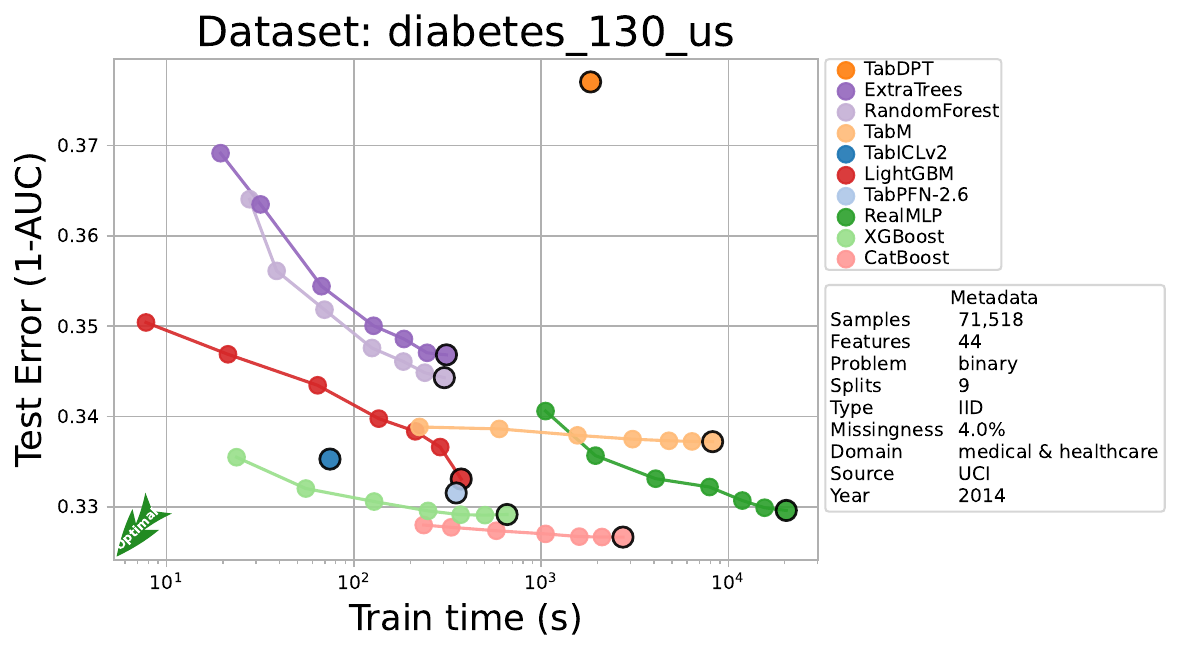}
  \end{minipage}

  \captionof{figure}{\textbf{diabetes\_130\_us}: per-method test error (left) and HPO Pareto trajectory (right).}
  \label{fig:perdataset_diabetes_130_us}
\end{center}
\space
\begin{center}
  \begin{minipage}[t]{0.48\textwidth}
    \centering
    \vspace{0pt}
    \scriptsize
\begin{tabular}{l@{\quad}lll}
\toprule
 & Default & Tuned & Tuned + Ens. \\
\midrule
Linear & 0.195 $\pm$ 0.036 & 0.202 $\pm$ 0.063 & 0.198 $\pm$ 0.056 \\
RandomForest & 0.092 $\pm$ 0.002 & 0.092 $\pm$ 0.002 & 0.092 $\pm$ 0.002 \\
ExtraTrees & 0.091 $\pm$ 0.001 & 0.090 $\pm$ 0.001 & 0.090 $\pm$ 0.001 \\
CatBoost & 0.084 $\pm$ 0.001 & 0.084 $\pm$ 0.001 & 0.084 $\pm$ 0.001 \\
LightGBM & 0.085 $\pm$ 0.001 & 0.085 $\pm$ 0.001 & 0.084 $\pm$ 0.001 \\
XGBoost & 0.085 $\pm$ 0.000 & 0.085 $\pm$ 0.001 & 0.085 $\pm$ 0.001 \\
RealMLP & 0.086 $\pm$ 0.001 & \underline{0.082 $\pm$ 0.001} & \underline{0.081 $\pm$ 0.001} \\
TabM & 0.082 $\pm$ 0.001 & \underline{0.081 $\pm$ 0.001} & \underline{0.081 $\pm$ 0.000} \\
TabDPT & 0.085 $\pm$ 0.002 & - & - \\
TabPFN-2.6 & 0.082 $\pm$ 0.001 & - & - \\
TabICLv2 & \underline{\textbf{\textcolor{green!50!black}{0.080 $\pm$ 0.001}}} & - & - \\
\bottomrule
\end{tabular}
\space
  \end{minipage}\hfill
  \begin{minipage}[t]{0.48\textwidth}
    \centering
    \vspace{0pt}
    \includegraphics[width=\linewidth]{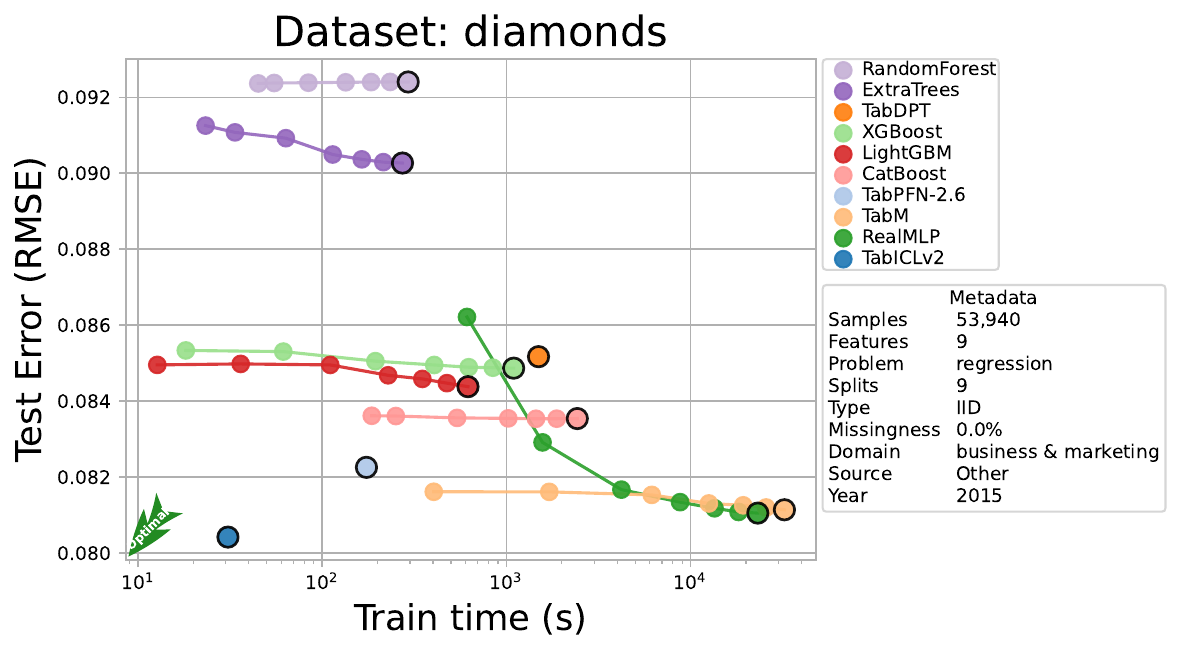}
  \end{minipage}

  \captionof{figure}{\textbf{diamonds}: per-method test error (left) and HPO Pareto trajectory (right).}
  \label{fig:perdataset_diamonds}
\end{center}
\space
\begin{center}
  \begin{minipage}[t]{0.48\textwidth}
    \centering
    \vspace{0pt}
    \scriptsize
\begin{tabular}{l@{\quad}lll}
\toprule
 & Default & Tuned & Tuned + Ens. \\
\midrule
Linear & 0.245 $\pm$ 0.030 & 0.247 $\pm$ 0.034 & 0.234 $\pm$ 0.031 \\
RandomForest & 0.174 $\pm$ 0.032 & 0.174 $\pm$ 0.032 & 0.173 $\pm$ 0.033 \\
ExtraTrees & 0.154 $\pm$ 0.032 & \underline{0.153 $\pm$ 0.032} & \underline{0.154 $\pm$ 0.032} \\
CatBoost & 0.164 $\pm$ 0.031 & 0.158 $\pm$ 0.031 & \underline{0.156 $\pm$ 0.031} \\
LightGBM & 0.187 $\pm$ 0.035 & 0.169 $\pm$ 0.034 & 0.170 $\pm$ 0.034 \\
XGBoost & 0.181 $\pm$ 0.032 & 0.181 $\pm$ 0.035 & 0.176 $\pm$ 0.033 \\
RealMLP & 0.188 $\pm$ 0.033 & 0.170 $\pm$ 0.033 & 0.159 $\pm$ 0.028 \\
TabM & 0.177 $\pm$ 0.033 & 0.170 $\pm$ 0.029 & 0.167 $\pm$ 0.029 \\
TabDPT & 0.182 $\pm$ 0.035 & - & - \\
TabPFN-2.6 & 0.150 $\pm$ 0.027 & - & - \\
TabICLv2 & \underline{\textbf{\textcolor{green!50!black}{0.140 $\pm$ 0.028}}} & - & - \\
\bottomrule
\end{tabular}
\space
  \end{minipage}\hfill
  \begin{minipage}[t]{0.48\textwidth}
    \centering
    \vspace{0pt}
    \includegraphics[width=\linewidth]{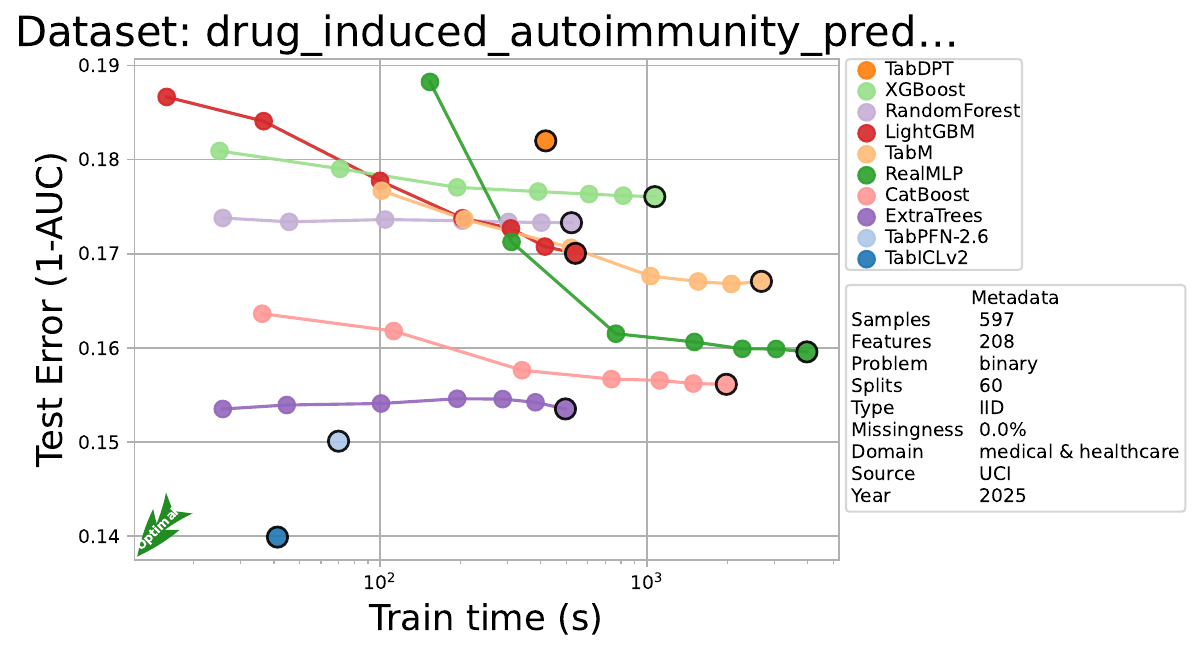}
  \end{minipage}

  \captionof{figure}{\textbf{drug\_induced\_autoimmunity\_prediction}: per-method test error (left) and HPO Pareto trajectory (right).}
  \label{fig:perdataset_drug_induced_autoimmunity_prediction}
\end{center}
\space
\begin{center}
  \begin{minipage}[t]{0.48\textwidth}
    \centering
    \vspace{0pt}
    \scriptsize
\begin{tabular}{l@{\quad}lll}
\toprule
 & Default & Tuned & Tuned + Ens. \\
\midrule
Linear & 17.06 $\pm$ 3.54 & 13.54 $\pm$ 0.60 & 13.33 $\pm$ 0.30 \\
RandomForest & 13.06 $\pm$ 0.13 & 13.04 $\pm$ 0.14 & 13.01 $\pm$ 0.14 \\
ExtraTrees & 13.01 $\pm$ 0.13 & 12.97 $\pm$ 0.12 & 12.96 $\pm$ 0.13 \\
CatBoost & \underline{12.71 $\pm$ 0.13} & \underline{12.66 $\pm$ 0.13} & \underline{\textbf{\textcolor{green!50!black}{12.64 $\pm$ 0.12}}} \\
LightGBM & 12.86 $\pm$ 0.11 & 12.72 $\pm$ 0.10 & 12.71 $\pm$ 0.10 \\
XGBoost & 12.79 $\pm$ 0.13 & 12.79 $\pm$ 0.13 & 12.78 $\pm$ 0.13 \\
RealMLP & 12.91 $\pm$ 0.10 & 12.76 $\pm$ 0.10 & 12.73 $\pm$ 0.10 \\
TabM & 12.80 $\pm$ 0.11 & 12.81 $\pm$ 0.11 & 12.78 $\pm$ 0.11 \\
TabDPT & 13.22 $\pm$ 0.17 & - & - \\
TabPFN-2.6 & 12.96 $\pm$ 0.12 & - & - \\
TabICLv2 & 12.88 $\pm$ 0.11 & - & - \\
\bottomrule
\end{tabular}
\space
  \end{minipage}\hfill
  \begin{minipage}[t]{0.48\textwidth}
    \centering
    \vspace{0pt}
    \includegraphics[width=\linewidth]{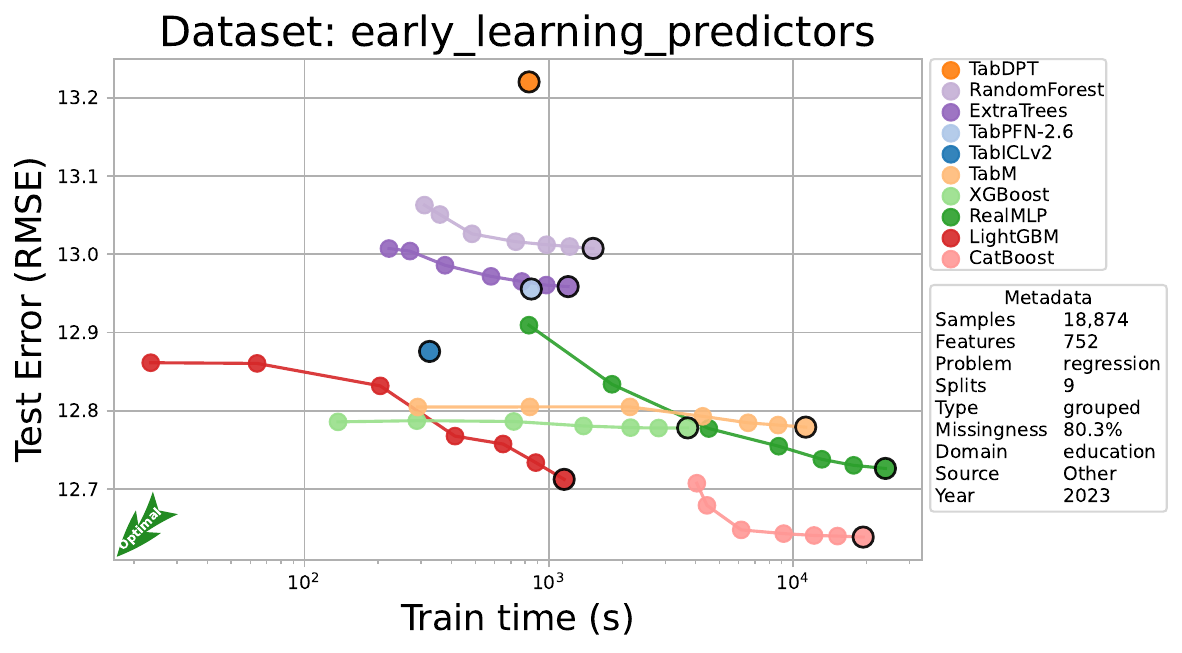}
  \end{minipage}

  \captionof{figure}{\textbf{early\_learning\_predictors}: per-method test error (left) and HPO Pareto trajectory (right).}
  \label{fig:perdataset_early_learning_predictors}
\end{center}
\space
\begin{center}
  \begin{minipage}[t]{0.48\textwidth}
    \centering
    \vspace{0pt}
    \scriptsize
\begin{tabular}{l@{\quad}lll}
\toprule
 & Default & Tuned & Tuned + Ens. \\
\midrule
Linear & 0.053 $\pm$ 0.018 & 0.050 $\pm$ 0.017 & 0.050 $\pm$ 0.017 \\
RandomForest & 0.028 $\pm$ 0.014 & \underline{0.029 $\pm$ 0.014} & 0.029 $\pm$ 0.014 \\
ExtraTrees & 0.027 $\pm$ 0.014 & \underline{0.029 $\pm$ 0.015} & \underline{0.029 $\pm$ 0.014} \\
CatBoost & 0.028 $\pm$ 0.014 & \underline{0.028 $\pm$ 0.013} & \underline{0.028 $\pm$ 0.013} \\
LightGBM & 0.041 $\pm$ 0.017 & 0.038 $\pm$ 0.017 & 0.038 $\pm$ 0.016 \\
XGBoost & 0.041 $\pm$ 0.016 & 0.037 $\pm$ 0.016 & 0.036 $\pm$ 0.016 \\
RealMLP & 0.037 $\pm$ 0.017 & 0.032 $\pm$ 0.015 & 0.033 $\pm$ 0.015 \\
TabM & 0.040 $\pm$ 0.017 & 0.037 $\pm$ 0.018 & 0.038 $\pm$ 0.017 \\
TabDPT & 0.033 $\pm$ 0.015 & - & - \\
TabPFN-2.6 & 0.031 $\pm$ 0.013 & - & - \\
TabICLv2 & \underline{\textbf{\textcolor{green!50!black}{0.025 $\pm$ 0.012}}} & - & - \\
\bottomrule
\end{tabular}
\space
  \end{minipage}\hfill
  \begin{minipage}[t]{0.48\textwidth}
    \centering
    \vspace{0pt}
    \includegraphics[width=\linewidth]{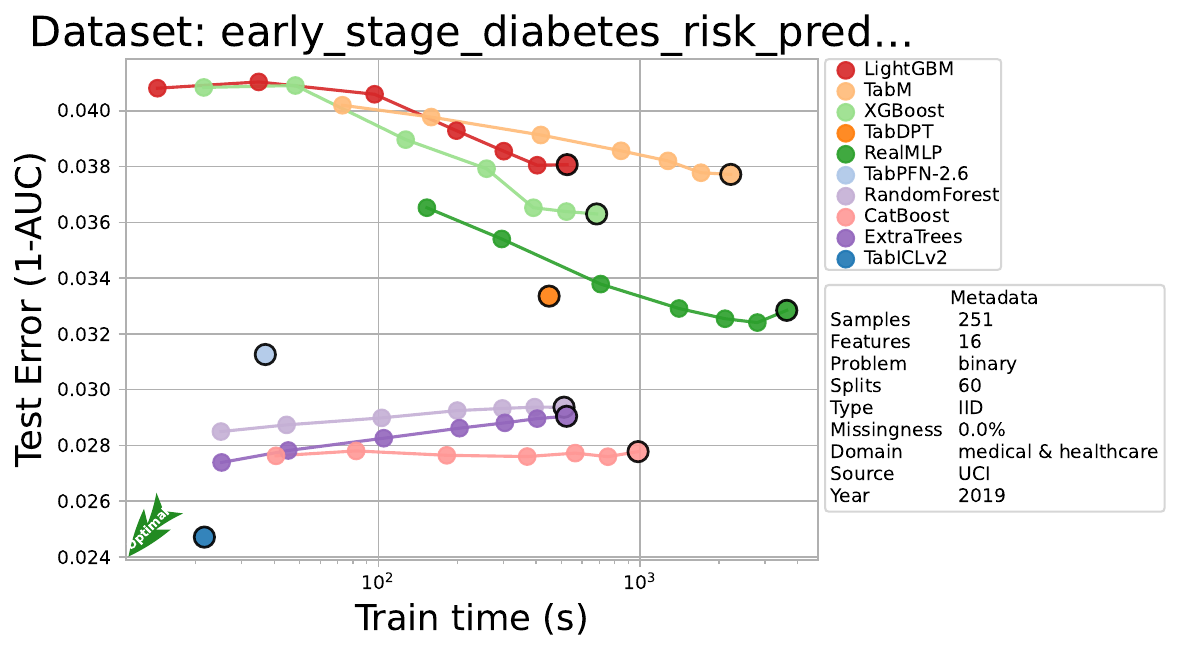}
  \end{minipage}

  \captionof{figure}{\textbf{early\_stage\_diabetes\_risk\_prediction}: per-method test error (left) and HPO Pareto trajectory (right).}
  \label{fig:perdataset_early_stage_diabetes_risk_prediction}
\end{center}
\space
\begin{center}
  \begin{minipage}[t]{0.48\textwidth}
    \centering
    \vspace{0pt}
    \scriptsize
\begin{tabular}{l@{\quad}lll}
\toprule
 & Default & Tuned & Tuned + Ens. \\
\midrule
Linear & 0.386 $\pm$ 0.057 & 0.388 $\pm$ 0.063 & 0.384 $\pm$ 0.062 \\
RandomForest & 0.396 $\pm$ 0.054 & 0.404 $\pm$ 0.064 & 0.401 $\pm$ 0.061 \\
ExtraTrees & 0.420 $\pm$ 0.038 & 0.394 $\pm$ 0.055 & 0.392 $\pm$ 0.055 \\
CatBoost & 0.385 $\pm$ 0.060 & \underline{\textbf{0.369 $\pm$ 0.056}} & \underline{\textbf{0.368 $\pm$ 0.055}} \\
LightGBM & 0.408 $\pm$ 0.059 & 0.384 $\pm$ 0.058 & 0.382 $\pm$ 0.054 \\
XGBoost & 0.443 $\pm$ 0.059 & 0.405 $\pm$ 0.053 & 0.406 $\pm$ 0.054 \\
RealMLP & 0.405 $\pm$ 0.097 & 0.382 $\pm$ 0.071 & \underline{0.371 $\pm$ 0.067} \\
TabM & 0.386 $\pm$ 0.059 & 0.395 $\pm$ 0.062 & 0.391 $\pm$ 0.063 \\
TabDPT & 0.372 $\pm$ 0.065 & - & - \\
TabPFN-2.6 & 0.393 $\pm$ 0.077 & - & - \\
TabICLv2 & \underline{\textbf{\textcolor{green!50!black}{0.362 $\pm$ 0.076}}} & - & - \\
\bottomrule
\end{tabular}
\space
  \end{minipage}\hfill
  \begin{minipage}[t]{0.48\textwidth}
    \centering
    \vspace{0pt}
    \includegraphics[width=\linewidth]{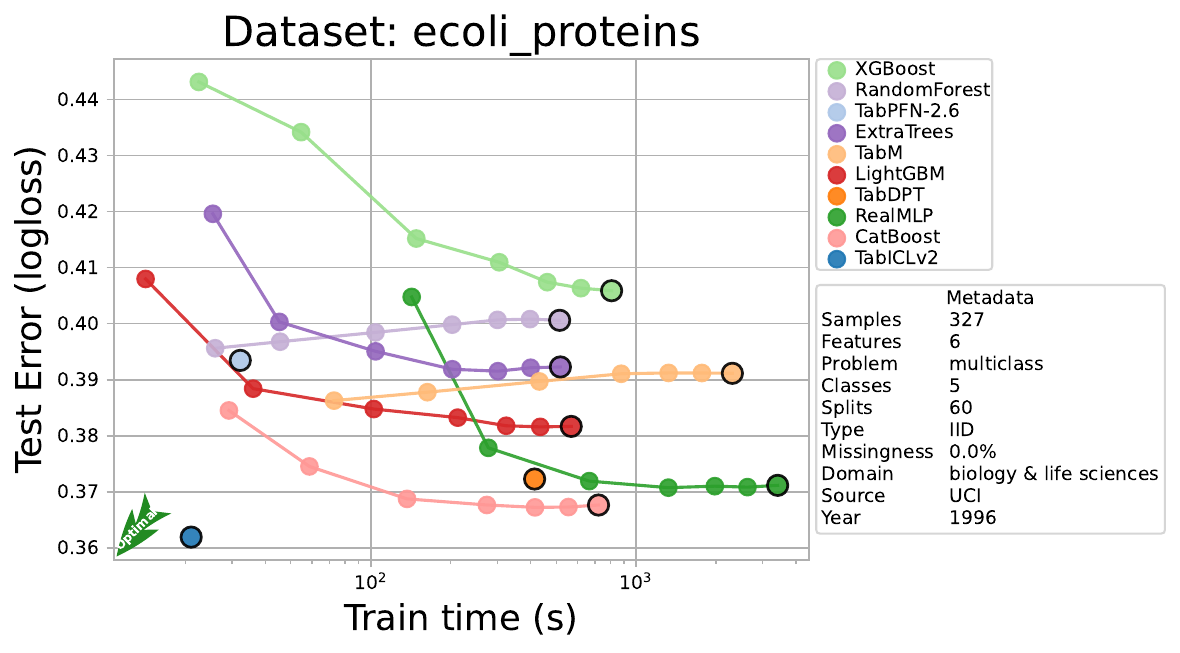}
  \end{minipage}

  \captionof{figure}{\textbf{ecoli\_proteins}: per-method test error (left) and HPO Pareto trajectory (right).}
  \label{fig:perdataset_ecoli_proteins}
\end{center}
\space
\begin{center}
  \begin{minipage}[t]{0.48\textwidth}
    \centering
    \vspace{0pt}
    \scriptsize
\begin{tabular}{l@{\quad}lll}
\toprule
 & Default & Tuned & Tuned + Ens. \\
\midrule
Linear & 0.295 $\pm$ 0.010 & 0.278 $\pm$ 0.007 & 0.276 $\pm$ 0.008 \\
RandomForest & \underline{\textbf{0.256 $\pm$ 0.007}} & \underline{\textbf{\textcolor{green!50!black}{0.253 $\pm$ 0.008}}} & \underline{\textbf{0.254 $\pm$ 0.006}} \\
ExtraTrees & \underline{0.258 $\pm$ 0.008} & \underline{\textbf{0.256 $\pm$ 0.009}} & \underline{\textbf{0.255 $\pm$ 0.008}} \\
CatBoost & \underline{\textbf{0.253 $\pm$ 0.006}} & \underline{\textbf{0.255 $\pm$ 0.007}} & \underline{\textbf{0.255 $\pm$ 0.006}} \\
LightGBM & \underline{\textbf{0.255 $\pm$ 0.006}} & \underline{\textbf{0.256 $\pm$ 0.004}} & \underline{\textbf{0.256 $\pm$ 0.005}} \\
XGBoost & \underline{\textbf{0.255 $\pm$ 0.005}} & \underline{\textbf{0.255 $\pm$ 0.004}} & \underline{\textbf{0.255 $\pm$ 0.005}} \\
RealMLP & \underline{\textbf{0.258 $\pm$ 0.006}} & \underline{\textbf{0.256 $\pm$ 0.006}} & \underline{\textbf{0.253 $\pm$ 0.007}} \\
TabM & \underline{\textbf{0.254 $\pm$ 0.008}} & \underline{\textbf{0.255 $\pm$ 0.008}} & \underline{\textbf{0.254 $\pm$ 0.006}} \\
TabDPT & 0.259 $\pm$ 0.007 & - & - \\
TabPFN-2.6 & \underline{\textbf{0.254 $\pm$ 0.008}} & - & - \\
TabICLv2 & \underline{\textbf{0.253 $\pm$ 0.008}} & - & - \\
\bottomrule
\end{tabular}
\space
  \end{minipage}\hfill
  \begin{minipage}[t]{0.48\textwidth}
    \centering
    \vspace{0pt}
    \includegraphics[width=\linewidth]{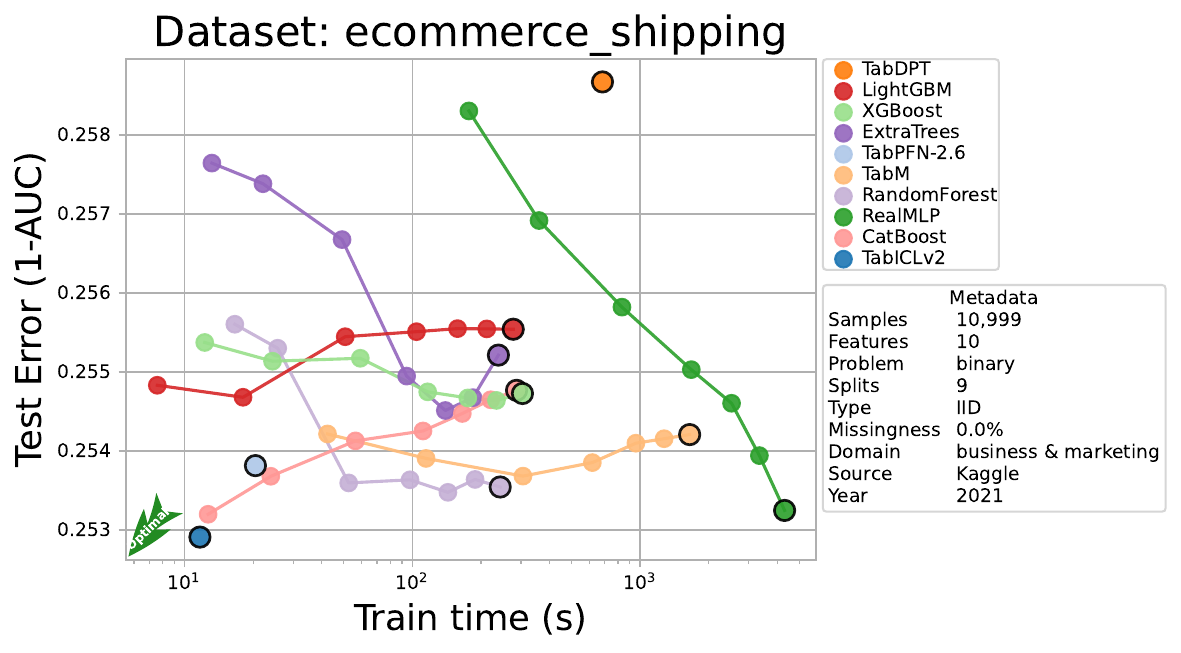}
  \end{minipage}

  \captionof{figure}{\textbf{ecommerce\_shipping}: per-method test error (left) and HPO Pareto trajectory (right).}
  \label{fig:perdataset_ecommerce_shipping}
\end{center}
\space
\begin{center}
  \begin{minipage}[t]{0.48\textwidth}
    \centering
    \vspace{0pt}
    \scriptsize
\begin{tabular}{l@{\quad}lll}
\toprule
 & Default & Tuned & Tuned + Ens. \\
\midrule
Linear & 3.053 & 2.221 & 2.208 \\
RandomForest & 3.390 & 2.920 & 2.947 \\
ExtraTrees & 2.464 & 2.460 & 2.475 \\
CatBoost & 2.493 & 2.548 & 2.480 \\
LightGBM & 2.694 & 2.428 & 2.358 \\
XGBoost & 2.835 & 2.520 & 2.523 \\
RealMLP & 1.858 & \textcolor{green!50!black}{1.702} & 1.772 \\
TabM & 2.151 & 2.059 & 2.039 \\
TabDPT & - & - & - \\
TabPFN-2.6 & - & - & - \\
TabICLv2 & 2.454 & - & - \\
\bottomrule
\end{tabular}
\space
  \end{minipage}\hfill
  \begin{minipage}[t]{0.48\textwidth}
    \centering
    \vspace{0pt}
    \includegraphics[width=\linewidth]{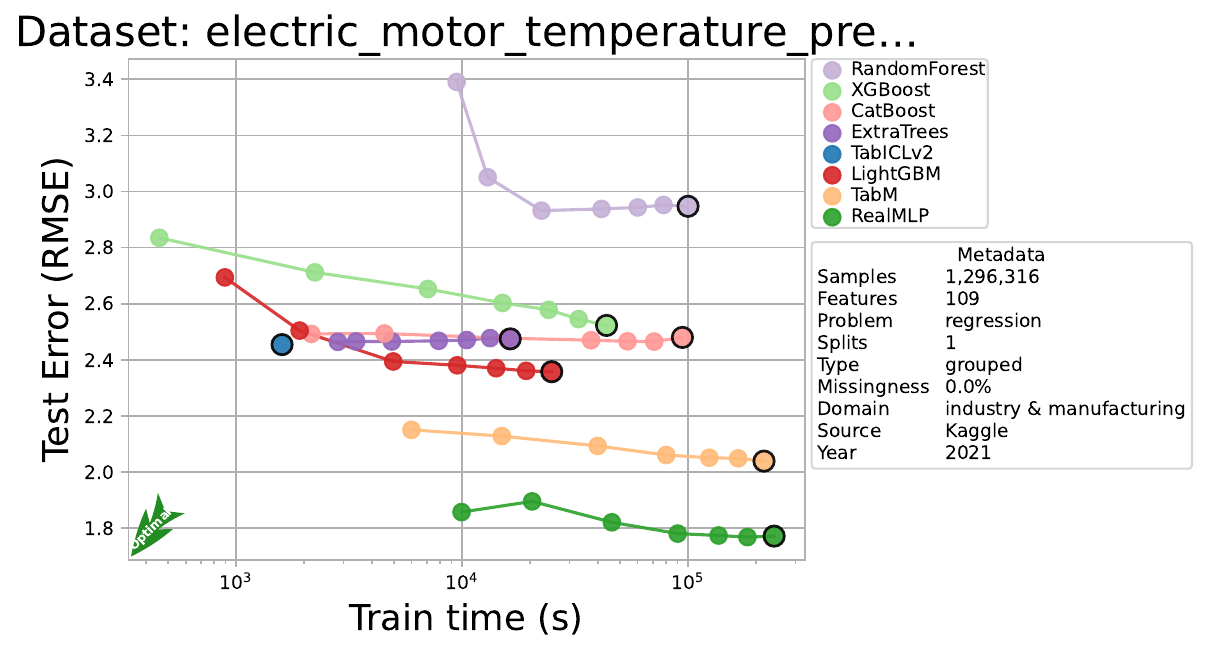}
  \end{minipage}

  \captionof{figure}{\textbf{electric\_motor\_temperature\_prediction}: per-method test error (left) and HPO Pareto trajectory (right).}
  \label{fig:perdataset_electric_motor_temperature_prediction}
\end{center}
\space
\begin{center}
  \begin{minipage}[t]{0.48\textwidth}
    \centering
    \vspace{0pt}
    \scriptsize
\begin{tabular}{l@{\quad}lll}
\toprule
 & Default & Tuned & Tuned + Ens. \\
\midrule
Linear & 0.035 $\pm$ 0.003 & 0.036 $\pm$ 0.004 & 0.034 $\pm$ 0.003 \\
RandomForest & 0.013 $\pm$ 0.001 & 0.013 $\pm$ 0.001 & 0.013 $\pm$ 0.001 \\
ExtraTrees & 0.013 $\pm$ 0.001 & 0.013 $\pm$ 0.002 & 0.011 $\pm$ 0.002 \\
CatBoost & 0.012 $\pm$ 0.001 & 0.012 $\pm$ 0.002 & 0.012 $\pm$ 0.002 \\
LightGBM & 0.011 $\pm$ 0.002 & 0.012 $\pm$ 0.002 & 0.010 $\pm$ 0.002 \\
XGBoost & 0.012 $\pm$ 0.002 & 0.011 $\pm$ 0.002 & 0.011 $\pm$ 0.002 \\
RealMLP & 0.012 $\pm$ 0.004 & \underline{0.010 $\pm$ 0.003} & \underline{0.007 $\pm$ 0.002} \\
TabM & 0.012 $\pm$ 0.002 & \underline{0.010 $\pm$ 0.002} & 0.010 $\pm$ 0.002 \\
TabDPT & 0.011 $\pm$ 0.002 & - & - \\
TabPFN-2.6 & \underline{\textbf{\textcolor{green!50!black}{0.005 $\pm$ 0.001}}} & - & - \\
TabICLv2 & 0.006 $\pm$ 0.001 & - & - \\
\bottomrule
\end{tabular}
\space
  \end{minipage}\hfill
  \begin{minipage}[t]{0.48\textwidth}
    \centering
    \vspace{0pt}
    \includegraphics[width=\linewidth]{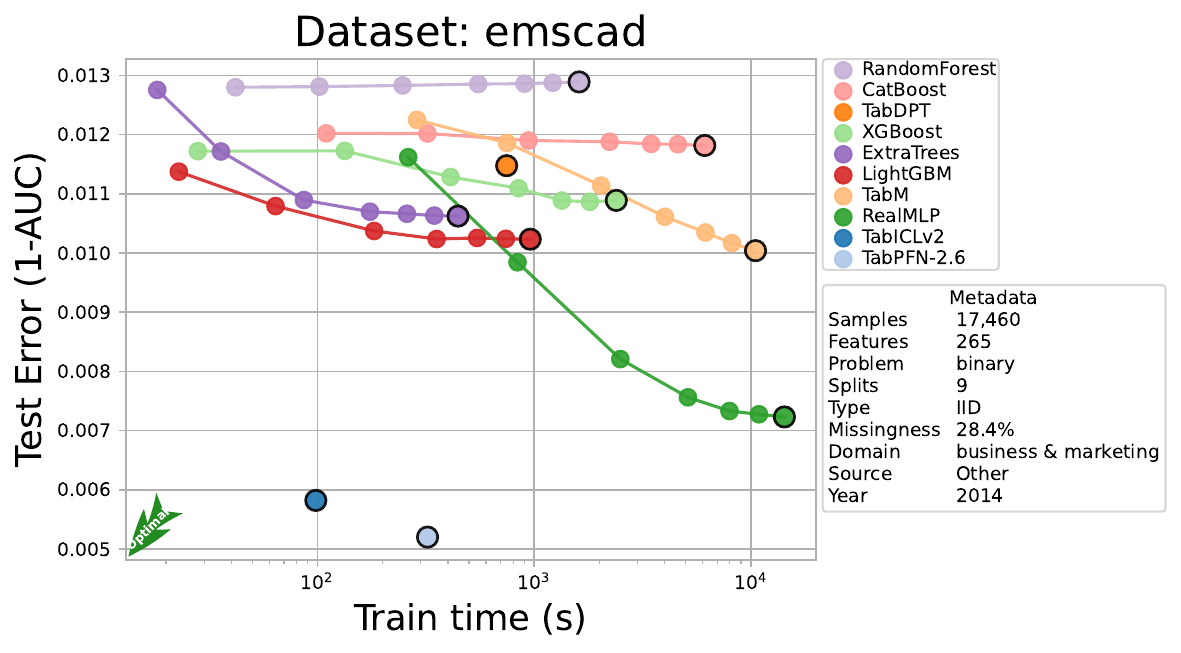}
  \end{minipage}

  \captionof{figure}{\textbf{emscad}: per-method test error (left) and HPO Pareto trajectory (right).}
  \label{fig:perdataset_emscad}
\end{center}
\space
\begin{center}
  \begin{minipage}[t]{0.48\textwidth}
    \centering
    \vspace{0pt}
    \scriptsize
\begin{tabular}{l@{\quad}lll}
\toprule
 & Default & Tuned & Tuned + Ens. \\
\midrule
Linear & 0.109 $\pm$ 0.016 & \underline{0.066 $\pm$ 0.025} & \underline{0.065 $\pm$ 0.025} \\
RandomForest & 0.158 $\pm$ 0.014 & 0.107 $\pm$ 0.027 & 0.107 $\pm$ 0.027 \\
ExtraTrees & 0.172 $\pm$ 0.013 & 0.095 $\pm$ 0.020 & 0.096 $\pm$ 0.019 \\
CatBoost & 0.084 $\pm$ 0.031 & \underline{0.069 $\pm$ 0.025} & \underline{0.069 $\pm$ 0.025} \\
LightGBM & 0.849 $\pm$ 0.066 & 0.069 $\pm$ 0.019 & 0.070 $\pm$ 0.019 \\
XGBoost & 0.121 $\pm$ 0.034 & 0.092 $\pm$ 0.023 & 0.092 $\pm$ 0.024 \\
RealMLP & 0.071 $\pm$ 0.018 & \underline{0.066 $\pm$ 0.019} & \underline{0.065 $\pm$ 0.019} \\
TabM & 0.074 $\pm$ 0.014 & 0.075 $\pm$ 0.015 & 0.075 $\pm$ 0.014 \\
TabDPT & 0.083 $\pm$ 0.021 & - & - \\
TabPFN-2.6 & 0.066 $\pm$ 0.029 & - & - \\
TabICLv2 & \underline{\textbf{\textcolor{green!50!black}{0.054 $\pm$ 0.020}}} & - & - \\
\bottomrule
\end{tabular}
\space
  \end{minipage}\hfill
  \begin{minipage}[t]{0.48\textwidth}
    \centering
    \vspace{0pt}
    \includegraphics[width=\linewidth]{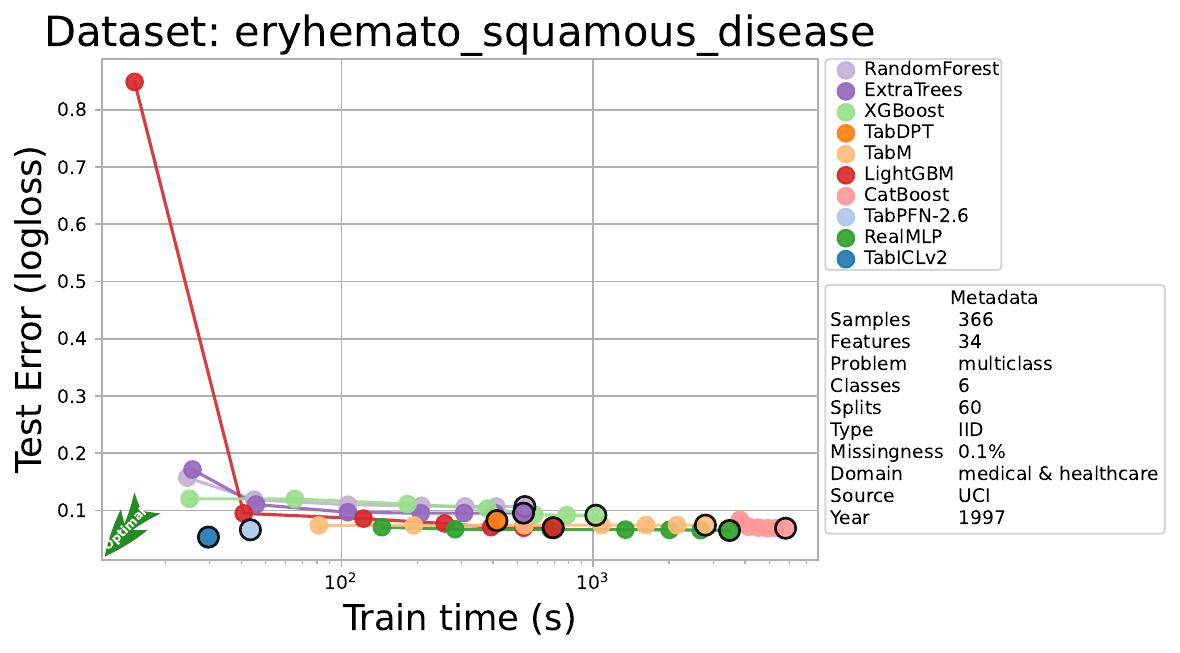}
  \end{minipage}

  \captionof{figure}{\textbf{eryhemato\_squamous\_disease}: per-method test error (left) and HPO Pareto trajectory (right).}
  \label{fig:perdataset_eryhemato_squamous_disease}
\end{center}
\space
\begin{center}
  \begin{minipage}[t]{0.48\textwidth}
    \centering
    \vspace{0pt}
    \scriptsize
\begin{tabular}{l@{\quad}lll}
\toprule
 & Default & Tuned & Tuned + Ens. \\
\midrule
Linear & 791.7 $\pm$ 27.5 & 763.9 $\pm$ 24.5 & 764.1 $\pm$ 24.5 \\
RandomForest & 745.0 $\pm$ 28.0 & \underline{733.0 $\pm$ 26.9} & 731.5 $\pm$ 26.9 \\
ExtraTrees & 738.3 $\pm$ 29.6 & 735.9 $\pm$ 29.7 & 734.3 $\pm$ 28.7 \\
CatBoost & 737.3 $\pm$ 27.0 & \underline{733.7 $\pm$ 25.3} & 730.2 $\pm$ 26.2 \\
LightGBM & 742.5 $\pm$ 26.7 & 740.9 $\pm$ 25.0 & 731.2 $\pm$ 26.3 \\
XGBoost & 751.9 $\pm$ 29.7 & 736.8 $\pm$ 26.1 & 733.6 $\pm$ 26.6 \\
RealMLP & 755.3 $\pm$ 27.0 & \underline{730.6 $\pm$ 29.1} & \underline{723.5 $\pm$ 28.3} \\
TabM & 750.9 $\pm$ 28.6 & 753.0 $\pm$ 28.1 & 748.1 $\pm$ 29.1 \\
TabDPT & 720.1 $\pm$ 30.1 & - & - \\
TabPFN-2.6 & 726.8 $\pm$ 27.9 & - & - \\
TabICLv2 & \underline{\textbf{\textcolor{green!50!black}{714.3 $\pm$ 30.7}}} & - & - \\
\bottomrule
\end{tabular}
\space
  \end{minipage}\hfill
  \begin{minipage}[t]{0.48\textwidth}
    \centering
    \vspace{0pt}
    \includegraphics[width=\linewidth]{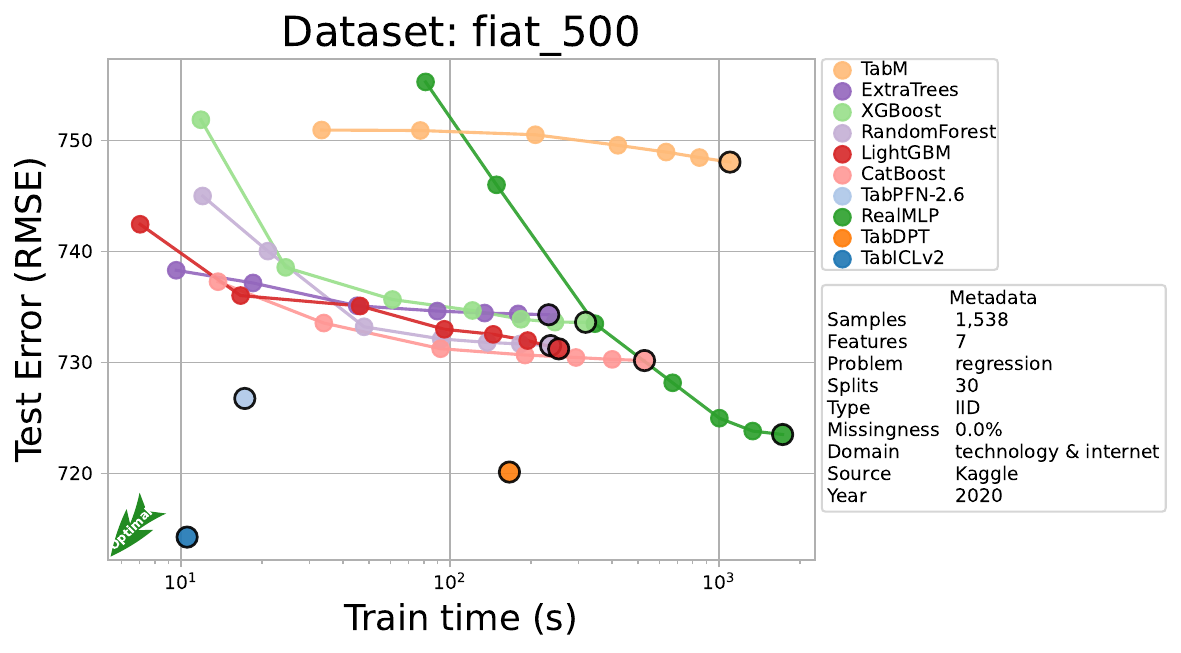}
  \end{minipage}

  \captionof{figure}{\textbf{fiat\_500}: per-method test error (left) and HPO Pareto trajectory (right).}
  \label{fig:perdataset_fiat_500}
\end{center}
\space
\begin{center}
  \begin{minipage}[t]{0.48\textwidth}
    \centering
    \vspace{0pt}
    \scriptsize
\begin{tabular}{l@{\quad}lll}
\toprule
 & Default & Tuned & Tuned + Ens. \\
\midrule
Linear & \underline{\textbf{0.181 $\pm$ 0.018}} & \underline{\textbf{0.181 $\pm$ 0.018}} & \underline{\textbf{0.181 $\pm$ 0.018}} \\
RandomForest & 0.219 $\pm$ 0.018 & 0.206 $\pm$ 0.018 & 0.205 $\pm$ 0.018 \\
ExtraTrees & 0.224 $\pm$ 0.016 & 0.185 $\pm$ 0.019 & 0.187 $\pm$ 0.018 \\
CatBoost & 0.185 $\pm$ 0.018 & 0.187 $\pm$ 0.018 & 0.187 $\pm$ 0.018 \\
LightGBM & 0.206 $\pm$ 0.018 & 0.183 $\pm$ 0.019 & 0.183 $\pm$ 0.019 \\
XGBoost & 0.202 $\pm$ 0.016 & 0.191 $\pm$ 0.018 & 0.192 $\pm$ 0.018 \\
RealMLP & 0.188 $\pm$ 0.017 & 0.184 $\pm$ 0.018 & \underline{0.182 $\pm$ 0.017} \\
TabM & 0.183 $\pm$ 0.018 & \underline{0.183 $\pm$ 0.018} & \underline{0.182 $\pm$ 0.018} \\
TabDPT & \underline{\textbf{0.180 $\pm$ 0.018}} & - & - \\
TabPFN-2.6 & \underline{\textbf{\textcolor{green!50!black}{0.179 $\pm$ 0.018}}} & - & - \\
TabICLv2 & 0.180 $\pm$ 0.017 & - & - \\
\bottomrule
\end{tabular}
\space
  \end{minipage}\hfill
  \begin{minipage}[t]{0.48\textwidth}
    \centering
    \vspace{0pt}
    \includegraphics[width=\linewidth]{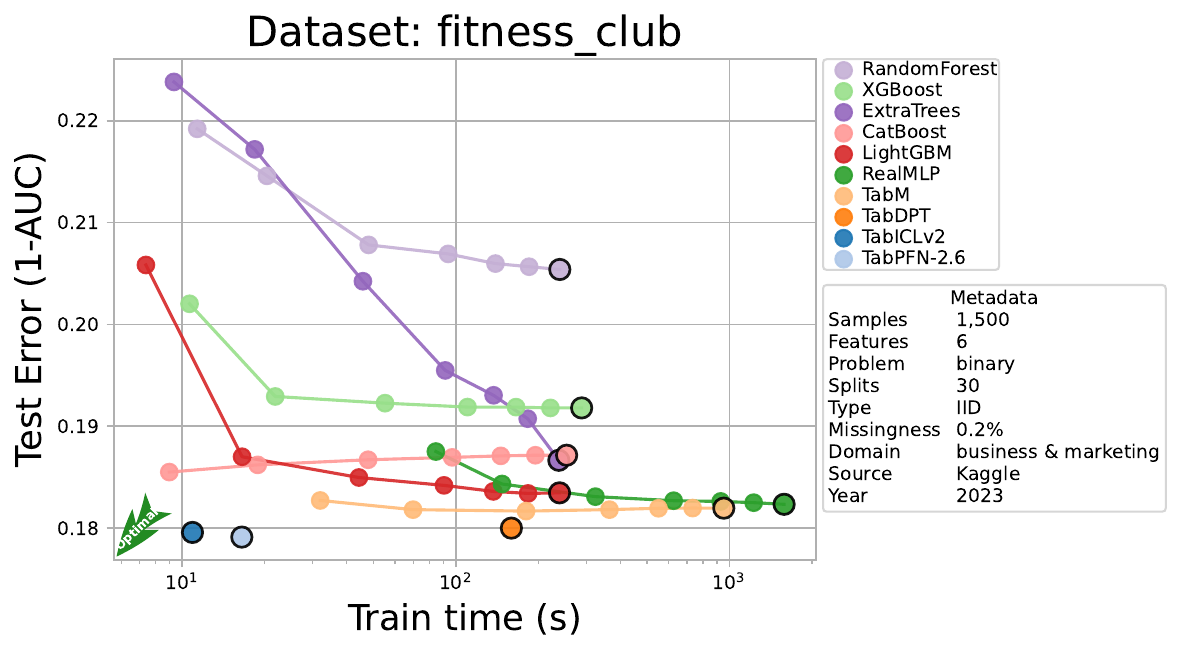}
  \end{minipage}

  \captionof{figure}{\textbf{fitness\_club}: per-method test error (left) and HPO Pareto trajectory (right).}
  \label{fig:perdataset_fitness_club}
\end{center}
\space
\begin{center}
  \begin{minipage}[t]{0.48\textwidth}
    \centering
    \vspace{0pt}
    \scriptsize
\begin{tabular}{l@{\quad}lll}
\toprule
 & Default & Tuned & Tuned + Ens. \\
\midrule
Linear & 8.557 $\pm$ 0.046 & 8.333 $\pm$ 0.068 & 8.295 $\pm$ 0.066 \\
RandomForest & 7.774 $\pm$ 0.042 & 7.586 $\pm$ 0.049 & 7.589 $\pm$ 0.047 \\
ExtraTrees & 8.108 $\pm$ 0.044 & 7.778 $\pm$ 0.058 & 7.762 $\pm$ 0.057 \\
CatBoost & \underline{7.376 $\pm$ 0.055} & \underline{\textbf{7.366 $\pm$ 0.055}} & \underline{\textbf{\textcolor{green!50!black}{7.366 $\pm$ 0.054}}} \\
LightGBM & 7.618 $\pm$ 0.056 & 7.513 $\pm$ 0.056 & 7.461 $\pm$ 0.054 \\
XGBoost & 7.392 $\pm$ 0.057 & 7.392 $\pm$ 0.057 & 7.396 $\pm$ 0.056 \\
RealMLP & 7.925 $\pm$ 0.062 & 7.452 $\pm$ 0.057 & 7.448 $\pm$ 0.056 \\
TabM & 7.784 $\pm$ 0.045 & 7.675 $\pm$ 0.054 & 7.675 $\pm$ 0.054 \\
TabDPT & 7.461 $\pm$ 0.070 & - & - \\
TabPFN-2.6 & \underline{7.381 $\pm$ 0.057} & - & - \\
TabICLv2 & 7.593 $\pm$ 0.053 & - & - \\
\bottomrule
\end{tabular}
\space
  \end{minipage}\hfill
  \begin{minipage}[t]{0.48\textwidth}
    \centering
    \vspace{0pt}
    \includegraphics[width=\linewidth]{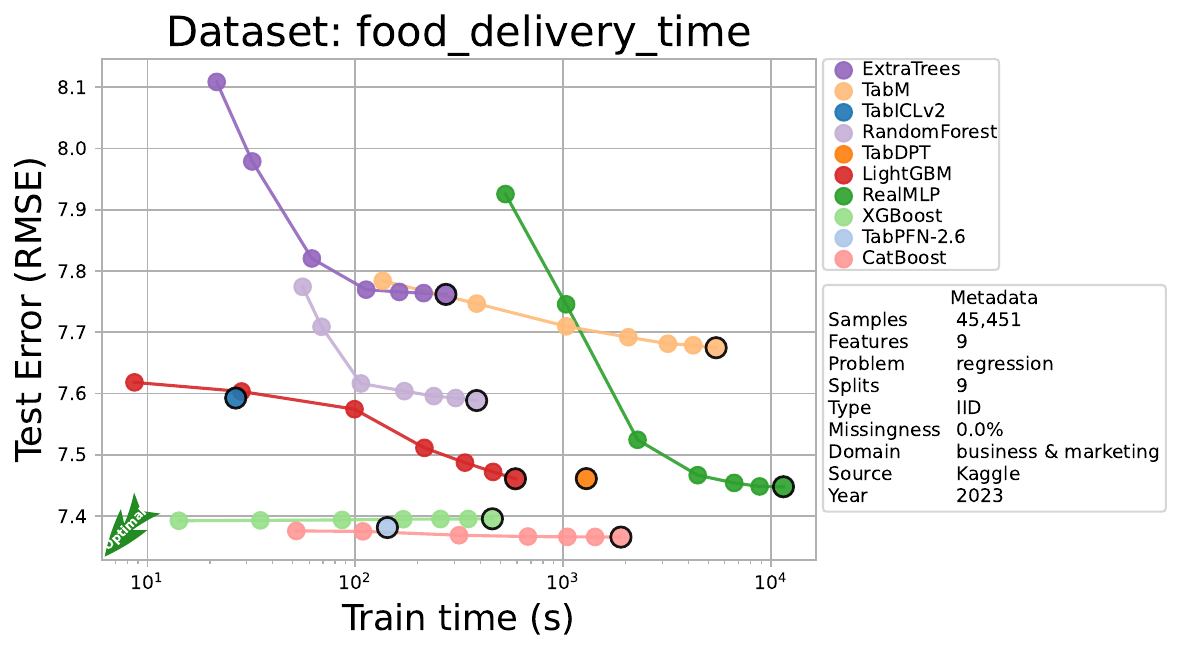}
  \end{minipage}

  \captionof{figure}{\textbf{food\_delivery\_time}: per-method test error (left) and HPO Pareto trajectory (right).}
  \label{fig:perdataset_food_delivery_time}
\end{center}
\space
\begin{center}
  \begin{minipage}[t]{0.48\textwidth}
    \centering
    \vspace{0pt}
    \scriptsize
\begin{tabular}{l@{\quad}lll}
\toprule
 & Default & Tuned & Tuned + Ens. \\
\midrule
Linear & 0.964 $\pm$ 0.129 & 0.988 $\pm$ 0.119 & 0.888 $\pm$ 0.101 \\
RandomForest & 0.669 $\pm$ 0.063 & 0.662 $\pm$ 0.071 & 0.655 $\pm$ 0.073 \\
ExtraTrees & 0.690 $\pm$ 0.055 & \underline{0.639 $\pm$ 0.079} & \underline{0.635 $\pm$ 0.072} \\
CatBoost & 0.636 $\pm$ 0.083 & \underline{0.626 $\pm$ 0.089} & \underline{0.623 $\pm$ 0.084} \\
LightGBM & 0.725 $\pm$ 0.070 & 0.676 $\pm$ 0.085 & 0.677 $\pm$ 0.084 \\
XGBoost & 0.746 $\pm$ 0.092 & 0.732 $\pm$ 0.082 & 0.726 $\pm$ 0.083 \\
RealMLP & 0.788 $\pm$ 0.124 & \underline{0.645 $\pm$ 0.105} & \underline{0.631 $\pm$ 0.084} \\
TabM & 0.680 $\pm$ 0.091 & 0.675 $\pm$ 0.096 & 0.662 $\pm$ 0.080 \\
TabDPT & 0.591 $\pm$ 0.088 & - & - \\
TabPFN-2.6 & \underline{\textbf{0.560 $\pm$ 0.098}} & - & - \\
TabICLv2 & \underline{\textbf{\textcolor{green!50!black}{0.552 $\pm$ 0.093}}} & - & - \\
\bottomrule
\end{tabular}
\space
  \end{minipage}\hfill
  \begin{minipage}[t]{0.48\textwidth}
    \centering
    \vspace{0pt}
    \includegraphics[width=\linewidth]{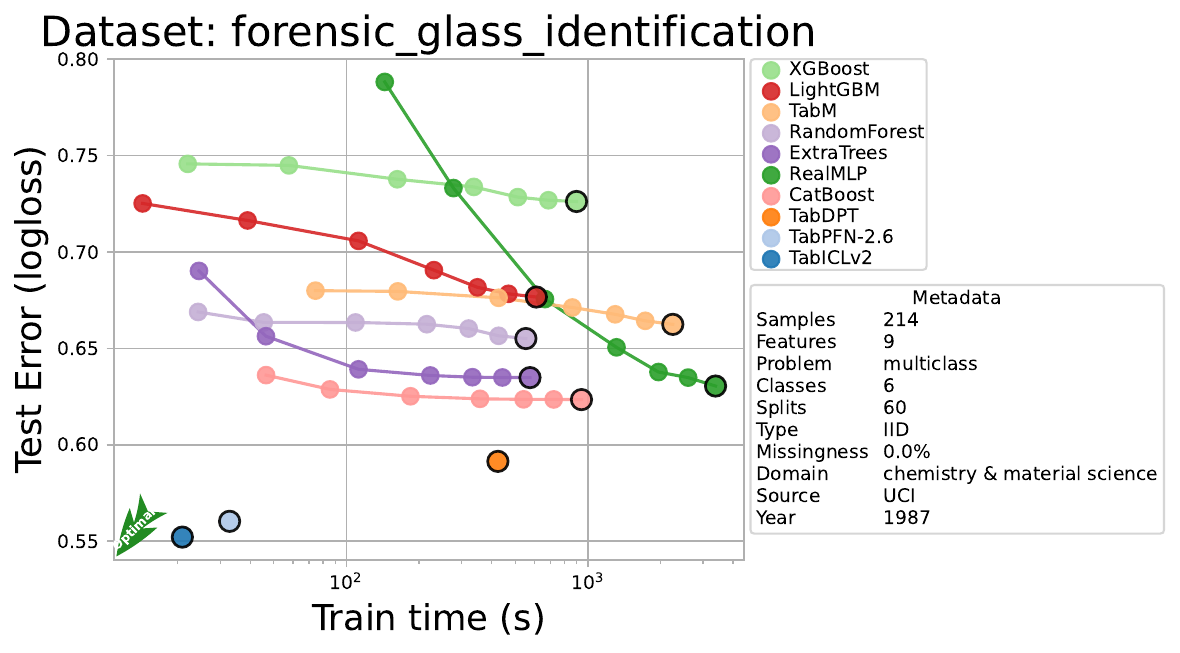}
  \end{minipage}

  \captionof{figure}{\textbf{forensic\_glass\_identification}: per-method test error (left) and HPO Pareto trajectory (right).}
  \label{fig:perdataset_forensic_glass_identification}
\end{center}
\space
\begin{center}
  \begin{minipage}[t]{0.48\textwidth}
    \centering
    \vspace{0pt}
    \scriptsize
\begin{tabular}{l@{\quad}lll}
\toprule
 & Default & Tuned & Tuned + Ens. \\
\midrule
Linear & 1.428 $\pm$ 0.075 & 1.415 $\pm$ 0.088 & \underline{1.404 $\pm$ 0.079} \\
RandomForest & 1.463 $\pm$ 0.070 & 1.440 $\pm$ 0.074 & 1.441 $\pm$ 0.073 \\
ExtraTrees & 1.444 $\pm$ 0.072 & 1.405 $\pm$ 0.080 & 1.407 $\pm$ 0.080 \\
CatBoost & 1.397 $\pm$ 0.080 & \underline{1.400 $\pm$ 0.079} & \underline{1.399 $\pm$ 0.080} \\
LightGBM & 1.399 $\pm$ 0.080 & 1.402 $\pm$ 0.080 & 1.401 $\pm$ 0.080 \\
XGBoost & 1.398 $\pm$ 0.079 & \underline{1.399 $\pm$ 0.080} & \underline{1.399 $\pm$ 0.080} \\
RealMLP & 1.404 $\pm$ 0.078 & 1.402 $\pm$ 0.080 & \underline{1.400 $\pm$ 0.079} \\
TabM & 1.399 $\pm$ 0.080 & \underline{1.400 $\pm$ 0.081} & \underline{1.400 $\pm$ 0.081} \\
TabDPT & \underline{\textbf{\textcolor{green!50!black}{1.392 $\pm$ 0.079}}} & - & - \\
TabPFN-2.6 & 1.403 $\pm$ 0.082 & - & - \\
TabICLv2 & 1.399 $\pm$ 0.079 & - & - \\
\bottomrule
\end{tabular}
\space
  \end{minipage}\hfill
  \begin{minipage}[t]{0.48\textwidth}
    \centering
    \vspace{0pt}
    \includegraphics[width=\linewidth]{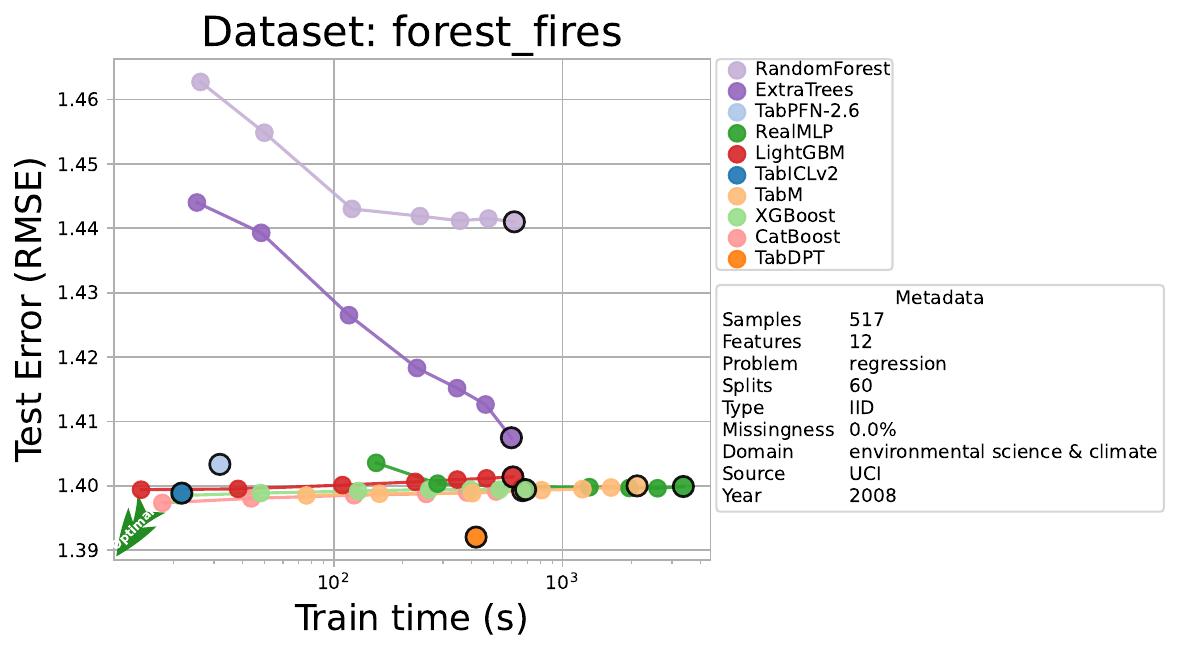}
  \end{minipage}

  \captionof{figure}{\textbf{forest\_fires}: per-method test error (left) and HPO Pareto trajectory (right).}
  \label{fig:perdataset_forest_fires}
\end{center}
\space
\begin{center}
  \begin{minipage}[t]{0.48\textwidth}
    \centering
    \vspace{0pt}
    \scriptsize
\begin{tabular}{l@{\quad}lll}
\toprule
 & Default & Tuned & Tuned + Ens. \\
\midrule
Linear & 0.142 $\pm$ 0.032 & 0.146 $\pm$ 0.030 & 0.130 $\pm$ 0.027 \\
RandomForest & 0.152 $\pm$ 0.032 & 0.154 $\pm$ 0.032 & 0.153 $\pm$ 0.031 \\
ExtraTrees & 0.162 $\pm$ 0.041 & 0.147 $\pm$ 0.031 & 0.147 $\pm$ 0.032 \\
CatBoost & 0.147 $\pm$ 0.030 & 0.142 $\pm$ 0.029 & 0.142 $\pm$ 0.030 \\
LightGBM & 0.152 $\pm$ 0.031 & 0.140 $\pm$ 0.028 & 0.136 $\pm$ 0.028 \\
XGBoost & 0.158 $\pm$ 0.030 & 0.149 $\pm$ 0.031 & 0.147 $\pm$ 0.030 \\
RealMLP & 0.122 $\pm$ 0.025 & \underline{0.125 $\pm$ 0.027} & \underline{0.113 $\pm$ 0.027} \\
TabM & 0.135 $\pm$ 0.028 & 0.134 $\pm$ 0.027 & 0.135 $\pm$ 0.027 \\
TabDPT & 0.107 $\pm$ 0.026 & - & - \\
TabPFN-2.6 & 0.104 $\pm$ 0.026 & - & - \\
TabICLv2 & \underline{\textbf{\textcolor{green!50!black}{0.100 $\pm$ 0.029}}} & - & - \\
\bottomrule
\end{tabular}
\space
  \end{minipage}\hfill
  \begin{minipage}[t]{0.48\textwidth}
    \centering
    \vspace{0pt}
    \includegraphics[width=\linewidth]{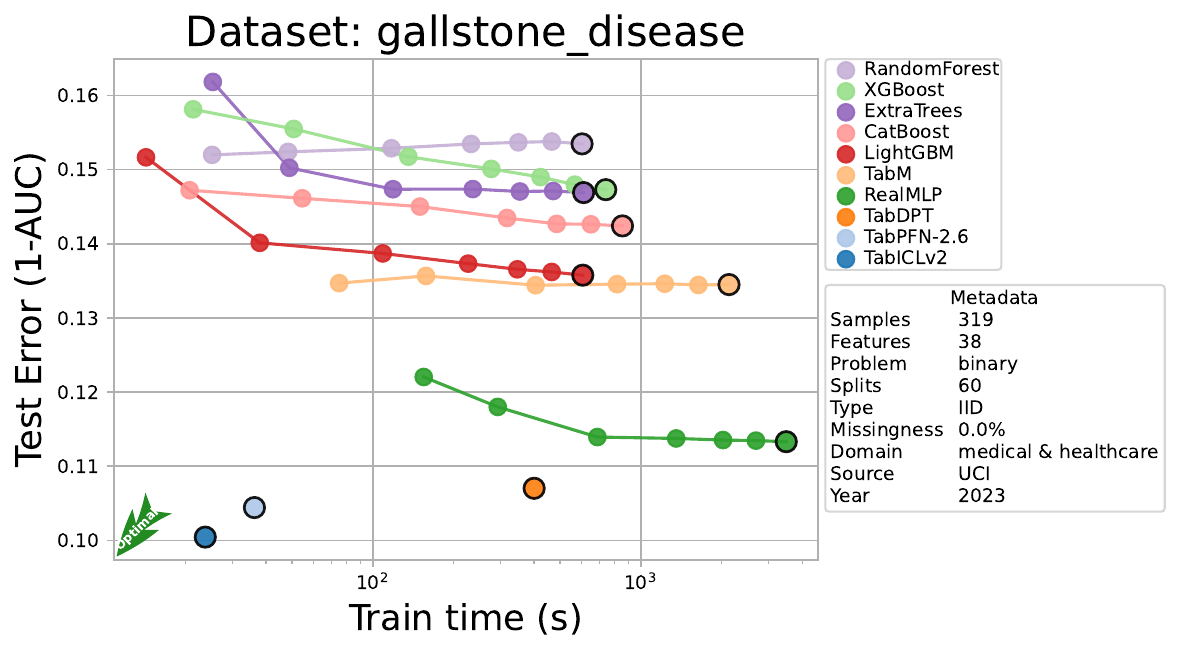}
  \end{minipage}

  \captionof{figure}{\textbf{gallstone\_disease}: per-method test error (left) and HPO Pareto trajectory (right).}
  \label{fig:perdataset_gallstone_disease}
\end{center}
\space
\begin{center}
  \begin{minipage}[t]{0.48\textwidth}
    \centering
    \vspace{0pt}
    \scriptsize
\begin{tabular}{l@{\quad}lll}
\toprule
 & Default & Tuned & Tuned + Ens. \\
\midrule
Linear & 0.267 $\pm$ 0.655 & 0.187 $\pm$ 0.122 & 0.217 $\pm$ 0.348 \\
RandomForest & 0.133 $\pm$ 0.033 & \underline{0.132 $\pm$ 0.032} & \underline{0.132 $\pm$ 0.031} \\
ExtraTrees & 0.133 $\pm$ 0.033 & \underline{0.133 $\pm$ 0.032} & \underline{0.134 $\pm$ 0.032} \\
CatBoost & 0.132 $\pm$ 0.031 & \underline{0.131 $\pm$ 0.031} & \underline{0.131 $\pm$ 0.031} \\
LightGBM & 0.132 $\pm$ 0.032 & \underline{0.133 $\pm$ 0.032} & \underline{0.131 $\pm$ 0.031} \\
XGBoost & 0.133 $\pm$ 0.032 & \underline{0.132 $\pm$ 0.032} & \underline{0.131 $\pm$ 0.031} \\
RealMLP & 0.142 $\pm$ 0.034 & \underline{0.135 $\pm$ 0.036} & \underline{0.134 $\pm$ 0.033} \\
TabM & \underline{\textbf{0.135 $\pm$ 0.037}} & \underline{0.134 $\pm$ 0.036} & \underline{\textbf{0.134 $\pm$ 0.037}} \\
TabDPT & \underline{\textbf{0.133 $\pm$ 0.032}} & - & - \\
TabPFN-2.6 & \underline{\textbf{\textcolor{green!50!black}{0.128 $\pm$ 0.029}}} & - & - \\
TabICLv2 & \underline{\textbf{0.134 $\pm$ 0.041}} & - & - \\
\bottomrule
\end{tabular}
\space
  \end{minipage}\hfill
  \begin{minipage}[t]{0.48\textwidth}
    \centering
    \vspace{0pt}
    \includegraphics[width=\linewidth]{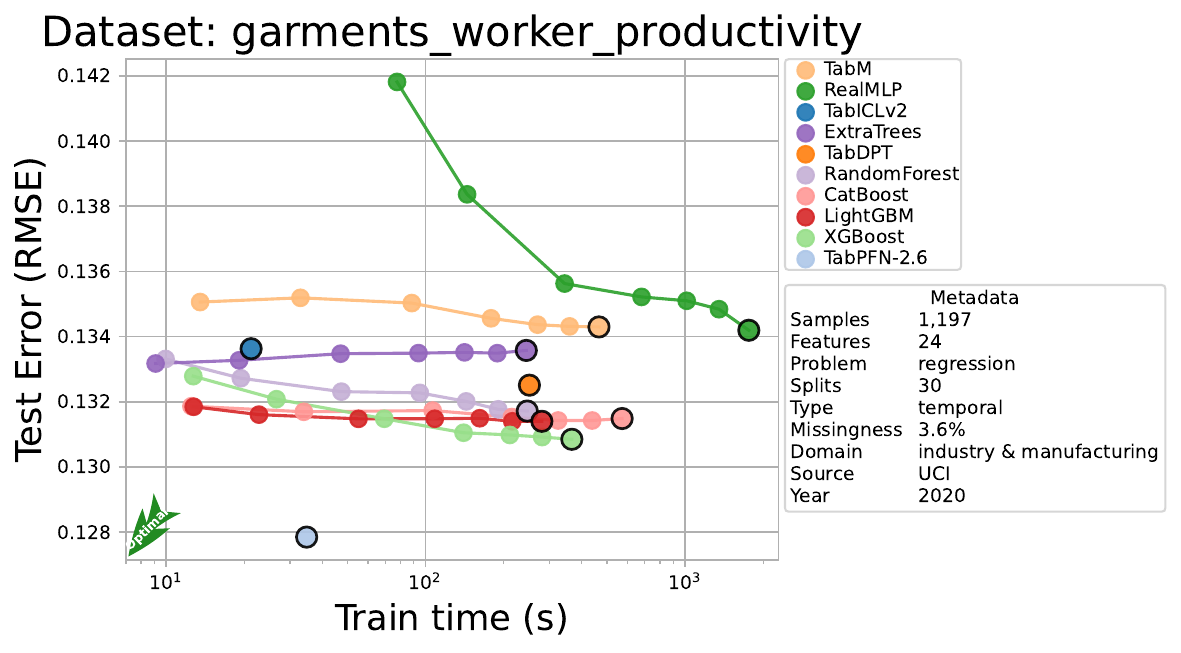}
  \end{minipage}

  \captionof{figure}{\textbf{garments\_worker\_productivity}: per-method test error (left) and HPO Pareto trajectory (right).}
  \label{fig:perdataset_garments_worker_productivity}
\end{center}
\space
\begin{center}
  \begin{minipage}[t]{0.48\textwidth}
    \centering
    \vspace{0pt}
    \scriptsize
\begin{tabular}{l@{\quad}lll}
\toprule
 & Default & Tuned & Tuned + Ens. \\
\midrule
Linear & \underline{\textbf{0.231 $\pm$ 0.138}} & \underline{\textbf{0.209 $\pm$ 0.110}} & \underline{\textbf{0.206 $\pm$ 0.114}} \\
RandomForest & 0.239 $\pm$ 0.125 & 0.251 $\pm$ 0.144 & 0.242 $\pm$ 0.136 \\
ExtraTrees & 0.218 $\pm$ 0.108 & 0.232 $\pm$ 0.134 & 0.236 $\pm$ 0.141 \\
CatBoost & 0.204 $\pm$ 0.123 & \underline{\textbf{0.184 $\pm$ 0.108}} & \underline{\textbf{0.189 $\pm$ 0.106}} \\
LightGBM & \underline{\textbf{0.225 $\pm$ 0.158}} & \underline{\textbf{0.191 $\pm$ 0.102}} & \underline{\textbf{0.182 $\pm$ 0.102}} \\
XGBoost & 0.236 $\pm$ 0.119 & \underline{0.216 $\pm$ 0.136} & \underline{\textbf{0.214 $\pm$ 0.131}} \\
RealMLP & \underline{\textbf{\textcolor{green!50!black}{0.171 $\pm$ 0.118}}} & \underline{\textbf{0.174 $\pm$ 0.091}} & \underline{\textbf{0.174 $\pm$ 0.092}} \\
TabM & \underline{\textbf{0.183 $\pm$ 0.106}} & \underline{\textbf{0.202 $\pm$ 0.126}} & \underline{\textbf{0.186 $\pm$ 0.107}} \\
TabDPT & 0.244 $\pm$ 0.091 & - & - \\
TabPFN-2.6 & \underline{\textbf{0.193 $\pm$ 0.092}} & - & - \\
TabICLv2 & 0.243 $\pm$ 0.082 & - & - \\
\bottomrule
\end{tabular}
\space
  \end{minipage}\hfill
  \begin{minipage}[t]{0.48\textwidth}
    \centering
    \vspace{0pt}
    \includegraphics[width=\linewidth]{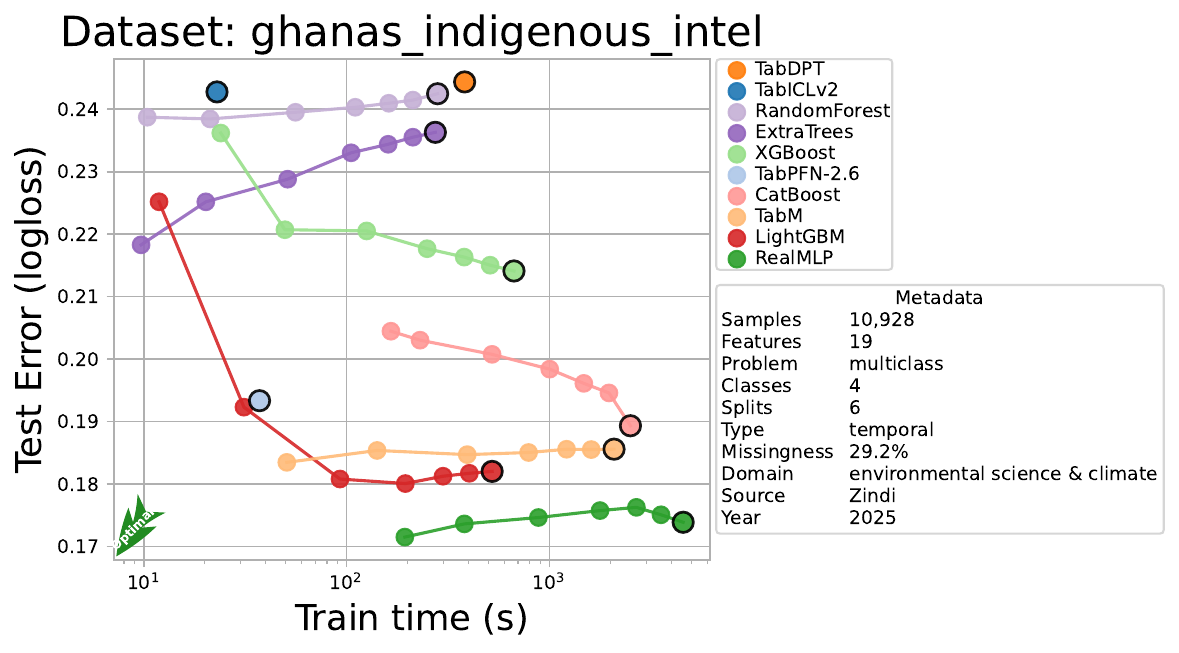}
  \end{minipage}

  \captionof{figure}{\textbf{ghanas\_indigenous\_intel}: per-method test error (left) and HPO Pareto trajectory (right).}
  \label{fig:perdataset_ghanas_indigenous_intel}
\end{center}
\space
\begin{center}
  \begin{minipage}[t]{0.48\textwidth}
    \centering
    \vspace{0pt}
    \scriptsize
\begin{tabular}{l@{\quad}lll}
\toprule
 & Default & Tuned & Tuned + Ens. \\
\midrule
Linear & 0.176 $\pm$ 0.003 & 0.175 $\pm$ 0.003 & 0.173 $\pm$ 0.003 \\
RandomForest & 0.225 $\pm$ 0.020 & 0.178 $\pm$ 0.003 & 0.177 $\pm$ 0.002 \\
ExtraTrees & 0.212 $\pm$ 0.011 & 0.181 $\pm$ 0.004 & 0.181 $\pm$ 0.004 \\
CatBoost & 0.157 $\pm$ 0.003 & 0.156 $\pm$ 0.004 & 0.156 $\pm$ 0.003 \\
LightGBM & \underline{0.153 $\pm$ 0.004} & \underline{\textbf{0.153 $\pm$ 0.004}} & \underline{\textbf{\textcolor{green!50!black}{0.152 $\pm$ 0.004}}} \\
XGBoost & 0.154 $\pm$ 0.004 & 0.154 $\pm$ 0.004 & 0.154 $\pm$ 0.003 \\
RealMLP & 0.156 $\pm$ 0.003 & 0.154 $\pm$ 0.004 & 0.154 $\pm$ 0.003 \\
TabM & \underline{0.154 $\pm$ 0.004} & \underline{0.153 $\pm$ 0.003} & \underline{\textbf{0.153 $\pm$ 0.004}} \\
TabDPT & 0.222 $\pm$ 0.010 & - & - \\
TabPFN-2.6 & 0.169 $\pm$ 0.006 & - & - \\
TabICLv2 & \underline{\textbf{0.153 $\pm$ 0.003}} & - & - \\
\bottomrule
\end{tabular}
\space
  \end{minipage}\hfill
  \begin{minipage}[t]{0.48\textwidth}
    \centering
    \vspace{0pt}
    \includegraphics[width=\linewidth]{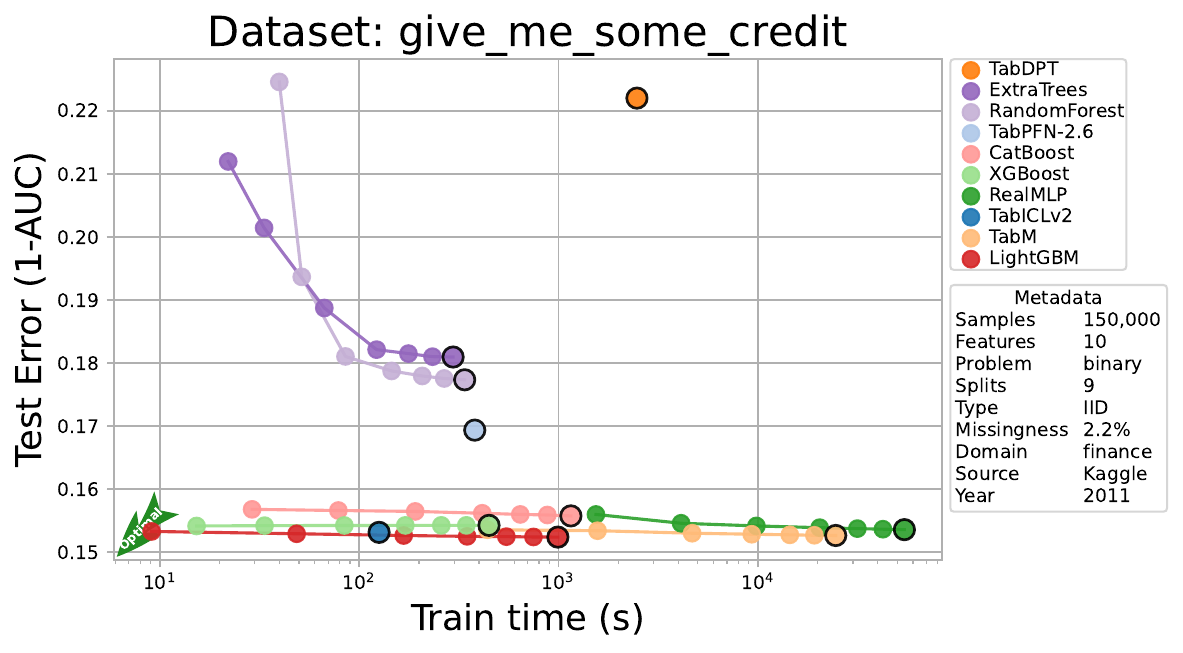}
  \end{minipage}

  \captionof{figure}{\textbf{give\_me\_some\_credit}: per-method test error (left) and HPO Pareto trajectory (right).}
  \label{fig:perdataset_give_me_some_credit}
\end{center}
\space
\begin{center}
  \begin{minipage}[t]{0.48\textwidth}
    \centering
    \vspace{0pt}
    \scriptsize
\begin{tabular}{l@{\quad}lll}
\toprule
 & Default & Tuned & Tuned + Ens. \\
\midrule
Linear & 0.051 $\pm$ 0.007 & 0.047 $\pm$ 0.006 & 0.047 $\pm$ 0.006 \\
RandomForest & 0.043 $\pm$ 0.007 & 0.042 $\pm$ 0.006 & 0.041 $\pm$ 0.006 \\
ExtraTrees & 0.047 $\pm$ 0.007 & 0.037 $\pm$ 0.006 & 0.037 $\pm$ 0.006 \\
CatBoost & 0.027 $\pm$ 0.006 & 0.026 $\pm$ 0.006 & 0.027 $\pm$ 0.006 \\
LightGBM & 0.028 $\pm$ 0.005 & 0.023 $\pm$ 0.006 & 0.023 $\pm$ 0.006 \\
XGBoost & 0.028 $\pm$ 0.006 & 0.026 $\pm$ 0.006 & 0.026 $\pm$ 0.006 \\
RealMLP & 0.016 $\pm$ 0.005 & \underline{0.016 $\pm$ 0.004} & \underline{0.015 $\pm$ 0.004} \\
TabM & 0.033 $\pm$ 0.006 & 0.018 $\pm$ 0.005 & 0.018 $\pm$ 0.005 \\
TabDPT & 0.007 $\pm$ 0.003 & - & - \\
TabPFN-2.6 & 0.008 $\pm$ 0.003 & - & - \\
TabICLv2 & \underline{\textbf{\textcolor{green!50!black}{0.005 $\pm$ 0.002}}} & - & - \\
\bottomrule
\end{tabular}
\space
  \end{minipage}\hfill
  \begin{minipage}[t]{0.48\textwidth}
    \centering
    \vspace{0pt}
    \includegraphics[width=\linewidth]{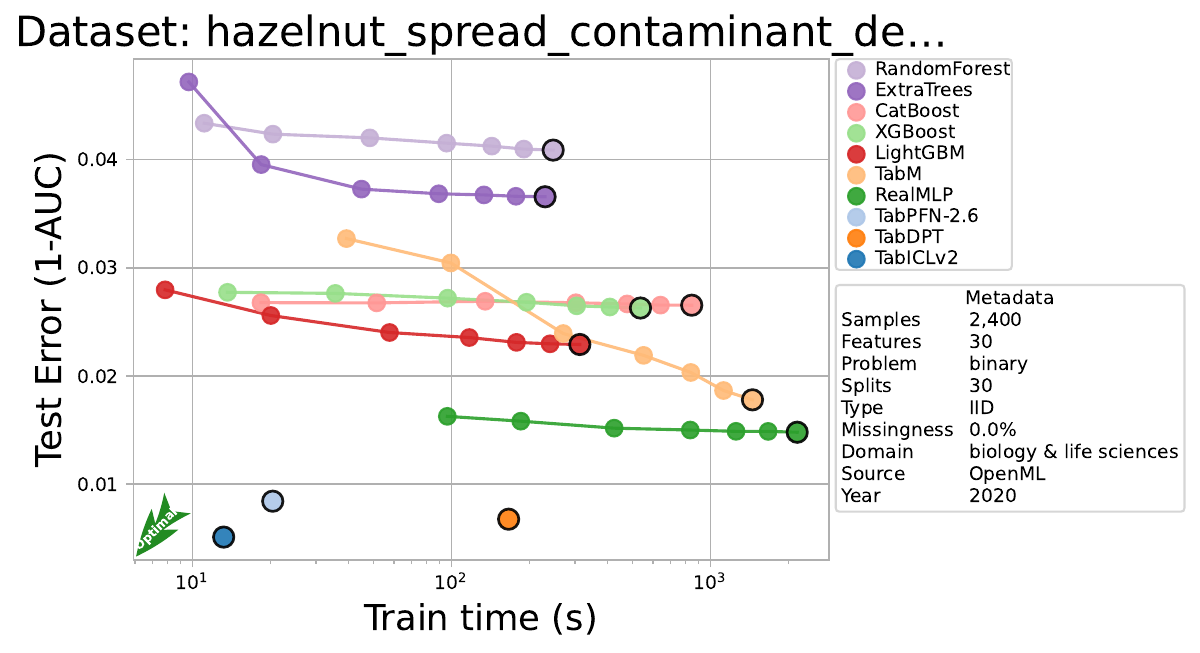}
  \end{minipage}

  \captionof{figure}{\textbf{hazelnut\_spread\_contaminant\_detection}: per-method test error (left) and HPO Pareto trajectory (right).}
  \label{fig:perdataset_hazelnut_spread_contaminant_detection}
\end{center}
\space
\begin{center}
  \begin{minipage}[t]{0.48\textwidth}
    \centering
    \vspace{0pt}
    \scriptsize
\begin{tabular}{l@{\quad}lll}
\toprule
 & Default & Tuned & Tuned + Ens. \\
\midrule
Linear & 0.446 $\pm$ 0.024 & 0.446 $\pm$ 0.024 & 0.446 $\pm$ 0.024 \\
RandomForest & 0.394 $\pm$ 0.026 & 0.379 $\pm$ 0.028 & 0.378 $\pm$ 0.028 \\
ExtraTrees & 0.390 $\pm$ 0.027 & 0.372 $\pm$ 0.031 & 0.372 $\pm$ 0.031 \\
CatBoost & 0.373 $\pm$ 0.031 & 0.372 $\pm$ 0.030 & 0.372 $\pm$ 0.030 \\
LightGBM & 0.377 $\pm$ 0.028 & \underline{0.370 $\pm$ 0.030} & \underline{0.368 $\pm$ 0.031} \\
XGBoost & 0.384 $\pm$ 0.026 & \underline{0.370 $\pm$ 0.030} & 0.371 $\pm$ 0.030 \\
RealMLP & 0.372 $\pm$ 0.031 & 0.374 $\pm$ 0.029 & 0.371 $\pm$ 0.030 \\
TabM & 0.373 $\pm$ 0.030 & 0.371 $\pm$ 0.031 & 0.371 $\pm$ 0.031 \\
TabDPT & 0.364 $\pm$ 0.032 & - & - \\
TabPFN-2.6 & \underline{\textbf{0.360 $\pm$ 0.037}} & - & - \\
TabICLv2 & \underline{\textbf{\textcolor{green!50!black}{0.360 $\pm$ 0.034}}} & - & - \\
\bottomrule
\end{tabular}
\space
  \end{minipage}\hfill
  \begin{minipage}[t]{0.48\textwidth}
    \centering
    \vspace{0pt}
    \includegraphics[width=\linewidth]{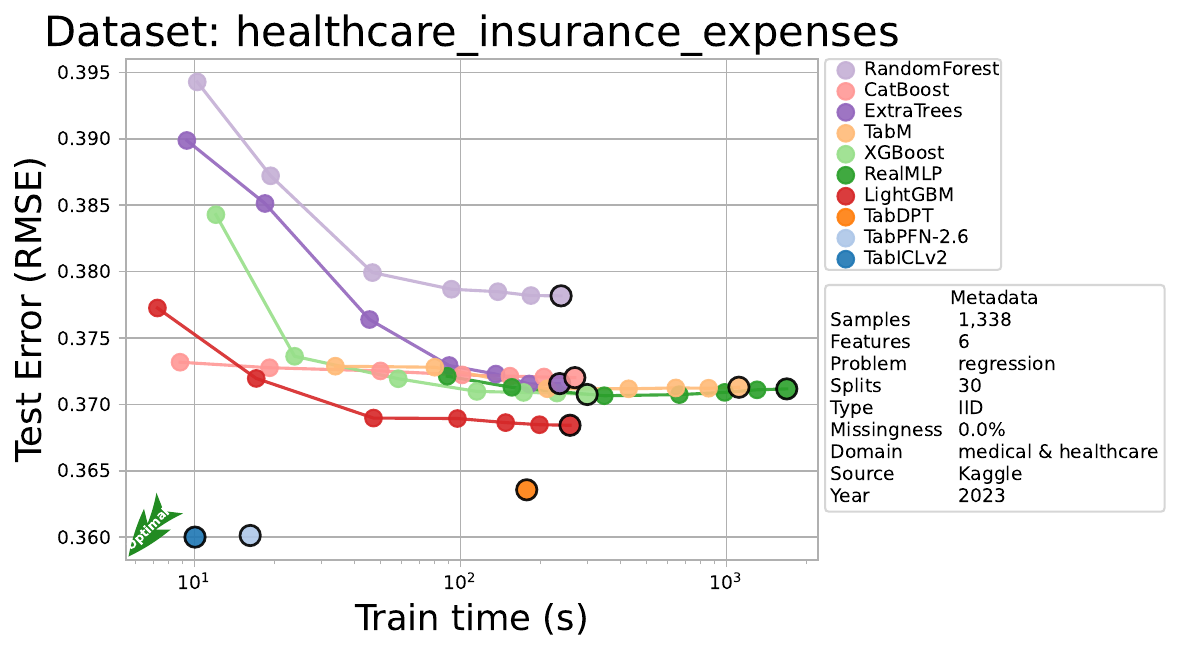}
  \end{minipage}

  \captionof{figure}{\textbf{healthcare\_insurance\_expenses}: per-method test error (left) and HPO Pareto trajectory (right).}
  \label{fig:perdataset_healthcare_insurance_expenses}
\end{center}
\space
\begin{center}
  \begin{minipage}[t]{0.48\textwidth}
    \centering
    \vspace{0pt}
    \scriptsize
\begin{tabular}{l@{\quad}lll}
\toprule
 & Default & Tuned & Tuned + Ens. \\
\midrule
Linear & \underline{\textbf{0.092 $\pm$ 0.025}} & \underline{\textbf{0.093 $\pm$ 0.025}} & \underline{\textbf{0.090 $\pm$ 0.025}} \\
RandomForest & 0.097 $\pm$ 0.027 & 0.100 $\pm$ 0.028 & 0.100 $\pm$ 0.027 \\
ExtraTrees & \underline{\textbf{\textcolor{green!50!black}{0.090 $\pm$ 0.025}}} & \underline{0.092 $\pm$ 0.026} & \underline{0.092 $\pm$ 0.026} \\
CatBoost & 0.098 $\pm$ 0.028 & \underline{0.094 $\pm$ 0.027} & 0.094 $\pm$ 0.026 \\
LightGBM & 0.113 $\pm$ 0.028 & \underline{0.094 $\pm$ 0.026} & \underline{\textbf{0.092 $\pm$ 0.026}} \\
XGBoost & 0.104 $\pm$ 0.028 & 0.101 $\pm$ 0.028 & 0.100 $\pm$ 0.027 \\
RealMLP & 0.097 $\pm$ 0.025 & \underline{\textbf{0.093 $\pm$ 0.026}} & \underline{\textbf{0.092 $\pm$ 0.026}} \\
TabM & 0.093 $\pm$ 0.026 & 0.095 $\pm$ 0.026 & 0.094 $\pm$ 0.027 \\
TabDPT & 0.093 $\pm$ 0.026 & - & - \\
TabPFN-2.6 & \underline{\textbf{0.090 $\pm$ 0.027}} & - & - \\
TabICLv2 & \underline{\textbf{0.091 $\pm$ 0.026}} & - & - \\
\bottomrule
\end{tabular}
\space
  \end{minipage}\hfill
  \begin{minipage}[t]{0.48\textwidth}
    \centering
    \vspace{0pt}
    \includegraphics[width=\linewidth]{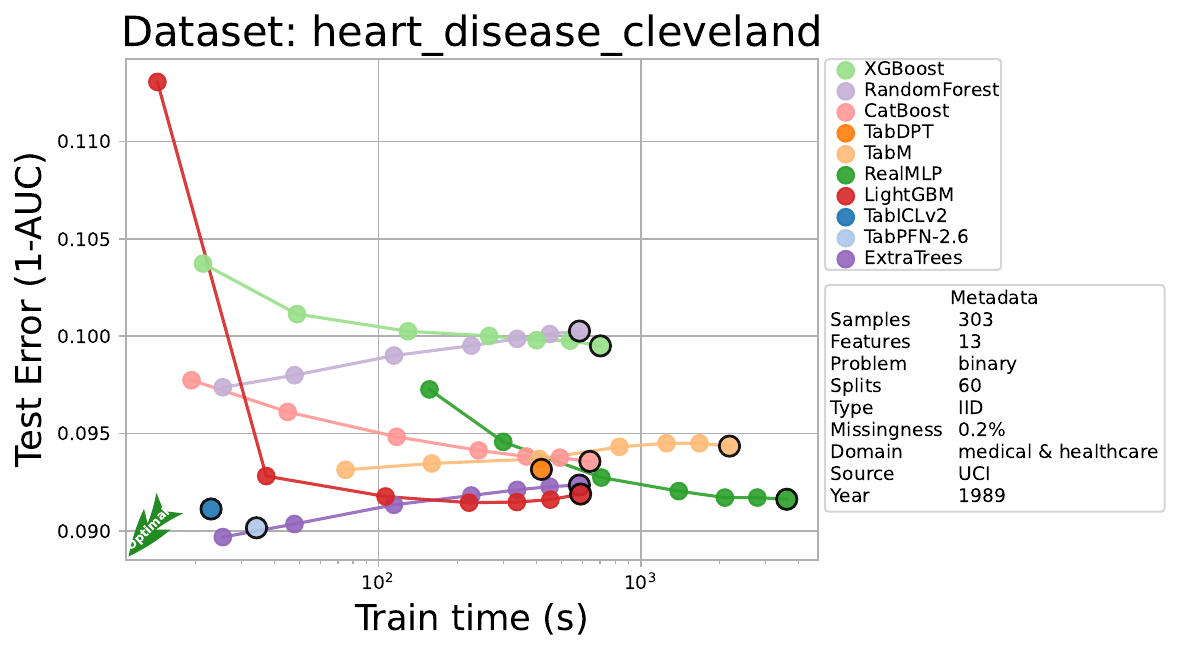}
  \end{minipage}

  \captionof{figure}{\textbf{heart\_disease\_cleveland}: per-method test error (left) and HPO Pareto trajectory (right).}
  \label{fig:perdataset_heart_disease_cleveland}
\end{center}
\space
\begin{center}
  \begin{minipage}[t]{0.48\textwidth}
    \centering
    \vspace{0pt}
    \scriptsize
\begin{tabular}{l@{\quad}lll}
\toprule
 & Default & Tuned & Tuned + Ens. \\
\midrule
Linear & 0.097 $\pm$ 0.026 & 0.096 $\pm$ 0.028 & \underline{\textbf{0.093 $\pm$ 0.026}} \\
RandomForest & 0.110 $\pm$ 0.023 & 0.109 $\pm$ 0.025 & 0.110 $\pm$ 0.024 \\
ExtraTrees & 0.102 $\pm$ 0.022 & 0.096 $\pm$ 0.023 & 0.096 $\pm$ 0.023 \\
CatBoost & 0.098 $\pm$ 0.021 & 0.100 $\pm$ 0.025 & 0.099 $\pm$ 0.024 \\
LightGBM & 0.154 $\pm$ 0.032 & 0.099 $\pm$ 0.026 & 0.099 $\pm$ 0.024 \\
XGBoost & 0.124 $\pm$ 0.025 & 0.101 $\pm$ 0.022 & 0.101 $\pm$ 0.022 \\
RealMLP & \underline{\textbf{0.094 $\pm$ 0.023}} & \underline{\textbf{\textcolor{green!50!black}{0.091 $\pm$ 0.023}}} & \underline{\textbf{0.091 $\pm$ 0.023}} \\
TabM & \underline{\textbf{0.094 $\pm$ 0.023}} & \underline{\textbf{0.093 $\pm$ 0.023}} & 0.095 $\pm$ 0.022 \\
TabDPT & 0.095 $\pm$ 0.021 & - & - \\
TabPFN-2.6 & \underline{\textbf{0.093 $\pm$ 0.024}} & - & - \\
TabICLv2 & \underline{\textbf{0.094 $\pm$ 0.023}} & - & - \\
\bottomrule
\end{tabular}
\space
  \end{minipage}\hfill
  \begin{minipage}[t]{0.48\textwidth}
    \centering
    \vspace{0pt}
    \includegraphics[width=\linewidth]{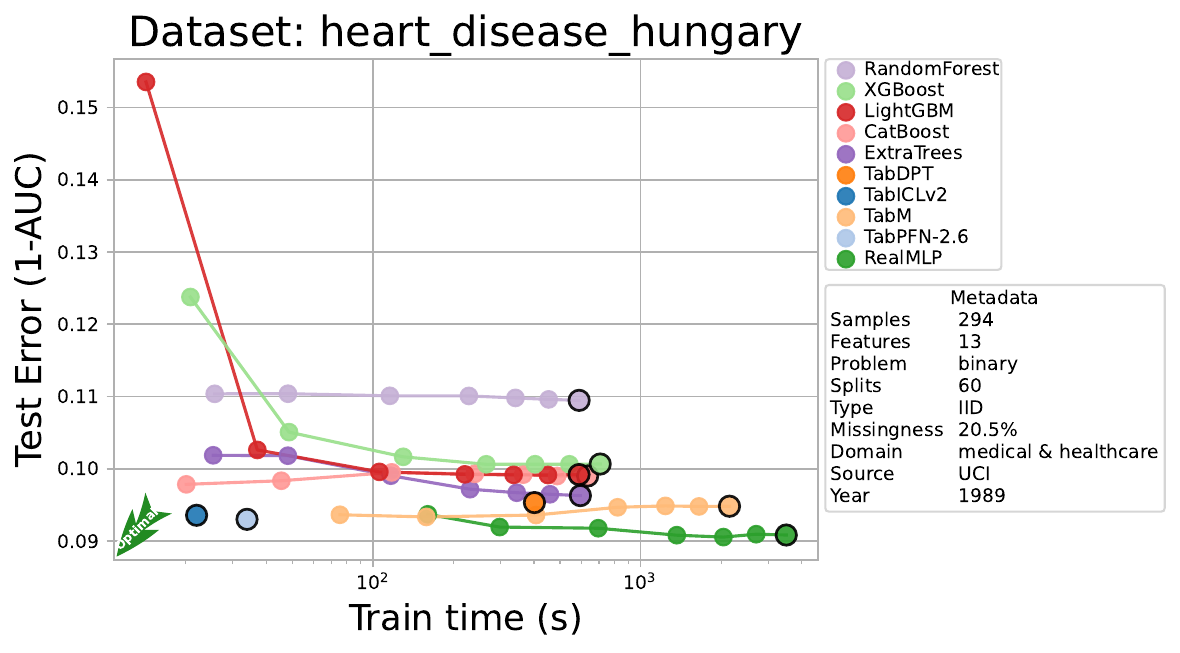}
  \end{minipage}

  \captionof{figure}{\textbf{heart\_disease\_hungary}: per-method test error (left) and HPO Pareto trajectory (right).}
  \label{fig:perdataset_heart_disease_hungary}
\end{center}
\space
\begin{center}
  \begin{minipage}[t]{0.48\textwidth}
    \centering
    \vspace{0pt}
    \scriptsize
\begin{tabular}{l@{\quad}lll}
\toprule
 & Default & Tuned & Tuned + Ens. \\
\midrule
Linear & 0.338 $\pm$ 0.056 & 0.334 $\pm$ 0.063 & 0.335 $\pm$ 0.061 \\
RandomForest & 0.322 $\pm$ 0.049 & 0.327 $\pm$ 0.051 & 0.327 $\pm$ 0.050 \\
ExtraTrees & 0.304 $\pm$ 0.047 & \underline{0.311 $\pm$ 0.049} & \underline{0.311 $\pm$ 0.050} \\
CatBoost & 0.305 $\pm$ 0.050 & \underline{0.321 $\pm$ 0.048} & \underline{0.318 $\pm$ 0.048} \\
LightGBM & 0.371 $\pm$ 0.064 & 0.333 $\pm$ 0.065 & 0.334 $\pm$ 0.064 \\
XGBoost & 0.368 $\pm$ 0.061 & 0.358 $\pm$ 0.063 & 0.349 $\pm$ 0.062 \\
RealMLP & 0.342 $\pm$ 0.057 & 0.348 $\pm$ 0.060 & 0.351 $\pm$ 0.055 \\
TabM & 0.329 $\pm$ 0.056 & 0.350 $\pm$ 0.064 & 0.349 $\pm$ 0.063 \\
TabDPT & \underline{\textbf{\textcolor{green!50!black}{0.294 $\pm$ 0.049}}} & - & - \\
TabPFN-2.6 & 0.319 $\pm$ 0.055 & - & - \\
TabICLv2 & 0.327 $\pm$ 0.051 & - & - \\
\bottomrule
\end{tabular}
\space
  \end{minipage}\hfill
  \begin{minipage}[t]{0.48\textwidth}
    \centering
    \vspace{0pt}
    \includegraphics[width=\linewidth]{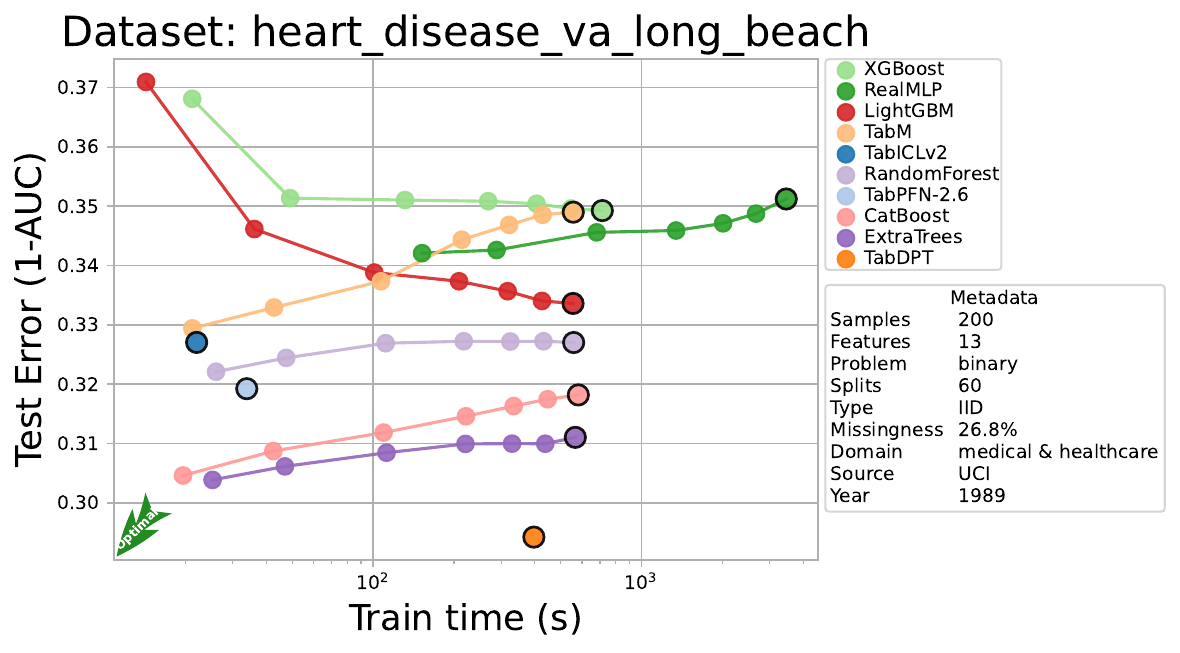}
  \end{minipage}

  \captionof{figure}{\textbf{heart\_disease\_va\_long\_beach}: per-method test error (left) and HPO Pareto trajectory (right).}
  \label{fig:perdataset_heart_disease_va_long_beach}
\end{center}
\space
\begin{center}
  \begin{minipage}[t]{0.48\textwidth}
    \centering
    \vspace{0pt}
    \scriptsize
\begin{tabular}{l@{\quad}lll}
\toprule
 & Default & Tuned & Tuned + Ens. \\
\midrule
Linear & 0.129 $\pm$ 0.029 & 0.129 $\pm$ 0.028 & 0.128 $\pm$ 0.027 \\
RandomForest & \underline{\textbf{0.086 $\pm$ 0.021}} & \underline{\textbf{0.087 $\pm$ 0.023}} & \underline{\textbf{0.087 $\pm$ 0.022}} \\
ExtraTrees & 0.106 $\pm$ 0.021 & 0.092 $\pm$ 0.023 & 0.091 $\pm$ 0.022 \\
CatBoost & 0.090 $\pm$ 0.021 & 0.092 $\pm$ 0.022 & 0.090 $\pm$ 0.022 \\
LightGBM & 0.090 $\pm$ 0.021 & 0.092 $\pm$ 0.023 & 0.091 $\pm$ 0.022 \\
XGBoost & 0.096 $\pm$ 0.025 & 0.090 $\pm$ 0.021 & \underline{0.089 $\pm$ 0.021} \\
RealMLP & 0.121 $\pm$ 0.026 & 0.099 $\pm$ 0.024 & 0.095 $\pm$ 0.020 \\
TabM & \underline{\textbf{0.088 $\pm$ 0.023}} & \underline{0.090 $\pm$ 0.024} & \underline{0.089 $\pm$ 0.024} \\
TabDPT & 0.094 $\pm$ 0.022 & - & - \\
TabPFN-2.6 & 0.089 $\pm$ 0.020 & - & - \\
TabICLv2 & \underline{\textbf{\textcolor{green!50!black}{0.084 $\pm$ 0.020}}} & - & - \\
\bottomrule
\end{tabular}
\space
  \end{minipage}\hfill
  \begin{minipage}[t]{0.48\textwidth}
    \centering
    \vspace{0pt}
    \includegraphics[width=\linewidth]{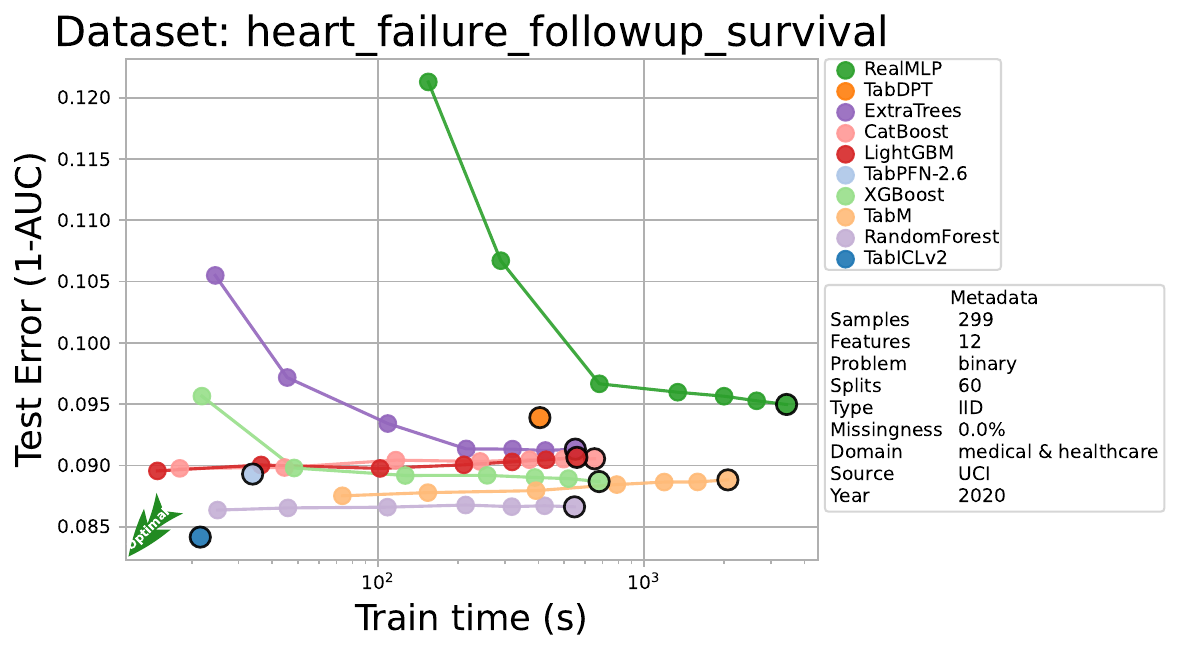}
  \end{minipage}

  \captionof{figure}{\textbf{heart\_failure\_followup\_survival}: per-method test error (left) and HPO Pareto trajectory (right).}
  \label{fig:perdataset_heart_failure_followup_survival}
\end{center}
\space
\begin{center}
  \begin{minipage}[t]{0.48\textwidth}
    \centering
    \vspace{0pt}
    \scriptsize
\begin{tabular}{l@{\quad}lll}
\toprule
 & Default & Tuned & Tuned + Ens. \\
\midrule
Linear & 0.214 $\pm$ 0.005 & 0.214 $\pm$ 0.006 & 0.214 $\pm$ 0.005 \\
RandomForest & 0.207 $\pm$ 0.004 & 0.207 $\pm$ 0.004 & 0.207 $\pm$ 0.004 \\
ExtraTrees & 0.207 $\pm$ 0.005 & 0.207 $\pm$ 0.004 & 0.206 $\pm$ 0.005 \\
CatBoost & 0.202 $\pm$ 0.005 & \underline{0.201 $\pm$ 0.005} & 0.201 $\pm$ 0.005 \\
LightGBM & 0.205 $\pm$ 0.004 & \underline{0.201 $\pm$ 0.005} & 0.201 $\pm$ 0.005 \\
XGBoost & 0.206 $\pm$ 0.004 & 0.203 $\pm$ 0.004 & 0.202 $\pm$ 0.004 \\
RealMLP & 0.202 $\pm$ 0.005 & \underline{0.200 $\pm$ 0.005} & \underline{0.199 $\pm$ 0.004} \\
TabM & 0.203 $\pm$ 0.004 & 0.202 $\pm$ 0.004 & 0.201 $\pm$ 0.004 \\
TabDPT & 0.204 $\pm$ 0.005 & - & - \\
TabPFN-2.6 & \underline{\textbf{0.198 $\pm$ 0.004}} & - & - \\
TabICLv2 & \underline{\textbf{\textcolor{green!50!black}{0.198 $\pm$ 0.004}}} & - & - \\
\bottomrule
\end{tabular}
\space
  \end{minipage}\hfill
  \begin{minipage}[t]{0.48\textwidth}
    \centering
    \vspace{0pt}
    \includegraphics[width=\linewidth]{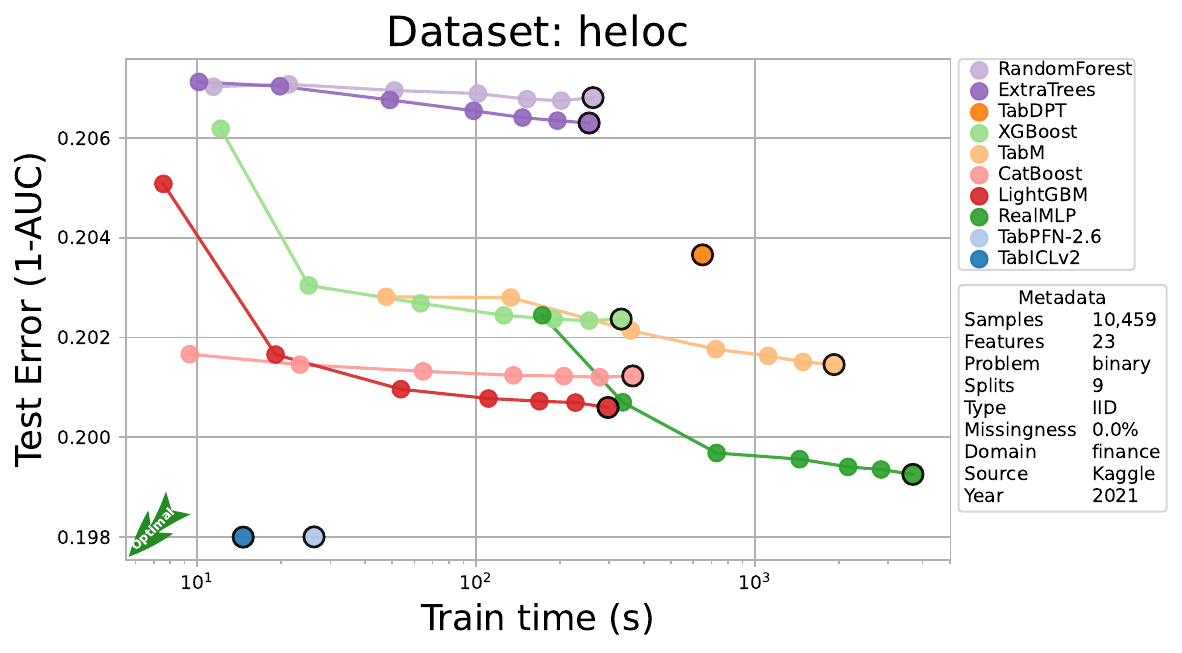}
  \end{minipage}

  \captionof{figure}{\textbf{heloc}: per-method test error (left) and HPO Pareto trajectory (right).}
  \label{fig:perdataset_heloc}
\end{center}
\space
\begin{center}
  \begin{minipage}[t]{0.48\textwidth}
    \centering
    \vspace{0pt}
    \scriptsize
\begin{tabular}{l@{\quad}lll}
\toprule
 & Default & Tuned & Tuned + Ens. \\
\midrule
Linear & 0.195 $\pm$ 0.044 & 0.196 $\pm$ 0.037 & 0.186 $\pm$ 0.034 \\
RandomForest & 0.162 $\pm$ 0.013 & 0.155 $\pm$ 0.023 & 0.154 $\pm$ 0.019 \\
ExtraTrees & 0.184 $\pm$ 0.010 & 0.147 $\pm$ 0.013 & 0.147 $\pm$ 0.013 \\
CatBoost & 0.150 $\pm$ 0.022 & 0.150 $\pm$ 0.024 & 0.149 $\pm$ 0.022 \\
LightGBM & 0.160 $\pm$ 0.025 & 0.144 $\pm$ 0.023 & 0.144 $\pm$ 0.022 \\
XGBoost & 0.164 $\pm$ 0.028 & 0.157 $\pm$ 0.027 & 0.156 $\pm$ 0.026 \\
RealMLP & 0.170 $\pm$ 0.039 & 0.144 $\pm$ 0.029 & 0.141 $\pm$ 0.026 \\
TabM & 0.150 $\pm$ 0.026 & \underline{0.136 $\pm$ 0.027} & \underline{0.134 $\pm$ 0.026} \\
TabDPT & 0.160 $\pm$ 0.030 & - & - \\
TabPFN-2.6 & \underline{\textbf{\textcolor{green!50!black}{0.125 $\pm$ 0.026}}} & - & - \\
TabICLv2 & 0.131 $\pm$ 0.034 & - & - \\
\bottomrule
\end{tabular}
\space
  \end{minipage}\hfill
  \begin{minipage}[t]{0.48\textwidth}
    \centering
    \vspace{0pt}
    \includegraphics[width=\linewidth]{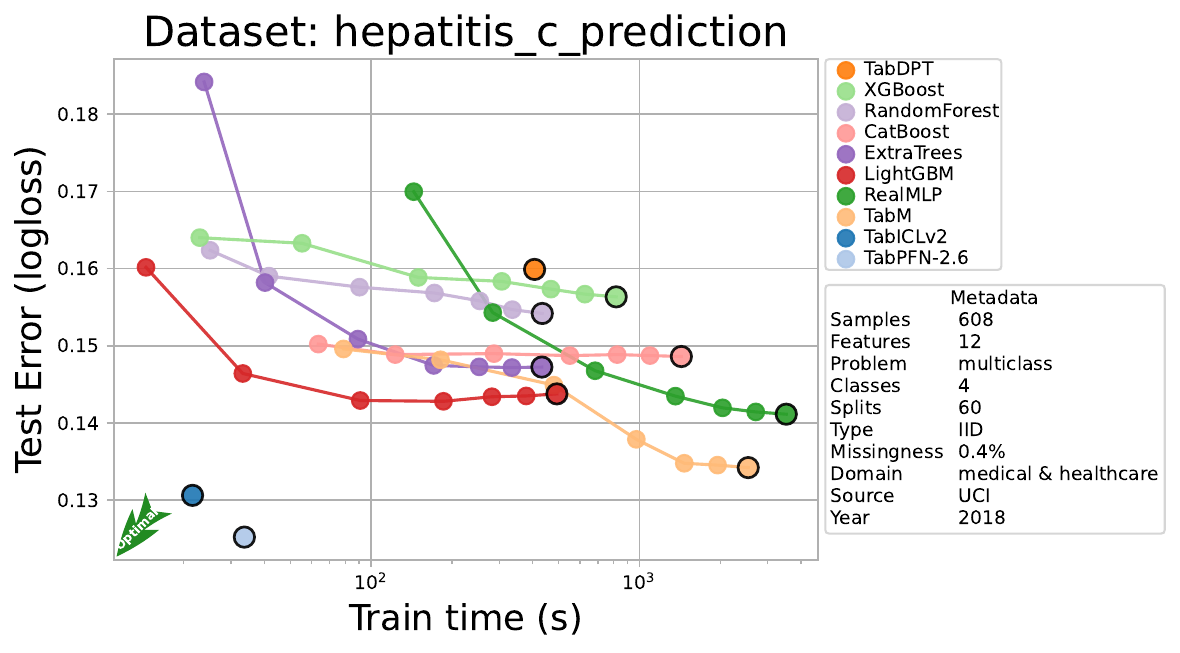}
  \end{minipage}

  \captionof{figure}{\textbf{hepatitis\_c\_prediction}: per-method test error (left) and HPO Pareto trajectory (right).}
  \label{fig:perdataset_hepatitis_c_prediction}
\end{center}
\space
\begin{center}
  \begin{minipage}[t]{0.48\textwidth}
    \centering
    \vspace{0pt}
    \scriptsize
\begin{tabular}{l@{\quad}lll}
\toprule
 & Default & Tuned & Tuned + Ens. \\
\midrule
Linear & 0.156 $\pm$ 0.058 & 0.143 $\pm$ 0.062 & 0.146 $\pm$ 0.059 \\
RandomForest & 0.130 $\pm$ 0.048 & 0.141 $\pm$ 0.055 & \underline{0.142 $\pm$ 0.054} \\
ExtraTrees & 0.128 $\pm$ 0.044 & \underline{0.138 $\pm$ 0.048} & \underline{0.135 $\pm$ 0.046} \\
CatBoost & 0.137 $\pm$ 0.049 & \underline{0.132 $\pm$ 0.050} & \underline{0.132 $\pm$ 0.050} \\
LightGBM & 0.170 $\pm$ 0.057 & 0.150 $\pm$ 0.056 & 0.150 $\pm$ 0.055 \\
XGBoost & 0.168 $\pm$ 0.061 & 0.140 $\pm$ 0.058 & 0.143 $\pm$ 0.059 \\
RealMLP & 0.146 $\pm$ 0.060 & 0.150 $\pm$ 0.066 & 0.145 $\pm$ 0.064 \\
TabM & 0.131 $\pm$ 0.058 & \underline{0.139 $\pm$ 0.057} & \underline{0.131 $\pm$ 0.056} \\
TabDPT & \underline{\textbf{\textcolor{green!50!black}{0.121 $\pm$ 0.045}}} & - & - \\
TabPFN-2.6 & 0.135 $\pm$ 0.054 & - & - \\
TabICLv2 & 0.127 $\pm$ 0.055 & - & - \\
\bottomrule
\end{tabular}
\space
  \end{minipage}\hfill
  \begin{minipage}[t]{0.48\textwidth}
    \centering
    \vspace{0pt}
    \includegraphics[width=\linewidth]{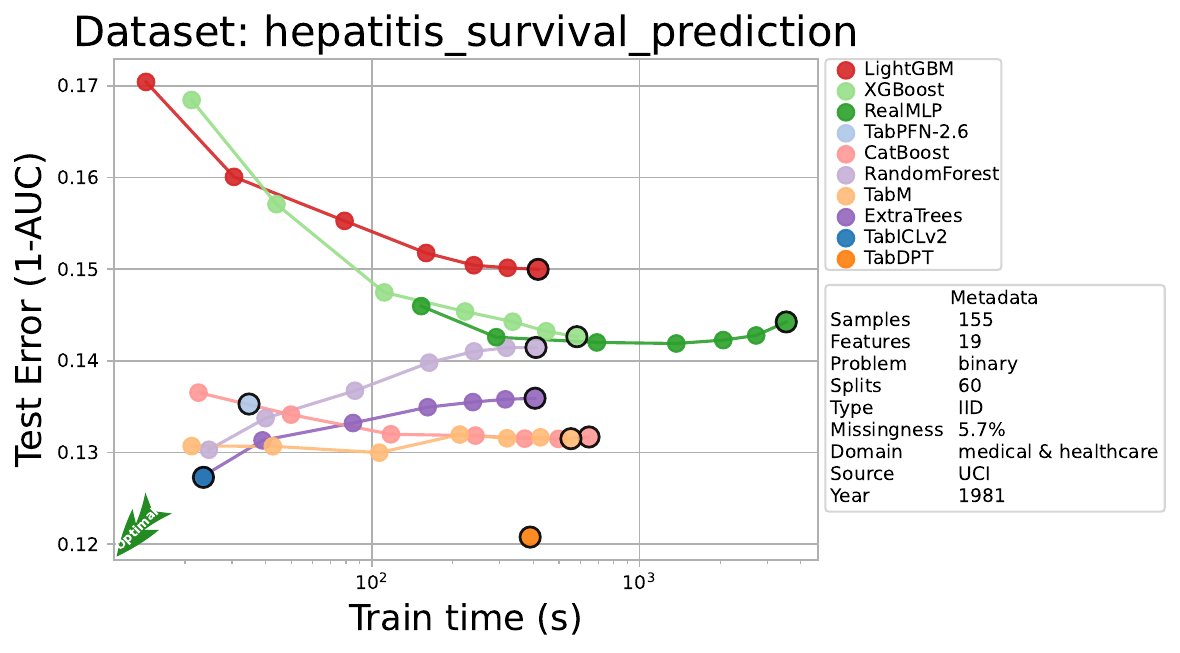}
  \end{minipage}

  \captionof{figure}{\textbf{hepatitis\_survival\_prediction}: per-method test error (left) and HPO Pareto trajectory (right).}
  \label{fig:perdataset_hepatitis_survival_prediction}
\end{center}
\space
\begin{center}
  \begin{minipage}[t]{0.48\textwidth}
    \centering
    \vspace{0pt}
    \scriptsize
\begin{tabular}{l@{\quad}lll}
\toprule
 & Default & Tuned & Tuned + Ens. \\
\midrule
Linear & 0.262 $\pm$ 0.026 & 0.266 $\pm$ 0.031 & 0.266 $\pm$ 0.031 \\
RandomForest & \underline{\textbf{0.212 $\pm$ 0.024}} & \underline{\textbf{0.212 $\pm$ 0.024}} & \underline{\textbf{0.213 $\pm$ 0.027}} \\
ExtraTrees & \underline{\textbf{0.208 $\pm$ 0.025}} & \underline{\textbf{0.215 $\pm$ 0.021}} & \underline{\textbf{0.214 $\pm$ 0.022}} \\
CatBoost & \underline{\textbf{0.210 $\pm$ 0.023}} & \underline{\textbf{0.215 $\pm$ 0.021}} & \underline{\textbf{0.214 $\pm$ 0.021}} \\
LightGBM & \underline{\textbf{\textcolor{green!50!black}{0.208 $\pm$ 0.033}}} & \underline{\textbf{0.210 $\pm$ 0.038}} & \underline{\textbf{0.209 $\pm$ 0.034}} \\
XGBoost & \underline{\textbf{0.208 $\pm$ 0.031}} & \underline{\textbf{0.213 $\pm$ 0.029}} & \underline{\textbf{0.209 $\pm$ 0.028}} \\
RealMLP & 0.233 $\pm$ 0.030 & 0.240 $\pm$ 0.043 & \underline{\textbf{0.220 $\pm$ 0.030}} \\
TabM & 0.237 $\pm$ 0.032 & \underline{0.232 $\pm$ 0.027} & \underline{0.232 $\pm$ 0.029} \\
TabDPT & \underline{\textbf{0.214 $\pm$ 0.025}} & - & - \\
TabPFN-2.6 & \underline{\textbf{0.212 $\pm$ 0.020}} & - & - \\
TabICLv2 & 0.250 $\pm$ 0.028 & - & - \\
\bottomrule
\end{tabular}
\space
  \end{minipage}\hfill
  \begin{minipage}[t]{0.48\textwidth}
    \centering
    \vspace{0pt}
    \includegraphics[width=\linewidth]{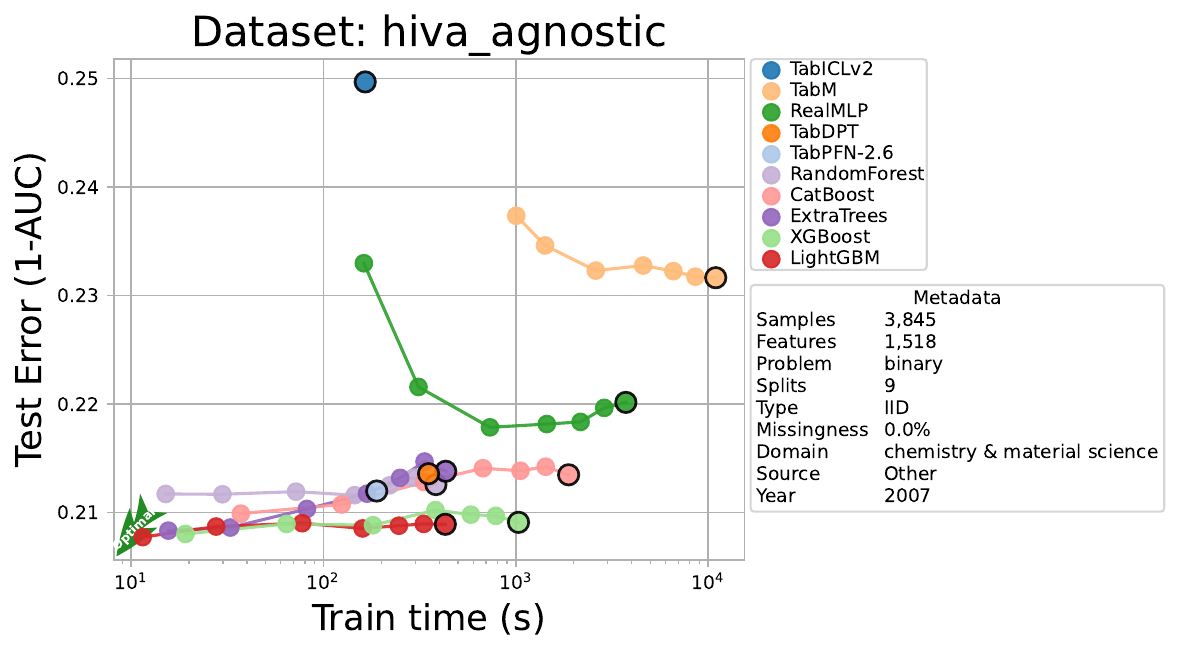}
  \end{minipage}

  \captionof{figure}{\textbf{hiva\_agnostic}: per-method test error (left) and HPO Pareto trajectory (right).}
  \label{fig:perdataset_hiva_agnostic}
\end{center}
\space
\begin{center}
  \begin{minipage}[t]{0.48\textwidth}
    \centering
    \vspace{0pt}
    \scriptsize
\begin{tabular}{l@{\quad}lll}
\toprule
 & Default & Tuned & Tuned + Ens. \\
\midrule
Linear & 0.226 $\pm$ 0.001 & 0.225 $\pm$ 0.001 & 0.222 $\pm$ 0.001 \\
RandomForest & 0.260 $\pm$ 0.002 & 0.252 $\pm$ 0.002 & 0.245 $\pm$ 0.002 \\
ExtraTrees & 0.257 $\pm$ 0.002 & 0.252 $\pm$ 0.002 & 0.244 $\pm$ 0.002 \\
CatBoost & \underline{0.209 $\pm$ 0.001} & 0.208 $\pm$ 0.001 & 0.208 $\pm$ 0.001 \\
LightGBM & 0.213 $\pm$ 0.001 & 0.208 $\pm$ 0.001 & 0.208 $\pm$ 0.001 \\
XGBoost & 0.212 $\pm$ 0.001 & \underline{\textbf{0.208 $\pm$ 0.001}} & \underline{\textbf{\textcolor{green!50!black}{0.208 $\pm$ 0.001}}} \\
RealMLP & 0.216 $\pm$ 0.001 & 0.209 $\pm$ 0.001 & \underline{\textbf{0.208 $\pm$ 0.001}} \\
TabM & 0.211 $\pm$ 0.001 & 0.209 $\pm$ 0.001 & 0.208 $\pm$ 0.001 \\
TabDPT & - & - & - \\
TabPFN-2.6 & - & - & - \\
TabICLv2 & 0.224 $\pm$ 0.002 & - & - \\
\bottomrule
\end{tabular}
\space
  \end{minipage}\hfill
  \begin{minipage}[t]{0.48\textwidth}
    \centering
    \vspace{0pt}
    \includegraphics[width=\linewidth]{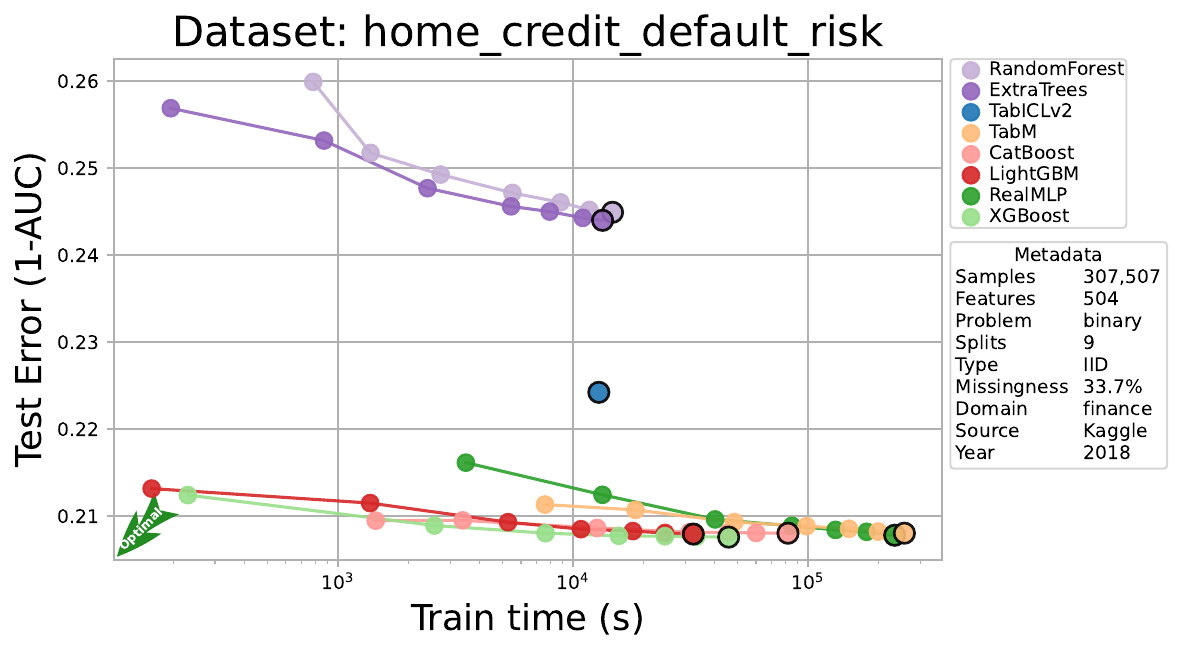}
  \end{minipage}

  \captionof{figure}{\textbf{home\_credit\_default\_risk}: per-method test error (left) and HPO Pareto trajectory (right).}
  \label{fig:perdataset_home_credit_default_risk}
\end{center}
\space
\begin{center}
  \begin{minipage}[t]{0.48\textwidth}
    \centering
    \vspace{0pt}
    \scriptsize
\begin{tabular}{l@{\quad}lll}
\toprule
 & Default & Tuned & Tuned + Ens. \\
\midrule
Linear & 0.160 & 0.157 & 0.153 \\
RandomForest & 0.172 & 0.172 & 0.169 \\
ExtraTrees & 0.181 & 0.181 & 0.169 \\
CatBoost & 0.134 & 0.130 & 0.131 \\
LightGBM & 0.146 & 0.138 & 0.136 \\
XGBoost & 0.133 & 0.129 & 0.128 \\
RealMLP & 0.145 & \textcolor{green!50!black}{0.127} & 0.128 \\
TabM & 0.130 & 0.131 & 0.130 \\
TabDPT & - & - & - \\
TabPFN-2.6 & - & - & - \\
TabICLv2 & 0.164 & - & - \\
\bottomrule
\end{tabular}
\space
  \end{minipage}\hfill
  \begin{minipage}[t]{0.48\textwidth}
    \centering
    \vspace{0pt}
    \includegraphics[width=\linewidth]{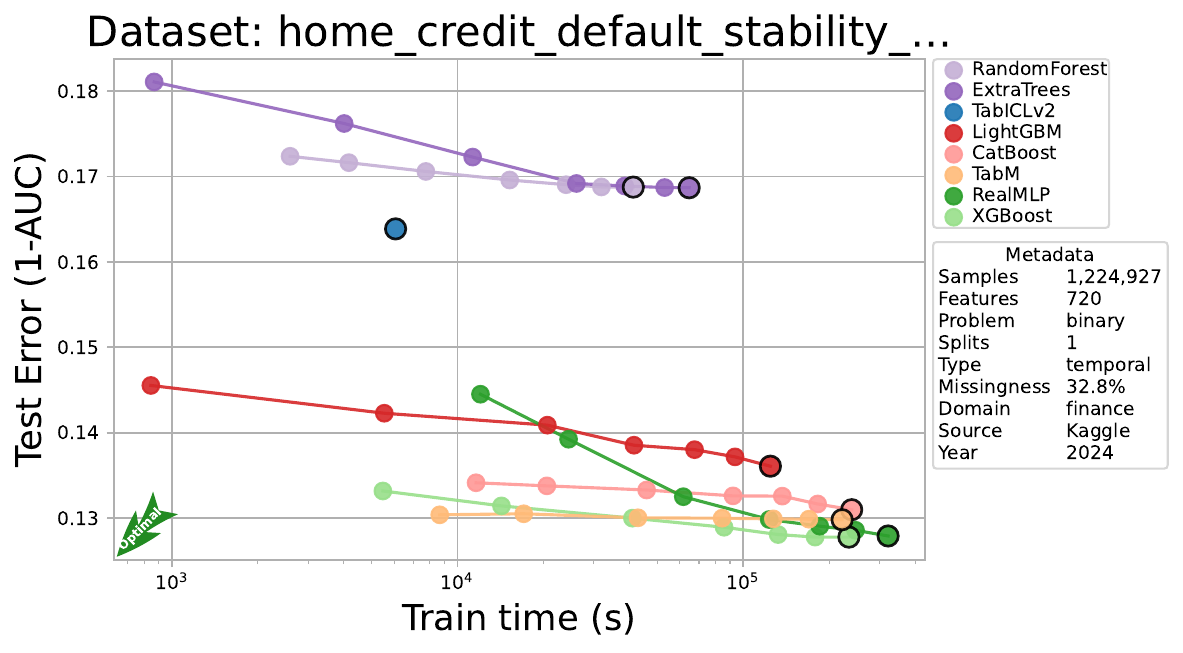}
  \end{minipage}

  \captionof{figure}{\textbf{home\_credit\_default\_stability\_1m}: per-method test error (left) and HPO Pareto trajectory (right).}
  \label{fig:perdataset_home_credit_default_stability_1m}
\end{center}
\space
\begin{center}
  \begin{minipage}[t]{0.48\textwidth}
    \centering
    \vspace{0pt}
    \scriptsize
\begin{tabular}{l@{\quad}lll}
\toprule
 & Default & Tuned & Tuned + Ens. \\
\midrule
Linear & 0.206 $\pm$ 0.010 & 0.200 $\pm$ 0.010 & 0.197 $\pm$ 0.009 \\
RandomForest & 0.042 $\pm$ 0.005 & 0.042 $\pm$ 0.005 & 0.042 $\pm$ 0.005 \\
ExtraTrees & 0.034 $\pm$ 0.004 & 0.034 $\pm$ 0.004 & 0.034 $\pm$ 0.004 \\
CatBoost & 0.032 $\pm$ 0.003 & 0.030 $\pm$ 0.004 & 0.029 $\pm$ 0.004 \\
LightGBM & 0.035 $\pm$ 0.004 & 0.032 $\pm$ 0.002 & 0.032 $\pm$ 0.003 \\
XGBoost & 0.039 $\pm$ 0.004 & 0.032 $\pm$ 0.003 & 0.032 $\pm$ 0.004 \\
RealMLP & 0.023 $\pm$ 0.005 & \underline{0.013 $\pm$ 0.003} & \underline{0.010 $\pm$ 0.002} \\
TabM & 0.033 $\pm$ 0.004 & 0.026 $\pm$ 0.004 & 0.025 $\pm$ 0.003 \\
TabDPT & 0.007 $\pm$ 0.002 & - & - \\
TabPFN-2.6 & \underline{\textbf{\textcolor{green!50!black}{0.001 $\pm$ 0.000}}} & - & - \\
TabICLv2 & 0.003 $\pm$ 0.001 & - & - \\
\bottomrule
\end{tabular}
\space
  \end{minipage}\hfill
  \begin{minipage}[t]{0.48\textwidth}
    \centering
    \vspace{0pt}
    \includegraphics[width=\linewidth]{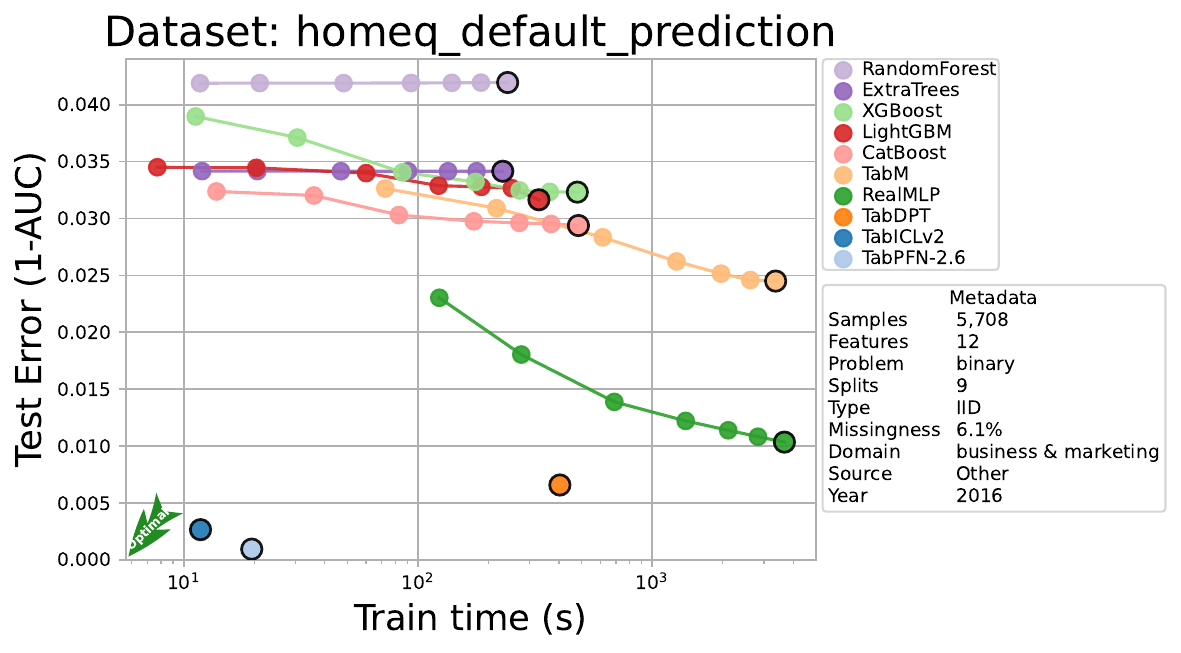}
  \end{minipage}

  \captionof{figure}{\textbf{homeq\_default\_prediction}: per-method test error (left) and HPO Pareto trajectory (right).}
  \label{fig:perdataset_homeq_default_prediction}
\end{center}
\space
\begin{center}
  \begin{minipage}[t]{0.48\textwidth}
    \centering
    \vspace{0pt}
    \scriptsize
\begin{tabular}{l@{\quad}lll}
\toprule
 & Default & Tuned & Tuned + Ens. \\
\midrule
Linear & 0.056 $\pm$ 0.000 & 0.050 $\pm$ 0.000 & 0.050 $\pm$ 0.000 \\
RandomForest & 0.043 $\pm$ 0.001 & 0.040 $\pm$ 0.001 & 0.039 $\pm$ 0.000 \\
ExtraTrees & 0.045 $\pm$ 0.001 & 0.039 $\pm$ 0.001 & 0.039 $\pm$ 0.001 \\
CatBoost & \underline{\textbf{0.032 $\pm$ 0.000}} & \underline{0.032 $\pm$ 0.001} & \underline{\textbf{\textcolor{green!50!black}{0.032 $\pm$ 0.001}}} \\
LightGBM & 0.034 $\pm$ 0.001 & 0.033 $\pm$ 0.001 & 0.033 $\pm$ 0.001 \\
XGBoost & 0.033 $\pm$ 0.001 & 0.033 $\pm$ 0.001 & 0.033 $\pm$ 0.001 \\
RealMLP & 0.034 $\pm$ 0.001 & 0.033 $\pm$ 0.001 & 0.033 $\pm$ 0.001 \\
TabM & 0.033 $\pm$ 0.001 & 0.033 $\pm$ 0.001 & 0.033 $\pm$ 0.001 \\
TabDPT & - & - & - \\
TabPFN-2.6 & - & - & - \\
TabICLv2 & 0.042 $\pm$ 0.001 & - & - \\
\bottomrule
\end{tabular}
\space
  \end{minipage}\hfill
  \begin{minipage}[t]{0.48\textwidth}
    \centering
    \vspace{0pt}
    \includegraphics[width=\linewidth]{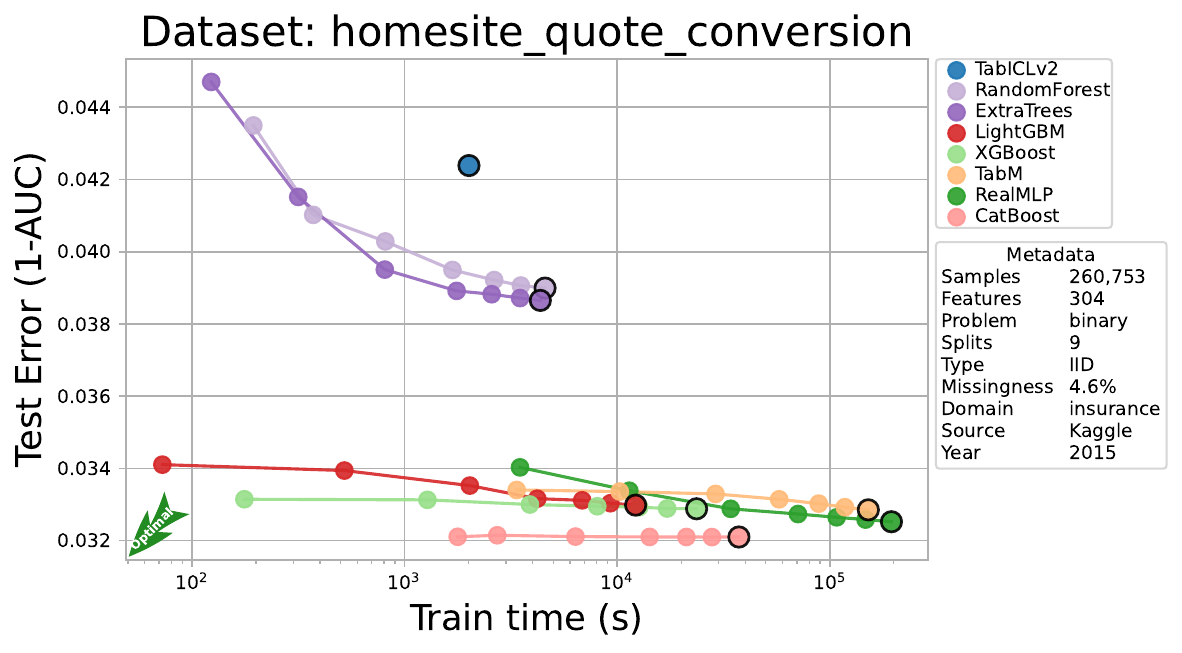}
  \end{minipage}

  \captionof{figure}{\textbf{homesite\_quote\_conversion}: per-method test error (left) and HPO Pareto trajectory (right).}
  \label{fig:perdataset_homesite_quote_conversion}
\end{center}
\space
\begin{center}
  \begin{minipage}[t]{0.48\textwidth}
    \centering
    \vspace{0pt}
    \scriptsize
\begin{tabular}{l@{\quad}lll}
\toprule
 & Default & Tuned & Tuned + Ens. \\
\midrule
Linear & 0.854 $\pm$ 0.081 & 0.750 $\pm$ 0.046 & 0.749 $\pm$ 0.044 \\
RandomForest & 0.725 $\pm$ 0.035 & 0.722 $\pm$ 0.034 & 0.722 $\pm$ 0.036 \\
ExtraTrees & 0.727 $\pm$ 0.033 & 0.730 $\pm$ 0.050 & 0.725 $\pm$ 0.041 \\
CatBoost & 0.732 $\pm$ 0.032 & 0.735 $\pm$ 0.033 & 0.733 $\pm$ 0.032 \\
LightGBM & 0.731 $\pm$ 0.033 & 0.714 $\pm$ 0.039 & 0.712 $\pm$ 0.037 \\
XGBoost & 0.734 $\pm$ 0.034 & 0.713 $\pm$ 0.038 & 0.713 $\pm$ 0.038 \\
RealMLP & 0.820 $\pm$ 0.051 & 0.717 $\pm$ 0.042 & 0.707 $\pm$ 0.037 \\
TabM & \underline{0.702 $\pm$ 0.033} & \underline{\textbf{\textcolor{green!50!black}{0.696 $\pm$ 0.037}}} & \underline{\textbf{0.696 $\pm$ 0.036}} \\
TabDPT & 0.726 $\pm$ 0.045 & - & - \\
TabPFN-2.6 & 0.720 $\pm$ 0.049 & - & - \\
TabICLv2 & \underline{\textbf{0.701 $\pm$ 0.049}} & - & - \\
\bottomrule
\end{tabular}
\space
  \end{minipage}\hfill
  \begin{minipage}[t]{0.48\textwidth}
    \centering
    \vspace{0pt}
    \includegraphics[width=\linewidth]{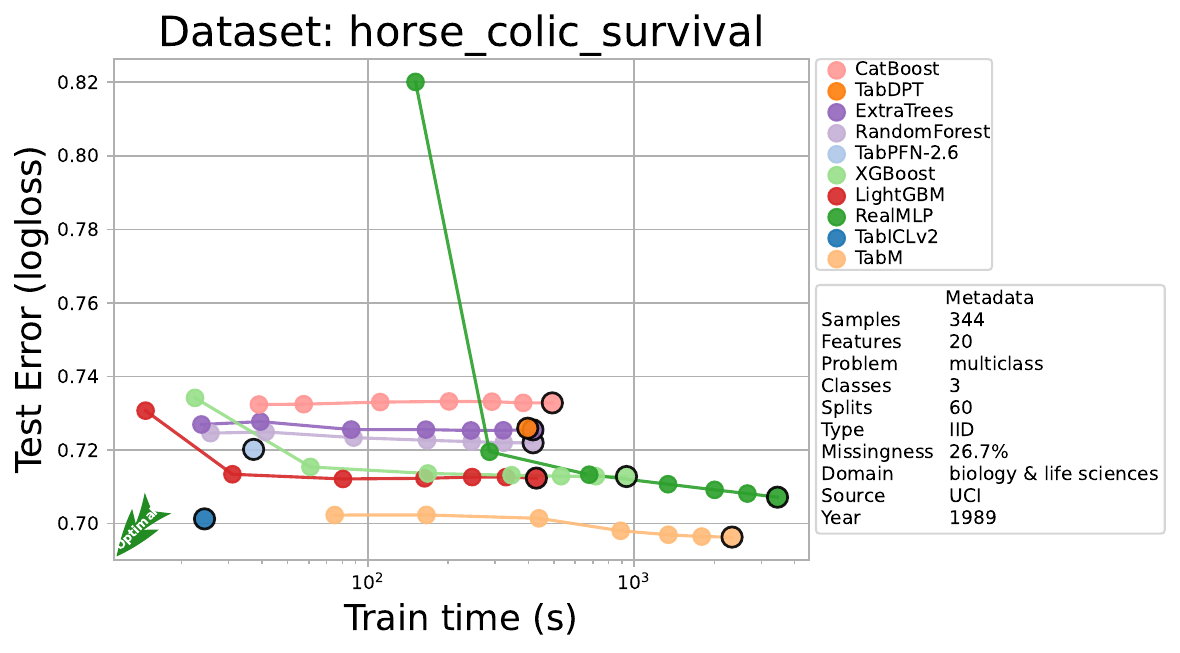}
  \end{minipage}

  \captionof{figure}{\textbf{horse\_colic\_survival}: per-method test error (left) and HPO Pareto trajectory (right).}
  \label{fig:perdataset_horse_colic_survival}
\end{center}
\space
\begin{center}
  \begin{minipage}[t]{0.48\textwidth}
    \centering
    \vspace{0pt}
    \scriptsize
\begin{tabular}{l@{\quad}lll}
\toprule
 & Default & Tuned & Tuned + Ens. \\
\midrule
Linear & 0.207 $\pm$ 0.020 & 0.207 $\pm$ 0.022 & 0.207 $\pm$ 0.022 \\
RandomForest & 0.278 $\pm$ 0.035 & 0.269 $\pm$ 0.035 & 0.273 $\pm$ 0.033 \\
ExtraTrees & 0.304 $\pm$ 0.033 & 0.275 $\pm$ 0.029 & 0.275 $\pm$ 0.030 \\
CatBoost & \underline{\textbf{\textcolor{green!50!black}{0.179 $\pm$ 0.017}}} & \underline{\textbf{0.181 $\pm$ 0.021}} & \underline{\textbf{0.182 $\pm$ 0.021}} \\
LightGBM & 0.203 $\pm$ 0.022 & 0.201 $\pm$ 0.022 & 0.199 $\pm$ 0.022 \\
XGBoost & 0.198 $\pm$ 0.026 & 0.195 $\pm$ 0.025 & 0.197 $\pm$ 0.025 \\
RealMLP & \underline{\textbf{0.183 $\pm$ 0.023}} & 0.208 $\pm$ 0.027 & \underline{0.193 $\pm$ 0.022} \\
TabM & 0.190 $\pm$ 0.025 & 0.200 $\pm$ 0.021 & 0.193 $\pm$ 0.024 \\
TabDPT & 0.324 $\pm$ 0.040 & - & - \\
TabPFN-2.6 & 0.213 $\pm$ 0.026 & - & - \\
TabICLv2 & 0.223 $\pm$ 0.024 & - & - \\
\bottomrule
\end{tabular}
\space
  \end{minipage}\hfill
  \begin{minipage}[t]{0.48\textwidth}
    \centering
    \vspace{0pt}
    \includegraphics[width=\linewidth]{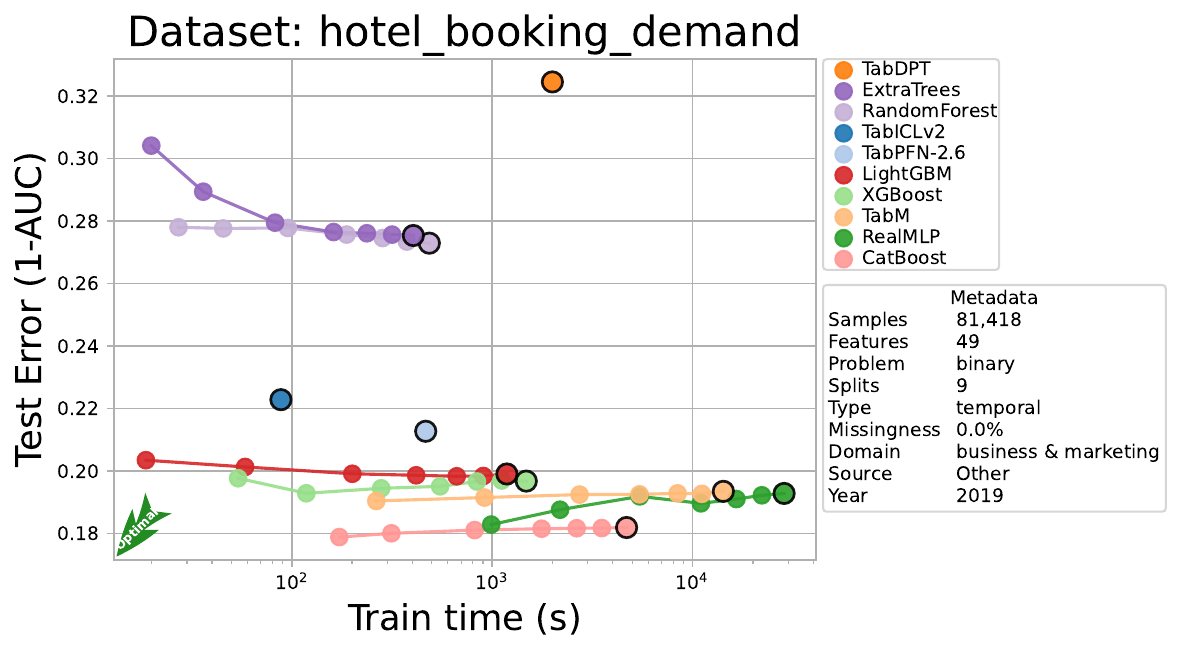}
  \end{minipage}

  \captionof{figure}{\textbf{hotel\_booking\_demand}: per-method test error (left) and HPO Pareto trajectory (right).}
  \label{fig:perdataset_hotel_booking_demand}
\end{center}
\space
\begin{center}
  \begin{minipage}[t]{0.48\textwidth}
    \centering
    \vspace{0pt}
    \scriptsize
\begin{tabular}{l@{\quad}lll}
\toprule
 & Default & Tuned & Tuned + Ens. \\
\midrule
Linear & 0.313 $\pm$ 0.005 & 0.313 $\pm$ 0.005 & 0.312 $\pm$ 0.005 \\
RandomForest & 0.224 $\pm$ 0.004 & 0.224 $\pm$ 0.004 & 0.224 $\pm$ 0.004 \\
ExtraTrees & 0.235 $\pm$ 0.004 & 0.231 $\pm$ 0.004 & 0.231 $\pm$ 0.004 \\
CatBoost & 0.208 $\pm$ 0.004 & 0.208 $\pm$ 0.004 & 0.207 $\pm$ 0.004 \\
LightGBM & 0.213 $\pm$ 0.004 & 0.210 $\pm$ 0.004 & 0.211 $\pm$ 0.004 \\
XGBoost & 0.211 $\pm$ 0.004 & 0.212 $\pm$ 0.004 & 0.211 $\pm$ 0.004 \\
RealMLP & 0.221 $\pm$ 0.005 & \underline{0.199 $\pm$ 0.004} & \underline{0.198 $\pm$ 0.004} \\
TabM & 0.209 $\pm$ 0.003 & 0.204 $\pm$ 0.003 & 0.202 $\pm$ 0.003 \\
TabDPT & 0.202 $\pm$ 0.003 & - & - \\
TabPFN-2.6 & 0.194 $\pm$ 0.004 & - & - \\
TabICLv2 & \underline{\textbf{\textcolor{green!50!black}{0.191 $\pm$ 0.004}}} & - & - \\
\bottomrule
\end{tabular}
\space
  \end{minipage}\hfill
  \begin{minipage}[t]{0.48\textwidth}
    \centering
    \vspace{0pt}
    \includegraphics[width=\linewidth]{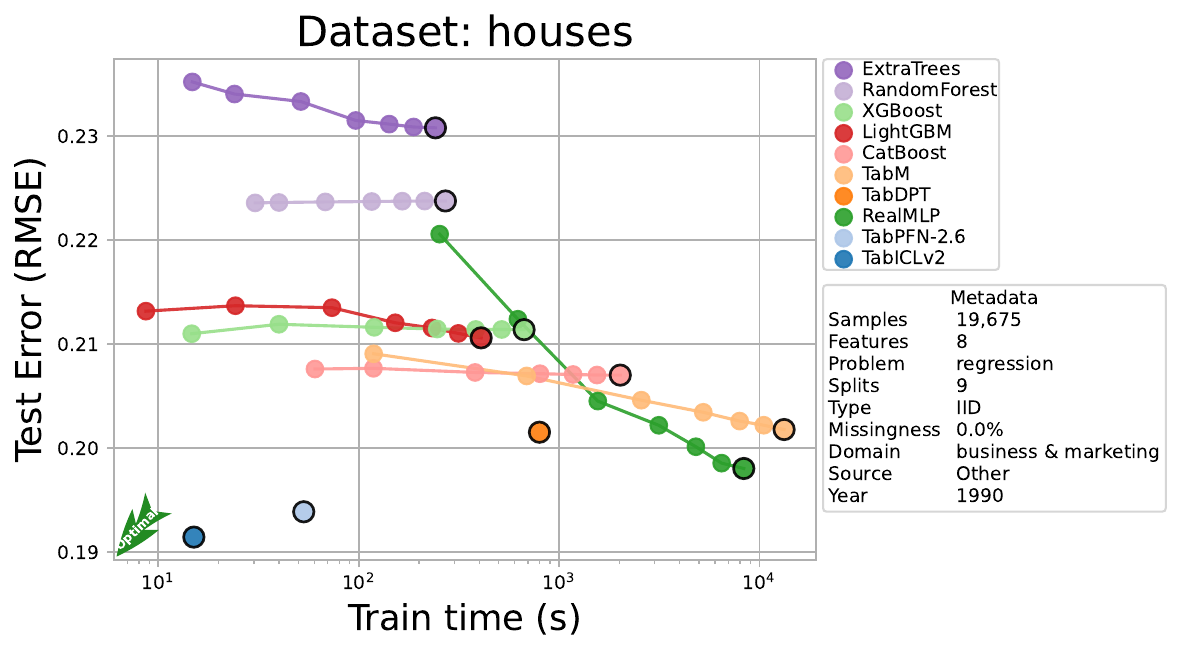}
  \end{minipage}

  \captionof{figure}{\textbf{houses}: per-method test error (left) and HPO Pareto trajectory (right).}
  \label{fig:perdataset_houses}
\end{center}
\space
\begin{center}
  \begin{minipage}[t]{0.48\textwidth}
    \centering
    \vspace{0pt}
    \scriptsize
\begin{tabular}{l@{\quad}lll}
\toprule
 & Default & Tuned & Tuned + Ens. \\
\midrule
Linear & 0.204 $\pm$ 0.004 & 0.203 $\pm$ 0.003 & 0.203 $\pm$ 0.003 \\
RandomForest & 0.208 $\pm$ 0.003 & 0.198 $\pm$ 0.002 & 0.198 $\pm$ 0.002 \\
ExtraTrees & 0.212 $\pm$ 0.004 & 0.200 $\pm$ 0.003 & 0.199 $\pm$ 0.003 \\
CatBoost & \underline{\textbf{0.195 $\pm$ 0.003}} & \underline{\textbf{0.196 $\pm$ 0.003}} & \underline{0.196 $\pm$ 0.003} \\
LightGBM & 0.199 $\pm$ 0.004 & 0.199 $\pm$ 0.004 & 0.197 $\pm$ 0.003 \\
XGBoost & \underline{\textbf{0.195 $\pm$ 0.003}} & \underline{0.196 $\pm$ 0.003} & \underline{\textbf{0.195 $\pm$ 0.003}} \\
RealMLP & 0.199 $\pm$ 0.003 & \underline{0.197 $\pm$ 0.004} & 0.197 $\pm$ 0.004 \\
TabM & 0.199 $\pm$ 0.003 & 0.199 $\pm$ 0.003 & 0.198 $\pm$ 0.003 \\
TabDPT & 0.197 $\pm$ 0.004 & - & - \\
TabPFN-2.6 & \underline{\textbf{0.194 $\pm$ 0.004}} & - & - \\
TabICLv2 & \underline{\textbf{\textcolor{green!50!black}{0.193 $\pm$ 0.004}}} & - & - \\
\bottomrule
\end{tabular}
\space
  \end{minipage}\hfill
  \begin{minipage}[t]{0.48\textwidth}
    \centering
    \vspace{0pt}
    \includegraphics[width=\linewidth]{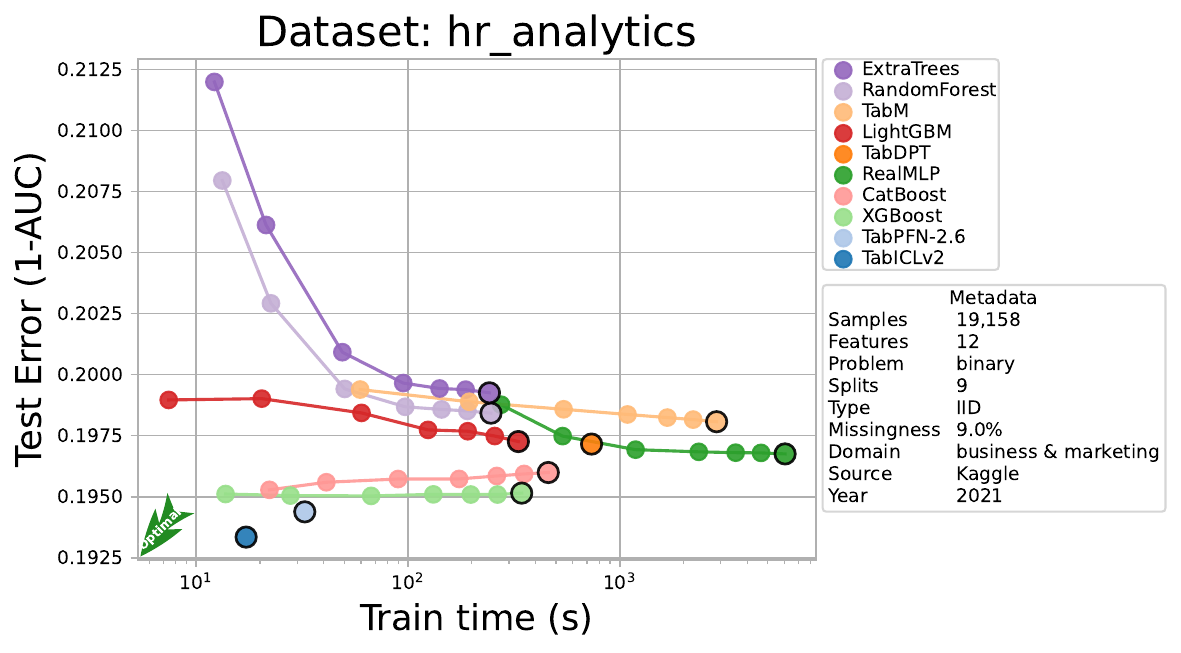}
  \end{minipage}

  \captionof{figure}{\textbf{hr\_analytics}: per-method test error (left) and HPO Pareto trajectory (right).}
  \label{fig:perdataset_hr_analytics}
\end{center}
\space
\begin{center}
  \begin{minipage}[t]{0.48\textwidth}
    \centering
    \vspace{0pt}
    \scriptsize
\begin{tabular}{l@{\quad}lll}
\toprule
 & Default & Tuned & Tuned + Ens. \\
\midrule
Linear & \underline{\textbf{0.141 $\pm$ 0.009}} & \underline{\textbf{0.140 $\pm$ 0.009}} & \underline{\textbf{0.136 $\pm$ 0.009}} \\
RandomForest & \underline{\textbf{0.145 $\pm$ 0.038}} & \underline{\textbf{0.140 $\pm$ 0.010}} & \underline{\textbf{0.133 $\pm$ 0.012}} \\
ExtraTrees & \underline{\textbf{0.123 $\pm$ 0.007}} & \underline{\textbf{0.123 $\pm$ 0.007}} & \underline{\textbf{0.121 $\pm$ 0.005}} \\
CatBoost & \underline{\textbf{0.096 $\pm$ 0.010}} & \underline{\textbf{0.092 $\pm$ 0.010}} & \underline{\textbf{0.091 $\pm$ 0.006}} \\
LightGBM & \underline{\textbf{0.120 $\pm$ 0.017}} & \underline{\textbf{0.105 $\pm$ 0.015}} & \underline{\textbf{0.093 $\pm$ 0.012}} \\
XGBoost & \underline{\textbf{0.103 $\pm$ 0.010}} & \underline{\textbf{\textcolor{green!50!black}{0.091 $\pm$ 0.012}}} & \underline{\textbf{0.091 $\pm$ 0.016}} \\
RealMLP & \underline{\textbf{0.110 $\pm$ 0.015}} & \underline{\textbf{0.104 $\pm$ 0.014}} & \underline{\textbf{0.093 $\pm$ 0.010}} \\
TabM & \underline{\textbf{0.111 $\pm$ 0.010}} & \underline{\textbf{0.110 $\pm$ 0.010}} & \underline{\textbf{0.103 $\pm$ 0.011}} \\
TabDPT & - & - & - \\
TabPFN-2.6 & - & - & - \\
TabICLv2 & \underline{\textbf{0.121 $\pm$ 0.014}} & - & - \\
\bottomrule
\end{tabular}
\space
  \end{minipage}\hfill
  \begin{minipage}[t]{0.48\textwidth}
    \centering
    \vspace{0pt}
    \includegraphics[width=\linewidth]{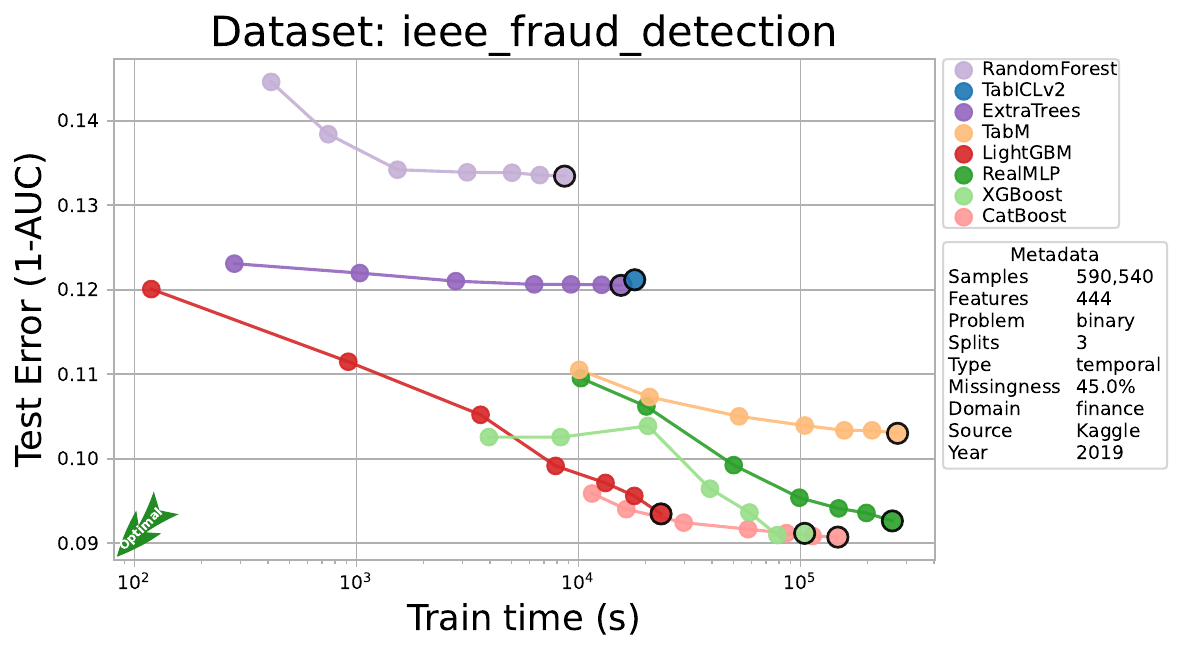}
  \end{minipage}

  \captionof{figure}{\textbf{ieee\_fraud\_detection}: per-method test error (left) and HPO Pareto trajectory (right).}
  \label{fig:perdataset_ieee_fraud_detection}
\end{center}
\space
\begin{center}
  \begin{minipage}[t]{0.48\textwidth}
    \centering
    \vspace{0pt}
    \scriptsize
\begin{tabular}{l@{\quad}lll}
\toprule
 & Default & Tuned & Tuned + Ens. \\
\midrule
Linear & 0.473 $\pm$ 0.007 & 0.472 $\pm$ 0.007 & 0.470 $\pm$ 0.008 \\
RandomForest & 0.454 $\pm$ 0.007 & 0.434 $\pm$ 0.007 & 0.436 $\pm$ 0.006 \\
ExtraTrees & 0.462 $\pm$ 0.008 & 0.449 $\pm$ 0.008 & 0.448 $\pm$ 0.008 \\
CatBoost & 0.352 $\pm$ 0.006 & \underline{0.353 $\pm$ 0.006} & 0.352 $\pm$ 0.006 \\
LightGBM & 0.364 $\pm$ 0.006 & 0.358 $\pm$ 0.007 & 0.356 $\pm$ 0.007 \\
XGBoost & 0.369 $\pm$ 0.006 & 0.359 $\pm$ 0.007 & 0.357 $\pm$ 0.007 \\
RealMLP & 0.367 $\pm$ 0.006 & \underline{0.351 $\pm$ 0.007} & \underline{\textbf{0.348 $\pm$ 0.007}} \\
TabM & 0.354 $\pm$ 0.007 & 0.353 $\pm$ 0.007 & 0.350 $\pm$ 0.007 \\
TabDPT & 0.440 $\pm$ 0.008 & - & - \\
TabPFN-2.6 & \underline{\textbf{\textcolor{green!50!black}{0.346 $\pm$ 0.005}}} & - & - \\
TabICLv2 & 0.352 $\pm$ 0.005 & - & - \\
\bottomrule
\end{tabular}
\space
  \end{minipage}\hfill
  \begin{minipage}[t]{0.48\textwidth}
    \centering
    \vspace{0pt}
    \includegraphics[width=\linewidth]{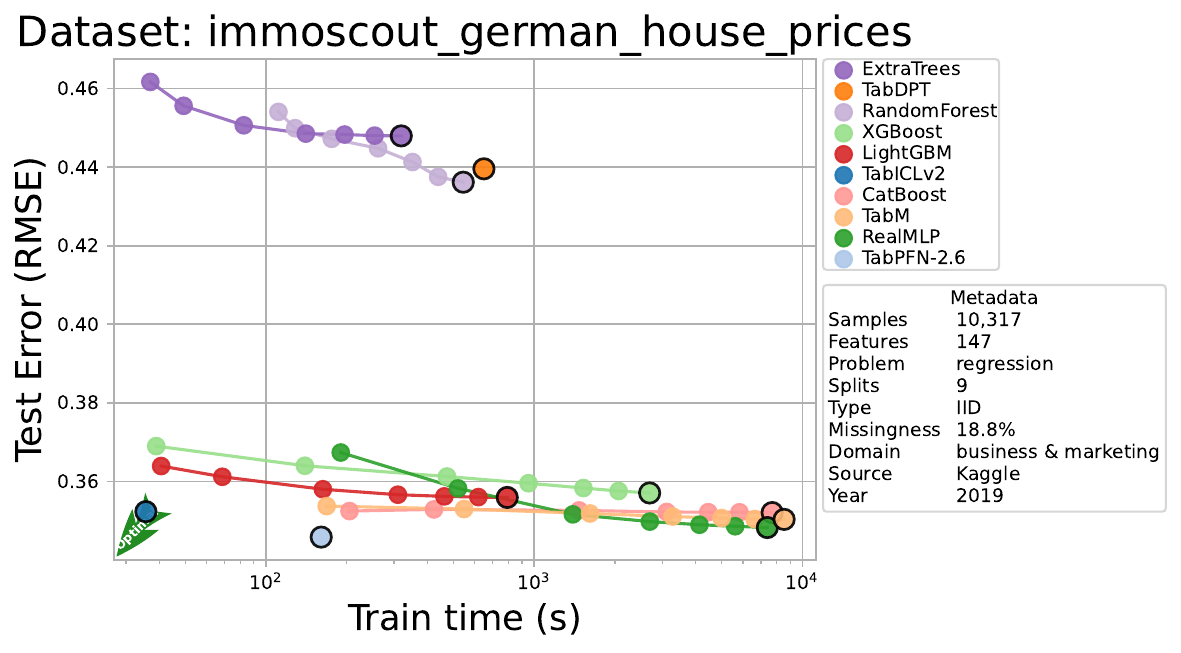}
  \end{minipage}

  \captionof{figure}{\textbf{immoscout\_german\_house\_prices}: per-method test error (left) and HPO Pareto trajectory (right).}
  \label{fig:perdataset_immoscout_german_house_prices}
\end{center}
\space
\begin{center}
  \begin{minipage}[t]{0.48\textwidth}
    \centering
    \vspace{0pt}
    \scriptsize
\begin{tabular}{l@{\quad}lll}
\toprule
 & Default & Tuned & Tuned + Ens. \\
\midrule
Linear & 0.264 $\pm$ 0.006 & 0.264 $\pm$ 0.006 & 0.264 $\pm$ 0.006 \\
RandomForest & 0.186 $\pm$ 0.007 & 0.182 $\pm$ 0.009 & 0.177 $\pm$ 0.007 \\
ExtraTrees & 0.202 $\pm$ 0.008 & 0.195 $\pm$ 0.008 & 0.189 $\pm$ 0.008 \\
CatBoost & 0.160 $\pm$ 0.005 & 0.156 $\pm$ 0.006 & 0.154 $\pm$ 0.006 \\
LightGBM & 0.162 $\pm$ 0.005 & 0.160 $\pm$ 0.006 & 0.157 $\pm$ 0.005 \\
XGBoost & 0.166 $\pm$ 0.006 & 0.158 $\pm$ 0.006 & 0.157 $\pm$ 0.005 \\
RealMLP & 0.162 $\pm$ 0.004 & 0.153 $\pm$ 0.006 & 0.150 $\pm$ 0.005 \\
TabM & 0.151 $\pm$ 0.005 & \underline{\textbf{0.149 $\pm$ 0.005}} & \underline{\textbf{\textcolor{green!50!black}{0.148 $\pm$ 0.005}}} \\
TabDPT & 0.188 $\pm$ 0.008 & - & - \\
TabPFN-2.6 & \underline{0.150 $\pm$ 0.006} & - & - \\
TabICLv2 & 0.154 $\pm$ 0.006 & - & - \\
\bottomrule
\end{tabular}
\space
  \end{minipage}\hfill
  \begin{minipage}[t]{0.48\textwidth}
    \centering
    \vspace{0pt}
    \includegraphics[width=\linewidth]{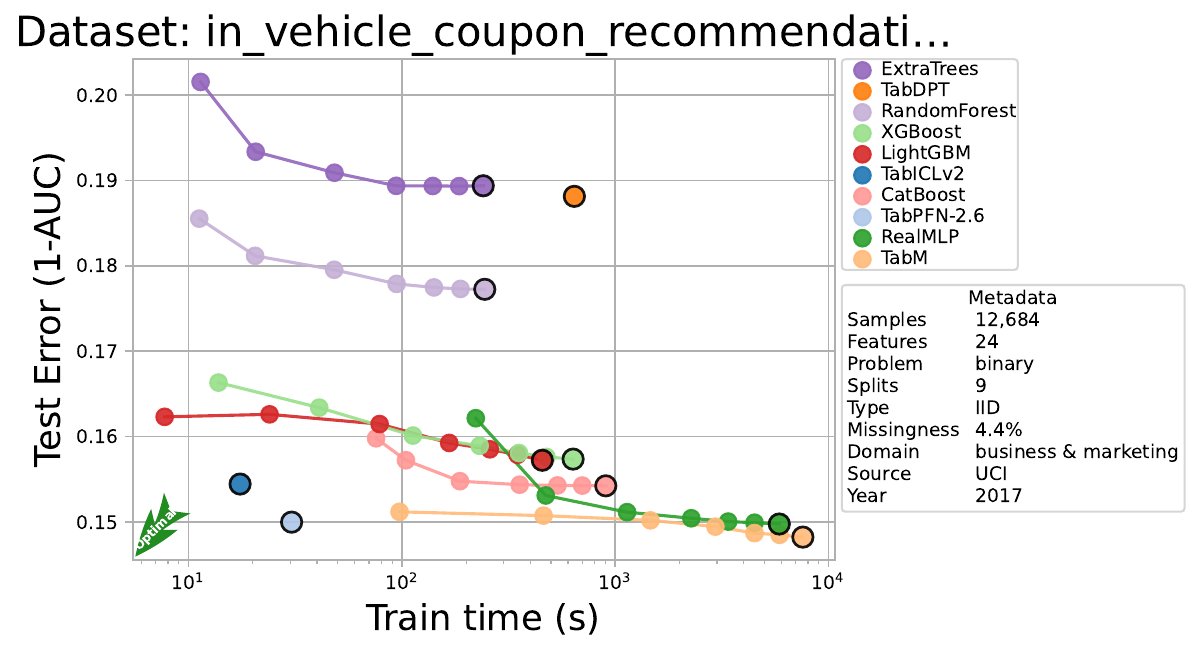}
  \end{minipage}

  \captionof{figure}{\textbf{in\_vehicle\_coupon\_recommendation}: per-method test error (left) and HPO Pareto trajectory (right).}
  \label{fig:perdataset_in_vehicle_coupon_recommendation}
\end{center}
\space
\begin{center}
  \begin{minipage}[t]{0.48\textwidth}
    \centering
    \vspace{0pt}
    \scriptsize
\begin{tabular}{l@{\quad}lll}
\toprule
 & Default & Tuned & Tuned + Ens. \\
\midrule
Linear & 0.257 $\pm$ 0.030 & 0.263 $\pm$ 0.029 & 0.255 $\pm$ 0.030 \\
RandomForest & 0.256 $\pm$ 0.029 & 0.261 $\pm$ 0.029 & 0.261 $\pm$ 0.029 \\
ExtraTrees & 0.241 $\pm$ 0.030 & \underline{0.247 $\pm$ 0.029} & \underline{0.245 $\pm$ 0.031} \\
CatBoost & 0.255 $\pm$ 0.032 & 0.252 $\pm$ 0.031 & 0.251 $\pm$ 0.032 \\
LightGBM & 0.273 $\pm$ 0.031 & 0.266 $\pm$ 0.032 & 0.265 $\pm$ 0.030 \\
XGBoost & 0.273 $\pm$ 0.032 & 0.269 $\pm$ 0.028 & 0.269 $\pm$ 0.029 \\
RealMLP & 0.262 $\pm$ 0.031 & 0.259 $\pm$ 0.031 & 0.253 $\pm$ 0.031 \\
TabM & 0.255 $\pm$ 0.031 & 0.256 $\pm$ 0.031 & 0.252 $\pm$ 0.031 \\
TabDPT & \underline{\textbf{0.236 $\pm$ 0.029}} & - & - \\
TabPFN-2.6 & 0.251 $\pm$ 0.031 & - & - \\
TabICLv2 & \underline{\textbf{\textcolor{green!50!black}{0.234 $\pm$ 0.028}}} & - & - \\
\bottomrule
\end{tabular}
\space
  \end{minipage}\hfill
  \begin{minipage}[t]{0.48\textwidth}
    \centering
    \vspace{0pt}
    \includegraphics[width=\linewidth]{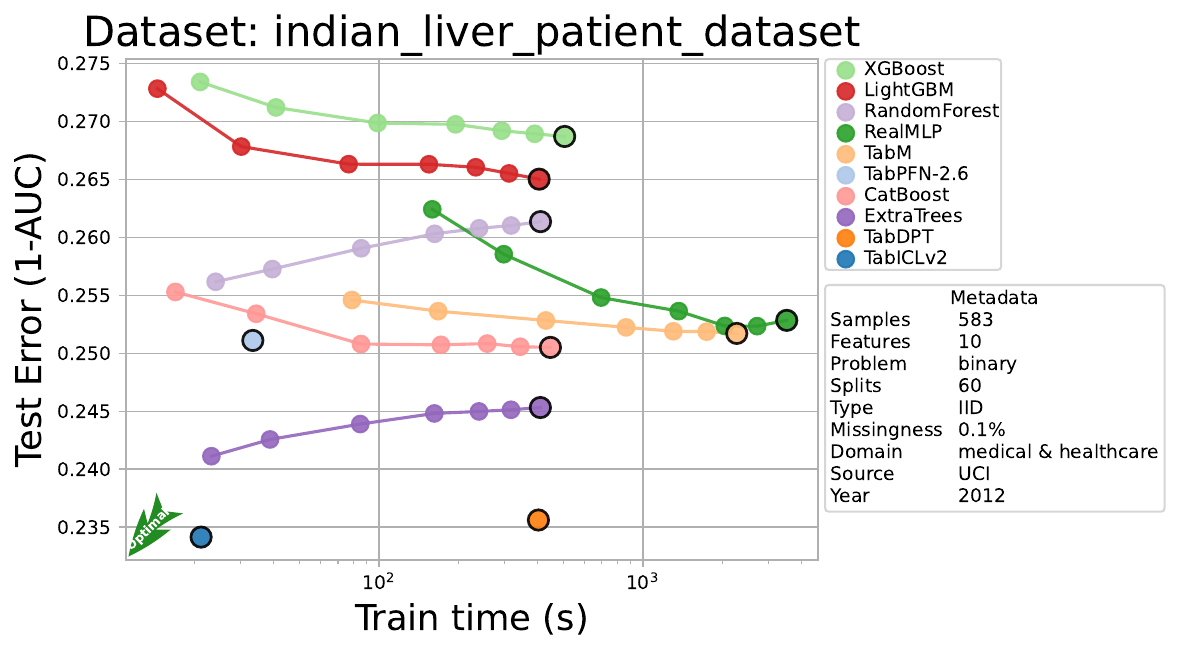}
  \end{minipage}

  \captionof{figure}{\textbf{indian\_liver\_patient\_dataset}: per-method test error (left) and HPO Pareto trajectory (right).}
  \label{fig:perdataset_indian_liver_patient_dataset}
\end{center}
\space
\begin{center}
  \begin{minipage}[t]{0.48\textwidth}
    \centering
    \vspace{0pt}
    \scriptsize
\begin{tabular}{l@{\quad}lll}
\toprule
 & Default & Tuned & Tuned + Ens. \\
\midrule
Linear & 0.074 $\pm$ 0.008 & 0.067 $\pm$ 0.007 & 0.066 $\pm$ 0.007 \\
RandomForest & 0.016 $\pm$ 0.003 & 0.017 $\pm$ 0.003 & 0.017 $\pm$ 0.003 \\
ExtraTrees & 0.015 $\pm$ 0.002 & 0.015 $\pm$ 0.003 & 0.014 $\pm$ 0.002 \\
CatBoost & 0.013 $\pm$ 0.002 & 0.011 $\pm$ 0.002 & 0.011 $\pm$ 0.002 \\
LightGBM & 0.014 $\pm$ 0.003 & 0.015 $\pm$ 0.003 & 0.013 $\pm$ 0.003 \\
XGBoost & 0.016 $\pm$ 0.003 & 0.014 $\pm$ 0.003 & 0.014 $\pm$ 0.003 \\
RealMLP & 0.010 $\pm$ 0.003 & \underline{0.007 $\pm$ 0.002} & \underline{0.007 $\pm$ 0.002} \\
TabM & 0.015 $\pm$ 0.004 & 0.009 $\pm$ 0.003 & 0.008 $\pm$ 0.002 \\
TabDPT & 0.005 $\pm$ 0.002 & - & - \\
TabPFN-2.6 & 0.005 $\pm$ 0.002 & - & - \\
TabICLv2 & \underline{\textbf{\textcolor{green!50!black}{0.004 $\pm$ 0.002}}} & - & - \\
\bottomrule
\end{tabular}
\space
  \end{minipage}\hfill
  \begin{minipage}[t]{0.48\textwidth}
    \centering
    \vspace{0pt}
    \includegraphics[width=\linewidth]{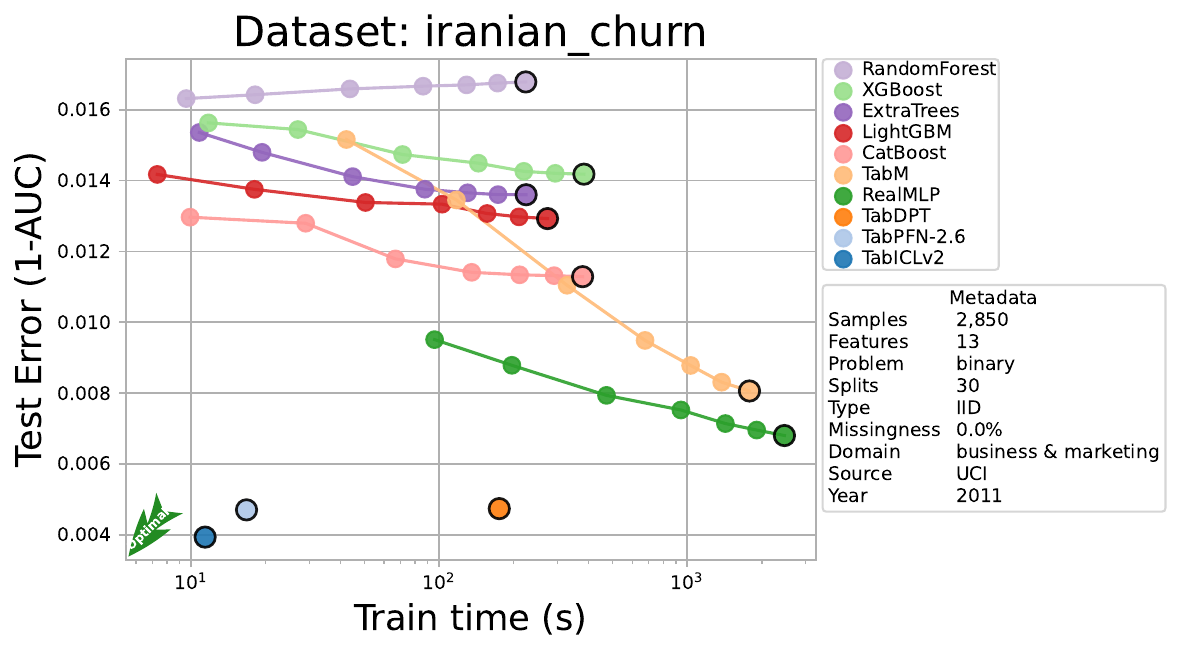}
  \end{minipage}

  \captionof{figure}{\textbf{iranian\_churn}: per-method test error (left) and HPO Pareto trajectory (right).}
  \label{fig:perdataset_iranian_churn}
\end{center}
\space
\begin{center}
  \begin{minipage}[t]{0.48\textwidth}
    \centering
    \vspace{0pt}
    \scriptsize
\begin{tabular}{l@{\quad}lll}
\toprule
 & Default & Tuned & Tuned + Ens. \\
\midrule
Linear & 0.277 $\pm$ 0.004 & 0.277 $\pm$ 0.004 & 0.277 $\pm$ 0.004 \\
RandomForest & 0.249 $\pm$ 0.007 & 0.251 $\pm$ 0.007 & 0.242 $\pm$ 0.005 \\
ExtraTrees & 0.243 $\pm$ 0.007 & \underline{0.240 $\pm$ 0.005} & \underline{0.236 $\pm$ 0.006} \\
CatBoost & 0.255 $\pm$ 0.003 & 0.251 $\pm$ 0.005 & 0.251 $\pm$ 0.003 \\
LightGBM & 0.252 $\pm$ 0.005 & 0.253 $\pm$ 0.005 & 0.251 $\pm$ 0.004 \\
XGBoost & 0.251 $\pm$ 0.004 & 0.251 $\pm$ 0.006 & 0.250 $\pm$ 0.005 \\
RealMLP & 0.269 $\pm$ 0.004 & 0.256 $\pm$ 0.007 & 0.242 $\pm$ 0.006 \\
TabM & 0.267 $\pm$ 0.005 & 0.266 $\pm$ 0.004 & 0.258 $\pm$ 0.004 \\
TabDPT & \underline{\textbf{0.224 $\pm$ 0.006}} & - & - \\
TabPFN-2.6 & 0.228 $\pm$ 0.004 & - & - \\
TabICLv2 & \underline{\textbf{\textcolor{green!50!black}{0.223 $\pm$ 0.005}}} & - & - \\
\bottomrule
\end{tabular}
\space
  \end{minipage}\hfill
  \begin{minipage}[t]{0.48\textwidth}
    \centering
    \vspace{0pt}
    \includegraphics[width=\linewidth]{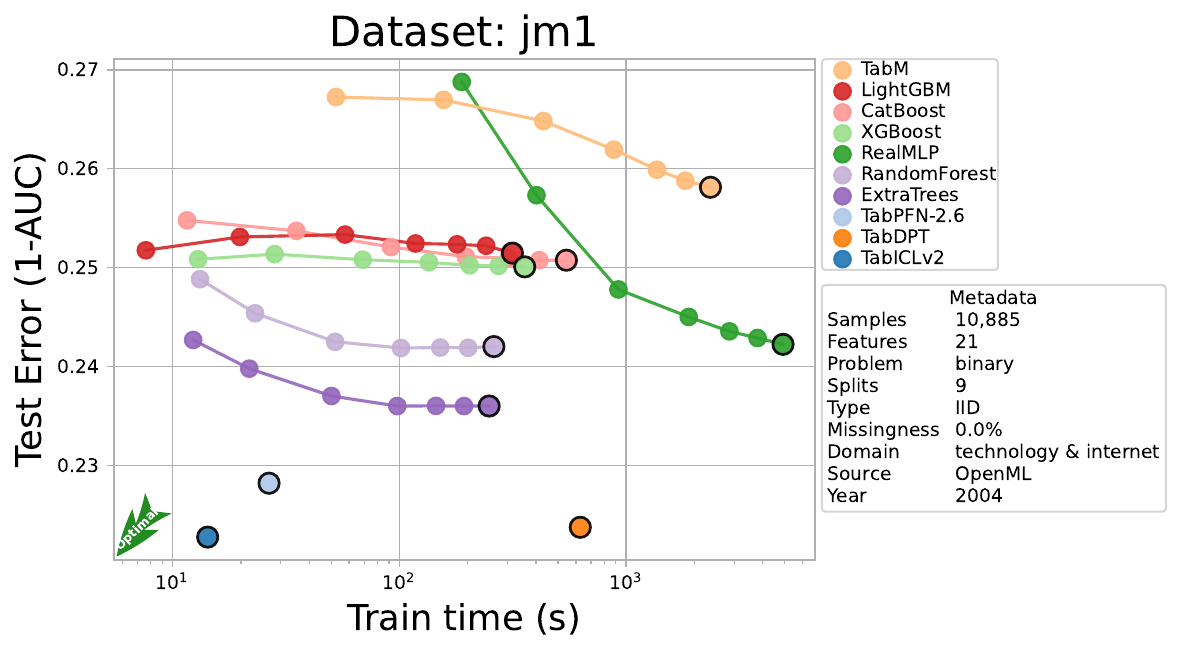}
  \end{minipage}

  \captionof{figure}{\textbf{jm1}: per-method test error (left) and HPO Pareto trajectory (right).}
  \label{fig:perdataset_jm1}
\end{center}
\space
\begin{center}
  \begin{minipage}[t]{0.48\textwidth}
    \centering
    \vspace{0pt}
    \scriptsize
\begin{tabular}{l@{\quad}lll}
\toprule
 & Default & Tuned & Tuned + Ens. \\
\midrule
Linear & 0.207 $\pm$ 0.011 & 0.184 $\pm$ 0.013 & 0.183 $\pm$ 0.012 \\
RandomForest & 0.288 $\pm$ 0.018 & 0.193 $\pm$ 0.005 & 0.193 $\pm$ 0.007 \\
ExtraTrees & 0.218 $\pm$ 0.014 & 0.188 $\pm$ 0.011 & 0.181 $\pm$ 0.010 \\
CatBoost & \underline{0.161 $\pm$ 0.010} & \underline{\textbf{\textcolor{green!50!black}{0.159 $\pm$ 0.011}}} & \underline{\textbf{0.159 $\pm$ 0.011}} \\
LightGBM & 0.203 $\pm$ 0.009 & 0.179 $\pm$ 0.011 & 0.175 $\pm$ 0.009 \\
XGBoost & 0.174 $\pm$ 0.011 & 0.167 $\pm$ 0.012 & 0.165 $\pm$ 0.010 \\
RealMLP & 0.195 $\pm$ 0.011 & 0.183 $\pm$ 0.007 & 0.180 $\pm$ 0.008 \\
TabM & 0.198 $\pm$ 0.007 & 0.185 $\pm$ 0.009 & 0.183 $\pm$ 0.009 \\
TabDPT & 0.266 $\pm$ 0.019 & - & - \\
TabPFN-2.6 & 0.173 $\pm$ 0.010 & - & - \\
TabICLv2 & 0.172 $\pm$ 0.012 & - & - \\
\bottomrule
\end{tabular}
\space
  \end{minipage}\hfill
  \begin{minipage}[t]{0.48\textwidth}
    \centering
    \vspace{0pt}
    \includegraphics[width=\linewidth]{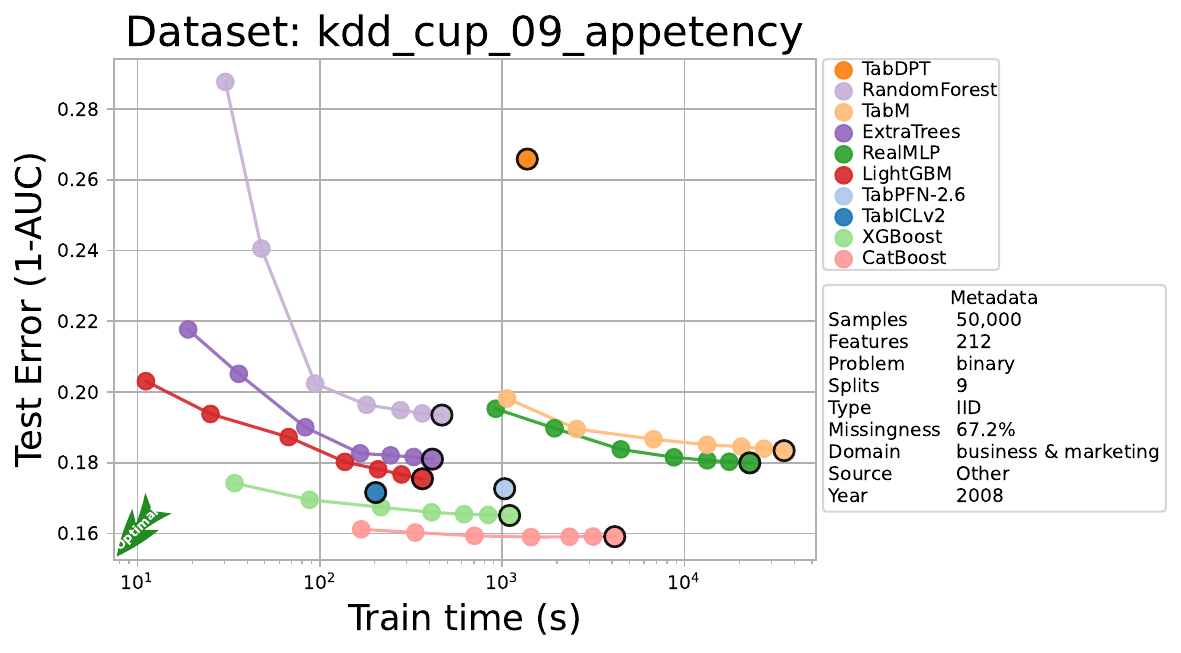}
  \end{minipage}

  \captionof{figure}{\textbf{kdd\_cup\_09\_appetency}: per-method test error (left) and HPO Pareto trajectory (right).}
  \label{fig:perdataset_kdd_cup_09_appetency}
\end{center}
\space
\begin{center}
  \begin{minipage}[t]{0.48\textwidth}
    \centering
    \vspace{0pt}
    \scriptsize
\begin{tabular}{l@{\quad}lll}
\toprule
 & Default & Tuned & Tuned + Ens. \\
\midrule
Linear & 0.254 $\pm$ 0.033 & 0.249 $\pm$ 0.031 & 0.249 $\pm$ 0.031 \\
RandomForest & 0.256 $\pm$ 0.018 & 0.249 $\pm$ 0.016 & 0.248 $\pm$ 0.014 \\
ExtraTrees & 0.249 $\pm$ 0.027 & 0.250 $\pm$ 0.025 & 0.247 $\pm$ 0.027 \\
CatBoost & \underline{\textbf{0.232 $\pm$ 0.029}} & \underline{\textbf{0.231 $\pm$ 0.027}} & \underline{\textbf{0.230 $\pm$ 0.027}} \\
LightGBM & 0.247 $\pm$ 0.022 & 0.238 $\pm$ 0.031 & 0.235 $\pm$ 0.028 \\
XGBoost & \underline{\textbf{0.231 $\pm$ 0.025}} & \underline{\textbf{0.234 $\pm$ 0.022}} & \underline{\textbf{0.231 $\pm$ 0.023}} \\
RealMLP & 0.237 $\pm$ 0.028 & \underline{\textbf{0.231 $\pm$ 0.027}} & \underline{\textbf{0.230 $\pm$ 0.027}} \\
TabM & 0.240 $\pm$ 0.030 & 0.239 $\pm$ 0.030 & 0.238 $\pm$ 0.031 \\
TabDPT & 0.248 $\pm$ 0.035 & - & - \\
TabPFN-2.6 & \underline{\textbf{\textcolor{green!50!black}{0.228 $\pm$ 0.031}}} & - & - \\
TabICLv2 & \underline{\textbf{0.232 $\pm$ 0.032}} & - & - \\
\bottomrule
\end{tabular}
\space
  \end{minipage}\hfill
  \begin{minipage}[t]{0.48\textwidth}
    \centering
    \vspace{0pt}
    \includegraphics[width=\linewidth]{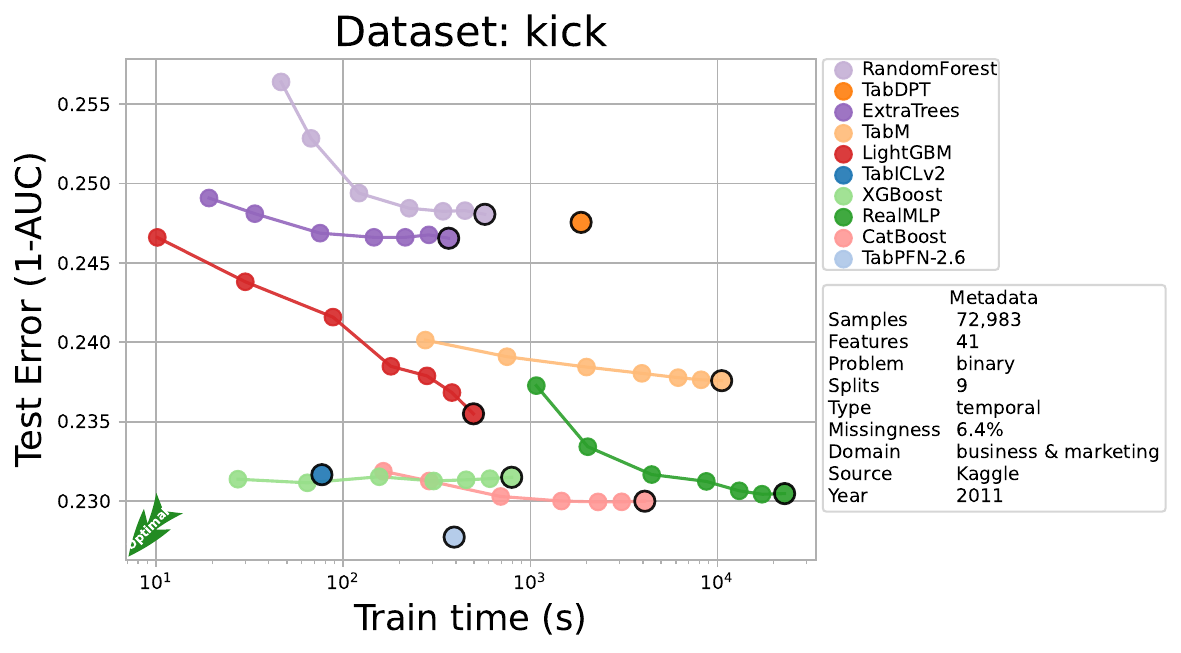}
  \end{minipage}

  \captionof{figure}{\textbf{kick}: per-method test error (left) and HPO Pareto trajectory (right).}
  \label{fig:perdataset_kick}
\end{center}
\space
\begin{center}
  \begin{minipage}[t]{0.48\textwidth}
    \centering
    \vspace{0pt}
    \scriptsize
\begin{tabular}{l@{\quad}lll}
\toprule
 & Default & Tuned & Tuned + Ens. \\
\midrule
Linear & \underline{\textbf{0.115 $\pm$ 0.018}} & \underline{\textbf{0.115 $\pm$ 0.019}} & \underline{\textbf{0.115 $\pm$ 0.018}} \\
RandomForest & \underline{\textbf{0.093 $\pm$ 0.005}} & \underline{\textbf{0.089 $\pm$ 0.003}} & \underline{\textbf{0.088 $\pm$ 0.003}} \\
ExtraTrees & \underline{\textbf{0.121 $\pm$ 0.000}} & \underline{\textbf{0.094 $\pm$ 0.006}} & \underline{\textbf{0.093 $\pm$ 0.006}} \\
CatBoost & \underline{\textbf{0.073 $\pm$ 0.006}} & \underline{\textbf{0.072 $\pm$ 0.005}} & \underline{\textbf{0.072 $\pm$ 0.005}} \\
LightGBM & \underline{\textbf{0.077 $\pm$ 0.006}} & \underline{\textbf{0.075 $\pm$ 0.007}} & \underline{\textbf{0.074 $\pm$ 0.006}} \\
XGBoost & \underline{\textbf{0.078 $\pm$ 0.008}} & \underline{\textbf{\textcolor{green!50!black}{0.072 $\pm$ 0.004}}} & \underline{\textbf{0.072 $\pm$ 0.005}} \\
RealMLP & \underline{\textbf{0.076 $\pm$ 0.008}} & \underline{\textbf{0.072 $\pm$ 0.005}} & \underline{\textbf{0.072 $\pm$ 0.006}} \\
TabM & \underline{\textbf{0.073 $\pm$ 0.004}} & \underline{\textbf{0.073 $\pm$ 0.006}} & \underline{\textbf{0.072 $\pm$ 0.004}} \\
TabDPT & - & - & - \\
TabPFN-2.6 & - & - & - \\
TabICLv2 & \underline{\textbf{0.079 $\pm$ 0.003}} & - & - \\
\bottomrule
\end{tabular}
\space
  \end{minipage}\hfill
  \begin{minipage}[t]{0.48\textwidth}
    \centering
    \vspace{0pt}
    \includegraphics[width=\linewidth]{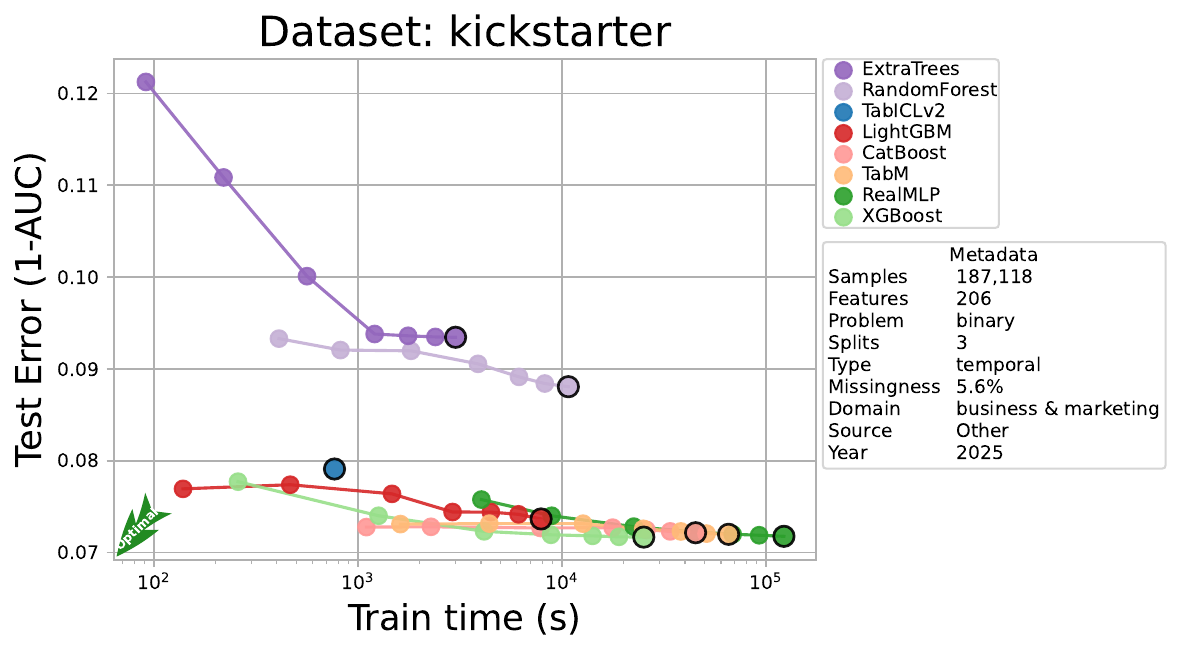}
  \end{minipage}

  \captionof{figure}{\textbf{kickstarter}: per-method test error (left) and HPO Pareto trajectory (right).}
  \label{fig:perdataset_kickstarter}
\end{center}
\space
\begin{center}
  \begin{minipage}[t]{0.48\textwidth}
    \centering
    \vspace{0pt}
    \scriptsize
\begin{tabular}{l@{\quad}lll}
\toprule
 & Default & Tuned & Tuned + Ens. \\
\midrule
Linear & 0.289 $\pm$ 0.003 & 0.288 $\pm$ 0.003 & 0.288 $\pm$ 0.003 \\
RandomForest & 0.254 $\pm$ 0.004 & 0.234 $\pm$ 0.004 & 0.233 $\pm$ 0.004 \\
ExtraTrees & 0.264 $\pm$ 0.004 & 0.233 $\pm$ 0.004 & 0.231 $\pm$ 0.004 \\
CatBoost & 0.221 $\pm$ 0.004 & 0.220 $\pm$ 0.004 & 0.220 $\pm$ 0.004 \\
LightGBM & 0.226 $\pm$ 0.005 & 0.225 $\pm$ 0.004 & 0.224 $\pm$ 0.004 \\
XGBoost & 0.225 $\pm$ 0.005 & 0.223 $\pm$ 0.004 & 0.222 $\pm$ 0.004 \\
RealMLP & 0.233 $\pm$ 0.005 & \underline{0.218 $\pm$ 0.004} & \underline{0.215 $\pm$ 0.004} \\
TabM & 0.223 $\pm$ 0.004 & 0.222 $\pm$ 0.004 & 0.222 $\pm$ 0.004 \\
TabDPT & 0.257 $\pm$ 0.003 & - & - \\
TabPFN-2.6 & 0.226 $\pm$ 0.012 & - & - \\
TabICLv2 & \underline{\textbf{\textcolor{green!50!black}{0.213 $\pm$ 0.004}}} & - & - \\
\bottomrule
\end{tabular}
\space
  \end{minipage}\hfill
  \begin{minipage}[t]{0.48\textwidth}
    \centering
    \vspace{0pt}
    \includegraphics[width=\linewidth]{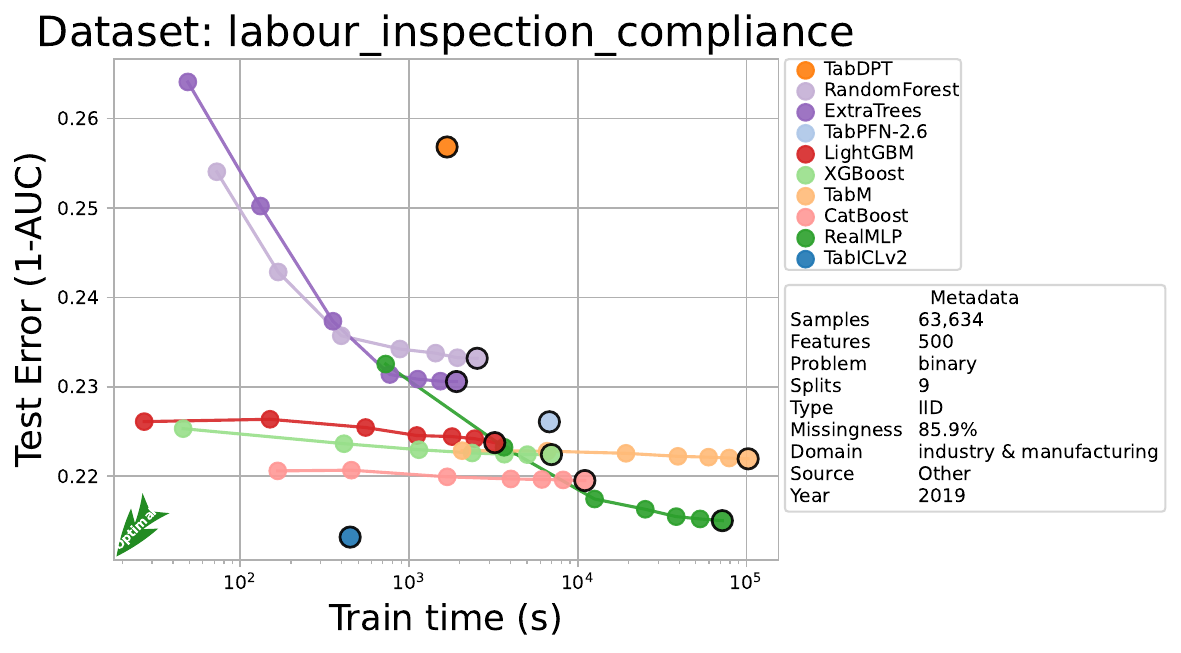}
  \end{minipage}

  \captionof{figure}{\textbf{labour\_inspection\_compliance}: per-method test error (left) and HPO Pareto trajectory (right).}
  \label{fig:perdataset_labour_inspection_compliance}
\end{center}
\space
\begin{center}
  \begin{minipage}[t]{0.48\textwidth}
    \centering
    \vspace{0pt}
    \scriptsize
\begin{tabular}{l@{\quad}lll}
\toprule
 & Default & Tuned & Tuned + Ens. \\
\midrule
Linear & 0.306 & 0.305 & 0.303 \\
RandomForest & 0.308 & 0.317 & 0.313 \\
ExtraTrees & 0.319 & 0.312 & 0.310 \\
CatBoost & 0.287 & 0.286 & 0.285 \\
LightGBM & 0.288 & 0.287 & 0.286 \\
XGBoost & 0.288 & 0.285 & \textcolor{green!50!black}{0.284} \\
RealMLP & 0.287 & 0.287 & 0.285 \\
TabM & 0.290 & 0.291 & 0.289 \\
TabDPT & - & - & - \\
TabPFN-2.6 & - & - & - \\
TabICLv2 & 0.315 & - & - \\
\bottomrule
\end{tabular}
\space
  \end{minipage}\hfill
  \begin{minipage}[t]{0.48\textwidth}
    \centering
    \vspace{0pt}
    \includegraphics[width=\linewidth]{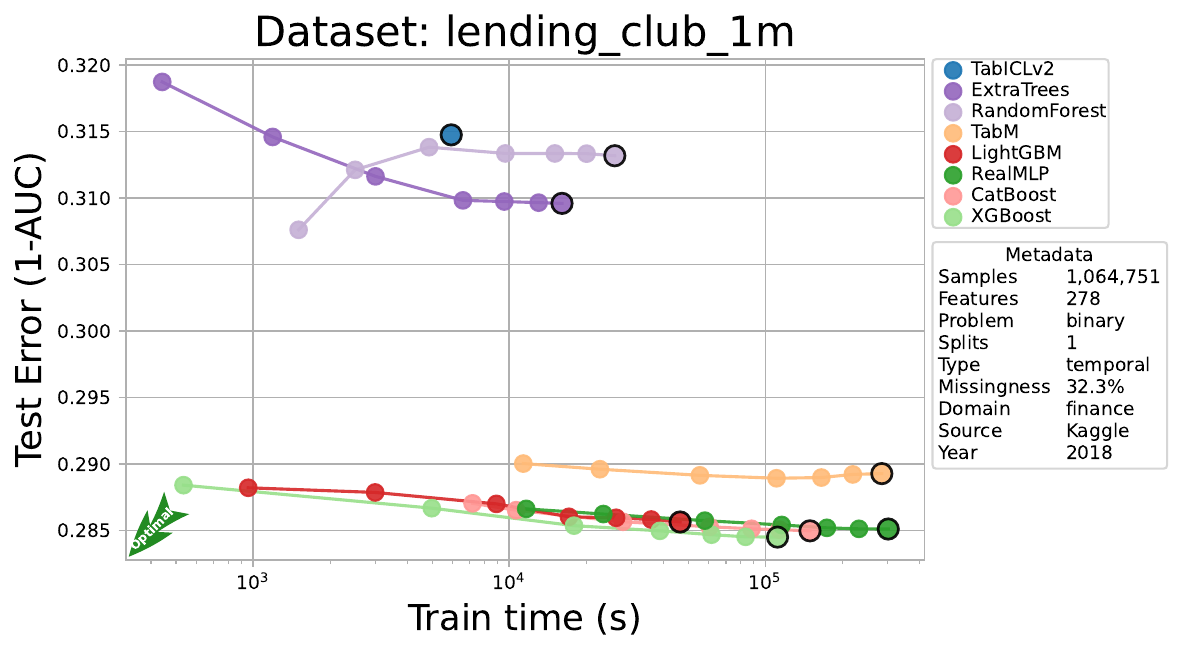}
  \end{minipage}

  \captionof{figure}{\textbf{lending\_club\_1m}: per-method test error (left) and HPO Pareto trajectory (right).}
  \label{fig:perdataset_lending_club_1m}
\end{center}
\space
\begin{center}
  \begin{minipage}[t]{0.48\textwidth}
    \centering
    \vspace{0pt}
    \scriptsize
\begin{tabular}{l@{\quad}lll}
\toprule
 & Default & Tuned & Tuned + Ens. \\
\midrule
Linear & 0.317 $\pm$ 0.048 & \underline{0.306 $\pm$ 0.048} & \underline{0.308 $\pm$ 0.049} \\
RandomForest & 0.332 $\pm$ 0.044 & 0.328 $\pm$ 0.049 & 0.328 $\pm$ 0.048 \\
ExtraTrees & 0.333 $\pm$ 0.047 & \underline{0.313 $\pm$ 0.053} & 0.315 $\pm$ 0.051 \\
CatBoost & \underline{\textbf{0.287 $\pm$ 0.043}} & \underline{0.304 $\pm$ 0.051} & \underline{0.303 $\pm$ 0.048} \\
LightGBM & 0.319 $\pm$ 0.043 & \underline{0.312 $\pm$ 0.048} & 0.318 $\pm$ 0.042 \\
XGBoost & 0.342 $\pm$ 0.044 & \underline{0.307 $\pm$ 0.046} & \underline{0.306 $\pm$ 0.045} \\
RealMLP & 0.323 $\pm$ 0.044 & 0.324 $\pm$ 0.046 & 0.313 $\pm$ 0.046 \\
TabM & 0.304 $\pm$ 0.044 & 0.313 $\pm$ 0.047 & 0.310 $\pm$ 0.044 \\
TabDPT & 0.288 $\pm$ 0.048 & - & - \\
TabPFN-2.6 & \underline{\textbf{\textcolor{green!50!black}{0.284 $\pm$ 0.045}}} & - & - \\
TabICLv2 & \underline{\textbf{0.284 $\pm$ 0.047}} & - & - \\
\bottomrule
\end{tabular}
\space
  \end{minipage}\hfill
  \begin{minipage}[t]{0.48\textwidth}
    \centering
    \vspace{0pt}
    \includegraphics[width=\linewidth]{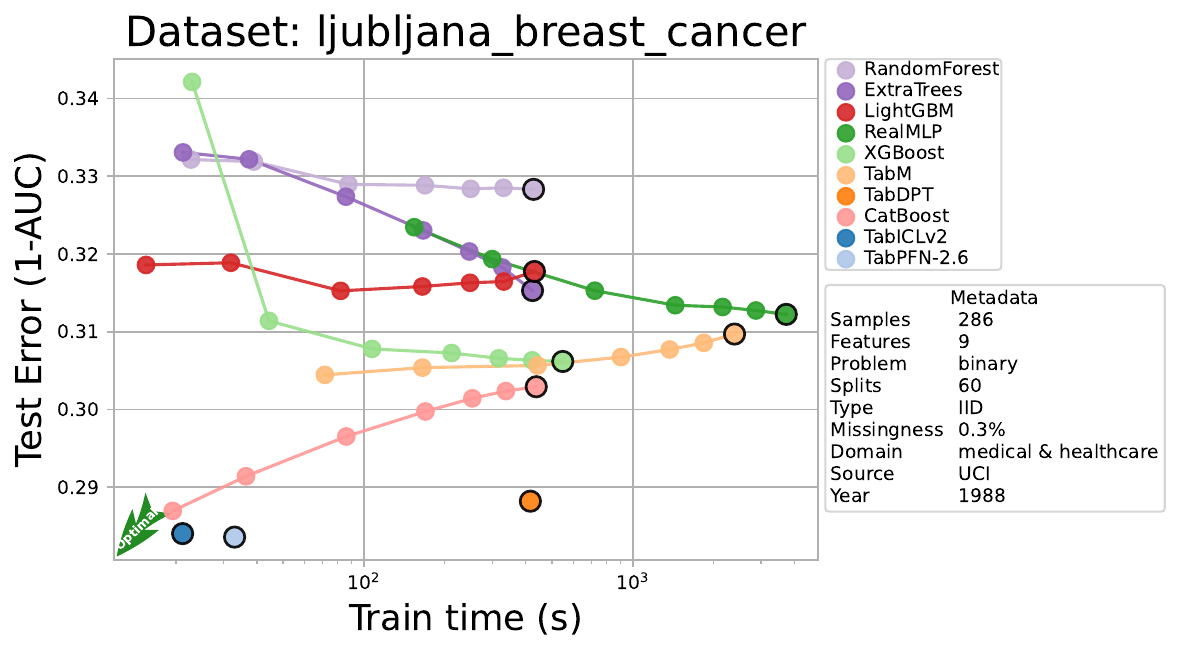}
  \end{minipage}

  \captionof{figure}{\textbf{ljubljana\_breast\_cancer}: per-method test error (left) and HPO Pareto trajectory (right).}
  \label{fig:perdataset_ljubljana_breast_cancer}
\end{center}
\space
\begin{center}
  \begin{minipage}[t]{0.48\textwidth}
    \centering
    \vspace{0pt}
    \scriptsize
\begin{tabular}{l@{\quad}lll}
\toprule
 & Default & Tuned & Tuned + Ens. \\
\midrule
Linear & 1.423 $\pm$ 0.105 & 1.399 $\pm$ 0.078 & 1.396 $\pm$ 0.078 \\
RandomForest & 1.483 $\pm$ 0.108 & 1.404 $\pm$ 0.086 & 1.401 $\pm$ 0.082 \\
ExtraTrees & 1.493 $\pm$ 0.114 & 1.399 $\pm$ 0.081 & 1.401 $\pm$ 0.078 \\
CatBoost & 1.449 $\pm$ 0.063 & 1.387 $\pm$ 0.067 & 1.381 $\pm$ 0.068 \\
LightGBM & 1.421 $\pm$ 0.083 & 1.355 $\pm$ 0.081 & 1.348 $\pm$ 0.081 \\
XGBoost & 1.460 $\pm$ 0.088 & 1.400 $\pm$ 0.083 & 1.395 $\pm$ 0.076 \\
RealMLP & 1.454 $\pm$ 0.087 & 1.343 $\pm$ 0.087 & 1.329 $\pm$ 0.082 \\
TabM & \underline{\textbf{\textcolor{green!50!black}{1.313 $\pm$ 0.080}}} & \underline{1.320 $\pm$ 0.082} & \underline{\textbf{1.315 $\pm$ 0.078}} \\
TabDPT & 1.391 $\pm$ 0.071 & - & - \\
TabPFN-2.6 & 1.336 $\pm$ 0.084 & - & - \\
TabICLv2 & 1.345 $\pm$ 0.094 & - & - \\
\bottomrule
\end{tabular}
\space
  \end{minipage}\hfill
  \begin{minipage}[t]{0.48\textwidth}
    \centering
    \vspace{0pt}
    \includegraphics[width=\linewidth]{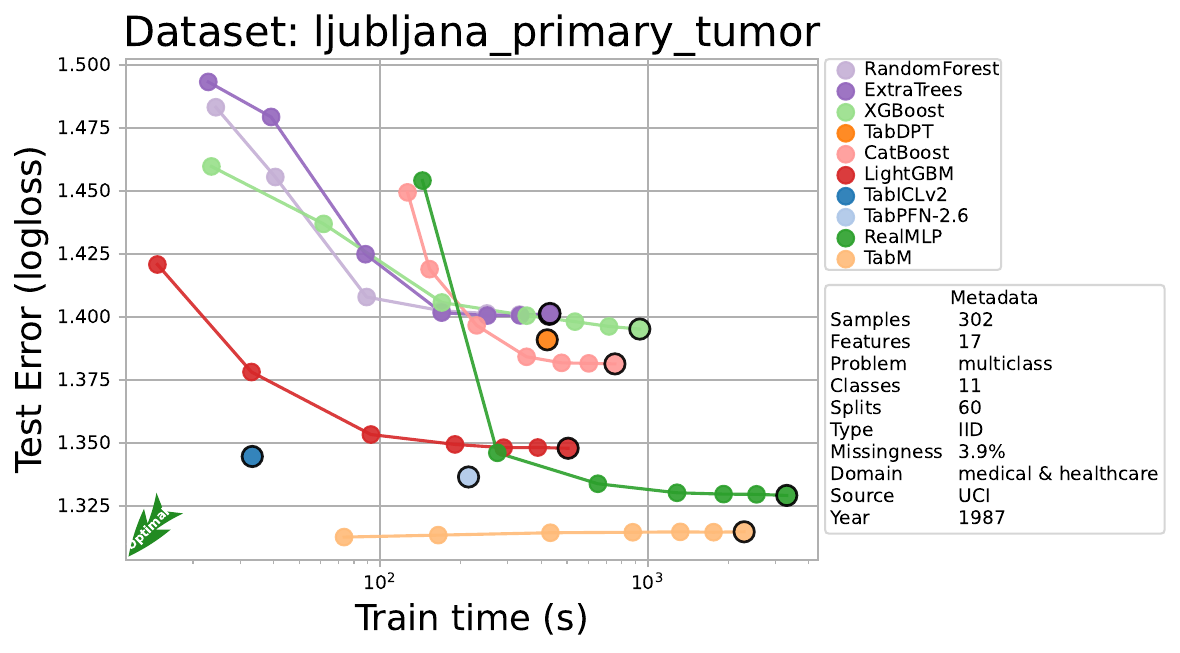}
  \end{minipage}

  \captionof{figure}{\textbf{ljubljana\_primary\_tumor}: per-method test error (left) and HPO Pareto trajectory (right).}
  \label{fig:perdataset_ljubljana_primary_tumor}
\end{center}
\space
\begin{center}
  \begin{minipage}[t]{0.48\textwidth}
    \centering
    \vspace{0pt}
    \scriptsize
\begin{tabular}{l@{\quad}lll}
\toprule
 & Default & Tuned & Tuned + Ens. \\
\midrule
Linear & 0.143 $\pm$ 0.072 & \underline{0.153 $\pm$ 0.078} & \underline{0.140 $\pm$ 0.067} \\
RandomForest & 0.307 $\pm$ 0.031 & 0.239 $\pm$ 0.045 & 0.240 $\pm$ 0.045 \\
ExtraTrees & 0.326 $\pm$ 0.028 & 0.182 $\pm$ 0.030 & 0.182 $\pm$ 0.030 \\
CatBoost & 0.224 $\pm$ 0.086 & 0.219 $\pm$ 0.079 & 0.217 $\pm$ 0.078 \\
LightGBM & 0.191 $\pm$ 0.050 & 0.168 $\pm$ 0.054 & 0.168 $\pm$ 0.054 \\
XGBoost & 0.229 $\pm$ 0.057 & 0.213 $\pm$ 0.057 & 0.213 $\pm$ 0.057 \\
RealMLP & 0.150 $\pm$ 0.078 & \underline{0.155 $\pm$ 0.087} & \underline{0.148 $\pm$ 0.078} \\
TabM & \underline{\textbf{0.133 $\pm$ 0.064}} & \underline{0.143 $\pm$ 0.065} & \underline{0.140 $\pm$ 0.062} \\
TabDPT & 0.144 $\pm$ 0.056 & - & - \\
TabPFN-2.6 & \underline{\textbf{\textcolor{green!50!black}{0.127 $\pm$ 0.057}}} & - & - \\
TabICLv2 & 0.172 $\pm$ 0.080 & - & - \\
\bottomrule
\end{tabular}
\space
  \end{minipage}\hfill
  \begin{minipage}[t]{0.48\textwidth}
    \centering
    \vspace{0pt}
    \includegraphics[width=\linewidth]{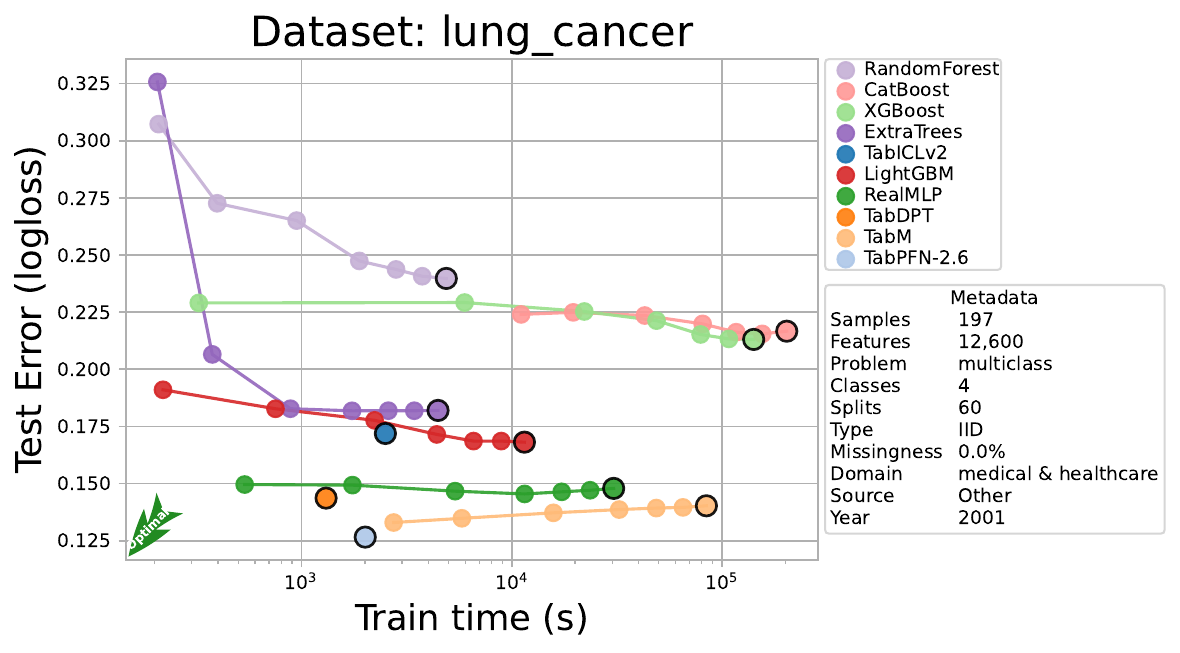}
  \end{minipage}

  \captionof{figure}{\textbf{lung\_cancer}: per-method test error (left) and HPO Pareto trajectory (right).}
  \label{fig:perdataset_lung_cancer}
\end{center}
\space
\begin{center}
  \begin{minipage}[t]{0.48\textwidth}
    \centering
    \vspace{0pt}
    \scriptsize
\begin{tabular}{l@{\quad}lll}
\toprule
 & Default & Tuned & Tuned + Ens. \\
\midrule
Linear & \underline{\textbf{\textcolor{green!50!black}{0.223 $\pm$ 0.052}}} & \underline{\textbf{0.224 $\pm$ 0.052}} & \underline{\textbf{0.224 $\pm$ 0.051}} \\
RandomForest & 0.266 $\pm$ 0.053 & 0.258 $\pm$ 0.057 & 0.256 $\pm$ 0.058 \\
ExtraTrees & 0.265 $\pm$ 0.051 & 0.265 $\pm$ 0.055 & 0.263 $\pm$ 0.055 \\
CatBoost & 0.249 $\pm$ 0.053 & 0.252 $\pm$ 0.050 & 0.251 $\pm$ 0.050 \\
LightGBM & 0.242 $\pm$ 0.059 & 0.241 $\pm$ 0.055 & 0.240 $\pm$ 0.054 \\
XGBoost & 0.250 $\pm$ 0.062 & 0.247 $\pm$ 0.060 & 0.244 $\pm$ 0.059 \\
RealMLP & 0.246 $\pm$ 0.053 & 0.245 $\pm$ 0.054 & 0.246 $\pm$ 0.050 \\
TabM & 0.246 $\pm$ 0.053 & 0.249 $\pm$ 0.053 & 0.249 $\pm$ 0.053 \\
TabDPT & \underline{\textbf{0.223 $\pm$ 0.054}} & - & - \\
TabPFN-2.6 & 0.241 $\pm$ 0.052 & - & - \\
TabICLv2 & 0.273 $\pm$ 0.053 & - & - \\
\bottomrule
\end{tabular}
\space
  \end{minipage}\hfill
  \begin{minipage}[t]{0.48\textwidth}
    \centering
    \vspace{0pt}
    \includegraphics[width=\linewidth]{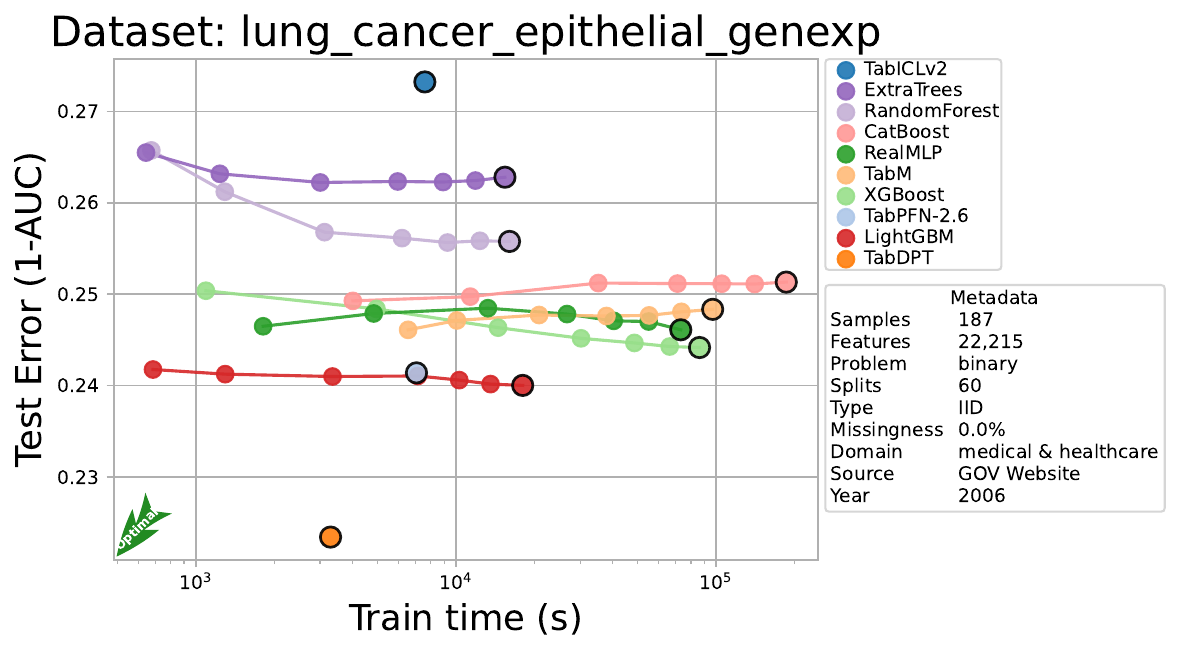}
  \end{minipage}

  \captionof{figure}{\textbf{lung\_cancer\_epithelial\_genexp}: per-method test error (left) and HPO Pareto trajectory (right).}
  \label{fig:perdataset_lung_cancer_epithelial_genexp}
\end{center}
\space
\begin{center}
  \begin{minipage}[t]{0.48\textwidth}
    \centering
    \vspace{0pt}
    \scriptsize
\begin{tabular}{l@{\quad}lll}
\toprule
 & Default & Tuned & Tuned + Ens. \\
\midrule
Linear & 0.163 & 0.163 & 0.163 \\
RandomForest & 0.162 & 0.162 & 0.162 \\
ExtraTrees & 0.162 & 0.162 & 0.162 \\
CatBoost & 0.157 & 0.157 & 0.157 \\
LightGBM & 0.157 & 0.157 & \textcolor{green!50!black}{0.157} \\
XGBoost & 0.157 & 0.157 & 0.157 \\
RealMLP & 0.158 & 0.157 & 0.157 \\
TabM & 0.158 & 0.158 & 0.158 \\
TabDPT & - & - & - \\
TabPFN-2.6 & - & - & - \\
TabICLv2 & 0.167 & - & - \\
\bottomrule
\end{tabular}
\space
  \end{minipage}\hfill
  \begin{minipage}[t]{0.48\textwidth}
    \centering
    \vspace{0pt}
    \includegraphics[width=\linewidth]{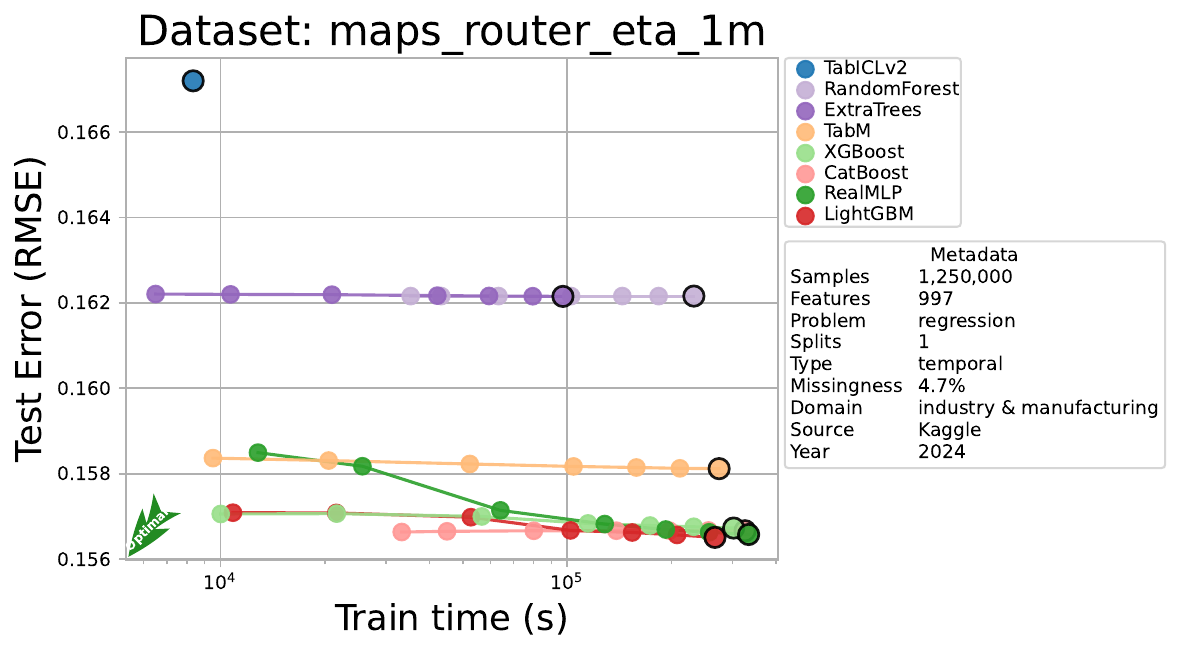}
  \end{minipage}

  \captionof{figure}{\textbf{maps\_router\_eta\_1m}: per-method test error (left) and HPO Pareto trajectory (right).}
  \label{fig:perdataset_maps_router_eta_1m}
\end{center}
\space
\begin{center}
  \begin{minipage}[t]{0.48\textwidth}
    \centering
    \vspace{0pt}
    \scriptsize
\begin{tabular}{l@{\quad}lll}
\toprule
 & Default & Tuned & Tuned + Ens. \\
\midrule
Linear & 0.096 $\pm$ 0.012 & 0.097 $\pm$ 0.013 & 0.096 $\pm$ 0.012 \\
RandomForest & 0.115 $\pm$ 0.013 & 0.116 $\pm$ 0.013 & 0.117 $\pm$ 0.014 \\
ExtraTrees & 0.117 $\pm$ 0.013 & 0.118 $\pm$ 0.013 & 0.115 $\pm$ 0.012 \\
CatBoost & 0.095 $\pm$ 0.011 & 0.097 $\pm$ 0.012 & 0.096 $\pm$ 0.011 \\
LightGBM & 0.100 $\pm$ 0.012 & \underline{0.090 $\pm$ 0.011} & 0.090 $\pm$ 0.011 \\
XGBoost & 0.103 $\pm$ 0.011 & 0.098 $\pm$ 0.010 & 0.097 $\pm$ 0.010 \\
RealMLP & 0.096 $\pm$ 0.012 & 0.093 $\pm$ 0.012 & 0.091 $\pm$ 0.012 \\
TabM & 0.097 $\pm$ 0.011 & \underline{0.091 $\pm$ 0.009} & \underline{0.087 $\pm$ 0.009} \\
TabDPT & 0.096 $\pm$ 0.014 & - & - \\
TabPFN-2.6 & 0.079 $\pm$ 0.010 & - & - \\
TabICLv2 & \underline{\textbf{\textcolor{green!50!black}{0.076 $\pm$ 0.010}}} & - & - \\
\bottomrule
\end{tabular}
\space
  \end{minipage}\hfill
  \begin{minipage}[t]{0.48\textwidth}
    \centering
    \vspace{0pt}
    \includegraphics[width=\linewidth]{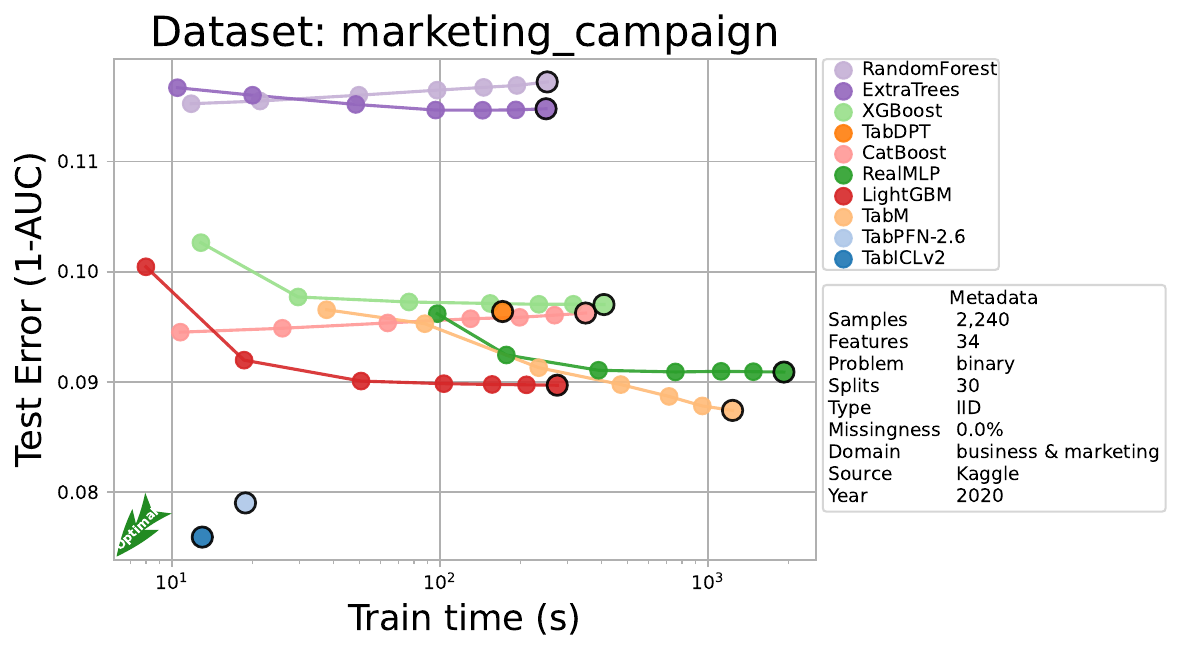}
  \end{minipage}

  \captionof{figure}{\textbf{marketing\_campaign}: per-method test error (left) and HPO Pareto trajectory (right).}
  \label{fig:perdataset_marketing_campaign}
\end{center}
\space
\begin{center}
  \begin{minipage}[t]{0.48\textwidth}
    \centering
    \vspace{0pt}
    \scriptsize
\begin{tabular}{l@{\quad}lll}
\toprule
 & Default & Tuned & Tuned + Ens. \\
\midrule
Linear & 0.795 $\pm$ 0.022 & 0.789 $\pm$ 0.026 & 0.782 $\pm$ 0.023 \\
RandomForest & 0.452 $\pm$ 0.043 & 0.468 $\pm$ 0.038 & 0.436 $\pm$ 0.037 \\
ExtraTrees & 0.457 $\pm$ 0.038 & 0.455 $\pm$ 0.048 & \underline{0.426 $\pm$ 0.034} \\
CatBoost & 0.462 $\pm$ 0.033 & 0.452 $\pm$ 0.033 & 0.449 $\pm$ 0.032 \\
LightGBM & 0.472 $\pm$ 0.037 & 0.458 $\pm$ 0.034 & 0.465 $\pm$ 0.029 \\
XGBoost & 0.456 $\pm$ 0.036 & 0.455 $\pm$ 0.032 & 0.453 $\pm$ 0.033 \\
RealMLP & 0.544 $\pm$ 0.041 & \underline{0.436 $\pm$ 0.040} & \underline{0.424 $\pm$ 0.034} \\
TabM & 0.510 $\pm$ 0.038 & 0.472 $\pm$ 0.043 & 0.466 $\pm$ 0.038 \\
TabDPT & 0.384 $\pm$ 0.041 & - & - \\
TabPFN-2.6 & 0.398 $\pm$ 0.044 & - & - \\
TabICLv2 & \underline{\textbf{\textcolor{green!50!black}{0.377 $\pm$ 0.040}}} & - & - \\
\bottomrule
\end{tabular}
\space
  \end{minipage}\hfill
  \begin{minipage}[t]{0.48\textwidth}
    \centering
    \vspace{0pt}
    \includegraphics[width=\linewidth]{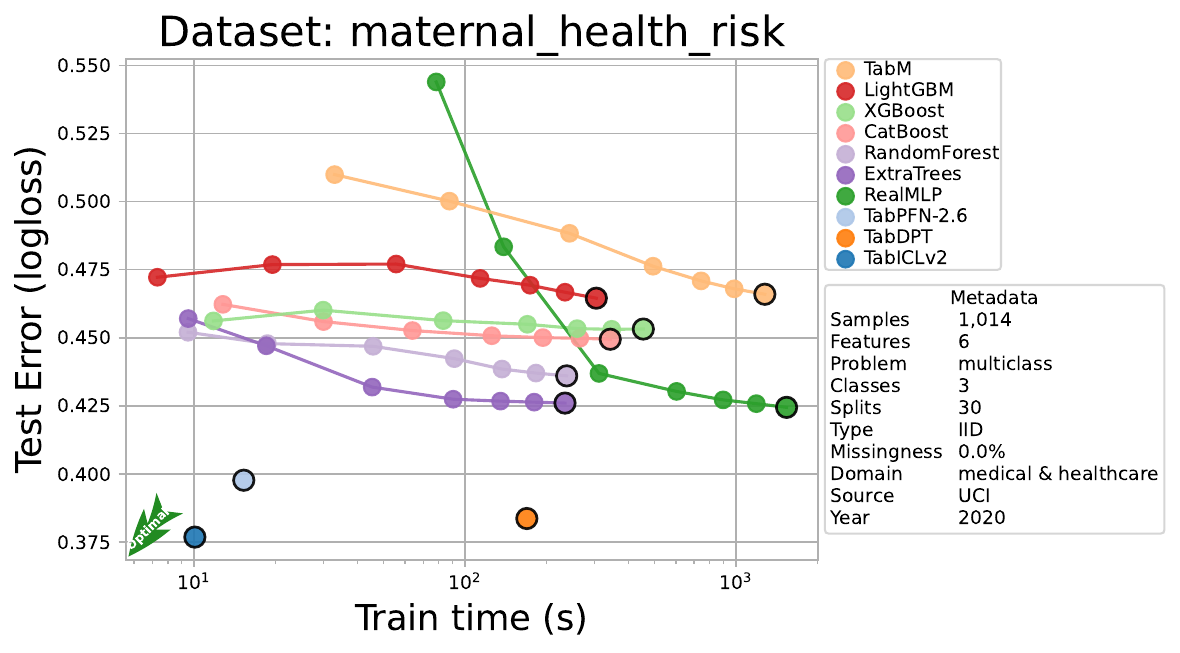}
  \end{minipage}

  \captionof{figure}{\textbf{maternal\_health\_risk}: per-method test error (left) and HPO Pareto trajectory (right).}
  \label{fig:perdataset_maternal_health_risk}
\end{center}
\space
\begin{center}
  \begin{minipage}[t]{0.48\textwidth}
    \centering
    \vspace{0pt}
    \scriptsize
\begin{tabular}{l@{\quad}lll}
\toprule
 & Default & Tuned & Tuned + Ens. \\
\midrule
Linear & 0.648 & 0.648 & 0.648 \\
RandomForest & 0.512 & 0.512 & 0.512 \\
ExtraTrees & 0.518 & 0.516 & 0.516 \\
CatBoost & 0.490 & 0.467 & 0.468 \\
LightGBM & 0.481 & 0.466 & 0.470 \\
XGBoost & 0.472 & 0.472 & 0.469 \\
RealMLP & 0.473 & \textcolor{green!50!black}{0.451} & 0.452 \\
TabM & 0.473 & 0.469 & 0.467 \\
TabDPT & - & - & - \\
TabPFN-2.6 & - & - & - \\
TabICLv2 & 0.464 & - & - \\
\bottomrule
\end{tabular}
\space
  \end{minipage}\hfill
  \begin{minipage}[t]{0.48\textwidth}
    \centering
    \vspace{0pt}
    \includegraphics[width=\linewidth]{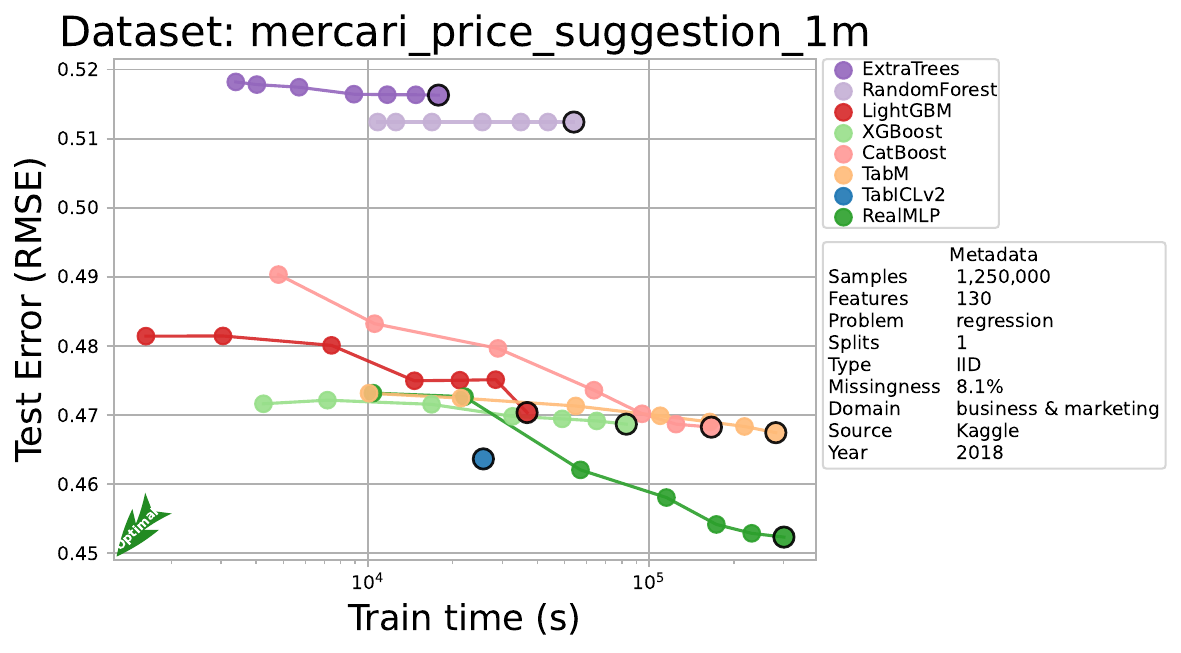}
  \end{minipage}

  \captionof{figure}{\textbf{mercari\_price\_suggestion\_1m}: per-method test error (left) and HPO Pareto trajectory (right).}
  \label{fig:perdataset_mercari_price_suggestion_1m}
\end{center}
\space
\begin{center}
  \begin{minipage}[t]{0.48\textwidth}
    \centering
    \vspace{0pt}
    \scriptsize
\begin{tabular}{l@{\quad}lll}
\toprule
 & Default & Tuned & Tuned + Ens. \\
\midrule
Linear & 8.220 $\pm$ 1.240 & 7.902 $\pm$ 1.162 & \underline{7.871 $\pm$ 1.136} \\
RandomForest & 8.918 $\pm$ 1.497 & 8.541 $\pm$ 1.177 & 8.574 $\pm$ 1.217 \\
ExtraTrees & 8.548 $\pm$ 1.138 & 7.906 $\pm$ 1.147 & 7.903 $\pm$ 1.158 \\
CatBoost & 7.725 $\pm$ 1.215 & \underline{7.743 $\pm$ 1.211} & \underline{7.748 $\pm$ 1.226} \\
LightGBM & 7.855 $\pm$ 1.267 & \underline{7.746 $\pm$ 1.235} & \underline{7.724 $\pm$ 1.216} \\
XGBoost & 8.428 $\pm$ 1.583 & \underline{7.797 $\pm$ 1.243} & 7.875 $\pm$ 1.290 \\
RealMLP & 7.813 $\pm$ 1.214 & \underline{7.748 $\pm$ 1.171} & \underline{7.742 $\pm$ 1.185} \\
TabM & 7.764 $\pm$ 1.187 & \underline{7.765 $\pm$ 1.285} & \underline{7.724 $\pm$ 1.253} \\
TabDPT & 7.889 $\pm$ 1.121 & - & - \\
TabPFN-2.6 & \underline{\textbf{\textcolor{green!50!black}{7.614 $\pm$ 1.234}}} & - & - \\
TabICLv2 & \underline{\textbf{7.671 $\pm$ 1.136}} & - & - \\
\bottomrule
\end{tabular}
\space
  \end{minipage}\hfill
  \begin{minipage}[t]{0.48\textwidth}
    \centering
    \vspace{0pt}
    \includegraphics[width=\linewidth]{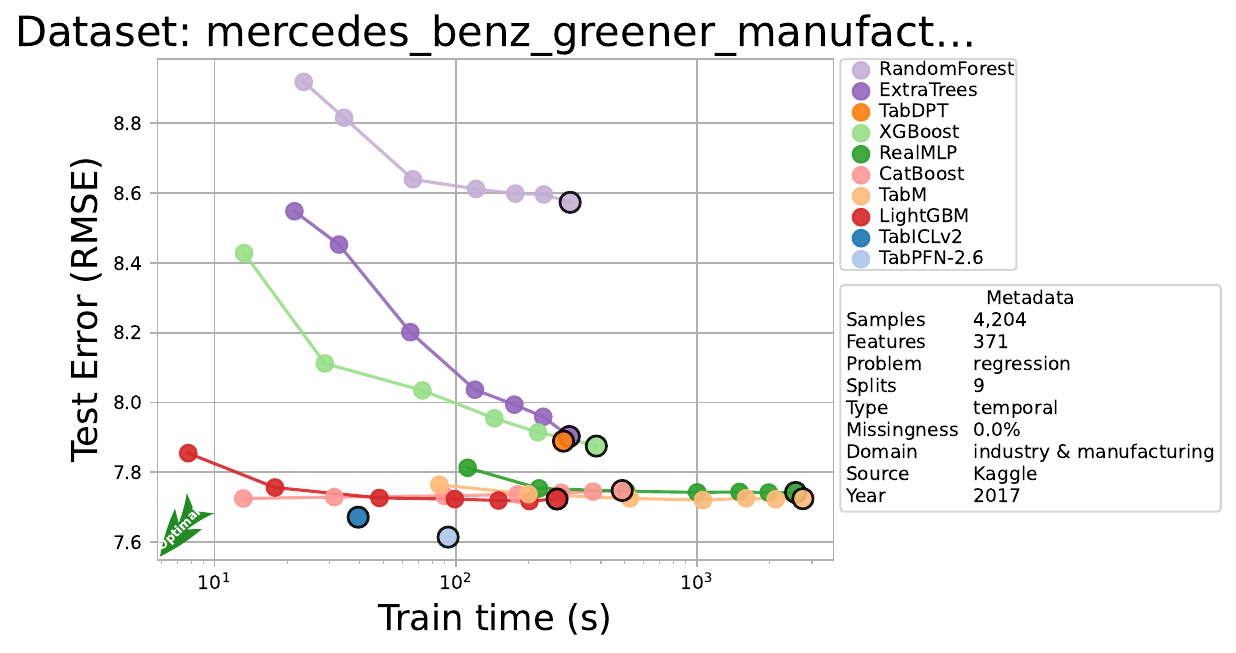}
  \end{minipage}

  \captionof{figure}{\textbf{mercedes\_benz\_greener\_manufacturing}: per-method test error (left) and HPO Pareto trajectory (right).}
  \label{fig:perdataset_mercedes_benz_greener_manufacturing}
\end{center}
\space
\begin{center}
  \begin{minipage}[t]{0.48\textwidth}
    \centering
    \vspace{0pt}
    \scriptsize
\begin{tabular}{l@{\quad}lll}
\toprule
 & Default & Tuned & Tuned + Ens. \\
\midrule
Linear & 0.269 $\pm$ 0.003 & 0.269 $\pm$ 0.003 & 0.264 $\pm$ 0.003 \\
RandomForest & 0.159 $\pm$ 0.004 & 0.157 $\pm$ 0.004 & 0.156 $\pm$ 0.003 \\
ExtraTrees & 0.156 $\pm$ 0.003 & 0.150 $\pm$ 0.003 & 0.150 $\pm$ 0.003 \\
CatBoost & 0.135 $\pm$ 0.002 & \underline{0.135 $\pm$ 0.002} & 0.134 $\pm$ 0.002 \\
LightGBM & 0.141 $\pm$ 0.002 & 0.139 $\pm$ 0.003 & 0.137 $\pm$ 0.003 \\
XGBoost & 0.143 $\pm$ 0.002 & 0.139 $\pm$ 0.003 & 0.138 $\pm$ 0.002 \\
RealMLP & 0.139 $\pm$ 0.003 & \underline{0.134 $\pm$ 0.003} & \underline{0.133 $\pm$ 0.003} \\
TabM & 0.136 $\pm$ 0.002 & \underline{0.135 $\pm$ 0.003} & \underline{0.133 $\pm$ 0.003} \\
TabDPT & 0.131 $\pm$ 0.003 & - & - \\
TabPFN-2.6 & 0.129 $\pm$ 0.003 & - & - \\
TabICLv2 & \underline{\textbf{\textcolor{green!50!black}{0.128 $\pm$ 0.003}}} & - & - \\
\bottomrule
\end{tabular}
\space
  \end{minipage}\hfill
  \begin{minipage}[t]{0.48\textwidth}
    \centering
    \vspace{0pt}
    \includegraphics[width=\linewidth]{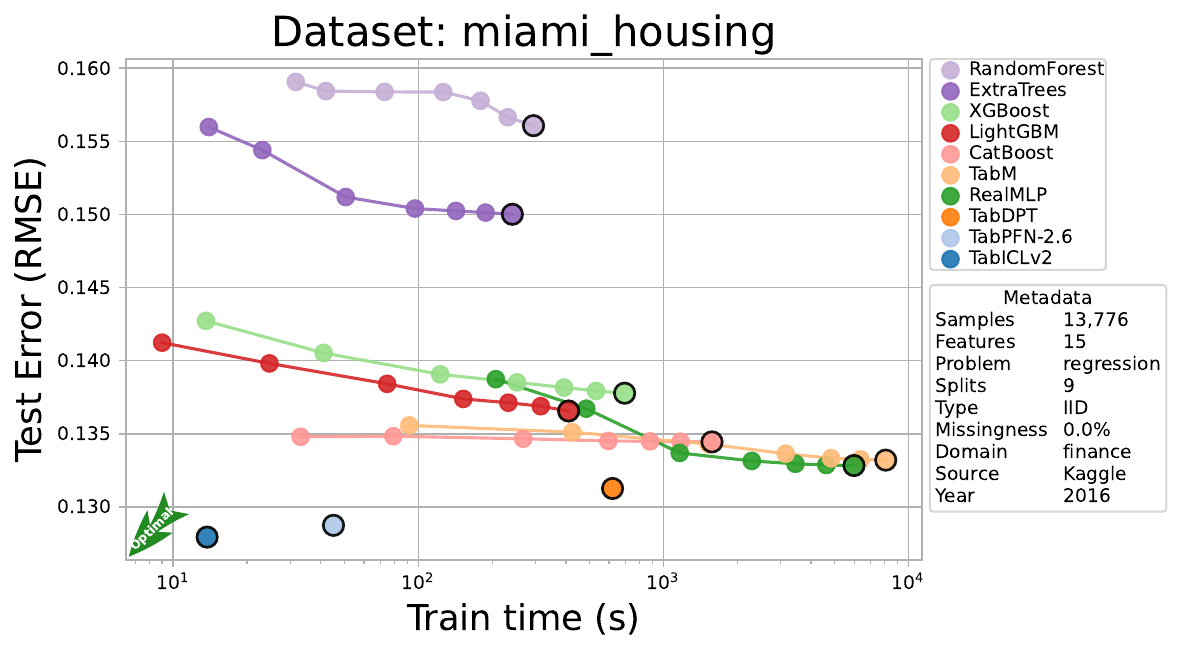}
  \end{minipage}

  \captionof{figure}{\textbf{miami\_housing}: per-method test error (left) and HPO Pareto trajectory (right).}
  \label{fig:perdataset_miami_housing}
\end{center}
\space
\begin{center}
  \begin{minipage}[t]{0.48\textwidth}
    \centering
    \vspace{0pt}
    \scriptsize
\begin{tabular}{l@{\quad}lll}
\toprule
 & Default & Tuned & Tuned + Ens. \\
\midrule
Linear & 0.585 $\pm$ 0.038 & 0.480 $\pm$ 0.025 & 0.479 $\pm$ 0.025 \\
RandomForest & 0.477 $\pm$ 0.017 & 0.478 $\pm$ 0.017 & 0.469 $\pm$ 0.017 \\
ExtraTrees & 0.498 $\pm$ 0.012 & 0.491 $\pm$ 0.020 & 0.471 $\pm$ 0.017 \\
CatBoost & 0.453 $\pm$ 0.018 & 0.455 $\pm$ 0.017 & 0.451 $\pm$ 0.017 \\
LightGBM & 0.500 $\pm$ 0.016 & 0.460 $\pm$ 0.021 & 0.457 $\pm$ 0.018 \\
XGBoost & 0.469 $\pm$ 0.018 & 0.438 $\pm$ 0.019 & 0.438 $\pm$ 0.019 \\
RealMLP & 0.490 $\pm$ 0.036 & \underline{\textbf{0.433 $\pm$ 0.020}} & \underline{\textbf{0.434 $\pm$ 0.018}} \\
TabM & 0.438 $\pm$ 0.018 & 0.439 $\pm$ 0.019 & \textbf{0.436 $\pm$ 0.018} \\
TabDPT & 0.508 $\pm$ 0.014 & - & - \\
TabPFN-2.6 & \underline{\textbf{\textcolor{green!50!black}{0.431 $\pm$ 0.021}}} & - & - \\
TabICLv2 & 0.438 $\pm$ 0.022 & - & - \\
\bottomrule
\end{tabular}
\space
  \end{minipage}\hfill
  \begin{minipage}[t]{0.48\textwidth}
    \centering
    \vspace{0pt}
    \includegraphics[width=\linewidth]{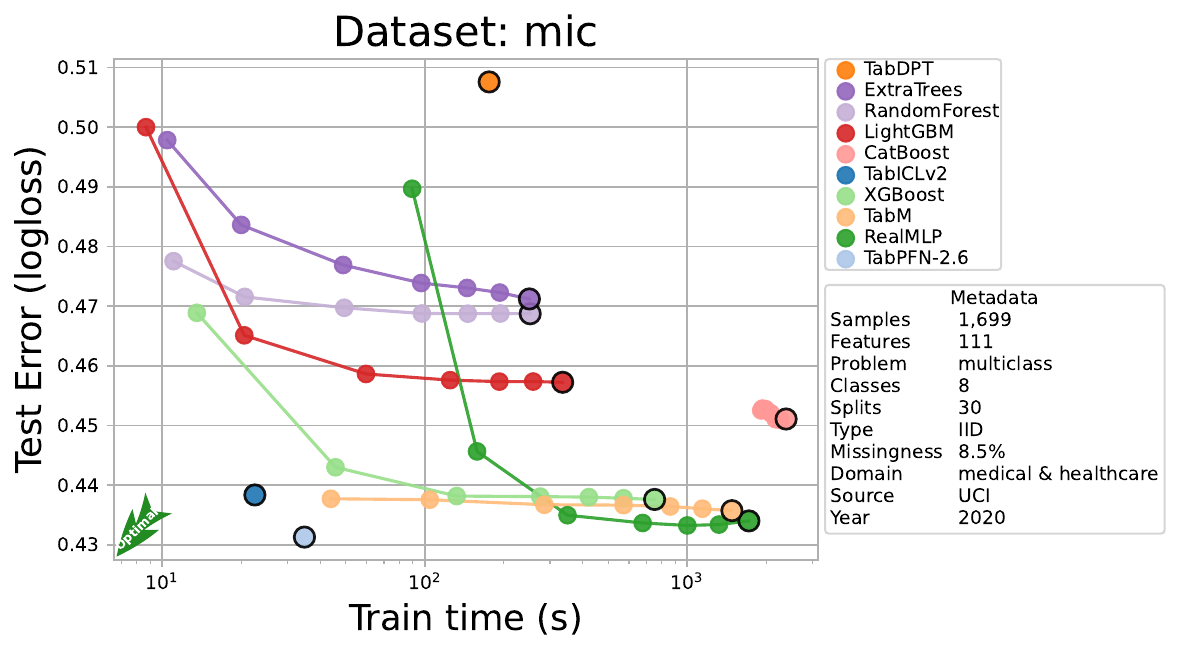}
  \end{minipage}

  \captionof{figure}{\textbf{mic}: per-method test error (left) and HPO Pareto trajectory (right).}
  \label{fig:perdataset_mic}
\end{center}
\space
\begin{center}
  \begin{minipage}[t]{0.48\textwidth}
    \centering
    \vspace{0pt}
    \scriptsize
\begin{tabular}{l@{\quad}lll}
\toprule
 & Default & Tuned & Tuned + Ens. \\
\midrule
Linear & 1.265 $\pm$ 0.347 & 1.234 $\pm$ 0.304 & 1.213 $\pm$ 0.300 \\
RandomForest & 1.240 $\pm$ 0.102 & 1.259 $\pm$ 0.138 & 1.237 $\pm$ 0.137 \\
ExtraTrees & 1.300 $\pm$ 0.081 & 1.226 $\pm$ 0.115 & 1.225 $\pm$ 0.113 \\
CatBoost & 1.372 $\pm$ 0.120 & 1.323 $\pm$ 0.133 & 1.321 $\pm$ 0.128 \\
LightGBM & 1.683 $\pm$ 0.120 & 1.294 $\pm$ 0.134 & 1.295 $\pm$ 0.135 \\
XGBoost & 1.669 $\pm$ 0.114 & 1.504 $\pm$ 0.140 & 1.505 $\pm$ 0.140 \\
RealMLP & 1.343 $\pm$ 0.194 & 1.196 $\pm$ 0.161 & 1.183 $\pm$ 0.152 \\
TabM & 1.153 $\pm$ 0.176 & \underline{1.121 $\pm$ 0.146} & \underline{1.118 $\pm$ 0.141} \\
TabDPT & 1.352 $\pm$ 0.158 & - & - \\
TabPFN-2.6 & 1.275 $\pm$ 0.101 & - & - \\
TabICLv2 & \underline{\textbf{\textcolor{green!50!black}{0.910 $\pm$ 0.190}}} & - & - \\
\bottomrule
\end{tabular}
\space
  \end{minipage}\hfill
  \begin{minipage}[t]{0.48\textwidth}
    \centering
    \vspace{0pt}
    \includegraphics[width=\linewidth]{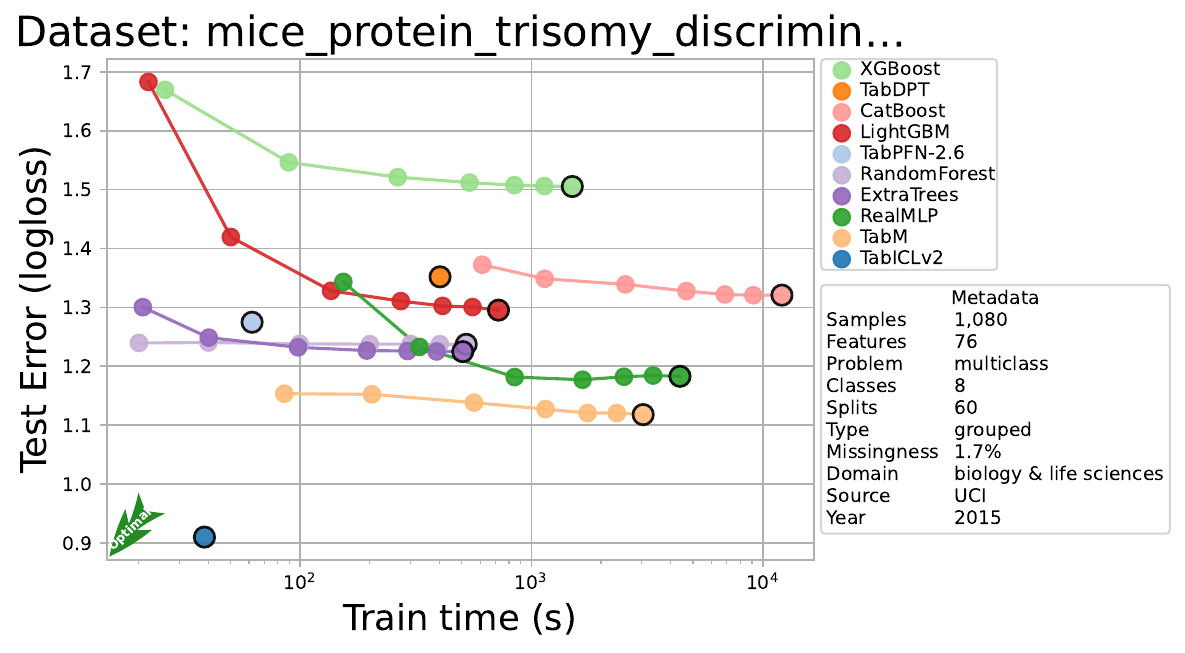}
  \end{minipage}

  \captionof{figure}{\textbf{mice\_protein\_trisomy\_discriminant}: per-method test error (left) and HPO Pareto trajectory (right).}
  \label{fig:perdataset_mice_protein_trisomy_discriminant}
\end{center}
\space
\begin{center}
  \begin{minipage}[t]{0.48\textwidth}
    \centering
    \vspace{0pt}
    \scriptsize
\begin{tabular}{l@{\quad}lll}
\toprule
 & Default & Tuned & Tuned + Ens. \\
\midrule
Linear & \underline{\textbf{0.413 $\pm$ 0.104}} & \underline{\textbf{0.415 $\pm$ 0.108}} & \underline{\textbf{\textcolor{green!50!black}{0.412 $\pm$ 0.108}}} \\
RandomForest & 0.843 $\pm$ 0.039 & 0.702 $\pm$ 0.083 & 0.689 $\pm$ 0.078 \\
ExtraTrees & 0.990 $\pm$ 0.033 & 0.716 $\pm$ 0.058 & 0.716 $\pm$ 0.058 \\
CatBoost & 0.538 $\pm$ 0.087 & 0.483 $\pm$ 0.082 & 0.484 $\pm$ 0.082 \\
LightGBM & 0.698 $\pm$ 0.085 & 0.512 $\pm$ 0.087 & 0.518 $\pm$ 0.082 \\
XGBoost & 0.897 $\pm$ 0.090 & 0.787 $\pm$ 0.091 & 0.787 $\pm$ 0.090 \\
RealMLP & 0.690 $\pm$ 0.070 & 0.613 $\pm$ 0.069 & 0.611 $\pm$ 0.064 \\
TabM & 0.428 $\pm$ 0.076 & 0.429 $\pm$ 0.076 & 0.429 $\pm$ 0.074 \\
TabDPT & 1.184 $\pm$ 0.184 & - & - \\
TabPFN-2.6 & 0.432 $\pm$ 0.087 & - & - \\
TabICLv2 & 0.563 $\pm$ 0.107 & - & - \\
\bottomrule
\end{tabular}
\space
  \end{minipage}\hfill
  \begin{minipage}[t]{0.48\textwidth}
    \centering
    \vspace{0pt}
    \includegraphics[width=\linewidth]{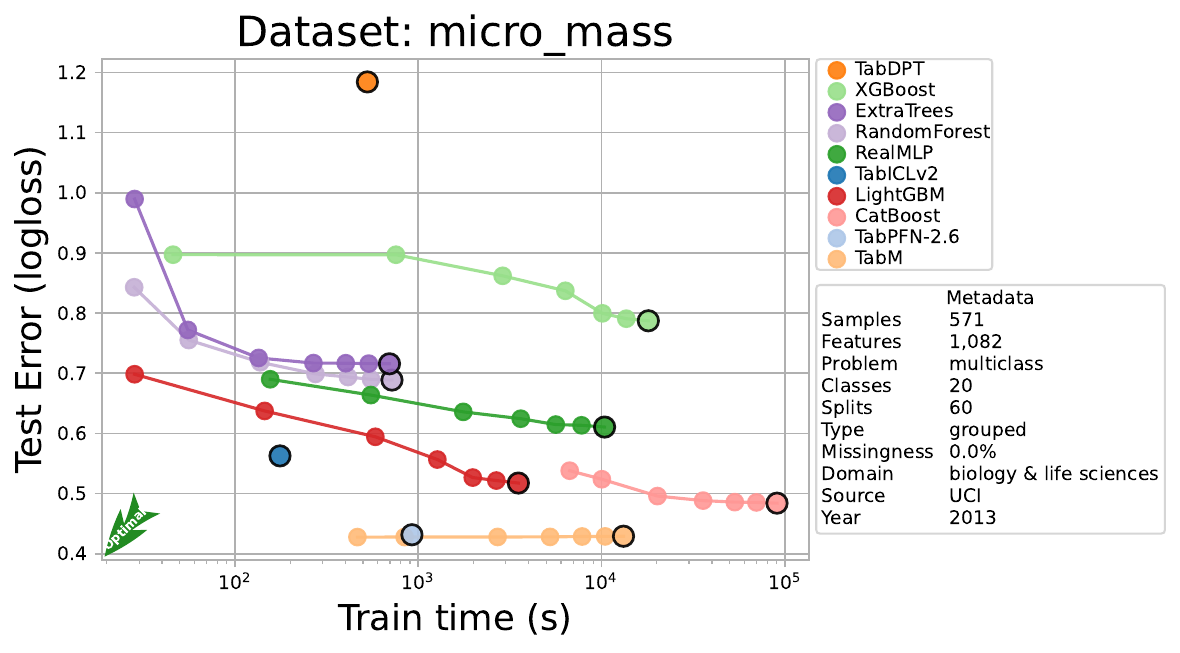}
  \end{minipage}

  \captionof{figure}{\textbf{micro\_mass}: per-method test error (left) and HPO Pareto trajectory (right).}
  \label{fig:perdataset_micro_mass}
\end{center}
\space
\begin{center}
  \begin{minipage}[t]{0.48\textwidth}
    \centering
    \vspace{0pt}
    \scriptsize
\begin{tabular}{l@{\quad}lll}
\toprule
 & Default & Tuned & Tuned + Ens. \\
\midrule
Linear & 0.172 $\pm$ 0.123 & 0.175 $\pm$ 0.130 & 0.179 $\pm$ 0.130 \\
RandomForest & \underline{0.144 $\pm$ 0.137} & 0.168 $\pm$ 0.158 & 0.166 $\pm$ 0.146 \\
ExtraTrees & 0.154 $\pm$ 0.131 & 0.168 $\pm$ 0.142 & 0.167 $\pm$ 0.142 \\
CatBoost & 0.234 $\pm$ 0.180 & 0.218 $\pm$ 0.174 & 0.220 $\pm$ 0.182 \\
LightGBM & 0.201 $\pm$ 0.151 & 0.159 $\pm$ 0.150 & 0.157 $\pm$ 0.149 \\
XGBoost & 0.193 $\pm$ 0.156 & 0.173 $\pm$ 0.157 & 0.168 $\pm$ 0.156 \\
RealMLP & \underline{0.143 $\pm$ 0.128} & 0.150 $\pm$ 0.121 & 0.145 $\pm$ 0.116 \\
TabM & \underline{\textbf{0.127 $\pm$ 0.121}} & \underline{0.119 $\pm$ 0.119} & \underline{\textbf{\textcolor{green!50!black}{0.117 $\pm$ 0.120}}} \\
TabDPT & 0.152 $\pm$ 0.114 & - & - \\
TabPFN-2.6 & 0.160 $\pm$ 0.125 & - & - \\
TabICLv2 & \underline{\textbf{0.126 $\pm$ 0.111}} & - & - \\
\bottomrule
\end{tabular}
\space
  \end{minipage}\hfill
  \begin{minipage}[t]{0.48\textwidth}
    \centering
    \vspace{0pt}
    \includegraphics[width=\linewidth]{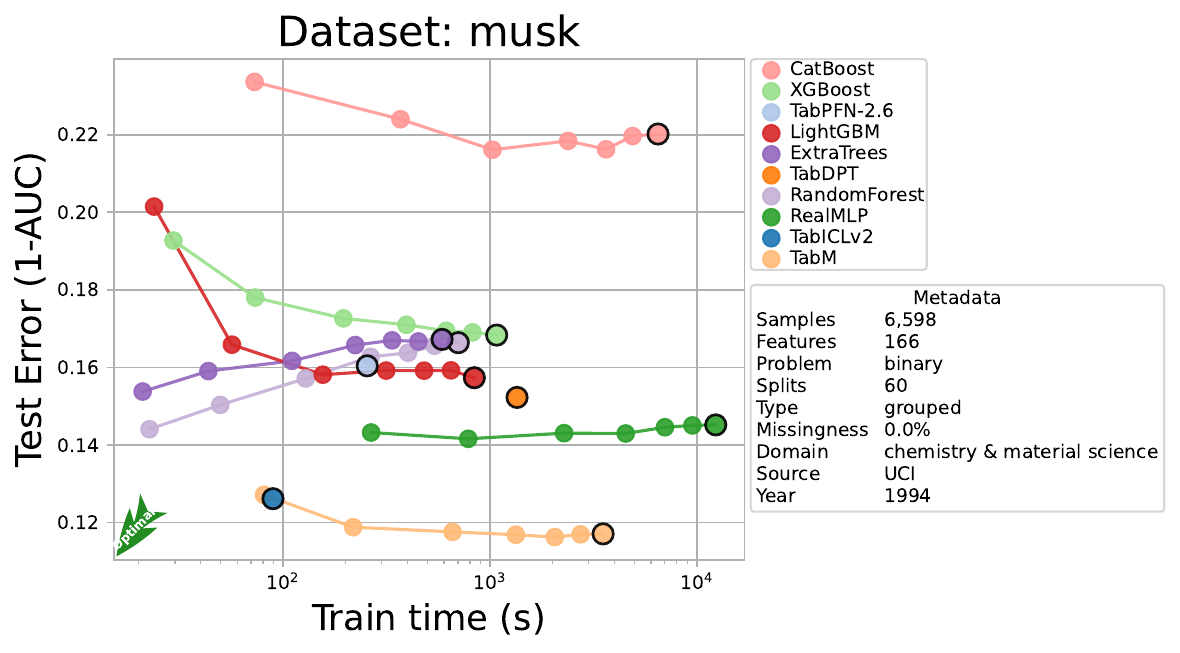}
  \end{minipage}

  \captionof{figure}{\textbf{musk}: per-method test error (left) and HPO Pareto trajectory (right).}
  \label{fig:perdataset_musk}
\end{center}
\space
\begin{center}
  \begin{minipage}[t]{0.48\textwidth}
    \centering
    \vspace{0pt}
    \scriptsize
\begin{tabular}{l@{\quad}lll}
\toprule
 & Default & Tuned & Tuned + Ens. \\
\midrule
Linear & 5.457 $\pm$ 0.258 & 5.447 $\pm$ 0.274 & 5.312 $\pm$ 0.312 \\
RandomForest & 5.249 $\pm$ 0.379 & 4.967 $\pm$ 0.365 & 5.009 $\pm$ 0.380 \\
ExtraTrees & 5.210 $\pm$ 0.396 & 5.093 $\pm$ 0.388 & 5.076 $\pm$ 0.387 \\
CatBoost & 5.245 $\pm$ 0.396 & 4.967 $\pm$ 0.364 & 4.961 $\pm$ 0.365 \\
LightGBM & 5.143 $\pm$ 0.382 & 5.022 $\pm$ 0.374 & 4.919 $\pm$ 0.372 \\
XGBoost & 4.984 $\pm$ 0.400 & 4.986 $\pm$ 0.348 & 4.872 $\pm$ 0.354 \\
RealMLP & 4.914 $\pm$ 0.369 & \underline{4.870 $\pm$ 0.354} & \underline{4.812 $\pm$ 0.357} \\
TabM & 5.085 $\pm$ 0.328 & \underline{4.872 $\pm$ 0.352} & 4.854 $\pm$ 0.345 \\
TabDPT & 5.377 $\pm$ 0.391 & - & - \\
TabPFN-2.6 & 4.605 $\pm$ 0.379 & - & - \\
TabICLv2 & \underline{\textbf{\textcolor{green!50!black}{4.559 $\pm$ 0.388}}} & - & - \\
\bottomrule
\end{tabular}
\space
  \end{minipage}\hfill
  \begin{minipage}[t]{0.48\textwidth}
    \centering
    \vspace{0pt}
    \includegraphics[width=\linewidth]{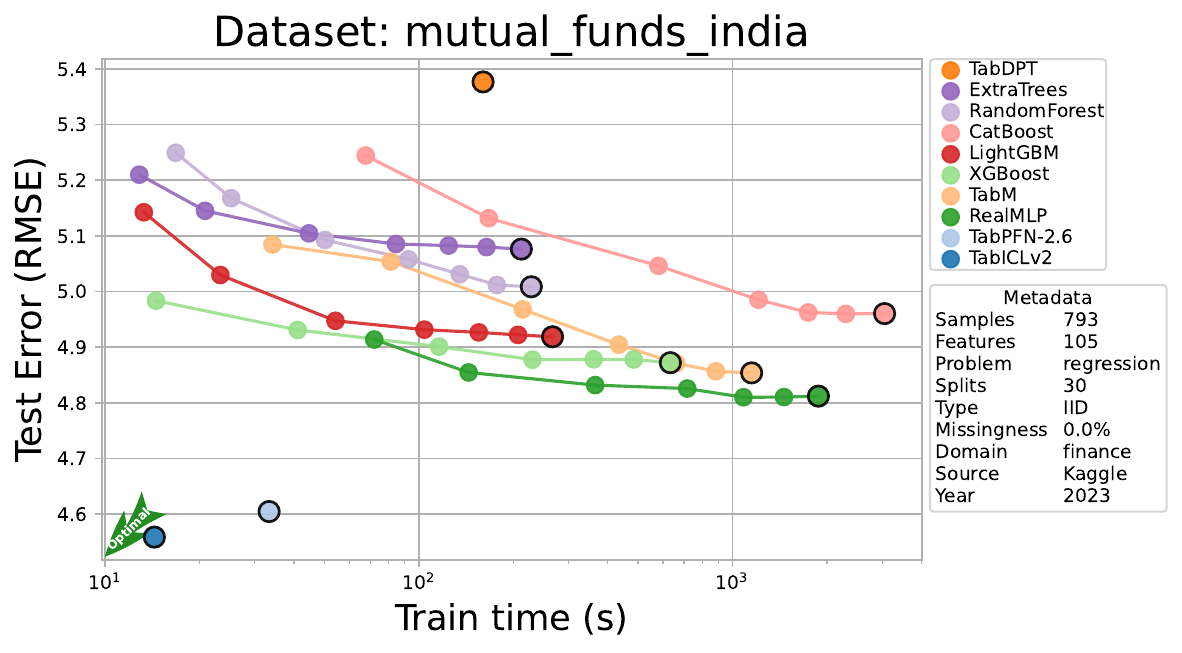}
  \end{minipage}

  \captionof{figure}{\textbf{mutual\_funds\_india}: per-method test error (left) and HPO Pareto trajectory (right).}
  \label{fig:perdataset_mutual_funds_india}
\end{center}
\space
\begin{center}
  \begin{minipage}[t]{0.48\textwidth}
    \centering
    \vspace{0pt}
    \scriptsize
\begin{tabular}{l@{\quad}lll}
\toprule
 & Default & Tuned & Tuned + Ens. \\
\midrule
Linear & 0.019 $\pm$ 0.002 & 0.019 $\pm$ 0.002 & 0.019 $\pm$ 0.002 \\
RandomForest & 0.021 $\pm$ 0.002 & 0.020 $\pm$ 0.002 & 0.019 $\pm$ 0.002 \\
ExtraTrees & 0.021 $\pm$ 0.002 & 0.017 $\pm$ 0.002 & 0.017 $\pm$ 0.002 \\
CatBoost & 0.014 $\pm$ 0.002 & \underline{0.014 $\pm$ 0.002} & \underline{0.014 $\pm$ 0.002} \\
LightGBM & 0.015 $\pm$ 0.002 & \underline{0.014 $\pm$ 0.002} & 0.014 $\pm$ 0.002 \\
XGBoost & 0.015 $\pm$ 0.002 & 0.016 $\pm$ 0.002 & 0.015 $\pm$ 0.002 \\
RealMLP & 0.015 $\pm$ 0.002 & \underline{0.014 $\pm$ 0.002} & \underline{0.014 $\pm$ 0.002} \\
TabM & 0.014 $\pm$ 0.002 & \underline{0.014 $\pm$ 0.002} & 0.014 $\pm$ 0.002 \\
TabDPT & 0.016 $\pm$ 0.002 & - & - \\
TabPFN-2.6 & 0.014 $\pm$ 0.002 & - & - \\
TabICLv2 & \underline{\textbf{\textcolor{green!50!black}{0.012 $\pm$ 0.002}}} & - & - \\
\bottomrule
\end{tabular}
\space
  \end{minipage}\hfill
  \begin{minipage}[t]{0.48\textwidth}
    \centering
    \vspace{0pt}
    \includegraphics[width=\linewidth]{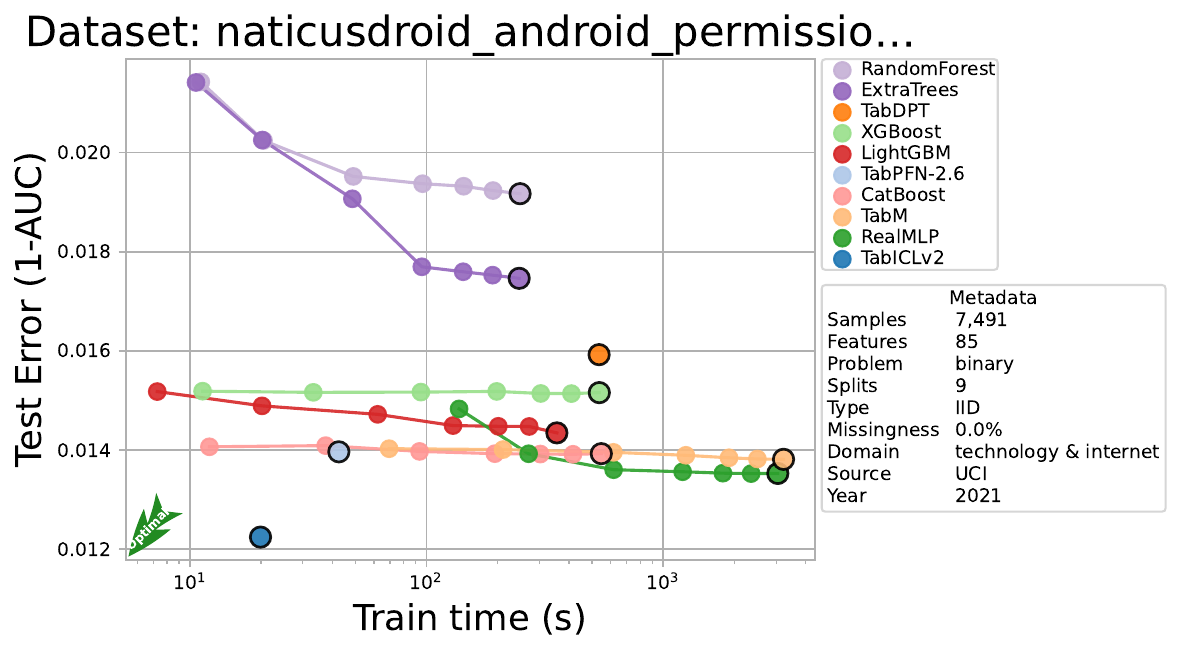}
  \end{minipage}

  \captionof{figure}{\textbf{naticusdroid\_android\_permissions\_dataset}: per-method test error (left) and HPO Pareto trajectory (right).}
  \label{fig:perdataset_naticusdroid_android_permissions_dataset}
\end{center}
\space
\begin{center}
  \begin{minipage}[t]{0.48\textwidth}
    \centering
    \vspace{0pt}
    \scriptsize
\begin{tabular}{l@{\quad}lll}
\toprule
 & Default & Tuned & Tuned + Ens. \\
\midrule
Linear & 4.440 $\pm$ 0.280 & \underline{4.365 $\pm$ 0.280} & \underline{4.364 $\pm$ 0.276} \\
RandomForest & 4.548 $\pm$ 0.253 & 4.406 $\pm$ 0.265 & 4.412 $\pm$ 0.266 \\
ExtraTrees & 4.489 $\pm$ 0.251 & 4.408 $\pm$ 0.261 & 4.406 $\pm$ 0.260 \\
CatBoost & 4.393 $\pm$ 0.276 & 4.397 $\pm$ 0.274 & 4.395 $\pm$ 0.273 \\
LightGBM & 4.454 $\pm$ 0.267 & \underline{4.385 $\pm$ 0.288} & 4.374 $\pm$ 0.280 \\
XGBoost & 4.508 $\pm$ 0.271 & 4.422 $\pm$ 0.269 & 4.423 $\pm$ 0.270 \\
RealMLP & 4.358 $\pm$ 0.281 & \underline{4.375 $\pm$ 0.283} & \underline{4.362 $\pm$ 0.280} \\
TabM & 4.350 $\pm$ 0.276 & \underline{4.371 $\pm$ 0.277} & \underline{4.366 $\pm$ 0.277} \\
TabDPT & \underline{\textbf{\textcolor{green!50!black}{4.280 $\pm$ 0.267}}} & - & - \\
TabPFN-2.6 & 4.341 $\pm$ 0.284 & - & - \\
TabICLv2 & 4.324 $\pm$ 0.279 & - & - \\
\bottomrule
\end{tabular}
\space
  \end{minipage}\hfill
  \begin{minipage}[t]{0.48\textwidth}
    \centering
    \vspace{0pt}
    \includegraphics[width=\linewidth]{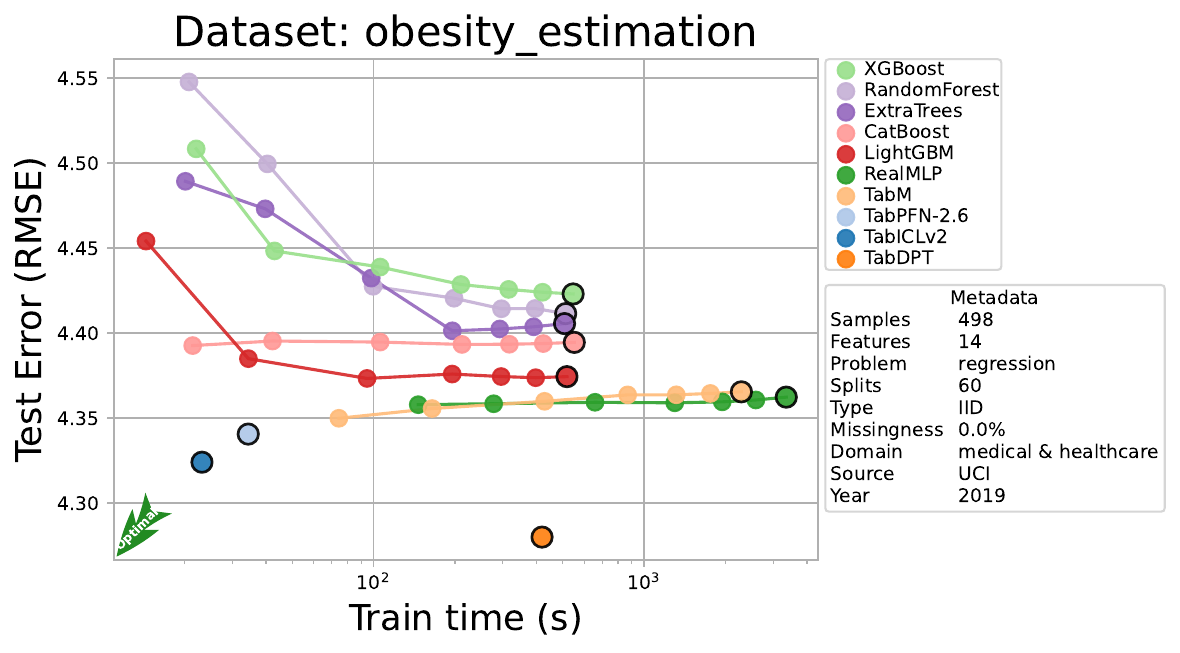}
  \end{minipage}

  \captionof{figure}{\textbf{obesity\_estimation}: per-method test error (left) and HPO Pareto trajectory (right).}
  \label{fig:perdataset_obesity_estimation}
\end{center}
\space
\begin{center}
  \begin{minipage}[t]{0.48\textwidth}
    \centering
    \vspace{0pt}
    \scriptsize
\begin{tabular}{l@{\quad}lll}
\toprule
 & Default & Tuned & Tuned + Ens. \\
\midrule
Linear & 0.088 $\pm$ 0.006 & 0.088 $\pm$ 0.006 & 0.083 $\pm$ 0.006 \\
RandomForest & 0.073 $\pm$ 0.005 & 0.070 $\pm$ 0.004 & 0.069 $\pm$ 0.004 \\
ExtraTrees & 0.080 $\pm$ 0.006 & 0.070 $\pm$ 0.005 & 0.070 $\pm$ 0.005 \\
CatBoost & 0.066 $\pm$ 0.003 & 0.066 $\pm$ 0.004 & 0.065 $\pm$ 0.003 \\
LightGBM & 0.065 $\pm$ 0.004 & \underline{0.064 $\pm$ 0.003} & \underline{0.064 $\pm$ 0.003} \\
XGBoost & 0.064 $\pm$ 0.004 & 0.065 $\pm$ 0.004 & \underline{0.064 $\pm$ 0.003} \\
RealMLP & 0.071 $\pm$ 0.004 & 0.067 $\pm$ 0.003 & 0.066 $\pm$ 0.004 \\
TabM & 0.065 $\pm$ 0.004 & \underline{0.064 $\pm$ 0.004} & \underline{0.064 $\pm$ 0.004} \\
TabDPT & 0.070 $\pm$ 0.005 & - & - \\
TabPFN-2.6 & \underline{\textbf{\textcolor{green!50!black}{0.062 $\pm$ 0.004}}} & - & - \\
TabICLv2 & 0.063 $\pm$ 0.004 & - & - \\
\bottomrule
\end{tabular}
\space
  \end{minipage}\hfill
  \begin{minipage}[t]{0.48\textwidth}
    \centering
    \vspace{0pt}
    \includegraphics[width=\linewidth]{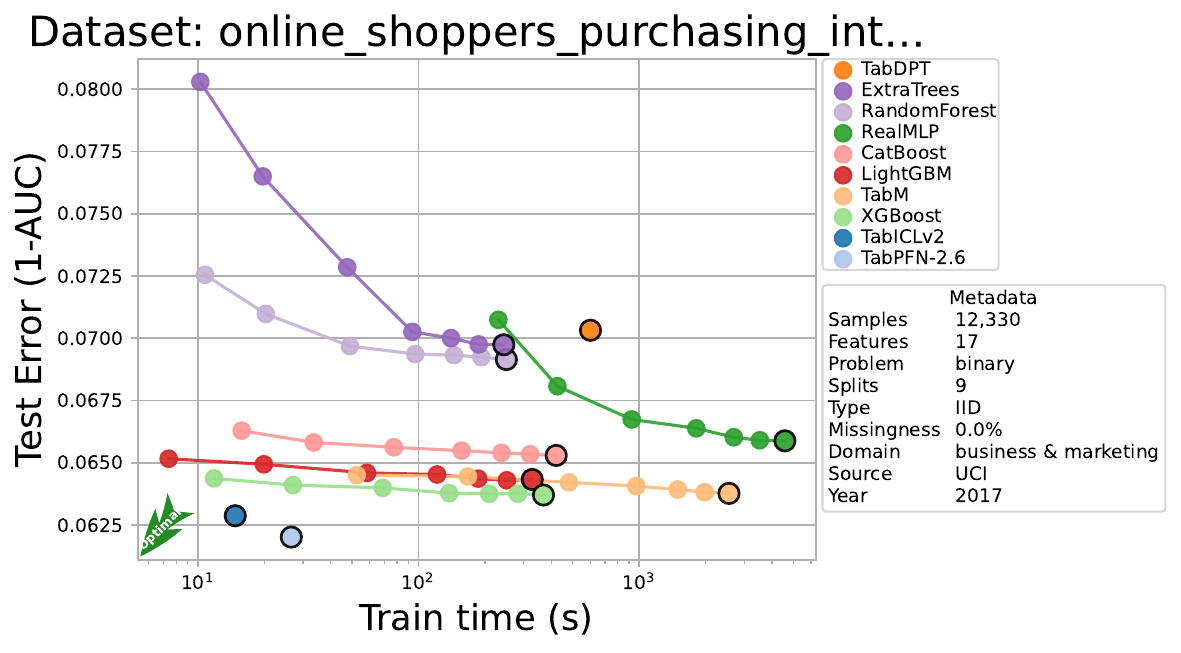}
  \end{minipage}

  \captionof{figure}{\textbf{online\_shoppers\_purchasing\_intention\_dataset}: per-method test error (left) and HPO Pareto trajectory (right).}
  \label{fig:perdataset_online_shoppers_purchasing_intention_dataset}
\end{center}
\space
\begin{center}
  \begin{minipage}[t]{0.48\textwidth}
    \centering
    \vspace{0pt}
    \scriptsize
\begin{tabular}{l@{\quad}lll}
\toprule
 & Default & Tuned & Tuned + Ens. \\
\midrule
Linear & 0.754 $\pm$ 0.005 & 0.640 $\pm$ 0.004 & 0.625 $\pm$ 0.003 \\
RandomForest & 0.587 $\pm$ 0.003 & 0.543 $\pm$ 0.004 & 0.546 $\pm$ 0.004 \\
ExtraTrees & 0.628 $\pm$ 0.003 & 0.545 $\pm$ 0.004 & 0.543 $\pm$ 0.003 \\
CatBoost & 0.475 $\pm$ 0.005 & 0.469 $\pm$ 0.005 & 0.466 $\pm$ 0.005 \\
LightGBM & 0.476 $\pm$ 0.005 & 0.463 $\pm$ 0.005 & 0.460 $\pm$ 0.005 \\
XGBoost & 0.482 $\pm$ 0.004 & 0.461 $\pm$ 0.004 & 0.461 $\pm$ 0.005 \\
RealMLP & 0.474 $\pm$ 0.005 & 0.449 $\pm$ 0.005 & \underline{0.440 $\pm$ 0.004} \\
TabM & 0.445 $\pm$ 0.005 & \underline{0.445 $\pm$ 0.005} & \underline{0.441 $\pm$ 0.004} \\
TabDPT & 0.494 $\pm$ 0.006 & - & - \\
TabPFN-2.6 & 0.463 $\pm$ 0.003 & - & - \\
TabICLv2 & \underline{\textbf{\textcolor{green!50!black}{0.421 $\pm$ 0.005}}} & - & - \\
\bottomrule
\end{tabular}
\space
  \end{minipage}\hfill
  \begin{minipage}[t]{0.48\textwidth}
    \centering
    \vspace{0pt}
    \includegraphics[width=\linewidth]{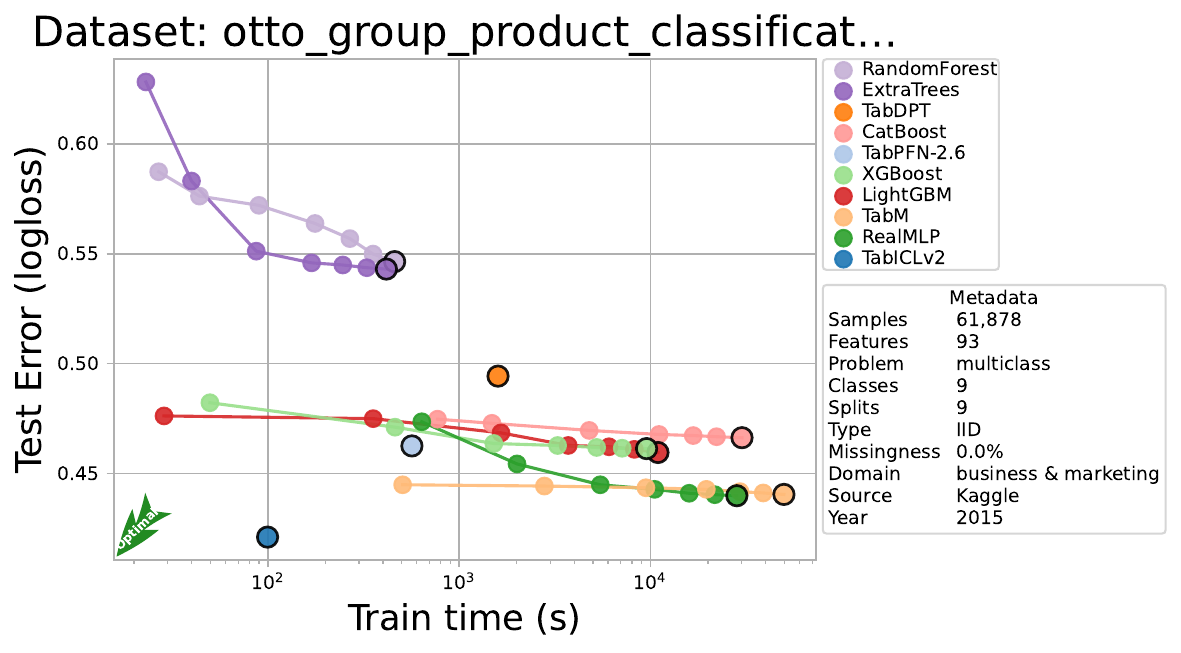}
  \end{minipage}

  \captionof{figure}{\textbf{otto\_group\_product\_classification\_challenge}: per-method test error (left) and HPO Pareto trajectory (right).}
  \label{fig:perdataset_otto_group_product_classification_challenge}
\end{center}
\space
\begin{center}
  \begin{minipage}[t]{0.48\textwidth}
    \centering
    \vspace{0pt}
    \scriptsize
\begin{tabular}{l@{\quad}lll}
\toprule
 & Default & Tuned & Tuned + Ens. \\
\midrule
Linear & \underline{\textbf{\textcolor{green!50!black}{0.373 $\pm$ 0.090}}} & \underline{0.383 $\pm$ 0.090} & \underline{\textbf{0.381 $\pm$ 0.089}} \\
RandomForest & 0.482 $\pm$ 0.061 & 0.507 $\pm$ 0.087 & 0.507 $\pm$ 0.087 \\
ExtraTrees & 0.465 $\pm$ 0.063 & 0.502 $\pm$ 0.090 & 0.506 $\pm$ 0.092 \\
CatBoost & 0.497 $\pm$ 0.079 & 0.491 $\pm$ 0.072 & 0.488 $\pm$ 0.074 \\
LightGBM & 0.512 $\pm$ 0.093 & 0.500 $\pm$ 0.085 & 0.503 $\pm$ 0.084 \\
XGBoost & 0.524 $\pm$ 0.110 & 0.513 $\pm$ 0.099 & 0.514 $\pm$ 0.096 \\
RealMLP & 0.452 $\pm$ 0.077 & 0.465 $\pm$ 0.082 & 0.458 $\pm$ 0.081 \\
TabM & 0.479 $\pm$ 0.068 & 0.478 $\pm$ 0.066 & 0.476 $\pm$ 0.066 \\
TabDPT & \underline{\textbf{0.384 $\pm$ 0.099}} & - & - \\
TabPFN-2.6 & 0.404 $\pm$ 0.091 & - & - \\
TabICLv2 & 0.511 $\pm$ 0.072 & - & - \\
\bottomrule
\end{tabular}
\space
  \end{minipage}\hfill
  \begin{minipage}[t]{0.48\textwidth}
    \centering
    \vspace{0pt}
    \includegraphics[width=\linewidth]{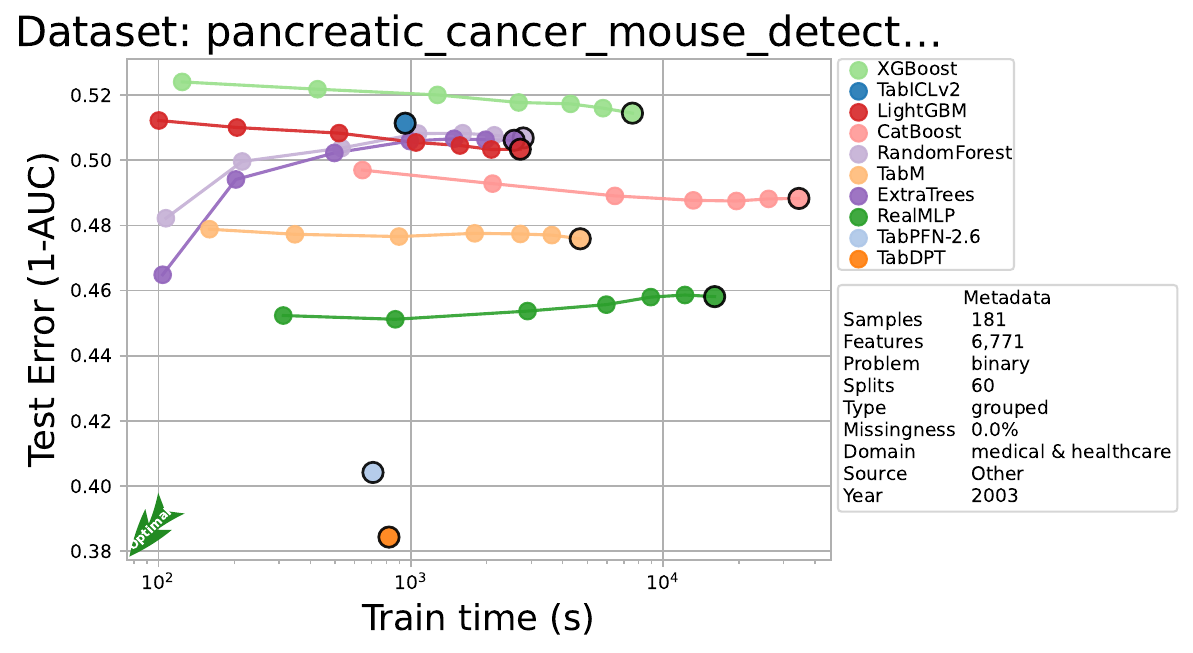}
  \end{minipage}

  \captionof{figure}{\textbf{pancreatic\_cancer\_mouse\_detection}: per-method test error (left) and HPO Pareto trajectory (right).}
  \label{fig:perdataset_pancreatic_cancer_mouse_detection}
\end{center}
\space
\begin{center}
  \begin{minipage}[t]{0.48\textwidth}
    \centering
    \vspace{0pt}
    \scriptsize
\begin{tabular}{l@{\quad}lll}
\toprule
 & Default & Tuned & Tuned + Ens. \\
\midrule
Linear & 0.200 $\pm$ 0.116 & 0.221 $\pm$ 0.131 & 0.227 $\pm$ 0.132 \\
RandomForest & 0.159 $\pm$ 0.113 & 0.179 $\pm$ 0.126 & 0.181 $\pm$ 0.125 \\
ExtraTrees & \underline{0.144 $\pm$ 0.114} & 0.154 $\pm$ 0.115 & 0.155 $\pm$ 0.117 \\
CatBoost & 0.177 $\pm$ 0.125 & 0.166 $\pm$ 0.120 & 0.167 $\pm$ 0.119 \\
LightGBM & 0.213 $\pm$ 0.156 & 0.166 $\pm$ 0.128 & 0.170 $\pm$ 0.131 \\
XGBoost & 0.213 $\pm$ 0.135 & 0.209 $\pm$ 0.135 & 0.207 $\pm$ 0.135 \\
RealMLP & 0.211 $\pm$ 0.141 & \underline{\textbf{0.126 $\pm$ 0.104}} & \underline{\textbf{\textcolor{green!50!black}{0.122 $\pm$ 0.100}}} \\
TabM & \underline{\textbf{0.132 $\pm$ 0.089}} & 0.153 $\pm$ 0.096 & 0.151 $\pm$ 0.092 \\
TabDPT & 0.232 $\pm$ 0.156 & - & - \\
TabPFN-2.6 & 0.182 $\pm$ 0.126 & - & - \\
TabICLv2 & 0.169 $\pm$ 0.138 & - & - \\
\bottomrule
\end{tabular}
\space
  \end{minipage}\hfill
  \begin{minipage}[t]{0.48\textwidth}
    \centering
    \vspace{0pt}
    \includegraphics[width=\linewidth]{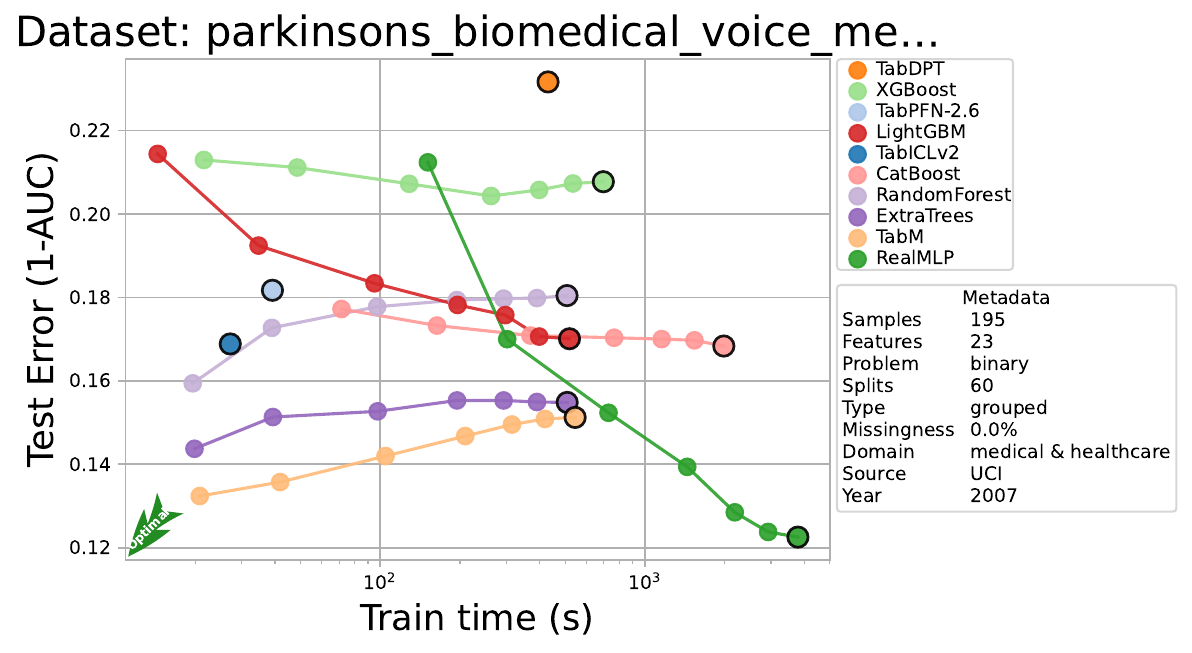}
  \end{minipage}

  \captionof{figure}{\textbf{parkinsons\_biomedical\_voice\_measurements}: per-method test error (left) and HPO Pareto trajectory (right).}
  \label{fig:perdataset_parkinsons_biomedical_voice_measurements}
\end{center}
\space
\begin{center}
  \begin{minipage}[t]{0.48\textwidth}
    \centering
    \vspace{0pt}
    \scriptsize
\begin{tabular}{l@{\quad}lll}
\toprule
 & Default & Tuned & Tuned + Ens. \\
\midrule
Linear & 5.195 $\pm$ 0.028 & 5.185 $\pm$ 0.027 & 5.143 $\pm$ 0.027 \\
RandomForest & 3.589 $\pm$ 0.027 & 3.481 $\pm$ 0.027 & 3.476 $\pm$ 0.026 \\
ExtraTrees & 3.577 $\pm$ 0.023 & 3.469 $\pm$ 0.024 & 3.470 $\pm$ 0.024 \\
CatBoost & 3.525 $\pm$ 0.019 & 3.434 $\pm$ 0.026 & 3.405 $\pm$ 0.020 \\
LightGBM & 3.485 $\pm$ 0.022 & 3.448 $\pm$ 0.022 & 3.453 $\pm$ 0.022 \\
XGBoost & 3.520 $\pm$ 0.022 & 3.408 $\pm$ 0.021 & 3.409 $\pm$ 0.021 \\
RealMLP & 3.449 $\pm$ 0.038 & \underline{3.185 $\pm$ 0.020} & \underline{3.106 $\pm$ 0.024} \\
TabM & 3.419 $\pm$ 0.020 & 3.290 $\pm$ 0.028 & 3.247 $\pm$ 0.023 \\
TabDPT & \underline{\textbf{\textcolor{green!50!black}{2.898 $\pm$ 0.022}}} & - & - \\
TabPFN-2.6 & 3.011 $\pm$ 0.023 & - & - \\
TabICLv2 & 2.944 $\pm$ 0.023 & - & - \\
\bottomrule
\end{tabular}
\space
  \end{minipage}\hfill
  \begin{minipage}[t]{0.48\textwidth}
    \centering
    \vspace{0pt}
    \includegraphics[width=\linewidth]{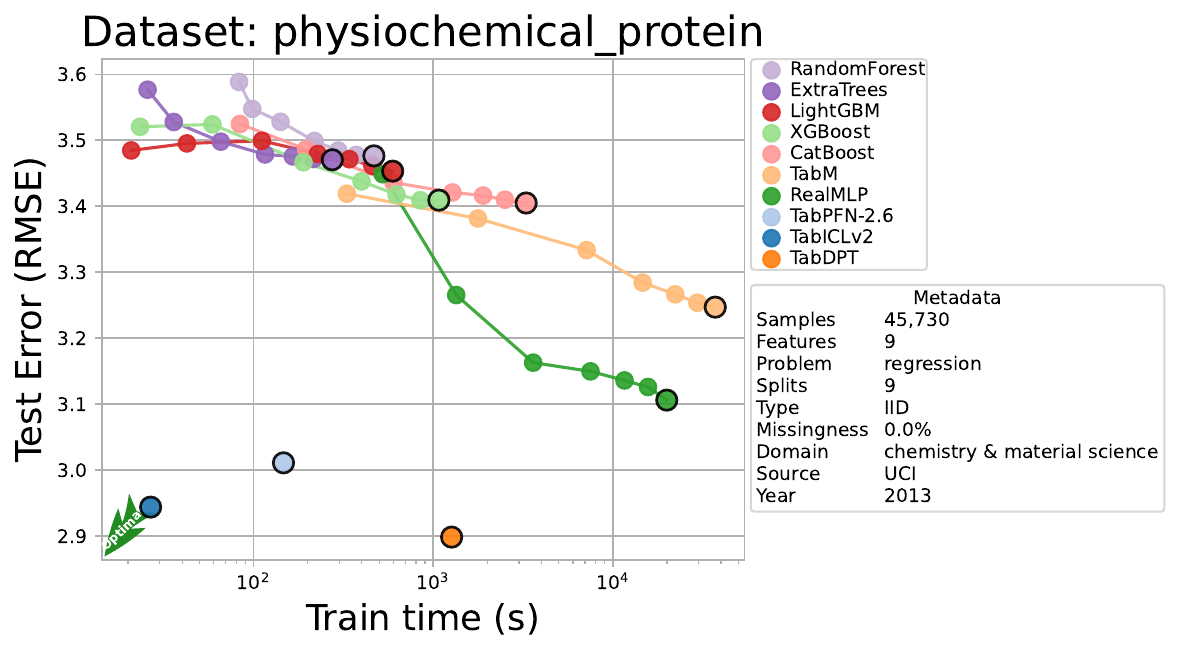}
  \end{minipage}

  \captionof{figure}{\textbf{physiochemical\_protein}: per-method test error (left) and HPO Pareto trajectory (right).}
  \label{fig:perdataset_physiochemical_protein}
\end{center}
\space
\begin{center}
  \begin{minipage}[t]{0.48\textwidth}
    \centering
    \vspace{0pt}
    \scriptsize
\begin{tabular}{l@{\quad}lll}
\toprule
 & Default & Tuned & Tuned + Ens. \\
\midrule
Linear & 0.134 $\pm$ 0.014 & 0.112 $\pm$ 0.013 & 0.105 $\pm$ 0.010 \\
RandomForest & 0.074 $\pm$ 0.009 & 0.069 $\pm$ 0.013 & 0.065 $\pm$ 0.011 \\
ExtraTrees & 0.121 $\pm$ 0.012 & 0.107 $\pm$ 0.010 & 0.107 $\pm$ 0.008 \\
CatBoost & 0.041 $\pm$ 0.008 & 0.042 $\pm$ 0.008 & 0.041 $\pm$ 0.008 \\
LightGBM & 0.048 $\pm$ 0.010 & 0.045 $\pm$ 0.010 & 0.045 $\pm$ 0.011 \\
XGBoost & 0.043 $\pm$ 0.010 & 0.043 $\pm$ 0.009 & 0.044 $\pm$ 0.009 \\
RealMLP & 0.043 $\pm$ 0.010 & 0.037 $\pm$ 0.008 & 0.037 $\pm$ 0.009 \\
TabM & 0.052 $\pm$ 0.009 & \underline{0.031 $\pm$ 0.007} & \underline{0.033 $\pm$ 0.008} \\
TabDPT & 0.034 $\pm$ 0.008 & - & - \\
TabPFN-2.6 & \underline{\textbf{\textcolor{green!50!black}{0.014 $\pm$ 0.006}}} & - & - \\
TabICLv2 & 0.018 $\pm$ 0.007 & - & - \\
\bottomrule
\end{tabular}
\space
  \end{minipage}\hfill
  \begin{minipage}[t]{0.48\textwidth}
    \centering
    \vspace{0pt}
    \includegraphics[width=\linewidth]{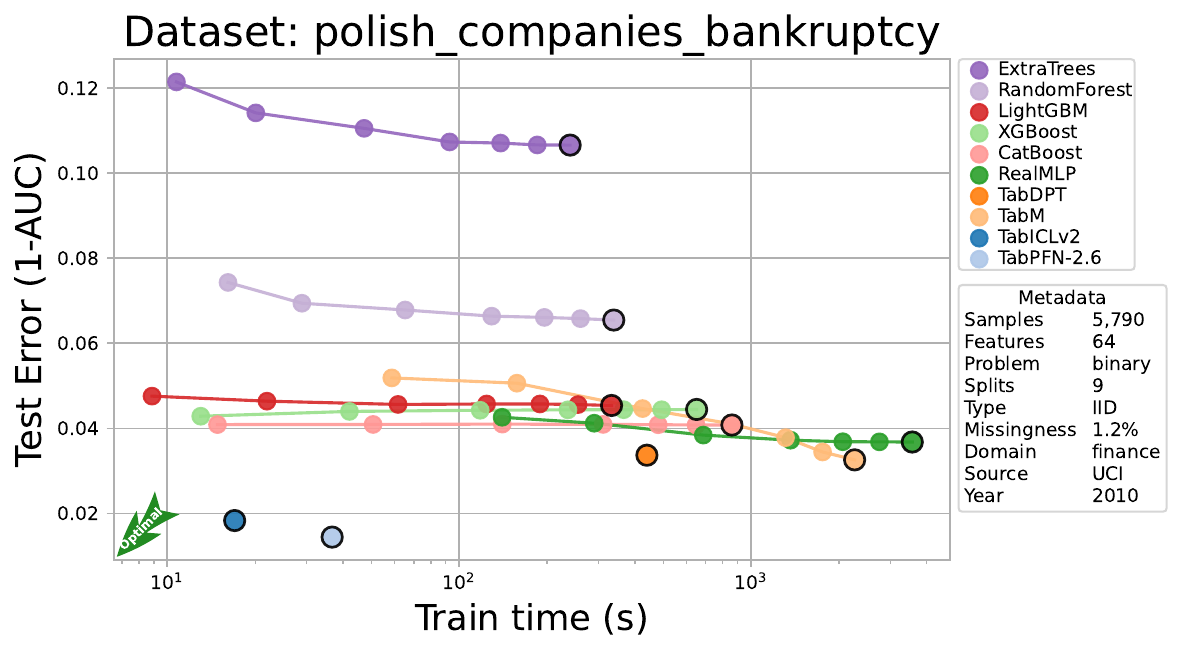}
  \end{minipage}

  \captionof{figure}{\textbf{polish\_companies\_bankruptcy}: per-method test error (left) and HPO Pareto trajectory (right).}
  \label{fig:perdataset_polish_companies_bankruptcy}
\end{center}
\space
\begin{center}
  \begin{minipage}[t]{0.48\textwidth}
    \centering
    \vspace{0pt}
    \scriptsize
\begin{tabular}{l@{\quad}lll}
\toprule
 & Default & Tuned & Tuned + Ens. \\
\midrule
Linear & \underline{\textbf{0.369 $\pm$ 0.001}} & \underline{\textbf{0.369 $\pm$ 0.001}} & \underline{\textbf{0.370 $\pm$ 0.001}} \\
RandomForest & \underline{\textbf{0.383 $\pm$ 0.002}} & \underline{\textbf{0.379 $\pm$ 0.004}} & \underline{\textbf{0.373 $\pm$ 0.004}} \\
ExtraTrees & \underline{\textbf{0.379 $\pm$ 0.002}} & \underline{\textbf{0.380 $\pm$ 0.003}} & \underline{\textbf{0.374 $\pm$ 0.004}} \\
CatBoost & \underline{\textbf{0.359 $\pm$ 0.001}} & \underline{\textbf{0.357 $\pm$ 0.000}} & \underline{\textbf{0.357 $\pm$ 0.000}} \\
LightGBM & \underline{\textbf{0.369 $\pm$ 0.001}} & \underline{\textbf{0.360 $\pm$ 0.001}} & \underline{\textbf{0.360 $\pm$ 0.000}} \\
XGBoost & \underline{\textbf{0.359 $\pm$ 0.001}} & \underline{\textbf{0.360 $\pm$ 0.002}} & \underline{\textbf{0.359 $\pm$ 0.000}} \\
RealMLP & \underline{\textbf{0.361 $\pm$ 0.001}} & \underline{\textbf{0.358 $\pm$ 0.001}} & \underline{\textbf{\textcolor{green!50!black}{0.357 $\pm$ 0.001}}} \\
TabM & \underline{\textbf{0.359 $\pm$ 0.001}} & \underline{\textbf{0.358 $\pm$ 0.001}} & \underline{\textbf{0.357 $\pm$ 0.001}} \\
TabDPT & - & - & - \\
TabPFN-2.6 & - & - & - \\
TabICLv2 & \underline{\textbf{0.366 $\pm$ 0.001}} & - & - \\
\bottomrule
\end{tabular}
\space
  \end{minipage}\hfill
  \begin{minipage}[t]{0.48\textwidth}
    \centering
    \vspace{0pt}
    \includegraphics[width=\linewidth]{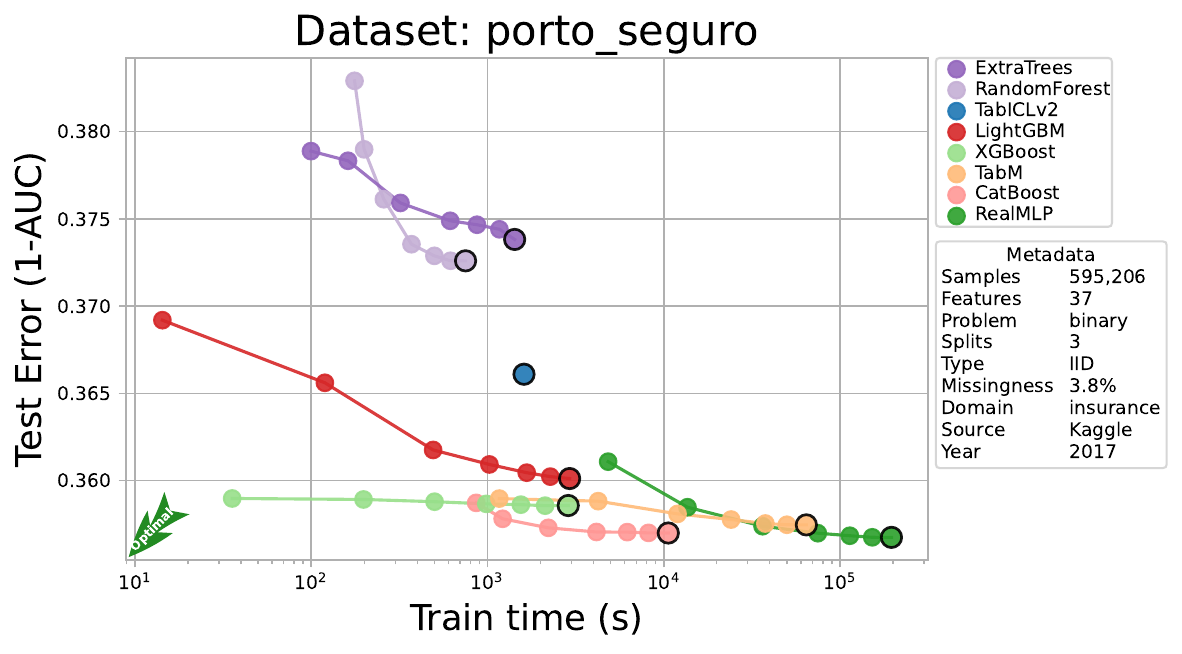}
  \end{minipage}

  \captionof{figure}{\textbf{porto\_seguro}: per-method test error (left) and HPO Pareto trajectory (right).}
  \label{fig:perdataset_porto_seguro}
\end{center}
\space
\begin{center}
  \begin{minipage}[t]{0.48\textwidth}
    \centering
    \vspace{0pt}
    \scriptsize
\begin{tabular}{l@{\quad}lll}
\toprule
 & Default & Tuned & Tuned + Ens. \\
\midrule
Linear & 0.567 $\pm$ 0.016 & 0.562 $\pm$ 0.015 & 0.559 $\pm$ 0.014 \\
RandomForest & 0.583 $\pm$ 0.014 & 0.578 $\pm$ 0.017 & 0.574 $\pm$ 0.016 \\
ExtraTrees & 0.589 $\pm$ 0.013 & 0.573 $\pm$ 0.016 & 0.567 $\pm$ 0.016 \\
CatBoost & 0.551 $\pm$ 0.016 & \underline{0.541 $\pm$ 0.019} & 0.540 $\pm$ 0.018 \\
LightGBM & 0.554 $\pm$ 0.018 & \underline{0.542 $\pm$ 0.020} & \underline{0.541 $\pm$ 0.018} \\
XGBoost & 0.556 $\pm$ 0.021 & 0.552 $\pm$ 0.018 & 0.550 $\pm$ 0.017 \\
RealMLP & 0.553 $\pm$ 0.017 & \underline{0.542 $\pm$ 0.019} & \underline{0.537 $\pm$ 0.017} \\
TabM & 0.545 $\pm$ 0.017 & \underline{0.545 $\pm$ 0.018} & 0.541 $\pm$ 0.017 \\
TabDPT & 0.555 $\pm$ 0.020 & - & - \\
TabPFN-2.6 & \underline{\textbf{0.533 $\pm$ 0.019}} & - & - \\
TabICLv2 & \underline{\textbf{\textcolor{green!50!black}{0.531 $\pm$ 0.017}}} & - & - \\
\bottomrule
\end{tabular}
\space
  \end{minipage}\hfill
  \begin{minipage}[t]{0.48\textwidth}
    \centering
    \vspace{0pt}
    \includegraphics[width=\linewidth]{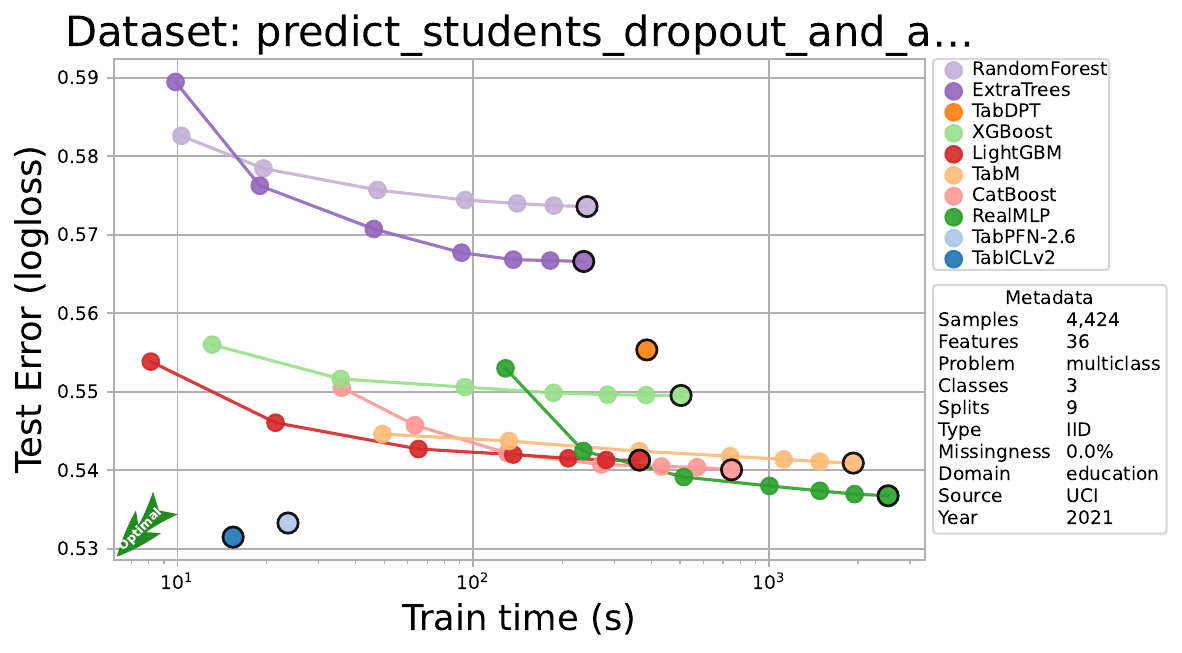}
  \end{minipage}

  \captionof{figure}{\textbf{predict\_students\_dropout\_and\_academic\_success}: per-method test error (left) and HPO Pareto trajectory (right).}
  \label{fig:perdataset_predict_students_dropout_and_academic_success}
\end{center}
\space
\begin{center}
  \begin{minipage}[t]{0.48\textwidth}
    \centering
    \vspace{0pt}
    \scriptsize
\begin{tabular}{l@{\quad}lll}
\toprule
 & Default & Tuned & Tuned + Ens. \\
\midrule
Linear & 0.024 $\pm$ 0.016 & 0.027 $\pm$ 0.016 & 0.025 $\pm$ 0.016 \\
RandomForest & 0.026 $\pm$ 0.014 & 0.028 $\pm$ 0.015 & 0.022 $\pm$ 0.012 \\
ExtraTrees & 0.026 $\pm$ 0.015 & 0.020 $\pm$ 0.011 & \underline{0.014 $\pm$ 0.008} \\
CatBoost & 0.020 $\pm$ 0.013 & \underline{0.018 $\pm$ 0.011} & 0.018 $\pm$ 0.012 \\
LightGBM & 0.023 $\pm$ 0.014 & \underline{0.017 $\pm$ 0.011} & 0.017 $\pm$ 0.011 \\
XGBoost & 0.032 $\pm$ 0.017 & 0.025 $\pm$ 0.015 & 0.025 $\pm$ 0.015 \\
RealMLP & 0.021 $\pm$ 0.016 & \underline{0.018 $\pm$ 0.015} & \underline{0.017 $\pm$ 0.015} \\
TabM & 0.025 $\pm$ 0.021 & \underline{0.022 $\pm$ 0.019} & 0.022 $\pm$ 0.019 \\
TabDPT & 0.022 $\pm$ 0.011 & - & - \\
TabPFN-2.6 & \underline{\textbf{\textcolor{green!50!black}{0.007 $\pm$ 0.005}}} & - & - \\
TabICLv2 & 0.056 $\pm$ 0.025 & - & - \\
\bottomrule
\end{tabular}
\space
  \end{minipage}\hfill
  \begin{minipage}[t]{0.48\textwidth}
    \centering
    \vspace{0pt}
    \includegraphics[width=\linewidth]{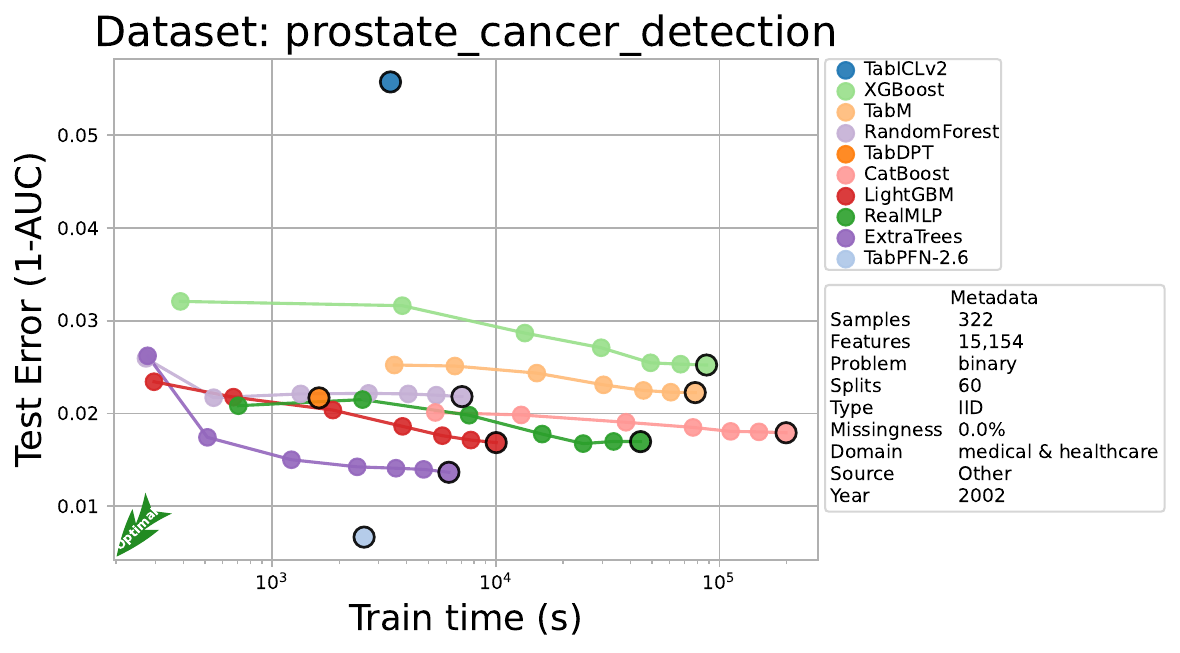}
  \end{minipage}

  \captionof{figure}{\textbf{prostate\_cancer\_detection}: per-method test error (left) and HPO Pareto trajectory (right).}
  \label{fig:perdataset_prostate_cancer_detection}
\end{center}
\space
\begin{center}
  \begin{minipage}[t]{0.48\textwidth}
    \centering
    \vspace{0pt}
    \scriptsize
\begin{tabular}{l@{\quad}lll}
\toprule
 & Default & Tuned & Tuned + Ens. \\
\midrule
Linear & 0.409 $\pm$ 0.007 & 0.407 $\pm$ 0.007 & 0.404 $\pm$ 0.006 \\
RandomForest & 0.417 $\pm$ 0.004 & 0.390 $\pm$ 0.006 & 0.389 $\pm$ 0.007 \\
ExtraTrees & 0.403 $\pm$ 0.003 & 0.387 $\pm$ 0.005 & 0.385 $\pm$ 0.006 \\
CatBoost & \underline{0.370 $\pm$ 0.006} & \underline{0.369 $\pm$ 0.006} & 0.369 $\pm$ 0.006 \\
LightGBM & 0.383 $\pm$ 0.006 & \underline{\textbf{0.368 $\pm$ 0.007}} & \underline{\textbf{\textcolor{green!50!black}{0.368 $\pm$ 0.006}}} \\
XGBoost & 0.374 $\pm$ 0.007 & 0.369 $\pm$ 0.006 & 0.369 $\pm$ 0.006 \\
RealMLP & 0.373 $\pm$ 0.004 & 0.372 $\pm$ 0.004 & 0.372 $\pm$ 0.005 \\
TabM & 0.376 $\pm$ 0.004 & 0.374 $\pm$ 0.005 & 0.374 $\pm$ 0.005 \\
TabDPT & 0.400 $\pm$ 0.005 & - & - \\
TabPFN-2.6 & 0.399 $\pm$ 0.006 & - & - \\
TabICLv2 & 0.397 $\pm$ 0.003 & - & - \\
\bottomrule
\end{tabular}
\space
  \end{minipage}\hfill
  \begin{minipage}[t]{0.48\textwidth}
    \centering
    \vspace{0pt}
    \includegraphics[width=\linewidth]{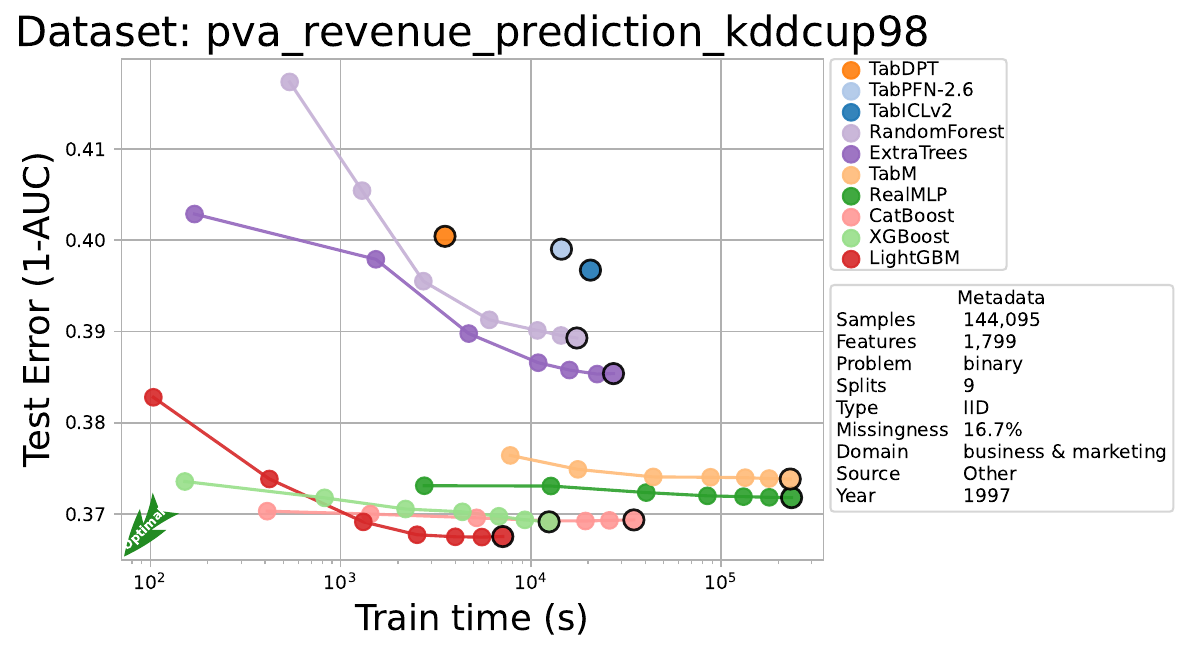}
  \end{minipage}

  \captionof{figure}{\textbf{pva\_revenue\_prediction\_kddcup98}: per-method test error (left) and HPO Pareto trajectory (right).}
  \label{fig:perdataset_pva_revenue_prediction_kddcup98}
\end{center}
\space
\begin{center}
  \begin{minipage}[t]{0.48\textwidth}
    \centering
    \vspace{0pt}
    \scriptsize
\begin{tabular}{l@{\quad}lll}
\toprule
 & Default & Tuned & Tuned + Ens. \\
\midrule
Linear & 1.290 $\pm$ 0.074 & 1.219 $\pm$ 0.071 & 1.221 $\pm$ 0.071 \\
RandomForest & 1.140 $\pm$ 0.072 & 1.124 $\pm$ 0.068 & 1.120 $\pm$ 0.067 \\
ExtraTrees & 1.108 $\pm$ 0.069 & 1.096 $\pm$ 0.066 & 1.095 $\pm$ 0.067 \\
CatBoost & 1.082 $\pm$ 0.065 & 1.083 $\pm$ 0.065 & 1.082 $\pm$ 0.065 \\
LightGBM & 1.134 $\pm$ 0.069 & 1.116 $\pm$ 0.073 & 1.117 $\pm$ 0.069 \\
XGBoost & 1.137 $\pm$ 0.075 & 1.115 $\pm$ 0.067 & 1.114 $\pm$ 0.067 \\
RealMLP & 1.068 $\pm$ 0.065 & \underline{1.068 $\pm$ 0.065} & \underline{1.067 $\pm$ 0.066} \\
TabM & 1.158 $\pm$ 0.066 & 1.137 $\pm$ 0.068 & 1.122 $\pm$ 0.065 \\
TabDPT & \underline{\textbf{\textcolor{green!50!black}{1.039 $\pm$ 0.069}}} & - & - \\
TabPFN-2.6 & 1.053 $\pm$ 0.071 & - & - \\
TabICLv2 & 1.050 $\pm$ 0.069 & - & - \\
\bottomrule
\end{tabular}
\space
  \end{minipage}\hfill
  \begin{minipage}[t]{0.48\textwidth}
    \centering
    \vspace{0pt}
    \includegraphics[width=\linewidth]{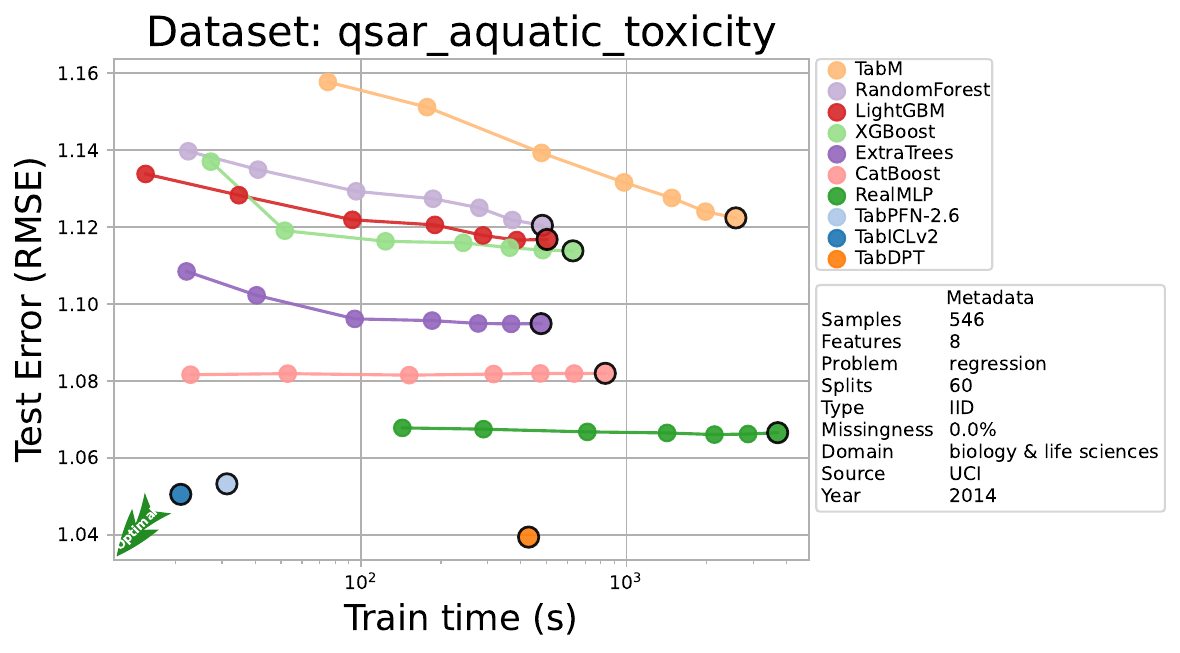}
  \end{minipage}

  \captionof{figure}{\textbf{qsar\_aquatic\_toxicity}: per-method test error (left) and HPO Pareto trajectory (right).}
  \label{fig:perdataset_qsar_aquatic_toxicity}
\end{center}
\space
\begin{center}
  \begin{minipage}[t]{0.48\textwidth}
    \centering
    \vspace{0pt}
    \scriptsize
\begin{tabular}{l@{\quad}lll}
\toprule
 & Default & Tuned & Tuned + Ens. \\
\midrule
Linear & 0.092 $\pm$ 0.012 & 0.078 $\pm$ 0.013 & 0.077 $\pm$ 0.012 \\
RandomForest & 0.069 $\pm$ 0.010 & 0.069 $\pm$ 0.011 & 0.071 $\pm$ 0.011 \\
ExtraTrees & 0.066 $\pm$ 0.009 & \underline{0.067 $\pm$ 0.011} & \underline{0.066 $\pm$ 0.010} \\
CatBoost & 0.069 $\pm$ 0.011 & 0.069 $\pm$ 0.011 & 0.068 $\pm$ 0.011 \\
LightGBM & 0.074 $\pm$ 0.011 & \underline{0.068 $\pm$ 0.012} & 0.068 $\pm$ 0.011 \\
XGBoost & 0.074 $\pm$ 0.012 & 0.069 $\pm$ 0.010 & 0.069 $\pm$ 0.011 \\
RealMLP & 0.072 $\pm$ 0.011 & 0.069 $\pm$ 0.012 & 0.067 $\pm$ 0.011 \\
TabM & 0.069 $\pm$ 0.011 & \underline{0.066 $\pm$ 0.012} & \underline{0.064 $\pm$ 0.010} \\
TabDPT & 0.064 $\pm$ 0.010 & - & - \\
TabPFN-2.6 & 0.063 $\pm$ 0.010 & - & - \\
TabICLv2 & \underline{\textbf{\textcolor{green!50!black}{0.059 $\pm$ 0.009}}} & - & - \\
\bottomrule
\end{tabular}
\space
  \end{minipage}\hfill
  \begin{minipage}[t]{0.48\textwidth}
    \centering
    \vspace{0pt}
    \includegraphics[width=\linewidth]{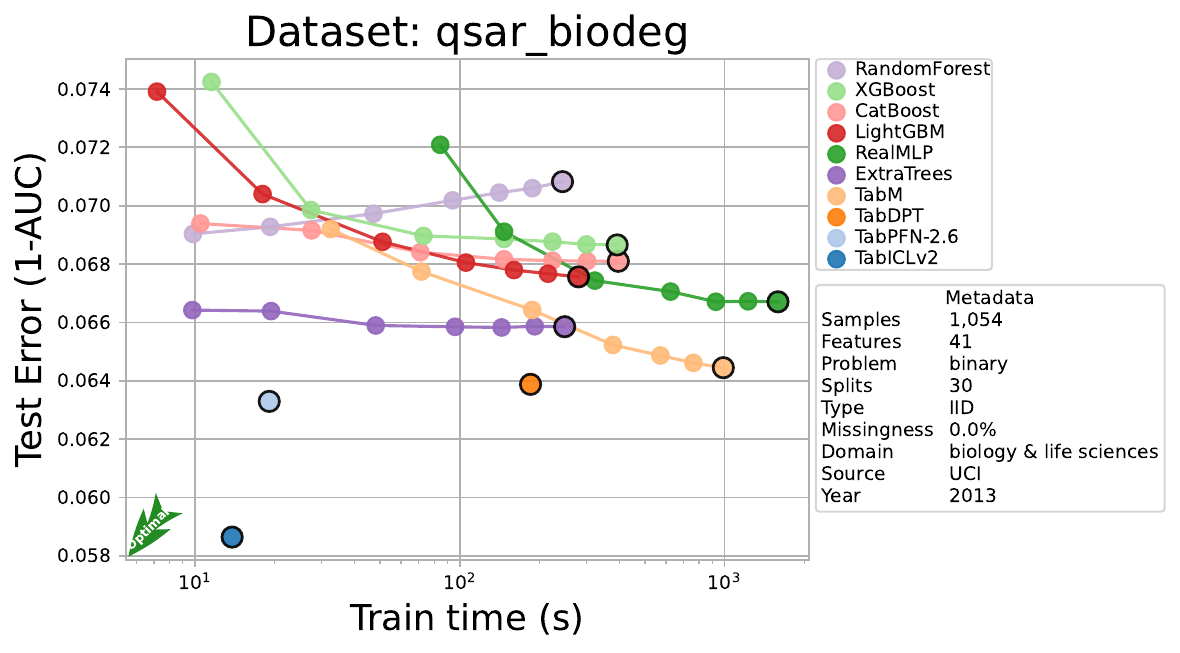}
  \end{minipage}

  \captionof{figure}{\textbf{qsar\_biodeg}: per-method test error (left) and HPO Pareto trajectory (right).}
  \label{fig:perdataset_qsar_biodeg}
\end{center}
\space
\begin{center}
  \begin{minipage}[t]{0.48\textwidth}
    \centering
    \vspace{0pt}
    \scriptsize
\begin{tabular}{l@{\quad}lll}
\toprule
 & Default & Tuned & Tuned + Ens. \\
\midrule
Linear & 0.947 $\pm$ 0.049 & 0.947 $\pm$ 0.049 & 0.947 $\pm$ 0.049 \\
RandomForest & 0.902 $\pm$ 0.053 & 0.887 $\pm$ 0.049 & 0.885 $\pm$ 0.050 \\
ExtraTrees & 0.874 $\pm$ 0.051 & \underline{0.871 $\pm$ 0.051} & \underline{0.867 $\pm$ 0.050} \\
CatBoost & 0.876 $\pm$ 0.052 & \underline{0.874 $\pm$ 0.052} & 0.874 $\pm$ 0.050 \\
LightGBM & 0.898 $\pm$ 0.045 & 0.890 $\pm$ 0.048 & 0.886 $\pm$ 0.048 \\
XGBoost & 0.906 $\pm$ 0.050 & 0.881 $\pm$ 0.053 & 0.879 $\pm$ 0.053 \\
RealMLP & 0.874 $\pm$ 0.048 & \underline{0.874 $\pm$ 0.050} & \underline{0.863 $\pm$ 0.049} \\
TabM & 0.909 $\pm$ 0.050 & 0.897 $\pm$ 0.055 & 0.893 $\pm$ 0.052 \\
TabDPT & \underline{\textbf{\textcolor{green!50!black}{0.853 $\pm$ 0.047}}} & - & - \\
TabPFN-2.6 & 0.857 $\pm$ 0.050 & - & - \\
TabICLv2 & \underline{\textbf{0.855 $\pm$ 0.052}} & - & - \\
\bottomrule
\end{tabular}
\space
  \end{minipage}\hfill
  \begin{minipage}[t]{0.48\textwidth}
    \centering
    \vspace{0pt}
    \includegraphics[width=\linewidth]{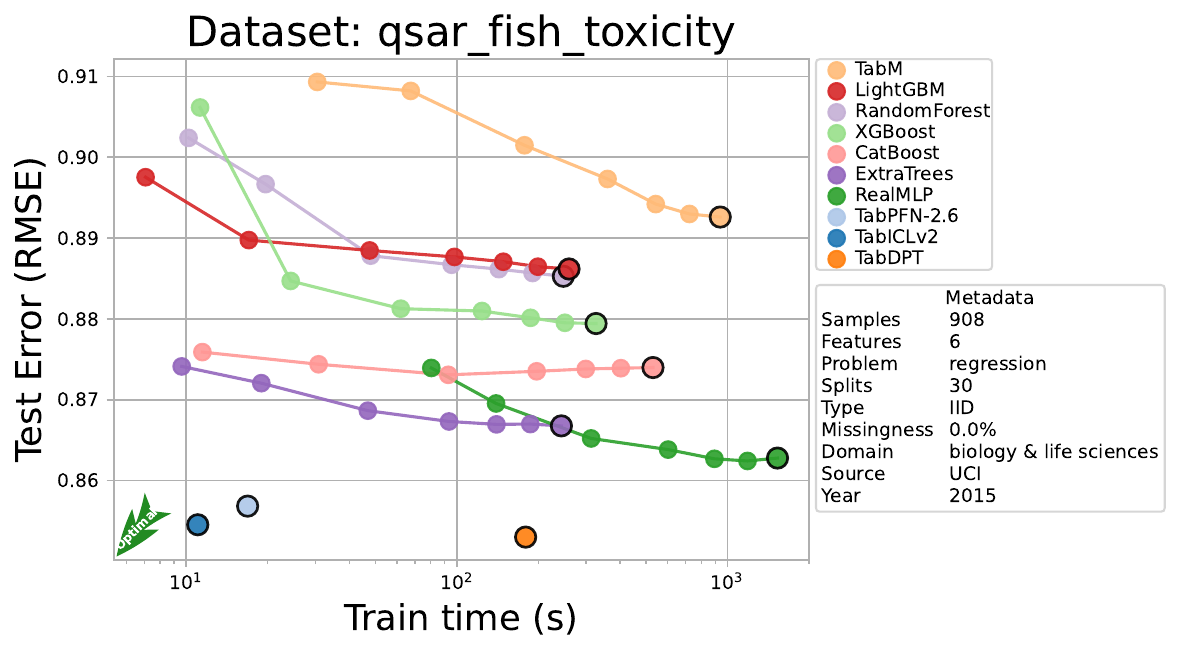}
  \end{minipage}

  \captionof{figure}{\textbf{qsar\_fish\_toxicity}: per-method test error (left) and HPO Pareto trajectory (right).}
  \label{fig:perdataset_qsar_fish_toxicity}
\end{center}
\space
\begin{center}
  \begin{minipage}[t]{0.48\textwidth}
    \centering
    \vspace{0pt}
    \scriptsize
\begin{tabular}{l@{\quad}lll}
\toprule
 & Default & Tuned & Tuned + Ens. \\
\midrule
Linear & 0.995 $\pm$ 0.019 & 0.923 $\pm$ 0.015 & 0.921 $\pm$ 0.013 \\
RandomForest & 0.789 $\pm$ 0.011 & 0.771 $\pm$ 0.014 & 0.772 $\pm$ 0.010 \\
ExtraTrees & 0.789 $\pm$ 0.011 & 0.774 $\pm$ 0.014 & 0.765 $\pm$ 0.013 \\
CatBoost & 0.749 $\pm$ 0.016 & 0.751 $\pm$ 0.013 & 0.748 $\pm$ 0.015 \\
LightGBM & 0.748 $\pm$ 0.016 & \underline{\textbf{\textcolor{green!50!black}{0.728 $\pm$ 0.014}}} & \underline{0.731 $\pm$ 0.013} \\
XGBoost & 0.763 $\pm$ 0.016 & 0.738 $\pm$ 0.012 & 0.735 $\pm$ 0.012 \\
RealMLP & \underline{0.737 $\pm$ 0.019} & 0.736 $\pm$ 0.017 & \underline{\textbf{0.730 $\pm$ 0.017}} \\
TabM & 0.739 $\pm$ 0.016 & 0.738 $\pm$ 0.018 & \underline{\textbf{0.734 $\pm$ 0.018}} \\
TabDPT & \underline{0.738 $\pm$ 0.020} & - & - \\
TabPFN-2.6 & \underline{\textbf{0.733 $\pm$ 0.017}} & - & - \\
TabICLv2 & 0.792 $\pm$ 0.014 & - & - \\
\bottomrule
\end{tabular}
\space
  \end{minipage}\hfill
  \begin{minipage}[t]{0.48\textwidth}
    \centering
    \vspace{0pt}
    \includegraphics[width=\linewidth]{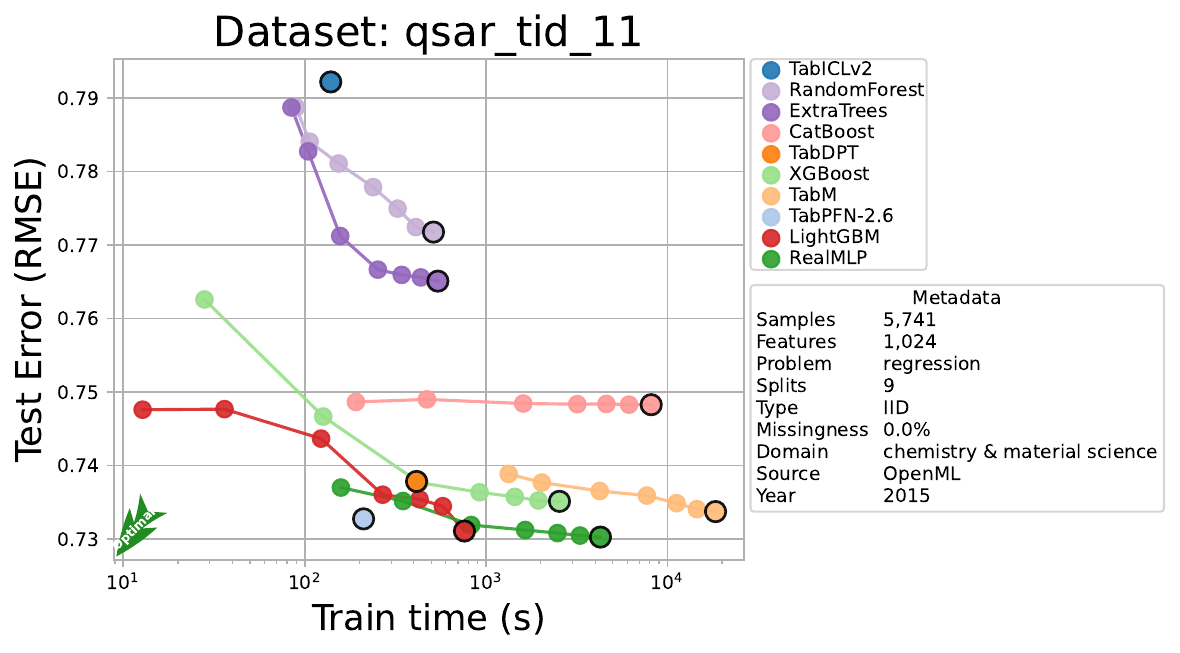}
  \end{minipage}

  \captionof{figure}{\textbf{qsar\_tid\_11}: per-method test error (left) and HPO Pareto trajectory (right).}
  \label{fig:perdataset_qsar_tid_11}
\end{center}
\space
\begin{center}
  \begin{minipage}[t]{0.48\textwidth}
    \centering
    \vspace{0pt}
    \scriptsize
\begin{tabular}{l@{\quad}lll}
\toprule
 & Default & Tuned & Tuned + Ens. \\
\midrule
Linear & 0.094 $\pm$ 0.023 & 0.084 $\pm$ 0.020 & 0.085 $\pm$ 0.020 \\
RandomForest & 0.079 $\pm$ 0.016 & \underline{0.081 $\pm$ 0.016} & 0.081 $\pm$ 0.018 \\
ExtraTrees & 0.082 $\pm$ 0.014 & 0.083 $\pm$ 0.015 & 0.082 $\pm$ 0.015 \\
CatBoost & 0.080 $\pm$ 0.018 & 0.081 $\pm$ 0.018 & 0.081 $\pm$ 0.017 \\
LightGBM & 0.092 $\pm$ 0.020 & \underline{0.076 $\pm$ 0.015} & \underline{0.076 $\pm$ 0.016} \\
XGBoost & 0.100 $\pm$ 0.020 & 0.084 $\pm$ 0.018 & 0.084 $\pm$ 0.018 \\
RealMLP & 0.091 $\pm$ 0.019 & 0.086 $\pm$ 0.021 & 0.086 $\pm$ 0.021 \\
TabM & 0.088 $\pm$ 0.020 & 0.087 $\pm$ 0.020 & 0.086 $\pm$ 0.020 \\
TabDPT & 0.094 $\pm$ 0.019 & - & - \\
TabPFN-2.6 & \underline{\textbf{0.071 $\pm$ 0.015}} & - & - \\
TabICLv2 & \underline{\textbf{\textcolor{green!50!black}{0.069 $\pm$ 0.017}}} & - & - \\
\bottomrule
\end{tabular}
\space
  \end{minipage}\hfill
  \begin{minipage}[t]{0.48\textwidth}
    \centering
    \vspace{0pt}
    \includegraphics[width=\linewidth]{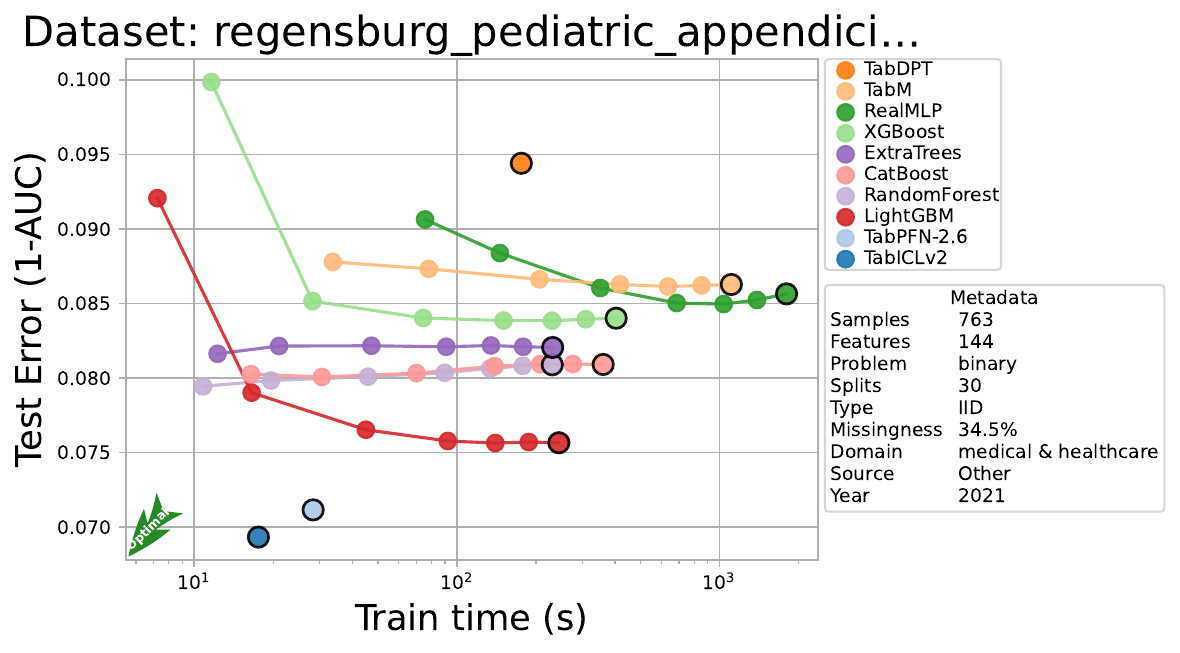}
  \end{minipage}

  \captionof{figure}{\textbf{regensburg\_pediatric\_appendicitis}: per-method test error (left) and HPO Pareto trajectory (right).}
  \label{fig:perdataset_regensburg_pediatric_appendicitis}
\end{center}
\space
\begin{center}
  \begin{minipage}[t]{0.48\textwidth}
    \centering
    \vspace{0pt}
    \scriptsize
\begin{tabular}{l@{\quad}lll}
\toprule
 & Default & Tuned & Tuned + Ens. \\
\midrule
Linear & \underline{\textbf{1450.3 $\pm$ 120.2}} & \underline{\textbf{1450.4 $\pm$ 120.5}} & \underline{\textbf{1450.4 $\pm$ 120.5}} \\
RandomForest & \underline{\textbf{1129.3 $\pm$ 130.2}} & \underline{\textbf{1128.8 $\pm$ 135.7}} & \underline{\textbf{1127.1 $\pm$ 135.0}} \\
ExtraTrees & \underline{\textbf{1197.7 $\pm$ 190.7}} & \underline{\textbf{1183.6 $\pm$ 193.1}} & \underline{\textbf{1182.2 $\pm$ 191.6}} \\
CatBoost & \underline{\textbf{1059.2 $\pm$ 129.4}} & \underline{\textbf{1021.1 $\pm$ 143.5}} & \underline{\textbf{1018.7 $\pm$ 131.6}} \\
LightGBM & \underline{\textbf{1027.3 $\pm$ 136.4}} & \underline{\textbf{1007.3 $\pm$ 136.9}} & \underline{\textbf{999.0 $\pm$ 141.9}} \\
XGBoost & \underline{\textbf{1048.2 $\pm$ 138.4}} & \underline{\textbf{1011.4 $\pm$ 107.5}} & \underline{\textbf{1009.2 $\pm$ 112.1}} \\
RealMLP & \underline{\textbf{1029.4 $\pm$ 242.0}} & \underline{\textbf{1006.1 $\pm$ 216.9}} & \underline{\textbf{\textcolor{green!50!black}{993.7 $\pm$ 217.1}}} \\
TabM & \underline{\textbf{1110.7 $\pm$ 103.4}} & \underline{\textbf{1090.7 $\pm$ 86.2}} & \underline{\textbf{1021.3 $\pm$ 104.4}} \\
TabDPT & - & - & - \\
TabPFN-2.6 & - & - & - \\
TabICLv2 & \underline{\textbf{2151.7 $\pm$ 169.8}} & - & - \\
\bottomrule
\end{tabular}
\space
  \end{minipage}\hfill
  \begin{minipage}[t]{0.48\textwidth}
    \centering
    \vspace{0pt}
    \includegraphics[width=\linewidth]{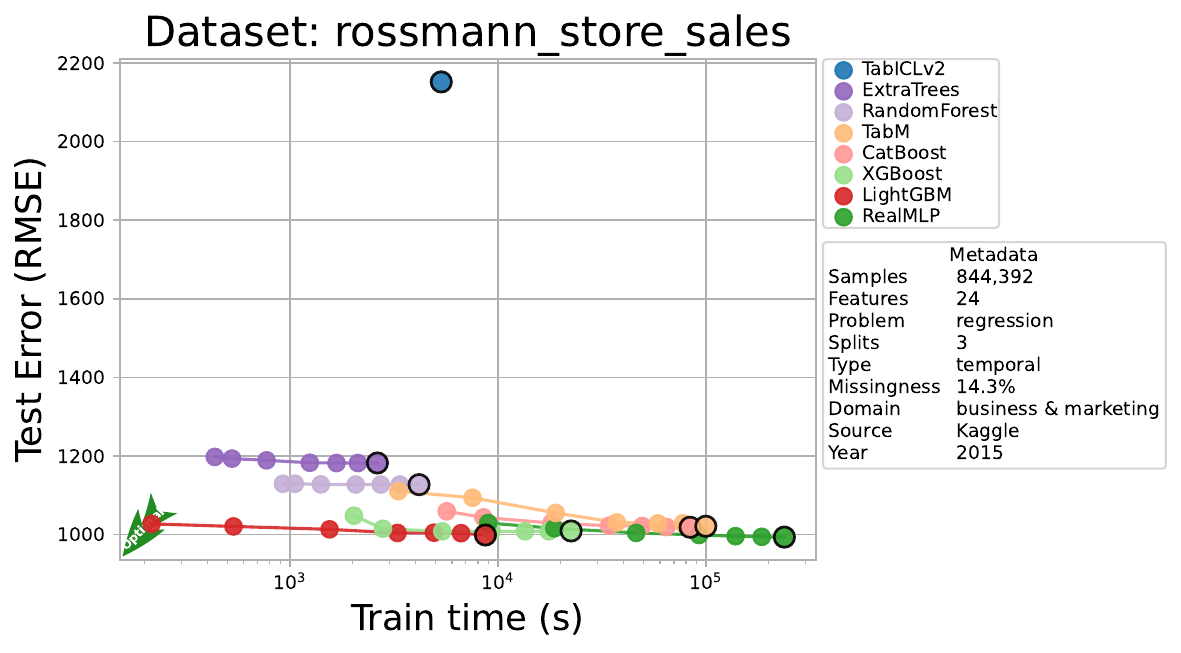}
  \end{minipage}

  \captionof{figure}{\textbf{rossmann\_store\_sales}: per-method test error (left) and HPO Pareto trajectory (right).}
  \label{fig:perdataset_rossmann_store_sales}
\end{center}
\space
\begin{center}
  \begin{minipage}[t]{0.48\textwidth}
    \centering
    \vspace{0pt}
    \scriptsize
\begin{tabular}{l@{\quad}lll}
\toprule
 & Default & Tuned & Tuned + Ens. \\
\midrule
Linear & 0.184 $\pm$ 0.003 & 0.183 $\pm$ 0.004 & 0.183 $\pm$ 0.004 \\
RandomForest & 0.225 $\pm$ 0.008 & 0.169 $\pm$ 0.007 & 0.167 $\pm$ 0.006 \\
ExtraTrees & 0.236 $\pm$ 0.008 & 0.178 $\pm$ 0.005 & 0.176 $\pm$ 0.005 \\
CatBoost & 0.162 $\pm$ 0.005 & 0.161 $\pm$ 0.005 & 0.161 $\pm$ 0.005 \\
LightGBM & 0.163 $\pm$ 0.005 & 0.161 $\pm$ 0.005 & 0.160 $\pm$ 0.005 \\
XGBoost & 0.163 $\pm$ 0.005 & 0.160 $\pm$ 0.005 & 0.160 $\pm$ 0.005 \\
RealMLP & 0.160 $\pm$ 0.005 & \underline{0.159 $\pm$ 0.005} & \underline{0.159 $\pm$ 0.005} \\
TabM & 0.161 $\pm$ 0.005 & \underline{0.159 $\pm$ 0.005} & \underline{0.159 $\pm$ 0.005} \\
TabDPT & 0.185 $\pm$ 0.005 & - & - \\
TabPFN-2.6 & \underline{\textbf{\textcolor{green!50!black}{0.158 $\pm$ 0.005}}} & - & - \\
TabICLv2 & 0.183 $\pm$ 0.011 & - & - \\
\bottomrule
\end{tabular}
\space
  \end{minipage}\hfill
  \begin{minipage}[t]{0.48\textwidth}
    \centering
    \vspace{0pt}
    \includegraphics[width=\linewidth]{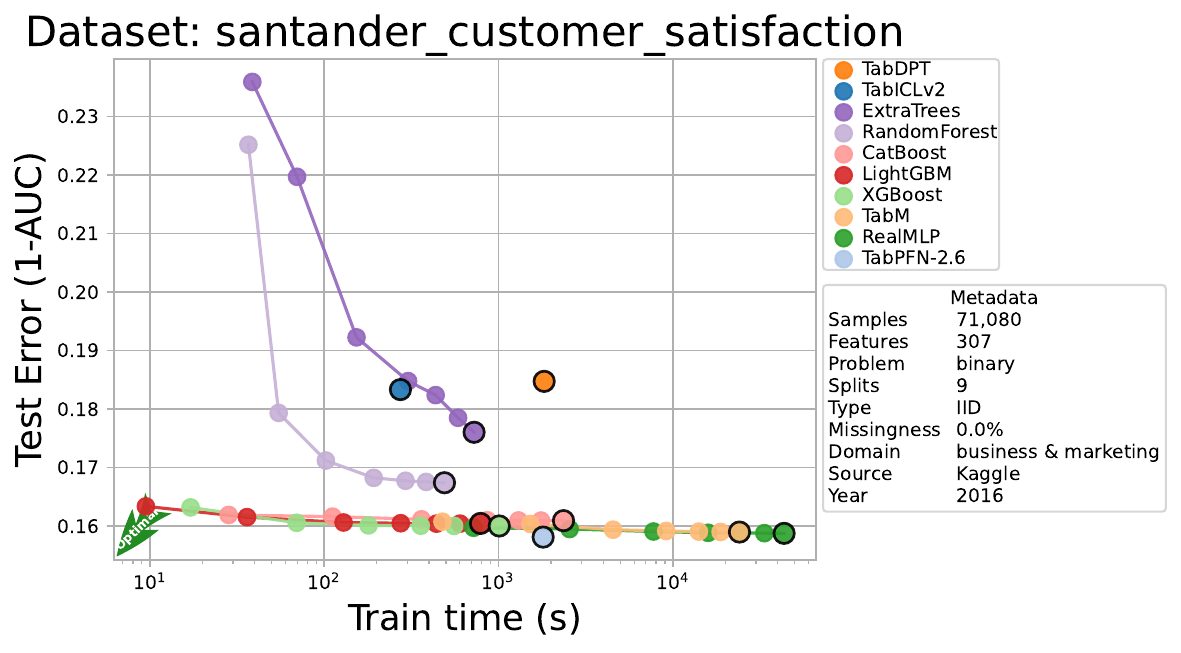}
  \end{minipage}

  \captionof{figure}{\textbf{santander\_customer\_satisfaction}: per-method test error (left) and HPO Pareto trajectory (right).}
  \label{fig:perdataset_santander_customer_satisfaction}
\end{center}
\space
\begin{center}
  \begin{minipage}[t]{0.48\textwidth}
    \centering
    \vspace{0pt}
    \scriptsize
\begin{tabular}{l@{\quad}lll}
\toprule
 & Default & Tuned & Tuned + Ens. \\
\midrule
Linear & 0.140 $\pm$ 0.002 & 0.140 $\pm$ 0.002 & 0.140 $\pm$ 0.002 \\
RandomForest & 0.141 $\pm$ 0.002 & 0.141 $\pm$ 0.002 & 0.143 $\pm$ 0.002 \\
ExtraTrees & 0.166 $\pm$ 0.001 & 0.175 $\pm$ 0.002 & 0.138 $\pm$ 0.002 \\
CatBoost & \underline{0.093 $\pm$ 0.002} & 0.092 $\pm$ 0.002 & 0.092 $\pm$ 0.002 \\
LightGBM & 0.097 $\pm$ 0.002 & \underline{\textbf{0.091 $\pm$ 0.002}} & \underline{\textbf{\textcolor{green!50!black}{0.091 $\pm$ 0.002}}} \\
XGBoost & 0.099 $\pm$ 0.002 & 0.093 $\pm$ 0.002 & 0.093 $\pm$ 0.002 \\
RealMLP & 0.113 $\pm$ 0.002 & 0.103 $\pm$ 0.001 & 0.102 $\pm$ 0.002 \\
TabM & 0.103 $\pm$ 0.002 & 0.103 $\pm$ 0.002 & 0.102 $\pm$ 0.002 \\
TabDPT & - & - & - \\
TabPFN-2.6 & - & - & - \\
TabICLv2 & 0.116 $\pm$ 0.002 & - & - \\
\bottomrule
\end{tabular}
\space
  \end{minipage}\hfill
  \begin{minipage}[t]{0.48\textwidth}
    \centering
    \vspace{0pt}
    \includegraphics[width=\linewidth]{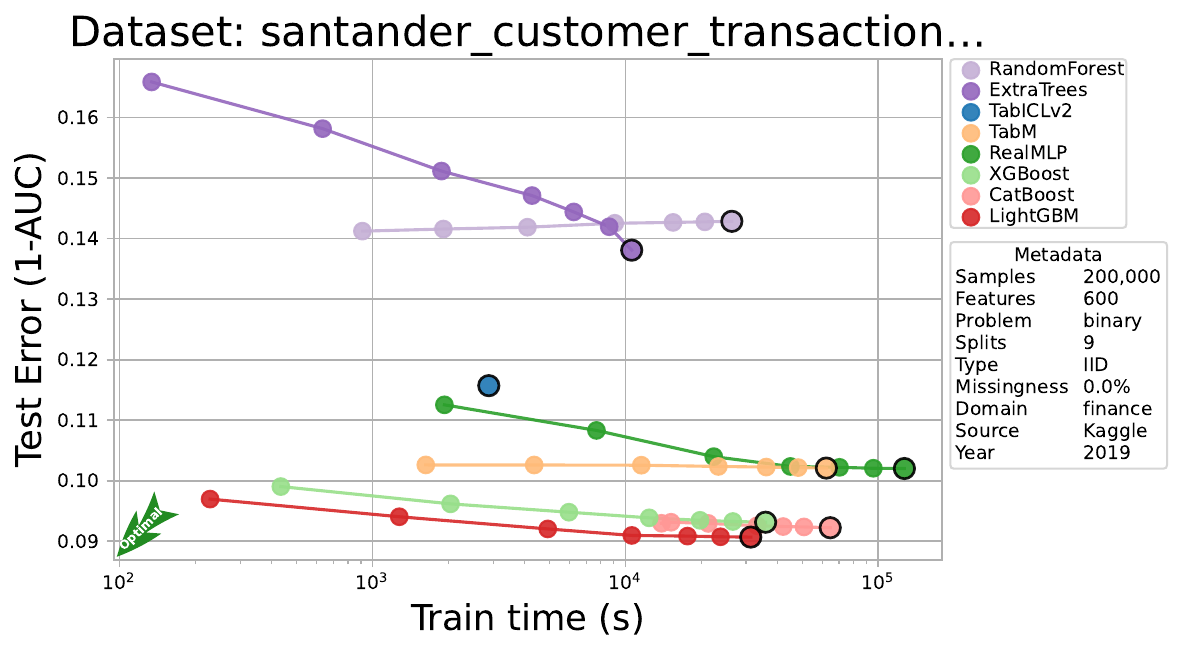}
  \end{minipage}

  \captionof{figure}{\textbf{santander\_customer\_transaction\_prediction}: per-method test error (left) and HPO Pareto trajectory (right).}
  \label{fig:perdataset_santander_customer_transaction_prediction}
\end{center}
\space
\begin{center}
  \begin{minipage}[t]{0.48\textwidth}
    \centering
    \vspace{0pt}
    \scriptsize
\begin{tabular}{l@{\quad}lll}
\toprule
 & Default & Tuned & Tuned + Ens. \\
\midrule
Linear & 1.486 $\pm$ 0.019 & 1.413 $\pm$ 0.017 & 1.404 $\pm$ 0.015 \\
RandomForest & 1.341 $\pm$ 0.023 & \underline{1.338 $\pm$ 0.019} & \underline{1.337 $\pm$ 0.020} \\
ExtraTrees & 1.344 $\pm$ 0.017 & 1.345 $\pm$ 0.016 & 1.344 $\pm$ 0.016 \\
CatBoost & 1.350 $\pm$ 0.014 & 1.350 $\pm$ 0.015 & 1.348 $\pm$ 0.016 \\
LightGBM & 1.349 $\pm$ 0.020 & 1.348 $\pm$ 0.019 & 1.344 $\pm$ 0.019 \\
XGBoost & 1.346 $\pm$ 0.020 & \underline{1.343 $\pm$ 0.023} & 1.342 $\pm$ 0.021 \\
RealMLP & 1.379 $\pm$ 0.017 & 1.362 $\pm$ 0.021 & 1.347 $\pm$ 0.019 \\
TabM & 1.363 $\pm$ 0.018 & 1.363 $\pm$ 0.018 & 1.360 $\pm$ 0.018 \\
TabDPT & 1.445 $\pm$ 0.022 & - & - \\
TabPFN-2.6 & 1.329 $\pm$ 0.019 & - & - \\
TabICLv2 & \underline{\textbf{\textcolor{green!50!black}{1.323 $\pm$ 0.020}}} & - & - \\
\bottomrule
\end{tabular}
\space
  \end{minipage}\hfill
  \begin{minipage}[t]{0.48\textwidth}
    \centering
    \vspace{0pt}
    \includegraphics[width=\linewidth]{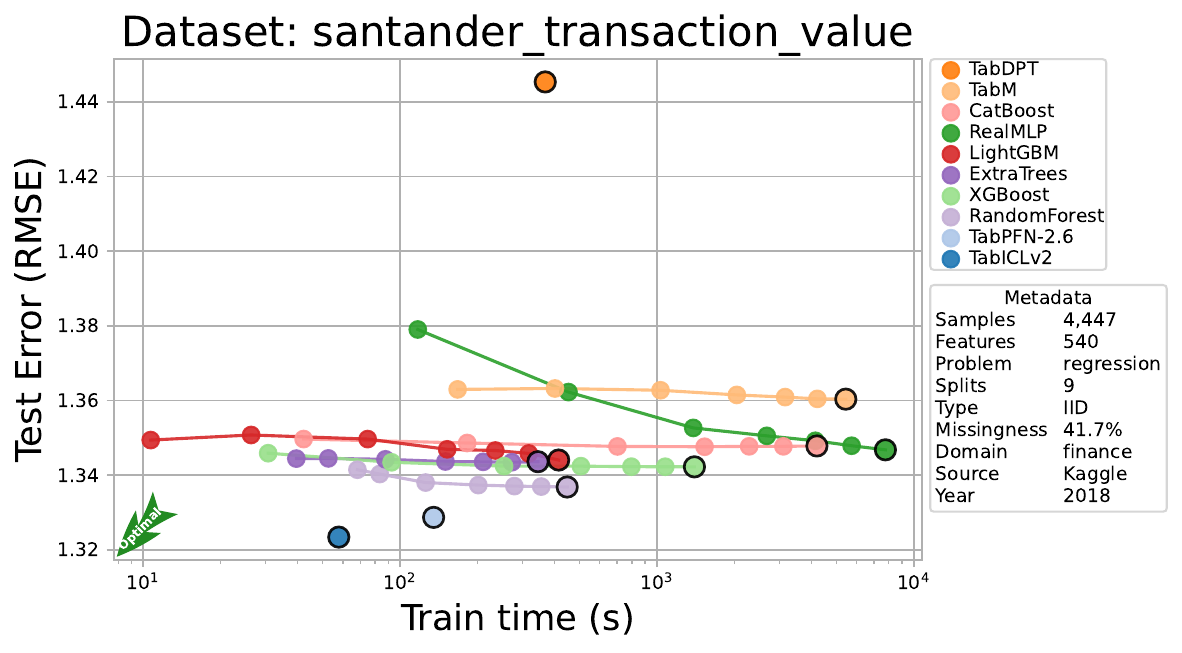}
  \end{minipage}

  \captionof{figure}{\textbf{santander\_transaction\_value}: per-method test error (left) and HPO Pareto trajectory (right).}
  \label{fig:perdataset_santander_transaction_value}
\end{center}
\space
\begin{center}
  \begin{minipage}[t]{0.48\textwidth}
    \centering
    \vspace{0pt}
    \scriptsize
\begin{tabular}{l@{\quad}lll}
\toprule
 & Default & Tuned & Tuned + Ens. \\
\midrule
Linear & 3.757 $\pm$ 0.485 & 2.666 $\pm$ 0.232 & 2.722 $\pm$ 0.259 \\
RandomForest & 2.189 $\pm$ 0.247 & 2.141 $\pm$ 0.217 & 2.138 $\pm$ 0.216 \\
ExtraTrees & 2.064 $\pm$ 0.183 & \underline{2.065 $\pm$ 0.173} & \underline{2.054 $\pm$ 0.177} \\
CatBoost & 2.111 $\pm$ 0.204 & 2.098 $\pm$ 0.217 & \underline{2.083 $\pm$ 0.205} \\
LightGBM & 2.236 $\pm$ 0.201 & \underline{2.046 $\pm$ 0.181} & \underline{2.058 $\pm$ 0.180} \\
XGBoost & 2.308 $\pm$ 0.228 & 2.115 $\pm$ 0.208 & 2.115 $\pm$ 0.202 \\
RealMLP & 2.110 $\pm$ 0.221 & 2.105 $\pm$ 0.210 & \underline{2.086 $\pm$ 0.200} \\
TabM & 2.101 $\pm$ 0.189 & 2.111 $\pm$ 0.195 & \underline{2.083 $\pm$ 0.194} \\
TabDPT & 2.413 $\pm$ 0.224 & - & - \\
TabPFN-2.6 & \underline{\textbf{\textcolor{green!50!black}{1.976 $\pm$ 0.173}}} & - & - \\
TabICLv2 & 2.136 $\pm$ 0.171 & - & - \\
\bottomrule
\end{tabular}
\space
  \end{minipage}\hfill
  \begin{minipage}[t]{0.48\textwidth}
    \centering
    \vspace{0pt}
    \includegraphics[width=\linewidth]{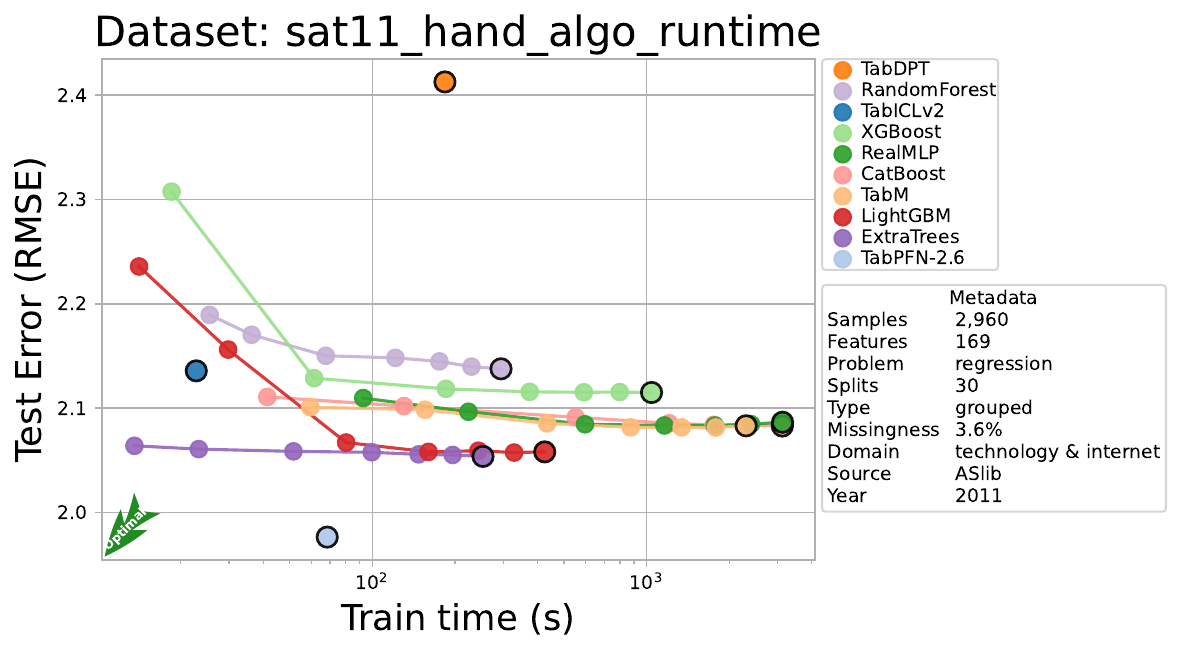}
  \end{minipage}

  \captionof{figure}{\textbf{sat11\_hand\_algo\_runtime}: per-method test error (left) and HPO Pareto trajectory (right).}
  \label{fig:perdataset_sat11_hand_algo_runtime}
\end{center}
\space
\begin{center}
  \begin{minipage}[t]{0.48\textwidth}
    \centering
    \vspace{0pt}
    \scriptsize
\begin{tabular}{l@{\quad}lll}
\toprule
 & Default & Tuned & Tuned + Ens. \\
\midrule
Linear & \underline{\textbf{0.579 $\pm$ 0.360}} & \underline{\textbf{0.276 $\pm$ 0.021}} & \underline{\textbf{0.277 $\pm$ 0.031}} \\
RandomForest & \underline{\textbf{0.273 $\pm$ 0.021}} & \underline{\textbf{0.273 $\pm$ 0.020}} & \underline{\textbf{0.267 $\pm$ 0.019}} \\
ExtraTrees & \underline{\textbf{0.262 $\pm$ 0.018}} & \underline{\textbf{0.262 $\pm$ 0.019}} & \underline{\textbf{0.260 $\pm$ 0.018}} \\
CatBoost & \underline{\textbf{0.252 $\pm$ 0.018}} & \underline{\textbf{0.255 $\pm$ 0.018}} & \underline{\textbf{0.254 $\pm$ 0.017}} \\
LightGBM & \underline{\textbf{0.254 $\pm$ 0.016}} & \underline{\textbf{0.253 $\pm$ 0.016}} & \underline{\textbf{0.251 $\pm$ 0.018}} \\
XGBoost & \underline{\textbf{0.254 $\pm$ 0.017}} & \underline{\textbf{0.250 $\pm$ 0.018}} & \underline{\textbf{0.250 $\pm$ 0.017}} \\
RealMLP & \underline{\textbf{0.252 $\pm$ 0.017}} & \underline{\textbf{0.248 $\pm$ 0.013}} & \underline{\textbf{0.247 $\pm$ 0.015}} \\
TabM & \underline{\textbf{0.260 $\pm$ 0.015}} & \underline{\textbf{0.256 $\pm$ 0.011}} & \underline{\textbf{0.253 $\pm$ 0.012}} \\
TabDPT & \underline{\textbf{0.261 $\pm$ 0.022}} & - & - \\
TabPFN-2.6 & \underline{\textbf{\textcolor{green!50!black}{0.243 $\pm$ 0.016}}} & - & - \\
TabICLv2 & \underline{\textbf{0.285 $\pm$ 0.037}} & - & - \\
\bottomrule
\end{tabular}
\space
  \end{minipage}\hfill
  \begin{minipage}[t]{0.48\textwidth}
    \centering
    \vspace{0pt}
    \includegraphics[width=\linewidth]{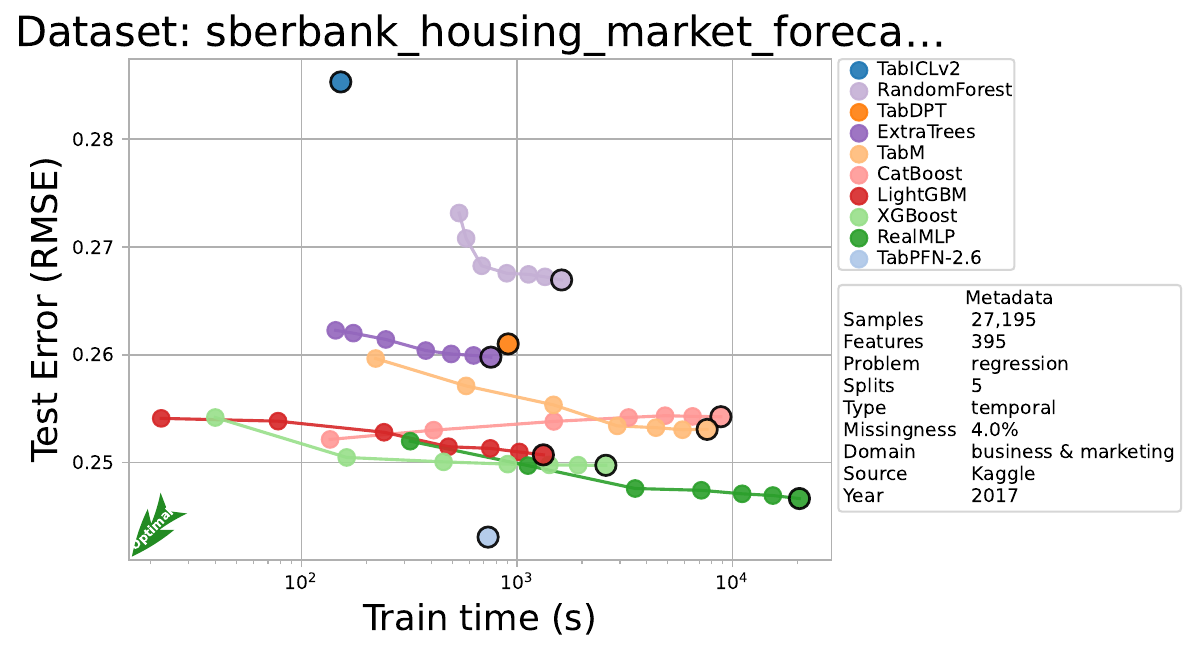}
  \end{minipage}

  \captionof{figure}{\textbf{sberbank\_housing\_market\_forecasting}: per-method test error (left) and HPO Pareto trajectory (right).}
  \label{fig:perdataset_sberbank_housing_market_forecasting}
\end{center}
\space
\begin{center}
  \begin{minipage}[t]{0.48\textwidth}
    \centering
    \vspace{0pt}
    \scriptsize
\begin{tabular}{l@{\quad}lll}
\toprule
 & Default & Tuned & Tuned + Ens. \\
\midrule
Linear & 0.147 $\pm$ 0.003 & 0.147 $\pm$ 0.003 & 0.136 $\pm$ 0.003 \\
RandomForest & 0.085 $\pm$ 0.002 & 0.077 $\pm$ 0.003 & 0.077 $\pm$ 0.003 \\
ExtraTrees & 0.131 $\pm$ 0.002 & 0.085 $\pm$ 0.002 & 0.081 $\pm$ 0.002 \\
CatBoost & 0.075 $\pm$ 0.002 & \underline{0.074 $\pm$ 0.002} & \underline{0.074 $\pm$ 0.002} \\
LightGBM & 0.088 $\pm$ 0.002 & 0.088 $\pm$ 0.002 & 0.084 $\pm$ 0.002 \\
XGBoost & 0.074 $\pm$ 0.002 & \underline{0.074 $\pm$ 0.002} & \underline{0.073 $\pm$ 0.002} \\
RealMLP & 0.104 $\pm$ 0.004 & 0.087 $\pm$ 0.002 & 0.088 $\pm$ 0.002 \\
TabM & 0.097 $\pm$ 0.003 & 0.085 $\pm$ 0.002 & 0.085 $\pm$ 0.002 \\
TabDPT & 0.085 $\pm$ 0.002 & - & - \\
TabPFN-2.6 & \underline{\textbf{\textcolor{green!50!black}{0.069 $\pm$ 0.002}}} & - & - \\
TabICLv2 & 0.071 $\pm$ 0.003 & - & - \\
\bottomrule
\end{tabular}
\space
  \end{minipage}\hfill
  \begin{minipage}[t]{0.48\textwidth}
    \centering
    \vspace{0pt}
    \includegraphics[width=\linewidth]{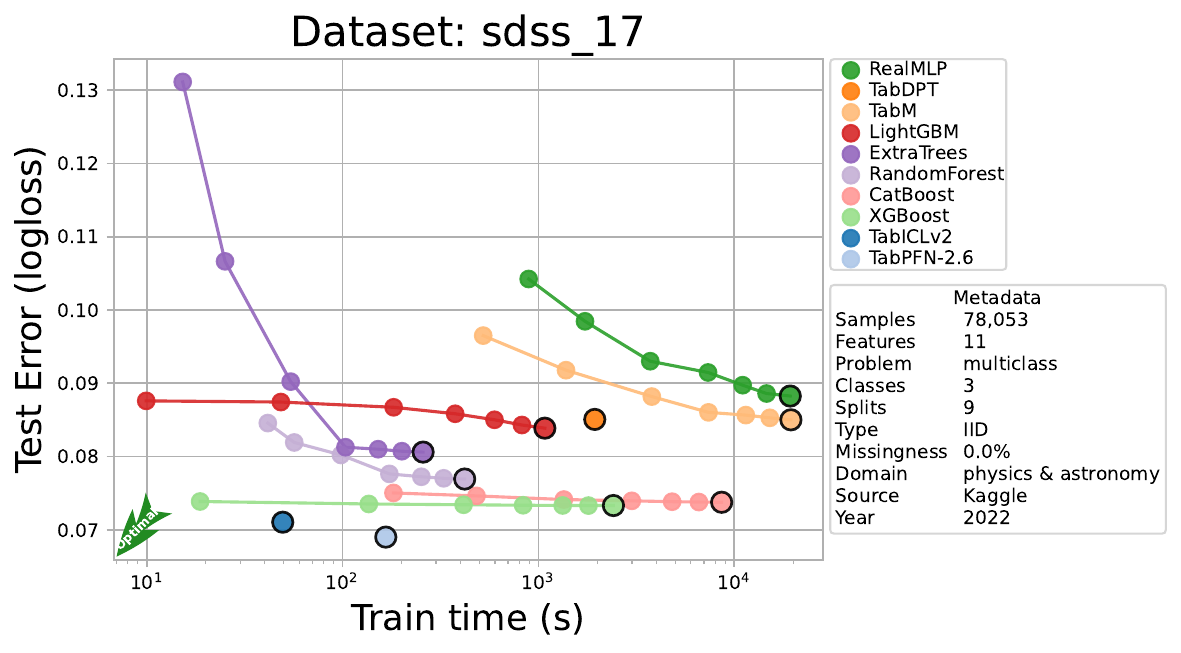}
  \end{minipage}

  \captionof{figure}{\textbf{sdss\_17}: per-method test error (left) and HPO Pareto trajectory (right).}
  \label{fig:perdataset_sdss_17}
\end{center}
\space
\begin{center}
  \begin{minipage}[t]{0.48\textwidth}
    \centering
    \vspace{0pt}
    \scriptsize
\begin{tabular}{l@{\quad}lll}
\toprule
 & Default & Tuned & Tuned + Ens. \\
\midrule
Linear & 0.242 $\pm$ 0.024 & 0.241 $\pm$ 0.023 & 0.238 $\pm$ 0.023 \\
RandomForest & 0.249 $\pm$ 0.022 & 0.236 $\pm$ 0.027 & 0.235 $\pm$ 0.025 \\
ExtraTrees & 0.251 $\pm$ 0.024 & 0.235 $\pm$ 0.026 & \underline{0.226 $\pm$ 0.024} \\
CatBoost & 0.225 $\pm$ 0.022 & \underline{0.228 $\pm$ 0.022} & \underline{0.228 $\pm$ 0.022} \\
LightGBM & 0.247 $\pm$ 0.025 & 0.231 $\pm$ 0.021 & 0.230 $\pm$ 0.021 \\
XGBoost & 0.240 $\pm$ 0.025 & \underline{0.231 $\pm$ 0.020} & \underline{0.230 $\pm$ 0.020} \\
RealMLP & 0.239 $\pm$ 0.026 & 0.238 $\pm$ 0.024 & 0.232 $\pm$ 0.023 \\
TabM & 0.234 $\pm$ 0.024 & 0.234 $\pm$ 0.026 & 0.233 $\pm$ 0.024 \\
TabDPT & \underline{\textbf{0.222 $\pm$ 0.023}} & - & - \\
TabPFN-2.6 & \underline{\textbf{\textcolor{green!50!black}{0.220 $\pm$ 0.024}}} & - & - \\
TabICLv2 & \underline{\textbf{0.227 $\pm$ 0.035}} & - & - \\
\bottomrule
\end{tabular}
\space
  \end{minipage}\hfill
  \begin{minipage}[t]{0.48\textwidth}
    \centering
    \vspace{0pt}
    \includegraphics[width=\linewidth]{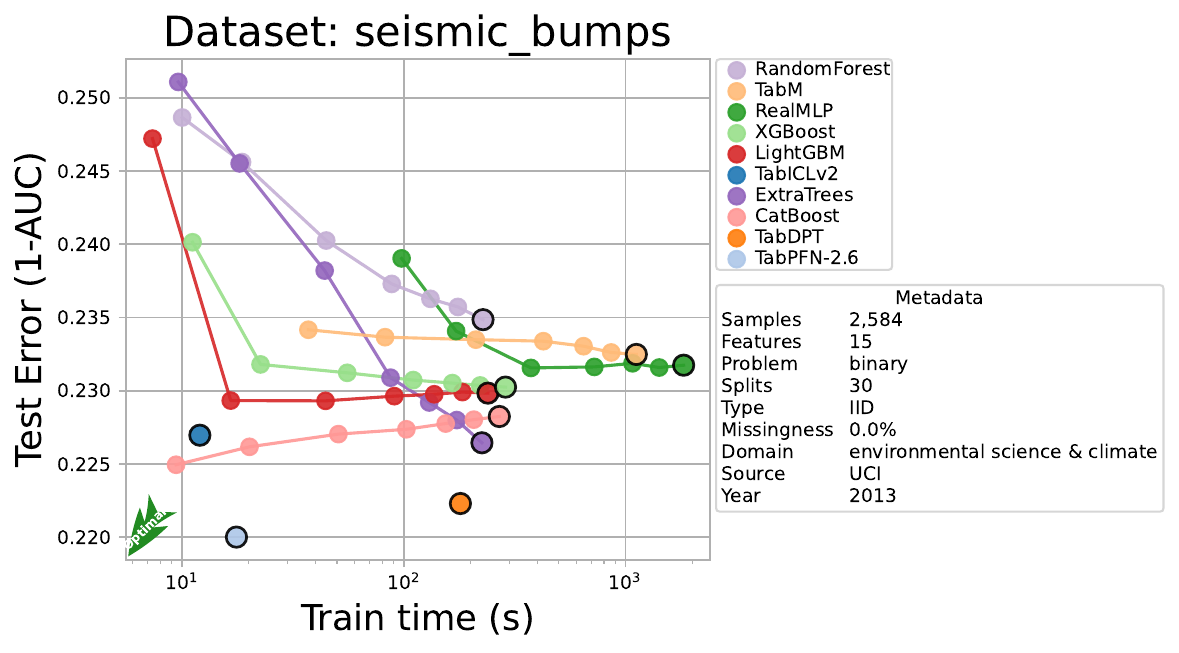}
  \end{minipage}

  \captionof{figure}{\textbf{seismic\_bumps}: per-method test error (left) and HPO Pareto trajectory (right).}
  \label{fig:perdataset_seismic_bumps}
\end{center}
\space
\begin{center}
  \begin{minipage}[t]{0.48\textwidth}
    \centering
    \vspace{0pt}
    \scriptsize
\begin{tabular}{l@{\quad}lll}
\toprule
 & Default & Tuned & Tuned + Ens. \\
\midrule
Linear & 0.304 & 0.280 & 0.280 \\
RandomForest & 0.196 & 0.196 & 0.195 \\
ExtraTrees & 0.205 & 0.205 & 0.202 \\
CatBoost & 0.190 & 0.188 & 0.188 \\
LightGBM & 0.187 & 0.179 & \textcolor{green!50!black}{0.178} \\
XGBoost & 0.185 & 0.181 & 0.181 \\
RealMLP & 0.186 & 0.183 & 0.182 \\
TabM & 0.181 & 0.181 & 0.178 \\
TabDPT & - & - & - \\
TabPFN-2.6 & - & - & - \\
TabICLv2 & 0.201 & - & - \\
\bottomrule
\end{tabular}
\space
  \end{minipage}\hfill
  \begin{minipage}[t]{0.48\textwidth}
    \centering
    \vspace{0pt}
    \includegraphics[width=\linewidth]{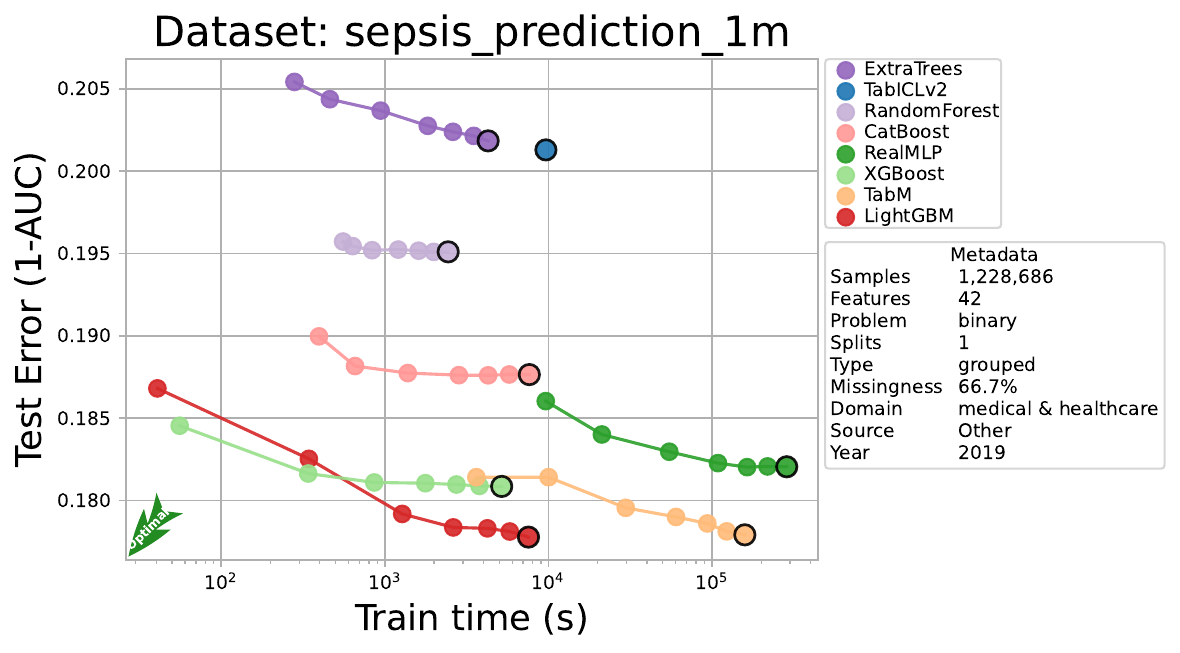}
  \end{minipage}

  \captionof{figure}{\textbf{sepsis\_prediction\_1m}: per-method test error (left) and HPO Pareto trajectory (right).}
  \label{fig:perdataset_sepsis_prediction_1m}
\end{center}
\space
\begin{center}
  \begin{minipage}[t]{0.48\textwidth}
    \centering
    \vspace{0pt}
    \scriptsize
\begin{tabular}{l@{\quad}lll}
\toprule
 & Default & Tuned & Tuned + Ens. \\
\midrule
Linear & 0.294 $\pm$ 0.001 & 0.294 $\pm$ 0.002 & 0.294 $\pm$ 0.002 \\
RandomForest & 0.311 $\pm$ 0.002 & 0.294 $\pm$ 0.002 & 0.294 $\pm$ 0.002 \\
ExtraTrees & 0.311 $\pm$ 0.002 & 0.294 $\pm$ 0.001 & 0.294 $\pm$ 0.002 \\
CatBoost & \underline{0.294 $\pm$ 0.002} & 0.293 $\pm$ 0.002 & 0.293 $\pm$ 0.002 \\
LightGBM & 0.297 $\pm$ 0.002 & \underline{\textbf{\textcolor{green!50!black}{0.293 $\pm$ 0.002}}} & \underline{\textbf{0.293 $\pm$ 0.002}} \\
XGBoost & 0.297 $\pm$ 0.002 & 0.294 $\pm$ 0.001 & 0.294 $\pm$ 0.002 \\
RealMLP & \underline{\textbf{0.293 $\pm$ 0.002}} & \underline{\textbf{0.293 $\pm$ 0.002}} & \underline{\textbf{0.293 $\pm$ 0.002}} \\
TabM & 0.294 $\pm$ 0.002 & 0.294 $\pm$ 0.002 & 0.293 $\pm$ 0.002 \\
TabDPT & 0.296 $\pm$ 0.002 & - & - \\
TabPFN-2.6 & \underline{\textbf{0.293 $\pm$ 0.002}} & - & - \\
TabICLv2 & 0.301 $\pm$ 0.007 & - & - \\
\bottomrule
\end{tabular}
\space
  \end{minipage}\hfill
  \begin{minipage}[t]{0.48\textwidth}
    \centering
    \vspace{0pt}
    \includegraphics[width=\linewidth]{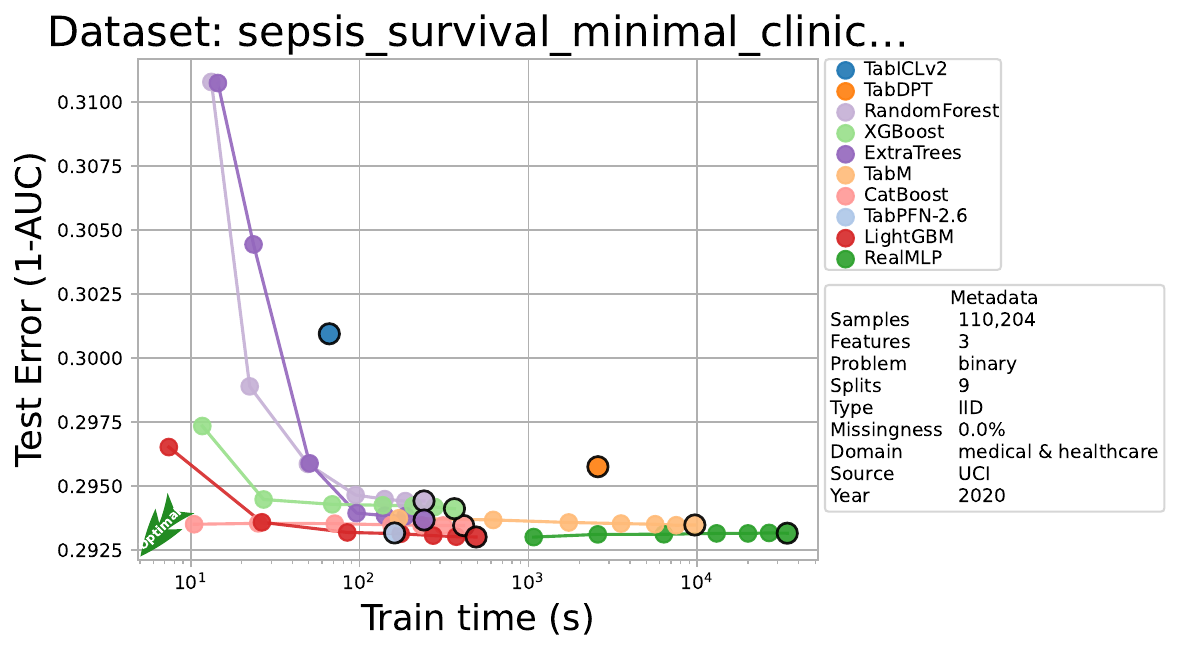}
  \end{minipage}

  \captionof{figure}{\textbf{sepsis\_survival\_minimal\_clinical\_records}: per-method test error (left) and HPO Pareto trajectory (right).}
  \label{fig:perdataset_sepsis_survival_minimal_clinical_records}
\end{center}
\space
\begin{center}
  \begin{minipage}[t]{0.48\textwidth}
    \centering
    \vspace{0pt}
    \scriptsize
\begin{tabular}{l@{\quad}lll}
\toprule
 & Default & Tuned & Tuned + Ens. \\
\midrule
Linear & 1.416 $\pm$ 0.062 & 1.420 $\pm$ 0.067 & 1.409 $\pm$ 0.069 \\
RandomForest & \underline{\textbf{1.344 $\pm$ 0.073}} & \textbf{1.342 $\pm$ 0.075} & \textbf{1.341 $\pm$ 0.074} \\
ExtraTrees & 1.360 $\pm$ 0.091 & 1.360 $\pm$ 0.091 & \underline{1.355 $\pm$ 0.091} \\
CatBoost & \underline{\textbf{1.315 $\pm$ 0.068}} & \underline{\textbf{1.315 $\pm$ 0.068}} & \underline{\textbf{1.315 $\pm$ 0.068}} \\
LightGBM & \underline{\textbf{1.321 $\pm$ 0.065}} & \underline{\textbf{\textcolor{green!50!black}{1.312 $\pm$ 0.069}}} & \underline{\textbf{1.315 $\pm$ 0.071}} \\
XGBoost & \underline{\textbf{1.323 $\pm$ 0.067}} & \underline{\textbf{1.317 $\pm$ 0.070}} & \underline{\textbf{1.316 $\pm$ 0.068}} \\
RealMLP & \underline{\textbf{1.349 $\pm$ 0.098}} & \underline{\textbf{1.332 $\pm$ 0.089}} & \underline{\textbf{1.334 $\pm$ 0.091}} \\
TabM & \underline{\textbf{1.439 $\pm$ 0.199}} & \underline{\textbf{1.447 $\pm$ 0.210}} & \underline{\textbf{1.413 $\pm$ 0.148}} \\
TabDPT & - & - & - \\
TabPFN-2.6 & - & - & - \\
TabICLv2 & \underline{\textbf{1.348 $\pm$ 0.081}} & - & - \\
\bottomrule
\end{tabular}
\space
  \end{minipage}\hfill
  \begin{minipage}[t]{0.48\textwidth}
    \centering
    \vspace{0pt}
    \includegraphics[width=\linewidth]{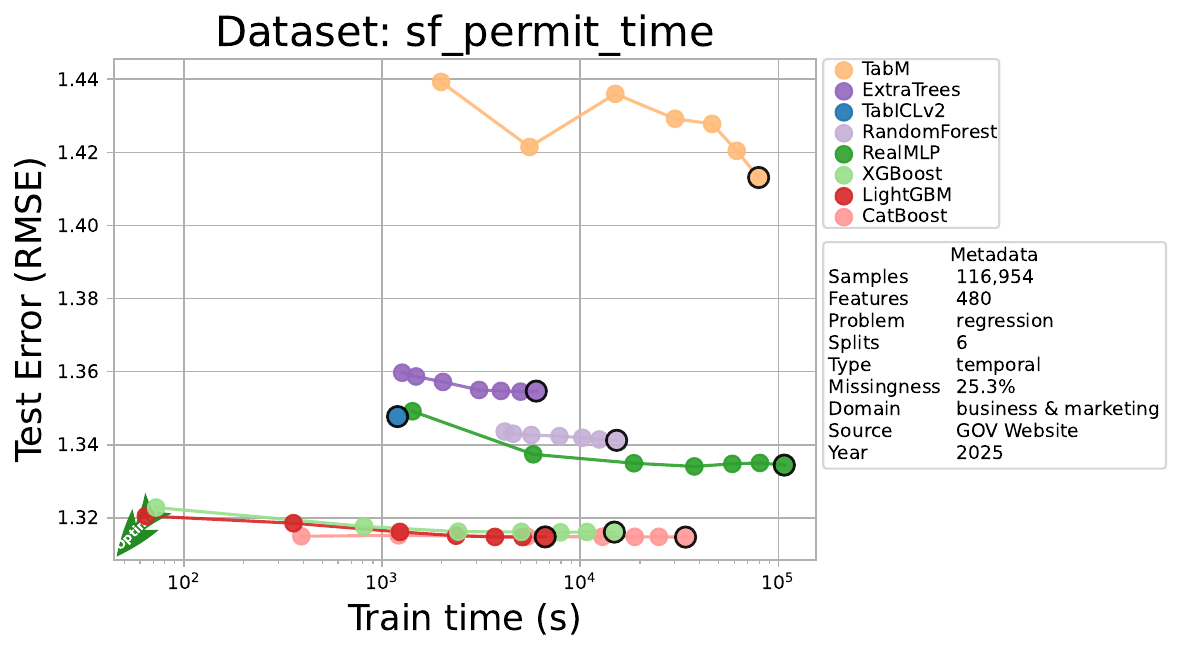}
  \end{minipage}

  \captionof{figure}{\textbf{sf\_permit\_time}: per-method test error (left) and HPO Pareto trajectory (right).}
  \label{fig:perdataset_sf_permit_time}
\end{center}
\space
\begin{center}
  \begin{minipage}[t]{0.48\textwidth}
    \centering
    \vspace{0pt}
    \scriptsize
\begin{tabular}{l@{\quad}lll}
\toprule
 & Default & Tuned & Tuned + Ens. \\
\midrule
Linear & \underline{\textbf{0.228 $\pm$ 0.036}} & \underline{0.231 $\pm$ 0.036} & \underline{\textbf{\textcolor{green!50!black}{0.227 $\pm$ 0.036}}} \\
RandomForest & 0.266 $\pm$ 0.032 & 0.258 $\pm$ 0.035 & 0.259 $\pm$ 0.035 \\
ExtraTrees & 0.253 $\pm$ 0.034 & 0.240 $\pm$ 0.037 & 0.241 $\pm$ 0.037 \\
CatBoost & 0.246 $\pm$ 0.036 & 0.244 $\pm$ 0.036 & 0.244 $\pm$ 0.035 \\
LightGBM & 0.257 $\pm$ 0.033 & 0.239 $\pm$ 0.035 & 0.239 $\pm$ 0.035 \\
XGBoost & 0.278 $\pm$ 0.031 & 0.261 $\pm$ 0.034 & 0.260 $\pm$ 0.034 \\
RealMLP & 0.239 $\pm$ 0.036 & 0.241 $\pm$ 0.036 & 0.240 $\pm$ 0.036 \\
TabM & 0.245 $\pm$ 0.036 & 0.250 $\pm$ 0.034 & 0.247 $\pm$ 0.036 \\
TabDPT & \underline{\textbf{0.229 $\pm$ 0.038}} & - & - \\
TabPFN-2.6 & 0.240 $\pm$ 0.035 & - & - \\
TabICLv2 & \underline{0.230 $\pm$ 0.037} & - & - \\
\bottomrule
\end{tabular}
\space
  \end{minipage}\hfill
  \begin{minipage}[t]{0.48\textwidth}
    \centering
    \vspace{0pt}
    \includegraphics[width=\linewidth]{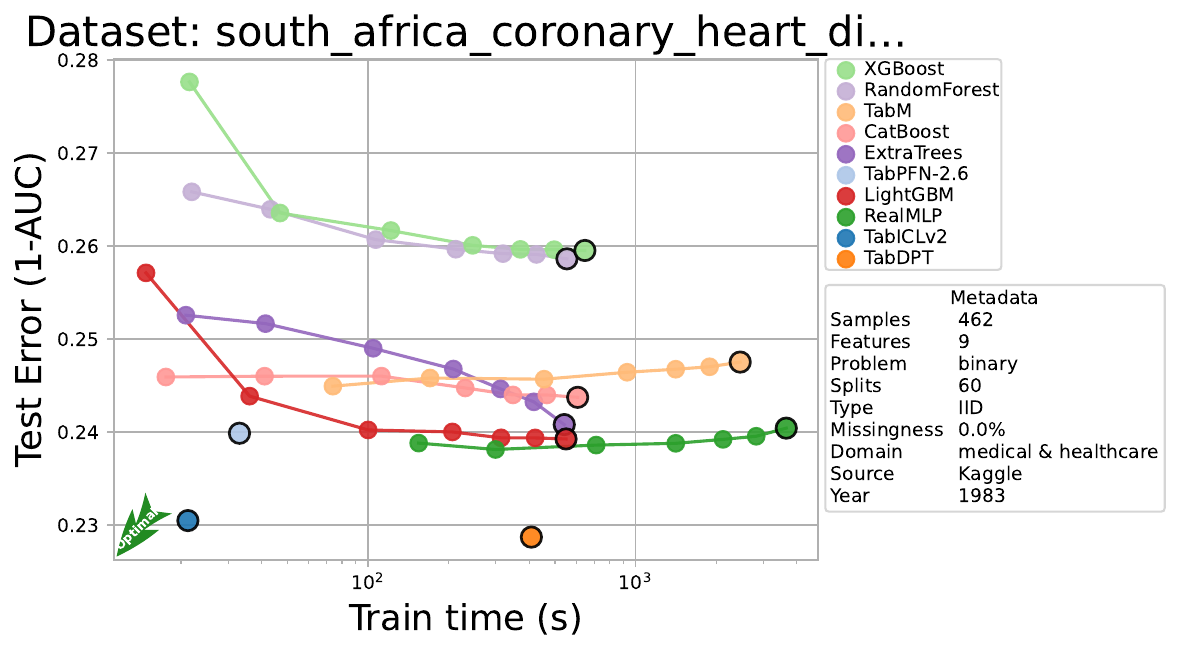}
  \end{minipage}

  \captionof{figure}{\textbf{south\_africa\_coronary\_heart\_disease}: per-method test error (left) and HPO Pareto trajectory (right).}
  \label{fig:perdataset_south_africa_coronary_heart_disease}
\end{center}
\space
\begin{center}
  \begin{minipage}[t]{0.48\textwidth}
    \centering
    \vspace{0pt}
    \scriptsize
\begin{tabular}{l@{\quad}lll}
\toprule
 & Default & Tuned & Tuned + Ens. \\
\midrule
Linear & 0.170 $\pm$ 0.013 & 0.150 $\pm$ 0.010 & 0.149 $\pm$ 0.009 \\
RandomForest & 0.327 $\pm$ 0.008 & 0.180 $\pm$ 0.015 & 0.177 $\pm$ 0.014 \\
ExtraTrees & 0.403 $\pm$ 0.009 & 0.174 $\pm$ 0.016 & 0.172 $\pm$ 0.014 \\
CatBoost & 0.111 $\pm$ 0.014 & 0.114 $\pm$ 0.015 & 0.109 $\pm$ 0.015 \\
LightGBM & 0.110 $\pm$ 0.014 & \underline{0.105 $\pm$ 0.014} & \underline{0.105 $\pm$ 0.014} \\
XGBoost & 0.117 $\pm$ 0.017 & \underline{0.107 $\pm$ 0.015} & 0.107 $\pm$ 0.015 \\
RealMLP & 0.128 $\pm$ 0.019 & \underline{0.105 $\pm$ 0.014} & \underline{0.104 $\pm$ 0.014} \\
TabM & 0.112 $\pm$ 0.015 & 0.112 $\pm$ 0.015 & 0.111 $\pm$ 0.015 \\
TabDPT & 0.200 $\pm$ 0.023 & - & - \\
TabPFN-2.6 & \underline{\textbf{0.098 $\pm$ 0.016}} & - & - \\
TabICLv2 & \underline{\textbf{\textcolor{green!50!black}{0.096 $\pm$ 0.016}}} & - & - \\
\bottomrule
\end{tabular}
\space
  \end{minipage}\hfill
  \begin{minipage}[t]{0.48\textwidth}
    \centering
    \vspace{0pt}
    \includegraphics[width=\linewidth]{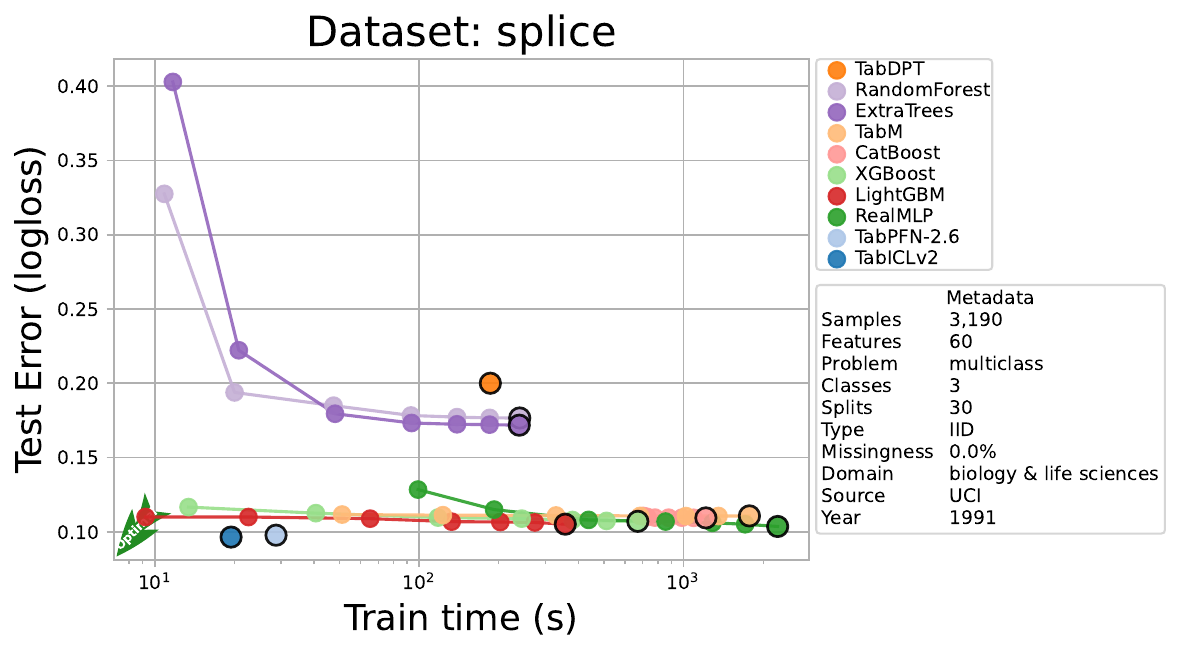}
  \end{minipage}

  \captionof{figure}{\textbf{splice}: per-method test error (left) and HPO Pareto trajectory (right).}
  \label{fig:perdataset_splice}
\end{center}
\space
\begin{center}
  \begin{minipage}[t]{0.48\textwidth}
    \centering
    \vspace{0pt}
    \scriptsize
\begin{tabular}{l@{\quad}lll}
\toprule
 & Default & Tuned & Tuned + Ens. \\
\midrule
Linear & 2.768 $\pm$ 0.175 & 2.769 $\pm$ 0.175 & 2.729 $\pm$ 0.175 \\
RandomForest & 2.712 $\pm$ 0.174 & \underline{2.676 $\pm$ 0.179} & \underline{2.676 $\pm$ 0.177} \\
ExtraTrees & 2.681 $\pm$ 0.167 & 2.698 $\pm$ 0.171 & \underline{2.683 $\pm$ 0.170} \\
CatBoost & 2.699 $\pm$ 0.183 & 2.696 $\pm$ 0.177 & 2.694 $\pm$ 0.176 \\
LightGBM & 2.713 $\pm$ 0.178 & 2.707 $\pm$ 0.173 & 2.694 $\pm$ 0.174 \\
XGBoost & 2.763 $\pm$ 0.175 & 2.705 $\pm$ 0.182 & 2.704 $\pm$ 0.181 \\
RealMLP & 2.687 $\pm$ 0.180 & 2.689 $\pm$ 0.183 & \underline{2.680 $\pm$ 0.183} \\
TabM & 2.692 $\pm$ 0.175 & \underline{2.683 $\pm$ 0.177} & \underline{2.682 $\pm$ 0.176} \\
TabDPT & \underline{\textbf{2.666 $\pm$ 0.178}} & - & - \\
TabPFN-2.6 & \underline{\textbf{2.661 $\pm$ 0.181}} & - & - \\
TabICLv2 & \underline{\textbf{\textcolor{green!50!black}{2.657 $\pm$ 0.190}}} & - & - \\
\bottomrule
\end{tabular}
\space
  \end{minipage}\hfill
  \begin{minipage}[t]{0.48\textwidth}
    \centering
    \vspace{0pt}
    \includegraphics[width=\linewidth]{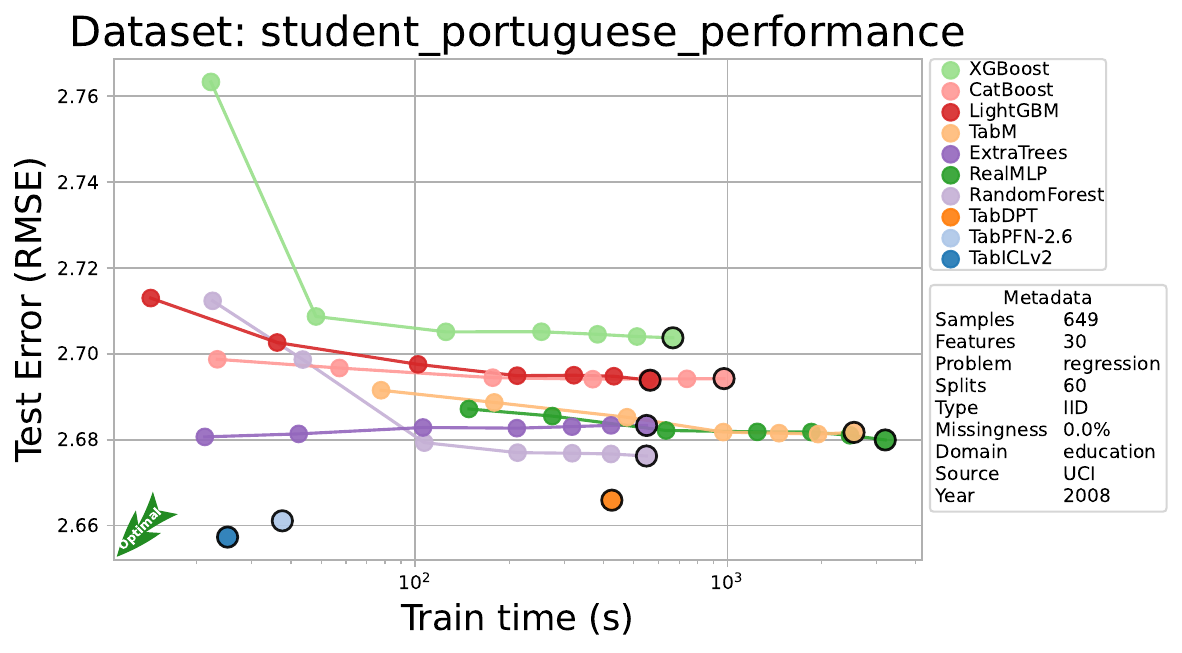}
  \end{minipage}

  \captionof{figure}{\textbf{student\_portuguese\_performance}: per-method test error (left) and HPO Pareto trajectory (right).}
  \label{fig:perdataset_student_portuguese_performance}
\end{center}
\space
\begin{center}
  \begin{minipage}[t]{0.48\textwidth}
    \centering
    \vspace{0pt}
    \scriptsize
\begin{tabular}{l@{\quad}lll}
\toprule
 & Default & Tuned & Tuned + Ens. \\
\midrule
Linear & 17.43 $\pm$ 0.09 & 17.42 $\pm$ 0.09 & 17.26 $\pm$ 0.08 \\
RandomForest & 9.65 $\pm$ 0.08 & 9.46 $\pm$ 0.11 & 9.48 $\pm$ 0.10 \\
ExtraTrees & 9.46 $\pm$ 0.13 & 9.34 $\pm$ 0.12 & 9.34 $\pm$ 0.12 \\
CatBoost & 9.33 $\pm$ 0.14 & 9.33 $\pm$ 0.13 & 9.31 $\pm$ 0.13 \\
LightGBM & 9.36 $\pm$ 0.11 & 9.34 $\pm$ 0.12 & 9.26 $\pm$ 0.12 \\
XGBoost & 9.41 $\pm$ 0.11 & 9.31 $\pm$ 0.12 & 9.31 $\pm$ 0.13 \\
RealMLP & 9.46 $\pm$ 0.17 & 9.38 $\pm$ 0.15 & \underline{9.22 $\pm$ 0.16} \\
TabM & 9.44 $\pm$ 0.17 & \underline{9.20 $\pm$ 0.18} & \underline{9.21 $\pm$ 0.17} \\
TabDPT & \underline{\textbf{8.93 $\pm$ 0.14}} & - & - \\
TabPFN-2.6 & 9.11 $\pm$ 0.16 & - & - \\
TabICLv2 & \underline{\textbf{\textcolor{green!50!black}{8.87 $\pm$ 0.17}}} & - & - \\
\bottomrule
\end{tabular}
\space
  \end{minipage}\hfill
  \begin{minipage}[t]{0.48\textwidth}
    \centering
    \vspace{0pt}
    \includegraphics[width=\linewidth]{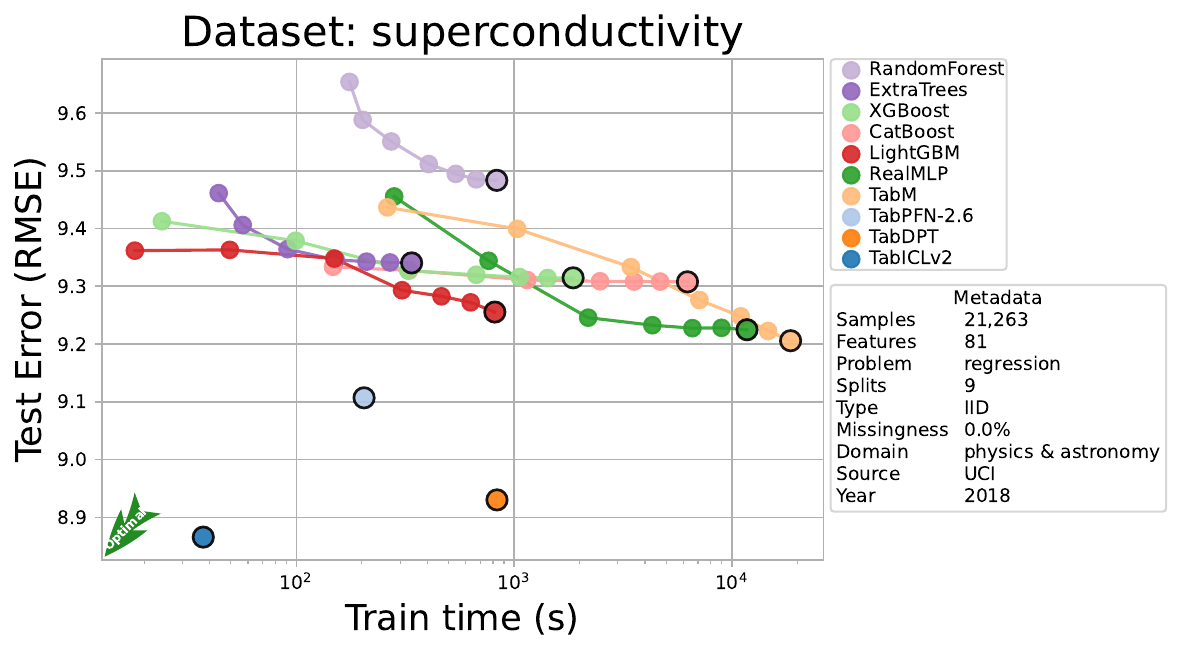}
  \end{minipage}

  \captionof{figure}{\textbf{superconductivity}: per-method test error (left) and HPO Pareto trajectory (right).}
  \label{fig:perdataset_superconductivity}
\end{center}
\space
\begin{center}
  \begin{minipage}[t]{0.48\textwidth}
    \centering
    \vspace{0pt}
    \scriptsize
\begin{tabular}{l@{\quad}lll}
\toprule
 & Default & Tuned & Tuned + Ens. \\
\midrule
Linear & 0.063 $\pm$ 0.015 & \underline{0.060 $\pm$ 0.015} & \underline{0.060 $\pm$ 0.014} \\
RandomForest & 0.062 $\pm$ 0.012 & 0.062 $\pm$ 0.012 & 0.063 $\pm$ 0.012 \\
ExtraTrees & 0.059 $\pm$ 0.010 & 0.059 $\pm$ 0.010 & 0.060 $\pm$ 0.010 \\
CatBoost & 0.054 $\pm$ 0.012 & 0.054 $\pm$ 0.012 & \underline{0.054 $\pm$ 0.012} \\
LightGBM & 0.059 $\pm$ 0.015 & \underline{0.052 $\pm$ 0.012} & \underline{0.052 $\pm$ 0.013} \\
XGBoost & 0.058 $\pm$ 0.015 & 0.054 $\pm$ 0.013 & \underline{0.054 $\pm$ 0.013} \\
RealMLP & 0.060 $\pm$ 0.012 & 0.057 $\pm$ 0.013 & \underline{0.054 $\pm$ 0.013} \\
TabM & 0.056 $\pm$ 0.012 & \underline{0.054 $\pm$ 0.012} & \underline{0.054 $\pm$ 0.011} \\
TabDPT & 0.056 $\pm$ 0.010 & - & - \\
TabPFN-2.6 & \underline{\textbf{\textcolor{green!50!black}{0.048 $\pm$ 0.009}}} & - & - \\
TabICLv2 & 0.051 $\pm$ 0.009 & - & - \\
\bottomrule
\end{tabular}
\space
  \end{minipage}\hfill
  \begin{minipage}[t]{0.48\textwidth}
    \centering
    \vspace{0pt}
    \includegraphics[width=\linewidth]{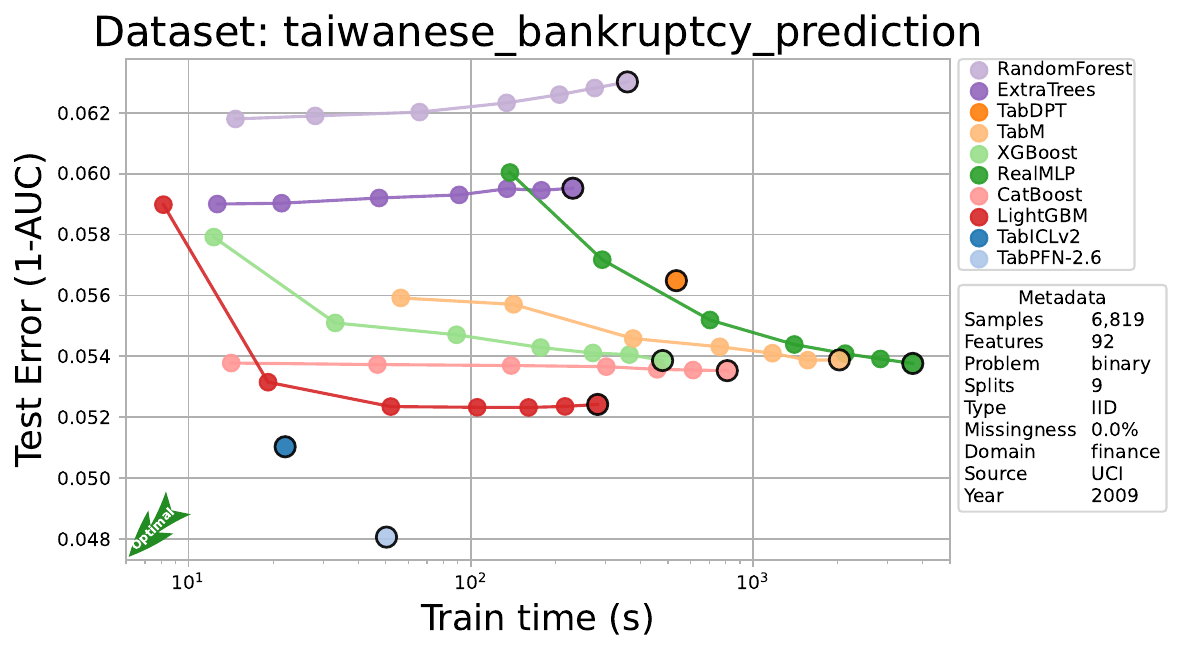}
  \end{minipage}

  \captionof{figure}{\textbf{taiwanese\_bankruptcy\_prediction}: per-method test error (left) and HPO Pareto trajectory (right).}
  \label{fig:perdataset_taiwanese_bankruptcy_prediction}
\end{center}
\space
\begin{center}
  \begin{minipage}[t]{0.48\textwidth}
    \centering
    \vspace{0pt}
    \scriptsize
\begin{tabular}{l@{\quad}lll}
\toprule
 & Default & Tuned & Tuned + Ens. \\
\midrule
Linear & 11.94 $\pm$ 1.46 & 11.63 $\pm$ 1.52 & \textbf{11.34 $\pm$ 1.54} \\
RandomForest & 11.85 $\pm$ 1.53 & 11.46 $\pm$ 1.49 & 11.48 $\pm$ 1.48 \\
ExtraTrees & 11.39 $\pm$ 1.52 & \underline{\textbf{11.24 $\pm$ 1.60}} & \underline{\textbf{11.23 $\pm$ 1.60}} \\
CatBoost & \underline{\textbf{\textcolor{green!50!black}{11.08 $\pm$ 1.70}}} & \underline{\textbf{11.09 $\pm$ 1.68}} & \underline{\textbf{11.08 $\pm$ 1.67}} \\
LightGBM & 11.15 $\pm$ 1.64 & \underline{\textbf{11.10 $\pm$ 1.60}} & \underline{\textbf{11.08 $\pm$ 1.60}} \\
XGBoost & \underline{\textbf{11.13 $\pm$ 1.71}} & 11.16 $\pm$ 1.68 & 11.15 $\pm$ 1.68 \\
RealMLP & \underline{\textbf{11.17 $\pm$ 1.47}} & \textbf{11.23 $\pm$ 1.55} & \underline{\textbf{11.09 $\pm$ 1.57}} \\
TabM & 11.29 $\pm$ 1.42 & \underline{\textbf{11.14 $\pm$ 1.52}} & \underline{\textbf{11.15 $\pm$ 1.51}} \\
TabDPT & 12.01 $\pm$ 1.45 & - & - \\
TabPFN-2.6 & 12.30 $\pm$ 1.66 & - & - \\
TabICLv2 & 11.79 $\pm$ 1.52 & - & - \\
\bottomrule
\end{tabular}
\space
  \end{minipage}\hfill
  \begin{minipage}[t]{0.48\textwidth}
    \centering
    \vspace{0pt}
    \includegraphics[width=\linewidth]{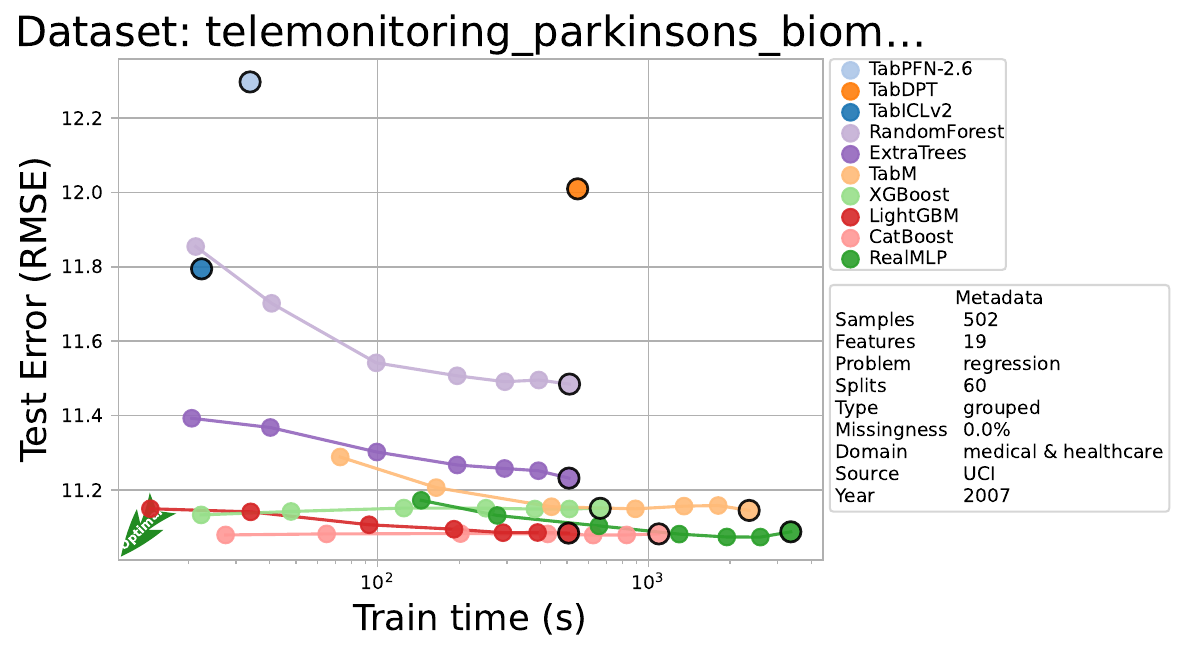}
  \end{minipage}

  \captionof{figure}{\textbf{telemonitoring\_parkinsons\_biomedical\_voice\_measurements}: per-method test error (left) and HPO Pareto trajectory (right).}
  \label{fig:perdataset_telemonitoring_parkinsons_biomedical_voice_measurements}
\end{center}
\space
\begin{center}
  \begin{minipage}[t]{0.48\textwidth}
    \centering
    \vspace{0pt}
    \scriptsize
\begin{tabular}{l@{\quad}lll}
\toprule
 & Default & Tuned & Tuned + Ens. \\
\midrule
Linear & 0.164 $\pm$ 0.044 & 0.169 $\pm$ 0.043 & 0.170 $\pm$ 0.045 \\
RandomForest & 0.022 $\pm$ 0.022 & 0.024 $\pm$ 0.027 & 0.024 $\pm$ 0.025 \\
ExtraTrees & 0.021 $\pm$ 0.012 & \underline{0.014 $\pm$ 0.007} & \underline{0.014 $\pm$ 0.007} \\
CatBoost & 0.019 $\pm$ 0.012 & \underline{\textbf{0.014 $\pm$ 0.010}} & \underline{0.016 $\pm$ 0.015} \\
LightGBM & 0.028 $\pm$ 0.024 & 0.027 $\pm$ 0.021 & 0.021 $\pm$ 0.015 \\
XGBoost & 0.022 $\pm$ 0.010 & 0.024 $\pm$ 0.013 & 0.022 $\pm$ 0.011 \\
RealMLP & 0.027 $\pm$ 0.025 & 0.024 $\pm$ 0.019 & \underline{0.016 $\pm$ 0.012} \\
TabM & 0.032 $\pm$ 0.026 & 0.029 $\pm$ 0.023 & 0.027 $\pm$ 0.022 \\
TabDPT & 0.018 $\pm$ 0.011 & - & - \\
TabPFN-2.6 & 0.013 $\pm$ 0.007 & - & - \\
TabICLv2 & \underline{\textbf{\textcolor{green!50!black}{0.011 $\pm$ 0.005}}} & - & - \\
\bottomrule
\end{tabular}
\space
  \end{minipage}\hfill
  \begin{minipage}[t]{0.48\textwidth}
    \centering
    \vspace{0pt}
    \includegraphics[width=\linewidth]{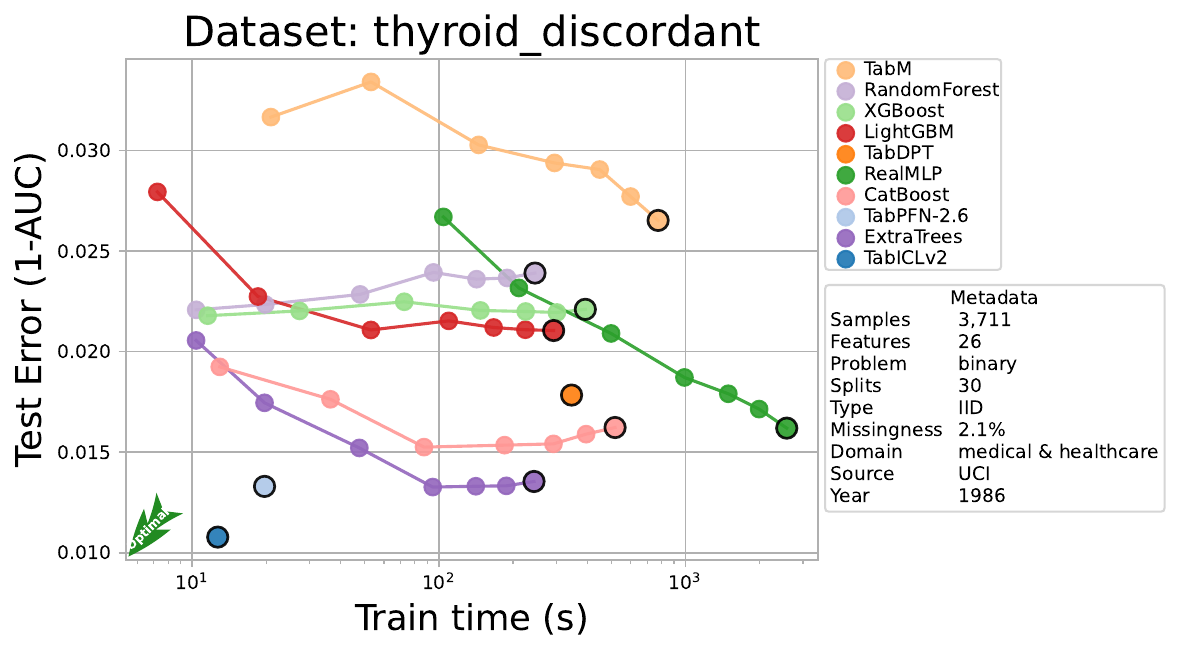}
  \end{minipage}

  \captionof{figure}{\textbf{thyroid\_discordant}: per-method test error (left) and HPO Pareto trajectory (right).}
  \label{fig:perdataset_thyroid_discordant}
\end{center}
\space
\begin{center}
  \begin{minipage}[t]{0.48\textwidth}
    \centering
    \vspace{0pt}
    \scriptsize
\begin{tabular}{l@{\quad}lll}
\toprule
 & Default & Tuned & Tuned + Ens. \\
\midrule
Linear & 0.166 $\pm$ 0.023 & 0.163 $\pm$ 0.024 & 0.163 $\pm$ 0.024 \\
RandomForest & 0.051 $\pm$ 0.010 & \underline{\textbf{0.040 $\pm$ 0.011}} & \underline{\textbf{0.040 $\pm$ 0.010}} \\
ExtraTrees & 0.058 $\pm$ 0.011 & 0.055 $\pm$ 0.011 & 0.051 $\pm$ 0.010 \\
CatBoost & 0.042 $\pm$ 0.008 & 0.042 $\pm$ 0.009 & \underline{0.041 $\pm$ 0.009} \\
LightGBM & 0.041 $\pm$ 0.009 & 0.042 $\pm$ 0.010 & 0.042 $\pm$ 0.010 \\
XGBoost & \underline{\textbf{0.040 $\pm$ 0.009}} & \underline{\textbf{0.040 $\pm$ 0.009}} & \underline{0.041 $\pm$ 0.009} \\
RealMLP & 0.054 $\pm$ 0.011 & 0.042 $\pm$ 0.009 & 0.042 $\pm$ 0.009 \\
TabM & 0.047 $\pm$ 0.011 & 0.044 $\pm$ 0.011 & 0.043 $\pm$ 0.010 \\
TabDPT & \underline{\textbf{\textcolor{green!50!black}{0.039 $\pm$ 0.010}}} & - & - \\
TabPFN-2.6 & 0.041 $\pm$ 0.009 & - & - \\
TabICLv2 & \underline{\textbf{0.040 $\pm$ 0.009}} & - & - \\
\bottomrule
\end{tabular}
\space
  \end{minipage}\hfill
  \begin{minipage}[t]{0.48\textwidth}
    \centering
    \vspace{0pt}
    \includegraphics[width=\linewidth]{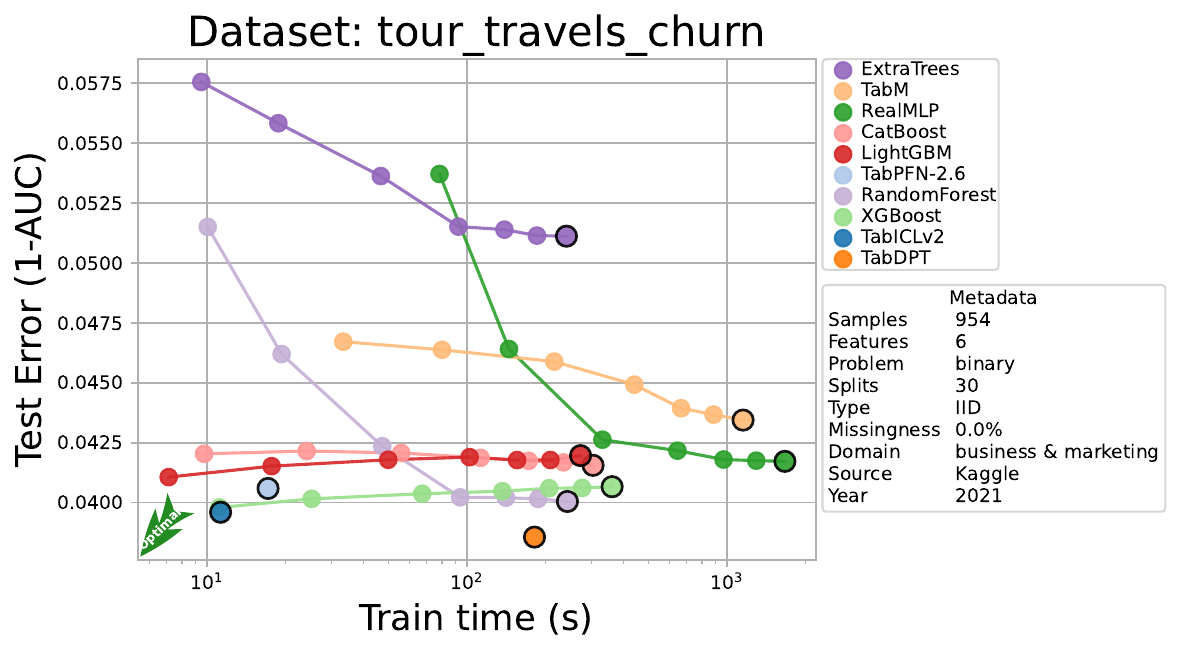}
  \end{minipage}

  \captionof{figure}{\textbf{tour\_travels\_churn}: per-method test error (left) and HPO Pareto trajectory (right).}
  \label{fig:perdataset_tour_travels_churn}
\end{center}
\space
\begin{center}
  \begin{minipage}[t]{0.48\textwidth}
    \centering
    \vspace{0pt}
    \scriptsize
\begin{tabular}{l@{\quad}lll}
\toprule
 & Default & Tuned & Tuned + Ens. \\
\midrule
Linear & 0.031 $\pm$ 0.003 & 0.025 $\pm$ 0.003 & 0.025 $\pm$ 0.003 \\
RandomForest & 0.032 $\pm$ 0.005 & 0.034 $\pm$ 0.006 & 0.033 $\pm$ 0.005 \\
ExtraTrees & 0.033 $\pm$ 0.005 & 0.033 $\pm$ 0.005 & 0.034 $\pm$ 0.005 \\
CatBoost & 0.019 $\pm$ 0.008 & 0.008 $\pm$ 0.005 & 0.008 $\pm$ 0.005 \\
LightGBM & 0.023 $\pm$ 0.005 & \underline{\textbf{\textcolor{green!50!black}{0.002 $\pm$ 0.000}}} & \underline{\textbf{0.002 $\pm$ 0.000}} \\
XGBoost & 0.025 $\pm$ 0.004 & 0.019 $\pm$ 0.004 & 0.019 $\pm$ 0.004 \\
RealMLP & 0.010 $\pm$ 0.004 & \underline{\textbf{0.006 $\pm$ 0.005}} & \underline{\textbf{0.006 $\pm$ 0.005}} \\
TabM & \underline{0.004 $\pm$ 0.001} & 0.003 $\pm$ 0.001 & 0.003 $\pm$ 0.001 \\
TabDPT & 0.015 $\pm$ 0.003 & - & - \\
TabPFN-2.6 & 0.014 $\pm$ 0.002 & - & - \\
TabICLv2 & 0.009 $\pm$ 0.002 & - & - \\
\bottomrule
\end{tabular}
\space
  \end{minipage}\hfill
  \begin{minipage}[t]{0.48\textwidth}
    \centering
    \vspace{0pt}
    \includegraphics[width=\linewidth]{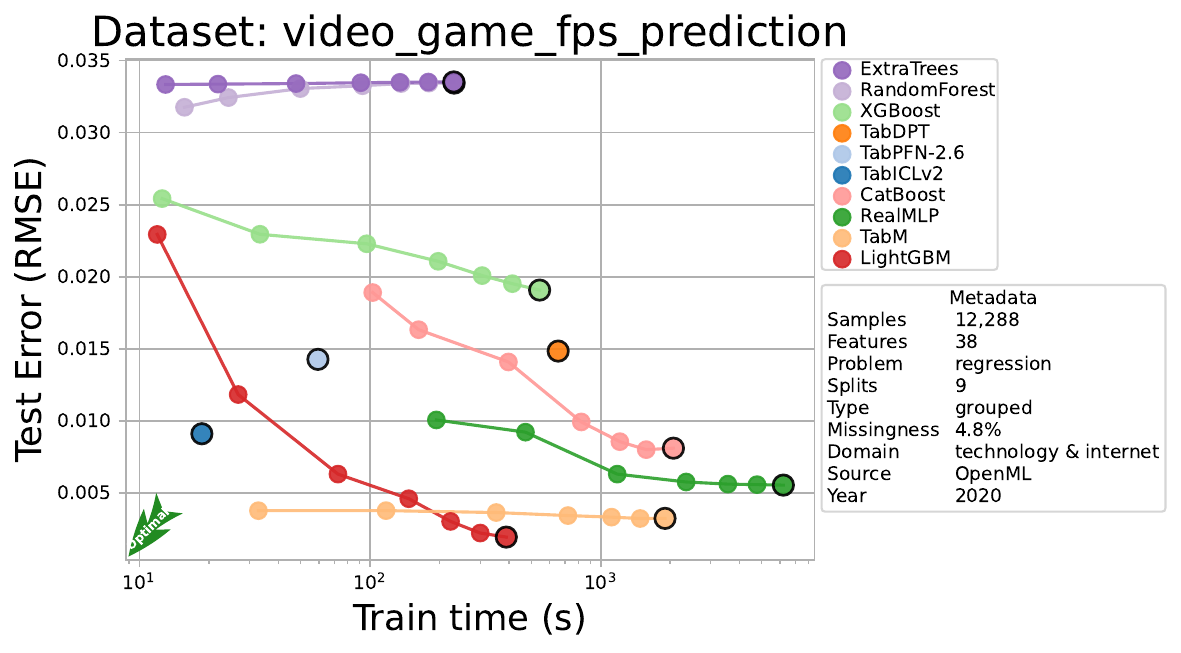}
  \end{minipage}

  \captionof{figure}{\textbf{video\_game\_fps\_prediction}: per-method test error (left) and HPO Pareto trajectory (right).}
  \label{fig:perdataset_video_game_fps_prediction}
\end{center}
\space
\begin{center}
  \begin{minipage}[t]{0.48\textwidth}
    \centering
    \vspace{0pt}
    \scriptsize
\begin{tabular}{l@{\quad}lll}
\toprule
 & Default & Tuned & Tuned + Ens. \\
\midrule
Linear & 0.432 $\pm$ 0.019 & 0.434 $\pm$ 0.017 & 0.423 $\pm$ 0.015 \\
RandomForest & 0.248 $\pm$ 0.029 & 0.236 $\pm$ 0.033 & 0.236 $\pm$ 0.032 \\
ExtraTrees & 0.227 $\pm$ 0.036 & 0.227 $\pm$ 0.036 & 0.228 $\pm$ 0.036 \\
CatBoost & \underline{0.220 $\pm$ 0.028} & \underline{0.221 $\pm$ 0.029} & 0.218 $\pm$ 0.027 \\
LightGBM & \underline{0.224 $\pm$ 0.026} & \underline{\textbf{0.226 $\pm$ 0.030}} & 0.222 $\pm$ 0.028 \\
XGBoost & 0.230 $\pm$ 0.028 & \underline{0.224 $\pm$ 0.031} & 0.223 $\pm$ 0.029 \\
RealMLP & \underline{\textbf{0.214 $\pm$ 0.030}} & \underline{0.214 $\pm$ 0.034} & \underline{\textbf{\textcolor{green!50!black}{0.208 $\pm$ 0.032}}} \\
TabM & \underline{0.222 $\pm$ 0.027} & \underline{\textbf{0.221 $\pm$ 0.033}} & \underline{\textbf{0.217 $\pm$ 0.034}} \\
TabDPT & 0.309 $\pm$ 0.023 & - & - \\
TabPFN-2.6 & 0.263 $\pm$ 0.018 & - & - \\
TabICLv2 & 0.246 $\pm$ 0.019 & - & - \\
\bottomrule
\end{tabular}
\space
  \end{minipage}\hfill
  \begin{minipage}[t]{0.48\textwidth}
    \centering
    \vspace{0pt}
    \includegraphics[width=\linewidth]{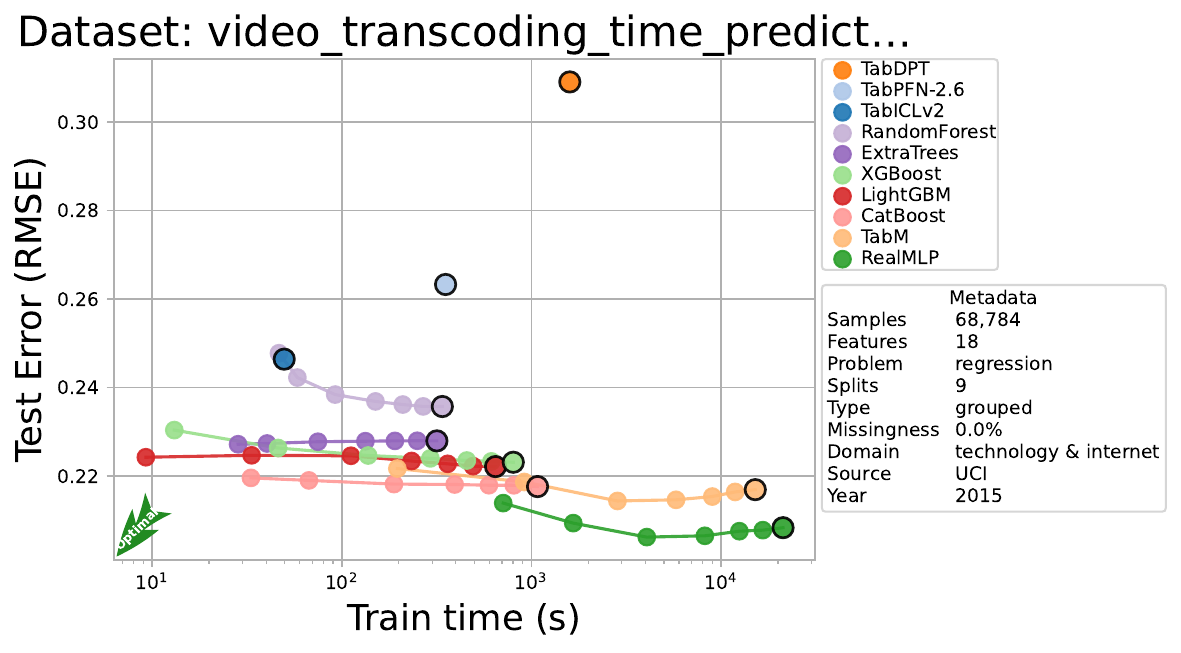}
  \end{minipage}

  \captionof{figure}{\textbf{video\_transcoding\_time\_prediction}: per-method test error (left) and HPO Pareto trajectory (right).}
  \label{fig:perdataset_video_transcoding_time_prediction}
\end{center}
\space
\begin{center}
  \begin{minipage}[t]{0.48\textwidth}
    \centering
    \vspace{0pt}
    \scriptsize
\begin{tabular}{l@{\quad}lll}
\toprule
 & Default & Tuned & Tuned + Ens. \\
\midrule
Linear & 0.356 $\pm$ 0.022 & 0.356 $\pm$ 0.025 & 0.354 $\pm$ 0.024 \\
RandomForest & 0.294 $\pm$ 0.025 & 0.261 $\pm$ 0.020 & 0.259 $\pm$ 0.019 \\
ExtraTrees & 0.295 $\pm$ 0.023 & 0.268 $\pm$ 0.022 & 0.258 $\pm$ 0.019 \\
CatBoost & 0.255 $\pm$ 0.019 & \underline{0.239 $\pm$ 0.022} & \underline{0.239 $\pm$ 0.021} \\
LightGBM & 0.258 $\pm$ 0.020 & 0.257 $\pm$ 0.021 & 0.257 $\pm$ 0.020 \\
XGBoost & 0.260 $\pm$ 0.022 & 0.256 $\pm$ 0.020 & 0.256 $\pm$ 0.020 \\
RealMLP & 0.259 $\pm$ 0.032 & \underline{0.239 $\pm$ 0.021} & \underline{0.241 $\pm$ 0.022} \\
TabM & 0.247 $\pm$ 0.026 & \underline{0.238 $\pm$ 0.026} & \underline{0.238 $\pm$ 0.025} \\
TabDPT & \underline{\textbf{0.223 $\pm$ 0.021}} & - & - \\
TabPFN-2.6 & \underline{\textbf{0.221 $\pm$ 0.022}} & - & - \\
TabICLv2 & \underline{\textbf{\textcolor{green!50!black}{0.221 $\pm$ 0.023}}} & - & - \\
\bottomrule
\end{tabular}
\space
  \end{minipage}\hfill
  \begin{minipage}[t]{0.48\textwidth}
    \centering
    \vspace{0pt}
    \includegraphics[width=\linewidth]{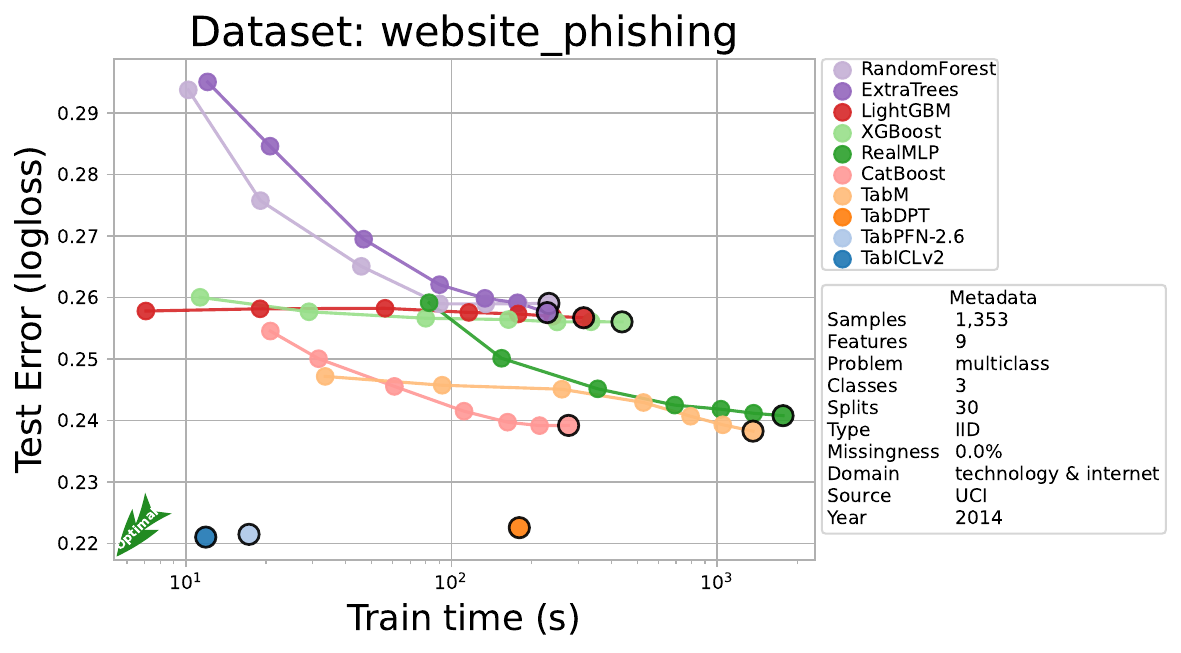}
  \end{minipage}

  \captionof{figure}{\textbf{website\_phishing}: per-method test error (left) and HPO Pareto trajectory (right).}
  \label{fig:perdataset_website_phishing}
\end{center}
\space
\begin{center}
  \begin{minipage}[t]{0.48\textwidth}
    \centering
    \vspace{0pt}
    \scriptsize
\begin{tabular}{l@{\quad}lll}
\toprule
 & Default & Tuned & Tuned + Ens. \\
\midrule
Linear & 0.160 $\pm$ 0.005 & 0.160 $\pm$ 0.005 & 0.159 $\pm$ 0.005 \\
RandomForest & 0.165 $\pm$ 0.001 & 0.157 $\pm$ 0.002 & 0.157 $\pm$ 0.001 \\
ExtraTrees & 0.172 $\pm$ 0.001 & 0.157 $\pm$ 0.001 & 0.155 $\pm$ 0.001 \\
CatBoost & 0.128 $\pm$ 0.001 & 0.127 $\pm$ 0.001 & 0.127 $\pm$ 0.001 \\
LightGBM & 0.133 $\pm$ 0.001 & 0.125 $\pm$ 0.001 & 0.125 $\pm$ 0.001 \\
XGBoost & 0.130 $\pm$ 0.001 & 0.125 $\pm$ 0.001 & 0.125 $\pm$ 0.001 \\
RealMLP & 0.126 $\pm$ 0.001 & \underline{0.123 $\pm$ 0.001} & 0.122 $\pm$ 0.001 \\
TabM & \underline{0.124 $\pm$ 0.001} & \underline{0.123 $\pm$ 0.001} & \underline{\textbf{\textcolor{green!50!black}{0.122 $\pm$ 0.001}}} \\
TabDPT & 0.190 $\pm$ 0.002 & - & - \\
TabPFN-2.6 & 0.126 $\pm$ 0.001 & - & - \\
TabICLv2 & 0.143 $\pm$ 0.002 & - & - \\
\bottomrule
\end{tabular}
\space
  \end{minipage}\hfill
  \begin{minipage}[t]{0.48\textwidth}
    \centering
    \vspace{0pt}
    \includegraphics[width=\linewidth]{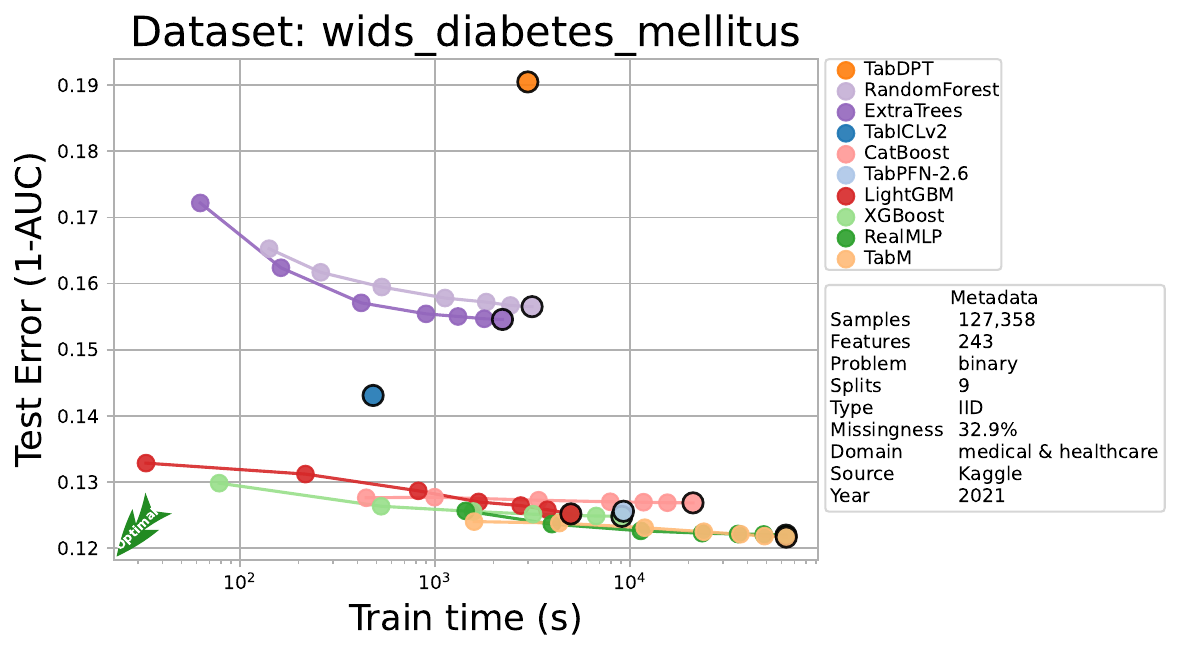}
  \end{minipage}

  \captionof{figure}{\textbf{wids\_diabetes\_mellitus}: per-method test error (left) and HPO Pareto trajectory (right).}
  \label{fig:perdataset_wids_diabetes_mellitus}
\end{center}
\space
\begin{center}
  \begin{minipage}[t]{0.48\textwidth}
    \centering
    \vspace{0pt}
    \scriptsize
\begin{tabular}{l@{\quad}lll}
\toprule
 & Default & Tuned & Tuned + Ens. \\
\midrule
Linear & 0.730 $\pm$ 0.006 & 0.730 $\pm$ 0.006 & 0.729 $\pm$ 0.006 \\
RandomForest & 0.619 $\pm$ 0.009 & 0.601 $\pm$ 0.008 & 0.605 $\pm$ 0.008 \\
ExtraTrees & 0.614 $\pm$ 0.009 & \underline{0.597 $\pm$ 0.010} & 0.598 $\pm$ 0.009 \\
CatBoost & 0.615 $\pm$ 0.009 & \underline{0.599 $\pm$ 0.009} & 0.599 $\pm$ 0.009 \\
LightGBM & 0.623 $\pm$ 0.010 & 0.610 $\pm$ 0.010 & 0.610 $\pm$ 0.009 \\
XGBoost & 0.617 $\pm$ 0.008 & 0.605 $\pm$ 0.010 & 0.606 $\pm$ 0.010 \\
RealMLP & 0.617 $\pm$ 0.009 & 0.604 $\pm$ 0.007 & \underline{0.594 $\pm$ 0.008} \\
TabM & 0.629 $\pm$ 0.009 & 0.615 $\pm$ 0.009 & 0.609 $\pm$ 0.008 \\
TabDPT & \underline{\textbf{\textcolor{green!50!black}{0.584 $\pm$ 0.008}}} & - & - \\
TabPFN-2.6 & 0.588 $\pm$ 0.007 & - & - \\
TabICLv2 & 0.587 $\pm$ 0.008 & - & - \\
\bottomrule
\end{tabular}
\space
  \end{minipage}\hfill
  \begin{minipage}[t]{0.48\textwidth}
    \centering
    \vspace{0pt}
    \includegraphics[width=\linewidth]{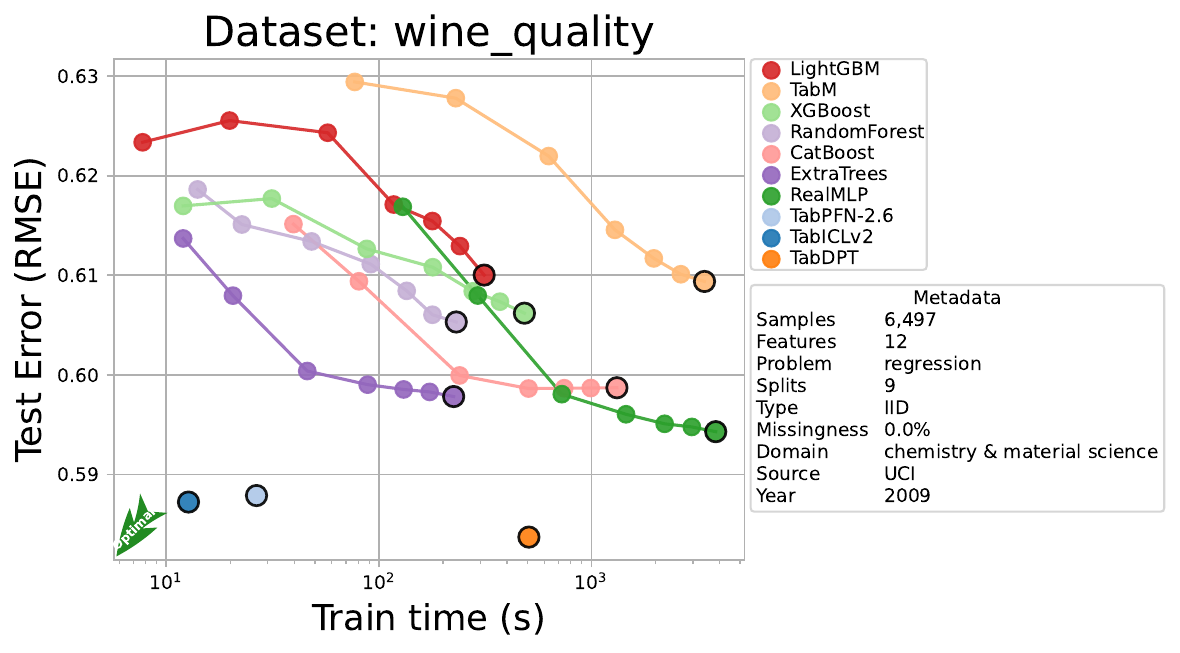}
  \end{minipage}

  \captionof{figure}{\textbf{wine\_quality}: per-method test error (left) and HPO Pareto trajectory (right).}
  \label{fig:perdataset_wine_quality}
\end{center}
\space
\begin{center}
  \begin{minipage}[t]{0.48\textwidth}
    \centering
    \vspace{0pt}
    \scriptsize
\begin{tabular}{l@{\quad}lll}
\toprule
 & Default & Tuned & Tuned + Ens. \\
\midrule
Linear & 0.471 $\pm$ 0.018 & 0.424 $\pm$ 0.018 & 0.417 $\pm$ 0.019 \\
RandomForest & 0.448 $\pm$ 0.024 & 0.427 $\pm$ 0.023 & 0.429 $\pm$ 0.023 \\
ExtraTrees & 0.435 $\pm$ 0.023 & 0.419 $\pm$ 0.023 & 0.419 $\pm$ 0.023 \\
CatBoost & 0.395 $\pm$ 0.024 & 0.393 $\pm$ 0.023 & 0.393 $\pm$ 0.024 \\
LightGBM & 0.409 $\pm$ 0.021 & 0.379 $\pm$ 0.023 & 0.379 $\pm$ 0.023 \\
XGBoost & 0.405 $\pm$ 0.022 & 0.398 $\pm$ 0.022 & 0.395 $\pm$ 0.022 \\
RealMLP & 0.371 $\pm$ 0.023 & \underline{0.365 $\pm$ 0.023} & \underline{0.363 $\pm$ 0.022} \\
TabM & 0.389 $\pm$ 0.023 & 0.377 $\pm$ 0.023 & 0.373 $\pm$ 0.023 \\
TabDPT & 0.386 $\pm$ 0.027 & - & - \\
TabPFN-2.6 & \underline{\textbf{\textcolor{green!50!black}{0.359 $\pm$ 0.019}}} & - & - \\
TabICLv2 & 0.385 $\pm$ 0.025 & - & - \\
\bottomrule
\end{tabular}
\space
  \end{minipage}\hfill
  \begin{minipage}[t]{0.48\textwidth}
    \centering
    \vspace{0pt}
    \includegraphics[width=\linewidth]{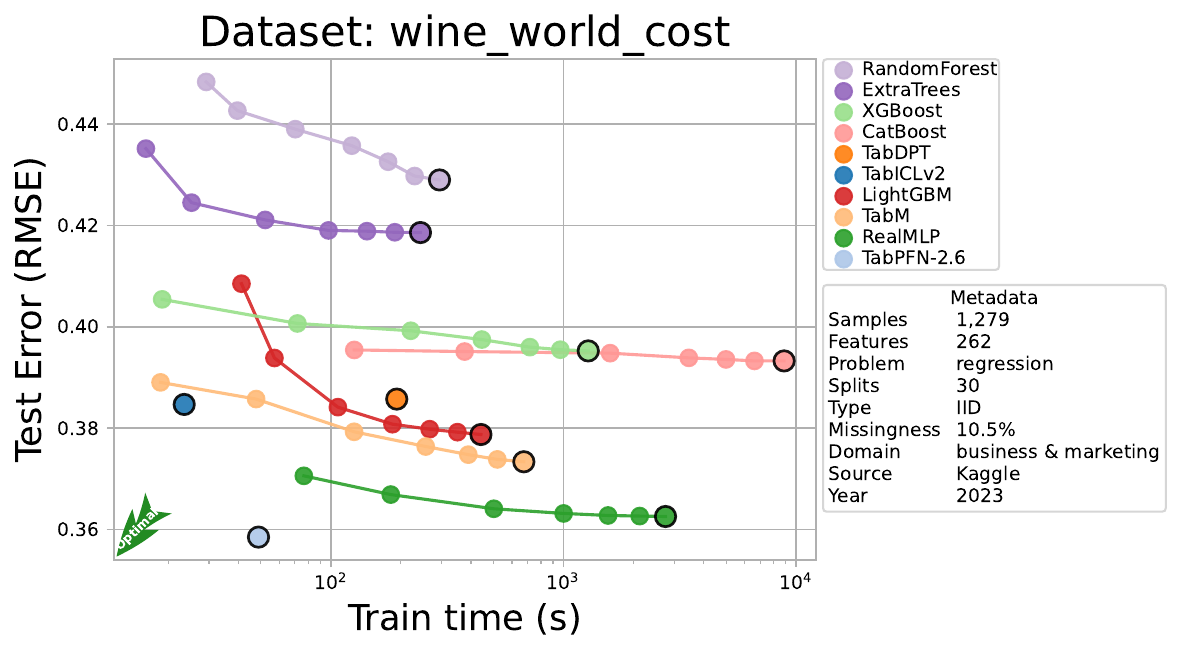}
  \end{minipage}

  \captionof{figure}{\textbf{wine\_world\_cost}: per-method test error (left) and HPO Pareto trajectory (right).}
  \label{fig:perdataset_wine_world_cost}
\end{center}
\space
\space

\clearpage
\newpage
\section{NeurIPS Paper Checklist}

\begin{enumerate}

\item {\bf Claims}
    \item[] Question: Do the main claims made in the abstract and introduction accurately reflect the paper's contributions and scope?
    \item[] Answer: \answerYes{}
    \item[] Justification: We claim in the abstract and introduction that we (1) introduce BeyondArena as benchmark and DataFoundry as a unified dataset curation framework and metadata schema, (2) curate 142 datasets spanning diverse disciplines and tabular learning settings, (3) support benchmarking across IID, temporal, and grouped prediction tasks, (4) evaluate models across varying sample sizes, feature dimensionalities, and feature types, and
    (5) current open-source TFMs, used via in-context learning, are competitive on datasets with up to 100k training rows, including temporal and high-dimensional ones, and (6) tuned traditional models lead on large and grouped data.   
    We deliver on these claims in \Cref{sec:data_curation} (1,2); \Cref{sec:experiment_setup} (1,3); and \Cref{sec:results} (4,5,6).
    \item[] Guidelines:
    \begin{itemize}
        \item The answer \answerNA{} means that the abstract and introduction do not include the claims made in the paper.
        \item The abstract and/or introduction should clearly state the claims made, including the contributions made in the paper and important assumptions and limitations. A \answerNo{} or \answerNA{} answer to this question will not be perceived well by the reviewers. 
        \item The claims made should match theoretical and experimental results, and reflect how much the results can be expected to generalize to other settings. 
        \item It is fine to include aspirational goals as motivation as long as it is clear that these goals are not attained by the paper. 
    \end{itemize}

\item {\bf Limitations}
    \item[] Question: Does the paper discuss the limitations of the work performed by the authors?
    \item[] Answer: \answerYes{} 
    \item[] Justification: We discuss limitations in \Cref{sec:conclusion}.
    \item[] Guidelines:
    \begin{itemize}
        \item The answer \answerNA{} means that the paper has no limitation while the answer \answerNo{} means that the paper has limitations, but those are not discussed in the paper. 
        \item The authors are encouraged to create a separate ``Limitations'' section in their paper.
        \item The paper should point out any strong assumptions and how robust the results are to violations of these assumptions (e.g., independence assumptions, noiseless settings, model well-specification, asymptotic approximations only holding locally). The authors should reflect on how these assumptions might be violated in practice and what the implications would be.
        \item The authors should reflect on the scope of the claims made, e.g., if the approach was only tested on a few datasets or with a few runs. In general, empirical results often depend on implicit assumptions, which should be articulated.
        \item The authors should reflect on the factors that influence the performance of the approach. For example, a facial recognition algorithm may perform poorly when image resolution is low or images are taken in low lighting. Or a speech-to-text system might not be used reliably to provide closed captions for online lectures because it fails to handle technical jargon.
        \item The authors should discuss the computational efficiency of the proposed algorithms and how they scale with dataset size.
        \item If applicable, the authors should discuss possible limitations of their approach to address problems of privacy and fairness.
        \item While the authors might fear that complete honesty about limitations might be used by reviewers as grounds for rejection, a worse outcome might be that reviewers discover limitations that aren't acknowledged in the paper. The authors should use their best judgment and recognize that individual actions in favor of transparency play an important role in developing norms that preserve the integrity of the community. Reviewers will be specifically instructed to not penalize honesty concerning limitations.
    \end{itemize}

\item {\bf Theory assumptions and proofs}
    \item[] Question: For each theoretical result, does the paper provide the full set of assumptions and a complete (and correct) proof?
    \item[] Answer: \answerNA{} %
    \item[] Justification: We present no theoretical results.
    \item[] Guidelines:
    \begin{itemize}
        \item The answer \answerNA{} means that the paper does not include theoretical results. 
        \item All the theorems, formulas, and proofs in the paper should be numbered and cross-referenced.
        \item All assumptions should be clearly stated or referenced in the statement of any theorems.
        \item The proofs can either appear in the main paper or the supplemental material, but if they appear in the supplemental material, the authors are encouraged to provide a short proof sketch to provide intuition. 
        \item Inversely, any informal proof provided in the core of the paper should be complemented by formal proofs provided in appendix or supplemental material.
        \item Theorems and Lemmas that the proof relies upon should be properly referenced. 
    \end{itemize}

    \item {\bf Experimental result reproducibility}
    \item[] Question: Does the paper fully disclose all the information needed to reproduce the main experimental results of the paper to the extent that it affects the main claims and/or conclusions of the paper (regardless of whether the code and data are provided or not)?
    \item[] Answer: \answerYes{} %
    \item[] Justification: We describe our dataset curation in \Cref{sec:data_curation}, the experimental setup in \Cref{sec:experiment_setup}, provide more details in the \Cref{appendix:experiment_setup_details}, and share all our code.
    \item[] Guidelines:
    \begin{itemize}
        \item The answer \answerNA{} means that the paper does not include experiments.
        \item If the paper includes experiments, a \answerNo{} answer to this question will not be perceived well by the reviewers: Making the paper reproducible is important, regardless of whether the code and data are provided or not.
        \item If the contribution is a dataset and\slash or model, the authors should describe the steps taken to make their results reproducible or verifiable. 
        \item Depending on the contribution, reproducibility can be accomplished in various ways. For example, if the contribution is a novel architecture, describing the architecture fully might suffice, or if the contribution is a specific model and empirical evaluation, it may be necessary to either make it possible for others to replicate the model with the same dataset, or provide access to the model. In general. releasing code and data is often one good way to accomplish this, but reproducibility can also be provided via detailed instructions for how to replicate the results, access to a hosted model (e.g., in the case of a large language model), releasing of a model checkpoint, or other means that are appropriate to the research performed.
        \item While NeurIPS does not require releasing code, the conference does require all submissions to provide some reasonable avenue for reproducibility, which may depend on the nature of the contribution. For example
        \begin{enumerate}
            \item If the contribution is primarily a new algorithm, the paper should make it clear how to reproduce that algorithm.
            \item If the contribution is primarily a new model architecture, the paper should describe the architecture clearly and fully.
            \item If the contribution is a new model (e.g., a large language model), then there should either be a way to access this model for reproducing the results or a way to reproduce the model (e.g., with an open-source dataset or instructions for how to construct the dataset).
            \item We recognize that reproducibility may be tricky in some cases, in which case authors are welcome to describe the particular way they provide for reproducibility. In the case of closed-source models, it may be that access to the model is limited in some way (e.g., to registered users), but it should be possible for other researchers to have some path to reproducing or verifying the results.
        \end{enumerate}
    \end{itemize}

\item {\bf Open access to data and code}
    \item[] Question: Does the paper provide open access to the data and code, with sufficient instructions to faithfully reproduce the main experimental results, as described in supplemental material?
    \item[] Answer: \answerYes{} %
    \item[] Justification: We share all our dataset curation code, benchmarking code, and dataset artifacts in the paper.
    \item[] Guidelines:
    \begin{itemize}
        \item The answer \answerNA{} means that paper does not include experiments requiring code.
        \item Please see the NeurIPS code and data submission guidelines (\url{https://neurips.cc/public/guides/CodeSubmissionPolicy}) for more details.
        \item While we encourage the release of code and data, we understand that this might not be possible, so \answerNo{} is an acceptable answer. Papers cannot be rejected simply for not including code, unless this is central to the contribution (e.g., for a new open-source benchmark).
        \item The instructions should contain the exact command and environment needed to run to reproduce the results. See the NeurIPS code and data submission guidelines (\url{https://neurips.cc/public/guides/CodeSubmissionPolicy}) for more details.
        \item The authors should provide instructions on data access and preparation, including how to access the raw data, preprocessed data, intermediate data, and generated data, etc.
        \item The authors should provide scripts to reproduce all experimental results for the new proposed method and baselines. If only a subset of experiments are reproducible, they should state which ones are omitted from the script and why.
        \item At submission time, to preserve anonymity, the authors should release anonymized versions (if applicable).
        \item Providing as much information as possible in supplemental material (appended to the paper) is recommended, but including URLs to data and code is permitted.
    \end{itemize}

\item {\bf Experimental setting/details}
    \item[] Question: Does the paper specify all the training and test details (e.g., data splits, hyperparameters, how they were chosen, type of optimizer) necessary to understand the results?
    \item[] Answer: \answerYes{} %
    \item[] Justification: We share all details in \Cref{sec:experiment_setup}.
    \item[] Guidelines:
    \begin{itemize}
        \item The answer \answerNA{} means that the paper does not include experiments.
        \item The experimental setting should be presented in the core of the paper to a level of detail that is necessary to appreciate the results and make sense of them.
        \item The full details can be provided either with the code, in appendix, or as supplemental material.
    \end{itemize}

\item {\bf Experiment statistical significance}
    \item[] Question: Does the paper report error bars suitably and correctly defined or other appropriate information about the statistical significance of the experiments?
    \item[] Answer: \answerYes{} %
    \item[] Justification: We report all results with error bars and provide a statistical analysis in \Cref{app:posthoc}.
    \item[] Guidelines:
    \begin{itemize}
        \item The answer \answerNA{} means that the paper does not include experiments.
        \item The authors should answer \answerYes{} if the results are accompanied by error bars, confidence intervals, or statistical significance tests, at least for the experiments that support the main claims of the paper.
        \item The factors of variability that the error bars are capturing should be clearly stated (for example, train/test split, initialization, random drawing of some parameter, or overall run with given experimental conditions).
        \item The method for calculating the error bars should be explained (closed form formula, call to a library function, bootstrap, etc.)
        \item The assumptions made should be given (e.g., Normally distributed errors).
        \item It should be clear whether the error bar is the standard deviation or the standard error of the mean.
        \item It is OK to report 1-sigma error bars, but one should state it. The authors should preferably report a 2-sigma error bar than state that they have a 96\% CI, if the hypothesis of Normality of errors is not verified.
        \item For asymmetric distributions, the authors should be careful not to show in tables or figures symmetric error bars that would yield results that are out of range (e.g., negative error rates).
        \item If error bars are reported in tables or plots, the authors should explain in the text how they were calculated and reference the corresponding figures or tables in the text.
    \end{itemize}

\item {\bf Experiments compute resources}
    \item[] Question: For each experiment, does the paper provide sufficient information on the computer resources (type of compute workers, memory, time of execution) needed to reproduce the experiments?
    \item[] Answer: \answerYes{} %
    \item[] Justification: We provide details on the compute resources in \Cref{sec:experiment_setup} and \Cref{appendix:experiment_setup_details}.
    \item[] Guidelines:
    \begin{itemize}
        \item The answer \answerNA{} means that the paper does not include experiments.
        \item The paper should indicate the type of compute workers CPU or GPU, internal cluster, or cloud provider, including relevant memory and storage.
        \item The paper should provide the amount of compute required for each of the individual experimental runs as well as estimate the total compute. 
        \item The paper should disclose whether the full research project required more compute than the experiments reported in the paper (e.g., preliminary or failed experiments that didn't make it into the paper). 
    \end{itemize}
    
\item {\bf Code of ethics}
    \item[] Question: Does the research conducted in the paper conform, in every respect, with the NeurIPS Code of Ethics \url{https://neurips.cc/public/EthicsGuidelines}?
    \item[] Answer: \answerYes{} %
    \item[] Justification: We believe our work conforms to the NeurIPS Code of Ethics.
    \item[] Guidelines:
    \begin{itemize}
        \item The answer \answerNA{} means that the authors have not reviewed the NeurIPS Code of Ethics.
        \item If the authors answer \answerNo, they should explain the special circumstances that require a deviation from the Code of Ethics.
        \item The authors should make sure to preserve anonymity (e.g., if there is a special consideration due to laws or regulations in their jurisdiction).
    \end{itemize}

\item {\bf Broader impacts}
    \item[] Question: Does the paper discuss both potential positive societal impacts and negative societal impacts of the work performed?
    \item[] Answer: \answerYes{} %
    \item[] Justification: We discuss societal impacts in \Cref{sec:conclusion}.
    \item[] Guidelines:
    \begin{itemize}
        \item The answer \answerNA{} means that there is no societal impact of the work performed.
        \item If the authors answer \answerNA{} or \answerNo, they should explain why their work has no societal impact or why the paper does not address societal impact.
        \item Examples of negative societal impacts include potential malicious or unintended uses (e.g., disinformation, generating fake profiles, surveillance), fairness considerations (e.g., deployment of technologies that could make decisions that unfairly impact specific groups), privacy considerations, and security considerations.
        \item The conference expects that many papers will be foundational research and not tied to particular applications, let alone deployments. However, if there is a direct path to any negative applications, the authors should point it out. For example, it is legitimate to point out that an improvement in the quality of generative models could be used to generate Deepfakes for disinformation. On the other hand, it is not needed to point out that a generic algorithm for optimizing neural networks could enable people to train models that generate Deepfakes faster.
        \item The authors should consider possible harms that could arise when the technology is being used as intended and functioning correctly, harms that could arise when the technology is being used as intended but gives incorrect results, and harms following from (intentional or unintentional) misuse of the technology.
        \item If there are negative societal impacts, the authors could also discuss possible mitigation strategies (e.g., gated release of models, providing defenses in addition to attacks, mechanisms for monitoring misuse, mechanisms to monitor how a system learns from feedback over time, improving the efficiency and accessibility of ML).
    \end{itemize}
    
\item {\bf Safeguards}
    \item[] Question: Does the paper describe safeguards that have been put in place for responsible release of data or models that have a high risk for misuse (e.g., pre-trained language models, image generators, or scraped datasets)?
    \item[] Answer: \answerNA{} %
    \item[] Justification: The code and curated data artifacts we release have no perceivable risk of misuse. We do not release models. 
    \item[] Guidelines:
    \begin{itemize}
        \item The answer \answerNA{} means that the paper poses no such risks.
        \item Released models that have a high risk for misuse or dual-use should be released with necessary safeguards to allow for controlled use of the model, for example by requiring that users adhere to usage guidelines or restrictions to access the model or implementing safety filters. 
        \item Datasets that have been scraped from the Internet could pose safety risks. The authors should describe how they avoided releasing unsafe images.
        \item We recognize that providing effective safeguards is challenging, and many papers do not require this, but we encourage authors to take this into account and make a best faith effort.
    \end{itemize}

\item {\bf Licenses for existing assets}
    \item[] Question: Are the creators or original owners of assets (e.g., code, data, models), used in the paper, properly credited and are the license and terms of use explicitly mentioned and properly respected?
    \item[] Answer: \answerYes{} %
    \item[] Justification: We credited all datasets we curated, see \Cref{tab:all-datasets}, and respected the licenses of all assets (e.g., code and models) that we used.
    \item[] Guidelines:
    \begin{itemize}
        \item The answer \answerNA{} means that the paper does not use existing assets.
        \item The authors should cite the original paper that produced the code package or dataset.
        \item The authors should state which version of the asset is used and, if possible, include a URL.
        \item The name of the license (e.g., CC-BY 4.0) should be included for each asset.
        \item For scraped data from a particular source (e.g., website), the copyright and terms of service of that source should be provided.
        \item If assets are released, the license, copyright information, and terms of use in the package should be provided. For popular datasets, \url{paperswithcode.com/datasets} has curated licenses for some datasets. Their licensing guide can help determine the license of a dataset.
        \item For existing datasets that are re-packaged, both the original license and the license of the derived asset (if it has changed) should be provided.
        \item If this information is not available online, the authors are encouraged to reach out to the asset's creators.
    \end{itemize}

\item {\bf New assets}
    \item[] Question: Are new assets introduced in the paper well documented and is the documentation provided alongside the assets?
    \item[] Answer: \answerYes{} %
    \item[] Justification: We introduce \df in \Cref{sec:data_curation} and share its documented code. 
    \item[] Guidelines:
    \begin{itemize}
        \item The answer \answerNA{} means that the paper does not release new assets.
        \item Researchers should communicate the details of the dataset\slash code\slash model as part of their submissions via structured templates. This includes details about training, license, limitations, etc. 
        \item The paper should discuss whether and how consent was obtained from people whose asset is used.
        \item At submission time, remember to anonymize your assets (if applicable). You can either create an anonymized URL or include an anonymized zip file.
    \end{itemize}

\item {\bf Crowdsourcing and research with human subjects}
    \item[] Question: For crowdsourcing experiments and research with human subjects, does the paper include the full text of instructions given to participants and screenshots, if applicable, as well as details about compensation (if any)? 
    \item[] Answer: \answerNA{} %
    \item[] Justification: The paper does not involve crowdsourcing nor research with human subjects.
    \item[] Guidelines:
    \begin{itemize}
        \item The answer \answerNA{} means that the paper does not involve crowdsourcing nor research with human subjects.
        \item Including this information in the supplemental material is fine, but if the main contribution of the paper involves human subjects, then as much detail as possible should be included in the main paper. 
        \item According to the NeurIPS Code of Ethics, workers involved in data collection, curation, or other labor should be paid at least the minimum wage in the country of the data collector. 
    \end{itemize}

\item {\bf Institutional review board (IRB) approvals or equivalent for research with human subjects}
    \item[] Question: Does the paper describe potential risks incurred by study participants, whether such risks were disclosed to the subjects, and whether Institutional Review Board (IRB) approvals (or an equivalent approval/review based on the requirements of your country or institution) were obtained?
    \item[] Answer: \answerNA{} %
    \item[] Justification: The paper does not involve crowdsourcing nor research with human subjects.
    \item[] Guidelines:
    \begin{itemize}
        \item The answer \answerNA{} means that the paper does not involve crowdsourcing nor research with human subjects.
        \item Depending on the country in which research is conducted, IRB approval (or equivalent) may be required for any human subjects research. If you obtained IRB approval, you should clearly state this in the paper. 
        \item We recognize that the procedures for this may vary significantly between institutions and locations, and we expect authors to adhere to the NeurIPS Code of Ethics and the guidelines for their institution. 
        \item For initial submissions, do not include any information that would break anonymity (if applicable), such as the institution conducting the review.
    \end{itemize}

\item {\bf Declaration of LLM usage}
    \item[] Question: Does the paper describe the usage of LLMs if it is an important, original, or non-standard component of the core methods in this research? Note that if the LLM is used only for writing, editing, or formatting purposes and does \emph{not} impact the core methodology, scientific rigor, or originality of the research, declaration is not required.
    \item[] Answer: \answerNA{} %
    \item[] Justification: Our contributions do not involve LLMs in any components. 
    \item[] Guidelines:
    \begin{itemize}
        \item The answer \answerNA{} means that the core method development in this research does not involve LLMs as any important, original, or non-standard components.
        \item Please refer to our LLM policy in the NeurIPS handbook for what should or should not be described.
    \end{itemize}

\end{enumerate}
\space

\end{appendices}
\space

\end{document}